\documentclass{elsarticle}

\usepackage{lineno,hyperref}
\modulolinenumbers[5]

\usepackage{xcolor}
\usepackage{placeins}
\usepackage{pifont}             
\newcommand{\cmark}{\ding{51}}  
\newcommand{\xmark}{\ding{55}}  
\usepackage{amsmath}
\usepackage{amssymb}
\usepackage{bbm} 
\usepackage{booktabs} 
\usepackage[export]{adjustbox} 

\begin{document}

\begin{frontmatter}

\title{Ground Truth Evaluation of Neural Network Explanations with CLEVR-XAI} 

\author{Leila Arras\corref{equalcontribution}}
\ead{leila.arrask@hhi.fraunhofer.de}
\author{Ahmed Osman\corref{equalcontribution}}
\ead{ahmed@osman.ai}
\author{Wojciech Samek}
\ead{wojciech.samek@hhi.fraunhofer.de}

\cortext[equalcontribution]{Equal contribution}

\address{Department of Artificial Intelligence, Fraunhofer Heinrich Hertz Institute, 10587 Berlin, Germany}

\begin{abstract}
The rise of deep learning in today's applications entailed an increasing need in explaining the model's decisions beyond prediction performances in order to foster trust and accountability. Recently, the field of explainable AI (XAI) has developed methods that provide such explanations for already trained neural networks. In computer vision tasks such explanations, termed \textit{heatmaps}, visualize the contributions of individual pixels to the prediction. So far XAI methods along with their heatmaps were mainly validated qualitatively via human-based assessment, or evaluated through auxiliary proxy tasks such as pixel perturbation, weak object localization or randomization tests. Due to the lack of an objective and commonly accepted quality measure for heatmaps, it was debatable which XAI method performs best and whether explanations can be trusted at all. In the present work, we tackle the problem by proposing a ground truth based evaluation framework for XAI methods based on the CLEVR visual question answering task. Our framework provides a (1) selective, (2) controlled and (3) realistic testbed for the evaluation of neural network explanations.
We compare ten different explanation methods, resulting in new insights about the quality and properties of XAI methods, sometimes contradicting with conclusions from previous comparative studies. The CLEVR-XAI dataset and the benchmarking code can be found at \url{https://github.com/ahmedmagdiosman/clevr-xai}.

\end{abstract}


\end{frontmatter}


\section{Introduction}

With the renaissance of neural networks in the last decade, the domains of application of deep learning have been continuously increasing.
Indeed these models were shown to reach excellent performance on various large-scale prediction tasks, e.g., in computer vision \citep{lu2014surpassing}, language understanding \citep{devlin2019bert} or medical diagnosis \citep{esteva2017dermatologist}.
At the same time, concerns were raised to whether such high performance is based on genuinely solving a given problem, or if it may partly rely on exploiting spurious correlations found in the data \cite{Geirhos:Nature2020,Hendricks:ECCV2018,Manjunatha:CVPR2019,Lap:Nature19}.
The field of explainable AI (XAI) \cite{SamXAI19,SamPIEEE21} has recently developed various techniques to uncover the decision making process of the model and help unmasking such Clever Hans\footnote{Clever Hans was a horse that was supposed to be able to perform simple calculations, but actually arrived at the correct solution by utilizing a spurious correlation, unintentionally reflected in the body language of the enquirer.} predictors \cite{Lap:Nature19}. From an end-user perspective it might also be desirable or even legally required \cite{Arrieta:InfoFusion2020,EU-GDPR}, especially for safety-critical systems, to accompany a model's decision with an {\it explanation} in order to trace it back to the decisive parts of the input. Beyond that, XAI bears also the potential to help improve model performance and efficiency \cite{Yeom:Arxiv2019}, or to enable new data-driven scientific discoveries \cite{schutt2017quantum,HorSREP19,Widrich:2020}.

In the vision domain, the explanation can take the form of a {\it heatmap}, where each pixel in an input image gets assigned a {\it relevance} value or score, indicating its relative contribution to the final decision.
Methods providing such heatmaps in a deterministic and unambiguous way on an already trained neural network (so-called \textit{post-hoc} explanation methods) include, amongst others, Class Saliency Map \cite{Simonyan:ICLR2014}, Grad-CAM \cite{Selvaraju:IJCV2020}, Gradient $\times$ Input \cite{Shrikumar:arxiv2016}, Integrated Gradients \cite{Sundararajan:ICML2017}, Layer-wise Relevance Propagation \cite{Bach:PLOS2015}, Excitation Backpropagation \cite{Zhang:IJCV18}, Guided Backpropagation \cite{Spring:ICLR2015}.
Another set of explanation methods include some aspects of randomness in the heatmap computation: they require generating additional perturbed training data samples or involve solving an \textit{ad-hoc} optimization problem to provide a single heatmap \cite{Ribeiro:KDD2016,Lundberg:NIPS2017,Fong:ICCV2017,Zintgraf:ICLR2017,Chen:ICML2018}.
In this work, we will mainly focus on the first set of methods\footnote{The only exception to this is what we additionally consider the SmoothGrad and VarGrad methods, since they were often used in previous comparative studies.}, because they exploit exactly the same amount of information as is available at prediction time, i.e., a single input data point and the trained neural network's parameters. Typically such explanation methods implement a custom backward pass through the network, or are based on gradient integration.

In previous works, visual explanations were mostly validated on real-world image classification tasks, either through pixel perturbation analyses \cite{Samek:TNNLS2017}, or by using the pixels' relevances as an object detection signal \cite{Zhang:IJCV18}. 
While these evaluations might be justified in a scenario where no ground truth explanation is available for a given task, they could also potentially create a mismatch between the explanation's primary goal (explain the {\it current} decision, which might be also based on the image's context or on dataset biases \cite{Geirhos:Nature2020}), and the evaluation criterion (track the {\it change} in model prediction when perturbing pixels, or \textit{localize} an object's bounding box).

In the present work, we propose instead to evaluate explanations {\it directly} against ground truth object coordinates using a restricted setup of synthetic, albeit realistically rendered, images of 3D geometric shapes. To the best of our knowledge, this is the first ground truth based and realistic testbed for the evaluation of neural network explanations proposed in the literature.
To this end, we leverage a synthetic visual question answering (VQA) task, namely the CLEVR diagnostic dataset \cite{Johnson:CLEVRDiagnosticDataset:2017}, which was initially proposed to diagnose the visual reasoning abilities of VQA models.
We argue that VQA is a more spatially grounded and selective setup for evaluating visual explanations than single-object image classification, since it has more variability: the number of objects, their location and size vary across images, while in image classification the object of interest would often lie in the middle of the image and occupy a great portion of the image. Further, in VQA, not every object present in the image is relevant to the prediction. Indeed the explanation is \textit{modulated} by the question which de-facto selects the important objects for a given prediction (i.e., the heatmap is not solely dependent on the image's content), while in standard image classification tasks every image has a single relevant object which shall be recognized to make the final classification decision.

More precisely, our contributions to the evaluation of XAI methods can be summarized as follows:
\begin{itemize}
\item We create a new large-scale visual question answering dataset, denoted as CLEVR-XAI, made of questions and the corresponding pixel-level ground truth masks, which can be used as a new benchmark for evaluating visual explanations. Our dataset comprises 39,761 simple questions (CLEVR-XAI-simple) and 100,000 complex questions (CLEVR-XAI-complex).
\item We propose two novel quantitative metrics suitable to evaluate visual explanations w.r.t.\ these ground truth masks: relevance mass accuracy and relevance rank accuracy. We expect our new metrics, together with our dataset\footnote{Our dataset and code will be made publicly available upon paper acceptance at \url{https://github.com/ahmedmagdiosman/clevr-xai}.}, will foster the evaluation of XAI methods on a well-defined common ground, which is crucially needed in the XAI research field \cite{Arrieta:InfoFusion2020}.
\item We systematically evaluate and compare ten different explanation methods on our new benchmark and metrics, using a trained Relation Network model \cite{Santoro:NIPS2017} (a state-of-the-art model on the CLEVR task), leading to new findings regarding the quality and properties of XAI methods, which are sometimes contradicting with the conclusions drawn from previous comparative studies in vision XAI.
\end{itemize}

Our work is divided as follows. 
Section \ref{sec:motivation} motivates our approach in relation to previous works. Section \ref{sec:dataset_and_metrics} introduces our new benchmark dataset, CLEVR-XAI as well as our novel quantitative metrics for heatmap evaluation: relevance mass accuracy and relevance rank accuracy. This Section also details the relevance pooling techniques we considered in our work. Section \ref{sec:XAI_methods} defines the XAI methods and their variants we tested in our empirical comparative study of visual explanations. In Section \ref{sec:experiments} and \ref{sec:discussion}, we present our experiments and discuss our results. Finally Section \ref{sec:conclusion} concludes our work.

\section{Previous Work on Evaluating Explanations}\label{sec:motivation}

A widely used approach to evaluate explanations in the vision domain was initially introduced as pixel-flipping or region-perturbation analysis \cite{Bach:PLOS2015,Samek:TNNLS2017} (variants of this analysis were also proposed in subsequent works \cite{Ancona:ICLR2018,Morcos:ICLR2018}).
It consists in repeatedly altering a sequence of pixels (or boxes of pixels) in an input image, accordingly to their {\it relevance} ordering, i.e., their importance for the prediction, and measure the effect of this perturbation on the model's prediction. The higher the effect, measured for example in terms of prediction performance drop, the more accurate was the relevance. 
Additionally, we notice that such evaluation ideally encompasses two baseline perturbation schemes: one random perturbation (corresponding to uninformative relevance), and a brute force search pixel perturbation scheme which optimizes the measured performance metric (the latter representing the best case ordering of the relevance, which can also be seen as a valid explanation and is related to occlusion-based relevance \cite{Zeiler:ECCV2014}).
One potential issue with the pixel perturbation evaluation is that the model might receive as input images that lie outside the actual training data distribution, i.e., out of data manifold samples. This can potentially lead to artifacts and unreliable model predictions.
In contrast, our evaluation approach is based on unmodified input images, from the same distribution as during model training.

Another perturbation-based evaluation technique consists in performing randomizations of the model weights or training data  \cite{Adebayo:NIPS2018} (in the latter case the model is re-trained on the same images but with randomly permuted class labels), and then measure the similarity between the original and the randomized explanations, in order to verify that the explanations are indeed sensitive to the model's parameters and the data generating process. While this type of analyses can serve as an auxiliary sanity check for explanations,
it doesn't constitute a direct assessment of the explanation's quality for a given real prediction. In a similar manner, further work proposes to measure the cosine similarity of explanations along the model's lower layers, when initializing the relevance at a given higher layer randomly, in order to test if backward-propagation based explanations converge to a single direction throughout the propagation process, which would render these explanations effectively independent of higher layers \cite{Sixt:ICML2020}. 
Recent work from \cite{Adebayo:NIPS2020} also modify the training or test data, or the model weights, and then compare the similarity of the resulting modified explanations with the original ones or some ground truth masks.
Lastly, authors of \cite{Hooker:NIPS2019} propose the Remove and Retrain evaluation method. Here a new training dataset is created by replacing the most important pixels in each image by an uninformative value (mean value per channel), accordingly to the pixels' relevance ranking. The model is then re-trained on that modified data, and the prediction accuracy of the re-trained model is compared to a random pixel importance baseline, the higher the decrease w.r.t.\ that baseline the more accurate the explanation.
However, a drawback of all previously mentioned perturbation-based evaluations is that they are based on input data, or on a model, that are different from the model of interest which is actually being deployed and explained at test time. In other words, they introduce a discrepancy between the original model's training/test configurations and the XAI evaluation setup, which can induce misleading or inaccurate results when comparing explanation methods.

A further commonly used approach for evaluating explanations in the visual domain is to use the pixel relevances for weakly supervised object localization (or segmentation) in real-world image classification tasks, e.g., by applying a threshold on the relevances, and then computing the Intersection over Union (IoU), or related metrics (such as the per-pixel accuracy),
w.r.t.\ object bounding box annotations (or segmentation masks) as a measure of relevance accuracy \cite{Simonyan:ICLR2014}. A closely related evaluation consists in verifying that the single pixel with the highest relevance lies inside the object corresponding to the model's predicted class, which is denoted as the pointing game accuracy \cite{Zhang:ECCV2016}. As noticed in previous work a naive baseline pointing to the center of the image already performs well on the latter type of evaluation \cite{Gu:ACCV2019}.
More generally such localization based evaluations could also be misleading, since they assume the model's classification decision is based solely on the object itself, and not on its context or background,
which on real-world image datasets can not be ensured. Indeed it was previously shown that state-of-the-art image classifiers fail to recognize objects in an unexpected location (e.g., a cow on a beach, rather than on grassland), or conversely they identify nonexistent objects solely based on the image's context (a lush hilly landscape is mistakenly recognized as a ``herd of sheep'') \cite{Geirhos:Nature2020,Beery:ECCV2018}. 
Moreover, the IoU type of metric favors a relevance distribution that fully covers the object's surface in the image, while a trained classifier might as well rely on parts of this area to make its decision. The two metrics we will introduce in Section \ref{sec:dataset_and_metrics} do not have this limitation and only assume the ``major'' part of the relevance lies on the object(s) of interest (in terms of mass or rank), but the relevances' spread does not need to fully match the object boundaries.

Besides, most previous works evaluating XAI methods in computer vision consider as a model a convolutional neural network (CNN) which was trained for image classification on the ImageNet dataset \cite{Russakovsky:ILSVRC15}. In this task the goal is to classify images into one object category.
However, as previously mentioned, such type of classification decision using real-world images can be based on various confounding features such as the object's texture, the image's background, dataset biases or image generation artifacts \cite{Geirhos:Nature2020,Lap:Nature19}. Therefore for this task evaluations based on object localization are not reliable. In contrast, we employ a \textit{synthetic} generated dataset of 3D rendered shapes. This allows us to ensure that the objects' background does not contain any information: it is made of the same uniformly colored plane surface for all images. In other words we can assume the relevant information for the considered prediction task is on the objects. Furthermore, we argue that VQA is a more \textit{selective} task than image classification for evaluating explanations. Indeed, for image classification the model acts as a feature detector: it will process all relevant parts of the image independently of the predicted class, up to the final linear layer that will integrate these features and assign a prediction score to each class. Hence, the image processing pipeline is greatly identical for all classes (up to the output layer): all important features in the image will be detected. Therefore it is not surprising that the heatmaps for different classes may look very similar. In contrast, in VQA, the hidden layer representations are modulated by the question, and thus different questions on the same image will induce different heatmaps. This makes VQA an ideal task to evaluate if an explanation can select different objects in the image depending on the question, reflecting the neural network's internal processing of both the image and the specific question. Lastly, the VQA model is naturally trained to deal with several objects present in the input image at the same time, whereas on single-object image classification tasks there is always one object to recognize and classify. Hence the property of ``class-discriminativeness'', which is often stated as a desirable property for an explanation method in the XAI literature (e.g., \cite{Selvaraju:ICCV2017,Smilkov:ICML2017,Gu:ACCV2019}), meaning the ability to focus on the right object when multiple objects are present in the image, might be a slightly overstated goal in the general case of a CNN trained on standard ImageNet object classification (if the CNN was instead trained for multi-label classification, this behavior might however be justified).   

More particularly, our VQA based evaluation benchmark is composed of 10,000 images, each containing 3 to 10 objects with different attributes (8 colors, 2 materials, 2 sizes and 3 shapes) placed at random locations in the image. Objects are also often situated near to one another, or occluded by one another: this forces the model to focus on the objects themselves, and largely reduces object detection strategies relying on the  ``empty space'' around objects, e.g., to recognize the objects' shape. This allows us to generate pixel-level ground truth masks that depend on the target objects of each question (our annotations are more fine-grained than the bounding boxes used in weak object localization). For each image we generate approx.\ 4 simple questions and 10 complex questions (more details on the ground truth generation for each type of question will be given in Section \ref{sec:dataset_and_metrics}).

Closely related to our work is the approach taken in \cite{Oramas:ICLR2019}, which is based on an image classification dataset of synthetic flowers with known discriminative features (these are mainly composed of the petal's color or the stem's color), and where the evaluation consists in calculating the IoU of the heatmaps w.r.t.\ ground truth masks of the corresponding flower components. In their considered task, all flowers have a unique size and each image contains a single flower to classify. Hence our VQA setup has more variability.

Finally, other works relied on the human-based visual assessment of the explanation's quality \cite{Sundararajan:ICML2017,Smilkov:ICML2017,Spring:ICLR2015,Lundberg:NIPS2017,Ribeiro:KDD2016,Zintgraf:ICLR2017,Adebayo:NIPS2020}. While these studies give complementary insights about the usefulness of explanations from an end-user perspective, and contribute to an intuitive understanding of
the properties of visual explanations, they can not replace an objective and systematic automatic evaluation of XAI in computer vision.

Table \ref{table:overview_evaluation_approaches} recapitulates the most commonly used approaches to quantitatively evaluate explanations in the vision domain, and contrast their main properties with our proposed new approach.

\begin{table}
	\begin{center}
		\caption{Evaluations of visual explanations (aka heatmaps) most commonly used in the XAI literature. Through the task we denote the model's training objective (and corresponding dataset). If the evaluation and model training are based on \textit{synthetic} datasets, this enables a tighter control over biases and artefacts (which can't be entirely controlled for in real-world collected images, or in human-generated annotations). We denote as \textit{selective} a learning task where each image contains several objects of potential interest, but which are not all relevant for the prediction, and thus require the model to distinguish and select the relevant one(s). Lastly, we differentiate if the evaluation is based on the \textit{same} model and \textit{same} data as during model training, as opposed to evaluations that rely either on a modified model (due to re-training or parameter randomization), or on modified inputs, and that induce a discrepancy between the training/test and evaluation configurations.
		\vspace*{0.2cm}}
		\label{table:overview_evaluation_approaches}

		\resizebox{1.02\columnwidth}{!}{
			\begin{tabular}{l|l|cccc|l}
				\hline
				Evaluation of XAI                   & Task (dataset)                        & \multicolumn{4}{c}{Properties}                                                    & {proposed/used in }        \\
			    {}                                  & {}                                    & synthetic           & selective           &  same model           &  same data      & {}                         \\
			    \hline
			    Pixel Perturbation                  & {}                                    & \xmark              &   \xmark            &  \cmark              &   \xmark       & \cite{Bach:PLOS2015,Samek:TNNLS2017,Shrikumar:ICML2017,Lundberg:NIPS2017,Fong:ICCV2017,Chen:ICML2018} \\
			    Data Randomization Test             & {}                                    & \xmark              &   \xmark            &  \xmark              &   \xmark       & \cite{Adebayo:NIPS2018,Adebayo:NIPS2020}                         \\
                Model Randomization Test            & object classification                 & \xmark              &   \xmark            &  \xmark              &   \cmark       & \cite{Adebayo:NIPS2018,Adebayo:ICLR2018,Sixt:ICML2020,Adebayo:NIPS2020}                         \\
                Remove and Retrain                  & (ImageNet etc.)                       & \xmark              &   \xmark            &  \xmark              &   \xmark       & \cite{Hooker:NIPS2019}                         \\
                Object Localization/Segmentation    & {}                                    & \xmark              &   \xmark            &  \cmark              &   \cmark       & \cite{Simonyan:ICLR2014,Zhang:ECCV2016,Mahendran:ECCV2016,Selvaraju:ICCV2017,Fong:ICCV2017}                         \\
                Pointing Game                       & {}                                    & \xmark              & \xmark              &  \cmark              &   \cmark       & \cite{Zhang:ECCV2016,Selvaraju:ICCV2017,Fong:ICCV2017}                         \\
                \hline
                Ours: CLEVR-XAI                       & VQA (CLEVR)                         & \cmark              &   \cmark            &  \cmark              &   \cmark       &                          \\
				\hline
			\end{tabular}
		}
	\end{center}
\end{table}

In the next Section we will introduce our CLEVR-XAI benchmark dataset as well as our novel quantitative metrics for evaluating explanations in this testbed. Furthermore, we define the different relevance pooling techniques we considered in our work.

\FloatBarrier

\section{Ground Truth Evaluation of Heatmaps}\label{sec:dataset_and_metrics}

\subsection{A new Benchmark for Visual Explanations: CLEVR-XAI}

Our new benchmark for evaluating explanations in the visual domain was built upon the original CLEVR dataset generator \cite{Johnson:CLEVRDiagnosticDataset:2017}, we call it CLEVR-XAI.

CLEVR is a synthetic VQA task that was designed to diagnose the reasoning abilities of VQA models by avoiding the biases present in real-world human-annotated datasets (such as \cite{Antol:ICCV2015,Goyal:CVPR2017,Agrawal:CVPR2018}), and allowing full control on the data generation pipeline. 
The CLEVR dataset is comprised of 3 splits of training/validation/test sets with  70,000/15,000/15,000 images, and resp.\ 699,989/149,991/149,988 questions (i.e., there are approx.\ 10 questions per image), and the prediction problem is framed as a classification task with 28 possible answers. 
CLEVR images contain 3D objects rendered under various lighting directions and positioned on a grey plane surface, each object having 4 types of attributes (color, material, size, and shape).
The image generation process encompasses the creation of a {\it scene graph} which contains all necessary information to describe the visual scene such as object locations, attributes, and inter-object relations.
Further, the question generation is done via a \textit{functional program} and various program templates. The functional program is made of a sequence of \textit{basic functions} such as querying and comparing attributes, counting objects, and checking the existence of a certain object. 
Once the functional program is built, it can be applied on a scene graph to yield a ground truth answer. 
Further, the program can be used to identify relevant objects along the question's processing pipeline.

In our evaluation framework, a model is first trained on the original CLEVR training set. And subsequently, the explanations obtained on this trained model with various XAI methods are evaluated on the separate CLEVR-XAI evaluation set, which is made of standard CLEVR-type questions that are augmented with the information from our generated ground truth masks (since those masks are not available in the original CLEVR dataset).
Our CLEVR-XAI evaluation set comprises approx.\ 140k questions\footnote{For comparison, the ImageNet image classification tasks contain 100k to 150k test images, and the ImageNet object detection task contains 40k test images \cite{Russakovsky:ILSVRC15}. Hence our CLEVR-XAI dataset can be considered as an evaluation at large scale.}, divided into two types: 
\begin{itemize}
\item 39,761 simple questions (CLEVR-XAI-simple): This subset contains only simple queries about object attributes with no inter-object relations. Hence these questions have only a {\it single} target object that can be used for the ground truth mask generation. 
\item 100,000 complex questions (CLEVR-XAI-complex): This subset contains \textit{all} types of questions as in the original CLEVR task \cite{Johnson:CLEVRDiagnosticDataset:2017}, i.e., they include spatial relations between objects as well as attribute comparison between objects and object counting. Therefore \textit{several} objects in the image can be relevant for such questions and act as ground truths.
\end{itemize}

Both subsets share the same set of 10,000 images (i.e., for each image, there are approx.\ 4 simple questions and 10 complex questions).

In the following we briefly describe the image generation process, which is common to both subsets. Then, for each subset we will detail the question generation process as well as the ground truth masks generation.

\paragraph{Image Generation} To generate the CLEVR-XAI images, we use the same
pipeline as in the CLEVR data generator \cite{Johnson:CLEVRDiagnosticDataset:2017}. Additionally, we use the scene graph
to create a segmentation mask of each object present in the scene (we achieved this
in practice by rendering a secondary image where the light sources are deactivated
and by assigning to each object a unique color). These segmentation masks will serve as a basis for generating the questions’ ground truth masks. 
In total we generated 10,000 images, where each image contains 3 to 10 objects with random attributes and locations.

\subsection{CLEVR-XAI-simple} 

\paragraph{Question Generation} We generate CLEVR-XAI-simple questions in the
same fashion as in the CLEVR data generator \cite{Johnson:CLEVRDiagnosticDataset:2017}, except that for this subset we consider only specific question types where the answer can be inferred by examining a single object in the image.
Our goal for this subset is indeed to consider the most restricted setup possible in order to remove any ambiguity in the ground truth mask generation. For instance, for the question {\it ``What material is the tiny cyan sphere?''}, the relevant object of the question must be the only tiny cyan sphere in the scene. Hence we use only simple questions that query an object's attribute, the corresponding question types are: \texttt{query\_shape}, \texttt{query\_color}, \texttt{query\_size}, \texttt{query\_material}.
For each question type, we randomly sample a question template, and generate this way 4 questions per image. Finally ill-posed questions are discarded (as was done in \cite{Johnson:CLEVRDiagnosticDataset:2017}). This leaves us with
a total of 39,761 questions.

\paragraph{Ground Truth Masks} Using the object segmentation masks created during the image generation step and each question’s functional program, we can automatically identify the single target object of each question. 
With this information, we generate two types of ground truth (GT) masks for evaluating explanations:
\begin{itemize}
    \item \textit{GT Single Object}: The first type of mask we generate is based solely on the target
object’s pixels, which are marked as True, while remaining objects’ pixels, as
well as the background, are set to False.
    \item \textit{GT All Objects}: The second ground truth mask we generate is less discriminative and encompasses all objects’ pixels in the scene, those pixels are set to True, while only the scene’s background is set to False. This latter mask will allow us to perform a weak sanity check on the explanation to verify whether the relevance is indeed assigned to objects and not to the background, since in our synthetic task, by construction, the background shall be uninformative. 
\end{itemize}

\begin{figure*}[!t]
	\small
	\centering
	\begin{tabular}{p{0.27\linewidth}p{0.27\linewidth}p{0.27\linewidth}}
		\toprule
		\hfil Image                                                                                       &           \hfil Question/Answer       &               \hfil  Program                                  \\
		\hfil \raisebox{-\height}{\includegraphics[width=2.9cm,height=2.9cm]{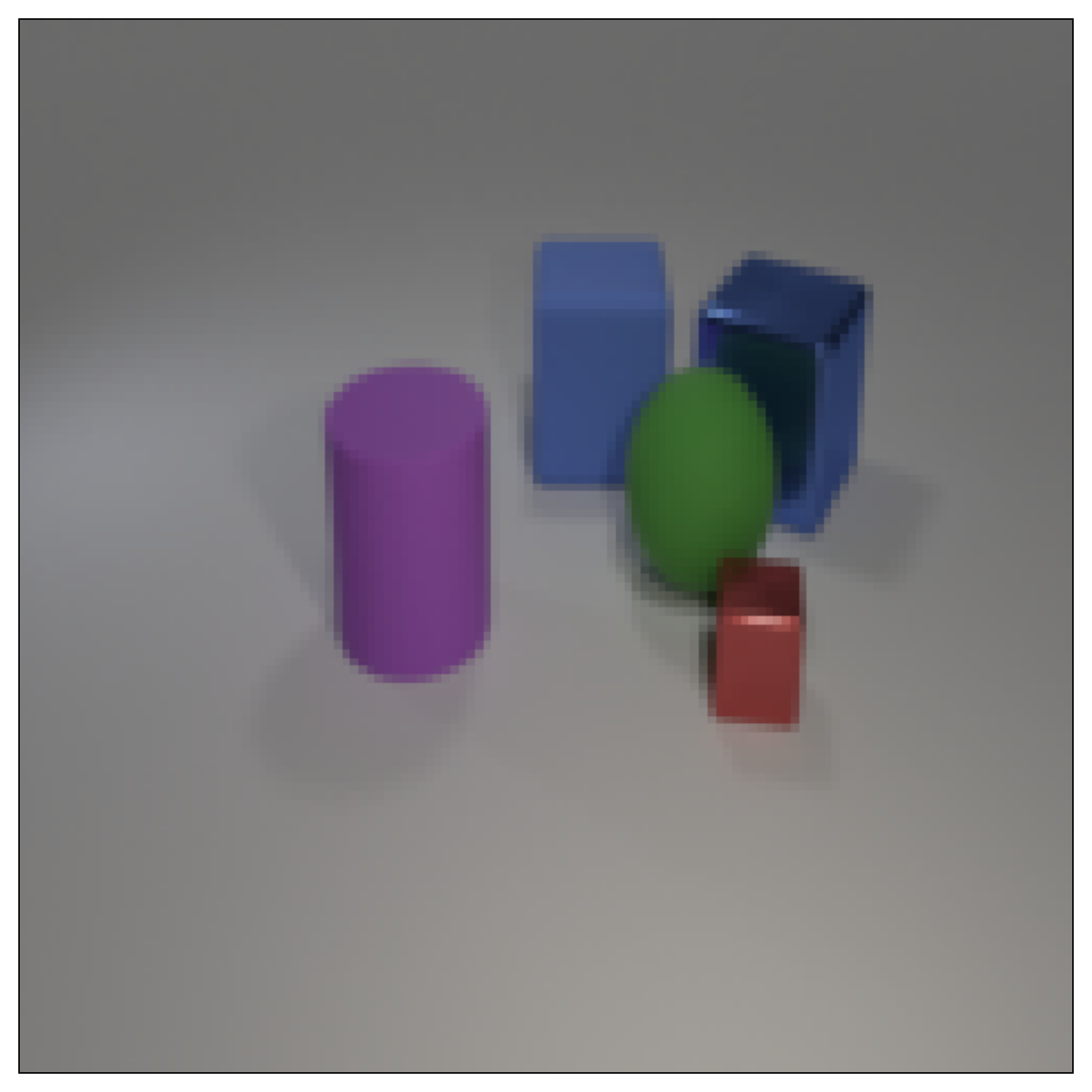}}                                                                              &
		\vspace*{\fill} \noindent{\raggedright \textbf{\scriptsize What is the sphere made of?}\vspace{0.2cm} \\ \textit{rubber}}  \vfill                  &
        \vspace*{\fill} \noindent{\raggedright  {\texttt {\scriptsize [scene, filter\_shape, unique, query\_material]}}} \vfill                                                               \\
        \midrule
        \hfil Ground Truth                                                                                &     \hfil \footnotesize GT Single Object         &      \hfil \footnotesize GT All Objects                 \\
		{} & \hfil \raisebox{-.5\height}{\includegraphics[width=2.9cm]{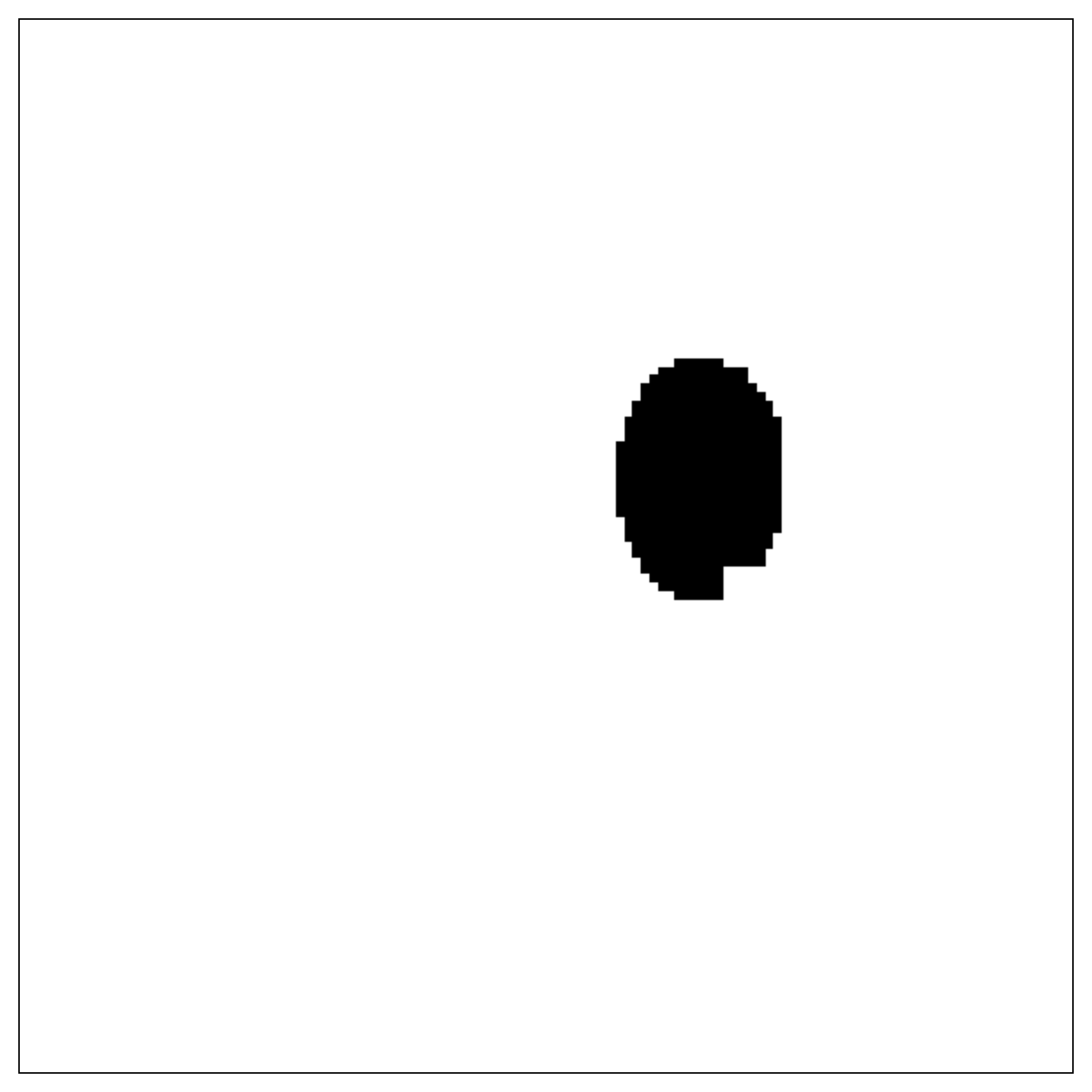}}                                                                                      &
		\hfil \raisebox{-.5\height}{\includegraphics[width=2.9cm]{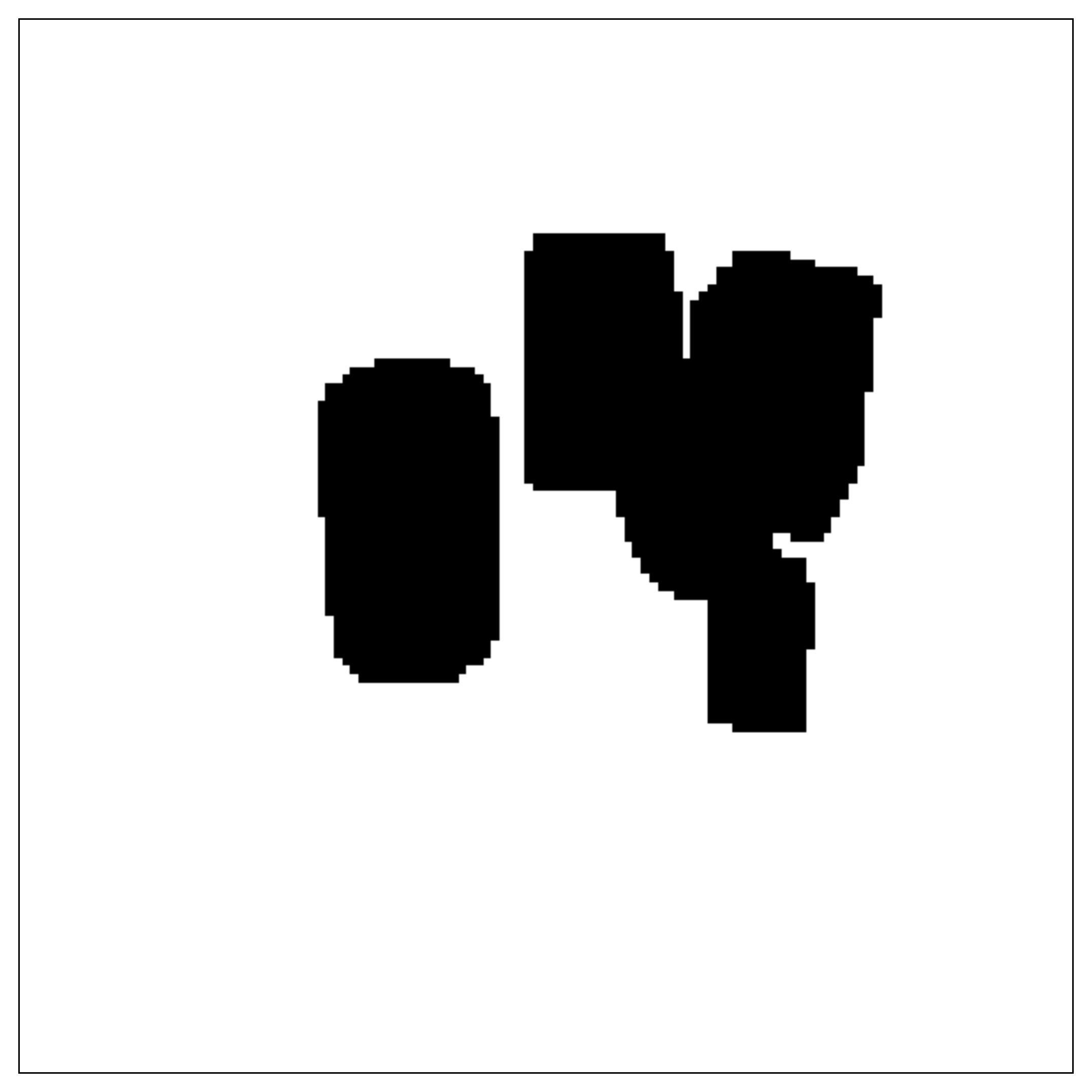}}                                                                                                \\
		\bottomrule
	\end{tabular}
	\caption{Example data point from CLEVR-XAI-simple. We create two ground truths: one containing the single target object of the question, and one containing all objects.}
	\label{fig:clevr-xai-simple_example}
\end{figure*}

Figure~\ref{fig:clevr-xai-simple_example} provides an example CLEVR-XAI-simple question with corresponding ground truth masks. 

\subsection{CLEVR-XAI-complex}

\paragraph{Question Generation} Unlike CLEVR-XAI-simple, CLEVR-XAI-complex questions are generated exactly like in the original CLEVR dataset \cite{Johnson:CLEVRDiagnosticDataset:2017}. We create in this way 10 questions per image, i.e., in total 100,000 questions.
The resulting question distribution is approx.\ the same as in the original CLEVR dataset, i.e., grouped by question family we obtain: 36\% \textit{Query Attribute}, 18\% \textit{Compare Attribute}, 9\% \textit{Compare Integer} (these are questions that compare the sizes of objects' sets), 24\% \textit{Count} and 13\% \textit{Exist} questions.
Hence, the CLEVR-XAI-complex subset of questions comprises far more complex questions than CLEVR-XAI-simple. This includes spatial relationships between objects, attribute comparison as well as logical operators (\texttt{and}, \texttt{or})  in the question formulation. The corresponding functional programs of the questions have a length between 2 and 24 (which means questions are made out of a sequence of up to 24 CLEVR basic functions). 
This drastically complicates the process of finding relevant objects in the scene for building the corresponding ground truth masks.
Since there is no unique way how to define the relevant objects for such complex questions,
we create 3 different ground truth masks per question, each based on different assumptions on how to process the question's functional program.

As introduced earlier, the functional program of each question is made of a sequence of basic functions (this sequence can further be chain- or tree-structured). In this sequence each function takes as input the output of the previous function.
The first function in the program is always the scene (it returns the entire set of objects present in the image), and the last function of the program returns the true answer to the VQA question. 
Input and output of each function can be attribute values, integer or boolean, but most importantly they can also be sets of objects (for a detailed list of all basic functions, their input and output, we refer the reader to the Supplementary Material in \cite{Johnson:CLEVRDiagnosticDataset:2017}). We build our ground truths on the sets of objects returned by the intermediate functions in the functional program.

\paragraph{Ground Truth Masks}
\begin{itemize}
    \item \textit{GT Unique}: This is the most discriminative ground truth mask we create. It is based on the union of the output of all \texttt{unique} functions occurring in the program.
    The function \texttt{unique} is used in the program when one object has to be uniquely identified among a set of objects present in the scene to answer the question. For instance, in the question  {\it ``How many objects are left from the grey cube?''}, there must be a unique grey cube in the image such that the question is not ill-posed. Such type of unique objects are selected to build our \textit{Unique} masks (note that a complex question can have several unique objects, for instance when the question contains a logical operator \texttt{and}/\texttt{or} in the program).
    \item \textit{GT Unique First-non-empty}: To create this ground truth we iterate through the functional program in reverse order and include to the ground truth the first non-empty set of objects we find which is returned by any intermediate function in the program (we exclude the \texttt{scene} function which the first function in the program since it is not discriminative and just returns all the objects of the scene).
    If the program is tree-structured (this can for instance occur when the function contains a logical operator), we aggregate the first non-empty sets from each of the two tree branches.
    Additionally, we incorporate to this ground truth all \texttt{unique} function outputs. In other words, the \textit{GT Unique First-non-empty} set of objects is a superset of the \textit{GT Unique}.
    \item \textit{GT Union}: This is the least discriminative ground truth mask we create that is still related to the question. It simply includes the union of all sets of objects that are returned by every function present in the program (excluding the first function, which is the \texttt{scene} function and just returns all the objects in the scene). Per construction the \textit{GT Union} is a superset of the \textit{GT Unique First-non-empty}.
    \item \textit{GT All Objects}: Like in CLEVR-XAI-simple, this mask contains all objects of the scene, and is independent of the question.
\end{itemize}

\begin{figure*}[!t]
	\small
	\centering
	\begin{tabular}{p{0.24\linewidth}p{0.24\linewidth}p{0.32\linewidth}}
		\toprule
		\hfil Image                                                                                       &           \hfil Question/Answer            &               \hfil  Program                                  \\
		\hfil \raisebox{-\height}{\includegraphics[width=2.9cm,height=2.9cm]{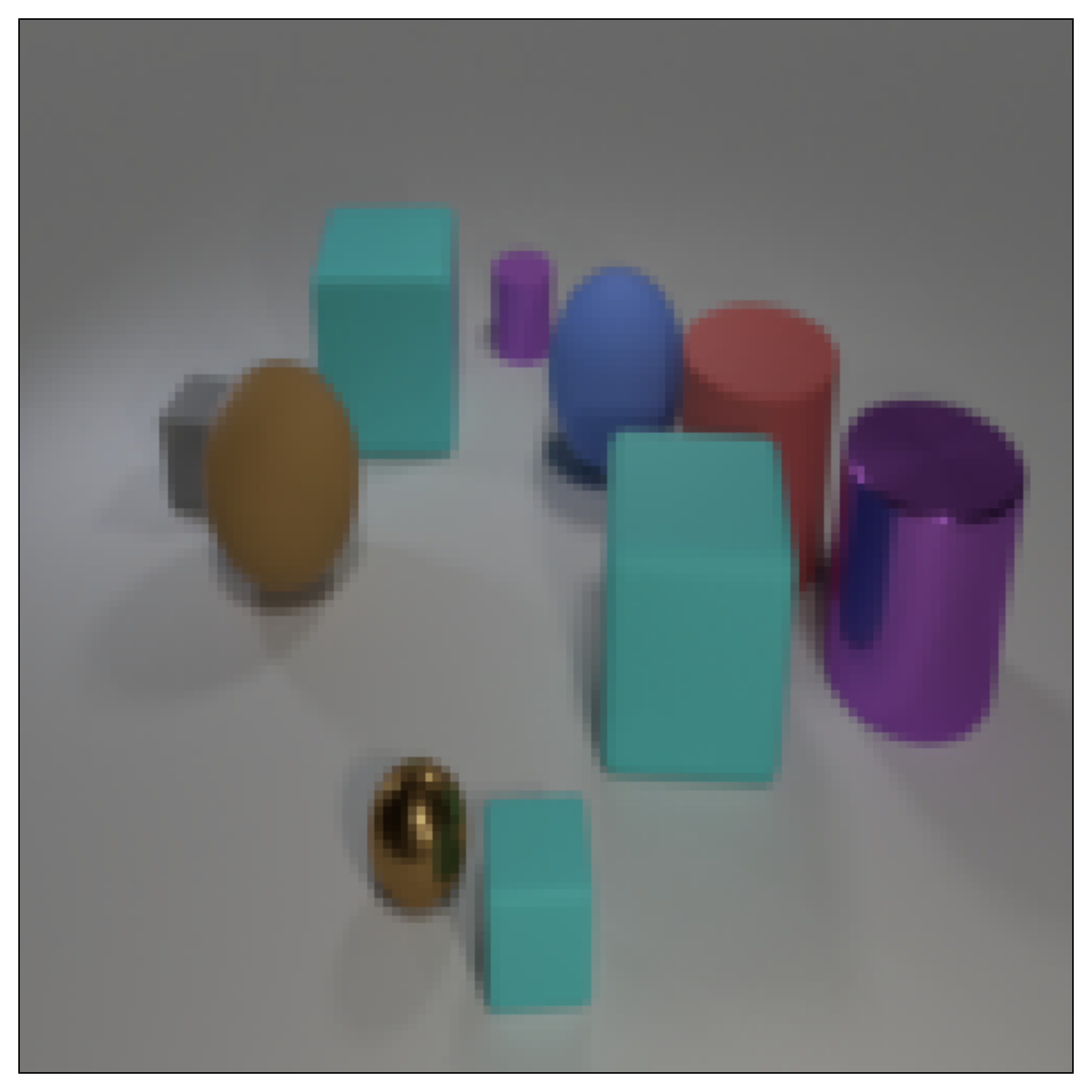}}                                                                              &
		\vspace*{\fill} \noindent{\raggedright \scriptsize \textbf{Is there any other thing that has the same size as the shiny sphere?}\vspace{0.2cm} \\ \textit{yes}}  \vfill                  &
        \vspace*{\fill} \noindent{\raggedright  {\texttt {\scriptsize [scene, filter\_material, filter\_shape, unique, same\_size, exist]}}} \vfill                                                               \\
        \midrule
        \hfil Ground Truth                                                                                &     \hfil \footnotesize GT Unique          &    \hfil  \footnotesize GT Unique First-non-empty                  \\
		{} & \hfil \raisebox{-.5\height}{\includegraphics[width=2.9cm]{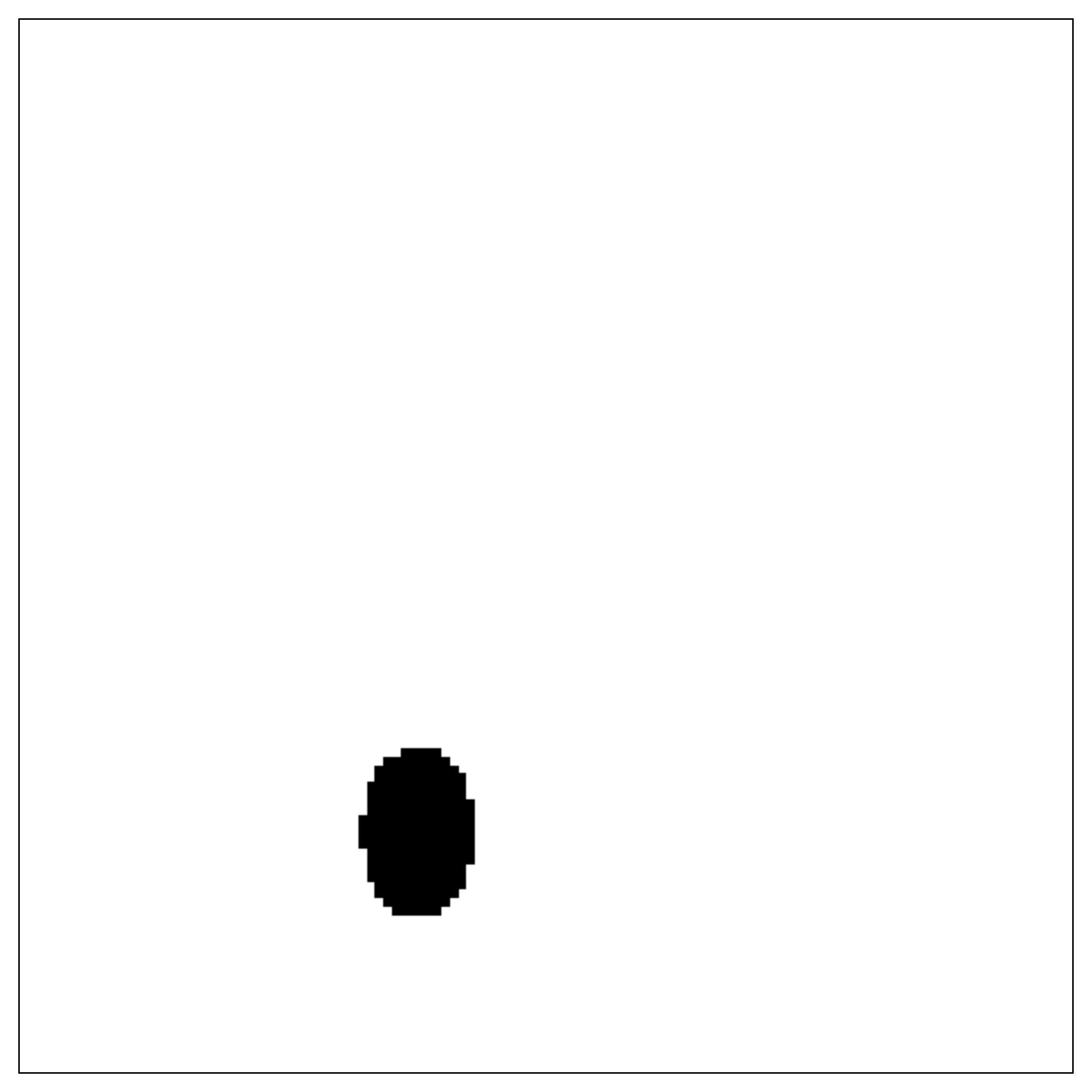}}                                                                                      &
		\hfil \raisebox{-.5\height}{\includegraphics[width=2.9cm]{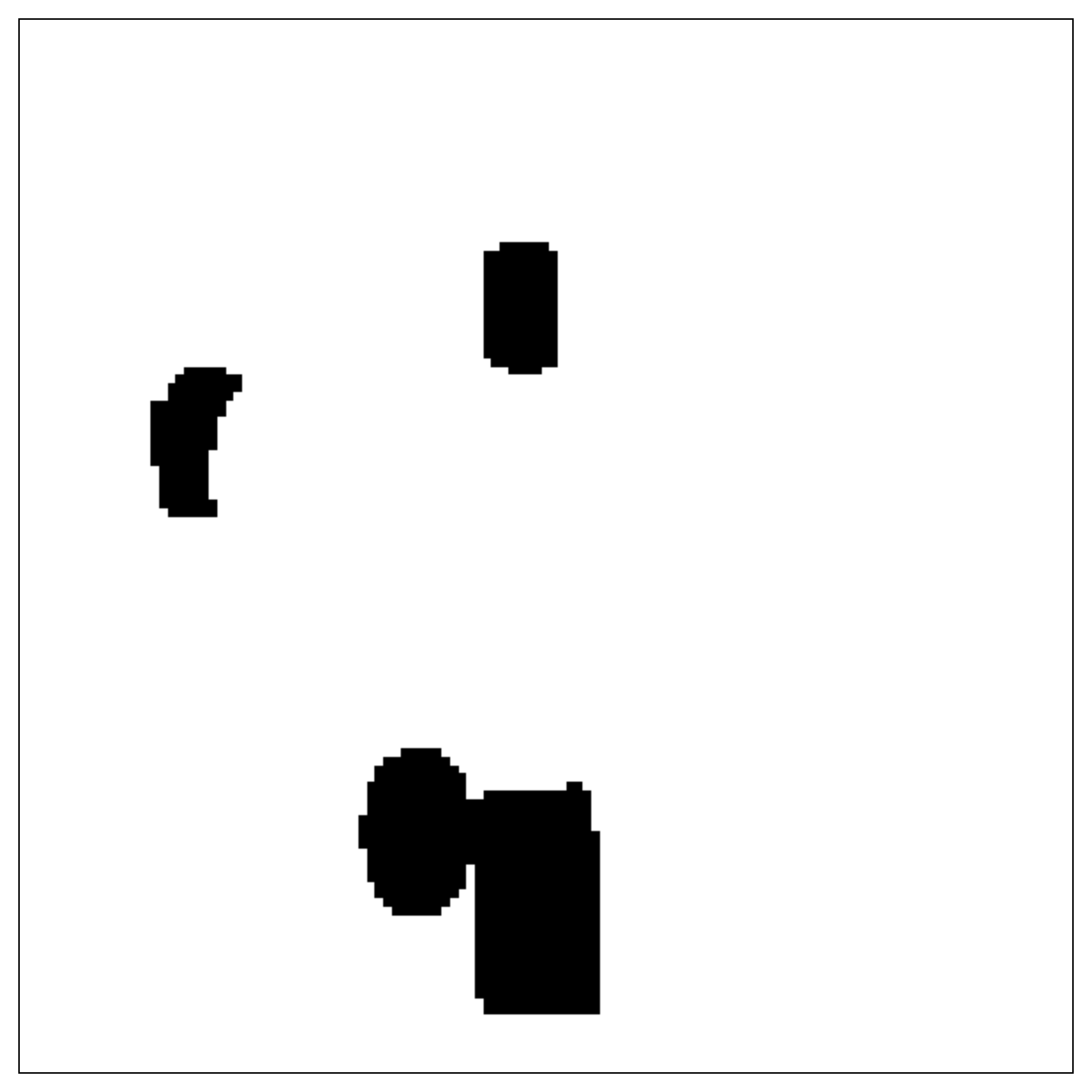}}                                                                                                \\
		{}                                                                                                &     \hfil \footnotesize GT Union & \hfil \footnotesize GT All Objects                                             \\
		{} & \hfil \raisebox{-.5\height}{\includegraphics[width=2.9cm]{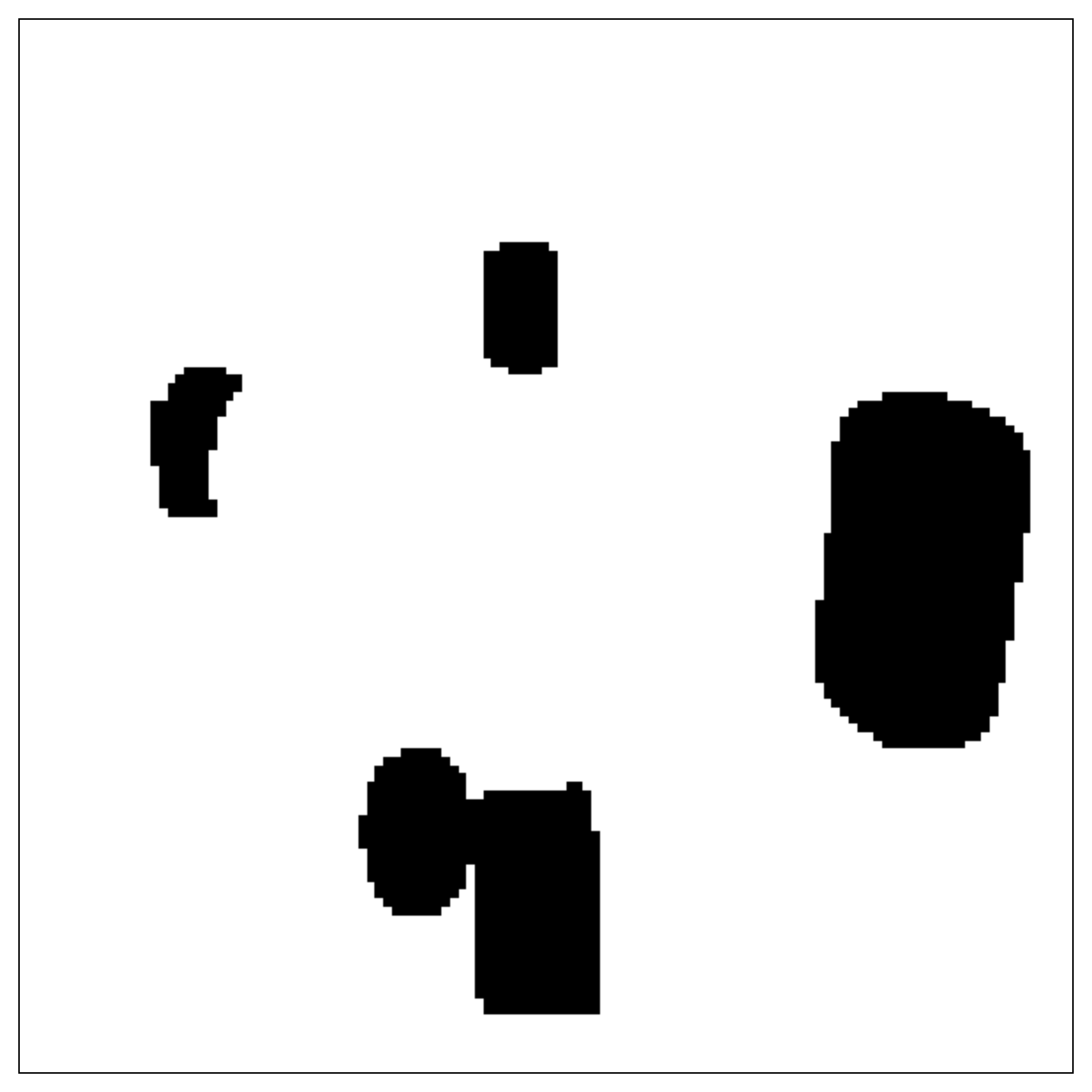}}                                                                                       &
		\hfil \raisebox{-.5\height}{\includegraphics[width=2.9cm]{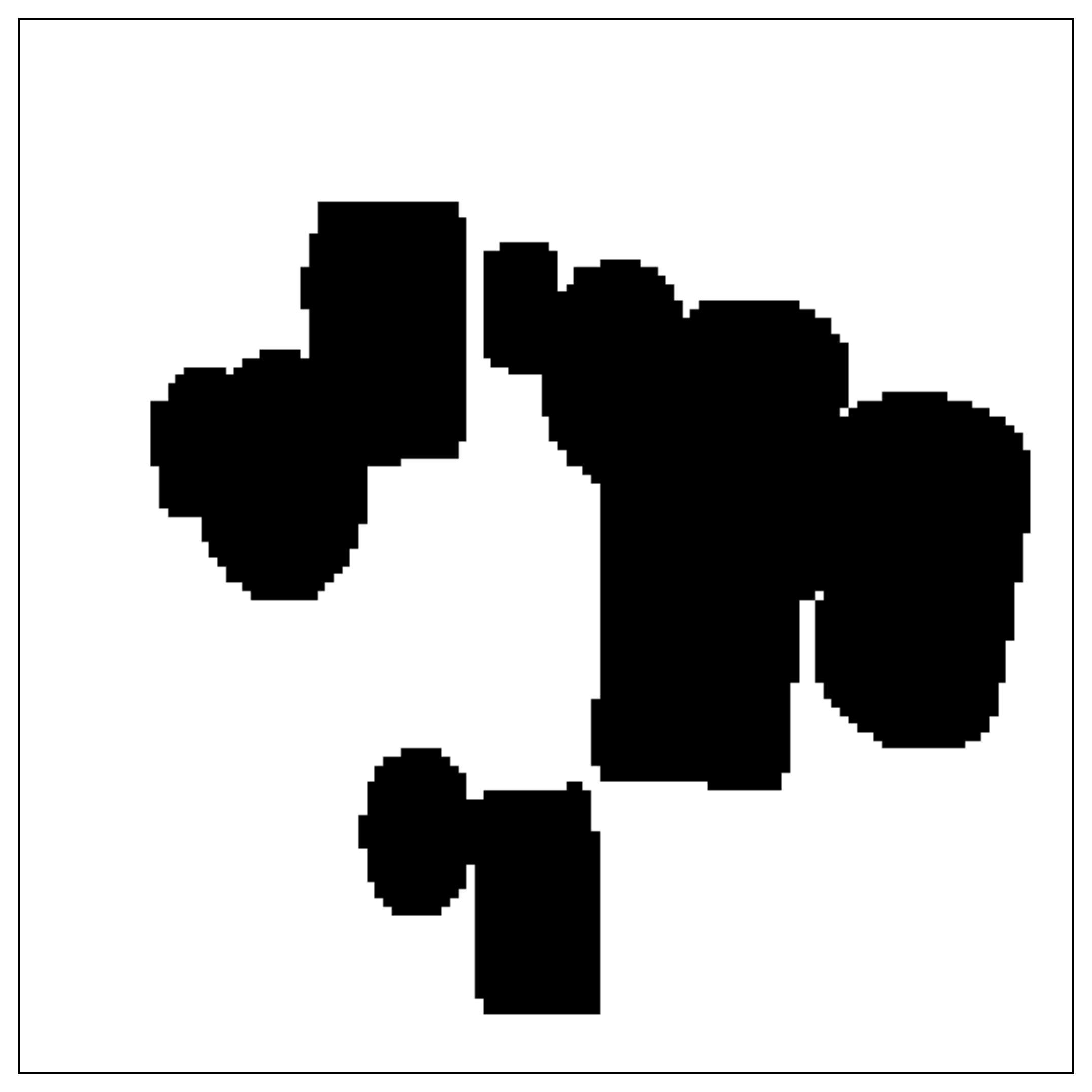}}                                                                                                          \\
		\bottomrule
	\end{tabular}
	\caption{Example data point from CLEVR-XAI-complex. The functional program is used to determine which objects in the scene are considered as ground truths.}
	\label{fig:clevr-xai-complex_example}
\end{figure*}

Figure~\ref{fig:clevr-xai-complex_example} provides an example CLEVR-XAI-complex question with corresponding ground truth masks. The given question is chain-structured (for an example with a tree-structured question see \ref{appendix:CLEVR-XAI_examples}). For this question the output of the \texttt{unique} function is the small brown sphere, it serves to build the \textit{GT Unique}. The \textit{GT Unique First-non-empty} further includes the first non-empty set of objects returned by the functions processed in reversed order of the program, in this case the output of the function \texttt{same\_size}: this corresponds to the other small objects present in the scene (the grey cube, the purple cylinder and the cyan cube). The \textit{GT Union} further contains the output of all intermediate functions in the program, for the given question this means the objects returned by the function \texttt{filter\_material} (which is a superset of the output of the second filter function \texttt{filter\_shape}), i.e., all shiny objects in the scene (thus it additionally encompasses the large purple cylinder). Finally the \textit{GT All Objects} simply contains all objects. The example question further illustrates the fact that objects can be situated next to one another, and partially occluded.

In the \ref{appendix:CLEVR-XAI_statistics} we further provide statistics on the CLEVR-XAI dataset (number of objects and number of pixels per ground truth mask).

\FloatBarrier

\subsection{Relevance Pooling}\label{sec:pooling}

For visual explanations, evaluation w.r.t.\ ground truth masks requires the heatmap to be a 2D positive-valued image with a single channel. Indeed, we care about the spatial location of the relevances rather than their channel (color) values. 
For most XAI methods, the original heatmap mirrors the shape of the model's input (i.e.,\ 3 channel values for RGB images). Thus, there are a number of ways to \textit{pool} the multiple channel values down to a single-channel heatmap. 
Since there is no consensus in the literature on how to perform this pooling step, in our work
we employ ten different pooling techniques. In the experimental Section \ref{sec:experiments}, we will report the results with the pooling technique that led the highest performance for each XAI method and evaluation metric.

The ten pooling techniques we utilize are the following:
\begin{equation*}
	\begin{aligned}
	    &\text{sum,pos:} \;      &R_{pool}        & = max(0,\sum_{i=1}^{C}R_i)        \\
		&\text{sum,abs:} \;      &R_{pool}        & = |\sum_{i=1}^{C}R_i|             \\
		&\text{l1-norm:} \;      &R_{pool}        & = \sum_{i=1}^{C}|R_i|             \\
		&\text{max-norm:} \;     &R_{pool}        & = max(|R_1|, |R_{2}|,...,|R_C|)   \\
		&\text{l2-norm:} \;      &R_{pool}        & = \sqrt{\sum_{i=1}^{C}{R_i}^2}    \\
		&\text{l2-norm-sq:} \;   &R_{pool}        & = \sum_{i=1}^{C}{R_i}^2           \\
		&\text{pos,sum:} \;      &R_{pool}        & = \sum_{i=1}^{C}max(0,R_i)        \\
		&\text{pos,max-norm:} \; &R_{pool}        & = max \big( max(0,R_1), max(0,R_{2}),...,max(0,R_C) \big)   \\
	    &\text{pos,l2-norm:} \;  &R_{pool}        & = \sqrt{\sum_{i=1}^{C}{\big(max(0,R_i)\big)}^2}    \\
	    &\text{pos,l2-norm-sq:} \;   &R_{pool}    & = \sum_{i=1}^{C}{\big(max(0,R_i)\big)}^2           \\
	\end{aligned}
\end{equation*}

where $R_{pool}$ is the pooled relevance at the current pixel, $i$ is the channel index which starts from $1$ to the number of channels $C$ ($C=3$ for RGB images) and  $R_i$ is the relevance value at channel $i$.
Note that the last 4 pooling techniques are just a variant of the first 6 pooling techniques where a ReLU preprocessing is applied to the relevance.

\subsection{Evaluation Metrics: Relevance Mass Accuracy and Relevance Rank Accuracy}\label{sec:metrics}
Given the already pooled heatmap, we propose two metrics to evaluate the accuracy of an explanation. These metrics only assume the ``major'' part of the relevance should lie inside the ground truth mask, either in terms of relevance mass or in terms of relevance ranking (hence our metrics do not require the heatmap to be binarized and to closely match the ground truth surface, as is the case for example for metrics typically used for weak object localization).
Both metrics deliver a number in the range $[0,\, 1]$, the higher this value the more accurate the relevance heatmap.

\paragraph{Relevance Mass Accuracy}The relevance mass accuracy is computed as the ratio of the sum of the relevance values lying within the ground truth mask over the sum of all relevance values over the entire image. In other words, it measures how much ``mass'' does the explanation method give to pixels within the ground truth.
It can be written as:
\begin{equation}
	\text{Mass Accuracy} = \frac{R_{within}}{R_{total}},
\end{equation}
\begin{equation}
    \begin{aligned}
	\mathrm{with}\quad R_{within} &= \sum_{\substack{k=1\\ \text{s.t.} \; p_k \; \in \; GT}}^{|GT|}  R_{p_k} \quad \mathrm{and} \quad
	R_{total}  &= \sum_{\substack{k=1\\}}^{N}  R_{p_k}
    \end{aligned}	
\end{equation}
where $R_{p_k}$ is the relevance value at pixel $p_k$, $GT$ is the set of pixel locations that lie within the ground truth mask, $|GT|$ is the number of pixels in this mask and $N$ is the total number of pixels in the image.

\paragraph{Relevance Rank Accuracy} The relevance rank accuracy measures how much of the high intensity relevances lie within the ground truth. It is calculated through the following steps. Let $K$ be the size of the ground truth mask. Get the $K$ highest relevance values. 
Then count how many of these values lie within the ground truth pixel locations, and divide by the size of the ground truth. Informally, this can be written as:
\begin{equation}
	P_{top \; K} = \{ p_1,p_2,...,p_K \;  | \; R_{p_1} > R_{p_2} > ... > R_{p_K} \}
\end{equation}
where $P_{top \; K}$ is the set of pixels with  relevance values $R_{p_1}, R_{p_2},..., R_{p_K}$ sorted in decreasing order until the $K$-th pixel. 

Then, the rank accuracy is computed as:
\begin{equation}
	\text{Rank Accuracy} = \frac{|{P_{top \; K}} \; \cap \; GT|}{|GT|}
\end{equation}

\section{Explanation Methods}\label{sec:XAI_methods}

In this work we consider ten different explanation methods and evaluate them against our CLEVR-XAI benchmark. In this Section we will briefly describe these methods and detail the specific variants (including hyperparameters) for each method we tested in our experiments. Note that in a general use-case one doesn't have access to an evaluation metric to test an explanation's quality. But since in this work we are interested in thoroughly comparing a set of XAI methods, we allow for tuning\footnote{By tuning we mean that we consider alternative variants (resp.\ hyperparameters) choices for each XAI method, and ultimately report the results for the best performing variant. This is different to how tuning is usually done in machine learning, where a validation set is used for tuning the model's hyperparameters and the results are reported on a separate test set.} those methods against our CLEVR-XAI benchmark. More broadly, we believe that an ``ideal'' XAI method shall not contain any free parameter and ought to be applicable in a straightforward manner without requiring any tuning.

In our evaluation we  mainly focus on \textit{post-hoc} and deterministic XAI methods (as opposed to sampling-based or optimization-based methods), i.e., we employ methods that, given an input and a trained neural network, always deliver the same explanation (the only exception to this is that we additionally consider the sampling-based SmoothGrad and VarGrad methods, since they were often used in previous works). Our XAI comparative study includes the same set of explanation methods as were previously compared in \cite{Adebayo:ICLR2018,Adebayo:NIPS2018,Hooker:NIPS2019}, resp.\ most of the methods that were compared in \cite{Ancona:ICLR2018,Sixt:ICML2020,Adebayo:NIPS2020}. It additionally contains methods which were omitted in some of these works, such as Layer-wise Relevance Propagation (LRP) \cite{Bach:PLOS2015} or Grad-CAM/Guided Grad-CAM \cite{Selvaraju:ICCV2017}.

We now briefly review how each of these methods computes the pixels' relevance values. To this end let us introduce some notations. We refer to $\mathbf{x} \in \mathbb{R}^{C \times H \times W}$ as the input image (with $C$ channels, height $H$ and width $W$), and to $x_i \in \mathbb{R}$ as a single input variable (i.e., $x_i$  corresponds to a specific pixel location in horizontal and vertical direction and to a specific channel).
Further we denote by $f_c(\cdot)$ the model's prediction function for some target class $c$ (i.e., the pre-softmax prediction score for class $c$). In our work we use as the target class the model's predicted answer.

Since our considered model is a VQA classifier, the prediction function $f_c(\cdot)$ takes as input both the image $\mathbf{x}$ and the question $\mathbf{q}$. However, in order to simplify notation, we omit the dependence on the 
question, as in this work we are only interested in the explanations obtained on the image-side. 
An explanation method delivers for each single input variable $x_i$
a scalar {\it relevance} value $R_{i} \in \mathbb{R}$.
These relevances are subsequently pooled along the channel axis (as described in Section \ref{sec:pooling}) to obtain a positive-valued heatmap of size $H\times W$, the latter finally serves as a basis for our quantitative evaluation using the metrics from Section \ref{sec:metrics} and the CLEVR-XAI ground truth masks.

\subsection{Gradient and Gradient$\times$Input}

One method to obtain the relevance $R_{i}$ is based on the partial derivative of the model's prediction function w.r.t.\ the input variable \cite{Simonyan:ICLR2014}, we denote it as Gradient:
\begin{equation}\label{eq:Gradient}
	R_{i} = \tfrac{\partial {f_c}}{\partial x_{i}}(\mathbf{x}) \\
\end{equation}

A similar method further multiplies the partial derivative with the input variable's value \cite{Shrikumar:arxiv2016}, we denote it as Gradient$\times$Input:
\begin{equation}\label{eq:GradientxInput}
	R_{i} = \tfrac{\partial {f_c}}{\partial x_{i}}(\mathbf{x})  \cdot x_i \\
\end{equation}

The partial derivatives can be obtained in one gradient backward pass through the network. 
For both methods, we additionally consider a variant which is based on the \textit{squared} partial derivative \cite{Dimopoulos:1995,Gevrey:2003}, instead of the simple partial derivative (i.e., $\textstyle \big(\tfrac{\partial {f_c}}{\partial x_{i}}(\mathbf{x})\big)^2$, resp.\  $\textstyle \big(\tfrac{\partial {f_c}}{\partial x_{i}}(\mathbf{x})\big)^2 \cdot x_i\;$ for Eq.~\ref{eq:Gradient} and \ref{eq:GradientxInput}).

\subsection{SmoothGrad and VarGrad}

Two other methods we consider are based on the mean or variance of the previous gradient-based relevance, when taking as input noisy versions of the input image.
Defining by $g_i$ the gradient-based relevance of a noisy version $k$ of the input image for variable $x_i$:
\begin{equation}\label{eq:noisy_gradient}
\textstyle	g_i^{[k]} = \tfrac{\partial {f_c}}{\partial x_{i}} \big(\mathbf{x} + (x_{\text{max}} - x_{\text{min}}) \cdot \mathcal{N}^{[k]}(0, \sigma^2)\big) \\
\end{equation}
where $x_{\text{max}}$ and $x_{\text{min}}$ represent the extremal values of the original image's single input variables, and $\mathcal{N}(0, \sigma^2)$ is the sampled additive gaussian noise with standard deviation $\sigma$ (each sample being indexed by $k$).

Then, the SmoothGrad relevance \cite{Smilkov:ICML2017} is computed as:
\begin{equation}\label{eq:SmoothGrad}
\textstyle 	R_{i} = \tfrac{1}{n} \sum_{k=1}^{n} g_i^{[k]} \\
\end{equation}
where $n$ is the number of random samples.

We additionally consider three other variants of the SmoothGrad method, by taking into account a multiplication with the input variable's value and/or squaring the gradient-based relevance (i.e., in Eq.~\ref{eq:SmoothGrad}, we use the following definitions of  $\tilde{g}_i^{[k]}$ in place of $g_i^{[k]}$: $\textstyle  \tilde{g}_i^{[k]} = g_i^{[k]} \cdot x_i \;$,  $\textstyle  \tilde{g}_i^{[k]} = (g_i^{[k]})^2 $ and  $\textstyle  \tilde{g}_i^{[k]} = (g_i^{[k]})^2 \cdot x_i$).

The VarGrad method \cite{Adebayo:ICLR2018,Seo:ICML2018} is similar to SmoothGrad, just that instead of a mean a variance is computed:
\begin{equation}\label{eq:VarGrad}
\textstyle	R_{i} = \tfrac{1}{n} \sum_{k=1}^{n} \big( g_i^{[k]} - (\tfrac{1}{n} \sum_{k=1}^{n} g_i^{[k]}) \big)^2 \\
\end{equation}

Both SmoothGrad and VarGrad have two hyperparameters: the gaussian noise's standard deviation $\sigma$ and the number of samples $n$. We tried the following values in our experiments: $\sigma= [0.02, 0.05, 0.10, 0.20, 0.30]$ and $n=[20, 50, 100, 300]$. (The SmoothGrad authors recommend to use $\sigma$ between 0.10 and 0.20 and $n=50$, according to their qualitative experiments \cite{Smilkov:ICML2017}.)

\subsection{Deconvnet and Guided Backpropagation}

Deconvnet and Guided Backpropagation are two methods that modify the standard gradient backward pass through the neural network's ReLU layers. 

Given a hidden layer neuron which is ReLU activated in the forward pass and defined as $\textstyle x_j^{l+1} = \text{ReLU} (x_j^l) = \text{max} (0, x_j^l)$ where $j$ stands for the neuron's index and $l$ for the layer's index.
Then in the standard gradient backward pass, given $R_j^{l+1} $ the gradient at the output of the ReLU layer, the gradient at the input is computed as:
\begin{equation}\label{eq:ReLU_Standard_Gradient}
	R_{j}^l = \mathbbm{1}_{x_j^l > 0}  \cdot R_j^{l+1} \\
\end{equation}

\noindent Instead in Deconvnet \cite{Zeiler:ECCV2014} the gradient at the input of the ReLU layer is computed as:
\begin{equation}\label{eq:ReLU_Deconvnet}
	R_{j}^l = \text{ReLU} (R_j^{l+1}) \\
\end{equation}

\noindent And in Guided Backpropagation \cite{Spring:ICLR2015} it is computed as:
\begin{equation}\label{eq:ReLU_GuideBackprop}
	R_{j}^l = \text{ReLU} (\mathbbm{1}_{x_j^l > 0}  \cdot R_j^{l+1} ) \\
\end{equation}

The latter method was also re-invented in \cite{Mahendran:ECCV2016} under the name of DeSaliNet. For both methods we tried two variants: applying the modified ReLU gradient backpropagation only to ReLU layers within convolutional layers, or applying it to all ReLUs present in the network.

\subsection{Grad-CAM and Guided Grad-CAM}

Grad-CAM \cite{Selvaraju:ICCV2017,Selvaraju:IJCV2020} is a method derived from the standard gradient-based relevance of the last convolutional layer in the network.

Following the authors' notation, denoting by $A^k$ the features maps of dimension $h \times w $ at the last convolutional layer (k being the feature map's index, and $K$ the total number of feature maps).
First the feature map's gradient-based relevance is global-average-pooled to obtain a feature map's importance weight $\alpha_k$, i.e.:
\begin{equation}\label{eq:Grad-CAM-weight}
\textstyle 	\alpha_k =  \tfrac{1}{h \times w} \sum_{i=1}^{h} \sum_{j=1}^{w}   \tfrac{\partial {f_c}}{\partial A_{ij}^k}(\mathbf{x}) \\
\end{equation}

Then the feature maps are linearly combined and passed through a ReLU processing to finally obtain the Grad-CAM heatmap $H$ of dimension $h \times w $:
\begin{equation}\label{eq:Grad-CAM-heatmap}
\textstyle	H =  \text{ReLU} (\sum_{k=1}^{K} \alpha_k \cdot A^k )\\
\end{equation}

To obtain a heatmap of the same dimension as the input image, the Grad-CAM heatmap is upsampled through bilinear interpolation.
Note that the resulting heatmap is already 2D and positive (thus no need to apply any further pooling technique from Section \ref{sec:pooling} when using this explanation method).
We further observe that Grad-CAM can be seen as a variation of Gradient$\times$Input, which is computed in the last convolutional layer, and where the gradient is first averaged in spatial location.

In our considered model (see Section \ref{sec:Model}), the last convolutional layer has the following structure: \texttt{conv} $\rightarrow$ \texttt{relu} $\rightarrow$ \texttt{batchnorm}. In our experiments we tried using each one of these layers for the Grad-CAM computation (the Grad-CAM authors used models with the following structure: \texttt{conv} $\rightarrow$ \texttt{relu} $\rightarrow$ \texttt{maxpool}, and implemented Grad-CAM on the \texttt{relu} layer). Further we found that for some data points the resulting Grad-CAM heatmaps are zero-valued, i.e., uninformative, we discard these data points from our analysis.

Additionally the authors of \cite{Selvaraju:ICCV2017,Selvaraju:IJCV2020} proposed the Guided Grad-CAM explanation method: it is simply as element-wise multiplication of the Grad-CAM and Guided Backpropagation input image relevances.

\subsection{Integrated Gradients (IG)}

Another method we consider is Integrated Gradients (IG) \cite{Sundararajan:ICML2017}.
The relevances $R_{i}$ of the input variables $x_i$ are based on approximating the following integral through a Riemann sum:
\begin{equation}
	R_{i} = (x_i- x_i^\prime) \cdot \int_{\alpha = 0}^{1} \tfrac{\partial {f_c} (\mathbf{x^\prime} + \alpha \cdot (\mathbf{x} - \mathbf{x^\prime}) )} {\partial x_{i}}  \; d\alpha , \\
\end{equation}
where $\mathbf{x^\prime}$ is a {\it baseline} image that needs to be chosen when applying the method.
Ideally it should be an image with near-zero prediction score and containing no signal.
The IG authors used a zero-valued image, i.e., a black image, for that purpose \cite{Sundararajan:ICML2017}.
In our experiments we also used a zero-valued baseline. Further we experimented with two other baselines: the mean image, as well as the mean channel values over the flattened images (which we computed over the CLEVR training set).
Since the CLEVR images contain a grey background, this resulted in two grey-valued baselines. 

Another choice to be made when employing the IG method is the number of steps used for integral approximation. We tried the values [300, 1000, 3000, 10000, 30000] for each image separately until we reached a sufficient precision for the integral approximation, and we used the Riemann sum with midpoint rule for numerical integration. 
Indeed one can compute the precision of the integral approximation
by exploiting the IG's completeness property:
\begin{equation}
	\textstyle \sum_{i} R_{i} = f_c(\mathbf{x}) -  f_c(\mathbf{x^\prime}) \\
\end{equation}

In our experiments we ensure that the resulting relative error is lower than 0.01 and discard from our analysis data points with higher errors (the original authors suggest a value of 0.05).

\subsection{Layer-wise Relevance Propagation (LRP)} 

Another method we consider is Layer-wise Relevance Propagation (LRP). It was initially introduced in \cite{Bach:PLOS2015}, and later justified via Deep Taylor Decomposition in \cite{Montavon:PR2017,Montavon:DSP2018}.
It consists in redistributing the model's prediction score $f_c(\mathbf{x})$ via a custom backward pass through the network that follows a local conservation principle. 
During  this backward pass each neuron in the network gets assigned its own \textit{relevance} value, up to the input layer neurons (i.e., the input variables).

In practice, through ReLU layers the relevance is backpropagated as the identity, while weighted linear connections (such as convolutional and fully-connected layers) serve to redistribute the relevance in {\it proportion} to neuron contributions from the forward pass. 
More precisely, given a linear layer with the forward pass equation $z_j = \sum_i z_i \cdot w_{ij} + b_j$, then two main LRP rules can be used to compute the neuron's relevance $R_i$, given the relevances of the connected higher-layer neurons $R_j$.
One is the $\epsilon$-rule:
\begin{equation}
	\textstyle R_{i} =  \sum_j \tfrac{z_i \cdot w_{ij}}{z_j \, + \, \epsilon \cdot {\text sign} (z_j)} \; \cdot R_j , \\
\end{equation}
where $\epsilon$ is a stabilizer (we use $\epsilon=0.001$). The other is
the $\alpha_{1},\beta_{0}$-rule:
\begin{equation}
	\textstyle R_{i} =  \sum_j \tfrac{(z_i \cdot w_{ij})^{+}}  {\sum_k (z_k \cdot w_{kj})^{+} + (b_j)^{+}} \; \cdot R_j , \\
\end{equation}
where $(\cdot)^{+}$ denotes the $\text{max}(0,\cdot)$ operation.
Note that when using the latter LRP propagation rule in all layers of the network LRP is equivalent to Excitation Backpropagation \cite{Zhang:ECCV2016,Zhang:IJCV18}.

We experimented with different LRP rules configurations: either using each rule uniformly in all layers of the network, or a combination of both rules. In the latter case we employed the $\epsilon$-rule in the classifier layers of the network (i.e., the last 3 layers of the network, see model description in Section \ref{sec:Model}), or else in all fully-connected layers (i.e., including the Relation Network module of the network), applying the $\alpha_{1},\beta_{0}$-rule in the remaining layers.
Additionally, we considered using a specific LRP rule for the input layer (namely the box-rule, the $w^2$-rule and the flat-rule, for a definition of these rules we refer to \cite{Montavon:ExplAIBook2020}), as well as subsuming consecutive convolutional/fully-connected and batchnorm layers into a single linear layer before applying LRP (according to a recommendation in \cite{Montavon:ExplAIBook2020}), otherwise we used for batchnorm layers the same rule as in the convolutional layers.

\section{Experiments}\label{sec:experiments}

\subsection{Model}\label{sec:Model}

The model we consider for the evaluation of XAI methods is a Relation Network (RN) model \cite{Santoro:NIPS2017}.

There are two reasons why we choose this model: i) it is a state-of-the-art model on the CLEVR dataset \cite{Johnson:CLEVRDiagnosticDataset:2017}, ii) it is a simple architecture made of very common layers (convolutional layers, batchnorm layers, fully-connected layers, element-wise summation, ReLU activations, plus an LSTM for processing the question). The latter is very important to not obfuscate the evaluation of explanation methods by the complexity of the neural network model, and first establish the suitability of XAI methods w.r.t. ground truths on widely used and standard neural network components.

More precisely, the RN model comprises a 4-layer CNN to extract feature maps from the image.
Then, the pixels in the last convolutional feature maps are pair-wise concatenated with the question representation obtained from the last hidden state of an LSTM.
These representations are passed through a 4-layer MLP of fully-connected layers, summed-up in place, and finally fed to a 3-layer MLP of fully-connected layers for classification.

Since the original authors \cite{Santoro:NIPS2017} did not release their code, we re-implement their model and train it from scratch on the original CLEVR training set  \cite{Johnson:CLEVRDiagnosticDataset:2017}, using the validation set for hyperparameter tuning. As an image preprocessing step, we rescaled the images  to the range [0, 1] (we did not center the data since the RN model contains batchnorm layers), further our input images have 3 channels (RGB).
For more details on the model specification and training we refer to the \ref{appendix:model}. Note also that since the RN model takes input images of size 128$\times$128, while the original CLEVR images have size 320$\times$480, we need to resize our Ground Truth masks to this former size too, see the \ref{appendix:GT_resizing} for more details on this step.

Our trained RN model reaches $93.3\%$ accuracy on the CLEVR test set (the original authors report $95.5\%$ accuracy). 

On the CLEVR-XAI-simple questions it reaches $98.2\%$ accuracy, 
and on CLEVR-XAI-complex $93.3\%$. In the latter case the performance differs per question type: for questions involving counting (\texttt{count}, \texttt{equal\_integer}, \texttt{greater\_than}, \texttt{less\_than}) the accuracy is lower, between $72.2\%$ and $93.0\%$, 
while on the remaining question types (\texttt{equal\_color}, \texttt{equal\_material}, \texttt{equal\_shape}, \texttt{equal\_size}, \texttt{query\_color}, \texttt{query\_material}, \texttt{query\_shape}, \texttt{query\_size}, \texttt{exist}) the accuracy is higher, between $94.2\%$ and $98.3\%$ (which is consistent with results from\cite{Santoro:NIPS2017}).

We now move to the evaluation of XAI methods on the CLEVR-XAI-simple and the CLEVR-XAI-complex questions, respectively.

\subsection{Simple Questions: CLEVR-XAI-simple}\label{sec:experiments-simple-questions}

\paragraph{Tuning XAI Methods}

In our evaluation paradigm  we start by tuning each XAI method, for each type of relevance accuracy metric separately, using the \textit{GT Single Object}. 
By tuning we mean selecting for each XAI method the variant/hyperparameter choice, as well as the pooling technique, that lead to the highest relevance accuracy.
The resulting method's best variant and pooling technique will further be applied on the CLEVR-XAI-complex questions and other ground truth masks to compare XAI methods.

We employ the \textit{GT Single Object} for tuning since it's the most restricted and controlled setup among all questions and ground truth masks we created.
Indeed here we consider only the case of simple queries about an object's attribute.
And, in this configuration, we may reasonably assume that a good explanation method shall focus
its relevance on the single target object of the question which must have triggered the model's response.

For this tuning step, we only use data points which are correctly classified by the model in order to avoid noise in the evaluation stemming from false predictions, and consider all pooling techniques that were introduced in Section \ref{sec:pooling} (i.e., we allow each XAI method to rely on a different pooling).

The resulting relevance accuracies are reported in Table \ref{table:mass_accuracy_simple_question_one_object} and \ref{table:rank_accuracy_simple_question_one_object}, left column (note that we list the XAI methods in the same order for all Tables in this work: this order generally reflects the relative quality of the XAI methods across our experiments). A detailed listing of the corresponding XAI methods variants/hyperparameters and pooling techniques can be found in \ref{appendix:XAI_methods_hyperparameters}.

A nice finding of this tuning step is that, for every XAI method, tuning with either \textit{mass} accuracy or \textit{rank} accuracy leads to the same best variant/hyperparameter choice for each method\footnote{Except for some gradient-based relevance techniques where using the unsquared gradient sometimes lead to better results for the rank accuracy metric, and for the Grad-CAM method where the used layer for the relevance computation differs per accuracy metric.}. 
Also the relative ordering of the XAI methods remains largely consistent across the two accuracy metrics.
This highlights the fact that both metrics are well-suited to measure the quality of an explanation method.
Concerning the most adequate pooling technique per accuracy metric, we find the $\text{l2-norm-sq}$ and  $\text{pos,l2-norm-sq}$ poolings to be the most appropriate poolings for the mass accuracy, and the $\text{max-norm}$ pooling stood out as the appropriate pooling for all XAI methods for the rank accuracy metric.

In the following we briefly review for each XAI method which of its variant led the best relevance accuracy, and in the subsequent subsection we will compare the relative performance of XAI methods.

For LRP applying the $\alpha_{1},\beta_{0}$-rule in every hidden layer and the box-rule in the input layer led to best results. Note that this setup is conform to the configuration
brought forward by the Deep Taylor Decomposition framework \cite{Montavon:PR2017,Montavon:DSP2018}. Although more recently a composite application of the LRP rules was shown to perform better in the particular case of CNNs for image classification for discriminativity reasons (according to \cite{Montavon:ExplAIBook2020,KohIJCNN20}), in our VQA setup the composite application of the $\alpha_{1},\beta_{0}$-rule and $\epsilon$-rule, where the latter is used on the classifier layers of the network (i.e., the last three layers) in addition to using the box-rule in the input layer, either led to slightly inferior results or results on-par with the best setup mentioned earlier, while applying the $\epsilon$-rule in all fully-connected layers and the $\alpha_{1},\beta_{0}$-rule in all convolutional layers consistently lead to worse results. Moreover, using the $\epsilon$-rule uniformly in all layers performed very poorly, although this rule was shown to perform well in other contexts, such as CNNs for text classification or recurrent neural networks for sequential data \cite{Arras:PLOSONE2017,Arras:17,Poerner:ACL2018,ArrXAI19}, which is probably due to the specific nature and spatial structure of neural network models in computer vision compared to other data domains. Lastly, we didn't find that combining the batchnorm layers with their nearest convolutional/fully-connected layer before applying LRP led to an improvement in the relevance's accuracy.

Further, we note that simply applying the LRP $\alpha_{1},\beta_{0}$-rule in all layers of the network, including the input layer, already performs very well: this is the result we report under Excitation Backprop \cite{Zhang:ECCV2016} in our tables. This particular LRP setup was also shown to perform very well in the context of neural network pruning for image classification CNNs \cite{Yeom:Arxiv2019}.

For IG the baseline image with mean channel values led to the best results, nearly followed by the mean image baseline. However, using a zero-valued baseline (as was done by the original authors in \cite{Sundararajan:ICML2017}) leads to very poor results: the resulting mass and rank accuracy are only 0.27 in this case. This highlights the crucial importance of choosing an appropriate baseline for the IG method (for a general discussion on the impact of the baseline choice we refer to \cite{sturmfels2020visualizing}).

Guided Backprop performed around 0.10 accuracy better  when applying the modified backpropagation rule to all ReLU layers in the network, rather than only to ReLUs within convolutional layers. For Deconvnet that's the opposite: applying the modification only within the CNN part of the network led to approx. 0.15 better accuracy (in the other case the accuracy is near zero).

Grad-CAM alone performs generally very poorly (and a bit worse when using the \texttt{conv} layer for its computation, rather than the \texttt{relu} or \texttt{batchnorm} layer). When augmented with Guided Backprop the Grad-CAM performance is highly boosted: Guided Grad-CAM reached its best performance when Grad-CAM was computed on the \texttt{relu} layer, and the Guided Backprop's backpropragation rule was applied to all ReLUs in the network.

SmoothGrad and VarGrad both reached their highest performance when the number of samples was set to 300 and the noise level to 0.05. If we had taken instead no more than 100 samples and a noise level of at least 0.10 (as suggested by the original authors in \cite{Smilkov:ICML2017}), then these methods would have lost at least 0.10 mass accuracy and 0.08 rank accuracy, rendering them not much better than the simpler Gradient and  Gradient$\times$Input methods. This illustrates the high sensitivity of the sampling-based XAI methods to the specific hyperparameter choices.

\paragraph{Comparing XAI Methods}

Now when comparing the XAI methods to one another using Table \ref{table:mass_accuracy_simple_question_one_object} and \ref{table:rank_accuracy_simple_question_one_object}, left column, as well as the results in subsequent Tables (which will be more detailed in the remainder of this work)  we find that Grad-CAM and Deconvnet are by far the worst performing methods. Guided Grad-CAM performs relatively well, but this is likely mainly due to the benefit added by its Guided Backprop component.
That Grad-CAM performs even worse that Gradient$\times$Input is not really surprising, since Grad-CAM's internal computation resembles an \textit{averaged} Gradient times Input in the CNN's last layer feature maps: the gradient information is diluted through the average pooling operation of Grad-CAM.

Gradient$\times$Input performs slightly worse that Gradient. In general we expect the former to be very sensitive to the specific image preprocessing: if we had instead center our data to have a mean of zero (in our case we merely rescaled the image values to the range [0,1]), then the images' background would have a near zero value, and consequently the Gradient$\times$Input relevance would have been near zero in these image regions too, which would have probably boosted the Gradient$\times$Input's performance.

SmoothGrad and VarGrad are a bit better than their unsmoothed counterparts Gradient and Gradient$\times$Input, however this is due to hyperparameter finetuning, and in a practical application it is unclear how to choose/tune those hyperparameters. Further we observe that VarGrad is generally no better than SmoothGrad.

Two methods that perform remarkably well in our evaluation, although their computation is very efficient (a single backward pass through the network) are LRP and Guided Backprop, though LRP generally outperforms the latter. 

IG also performs well, however it's computation is more expensive (multiple backward passes are required), and in general, it remains an open question how to choose a good baseline image for this XAI method. Note also that if we compare the IG and LRP results on the most selective setups we considered, i.e., the \textit{GT Single Object} on CLEVR-XAI-simple and the \textit{GT Unique} and \textit{GT Unique First-non-empty} on CLEVR-XAI-complex, then the LRP method is at least 0.11 better in mass accuracy, and at least 0.08 better in rank accuracy, than IG.

Now if we look at the variances of the accuracy results, and if we exclude from this analysis the Deconvnet and the Grad-CAM methods (since these methods anyway perform poorly), then the two methods with the lowest variance are LRP and Guided Backprop, while the gradient-based methods show the highest variances. IG, SmoothGrad, VarGrad, Gradient and Gradient$\times$Input show especially high variance in terms of relevance mass accuracy, while SmoothGrad has particularly high variance on the relevance rank accuracy.
The only exception to this behavior for the IG method is when we look at the \textit{GT All Objects} results in Table~\ref{table:mass_accuracy_simple_question_all_objects}-\ref{table:rank_accuracy_simple_question_all_objects} and \ref{table:mass_accuracy_complex_question_all_GTs}-\ref{table:rank_accuracy_complex_question_all_GTs}: here LRP and IG both have the lowest variance, and also perform on-par in terms of relevance accuracy mean and median.

More generally, we observe that for the mass metric the variances are higher than for the rank metric. 
This is to be expected since for the rank metric only the pixels' relevance ordering matters, and if a few outliers exist in the relevance heatmap with very high values which are located outside the ground truth mask, this metric will be less affected by those noisy values. In this way the rank accuracy will rather reflect the global ordering pattern of the relevance distribution. On the contrary the mass accuracy will be more sensitive to reveal the local position of a few pixels with very high relevance values.

Additionally to the relevance accuracy results presented here, in the \ref{appendix:heatmaps} we provide some further qualitative results, namely example heatmaps for a few data points from our CLEVR-XAI dataset.

\paragraph{Impact of Model Confidence and Question Difficulty}

We explored whether the relevance accuracy of an XAI method depends on the model's prediction confidence, and the difficulty of the considered VQA question.

In the first case we selected data points for which the model's predicted softmax probability is very high (higher than 0.99999), this gives us the results in Table \ref{table:mass_accuracy_simple_question_one_object} and \ref{table:rank_accuracy_simple_question_one_object}, middle column.

For the second analysis we retained only questions for which the \textit{GT Single Object} mask has at least 1000 pixels. This excluded objects of the category \textit{small} (in the CLEVR task objects generally have two possible size attributes: large or small). We hereby also seek to exclude \textit{large} objects that may be occluded, which would render the corresponding question more difficult to answer (the CLEVR scenes indeed often contain occluded objects). Our results for this setup are reported in Table \ref{table:mass_accuracy_simple_question_one_object} and \ref{table:rank_accuracy_simple_question_one_object}, right column.

We observe that in both cases the relevance accuracy generally tends to increase (i.e., moving from the left to the middle column, or from the left to right column increases the accuracy's mean and median), while the accuracy's standard deviation decreases. This demonstrates that the relevance accuracy is not a static value that only depends on the XAI method's quality, but also reflects the model's confidence in the prediction or the difficulty of a given question.

\paragraph{Validating our Evaluation Approach}

So far in our evaluation we have assumed that the ``major'' part of the relevance shall lie within an object for a VQA question about that object. This assumption may potentially be questioned or challenged.
Or at least there is one type of question where the model's predicted answer can only be found on the object itself: these are queries about an object's color.
If the role of an XAI method is to highlight the image region that triggered the model's answer, then we can safely expect the relevance to focus on the \textit{GT Single Object} mask for such types of questions.

To analyze this particular use case we conducted an experiment where this time we tune the XAI methods on the \textit{GT Single Object} using correctly predicted data points for which the question type is restricted to be \texttt{query\_color}. This gives us the results in Table \ref{table:mass_accuracy_simple_question_one_object_query_color} and \ref{table:rank_accuracy_simple_question_one_object_query_color} (which are to be compared with the left column of Table \ref{table:mass_accuracy_simple_question_one_object} and \ref{table:rank_accuracy_simple_question_one_object}).

We find that this setup generally doesn't affect the best variant/hyperparameter choice and pooling technique for each XAI method\footnote{The only noticeable difference we observe is for the LRP method: this time the composite application of the $\alpha_{1},\beta_{0}$-rule and $\epsilon$-rule, where the latter is applied to the classifier layers of the network (i.e., the last three layers) performs on-par with a uniform application of the $\alpha_{1},\beta_{0}$-rule on all hidden layers, in both cases using the box-rule for the input layer. While when considering all question types the composite configuration performed worse than the uniform application of the $\alpha_{1},\beta_{0}$-rule.}, and also doesn't modify much the absolute accuracy results (e.g., for the three best performing XAI methods, the accuracy increase is no more than 0.02).

This experiment demonstrates that the VQA model mainly focuses on the target object of the question, no matter which type of question is being asked, and validates our prior assumption that the target objects' of the questions can be utilized as ground truth masks to evaluate XAI methods.

\paragraph{Weak Sanity Check}

Now that we have compared and evaluated the XAI methods on the \textit{GT Single Object}, we report their performance on the \textit{GT All Objects} masks.
We recall that the latter masks are not selective and independent of the question: they simply contain all the objects in the scene.
By using the latter GT we can perform a weak sanity check on the XAI methods: how much relevance lies on the objects at all vs. the background. Given that our VQA task is controlled, the background shall be uninformative to the prediction and consequently very little relevance shall be assigned to it.
Our results are reported in Table \ref{table:mass_accuracy_simple_question_all_objects} and \ref{table:rank_accuracy_simple_question_all_objects}.

For this experiment, IG, Excitation Backprop and LRP perform best: these methods assign more than 90\% of the relevance's mass on objects, and more than 72\% of the most relevant pixels lie within objects. Further, for these methods, the standard deviation is more than halfed compared to the \textit{GT Single Object} (Table \ref{table:mass_accuracy_simple_question_one_object} and \ref{table:rank_accuracy_simple_question_one_object}, left column).
This indicates that for these methods, when the relevance is not located on the target object of the question, then it is mainly  situated on other objects and not on the background.
There is also a large gap between these methods and the remaining XAI methods: other methods present a mean mass accuracy of at most 0.73 and mean rank accuracy of at most 0.49.
Finally, overall, the relative ordering of the XAI methods w.r.t. the accuracies on this weaker sanity check remains largely unchanged.





\begin{table}
	\begin{center}
		\caption{ Relevance \textit{mass} accuracy on \textbf{CLEVR-XAI-simple}, using the single target object of the question as Ground Truth, i.e., \textit{GT Single Object}.
		          We consider 3 sets of questions: all correctly predicted, correctly predicted with high confidence (softmax proba $>$ 0.99999), and correctly predicted with large target object (greater than 1000 pixels).}
		\label{table:mass_accuracy_simple_question_one_object}

		\resizebox{0.99\columnwidth}{!}{
			\begin{tabular}{l|cc|cc|cc}
				\hline
				\textbf{}                                               & \multicolumn{2}{c}{correctly predicted}       & \multicolumn{2}{c}{correctly predicted}       & \multicolumn{2}{c}{correctly predicted}   \\
				\textbf{Relevance Mass Accuracy}                        & \multicolumn{2}{c}{\textbf{all}}                       & \multicolumn{2}{c}{\textbf{\& proba$>$0.99999}}     & \multicolumn{2}{c}{\textbf{\& nb pixels$>$1000}}   \\
			    \textbf{}                                               & mean (std)    & median                        & mean (std)        & median                    & mean (std)   & median    \\
			 	\hline
                LRP \cite{Bach:PLOS2015}                                &  \textbf{0.85} (0.17)  &  \textbf{0.91}     &  \textbf{0.90} (0.09)  &  \textbf{0.93}     &  \textbf{0.91} (0.09)  &  \textbf{0.94}    \\   
                Excitation Backprop \cite{Zhang:ECCV2016}               &  0.80 (0.20)  &  0.87     &  0.85 (0.14)  &  0.90     &  0.88 (0.13)  &  0.92    \\   
                IG \cite{Sundararajan:ICML2017}                         &  0.71 (0.27)  &  0.81     &  0.75 (0.25)  &  0.85     &  0.80 (0.24)  &  0.90    \\   
                Guided Backprop \cite{Spring:ICLR2015}                  &  0.58 (0.20)  &  0.62     &  0.63 (0.16)  &  0.66     &  0.76 (0.13)  &  0.78    \\   
                Guided Grad-CAM \cite{Selvaraju:ICCV2017}               &  0.58 (0.24)  &  0.63     &  0.64 (0.21)  &  0.68     &  0.83 (0.13)  &  0.86    \\   
                SmoothGrad \cite{Smilkov:ICML2017}                      &  0.60 (0.33)  &  0.69     &  0.61 (0.33)  &  0.72     &  0.80 (0.25)  &  0.91    \\   
                VarGrad \cite{Adebayo:ICLR2018}                         &  0.58 (0.34)  &  0.68     &  0.60 (0.34)  &  0.71     &  0.80 (0.25)  &  0.91    \\   
                Gradient \cite{Simonyan:ICLR2014}                       &  0.49 (0.35)  &  0.49     &  0.51 (0.34)  &  0.53     &  0.82 (0.25)  &  0.93    \\   
                Gradient$\times$Input \cite{Shrikumar:arxiv2016}        &  0.43 (0.34)  &  0.37     &  0.44 (0.34)  &  0.39     &  0.77 (0.27)  &  0.89    \\   
                Deconvnet \cite{Zeiler:ECCV2014}                        &  0.18 (0.16)  &  0.13     &  0.18 (0.16)  &  0.13     &  0.42 (0.21)  &  0.41    \\   
                Grad-CAM \cite{Selvaraju:ICCV2017}                      &  0.09 (0.10)  &  0.05     &  0.10 (0.10)  &  0.07     &  0.29 (0.10)  &  0.28    \\   
				\hline
			\end{tabular}
		}
	\end{center}
\end{table}

\begin{table}
	\begin{center}
		\caption{ Relevance \textit{rank} accuracy on \textbf{CLEVR-XAI-simple}, using the single target object of the question as Ground Truth, i.e., \textit{GT Single Object}.
		          We consider 3 sets of questions: all correctly predicted, correctly predicted with high confidence (softmax proba $>$ 0.99999), and correctly predicted with large target object (greater than 1000 pixels).}
		\label{table:rank_accuracy_simple_question_one_object}

		\resizebox{0.99\columnwidth}{!}{
			\begin{tabular}{l|cc|cc|cc}
				\hline
				\textbf{}                                               & \multicolumn{2}{c}{correctly predicted}       & \multicolumn{2}{c}{correctly predicted}       & \multicolumn{2}{c}{correctly predicted}   \\
				\textbf{Relevance Rank Accuracy}                        & \multicolumn{2}{c}{\textbf{all}}                       & \multicolumn{2}{c}{\textbf{\& proba$>$0.99999}}     & \multicolumn{2}{c}{\textbf{\& nb pixels$>$1000}}   \\
			    \textbf{}                                               & mean (std)    & median                        & mean (std)    & median                        & mean (std)   & median    \\
			 	\hline

                LRP \cite{Bach:PLOS2015}                                &  \textbf{0.72} (0.15)  &  \textbf{0.75}     &  \textbf{0.76} (0.10)  &  \textbf{0.79}     &  \textbf{0.73} (0.11)  &  0.75    \\   
                Excitation Backprop \cite{Zhang:ECCV2016}               &  0.69 (0.18)  &  0.74     &  0.74 (0.13)  &  0.78     &  0.73 (0.13)  &  \textbf{0.76}    \\   
                IG \cite{Sundararajan:ICML2017}                         &  0.53 (0.18)  &  0.55     &  0.56 (0.17)  &  0.59     &  0.65 (0.13)  &  0.67    \\   
                Guided Backprop \cite{Spring:ICLR2015}                  &  0.53 (0.14)  &  0.55     &  0.57 (0.11)  &  0.58     &  0.63 (0.09)  &  0.64    \\   
                Guided Grad-CAM \cite{Selvaraju:ICCV2017}               &  0.52 (0.17)  &  0.55     &  0.56 (0.14)  &  0.59     &  0.68 (0.10)  &  0.69    \\   
                SmoothGrad \cite{Smilkov:ICML2017}                      &  0.49 (0.19)  &  0.53     &  0.51 (0.18)  &  0.56     &  0.58 (0.15)  &  0.61    \\   
                VarGrad \cite{Adebayo:ICLR2018}                         &  0.46 (0.24)  &  0.51     &  0.47 (0.24)  &  0.53     &  0.64 (0.16)  &  0.67    \\   
                Gradient \cite{Simonyan:ICLR2014}                       &  0.34 (0.18)  &  0.34     &  0.35 (0.18)  &  0.36     &  0.55 (0.12)  &  0.56    \\   
                Gradient$\times$Input \cite{Shrikumar:arxiv2016}        &  0.31 (0.17)  &  0.30     &  0.31 (0.17)  &  0.31     &  0.52 (0.12)  &  0.53    \\   
                Deconvnet \cite{Zeiler:ECCV2014}                        &  0.21 (0.15)  &  0.18     &  0.21 (0.15)  &  0.18     &  0.37 (0.15)  &  0.38    \\   
                Grad-CAM \cite{Selvaraju:ICCV2017}                      &  0.17 (0.23)  &  0.01     &  0.19 (0.24)  &  0.06     &  0.55 (0.20)  &  0.58    \\   
				\hline
			\end{tabular}
		}
	\end{center}
\end{table}

\FloatBarrier

\begin{table}
	\begin{center}
		\caption{ Relevance \textit{mass} accuracy on \textbf{CLEVR-XAI-simple}, using the single target object of the question as Ground Truth, i.e., \textit{GT Single Object}.
		          Here we consider only questions that are correctly predicted and query an object's color.}
		\label{table:mass_accuracy_simple_question_one_object_query_color}

		\resizebox{0.5\columnwidth}{!}{
			\begin{tabular}{l|cc}
				\hline
				\textbf{}                                           & \multicolumn{2}{c}{correctly predicted}   \\
				\textbf{Relevance Mass Accuracy}                    & \multicolumn{2}{c}{\textbf{\& query color}}        \\
			    \textbf{}                                           & mean (std)    & median                    \\
			 	\hline

                LRP \cite{Bach:PLOS2015}                                &  \textbf{0.87} (0.15)  & \textbf{0.92}    \\   
                Excitation Backprop \cite{Zhang:ECCV2016}               &  0.82 (0.18)  &  0.89    \\   
                IG \cite{Sundararajan:ICML2017}                         &  0.70 (0.28)  &  0.79    \\   
                Guided Backprop \cite{Spring:ICLR2015}                  &  0.64 (0.16)  &  0.67    \\   
                Guided Grad-CAM \cite{Selvaraju:ICCV2017}               &  0.60 (0.24)  &  0.66    \\   
                SmoothGrad \cite{Smilkov:ICML2017}                      &  0.55 (0.35)  &  0.64    \\   
                VarGrad \cite{Adebayo:ICLR2018}                         &  0.54 (0.35)  &  0.63    \\   
                Gradient \cite{Simonyan:ICLR2014}                       &  0.49 (0.35)  &  0.48    \\   
                Gradient$\times$Input \cite{Shrikumar:arxiv2016}        &  0.42 (0.35)  &  0.35    \\   
                Deconvnet \cite{Zeiler:ECCV2014}                        &  0.24 (0.18)  &  0.20    \\   
                Grad-CAM \cite{Selvaraju:ICCV2017}                      &  0.09 (0.10)  &  0.05    \\   
				\hline
			\end{tabular}
		}
	\end{center}
\end{table}

\begin{table}
	\begin{center}
		\caption{ Relevance \textit{rank} accuracy on \textbf{CLEVR-XAI-simple}, using the single target object of the question as Ground Truth, i.e., \textit{GT Single Object}.
		          Here we consider only questions that are correctly predicted and query an object's color.}
		\label{table:rank_accuracy_simple_question_one_object_query_color}

		\resizebox{0.5\columnwidth}{!}{
			\begin{tabular}{l|cc}
				\hline
				\textbf{}                                           & \multicolumn{2}{c}{correctly predicted}   \\
				\textbf{Relevance Rank Accuracy}                    & \multicolumn{2}{c}{\textbf{\& query color}}        \\
			    \textbf{}                                           & mean (std)    & median                    \\
			 	\hline

                LRP \cite{Bach:PLOS2015}                                &  \textbf{0.73} (0.13)  &  \textbf{0.76}    \\   
                Excitation Backprop \cite{Zhang:ECCV2016}               &  0.71 (0.16)  &  0.76    \\   
                IG \cite{Sundararajan:ICML2017}                         &  0.54 (0.18)  &  0.55    \\   
                Guided Backprop \cite{Spring:ICLR2015}                  &  0.56 (0.11)  &  0.58    \\   
                Guided Grad-CAM \cite{Selvaraju:ICCV2017}               &  0.53 (0.18)  &  0.57    \\   
                SmoothGrad \cite{Smilkov:ICML2017}                      &  0.49 (0.20)  &  0.53    \\   
                VarGrad \cite{Adebayo:ICLR2018}                         &  0.43 (0.24)  &  0.48    \\   
                Gradient \cite{Simonyan:ICLR2014}                       &  0.37 (0.18)  &  0.37    \\   
                Gradient$\times$Input \cite{Shrikumar:arxiv2016}        &  0.33 (0.18)  &  0.33    \\   
                Deconvnet \cite{Zeiler:ECCV2014}                        &  0.26 (0.16)  &  0.25    \\   
                Grad-CAM \cite{Selvaraju:ICCV2017}                      &  0.16 (0.23)  &  0.00    \\   
				\hline
			\end{tabular}
		}
	\end{center}
\end{table}

\FloatBarrier

\begin{table}
	\begin{center}
		\caption{ Relevance \textit{mass} accuracy on \textbf{CLEVR-XAI-simple}, using all objects in the image as Ground Truth, i.e., \textit{GT All Objects}.
		          We consider all correctly predicted questions.}
		\label{table:mass_accuracy_simple_question_all_objects}

		\resizebox{0.5\columnwidth}{!}{
			\begin{tabular}{l|cc}
				\hline
				\textbf{}                                           & \multicolumn{2}{c}{all correctly predicted}   \\
				\textbf{Relevance Mass Accuracy}                    & \multicolumn{2}{c}{{\textbf{GT All Objects}}}        \\
			    \textbf{}                                           & mean (std)    & median                    \\
			 	\hline

                LRP \cite{Bach:PLOS2015}                                &  0.96 (0.03)  &  0.97    \\   
                Excitation Backprop \cite{Zhang:ECCV2016}               &  0.90 (0.10)  &  0.94    \\   
                IG \cite{Sundararajan:ICML2017}                         &  \textbf{0.97} (0.03)  &  \textbf{0.98}    \\   
                Guided Backprop \cite{Spring:ICLR2015}                  &  0.72 (0.12)  &  0.74    \\   
                Guided Grad-CAM \cite{Selvaraju:ICCV2017}               &  0.73 (0.14)  &  0.74    \\   
                SmoothGrad \cite{Smilkov:ICML2017}                      &  0.73 (0.25)  &  0.81    \\   
                VarGrad \cite{Adebayo:ICLR2018}                         &  0.72 (0.26)  &  0.80    \\   
                Gradient \cite{Simonyan:ICLR2014}                       &  0.66 (0.29)  &  0.73    \\   
                Gradient$\times$Input \cite{Shrikumar:arxiv2016}        &  0.56 (0.31)  &  0.58    \\   
                Deconvnet \cite{Zeiler:ECCV2014}                        &  0.29 (0.17)  &  0.26    \\   
                Grad-CAM \cite{Selvaraju:ICCV2017}                      &  0.32 (0.11)  &  0.33    \\   
				\hline
			\end{tabular}
		}
	\end{center}
\end{table}

\begin{table}
	\begin{center}
		\caption{ Relevance \textit{rank} accuracy on \textbf{CLEVR-XAI-simple}, using all objects in the image as Ground Truth, i.e., \textit{GT All Objects}.
		          We consider all correctly predicted questions.}
		\label{table:rank_accuracy_simple_question_all_objects}

		\resizebox{0.5\columnwidth}{!}{
			\begin{tabular}{l|cc}
				\hline
				\textbf{}                                           & \multicolumn{2}{c}{all correctly predicted}   \\
				\textbf{Relevance Rank Accuracy}                    & \multicolumn{2}{c}{{\textbf{GT All Objects}}}        \\
			    \textbf{}                                           & mean (std)    & median                    \\
			 	\hline

                LRP \cite{Bach:PLOS2015}                                &  0.74 (0.08)  &  0.75    \\   
                Excitation Backprop \cite{Zhang:ECCV2016}               & \textbf{0.76} (0.08)  &  \textbf{0.77}    \\   
                IG \cite{Sundararajan:ICML2017}                         &  0.72 (0.07)  &  0.72    \\   
                Guided Backprop \cite{Spring:ICLR2015}                  &  0.48 (0.09)  &  0.48    \\   
                Guided Grad-CAM \cite{Selvaraju:ICCV2017}               &  0.49 (0.10)  &  0.49    \\   
                SmoothGrad \cite{Smilkov:ICML2017}                      &  0.34 (0.12)  &  0.32    \\   
                VarGrad \cite{Adebayo:ICLR2018}                         &  0.29 (0.15)  &  0.26    \\   
                Gradient \cite{Simonyan:ICLR2014}                       &  0.36 (0.09)  &  0.36    \\   
                Gradient$\times$Input \cite{Shrikumar:arxiv2016}        &  0.33 (0.09)  &  0.33    \\   
                Deconvnet \cite{Zeiler:ECCV2014}                        &  0.23 (0.09)  &  0.22    \\   
                Grad-CAM \cite{Selvaraju:ICCV2017}                      &  0.39 (0.15)  &  0.40    \\   
				\hline
			\end{tabular}
		}
	\end{center}
\end{table}

\FloatBarrier

\subsection{Complex Questions: CLEVR-XAI-complex}\label{sec:experiments-complex-questions}

Using for each XAI method the best performing variant as was previously tuned on the simple questions \textit{GT Single Object}, we report in this Section their accuracy on the different Ground Truths masks we generated for complex questions.
Further, we consider only questions that are correctly predicted and that do not involving counting (since the VQA model was shown to perform less good in terms of prediction performance on these types of questions, see \ref{sec:Model}) in order to avoid noise in the evaluation of the XAI methods.
Additionally we discard questions of the type \texttt{exist}, where the true answer is \texttt{no}, since for these questions obviously no object in the scene matches the ground truth answer, and hence such questions could be ambiguous to use for the XAI evaluation.
Our results are reported in Table \ref{table:mass_accuracy_complex_question_all_GTs} and \ref{table:rank_accuracy_complex_question_all_GTs}.

\paragraph{Impact of Ground Truth Selectivity}
We observe that the more selective the Ground Truth, i.e., the less objects it contains, the lower the relevance accuracy. However interestingly, even for the most selective Ground Truth, i.e., \textit{GT Unique}, the relevance accuracy is still high in absolute value, and only less than 0.08 accuracy lower than when using the \textit{GT Single Object} on simple questions, for the XAI methods LRP, IG and Guided Backprop.
This highlights the fact that even on complex questions these XAI methods are able to reveal the unique objects in the scene that are important for a given VQA question. 
Further, the relative ordering of the XAI methods across the different Ground Truths remains largely unchanged.
This illustrates the fact that our CLEVR-XAI-complex set of Ground Truths is also appropriate to evaluate and compare XAI methods.

\paragraph{Comparison with Simple Questions}

We note that the relative ordering of the XAI methods on complex questions is also largely consistent with their ordering on the simple questions.
This supports the fact that both CLEVR-XAI-simple and CLEVR-XAI-complex subsets of questions are pertinent for delivering insights on 
which objects in the scene the XAI methods are focusing when explaining a given prediction.

\begin{table}
	\begin{center}
		\caption{ Relevance \textit{mass} accuracy on \textbf{CLEVR-XAI-complex}, using different types of Ground Truths.
		          We consider all correctly predicted questions that do not involve counting.}
		\label{table:mass_accuracy_complex_question_all_GTs}

		\resizebox{1.05\columnwidth}{!}{
			\begin{tabular}{l|cc|cc|cc|cc}
				\hline
				\textbf{}                                           & \multicolumn{8}{c}{correctly predicted \& no counting}  \\
				\textbf{Relevance Mass Accuracy}                    & \multicolumn{2}{c}{\footnotesize {\textbf{GT Unique}}}    & \multicolumn{2}{c}{\footnotesize {\textbf{GT Unique First-non-empty}}}    & \multicolumn{2}{c}{\footnotesize {\textbf{GT Union}}} & \multicolumn{2}{c}{\footnotesize {\textbf{GT All Objects}}}   \\
			    \textbf{}                                           & mean (std)    & median    & mean (std)    & median    & mean (std)   & median   & mean (std)   & median \\
			 	\hline
                LRP \cite{Bach:PLOS2015}                                &  \textbf{0.82} (0.19)  &  \textbf{0.90}     &  \textbf{0.84} (0.17)  &  \textbf{0.90}     &  \textbf{0.91} (0.12)  &  \textbf{0.95}    &  0.96 (0.03)  &  0.97 \\   
                Excitation Backprop \cite{Zhang:ECCV2016}               &  0.78 (0.21)  &  0.85     &  0.80 (0.19)  &  0.86     &  0.86 (0.15)  &  0.91    &  0.91 (0.08)  &  0.94 \\   
                IG \cite{Sundararajan:ICML2017}                         &  0.67 (0.29)  &  0.77     &  0.70 (0.27)  &  0.79     &  0.87 (0.19)  &  0.95    &  \textbf{0.97} (0.03)  &  \textbf{0.98} \\   
                Guided Backprop \cite{Spring:ICLR2015}                  &  0.65 (0.17)  &  0.68     &  0.65 (0.16)  &  0.69     &  0.71 (0.14)  &  0.74    &  0.75 (0.10)  &  0.76 \\   
                Guided Grad-CAM \cite{Selvaraju:ICCV2017}               &  0.64 (0.20)  &  0.69     &  0.65 (0.20)  &  0.69     &  0.71 (0.16)  &  0.74    &  0.76 (0.11)  &  0.77 \\   
                SmoothGrad \cite{Smilkov:ICML2017}                      &  0.49 (0.34)  &  0.52     &  0.51 (0.33)  &  0.55     &  0.61 (0.30)  &  0.69    &  0.65 (0.28)  &  0.73 \\   
                VarGrad \cite{Adebayo:ICLR2018}                         &  0.48 (0.34)  &  0.51     &  0.50 (0.33)  &  0.54     &  0.60 (0.30)  &  0.68    &  0.64 (0.29)  &  0.72 \\   
                Gradient \cite{Simonyan:ICLR2014}                       &  0.45 (0.34)  &  0.43     &  0.46 (0.33)  &  0.45     &  0.56 (0.31)  &  0.60    &  0.62 (0.29)  &  0.67 \\   
                Gradient$\times$Input \cite{Shrikumar:arxiv2016}        &  0.39 (0.33)  &  0.31     &  0.40 (0.32)  &  0.33     &  0.48 (0.31)  &  0.46    &  0.53 (0.30)  &  0.53 \\   
                Deconvnet \cite{Zeiler:ECCV2014}                        &  0.17 (0.14)  &  0.13     &  0.18 (0.14)  &  0.14     &  0.24 (0.16)  &  0.20    &  0.27 (0.16)  &  0.24 \\   
                Grad-CAM \cite{Selvaraju:ICCV2017}                      &  0.15 (0.12)  &  0.12     &  0.16 (0.12)  &  0.13     &  0.27 (0.14)  &  0.27    &  0.34 (0.11)  &  0.35 \\   
				\hline
			\end{tabular}
		}
	\end{center}
\end{table}

\begin{table}
	\begin{center}
		\caption{ Relevance \textit{rank} accuracy on \textbf{CLEVR-XAI-complex}, using different types of Ground Truths.
		          We consider all correctly predicted questions that do not involve counting.}
		\label{table:rank_accuracy_complex_question_all_GTs}

		\resizebox{1.05\columnwidth}{!}{
			\begin{tabular}{l|cc|cc|cc|cc}
				\hline
				\textbf{}                                           & \multicolumn{8}{c}{correctly predicted \& no counting}  \\
				\textbf{Relevance Rank Accuracy}                    & \multicolumn{2}{c}{\footnotesize {\textbf{GT Unique}}}    & \multicolumn{2}{c}{\footnotesize {\textbf{GT Unique First-non-empty}}}    & \multicolumn{2}{c}{\footnotesize {\textbf{GT Union}}} & \multicolumn{2}{c}{\footnotesize {\textbf{GT All Objects}}}   \\
			    \textbf{}                                           &  mean (std)    & median    & mean (std)    & median    & mean (std)   & median    & mean (std)   & median \\
			 	\hline

                LRP \cite{Bach:PLOS2015}                                &  \textbf{0.64} (0.16)  &  \textbf{0.66}     &  \textbf{0.64} (0.15)  &  \textbf{0.66}     &  \textbf{0.70} (0.13)  &  \textbf{0.73}   &  0.78 (0.07)  &  0.78 \\   
                Excitation Backprop \cite{Zhang:ECCV2016}               &  0.62 (0.17)  &  0.65     &  0.63 (0.16)  &  0.65     &  0.70 (0.14)  &  0.73    &  \textbf{0.78} (0.07)  &  \textbf{0.79} \\   
                IG \cite{Sundararajan:ICML2017}                         &  0.49 (0.19)  &  0.51     &  0.51 (0.17)  &  0.53     &  0.65 (0.14)  &  0.69    &  0.74 (0.06)  &  0.75 \\   
                Guided Backprop \cite{Spring:ICLR2015}                  &  0.50 (0.13)  &  0.52     &  0.50 (0.12)  &  0.51     &  0.51 (0.11)  &  0.52    &  0.54 (0.08)  &  0.55 \\   
                Guided Grad-CAM \cite{Selvaraju:ICCV2017}               &  0.50 (0.15)  &  0.51     &  0.49 (0.14)  &  0.51     &  0.51 (0.12)  &  0.53    &  0.55 (0.09)  &  0.56 \\   
                SmoothGrad \cite{Smilkov:ICML2017}                      &  0.33 (0.17)  &  0.32     &  0.34 (0.17)  &  0.33     &  0.35 (0.14)  &  0.35    &  0.35 (0.12)  &  0.35 \\   
                VarGrad \cite{Adebayo:ICLR2018}                         &  0.31 (0.20)  &  0.30     &  0.31 (0.20)  &  0.30     &  0.32 (0.17)  &  0.31    &  0.31 (0.15)  &  0.30 \\   
                Gradient \cite{Simonyan:ICLR2014}                       &  0.28 (0.14)  &  0.28     &  0.29 (0.14)  &  0.29     &  0.34 (0.12)  &  0.34    &  0.38 (0.09)  &  0.38 \\   
                Gradient$\times$Input \cite{Shrikumar:arxiv2016}        &  0.25 (0.14)  &  0.25     &  0.26 (0.13)  &  0.25     &  0.30 (0.11)  &  0.31    &  0.34 (0.09)  &  0.34 \\   
                Deconvnet \cite{Zeiler:ECCV2014}                        &  0.17 (0.12)  &  0.15     &  0.17 (0.11)  &  0.16     &  0.21 (0.10)  &  0.20    &  0.23 (0.10)  &  0.22 \\   
                Grad-CAM \cite{Selvaraju:ICCV2017}                      &  0.22 (0.22)  &  0.18     &  0.23 (0.22)  &  0.20     &  0.34 (0.20)  &  0.36    &  0.42 (0.16)  &  0.43 \\   
				\hline
			\end{tabular}
		}
	\end{center}
\end{table}

\FloatBarrier

\section{Discussion}\label{sec:discussion}

\paragraph{Comparison with previous works}

We could not confirm the findings from previous works, our results sometimes even contradict with the conclusions from previous comparative studies.

For example \cite{Adebayo:NIPS2018} perform randomization tests on the model parameters and training  data to evaluate and compare XAI methods. The authors of this work find that Guided Backprop and Guided GradCAM fail to pass their tests, and according to them this implies that these methods are incapable of supporting tasks that require explanations to be faithful to the model or data generating process. On the other hand they found  that Gradient and GradCAM successfully pass their tests.
In our controlled VQA framework we find exactly the opposite: GradCAM performs very poorly, gradient-based relevances also do not perform well, while Guided Backprop and Guided GradCAM are among the top performing XAI methods. 
We suspect the tested objectives from  \cite{Adebayo:NIPS2018} to be too far away from the actual prediction problem, and presumably prone to perturbation artefacts, to allow for a fine-grained and precise comparison of XAI methods. Indeed in their evaluation they track the change in explanation under perturbation of the model (either through parameter randomization or re-training), while we keep a clean unmodified configuration based on the same model and data as during model training.

Another work which is based on the Remove And Retrain paradigm \cite{Hooker:NIPS2019} concludes that Gradient, Integrated Gradients and Guided Backprop are no better or on par with a random estimate of feature importance. This again contradicts our results: in our study Integrated Gradients and Guided Backprop are among the best performing methods, and hence far from random feature importance assignments. We suspect again the evaluation protocol of this work to be too disconnected from the actual prediction task and the original model, to serve as an evaluation of the pertinence of XAI methods for a given prediction.

Lastly in  \cite{Sixt:ICML2020} the authors find that several modified backpropagation based methods, including LRP and Guided Backprop, which in our experiments are among the top performing methods, are independent of the parameters of higher layers in the model due to a convergence of the explanation to a rank-1 matrix, and hence that they can not explain a network's prediction faithfully. In our VQA setup we instead find empirically a high correlation between the network's prediction and the target objects of the question, which even increases with the model's confidence. 
Maybe the findings from \cite{Sixt:ICML2020} become predominant in the asymptotic case of a neural network with a very high number of layers, and may indicate directions for further XAI methods improvements. However, in our practical use case of a neural network with 12 layers, we could not confirm that this phenomenon alters the quality of modified backpropagation based explanations in comparison to gradient-based methods.

\paragraph{Limitations}

We are aware that XAI evaluations w.r.t. Ground Truths should always been carried out with care.
There is probably in many cases no perfect Ground Truth.
In our study for example, the ground truth masks do not take into the objects' shadows, while in principle the VQA model could use them for answering certain questions (such as queries about the objects' shapes).
However, for our analysis we preferred employing a dataset with realistic looking illumination conditions and shadows, rather than using a more artificial task (such as the sort-of-CLEVR toy task from \cite{Santoro:NIPS2017} for example) to keep the evaluation grounded in a practical and natural visual environment.
Further, we believe the specific dataset we considered, which is based on the CLEVR task \cite{Johnson:CLEVRDiagnosticDataset:2017}, is well-suited for vision XAI evaluation due to its complexity and variability: objects are often occluded and at different locations, this forces the model to focus on the objects themselves and not on the background to detect e.g. the objects' shapes (this would be less the case in a task with no overlapping object boundaries, where the model could develop a strategy that always includes a portion of background around objects for shape recognition).

Finally, although we acknowledge that our evaluation setup is not perfect, we firmly believe it is more reliable and trustworthy than previously conducted evaluation studies in XAI.

\paragraph{Outlook}

It remains an open question to whether the relevance accuracy upper bound of 1.0 can be reached by any XAI method (not tested in this study) on our CLEVR-XAI benchmark dataset, when using one of the selective Ground Truth masks we generated (on the generic \textit{GT All Objects} this upper bound is already nearly reached by the IG and LRP methods for the mass accuracy).
Indeed one possible reason for having an accuracy lower than 1.0 comes the classifier's uncertainty (we showed that the higher the model's confidence, the higher the relevance accuracy) on some more difficult to predict data points from our dataset. It could be that the VQA model is hesitating or confused for such data points, and consequently the relevance gets more diffused across multiple objects in the scene, rather than concentrated on the target objects of the question.
Nevertheless, we believe this does not invalidate our evaluation approach, and our benchmark dataset is still suited to compare the \textit{relative} performance of XAI methods.

Further, it would be interesting to explore whether the advantages of LRP and Guided Backprop could be combined to build an even stronger XAI method, or to analyze possible theoretical connections between those methods. Indeed both methods are efficient (they require a single backward pass), further they do not require any numerical hyperparameter to be chosen (like this is the case, e.g., in SmoothGrad), or to determine a baseline image (like this is required for IG).
IG also performs well, but is far more expensive to compute. Moreover, methods such as SmoothGrad and IG rely on perturbed input images for the averaging, resp. the integration of gradients. Such perturbed data points may lie outside the actual training data's manifold, bearing the risk of leading to artefacts and unreliable model behaviors.

Lastly, for future work our evaluation approach could be tested on trainable XAI methods (in this work we mainly tested deterministic XAI methods).
Also it would interesting to train a Deep Taylor Decomposition conform neural network model \cite{Montavon:PR2017} on the CLEVR task, i.e., a Relation Network model \cite{Santoro:NIPS2017} constrained to have negative biases and positive-valued predictions, to check if this increases the performance of the LRP explanations (in the present work we used a standard Relation Network model trained with no particular constraints).

\section{Conclusion}\label{sec:conclusion}

In the present work we have proposed a novel VQA-based evaluation paradigm for computer vision XAI. 
Our benchmark dataset (CLEVR-XAI) with corresponding Ground Truth annotations and metrics can be used to analyze and improve current explanation methods for neural networks.

Compared to previous evaluation approaches, which were mainly based on CNN object classifiers, our framework has the advantage of being altogether realistic, selective and controlled,
and to further rely on the same model and data as during model training.
Hereby we hope to pave the way for a more reliable and trustworthy assessment of XAI methods in computer vision.

Among the methods we tested in our comparative study, we found Layer-wise Relevance Propagation, Integrated Gradients and Guided Backpropagation to be the most accurate XAI methods, while Deconvnet and Grad-CAM were the least accurate ones to back up the model's predictions.


\bibliography{bibliography}

\begin{thebibliography}{10}
\expandafter\ifx\csname url\endcsname\relax
  \def\url#1{\texttt{#1}}\fi
\expandafter\ifx\csname urlprefix\endcsname\relax\def\urlprefix{URL }\fi
\expandafter\ifx\csname href\endcsname\relax
  \def\href#1#2{#2} \def\path#1{#1}\fi

\bibitem{lu2014surpassing}
C.~Lu, X.~Tang, Surpassing human-level face verification performance on {LFW}
  with {G}aussian{F}ace, in: AAAI Conference on Artificial Intelligence, 2015,
  pp. 3811--3819.

\bibitem{devlin2019bert}
J.~Devlin, M.-W. Chang, K.~Lee, K.~Toutanova, {BERT}: Pre-training of deep
  bidirectional transformers for language understanding, in: Proceedings of the
  North {A}merican Chapter of the Association for Computational Linguistics:
  Human Language Technologies (NAACL), 2019, pp. 4171--4186.

\bibitem{esteva2017dermatologist}
A.~Esteva, B.~Kuprel, R.~A. Novoa, J.~Ko, S.~M. Swetter, H.~M. Blau, S.~Thrun,
  Dermatologist-level classification of skin cancer with deep neural networks,
  Nature 542 (2017) 115--118.

\bibitem{Geirhos:Nature2020}
R.~Geirhos, J.~Jacobsen, C.~Michaelis, R.~S. Zemel, W.~Brendel, M.~Bethge,
  F.~A. Wichmann, {Shortcut Learning in Deep Neural Networks}, Nature Machine
  Intelligence 2 (2020) 665--673.

\bibitem{Hendricks:ECCV2018}
L.~A. Hendricks, K.~Burns, K.~Saenko, T.~Darrell, A.~Rohrbach, Women also
  snowboard: Overcoming bias in captioning models, in: Proceedings of the
  European Conference on Computer Vision (ECCV), 2018, pp. 793--811.

\bibitem{Manjunatha:CVPR2019}
V.~{Manjunatha}, N.~{Saini}, L.~S. {Davis}, Explicit bias discovery in visual
  question answering models, in: Proceedings of the Conference on Computer
  Vision and Pattern Recognition (CVPR), 2019, pp. 9554--9563.

\bibitem{Lap:Nature19}
S.~Lapuschkin, S.~W{\"a}ldchen, A.~Binder, G.~Montavon, W.~Samek, K.-R.
  M{\"u}ller, Unmasking clever hans predictors and assessing what machines
  really learn, Nature Communications 10 (2019) 1096.

\bibitem{SamXAI19}
W.~Samek, G.~Montavon, A.~Vedaldi, L.~K. Hansen, K.-R. M{\"u}ller, Explainable
  AI: Interpreting, Explaining and Visualizing Deep Learning, Vol. 11700 of
  Lecture Notes in Computer Science, Springer, 2019.

\bibitem{SamPIEEE21}
W.~Samek, G.~Montavon, S.~Lapuschkin, C.~J. Anders, K.-R. M{\"u}ller,
  Explaining deep neural networks and beyond: A review of methods and
  applications, Proceedings of the IEEE (2021).

\bibitem{Arrieta:InfoFusion2020}
A.~{Barredo Arrieta}, N.~Díaz-Rodríguez, J.~{Del Ser}, A.~Bennetot, S.~Tabik,
  A.~Barbado, S.~Garcia, S.~Gil-Lopez, D.~Molina, R.~Benjamins, R.~Chatila,
  F.~Herrera, {Explainable Artificial Intelligence (XAI): Concepts, taxonomies,
  opportunities and challenges toward responsible AI}, Information Fusion 58
  (2020) 82--115.

\bibitem{EU-GDPR}
EU-GDPR, {Regulation (EU) 2016/679 of the European Parliament and of the
  Council of 27 April 2016 on the protection of natural persons with regard to
  the processing of personal data and on the free movement of such data, and
  repealing Directive 95/46/EC (General Data Protection Regulation)}, Official
  Journal of the European Union L 119 59 (2016) 1--88.

\bibitem{Yeom:Arxiv2019}
S.-K. Yeom, P.~Seegerer, S.~Lapuschkin, A.~Binder, S.~Wiedemann, K.-R.
  M{\"u}ller, W.~Samek, {Pruning by Explaining: A Novel Criterion for Deep
  Neural Network Pruning}, Pattern Recognition (2021).

\bibitem{schutt2017quantum}
K.~T. Sch{\"u}tt, F.~Arbabzadah, S.~Chmiela, K.~R. M{\"u}ller, A.~Tkatchenko,
  Quantum-chemical insights from deep tensor neural networks, Nature
  Communications 8 (2017) 13890.

\bibitem{HorSREP19}
F.~Horst, S.~Lapuschkin, W.~Samek, K.-R. M{\"u}ller, W.~I. Sch{\"o}llhorn,
  Explaining the unique nature of individual gait patterns with deep learning,
  Scientific Reports 9 (2019) 2391.

\bibitem{Widrich:2020}
M.~Widrich, B.~Sch{\"a}fl, M.~Pavlovi{\'c}, H.~Ramsauer, L.~Gruber,
  M.~Holzleitner, J.~Brandstetter, G.~K. Sandve, V.~Greiff, S.~Hochreiter,
  G.~Klambauer, {Modern Hopfield Networks and Attention for Immune Repertoire
  Classification}, in: Advances in Neural Information Processing Systems
  (NeurIPS), 2020.

\bibitem{Simonyan:ICLR2014}
K.~Simonyan, A.~Vedaldi, A.~Zisserman, Deep {I}nside {C}onvolutional
  {N}etworks: {V}isualising {I}mage {C}lassification {M}odels and {S}aliency
  {M}aps, in: International Conference on Learning Representations (ICLR),
  2014.

\bibitem{Selvaraju:IJCV2020}
R.~R. Selvaraju, M.~Cogswell, A.~Das, R.~Vedantam, D.~Parikh, D.~Batra,
  Grad-cam: Visual explanations from deep networks via gradient-based
  localization, International Journal of Computer Vision 128~(2) (2020)
  336--359.

\bibitem{Shrikumar:arxiv2016}
A.~Shrikumar, P.~Greenside, A.~Shcherbina, A.~Kundaje, Not just a black box:
  Interpretable deep learning by propagating activation differences, arXiv
  preprint arXiv:1605.01713 (2016).

\bibitem{Sundararajan:ICML2017}
M.~Sundararajan, A.~Taly, Q.~Yan, Axiomatic attribution for deep networks, in:
  Proceedings of the International Conference on Machine Learning {(ICML)},
  2017, pp. 3319--3328.

\bibitem{Bach:PLOS2015}
S.~Bach, A.~Binder, G.~Montavon, F.~Klauschen, K.-R. M{\"u}ller, W.~Samek, On
  {P}ixel-{W}ise {E}xplanations for {N}on-{L}inear {C}lassifier {D}ecisions by
  {L}ayer-{W}ise {R}elevance {P}ropagation, PLoS ONE 10~(7) (2015) e0130140.

\bibitem{Zhang:IJCV18}
J.~Zhang, S.~A. Bargal, Z.~Lin, J.~Brandt, X.~Shen, S.~Sclaroff, Top-down
  neural attention by excitation backprop, International Journal of Computer
  Vision 126~(10) (2018) 1084--1102.

\bibitem{Spring:ICLR2015}
J.~T. Springenberg, A.~Dosovitskiy, T.~Brox, M.~Riedmiller, Striving for
  simplicity: The all convolutional net, in: International Conference on
  Learning Representations (ICLR), 2015.

\bibitem{Ribeiro:KDD2016}
M.~T. Ribeiro, S.~Singh, C.~Guestrin, ``{W}hy {S}hould {I} {T}rust {Y}ou?":
  {E}xplaining the {P}redictions of {A}ny {C}lassifier, in: Proceedings of the
  International Conference on Knowledge Discovery and Data Mining (KDD), 2016,
  pp. 1135--1144.

\bibitem{Lundberg:NIPS2017}
S.~M. Lundberg, S.-I. Lee, A unified approach to interpreting model
  predictions, in: Advances in Neural Information Processing Systems (NIPS),
  2017, pp. 4765--4774.

\bibitem{Fong:ICCV2017}
R.~C. {Fong}, A.~{Vedaldi}, Interpretable explanations of black boxes by
  meaningful perturbation, in: Proceedings of the International Conference on
  Computer Vision (ICCV), 2017.

\bibitem{Zintgraf:ICLR2017}
L.~M. Zintgraf, T.~S. Cohen, T.~Adel, M.~Welling, Visualizing deep neural
  network decisions: Prediction difference analysis, in: Proceedings of the
  International Conference on Learning Representations (ICLR), 2017.

\bibitem{Chen:ICML2018}
J.~Chen, L.~Song, M.~Wainwright, M.~Jordan, {Learning to Explain: An
  Information-Theoretic Perspective on Model Interpretation}, in: Proceedings
  of the International Conference on Machine Learning (ICML), 2018, pp.
  883--892.

\bibitem{Samek:TNNLS2017}
W.~Samek, A.~Binder, G.~Montavon, S.~Lapuschkin, K.-R. M{\"u}ller, Evaluating
  the visualization of what a {D}eep {N}eural {N}etwork has learned, IEEE
  Transactions on Neural Networks and Learning Systems 28~(11) (2017)
  2660--2673.

\bibitem{Johnson:CLEVRDiagnosticDataset:2017}
J.~{Johnson}, B.~{Hariharan}, L.~v.~d. {Maaten}, L.~{Fei-Fei}, C.~L. {Zitnick},
  R.~{Girshick}, {CLEVR}: A diagnostic dataset for compositional language and
  elementary visual reasoning, in: Proceedings of the Conference on Computer
  Vision and Pattern Recognition (CVPR), 2017, pp. 1988--1997.

\bibitem{Santoro:NIPS2017}
A.~Santoro, D.~Raposo, D.~G. Barrett, M.~Malinowski, R.~Pascanu, P.~Battaglia,
  T.~Lillicrap, A simple neural network module for relational reasoning, in:
  Advances in Neural Information Processing Systems (NIPS), 2017, pp.
  4967--4976.

\bibitem{Ancona:ICLR2018}
M.~Ancona, E.~Ceolini, C.~{\"{O}}ztireli, M.~Gross, {Towards better
  understanding of gradient-based attribution methods for Deep Neural
  Networks}, in: International Conference on Learning Representations (ICLR),
  2018.

\bibitem{Morcos:ICLR2018}
A.~S. Morcos, D.~G. Barrett, N.~C. Rabinowitz, M.~Botvinick, {On the importance
  of single directions for generalization}, in: International Conference on
  Learning Representations (ICLR), 2018.

\bibitem{Zeiler:ECCV2014}
M.~D. Zeiler, R.~Fergus, Visualizing and understanding convolutional networks,
  in: Proceedings of the European Conference on Computer Vision (ECCV), 2014,
  pp. 818--833.

\bibitem{Adebayo:NIPS2018}
J.~Adebayo, J.~Gilmer, M.~Muelly, I.~Goodfellow, M.~Hardt, B.~Kim, {Sanity
  Checks for Saliency Maps}, in: Advances in Neural Information Processing
  Systems (NeurIPS), 2018, pp. 9505--9515.

\bibitem{Sixt:ICML2020}
L.~Sixt, M.~Granz, T.~Landgraf, {When Explanations Lie: Why Many Modified {BP}
  Attributions Fail}, in: Proceedings of the International Conference on
  Machine Learning (ICML), 2020, pp. 9046--9057.

\bibitem{Adebayo:NIPS2020}
J.~Adebayo, M.~Muelly, I.~Liccardi, B.~Kim, Debugging tests for model
  explanations, in: Advances in Neural Information Processing Systems
  (NeurIPS), 2020.

\bibitem{Hooker:NIPS2019}
S.~Hooker, D.~Erhan, P.~Kindermans, B.~Kim, A benchmark for interpretability
  methods in deep neural networks, in: Advances in Neural Information
  Processing Systems (NeurIPS), 2019, pp. 9737--9748.

\bibitem{Zhang:ECCV2016}
J.~Zhang, Z.~Lin, J.~Brandt, X.~Shen, S.~Sclaroff, Top-down neural attention by
  excitation backprop, in: Proceedings of the European Conference on Computer
  Vision (ECCV), 2016, pp. 543--559.

\bibitem{Gu:ACCV2019}
J.~Gu, Y.~Yang, V.~Tresp, Understanding individual decisions of cnns via
  contrastive backpropagation, in: Proceedings of the Asian Conference on
  Computer Vision (ACCV), 2019, pp. 119--134.

\bibitem{Beery:ECCV2018}
S.~Beery, G.~Van~Horn, P.~Perona, {Recognition in Terra Incognita}, in:
  Proceedings of the European Conference on Computer Vision (ECCV), 2018.

\bibitem{Russakovsky:ILSVRC15}
O.~Russakovsky, J.~Deng, H.~Su, J.~Krause, S.~Satheesh, S.~Ma, Z.~Huang,
  A.~Karpathy, A.~Khosla, M.~Bernstein, A.~C. Berg, L.~Fei-Fei, {ImageNet Large
  Scale Visual Recognition Challenge}, International Journal of Computer Vision
  115~(3) (2015) 211--252.

\bibitem{Selvaraju:ICCV2017}
R.~R. Selvaraju, M.~Cogswell, A.~Das, R.~Vedantam, D.~Parikh, D.~Batra,
  Grad-cam: Visual explanations from deep networks via gradient-based
  localization, in: Proceedings of the International Conference on Computer
  Vision (ICCV), 2017.

\bibitem{Smilkov:ICML2017}
D.~Smilkov, N.~Thorat, B.~Kim, F.~Vi{\'{e}}gas, M.~Wattenberg, Smoothgrad:
  removing noise by adding noise, in: Proceedings of the International
  Conference on Machine Learning Workshop on Visualization for Deep Learning,
  2017.

\bibitem{Oramas:ICLR2019}
J.~Oramas, K.~Wang, T.~Tuytelaars, {Visual Explanation by Interpretation:
  Improving Visual Feedback Capabilities of Deep Neural Networks}, in:
  International Conference on Learning Representations (ICLR), 2019.

\bibitem{Shrikumar:ICML2017}
A.~Shrikumar, P.~Greenside, A.~Kundaje, Learning important features through
  propagating activation differences, in: Proceedings of the International
  Conference on Machine Learning (ICML), 2017, pp. 3145--3153.

\bibitem{Adebayo:ICLR2018}
J.~Adebayo, J.~Gilmer, I.~J. Goodfellow, B.~Kim, {Local Explanation Methods for
  Deep Neural Networks Lack Sensitivity to Parameter Values}, in: International
  Conference on Learning Representations (ICLR), 2018.

\bibitem{Mahendran:ECCV2016}
A.~Mahendran, A.~Vedaldi, Salient deconvolutional networks, in: Proceedings of
  the European Conference on Computer Vision (ECCV), 2016, pp. 120--135.

\bibitem{Antol:ICCV2015}
S.~{Antol}, A.~{Agrawal}, J.~{Lu}, M.~{Mitchell}, D.~{Batra}, C.~L. {Zitnick},
  D.~{Parikh}, {VQA: Visual Question Answering}, in: International Conference
  on Computer Vision (ICCV), 2015, pp. 2425--2433.

\bibitem{Goyal:CVPR2017}
Y.~{Goyal}, T.~{Khot}, D.~{Summers-Stay}, D.~{Batra}, D.~{Parikh}, Making the v
  in vqa matter: Elevating the role of image understanding in visual question
  answering, in: Proceedings of the Conference on Computer Vision and Pattern
  Recognition (CVPR), 2017, pp. 6325--6334.

\bibitem{Agrawal:CVPR2018}
A.~Agrawal, D.~Batra, D.~Parikh, A.~Kembhavi, Don't just assume; look and
  answer: Overcoming priors for visual question answering, in: Proceedings of
  the Conference on Computer Vision and Pattern Recognition (CVPR), 2018, pp.
  4971--4980.

\bibitem{Dimopoulos:1995}
Y.~Dimopoulos, P.~Bourret, S.~Lek, Use of some sensitivity criteria for
  choosing networks with good generalization ability, Neural Processing Letters
  2~(6) (1995) 1--4.

\bibitem{Gevrey:2003}
M.~Gevrey, I.~Dimopoulos, S.~Lek, Review and comparison of methods to study the
  contribution of variables in artificial neural network models, Ecological
  Modelling 160~(3) (2003) 249--264.

\bibitem{Seo:ICML2018}
J.~Seo, J.~Choe, J.~Koo, S.~Jeon, B.~Kim, T.~Jeon, Noise-adding methods of
  saliency map as series of higher order partial derivative, in: Proceedings of
  the ICML Workshop on Human Interpretability in Machine Learning, 2018.

\bibitem{Montavon:PR2017}
G.~Montavon, S.~Lapuschkin, A.~Binder, W.~Samek, K.-R. M{\"u}ller, Explaining
  nonlinear classification decisions with deep {T}aylor decomposition, Pattern
  Recognition 65 (2017) 211--222.

\bibitem{Montavon:DSP2018}
G.~Montavon, W.~Samek, K.-R. M{\"u}ller, {Methods for Interpreting and
  Understanding Deep Neural Networks}, Digital Signal Processing 73 (2018)
  1--15.

\bibitem{Montavon:ExplAIBook2020}
G.~Montavon, A.~Binder, S.~Lapuschkin, W.~Samek, K.-R. M{\"u}ller, Layer-wise
  relevance propagation: An overview, in: Explainable AI: Interpreting,
  Explaining and Visualizing Deep Learning, Vol. 11700 of Lecture Notes in
  Computer Science, Springer, 2019, pp. 193--209.

\bibitem{KohIJCNN20}
M.~Kohlbrenner, A.~Bauer, S.~Nakajima, A.~Binder, W.~Samek, S.~Lapuschkin,
  {Towards best practice in explaining neural network decisions with LRP}, in:
  IEEE International Joint Conference on Neural Networks (IJCNN), 2020, pp.
  1--7.

\bibitem{Arras:PLOSONE2017}
L.~Arras, F.~Horn, G.~Montavon, K.-R. M\"{u}ller, W.~Samek, {"What is relevant
  in a text document?": An interpretable machine learning approach}, PLoS ONE
  12~(7) (2017) e0181142.

\bibitem{Arras:17}
L.~Arras, G.~Montavon, K.-R. M{\"u}ller, W.~Samek, Explaining recurrent neural
  network predictions in sentiment analysis, in: Proceedings of the EMNLP'17
  Workshop on Computational Approaches to Subjectivity, Sentiment \& Social
  Media Analysis (WASSA), 2017, pp. 159--168.

\bibitem{Poerner:ACL2018}
N.~Poerner, B.~Roth, H.~Sch{\"u}tze, {Evaluating neural network explanation
  methods using hybrid documents and morphosyntactic agreement}, in:
  {Proceedings of the 56th Annual Meeting of the Association for Computational
  Linguistics (ACL)}, 2018, pp. 340--350.

\bibitem{ArrXAI19}
L.~Arras, J.~Arjona-Medina, M.~Widrich, G.~Montavon, M.~Gillhofer, K.-R.
  M{\"u}ller, S.~Hochreiter, W.~Samek, {Explaining and Interpreting LSTMs}, in:
  Explainable AI: Interpreting, Explaining and Visualizing Deep Learning, Vol.
  11700 of Lecture Notes in Computer Science, 2019, pp. 211--238.

\bibitem{sturmfels2020visualizing}
P.~Sturmfels, S.~Lundberg, S.-I. Lee, Visualizing the impact of feature
  attribution baselines, Distill (2020).
\newblock \href {https://doi.org/10.23915/distill.00022}
  {\path{doi:10.23915/distill.00022}}.

\end{thebibliography}

\appendix

\section{Additional CLEVR-XAI-complex example}\label{appendix:CLEVR-XAI_examples}

\begin{figure*}[!t]
	\small
	\centering
	\begin{tabular}{p{0.27\linewidth}p{0.27\linewidth}p{0.35\linewidth}}
		\toprule
		\hfil Image                                                                                       &           \hfil Question/Answer       &               \hfil  Program                                  \\
		\hfil \raisebox{-\height}{\includegraphics[width=2.9cm,height=2.9cm]{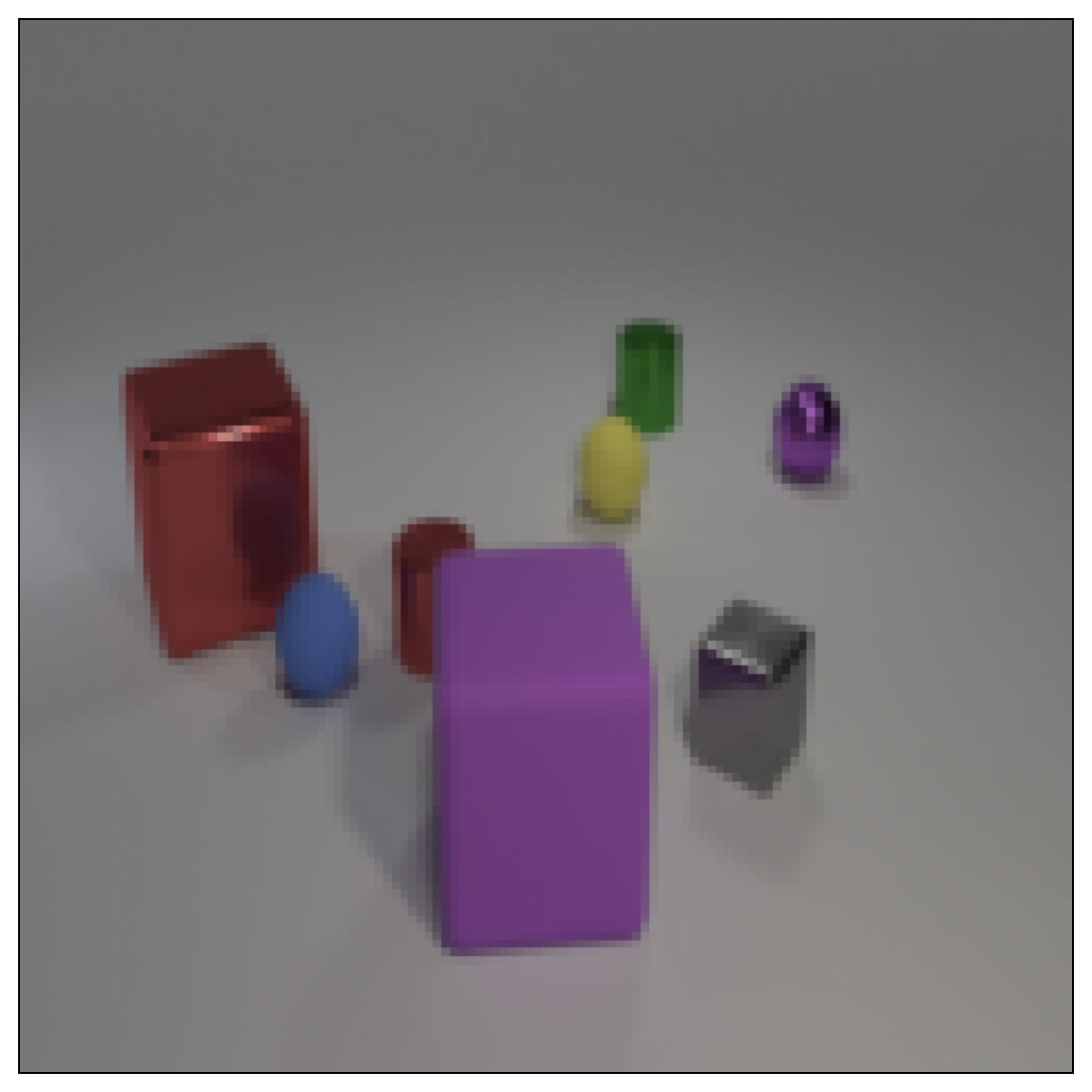}}                                                                              &
		\vspace*{\fill} \noindent{\raggedright  \textbf{\scriptsize What number of objects are large purple blocks or green metallic cubes?}\vspace{0.2cm} \\ \textit{one}}  \vfill                  &
        \vspace*{\fill} \noindent{\raggedright  {\texttt {\scriptsize branch1 = [scene, filter\_size, filter\_color, filter\_shape] \\ branch2 = [scene, filter\_color, filter\_material, filter\_shape] \\ program = [(branch1,branch2), union, count]}}} \vfill                                                               \\
        \midrule
        \hfil Ground Truth                                                                                          &     \hfil \scriptsize GT Unique          &      \hfil \scriptsize GT Unique First-non-empty                  \\
		{} & \hfil {\scriptsize       undefined}                                                                             &
		\hfil \raisebox{-.5\height}{\includegraphics[width=2.9cm]{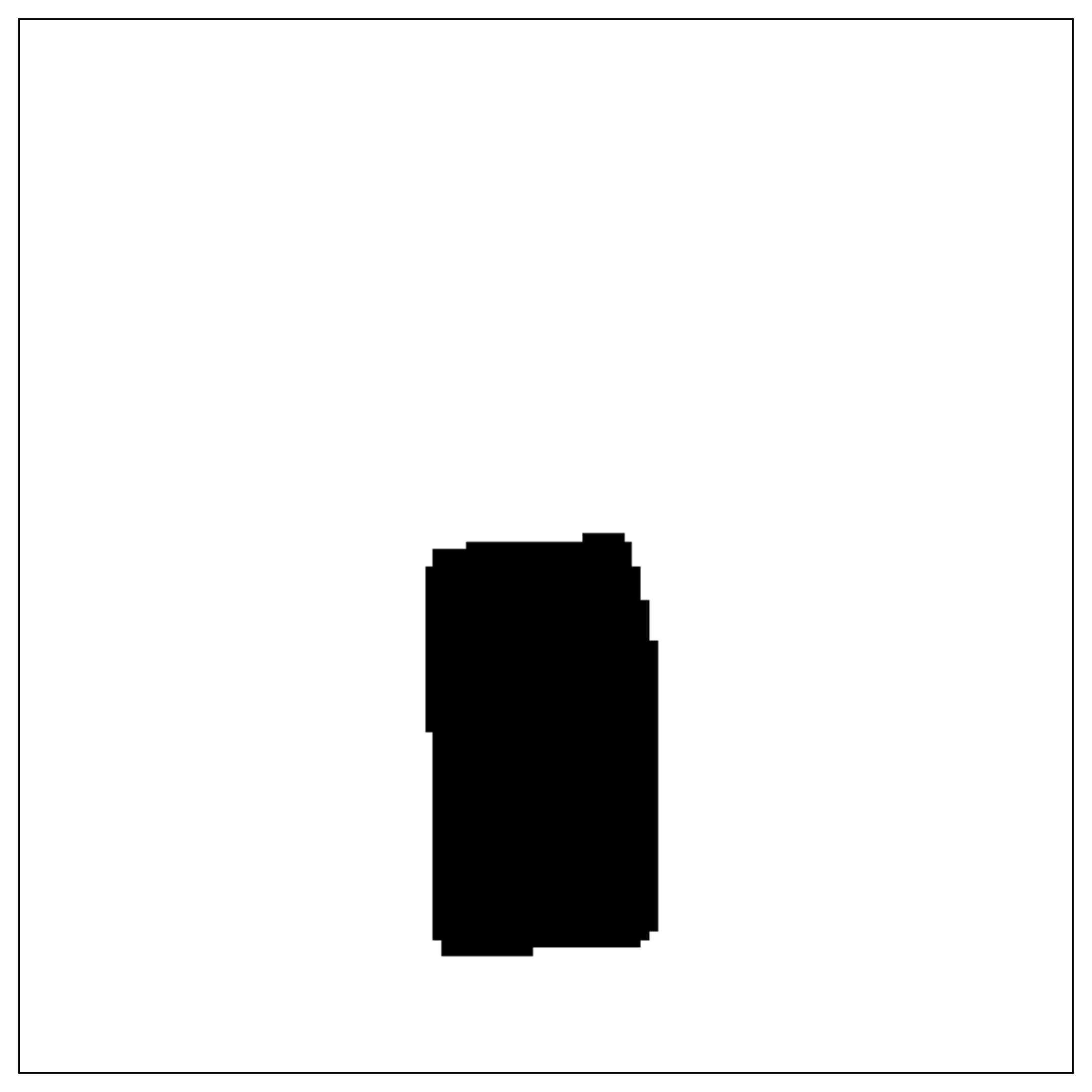}}                                                                                                \\
		{}                                                                                                &     \hfil \scriptsize GT Union & \hfil \scriptsize GT All Objects                                             \\
		{} & \hfil \raisebox{-.5\height}{\includegraphics[width=2.9cm]{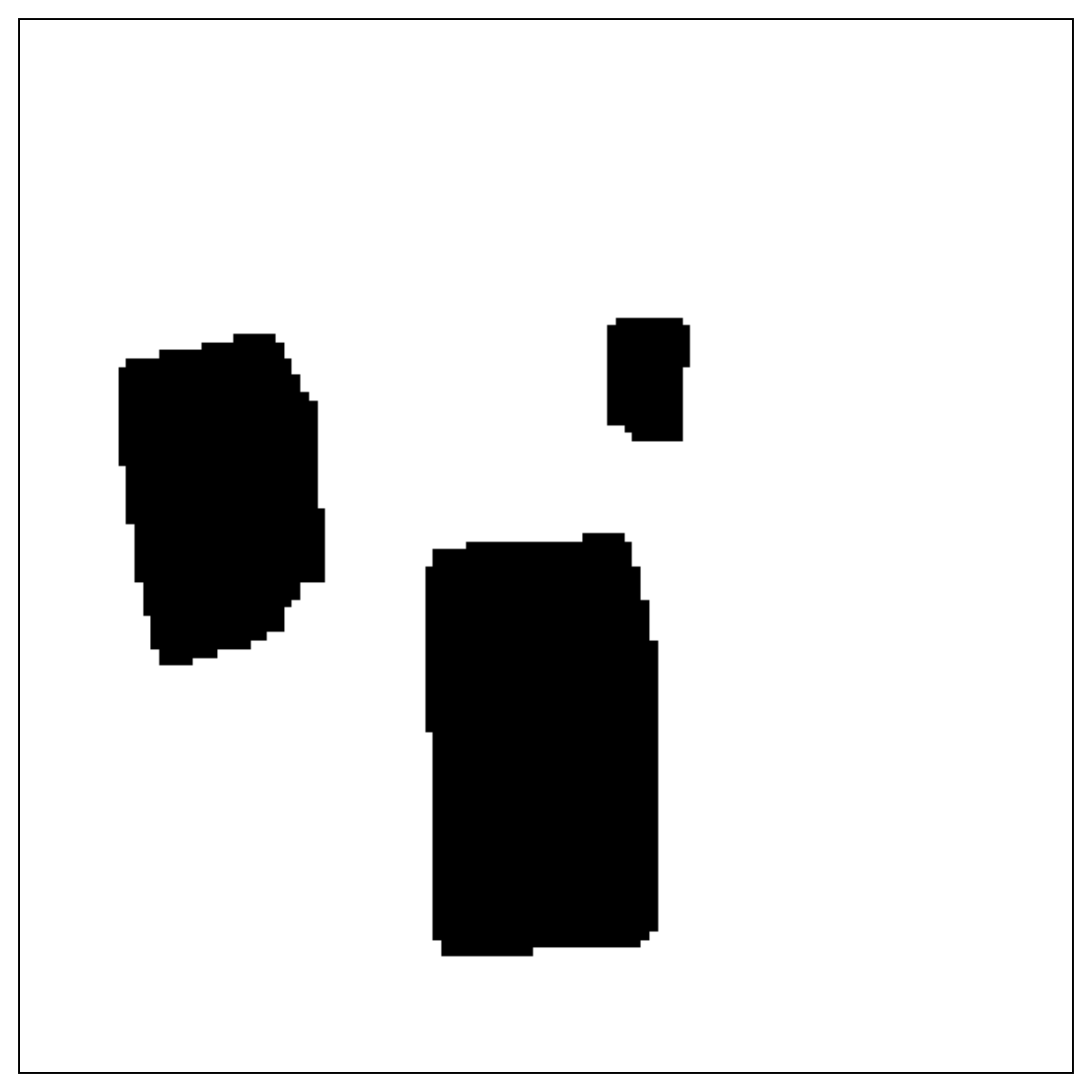}}                                                                                       &
		\hfil \raisebox{-.5\height}{\includegraphics[width=2.9cm]{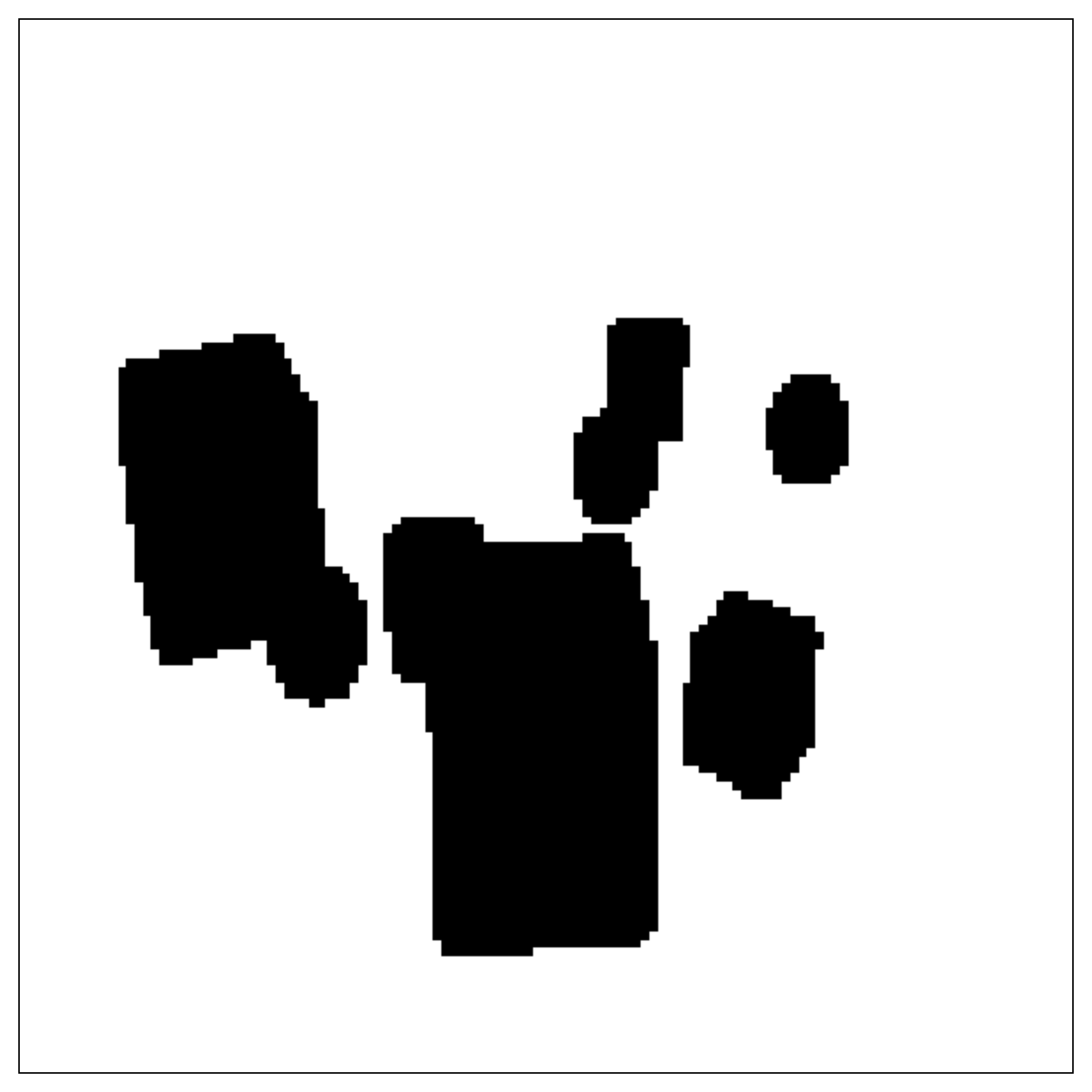}}                                                                                                          \\
		\bottomrule
	\end{tabular}
	\caption{Example data point from CLEVR-XAI-complex. The functional program is used to determine which objects in the scene are considered as ground truths.}
	\label{fig:clevr-xai-complex_example_tree-structured}
\end{figure*}

Figure~\ref{fig:clevr-xai-complex_example_tree-structured} provides a CLEVR-XAI-complex example data point for a question with a tree-structured functional program. Note that the program does not contain any \texttt{unique} function, therefore the \textit{GT Unique} mask is undefined for this question. The \textit{GT Unique First-non-empty} mask includes the first non-empty set of objects returned by the functions in the program processed in reversed order, i.e., the output of the \texttt{union} function, which corresponds to the big purple cube. The \textit{GT Union} further contains the output of all intermediate functions in the program, in particular this corresponds to the output of the first filter function in each program branch (\texttt{filter\_size} in  \texttt{branch1} and \texttt{filter\_color} in  \texttt{branch2}), i.e., all large objects and all green objects (thus the two large cubes and the small green cylinder). The \textit{GT All Objects} simply contains all the objects in the scene.

Note that in general, as illustrated with the previous example, for CLEVR-XAI-complex questions some ground truth masks can be undefined or empty. This is why in Table~\ref{table:clevr-xai-statistics} for three ground truths there are less than 100,000 data points.

\FloatBarrier

\section{CLEVR-XAI dataset statistics}\label{appendix:CLEVR-XAI_statistics}

In Table~\ref{table:clevr-xai-statistics} we provide statistics for the CLEVR-XAI benchmark dataset.
The program length is the number of basic functions present in the question's functional program and is an indicator of the question's complexity.
For the CLEVR-XAI-simple's \textit{GT Single Object} we additionally distinguish between two cases depending on the size attribute of the target object of the question: note the high variability in the number of pixels within each size category, indicating that objects can be partially occluded and located at different places (foreground or background of the scene).

Lastly the number of pixels per ground truth we report were calculated on images of size 128$\times$128 (since the neural network we use in our experiments takes as input images of size 128$\times$128, for more details on the ground truth mask resizing see \ref{appendix:GT_resizing}). In our dataset release we also provide the possibility to use ground truths of size 320$\times$480 (the original size of the CLEVR dataset images \cite{Johnson:CLEVRDiagnosticDataset:2017}) or any other chosen image size.

\setlength{\tabcolsep}{4pt}
\begin{table}
	\begin{center}
		\caption{CLEVR-XAI dataset statistics}
		\label{table:clevr-xai-statistics}

		\resizebox{1.0\columnwidth}{!}{
			\begin{tabular}{l|c|cccc|cccc}
				\toprule
				\noalign{\smallskip}
				\multicolumn{10}{l}{\textbf{CLEVR-XAI-simple (39,761 data points)}}  \\
				\noalign{\smallskip}
				\multicolumn{10}{l}{program length: 4 min, 6 max, 5.1 mean, 0.7 std} \\
                \multicolumn{2}{c}{}  & \multicolumn{4}{c}{number of pixels}  & \multicolumn{4}{c}{number of objects}  \\
                Ground Truth               &  number of questions                 & min & max & mean & std                & min & max & mean & std     \\
                \midrule
                \textit{GT Single Object} & 39,761  & 40 & 2040 & 428 & 321  & 1 & 1 & 1 & 0 \\
                {  - small target object} & 20,393  & 40 & 548 & 189 & 75  & 1 & 1 & 1 & 0 \\
                {  - large target object} & 19,368  & 125 & 2040 & 679 & 286  & 1 & 1 & 1 & 0 \\
				\hline
				\textit{GT All Objects} & 39,761  & 347 & 6789 & 2654 & 1071  & 3 & 10 & 6.5 & 2.3 \\
				\bottomrule
				\noalign{\smallskip}
				\noalign{\smallskip}
				\multicolumn{10}{l}{\textbf{CLEVR-XAI-complex (100,000 data points)}}  \\
				\noalign{\smallskip}
				\multicolumn{10}{l}{program length: 2 min, 24 max, 11.1 mean, 4.3 std} \\
                \multicolumn{2}{c}{}  & \multicolumn{4}{c}{number of pixels}  & \multicolumn{4}{c}{number of objects}  \\
                Ground Truth                &  number of questions                 & min & max & mean & std                & min & max & mean & std     \\
                \midrule
                \textit{GT Unique} & 89,873  & 43 & 4320 & 837 & 572 & 1 & 4 & 2.0 & 0.9  \\
                \hline
                \textit{GT Unique First-non-empty} & 99,786  & 43 & 5908 & 1042 & 676  & 1 & 10 & 2.5 & 1.1 \\
                \hline
                \textit{GT Union} & 99,786  & 45 & 6789 & 1958 & 1116  & 1 & 10 & 4.8 & 2.3 \\
                \hline
                \textit{GT All Objects} & 100,000  & 347 & 6789 & 2650 & 1072  & 3 & 10 & 6.5 & 2.3  \\
                \bottomrule
			\end{tabular}
		}

	\end{center}
\end{table}
\setlength{\tabcolsep}{1.4pt}

\FloatBarrier

\section{Neural Network used for the XAI Evaluation}\label{appendix:model}

Here we provide more details on the Relation Network (RN) based model architecture and training \cite{Santoro:NIPS2017}, which we subsequently used for the empirical evaluation of XAI methods.

The CNN part of the network is made of 4 layers, each with the following structure: \texttt{conv} $\rightarrow$ \texttt{relu} $\rightarrow$ \texttt{batchnorm}.
Each convolutional layer has 24 kernels of size 3$\times$3, stride 2, and no padding.

The LSTM part of the network is a unidirectional LSTM with word embeddings of size 32, and a hidden layer of size 128.

The Relation Network part of the model is made of 4 fully-connected layers of size 256, each followed with ReLU activation, and a final element-wise summation layer.

The classifier part of the network contains 3 fully-connected layers, where the first two layers have size 256 and are followed each by ReLU activation. Additionally, the second layer uses dropout ($p = 0.5$).
The output layer has size 28 (the number of possible answers for the CLEVR VQA task \cite{Johnson:CLEVRDiagnosticDataset:2017}).

For preprocessing the questions, we removed punctuation and applied lowercasing, this leaves us with a vocabulary of size 80.

For preprocessing the images, we rescaled the pixel values to the range  [0, 1], and resized the images to the size 128$\times$128 (the original CLEVR images have size 320$\times$480).

Training was done with the Adam optimizer, using a batch size of 64, an initial learning rate of 2.5e-4, clipping the gradient norm to 5.0, l2-norm regularization of 4e-5, and decreasing the learning rate by a factor of 0.95 once the validation accuracy does not improve within 10 epochs. Training was done for a maximum of 1200 epochs.

During training, we also applied data augmentation (random cropping and random rotation of the images), as described in the original publication \cite{Santoro:NIPS2017}.

\section{Resizing of Ground Truth Masks}\label{appendix:GT_resizing}

Note that since the RN model we use in our experiments takes as input images of size 128$\times$128, the ground truth masks used for the XAI evaluation also have to be resized to this size (initially these masks have the same size as the original CLEVR dataset's images \cite{Johnson:CLEVRDiagnosticDataset:2017}, i.e. 320$\times$480).
To this end we proceeded in the following way: we resized the masks using the same operation as was used for resizing the standard input images, starting with masks having a value of 1.0 on the ground truth pixels, and 0.0 elsewhere.
Then, after resizing the masks, we set all pixels having a non-zero value to True, and the remaining pixels to False. This way we ensure that the resulting masks also include the objects' borders (which are slightly blurred and dilated due to the resizing operation).

\section{XAI methods hyperparameters}\label{appendix:XAI_methods_hyperparameters}

In table~\ref{table:xai-methods-hyperparameters} we recapitulate the hyperparameters and pooling techniques we used for each XAI method, after tuning them for each relevance accuracy metric, as was described in Section~\ref{sec:experiments-simple-questions}.

\setlength{\tabcolsep}{4pt}
\begin{table}
	\begin{center}
		\caption{XAI methods hyperparameters}
		\label{table:xai-methods-hyperparameters}

		\resizebox{1.0\columnwidth}{!}{
			\begin{tabular}{p{0.35\textwidth}p{0.85\textwidth}p{0.2\textwidth}}
				\toprule
				\noalign{\smallskip}
				
				\multicolumn{3}{l}{\textbf{Relevance Mass Accuracy}}  \\
                Method                                              &  Hyperparameters                                                                                                  & Pooling    \\
                \midrule
                LRP \cite{Bach:PLOS2015}                            & $\alpha_{1},\beta_{0}$-rule in the hidden layers, box-rule in the input layer                                     & l2-norm-sq \\
                Excitation Backprop \cite{Zhang:ECCV2016}           & $\alpha_{1},\beta_{0}$-rule in all layers                                                                         & l2-norm-sq \\
                IG \cite{Sundararajan:ICML2017}                     & mean channel values as baseline image, nb of integration steps $\in$ [300, 1000, 3000, 10000, 30000] such that relative error $<$ 0.01 & pos,l2-norm-sq \\
                Guided Backprop \cite{Spring:ICLR2015}              & modified gradient backward pass for all ReLU layers in the network                                               & l2-norm-sq \\
                Guided Grad-CAM \cite{Selvaraju:ICCV2017}           & element-wise multiplication of Guided Backprop with modified gradient backward pass for all ReLU layers in the network \& Grad-CAM applied to the ReLU layer output in the last convolutional layer of the network & l2-norm-sq \\
                SmoothGrad \cite{Smilkov:ICML2017}                  & nb of samples 300, noise level 0.05, gradient squared                                                              & l2-norm-sq \\
                VarGrad \cite{Adebayo:ICLR2018}                     & nb of samples 300, noise level 0.05                                                                                & l2-norm-sq \\
                Gradient \cite{Simonyan:ICLR2014}                   & gradient squared                                                                                                   & l2-norm-sq \\
                Gradient$\times$Input \cite{Shrikumar:arxiv2016}    & gradient squared                                                                                                   & l2-norm-sq \\
                Deconvnet \cite{Zeiler:ECCV2014}                    & modified gradient backward pass only for ReLU layers within the convolutional neural network part of the network   & pos,l2-norm-sq \\
                 Grad-CAM \cite{Selvaraju:ICCV2017}                 & Grad-CAM applied to the batchnorm layer output in the last convolutional layer of the network                      & none\\
				\bottomrule
				\noalign{\smallskip}
				\noalign{\smallskip}
				
				\multicolumn{3}{l}{\textbf{Relevance Rank Accuracy}}  \\
                Method                                              &  Hyperparameters                                                                                                  & Pooling    \\
                \midrule
                LRP \cite{Bach:PLOS2015}                            & $\alpha_{1},\beta_{0}$-rule in the hidden layers, box-rule in the input layer                                     & max-norm \\
                Excitation Backprop \cite{Zhang:ECCV2016}           & $\alpha_{1},\beta_{0}$-rule in all layers                                                                         & max-norm\\
                IG \cite{Sundararajan:ICML2017}                     & mean channel values as baseline image, nb of integration steps $\in$ [300, 1000, 3000, 10000, 30000] such that relative error $<$ 0.01 & max-norm \\
                Guided Backprop \cite{Spring:ICLR2015}              & modified gradient backward pass for all ReLU layers in the network                                               & max-norm \\
                Guided Grad-CAM \cite{Selvaraju:ICCV2017}           & element-wise multiplication of Guided Backprop with modified gradient backward pass for all ReLU layers in the network \& Grad-CAM applied to the ReLU layer output in the last convolutional layer of the network & max-norm \\
                SmoothGrad \cite{Smilkov:ICML2017}                  & nb of samples 300, noise level 0.05, gradient unsquared                                                              & max-norm \\
                VarGrad \cite{Adebayo:ICLR2018}                     & nb of samples 300, noise level 0.05                                                                                & max-norm \\
                Gradient \cite{Simonyan:ICLR2014}                   & gradient unsquared                                                                                                   & max-norm \\
                Gradient$\times$Input \cite{Shrikumar:arxiv2016}    & gradient squared                                                                                                   & max-norm \\
                Deconvnet \cite{Zeiler:ECCV2014}                    & modified gradient backward pass only for ReLU layers within the convolutional neural network part of the network   & max-norm\\
                 Grad-CAM \cite{Selvaraju:ICCV2017}                 & Grad-CAM applied to the ReLU layer output in the last convolutional layer of the network                      & none\\
                \bottomrule
			\end{tabular}
		}

	\end{center}
\end{table}
\setlength{\tabcolsep}{1.4pt}

\FloatBarrier

\section{Heatmaps}\label{appendix:heatmaps}

In Figures~\ref{table:heatmap-simple-correct-3601}-\ref{table:heatmap-complex-correct-49997} we provide example heatmaps for CLEVR-XAI questions. In all cases we use as the target class for the XAI method the neural network's \textit{predicted} class.

We selected the example data points automatically in the following way: we conducted a search over the CLEVR-XAI-simple and CLEVR-XAI-complex subsets, resp., and retrieved the points with the highest predicted softmax probabilities.
For Fig.~\ref{table:heatmap-simple-correct-3601}-~\ref{table:heatmap-simple-correct-17094} we considered all correctly predicted data points from CLEVR-XAI-simple. For Fig.~\ref{table:heatmap-simple-false-33038}-~\ref{table:heatmap-simple-false-24357} we considered all falsely predicted data points from CLEVR-XAI-simple that query an object's color. And for Fig.~\ref{table:heatmap-complex-correct-65806}-~\ref{table:heatmap-complex-correct-49997} we considered all correctly predicted data points from CLEVR-XAI-complex.

For the heatmap visualization we depict the raw heatmap (after pooling the relevance along the channel axis according to the pooling technique that was tuned towards the relevance \textit{mass} accuracy, as was described in Section~\ref{sec:experiments-simple-questions}) and color code the relevance value in red (normalizing the color to the maximum value per heatmap), using Matplotlib's \texttt{seismic} colormap.
Additionally we visualize the original image overlayed with the heatmap. For the latter visualization we first apply gaussian blur to the heatmap with standard deviation 0.02 times the image dimension, and subsequently color code the heatmap with Matplotlib's \texttt{gist\_ncar} colormap (normalized to the extremal relevance values per heatmap).

Especially Fig.~\ref{table:heatmap-simple-false-33038} and \ref{table:heatmap-simple-false-24357} illustrate why falsely predicted data points should not be taken into account when evaluating XAI methods: indeed for these questions the model focused on the wrong object in the image to produce the answer. For Fig.~\ref{table:heatmap-simple-false-33038} it obviously focused on the small brown cylinder rather than on the small yellow ball. For Fig.~\ref{table:heatmap-simple-false-24357} it appears to have focused on the large blue cube rather than on the small purple cube. On Fig.~\ref{table:heatmap-simple-false-22552}, however, it seems the neural network was able to detect the right object (the small red ball), but apparently it mistakenly interpreted its color as being brown.

\begin{table}
        \scriptsize
		\caption{Heatmaps for a correctly predicted CLEVR-XAI-simple question (raw heatmap and heatmap overlayed with original image), and corresponding relevance \textit{mass} accuracy.}
		\label{table:heatmap-simple-correct-3601}
\begin{tabular}{lllc}
\midrule
\begin{tabular}{@{}l@{}}What is the material \\ of the large ball? \\ \textit{metal} \end{tabular}  & \includegraphics[width=.18\linewidth,valign=m]{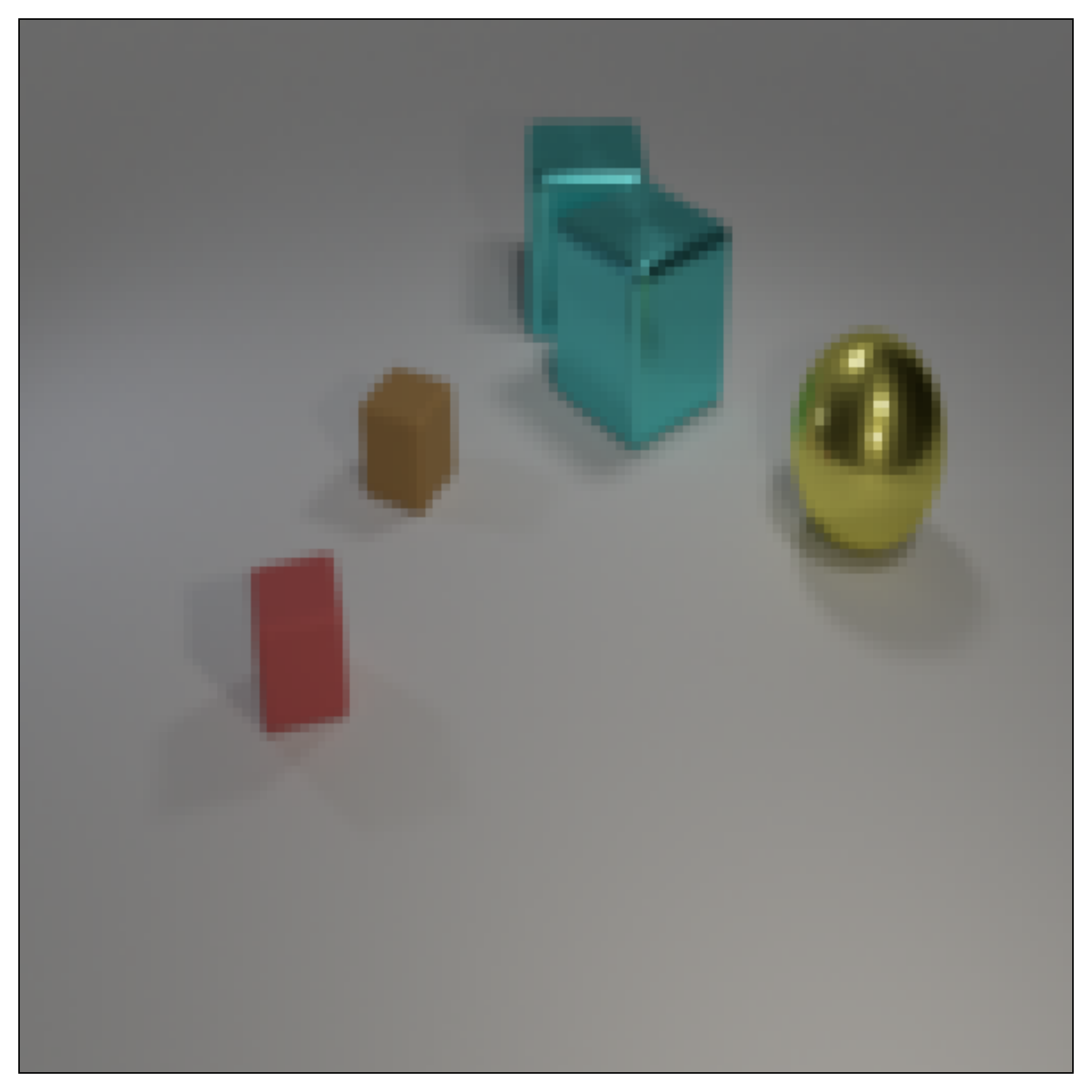} &
\includegraphics[width=.18\linewidth,valign=m]{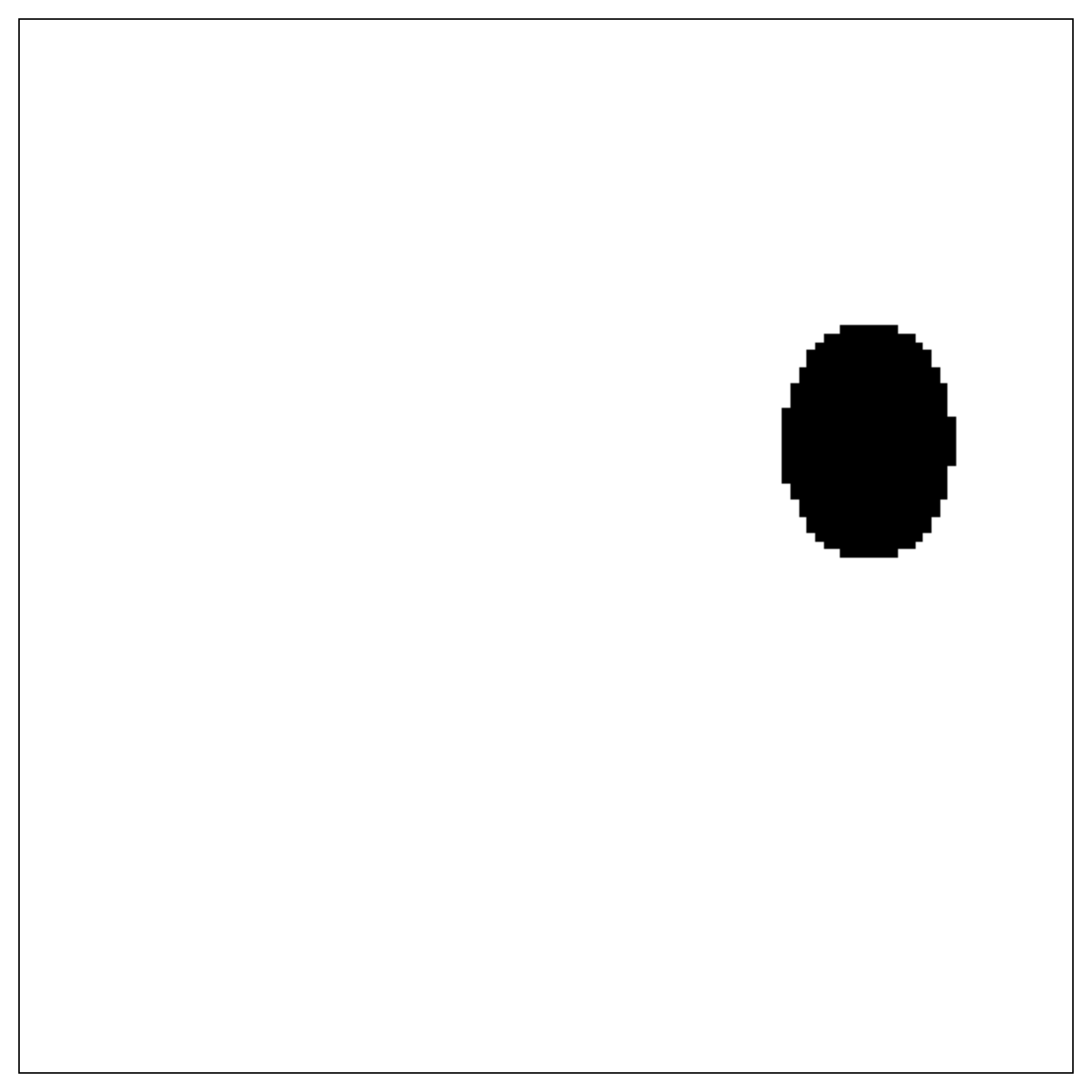} & GT Single Object \\
\midrule
LRP \cite{Bach:PLOS2015}                            & \includegraphics[width=.12\linewidth,valign=m]{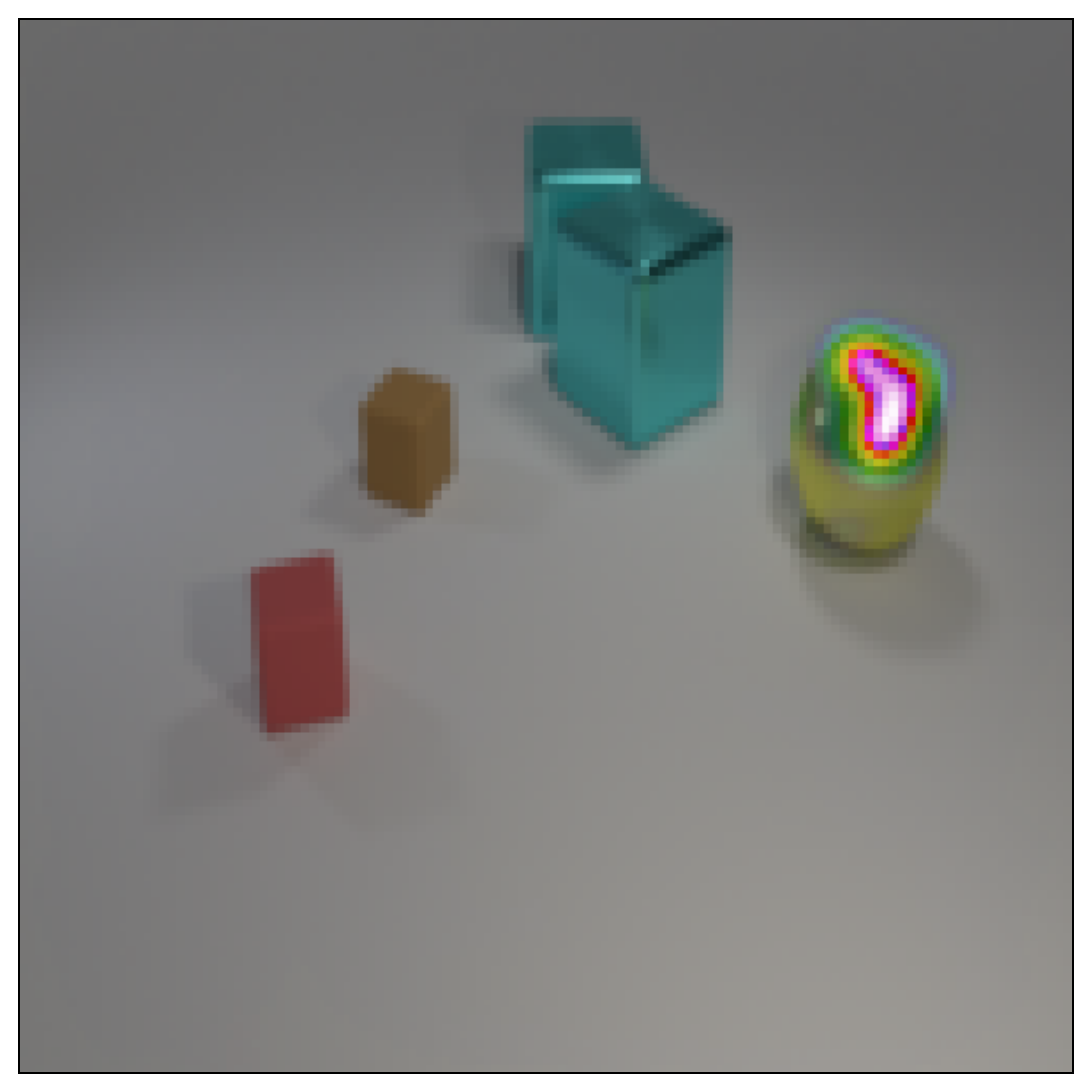} & \includegraphics[width=.12\linewidth,valign=m]{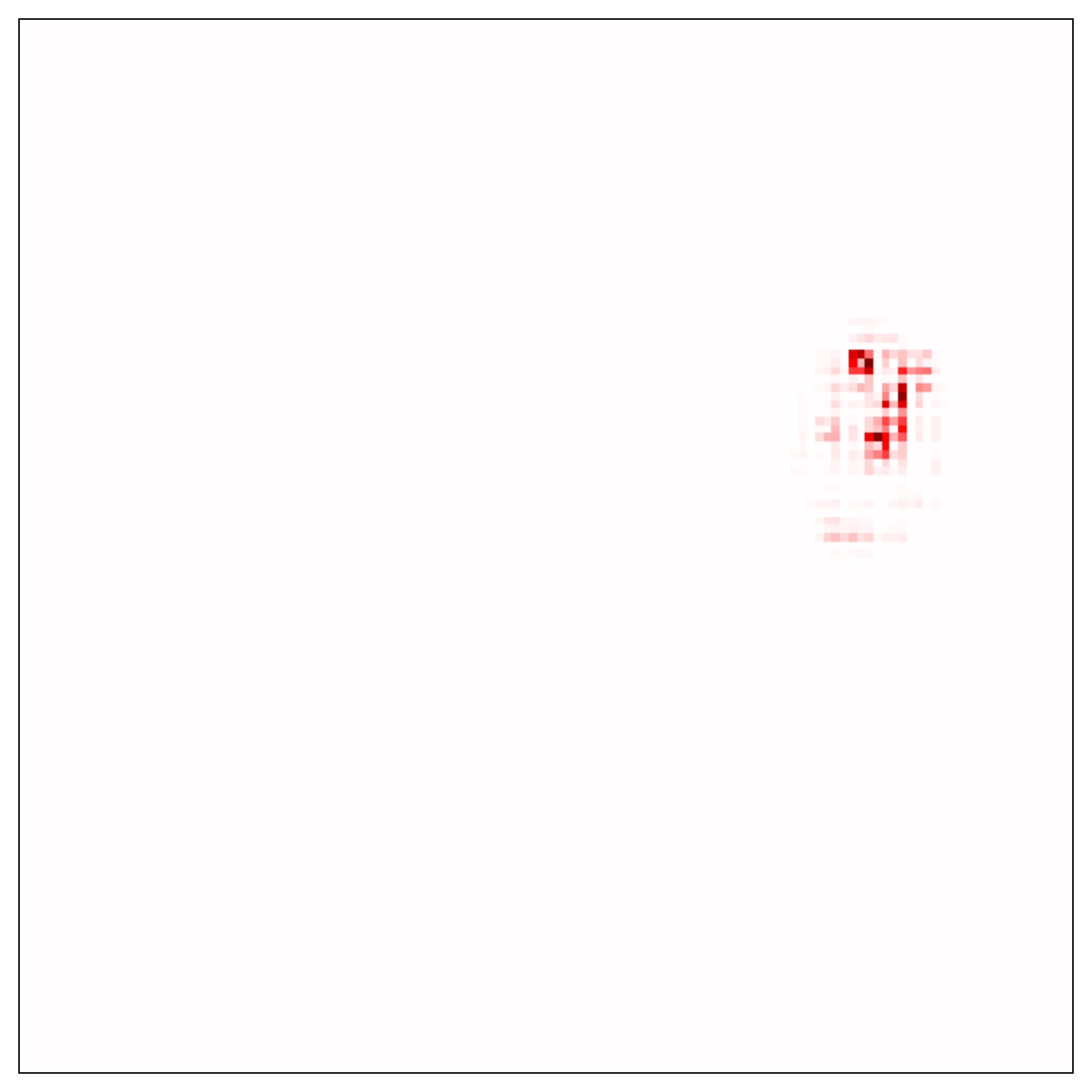} & 0.98 \\
Excitation Backprop \cite{Zhang:ECCV2016}           & \includegraphics[width=.12\linewidth,valign=m]{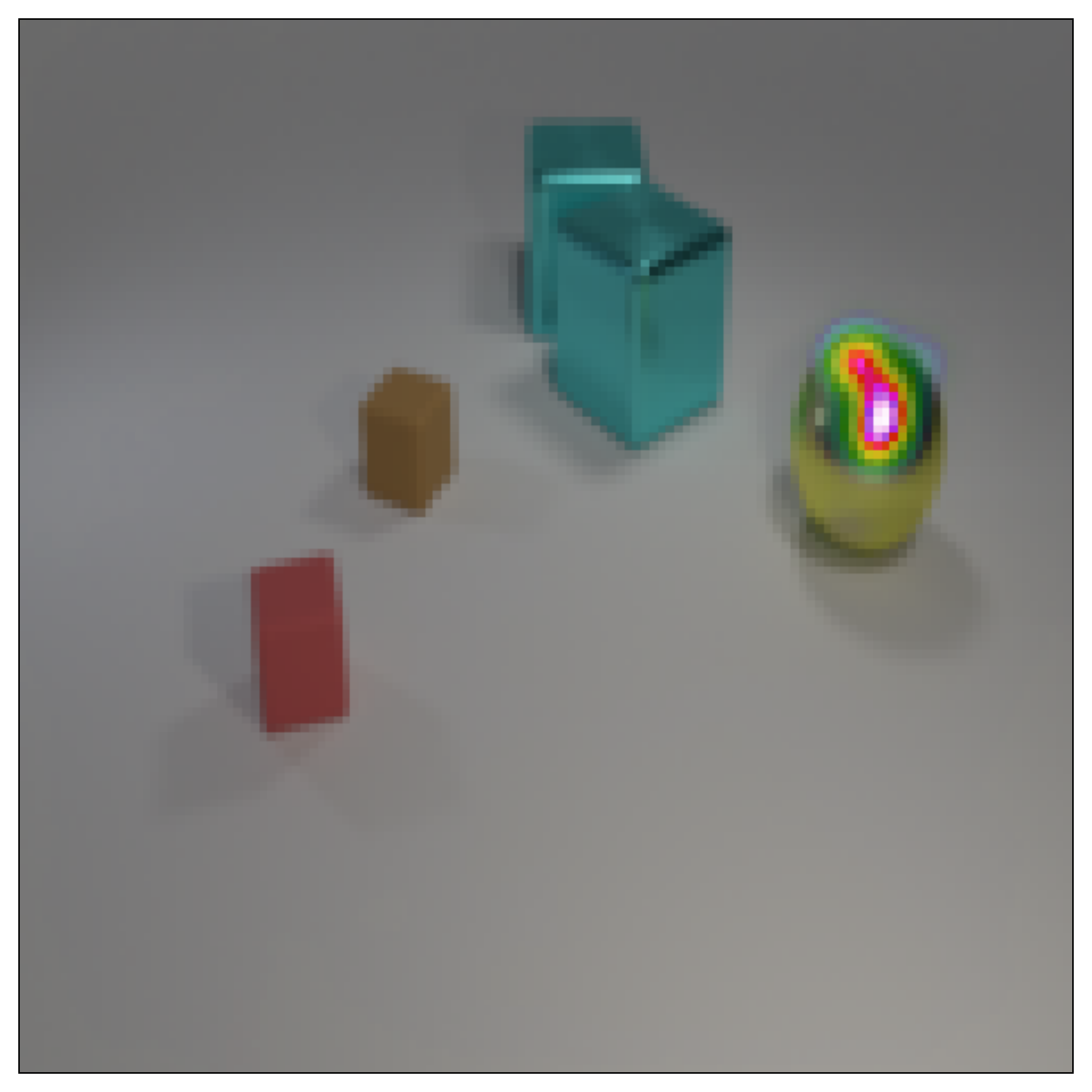} & \includegraphics[width=.12\linewidth,valign=m]{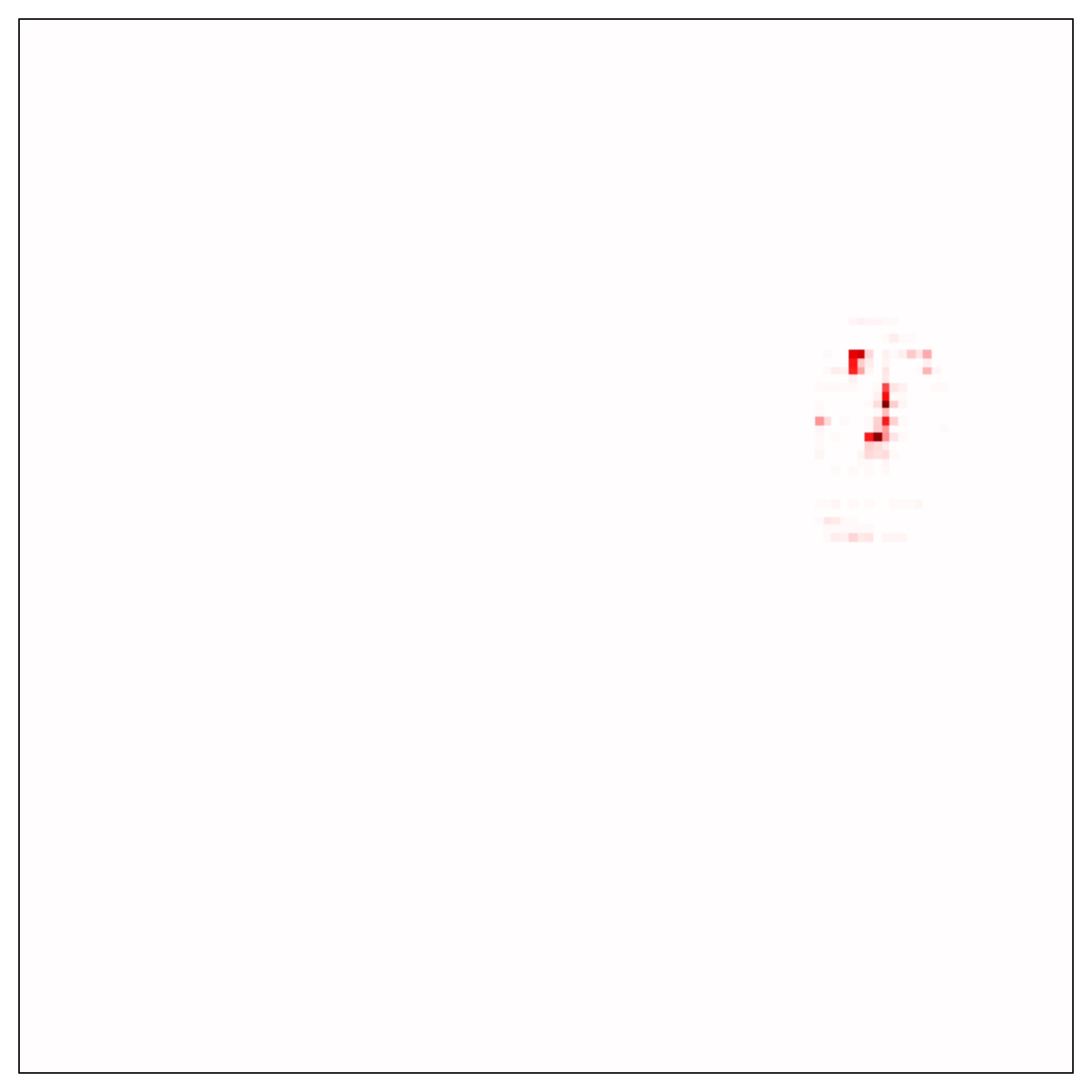} & 0.97 \\
IG \cite{Sundararajan:ICML2017}                     & \includegraphics[width=.12\linewidth,valign=m]{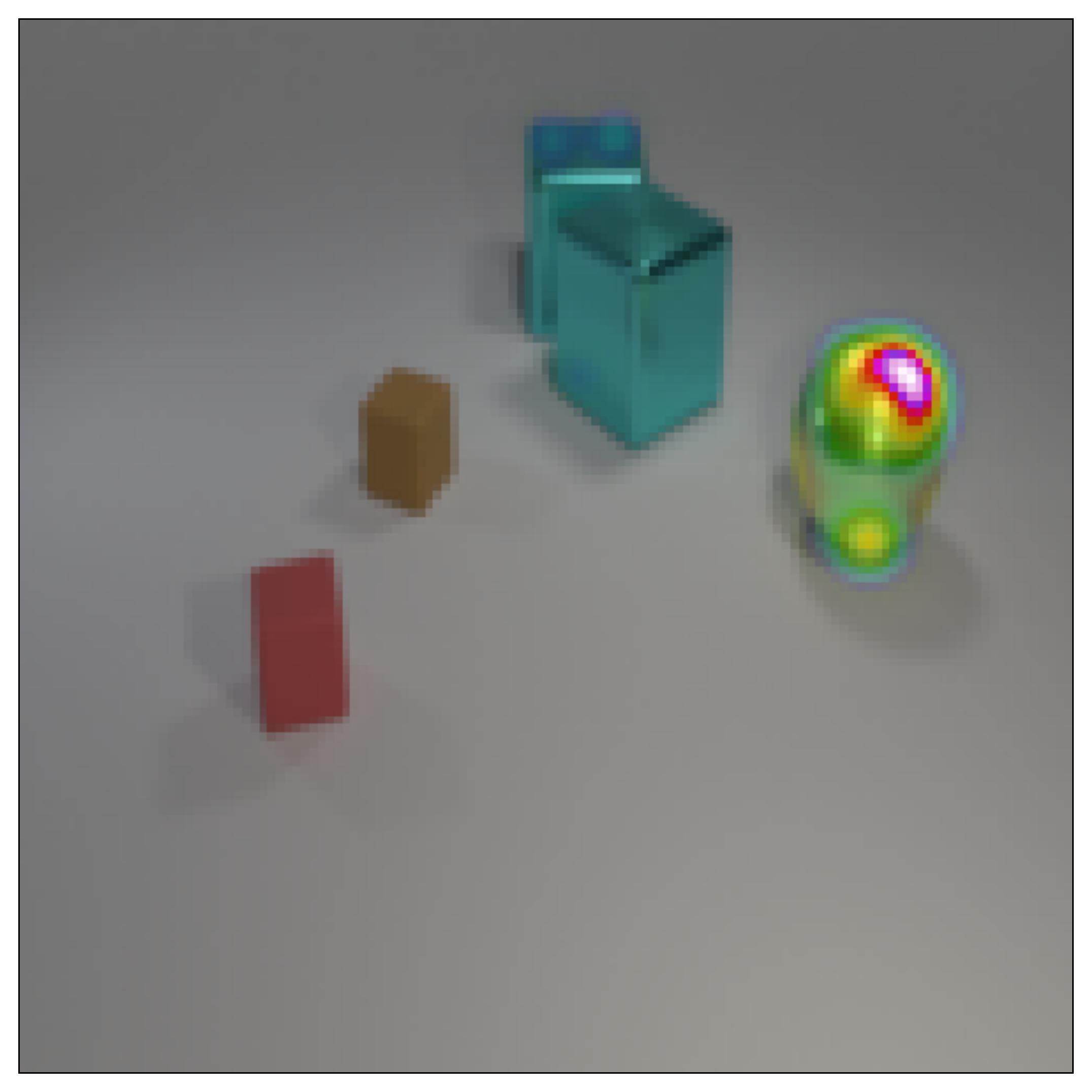} & \includegraphics[width=.12\linewidth,valign=m]{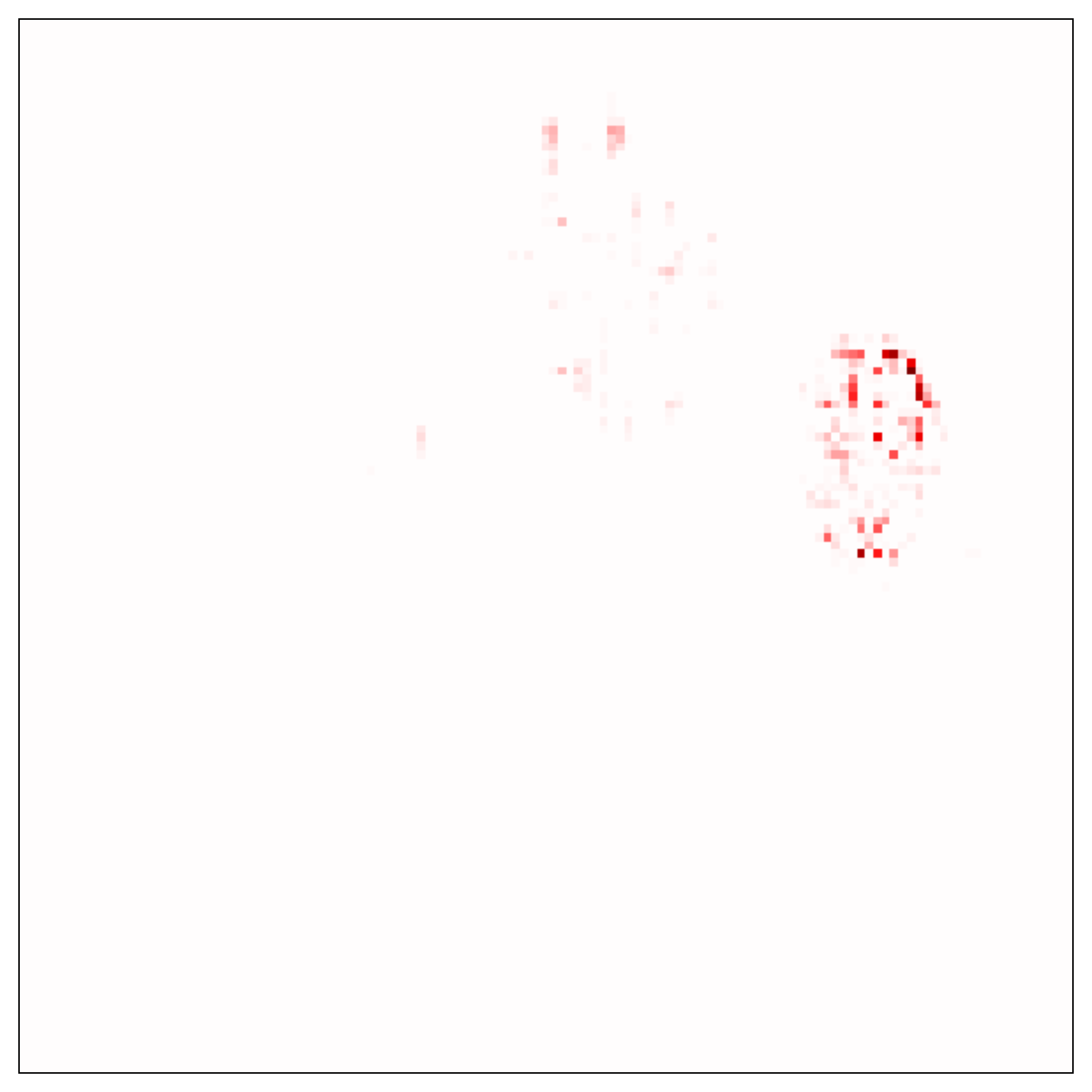} & 0.80 \\
Guided Backprop \cite{Spring:ICLR2015}              & \includegraphics[width=.12\linewidth,valign=m]{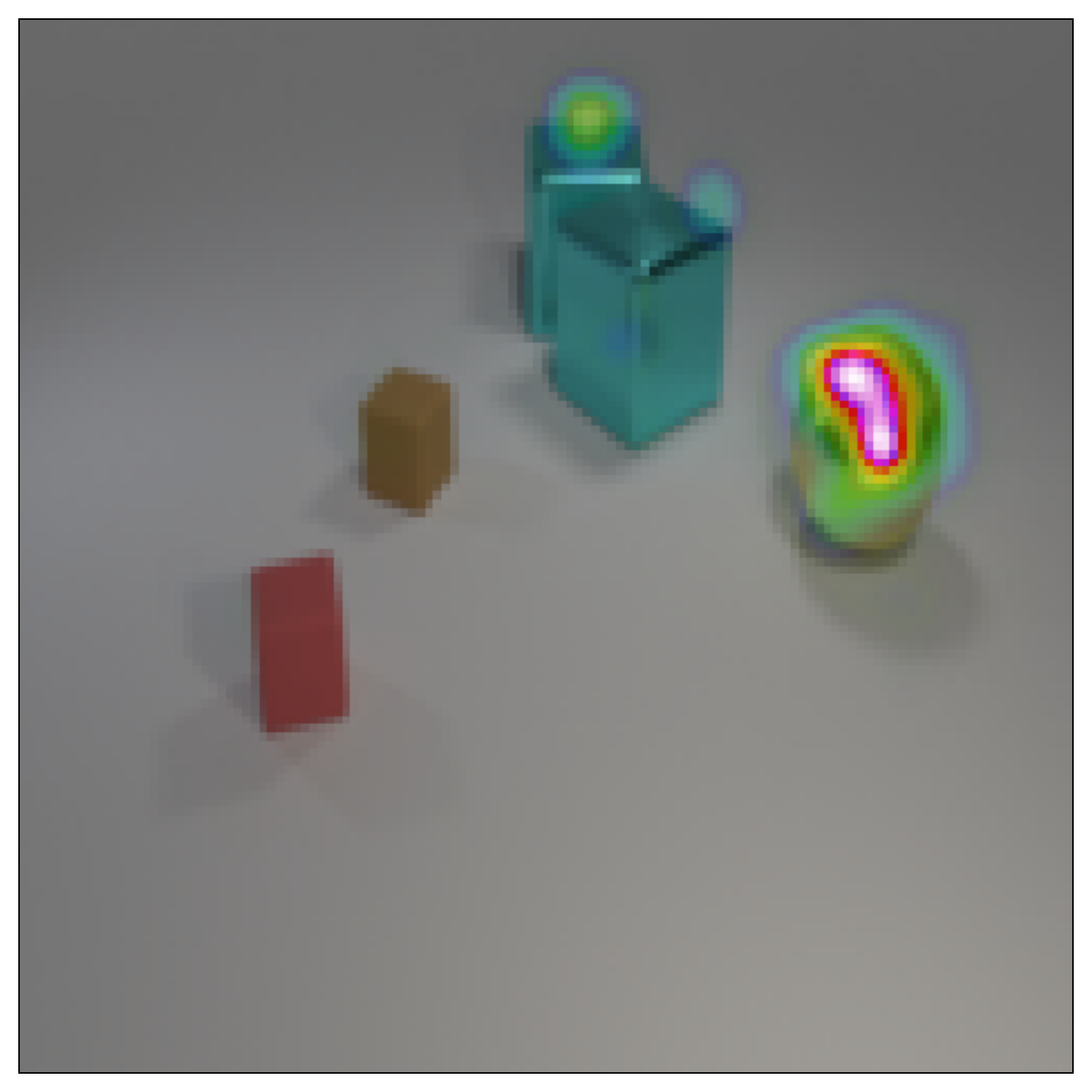} & \includegraphics[width=.12\linewidth,valign=m]{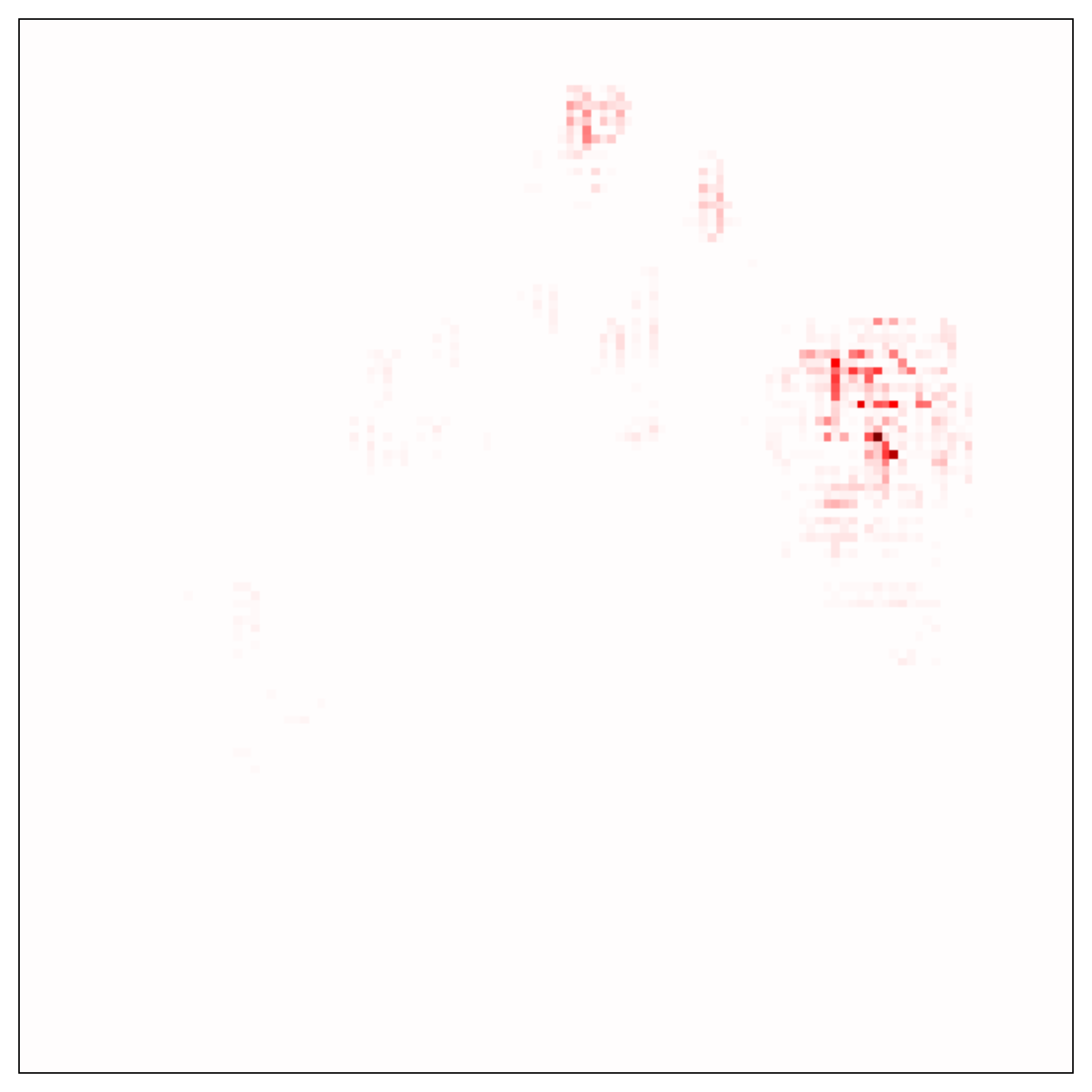} & 0.58 \\
Guided Grad-CAM \cite{Selvaraju:ICCV2017}           & \includegraphics[width=.12\linewidth,valign=m]{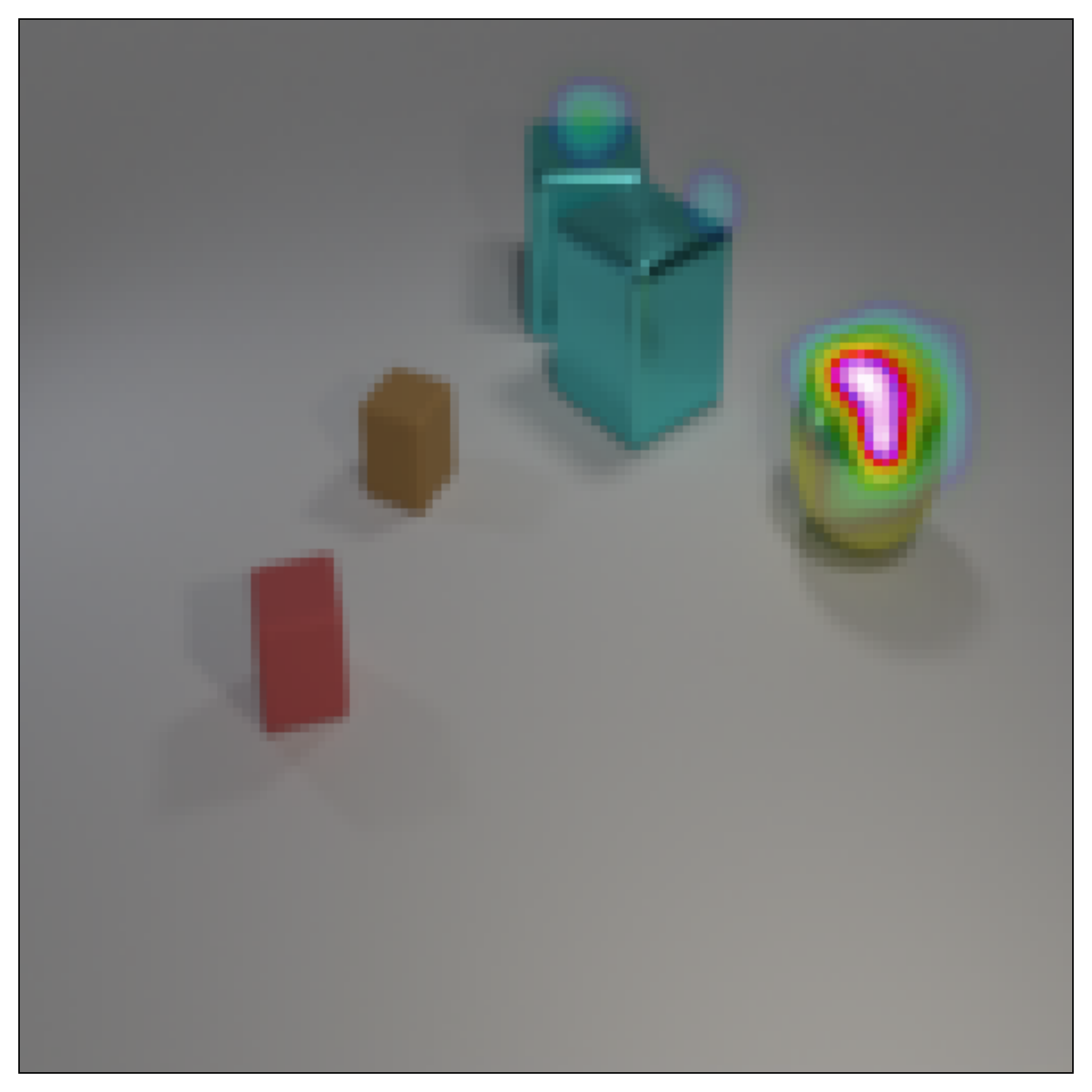} & \includegraphics[width=.12\linewidth,valign=m]{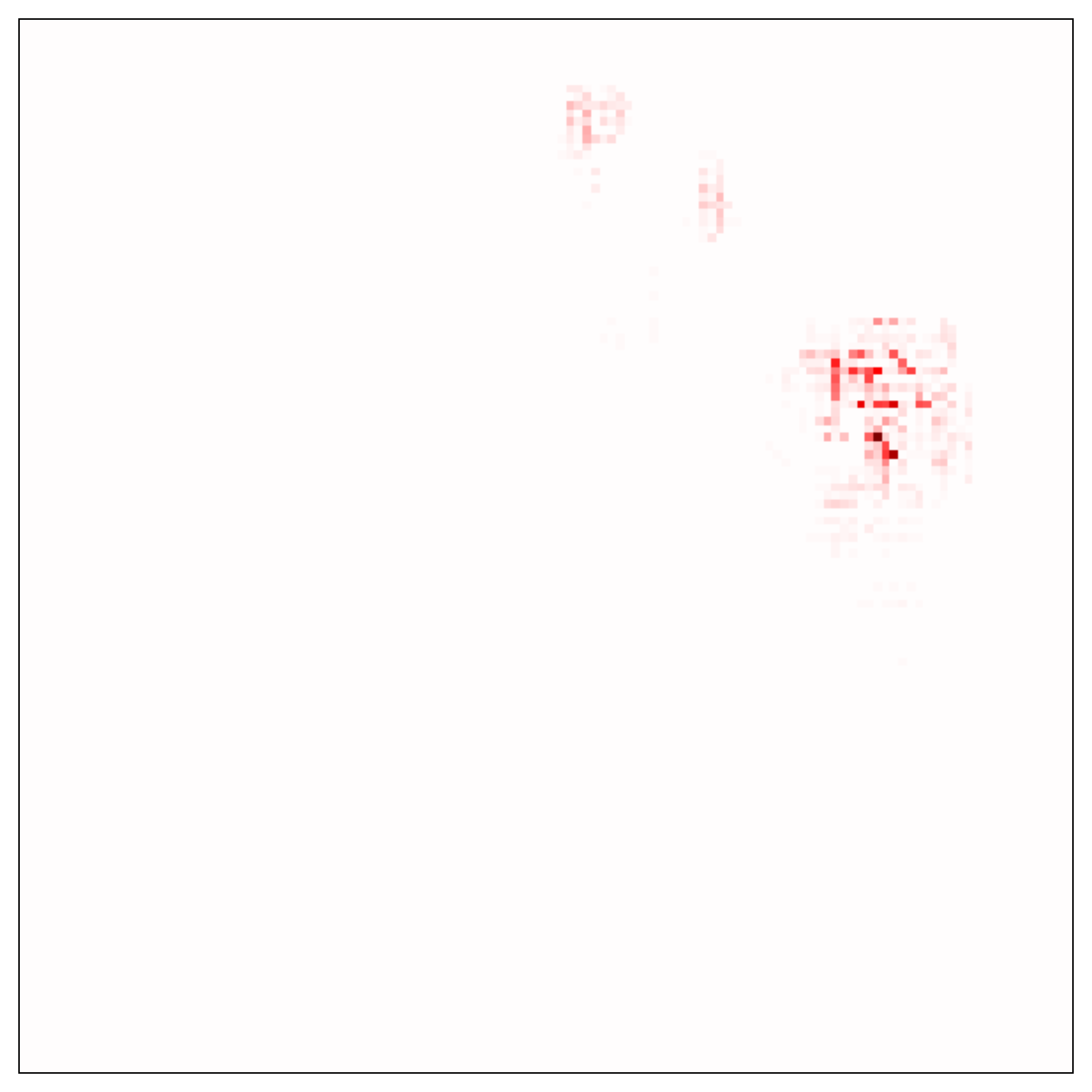} & 0.72 \\
SmoothGrad \cite{Smilkov:ICML2017}                  & \includegraphics[width=.12\linewidth,valign=m]{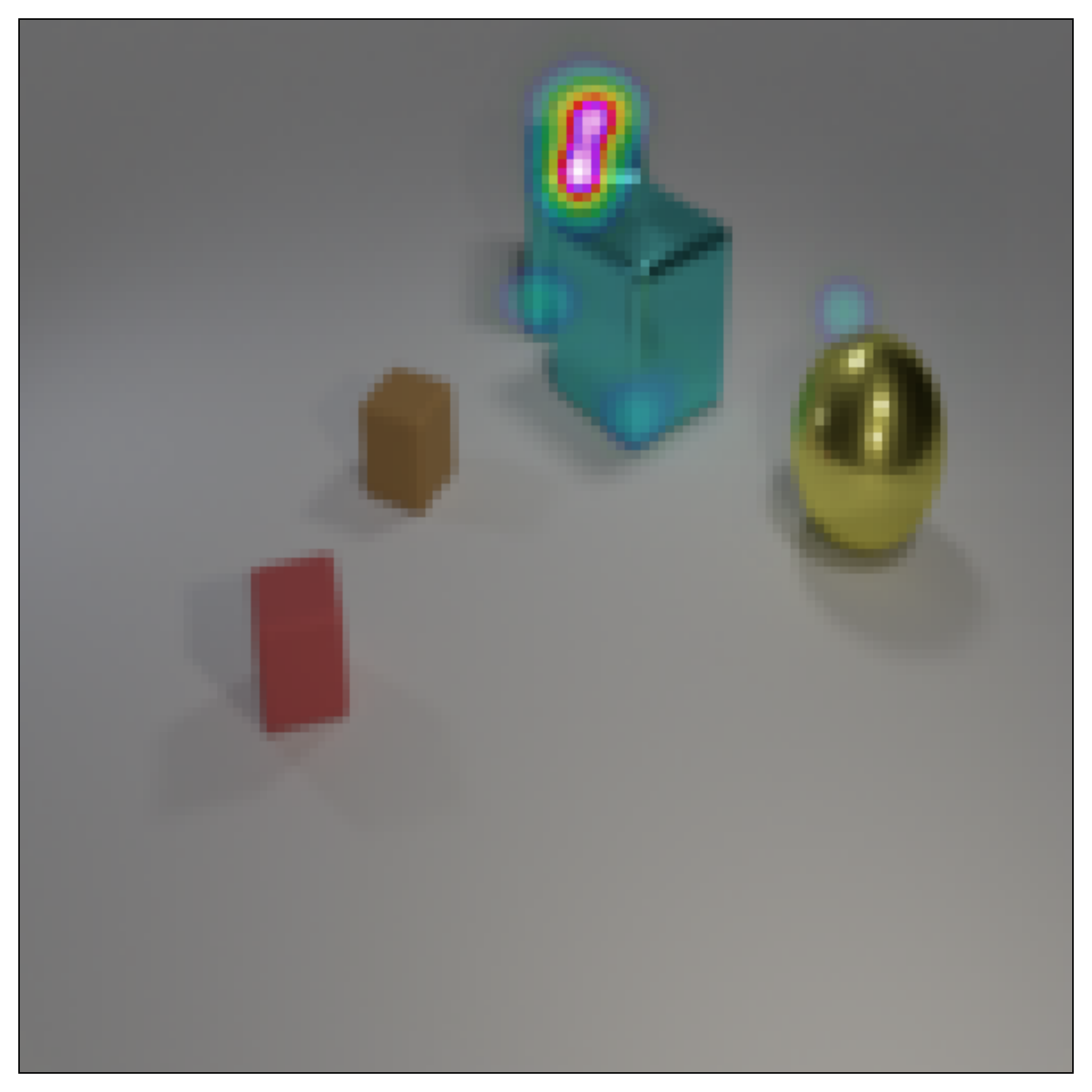} & \includegraphics[width=.12\linewidth,valign=m]{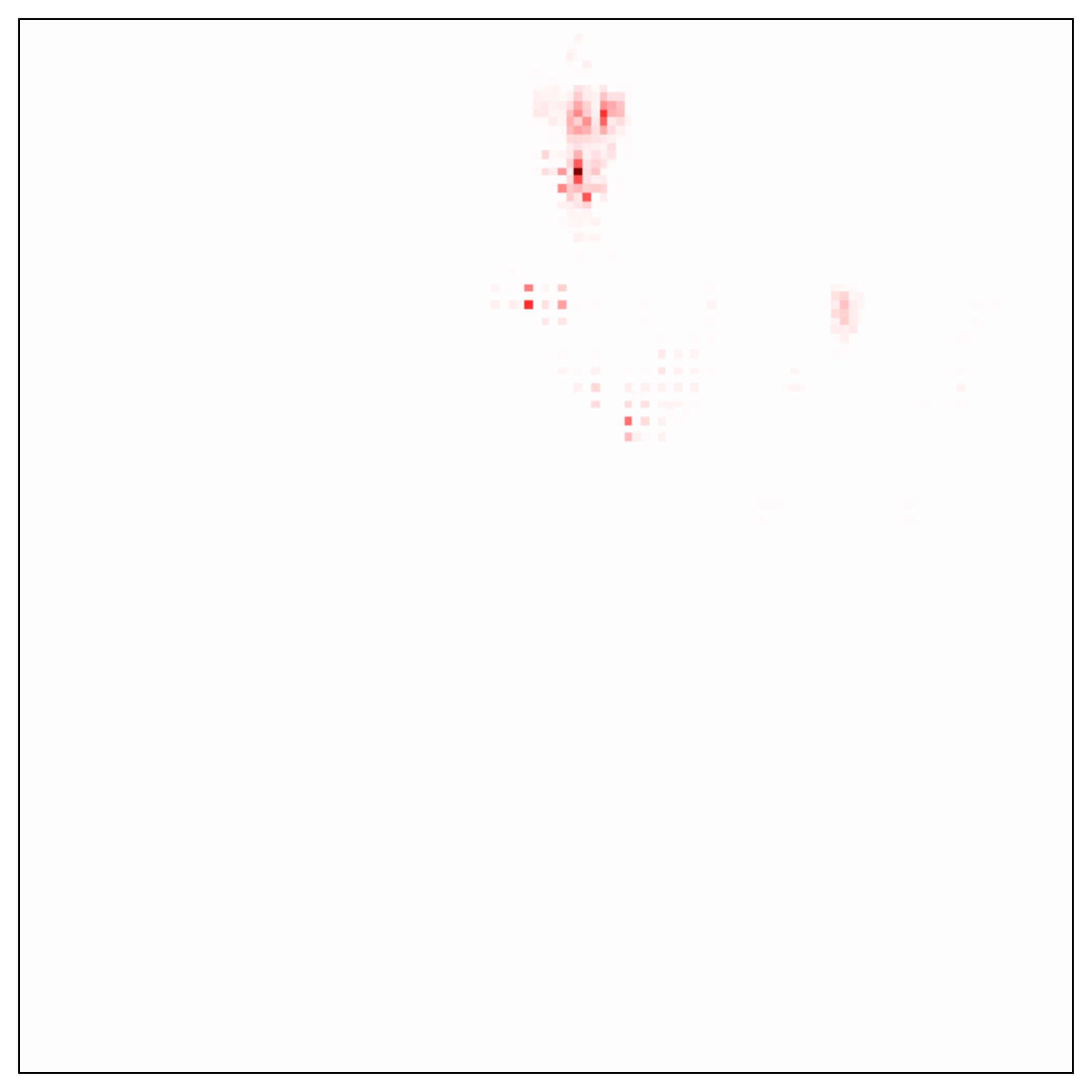} & 0.03 \\
VarGrad \cite{Adebayo:ICLR2018}                     & \includegraphics[width=.12\linewidth,valign=m]{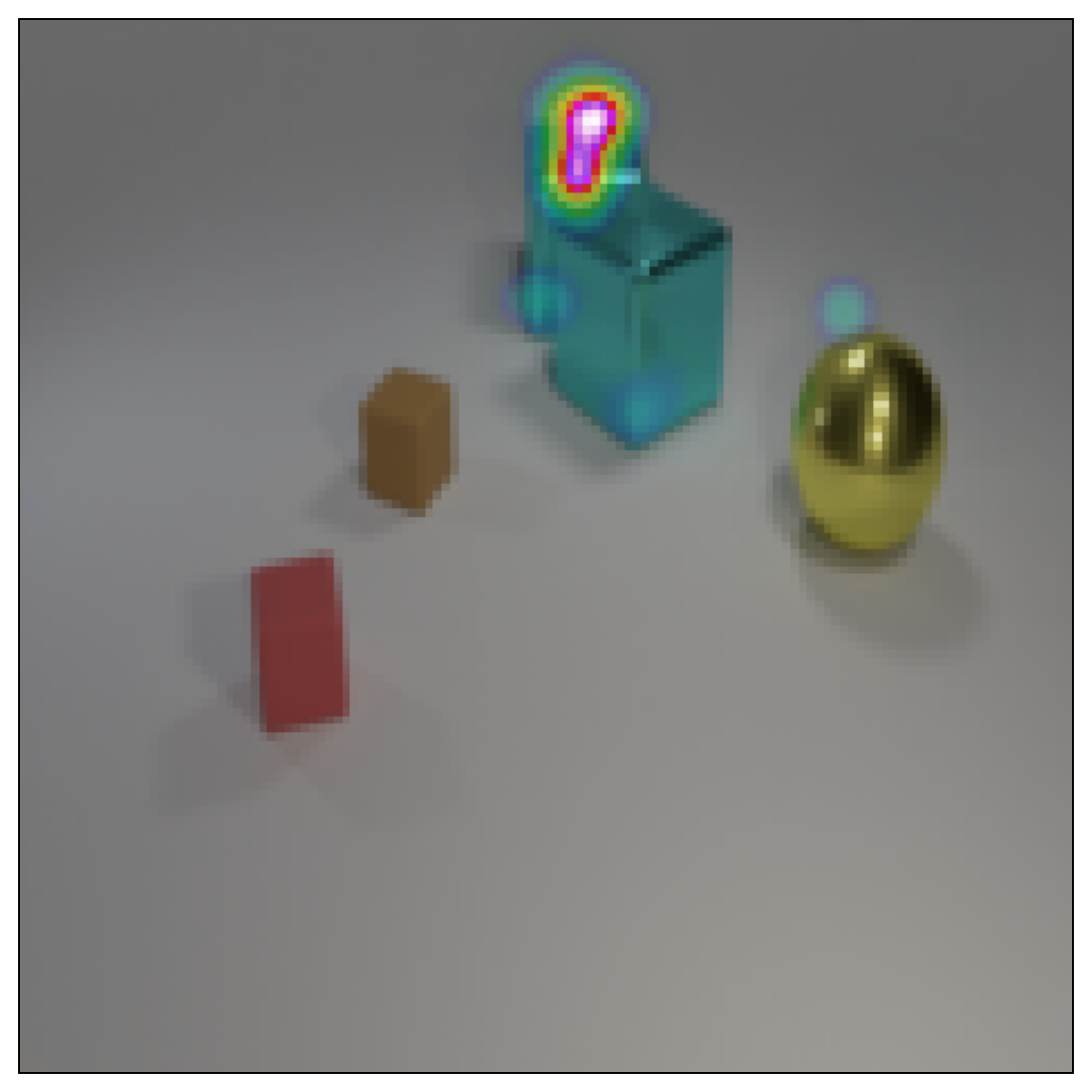} & \includegraphics[width=.12\linewidth,valign=m]{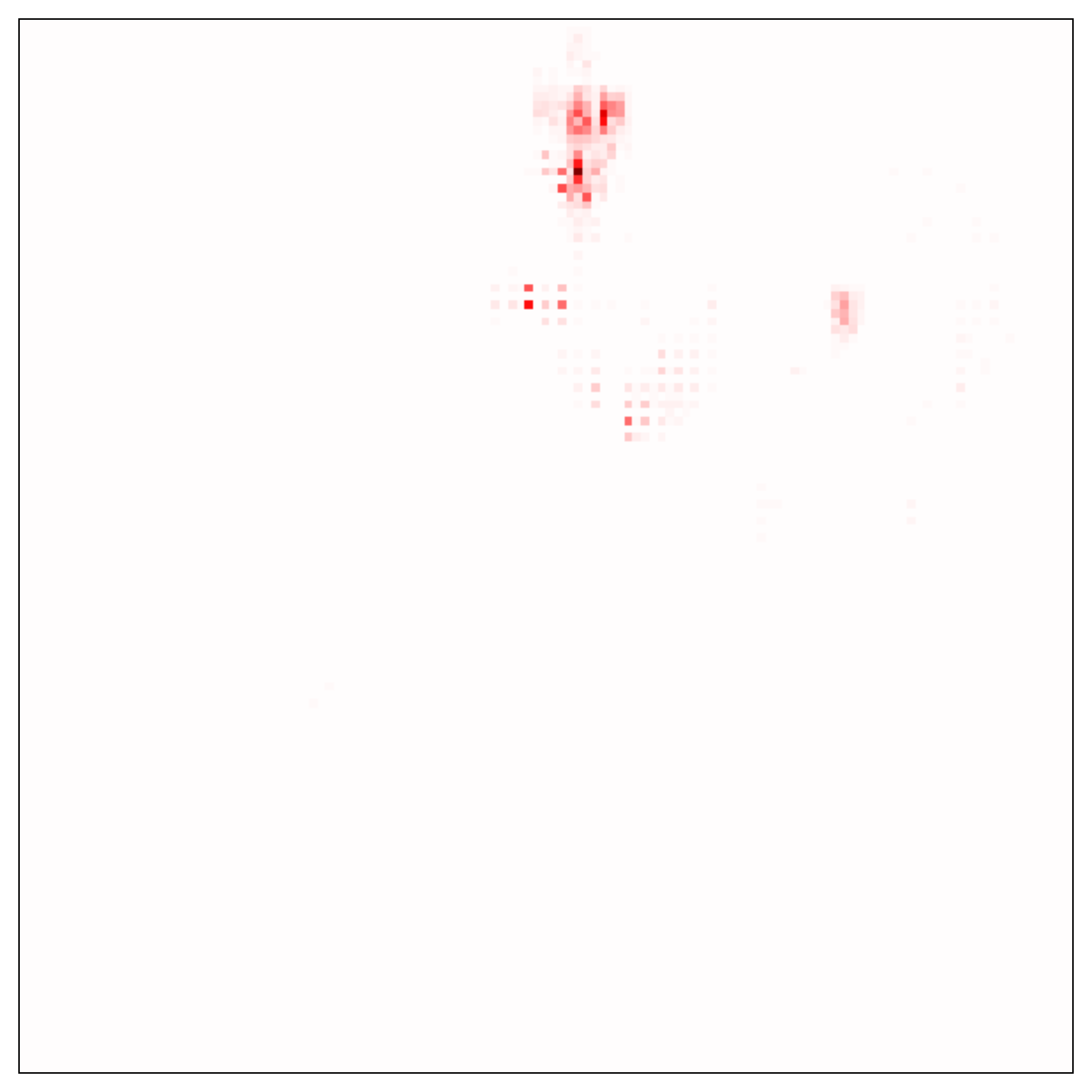} & 0.02 \\
Gradient \cite{Simonyan:ICLR2014}                   & \includegraphics[width=.12\linewidth,valign=m]{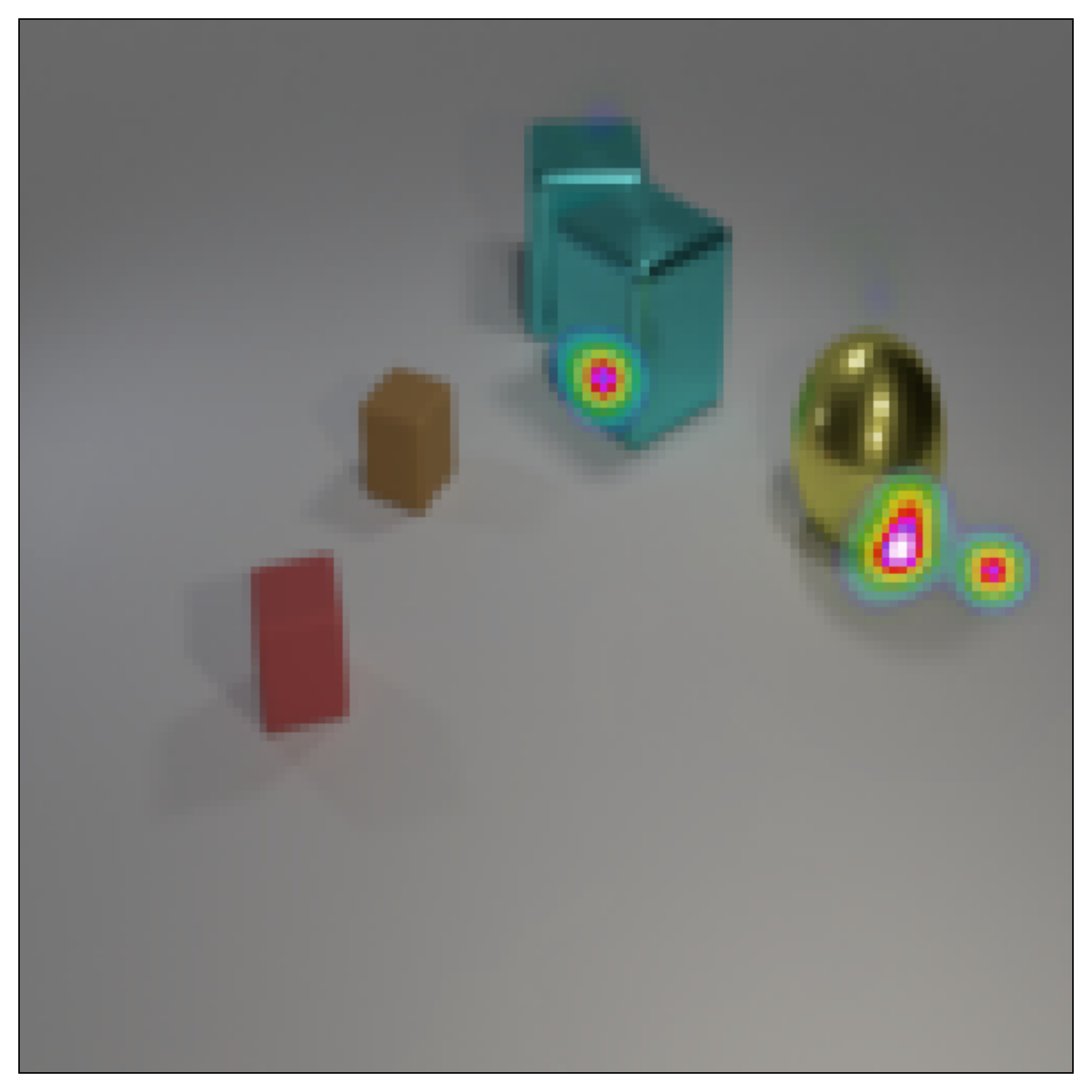} & \includegraphics[width=.12\linewidth,valign=m]{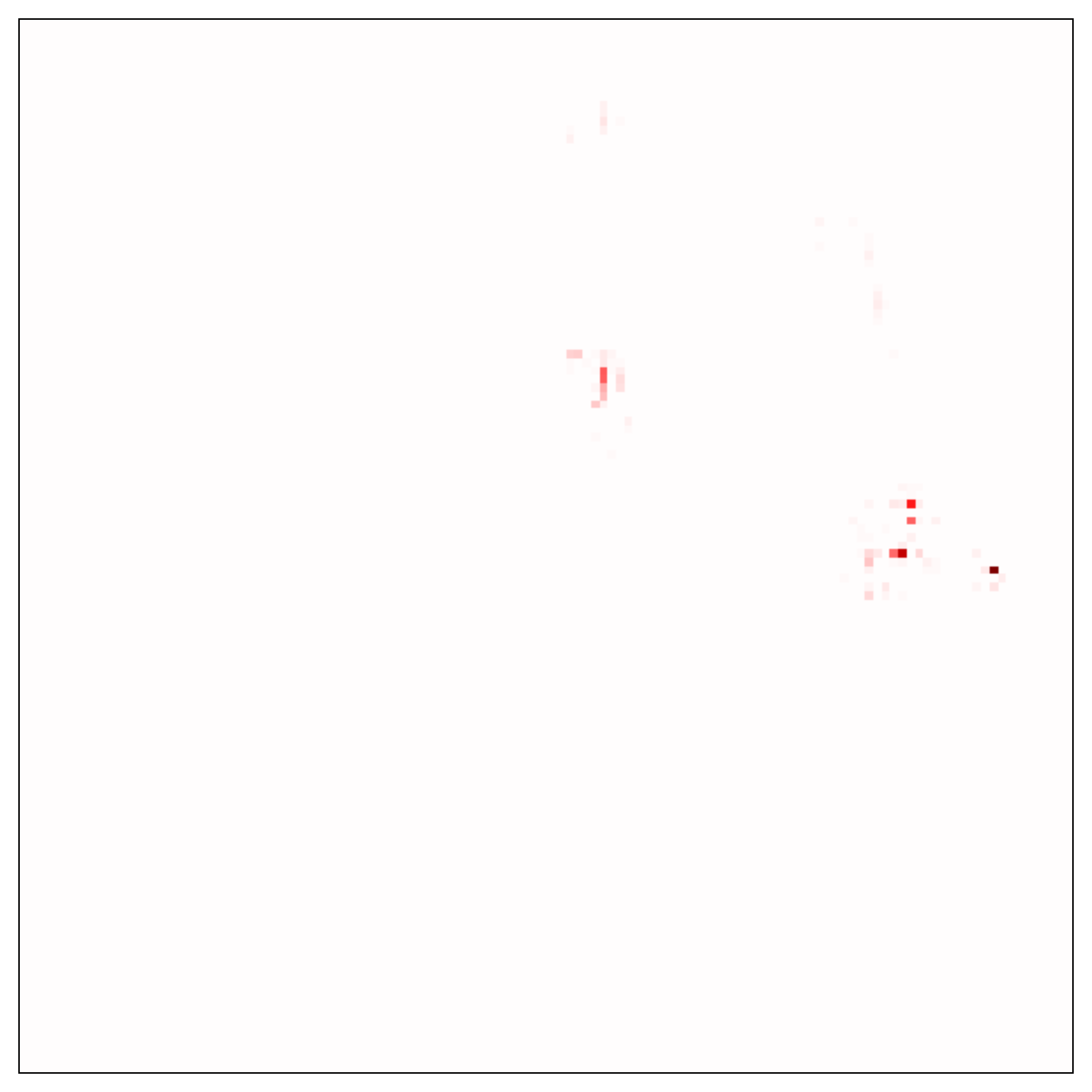} & 0.25 \\
Gradient$\times$Input \cite{Shrikumar:arxiv2016}    & \includegraphics[width=.12\linewidth,valign=m]{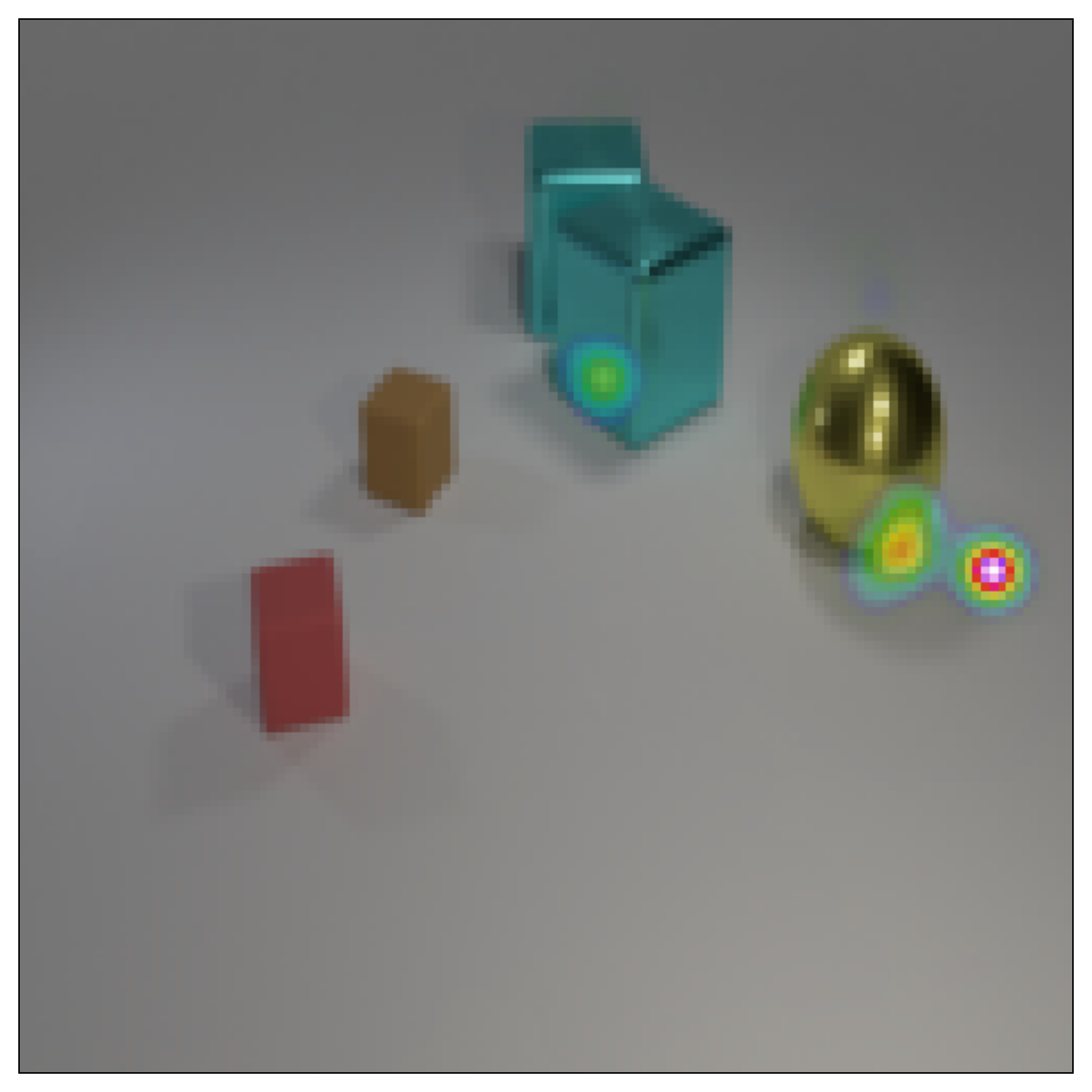} & \includegraphics[width=.12\linewidth,valign=m]{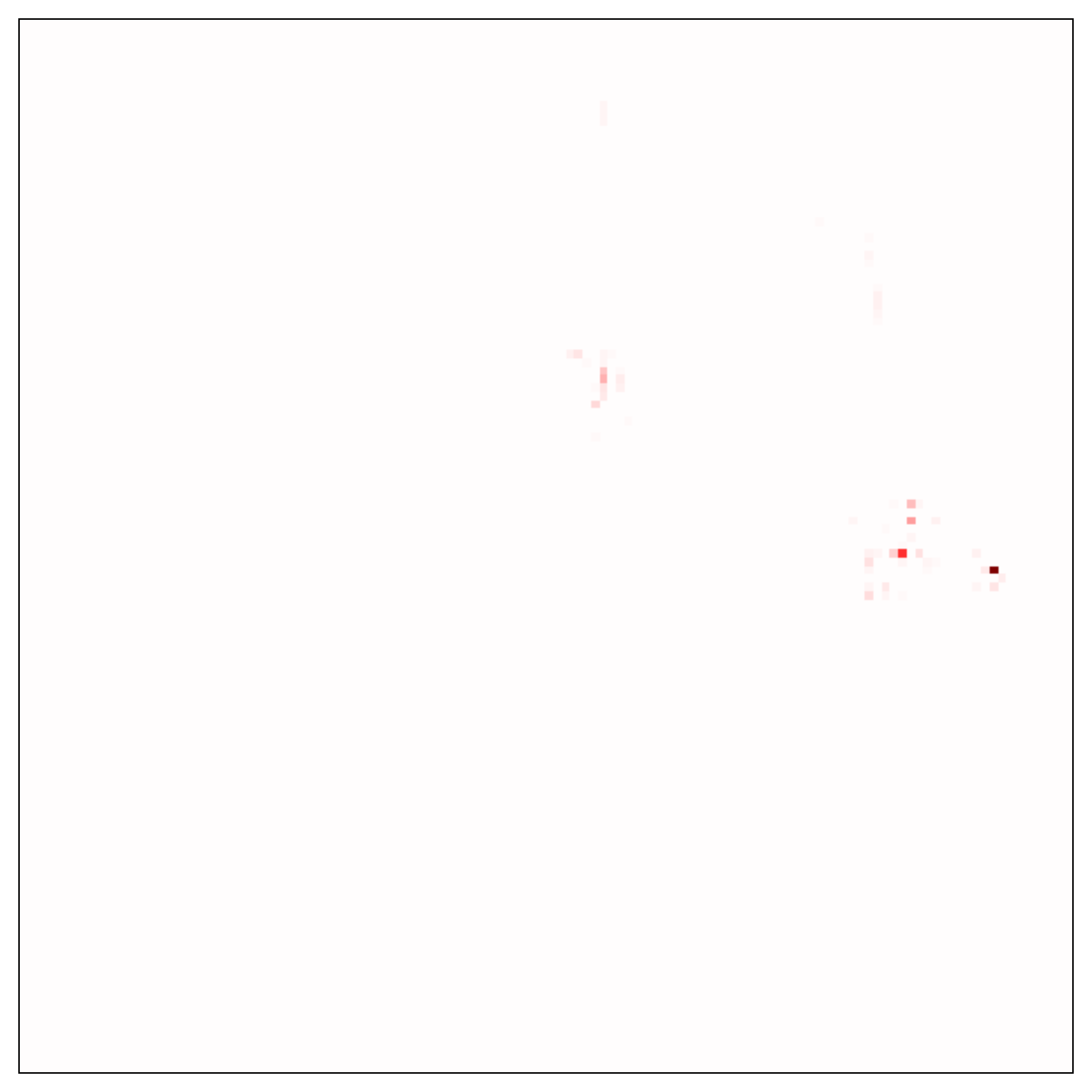} & 0.17 \\
Deconvnet \cite{Zeiler:ECCV2014}                    & \includegraphics[width=.12\linewidth,valign=m]{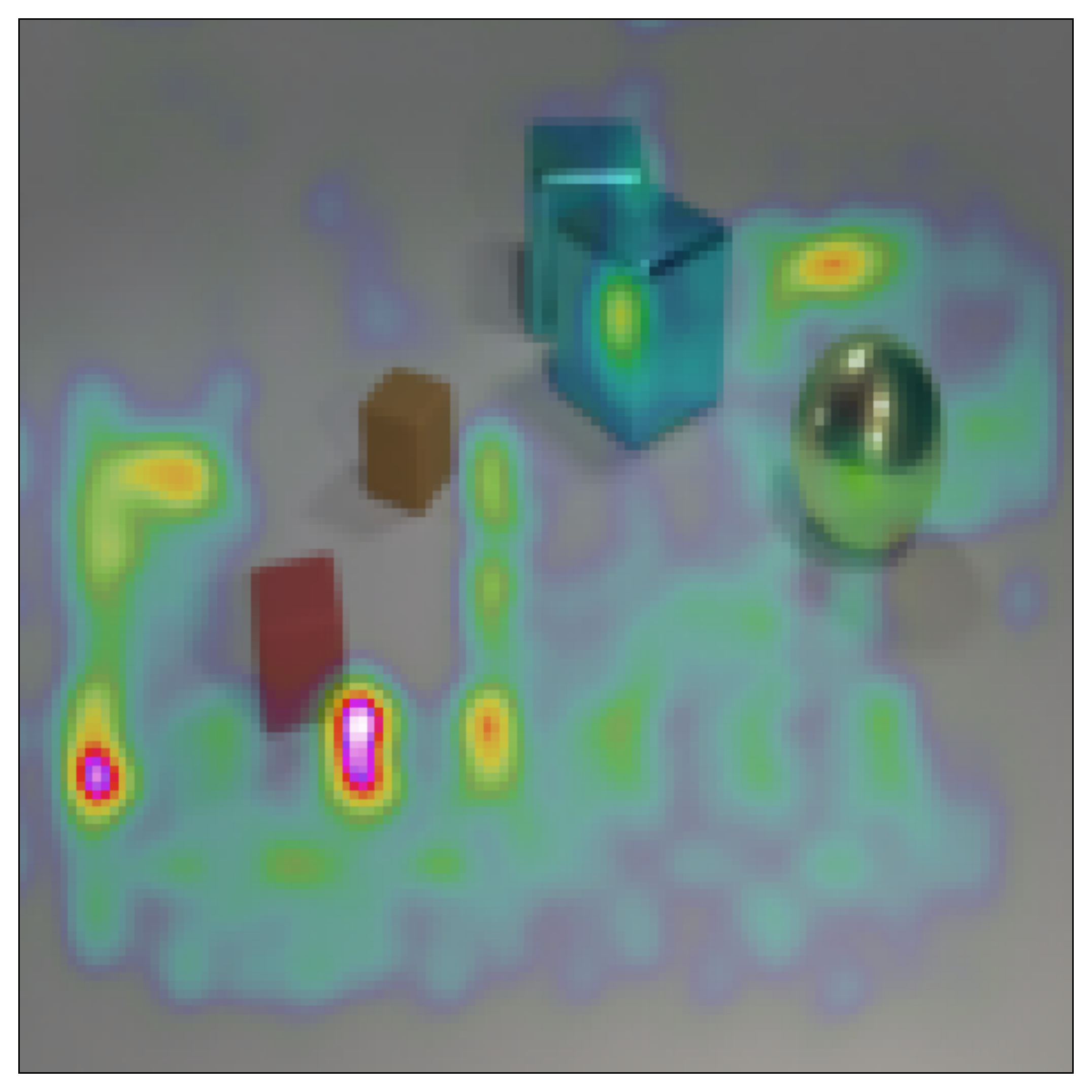} & \includegraphics[width=.12\linewidth,valign=m]{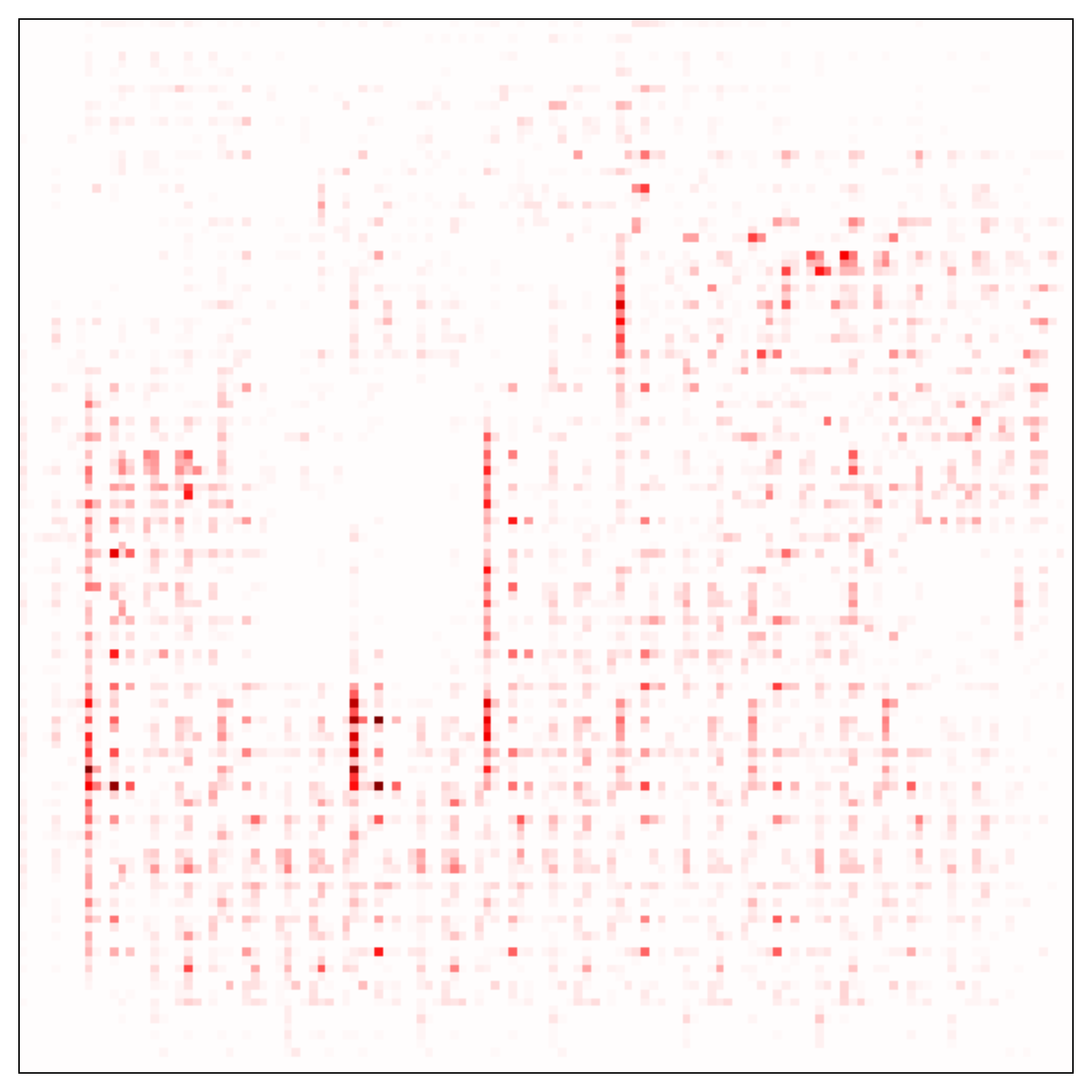} & 0.04 \\
Grad-CAM \cite{Selvaraju:ICCV2017}                  & \includegraphics[width=.12\linewidth,valign=m]{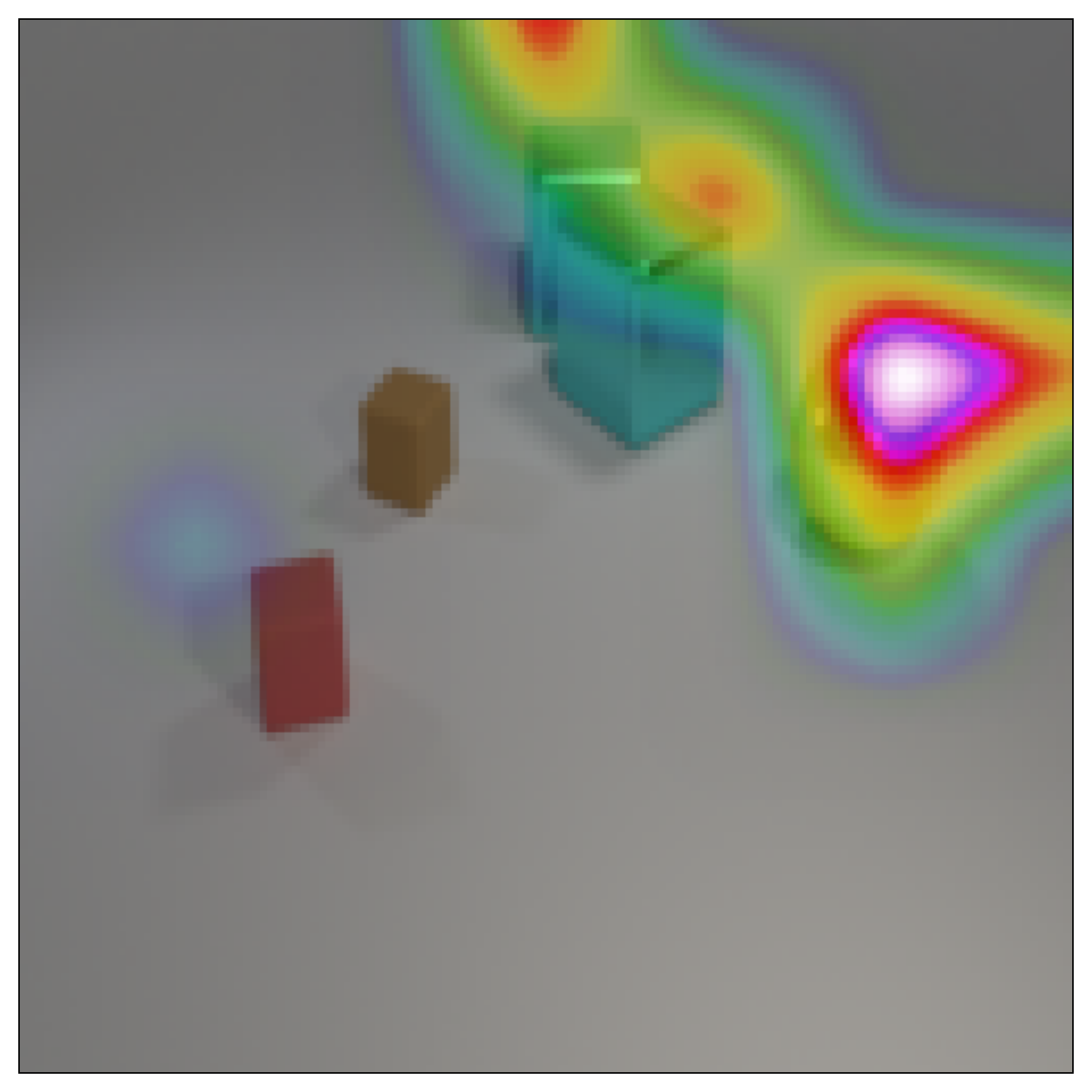} & \includegraphics[width=.12\linewidth,valign=m]{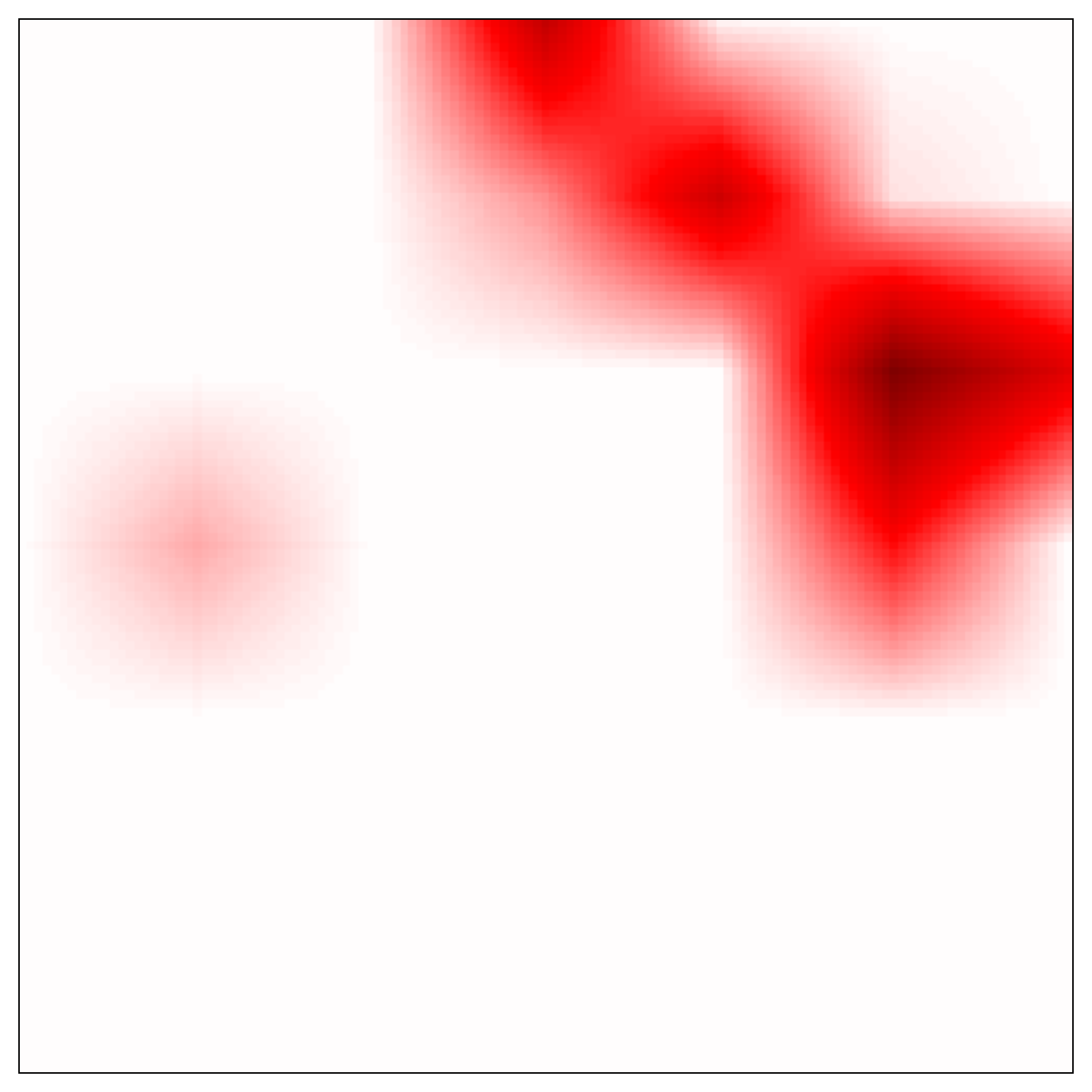} & 0.19 \\
\end{tabular}
\end{table}

\begin{table}
        \scriptsize
		\caption{Heatmaps for a correctly predicted CLEVR-XAI-simple question (raw heatmap and heatmap overlayed with original image), and corresponding relevance \textit{mass} accuracy.}
		\label{table:heatmap-simple-correct-30667}
\begin{tabular}{lllc}
\midrule
\begin{tabular}{@{}l@{}}What is the large cyan \\ ball made of?\\ \textit{metal} \end{tabular}  & \includegraphics[width=.18\linewidth,valign=m]{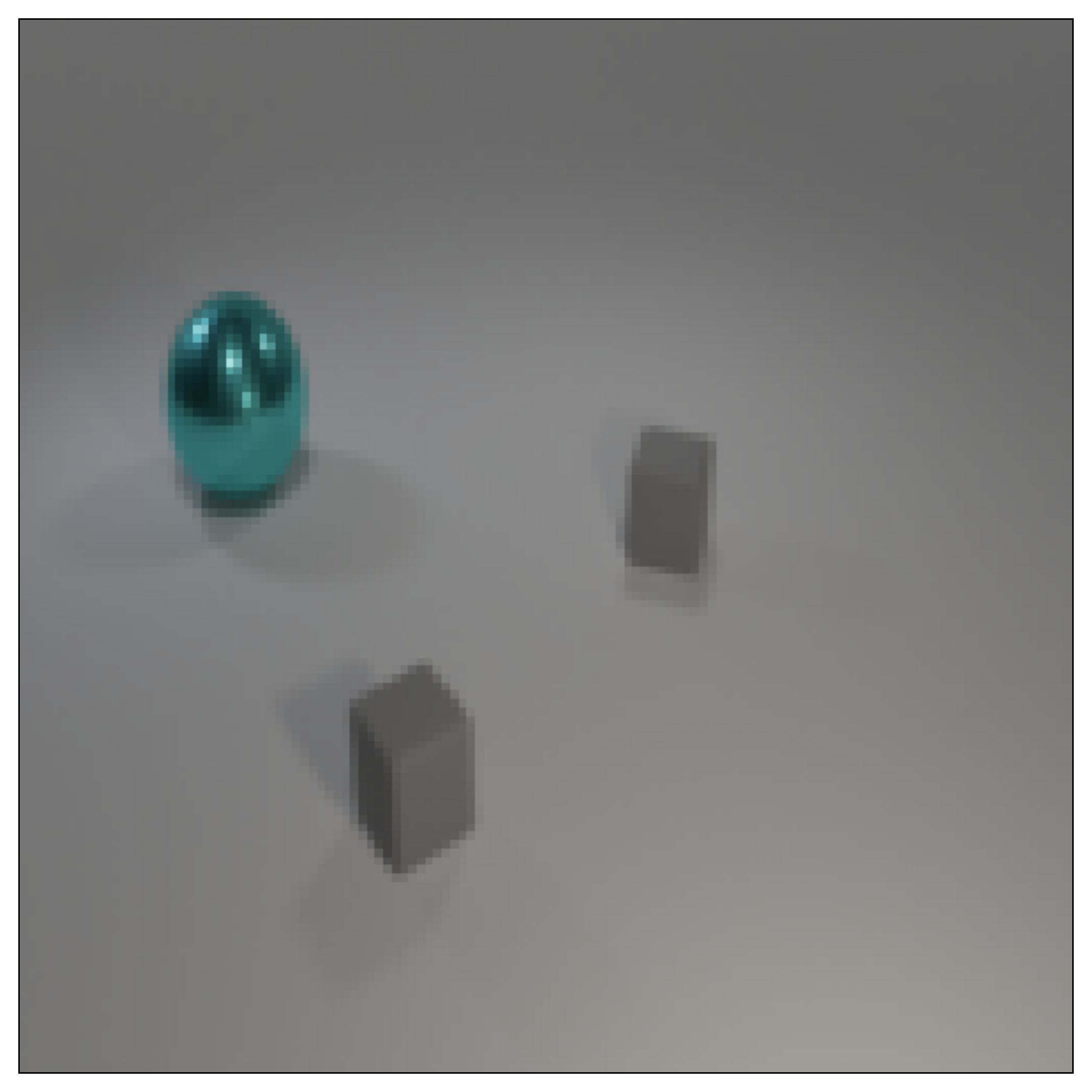} &
\includegraphics[width=.18\linewidth,valign=m]{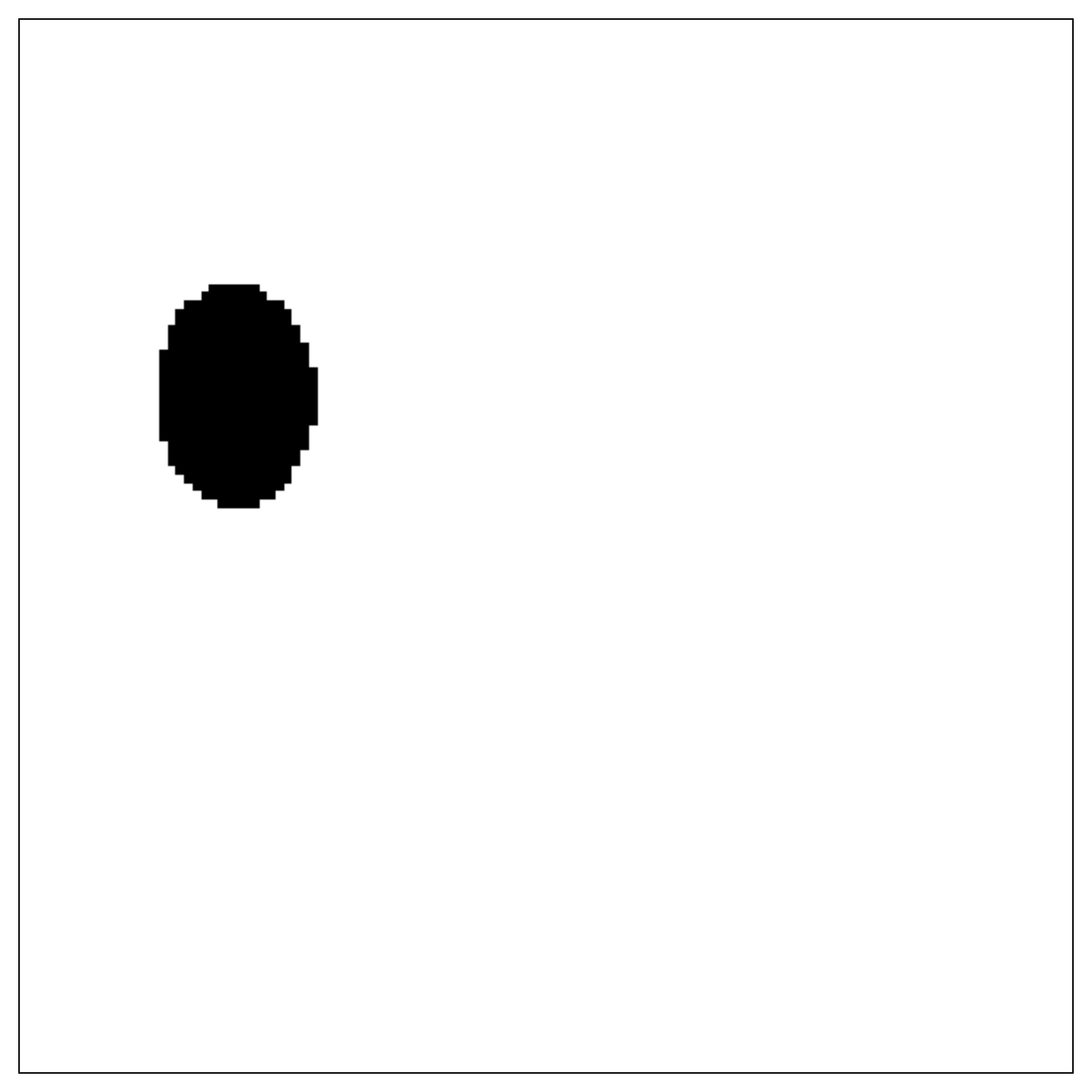} & GT Single Object \\
\midrule
LRP \cite{Bach:PLOS2015}                            & \includegraphics[width=.12\linewidth,valign=m]{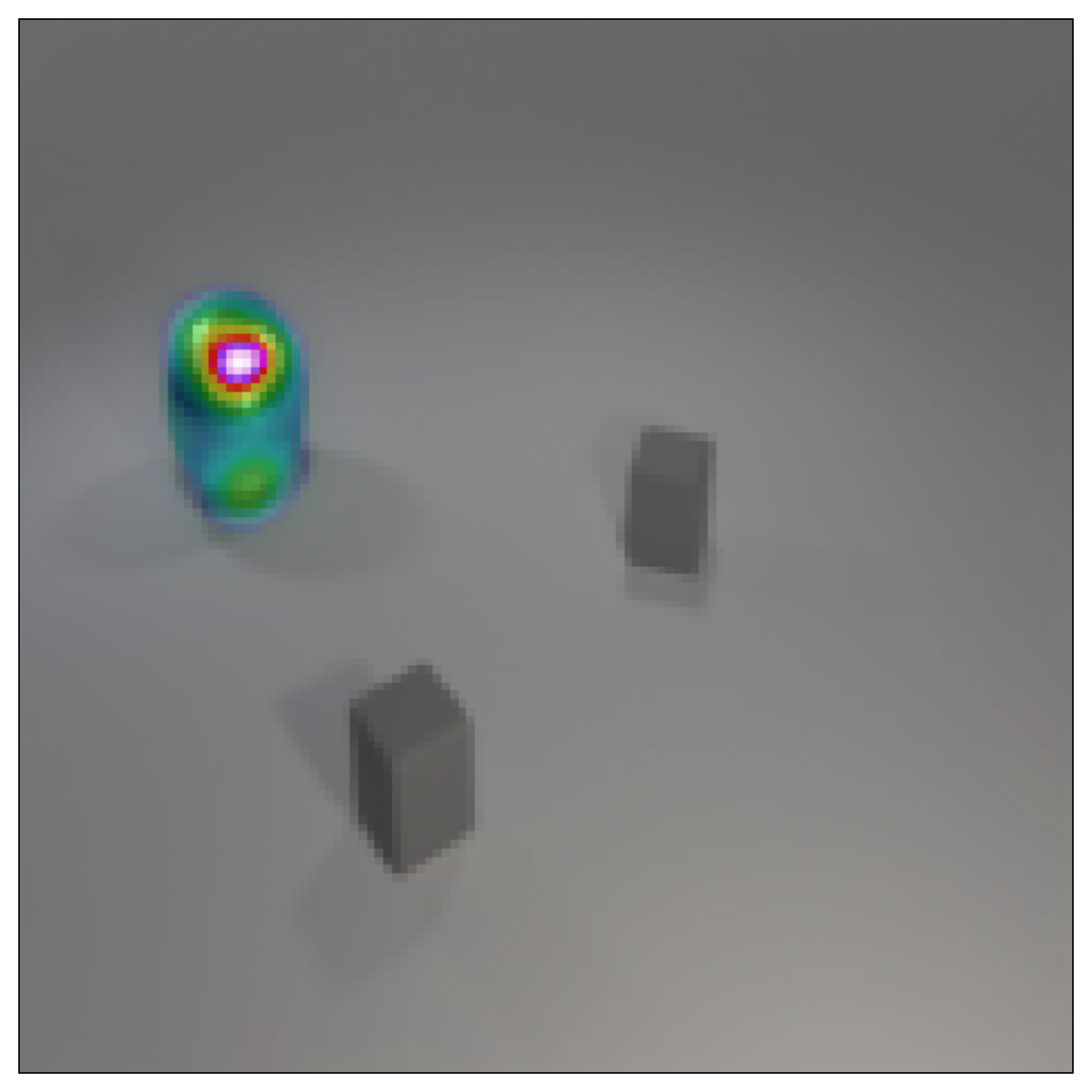} & \includegraphics[width=.12\linewidth,valign=m]{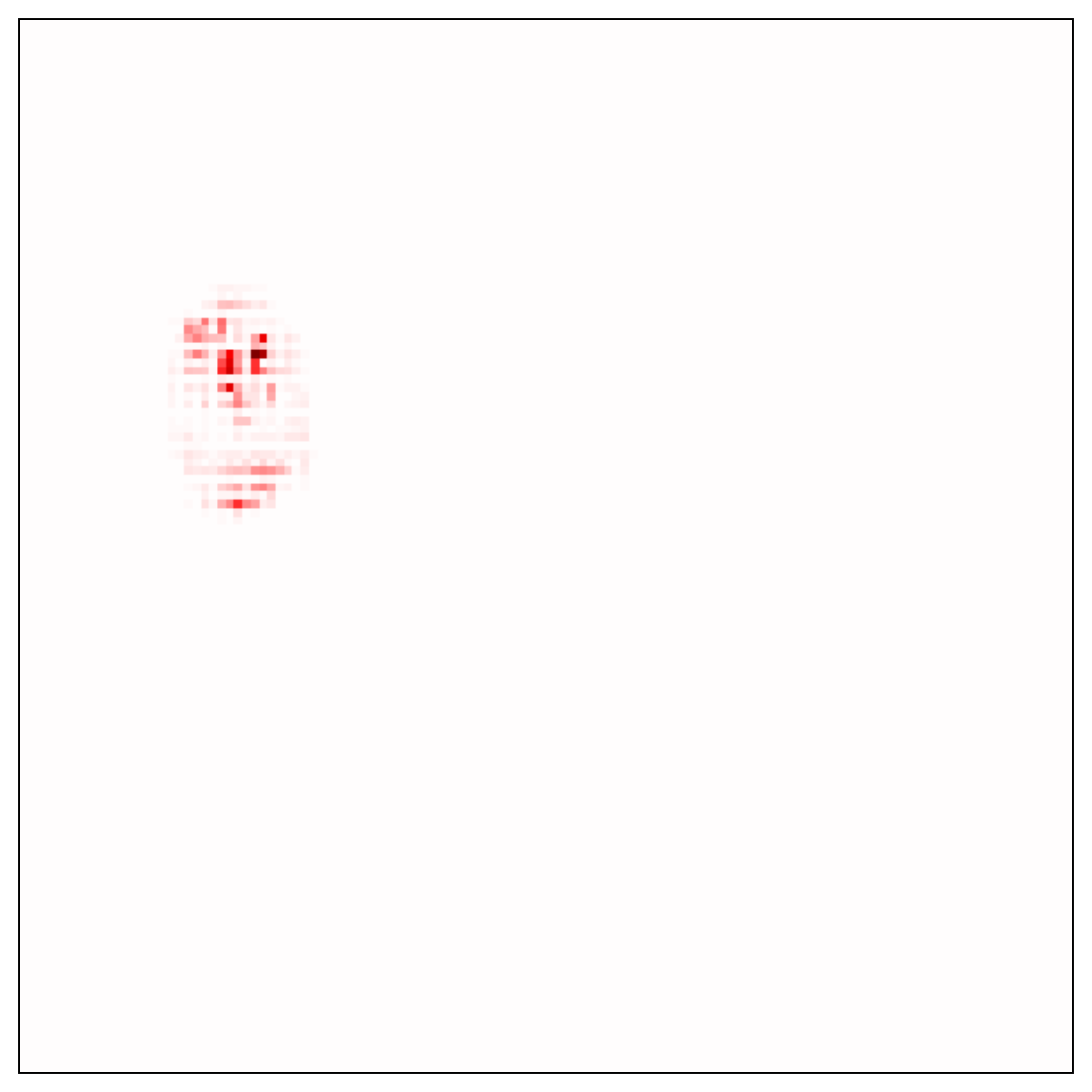} & 0.97 \\
Excitation Backprop \cite{Zhang:ECCV2016}           & \includegraphics[width=.12\linewidth,valign=m]{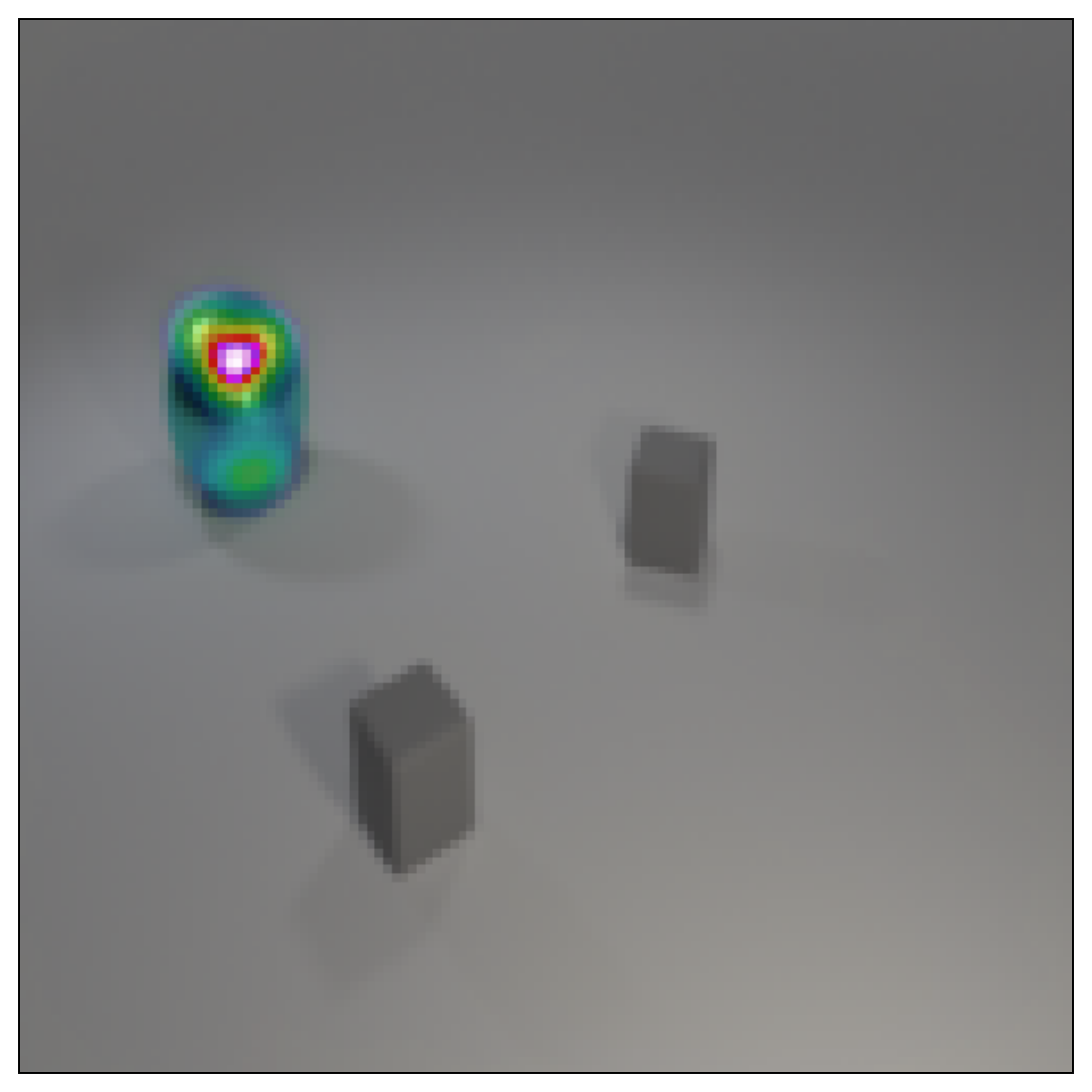} & \includegraphics[width=.12\linewidth,valign=m]{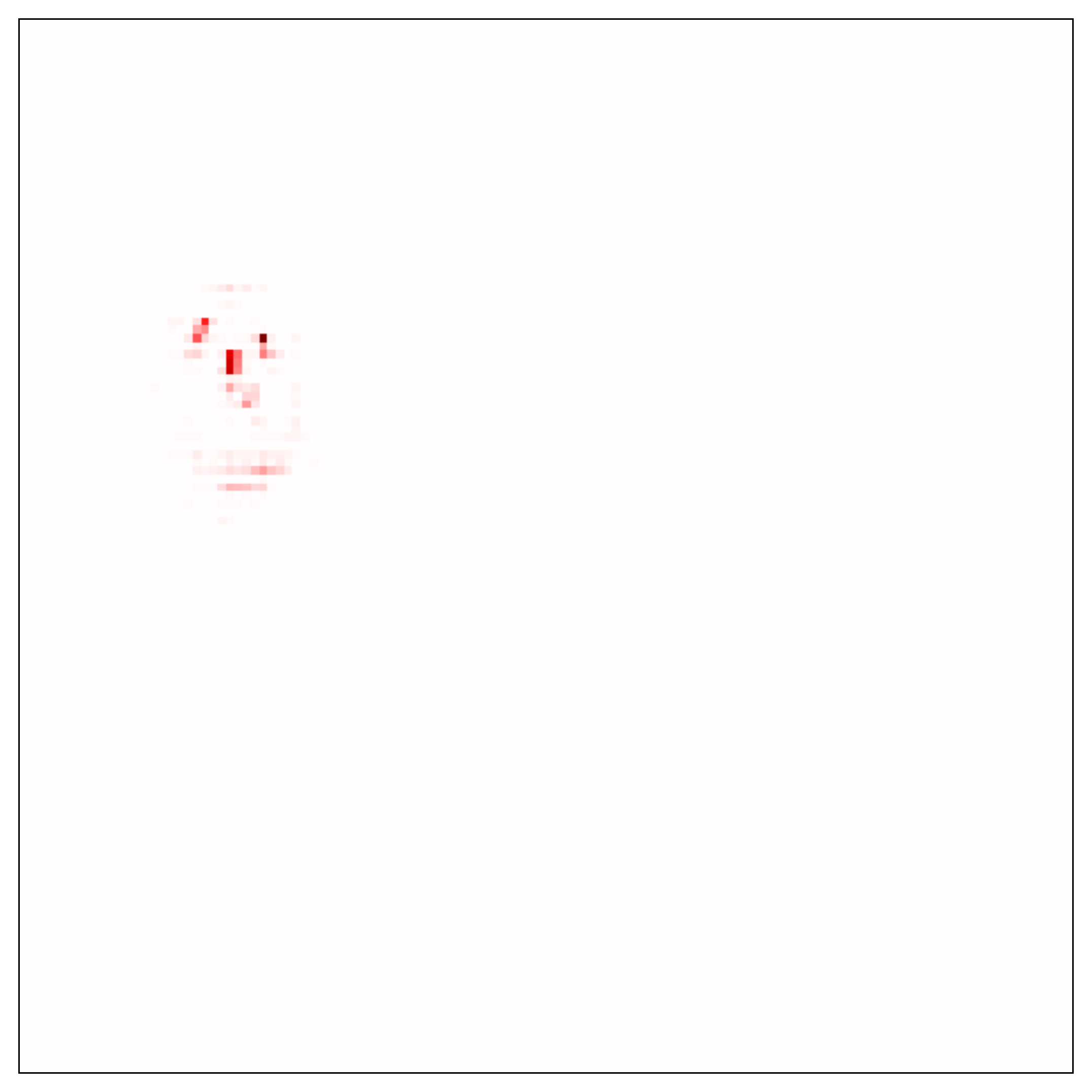} & 0.97 \\
IG \cite{Sundararajan:ICML2017}                     & \includegraphics[width=.12\linewidth,valign=m]{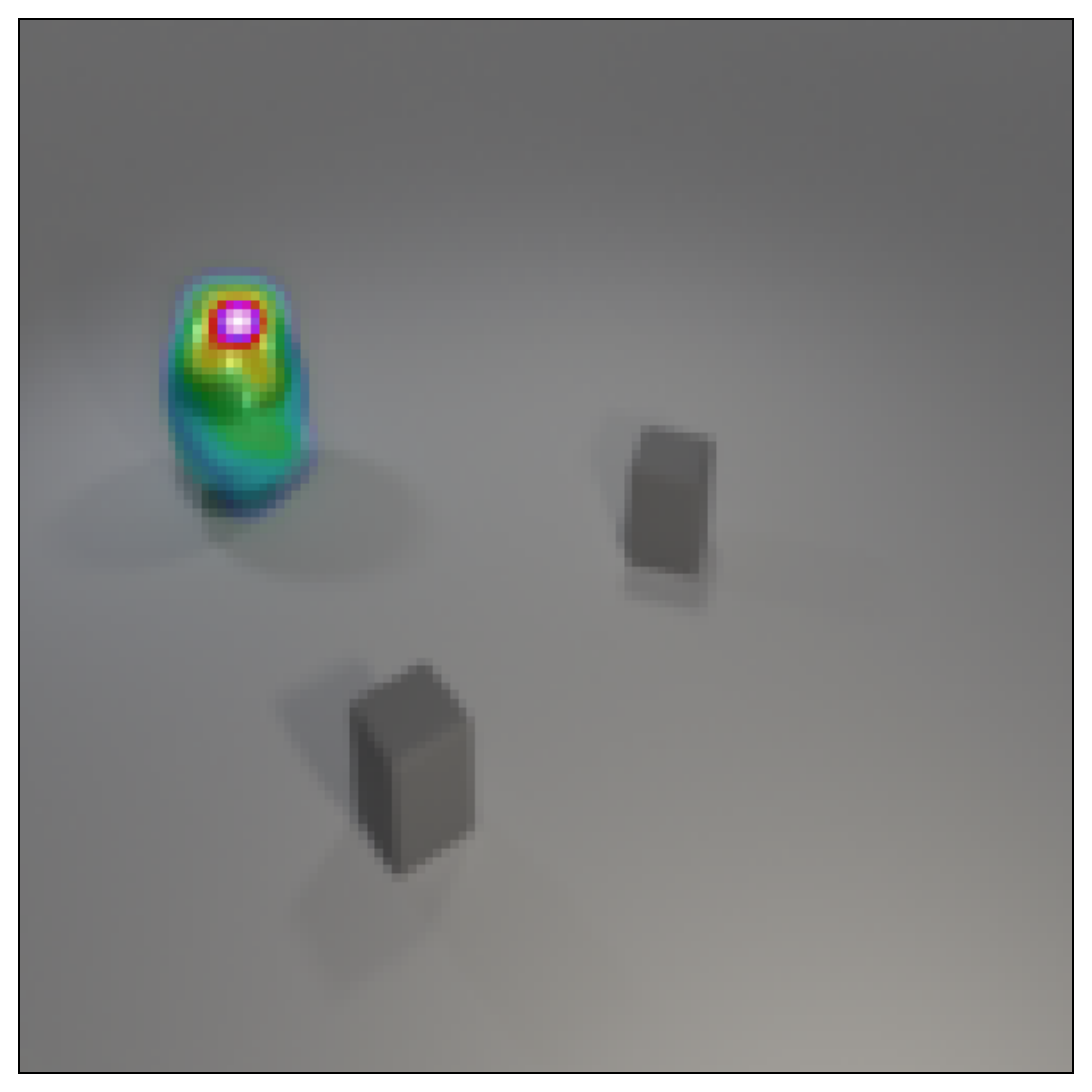} & \includegraphics[width=.12\linewidth,valign=m]{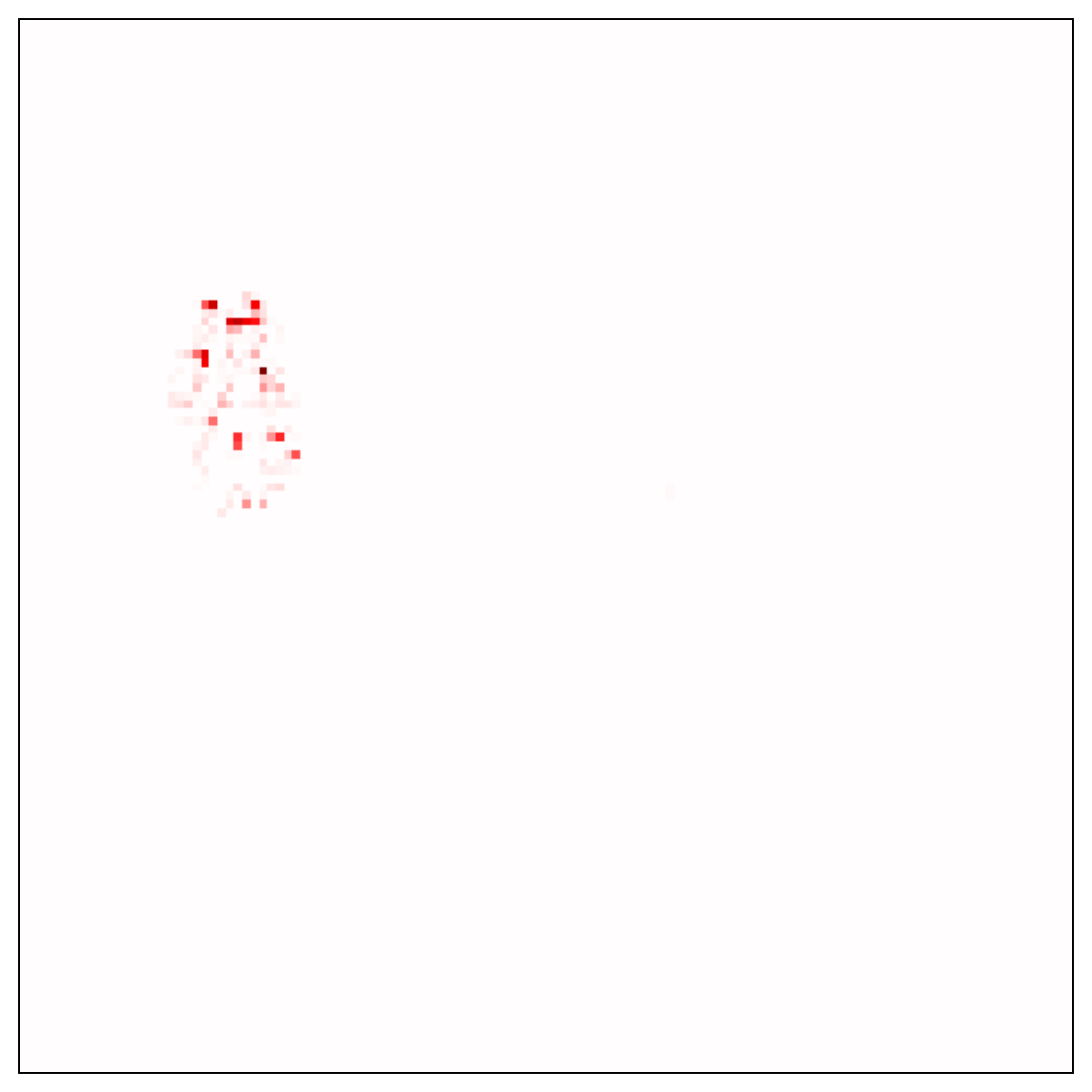} & 0.96 \\
Guided Backprop \cite{Spring:ICLR2015}              & \includegraphics[width=.12\linewidth,valign=m]{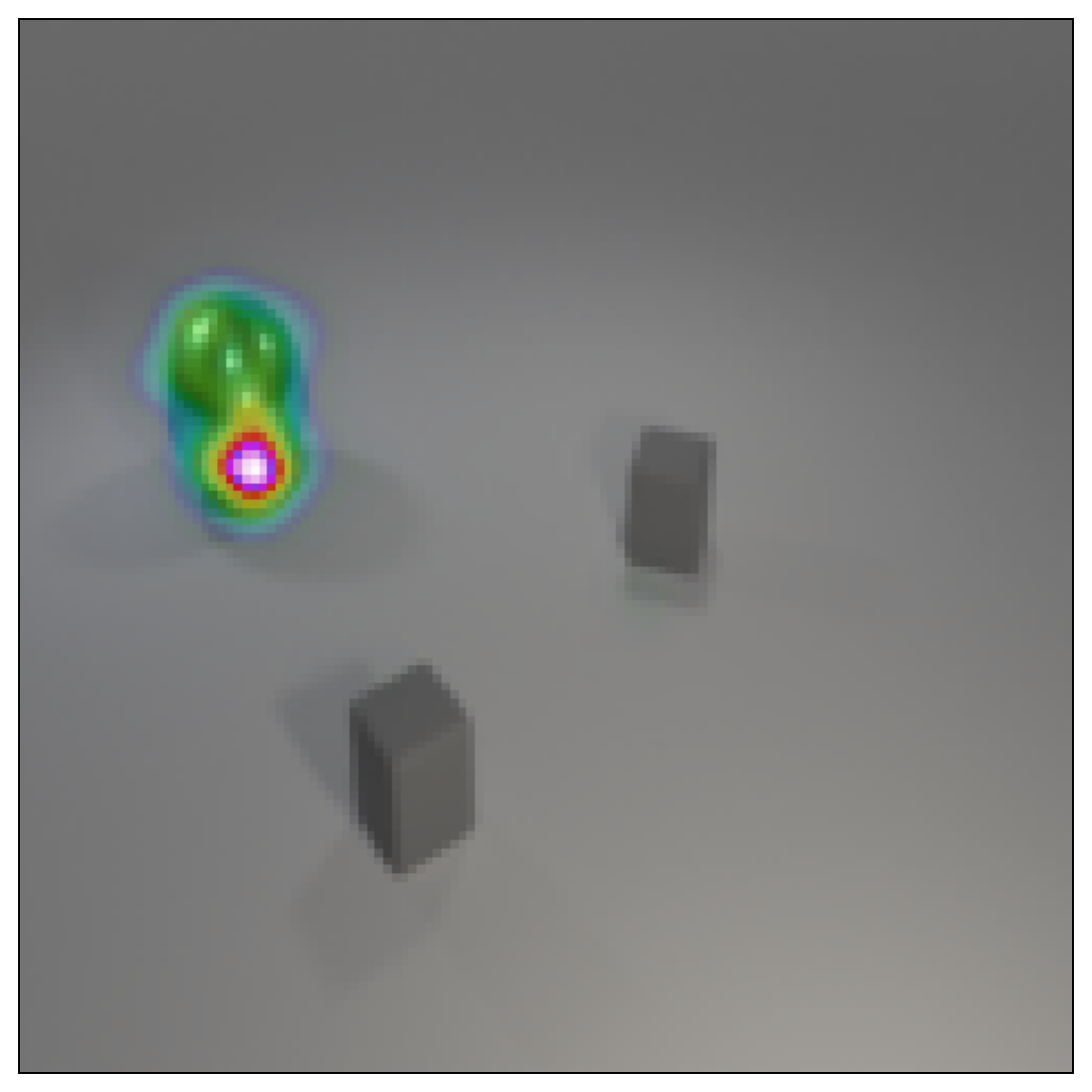} & \includegraphics[width=.12\linewidth,valign=m]{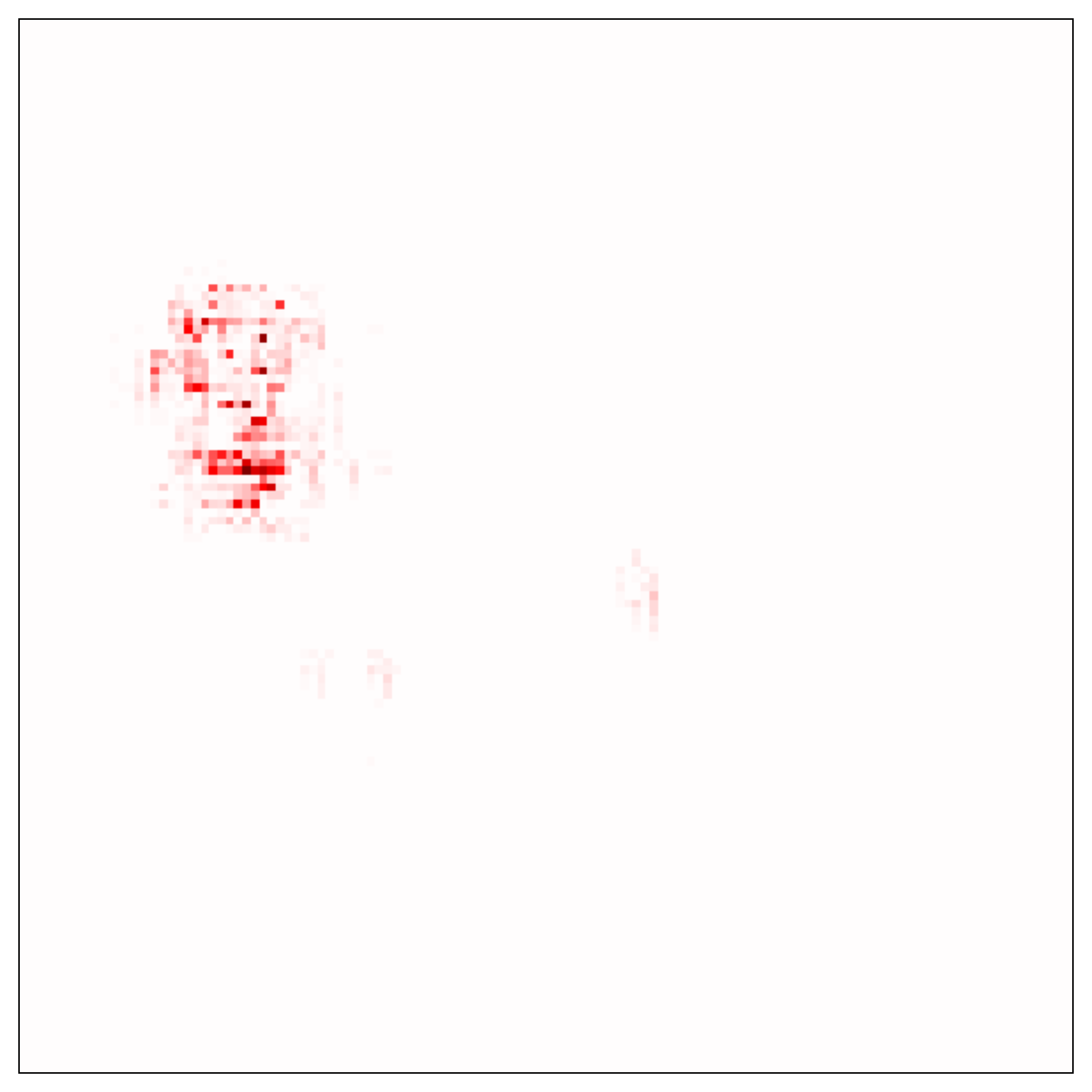} & 0.79 \\
Guided Grad-CAM \cite{Selvaraju:ICCV2017}           & \includegraphics[width=.12\linewidth,valign=m]{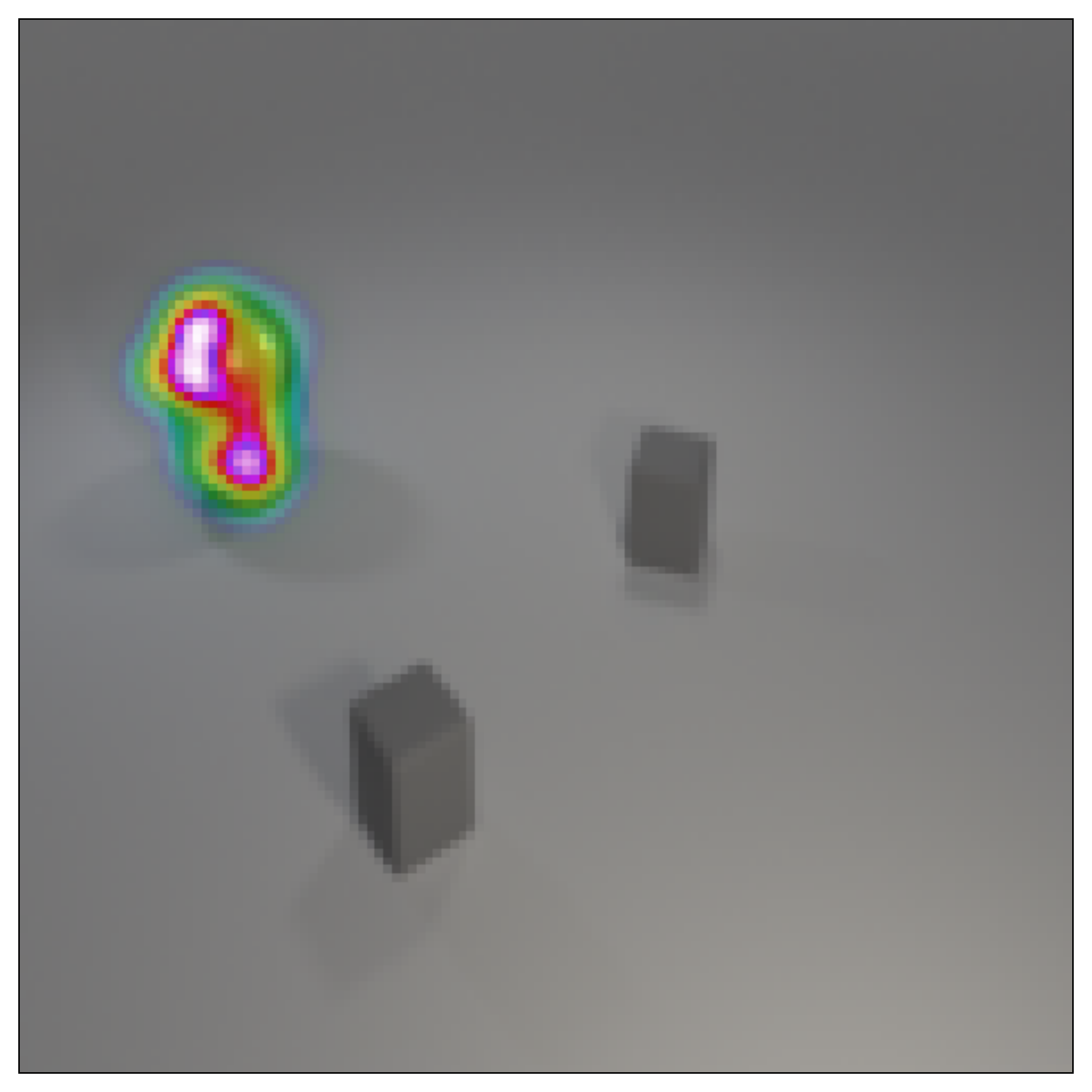} & \includegraphics[width=.12\linewidth,valign=m]{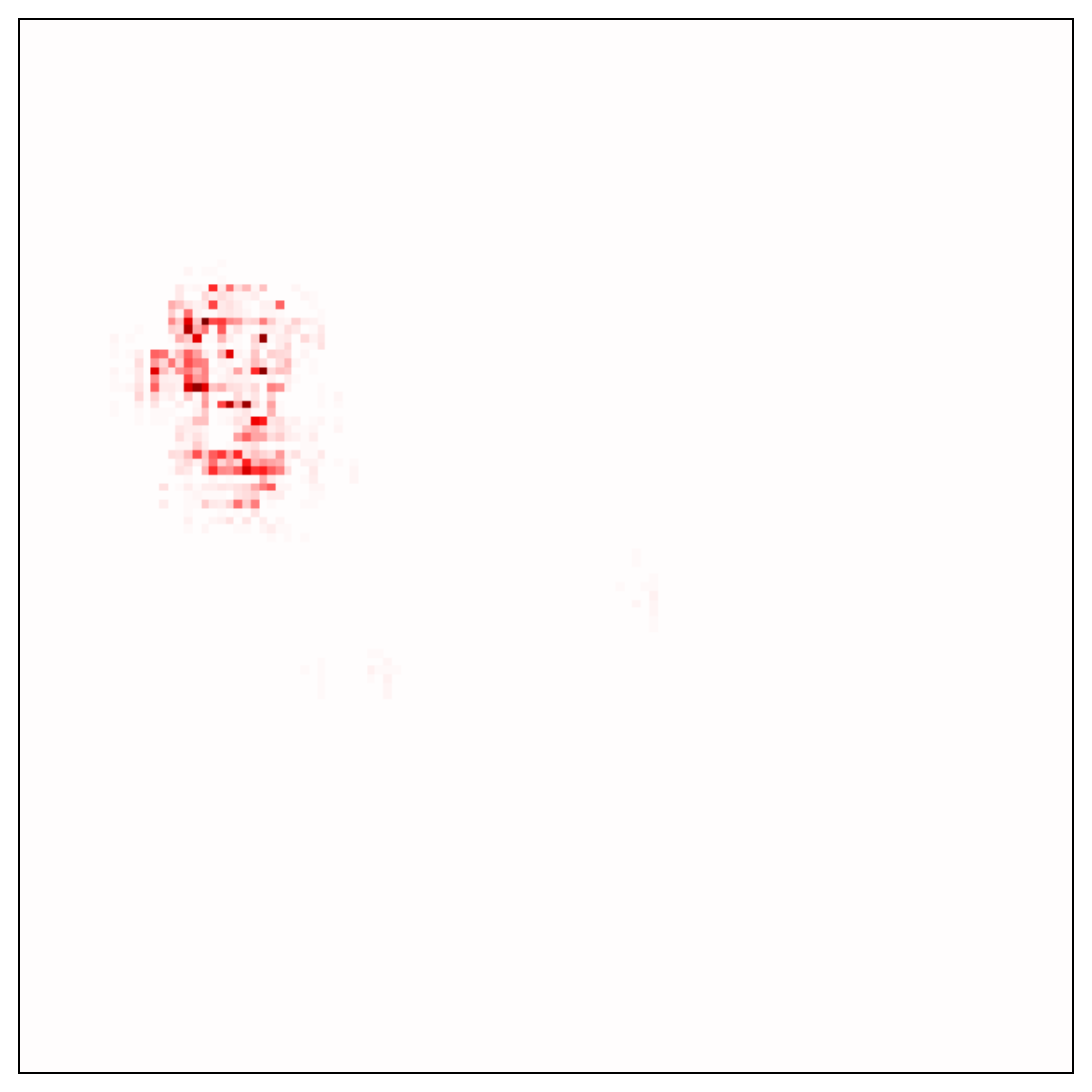} & 0.85 \\
SmoothGrad \cite{Smilkov:ICML2017}                  & \includegraphics[width=.12\linewidth,valign=m]{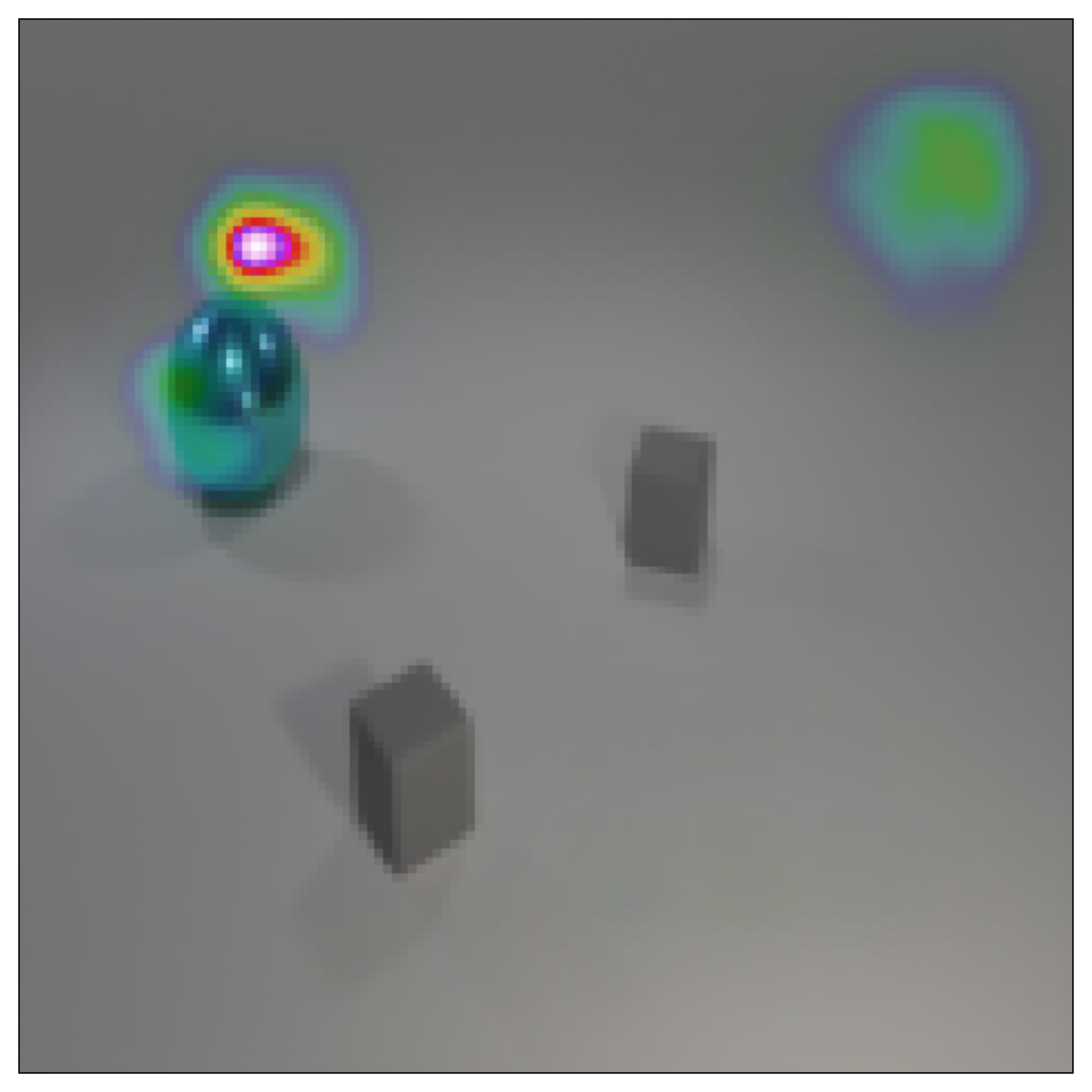} & \includegraphics[width=.12\linewidth,valign=m]{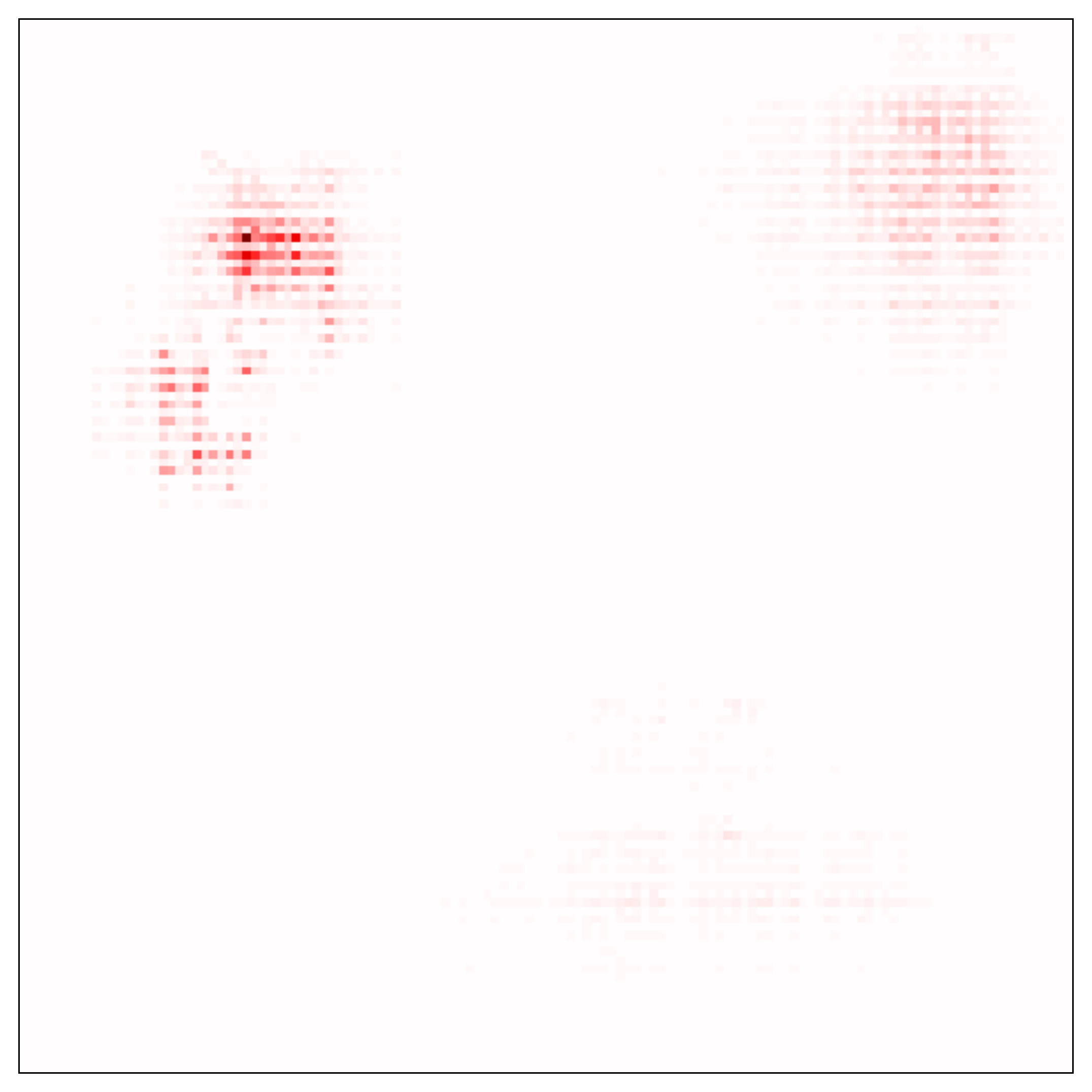} & 0.14 \\
VarGrad \cite{Adebayo:ICLR2018}                     & \includegraphics[width=.12\linewidth,valign=m]{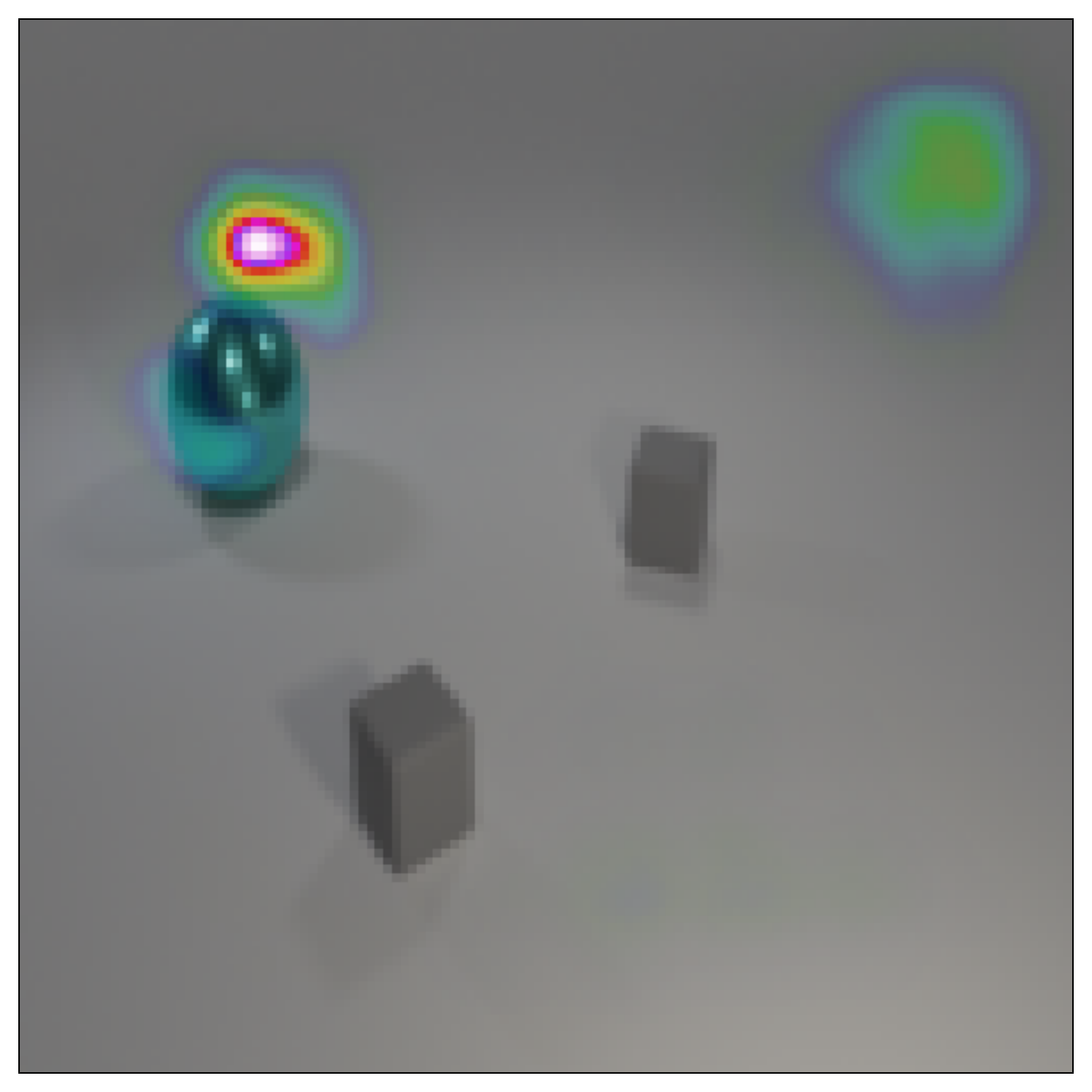} & \includegraphics[width=.12\linewidth,valign=m]{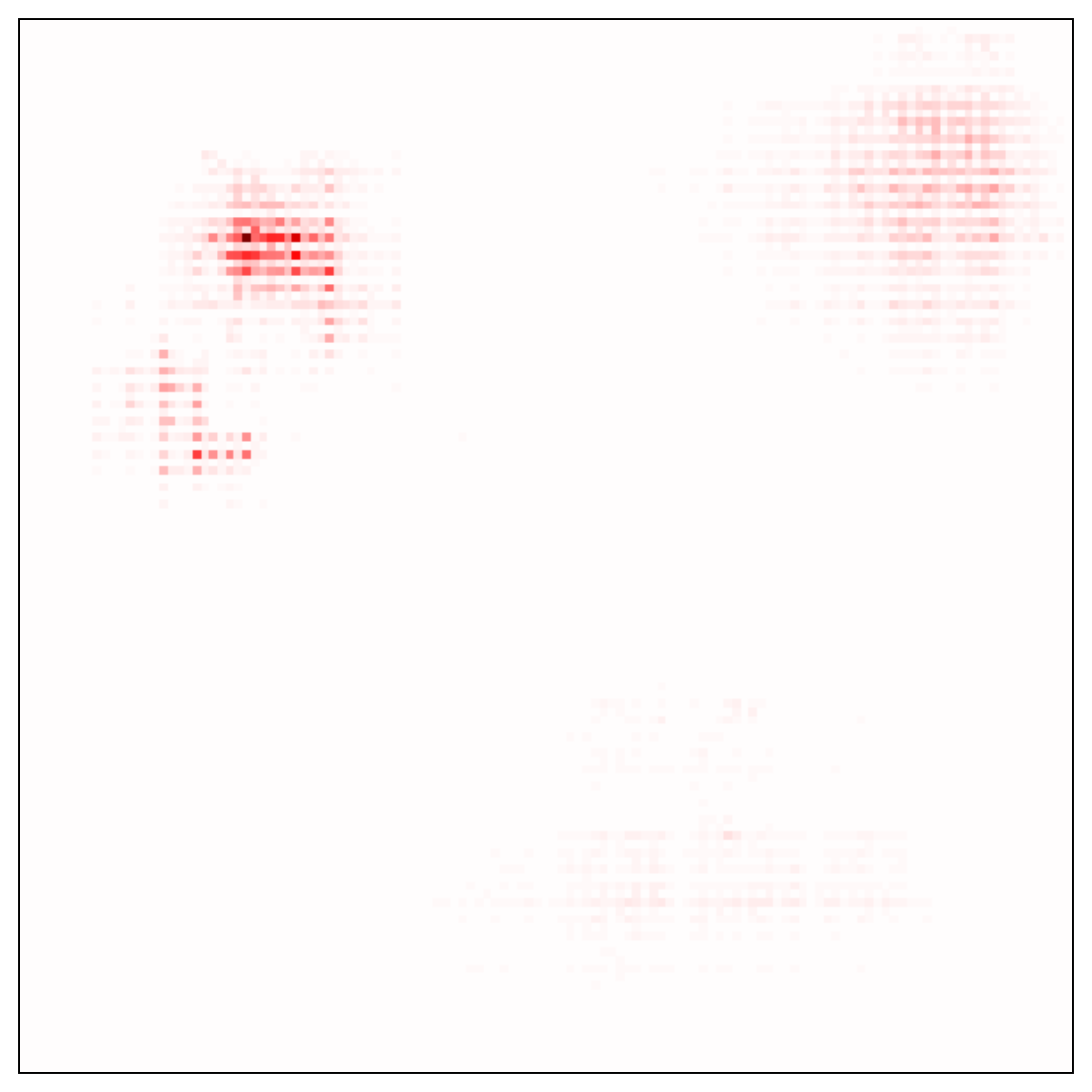} & 0.09 \\
Gradient \cite{Simonyan:ICLR2014}                   & \includegraphics[width=.12\linewidth,valign=m]{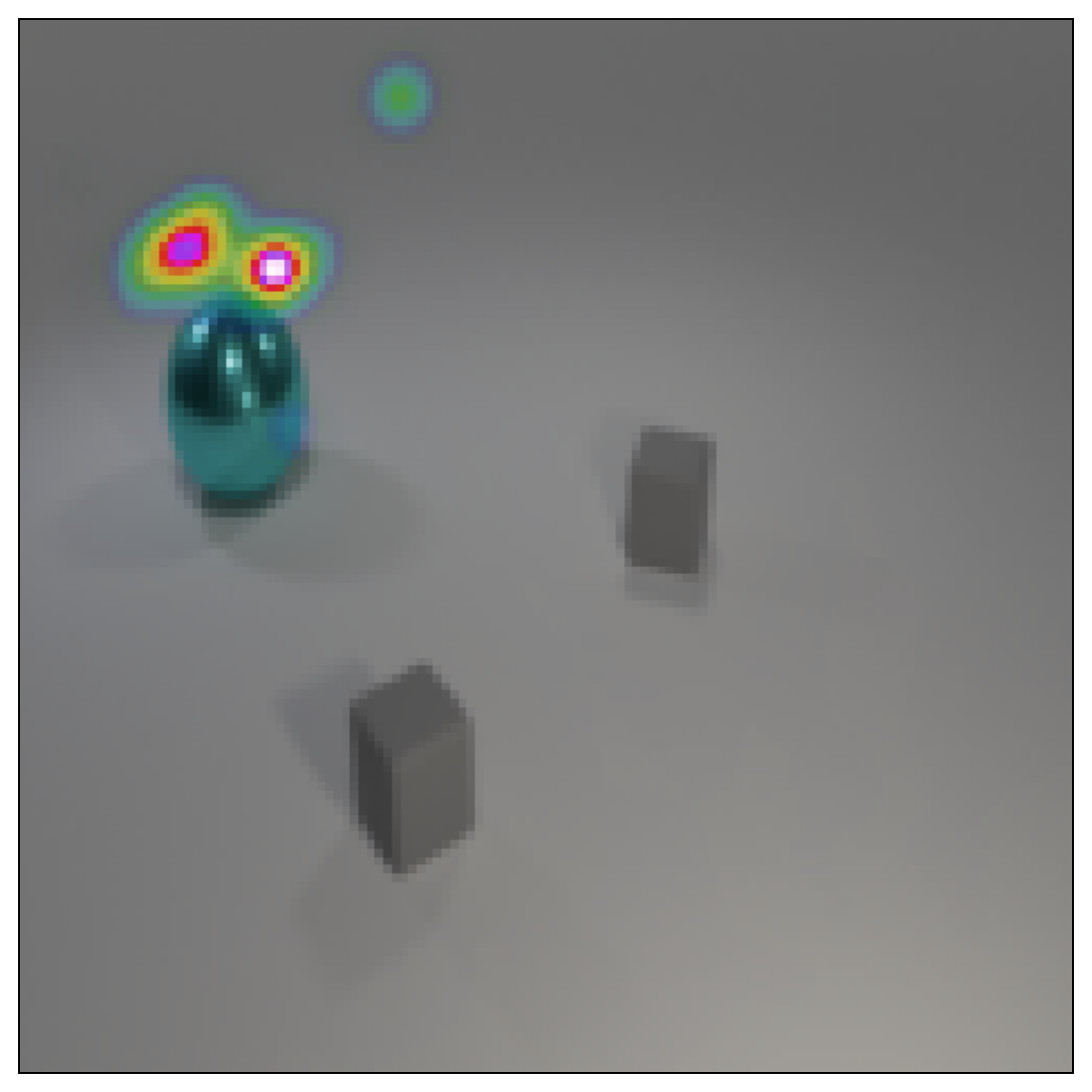} & \includegraphics[width=.12\linewidth,valign=m]{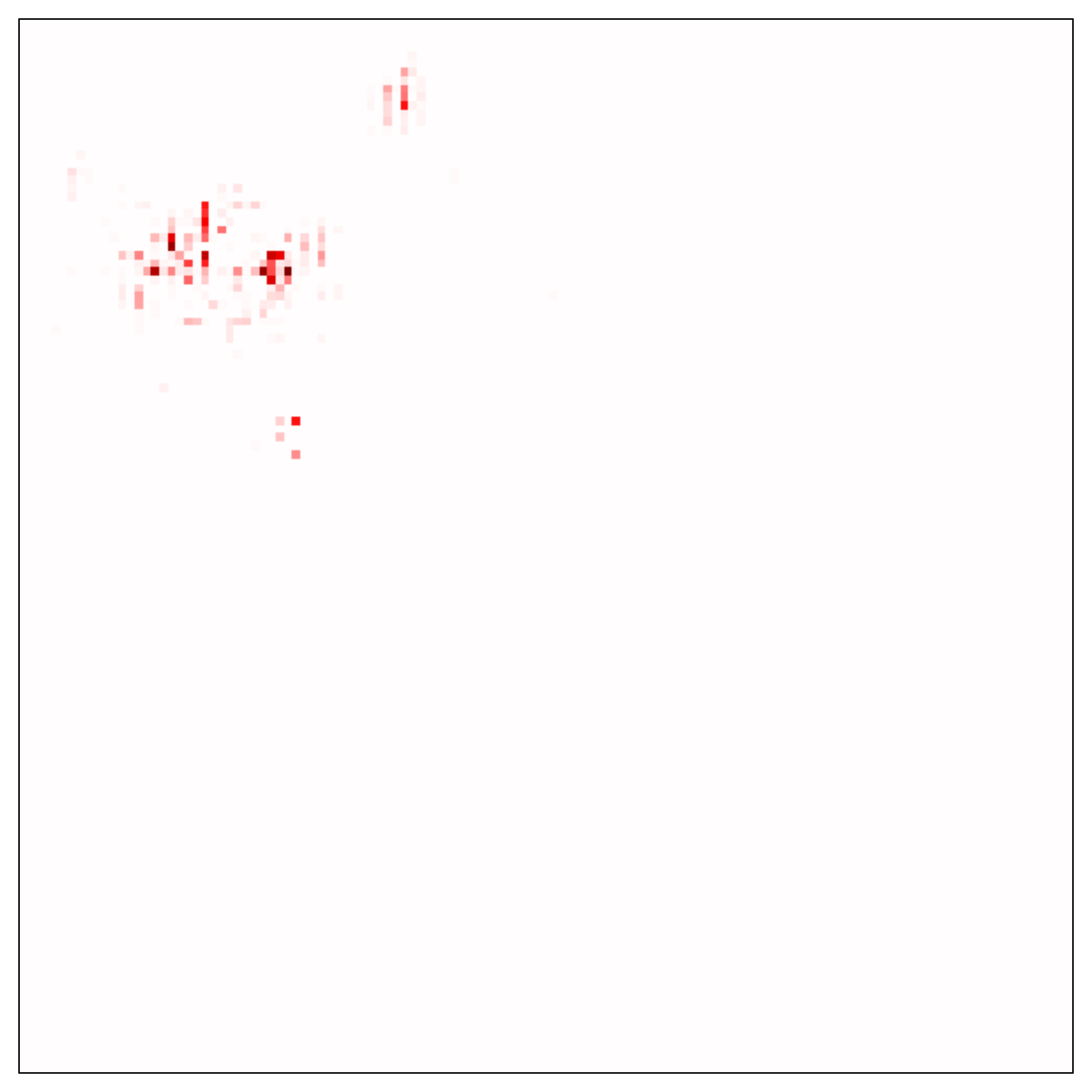} & 0.10 \\
Gradient$\times$Input \cite{Shrikumar:arxiv2016}    & \includegraphics[width=.12\linewidth,valign=m]{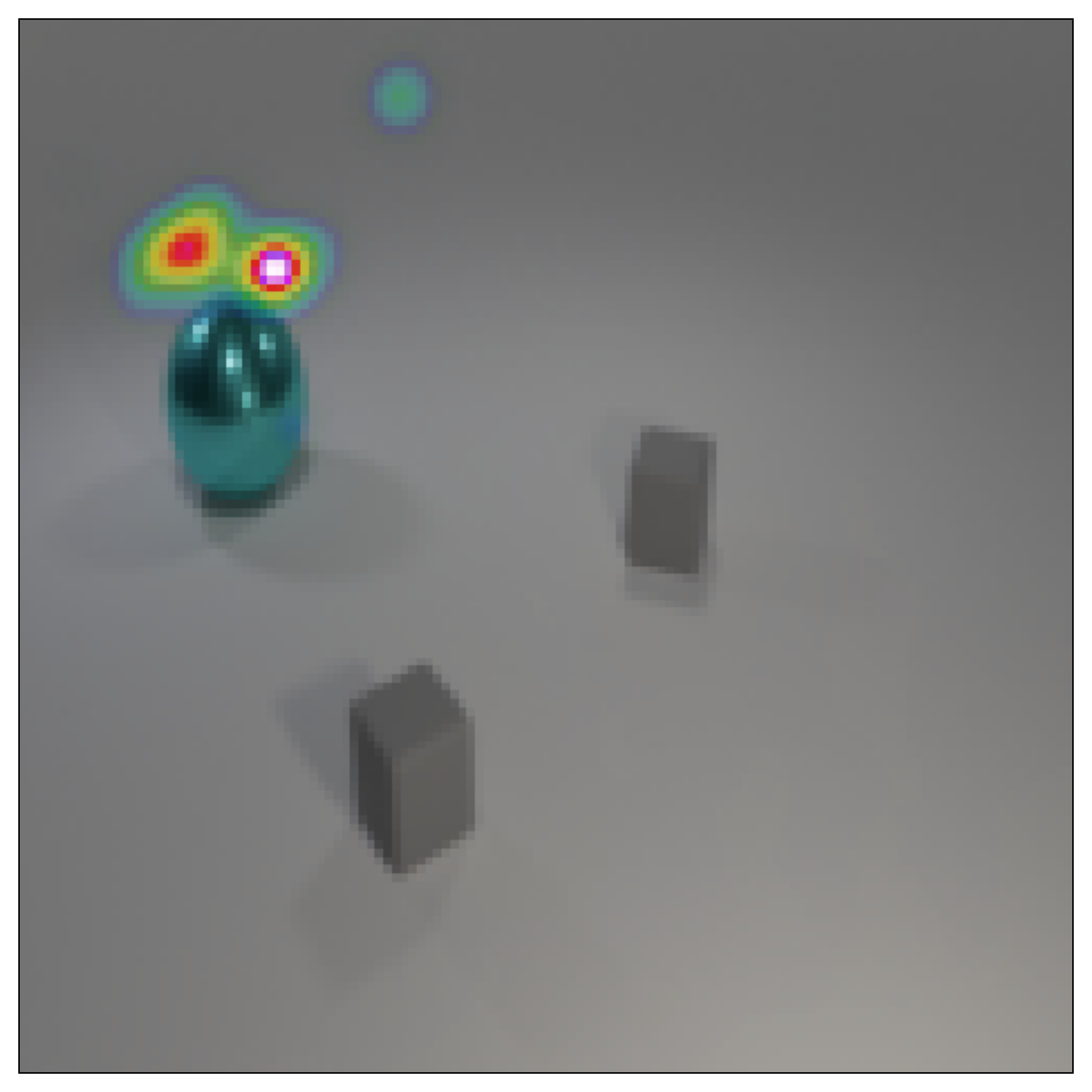} & \includegraphics[width=.12\linewidth,valign=m]{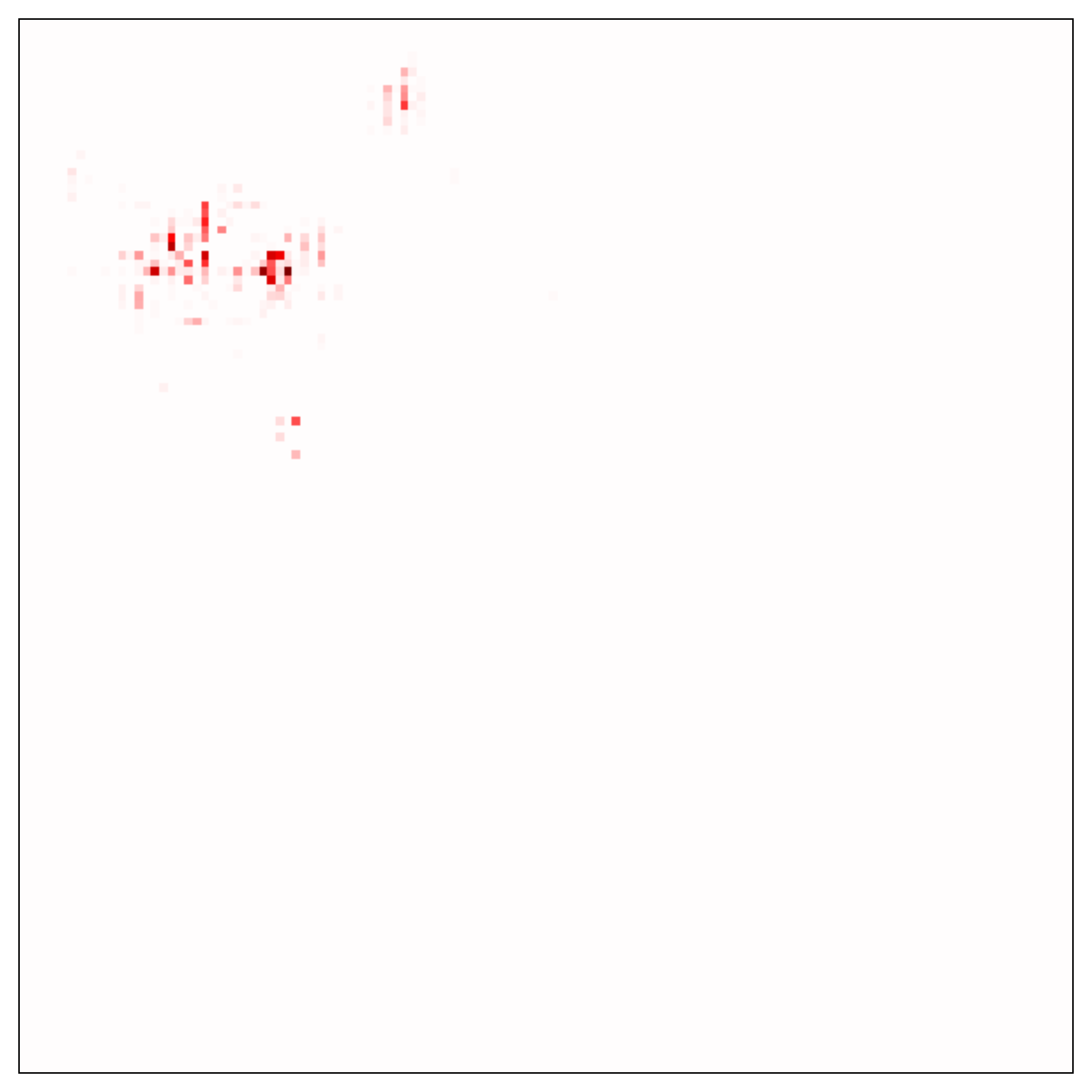} & 0.07 \\
Deconvnet \cite{Zeiler:ECCV2014}                    & \includegraphics[width=.12\linewidth,valign=m]{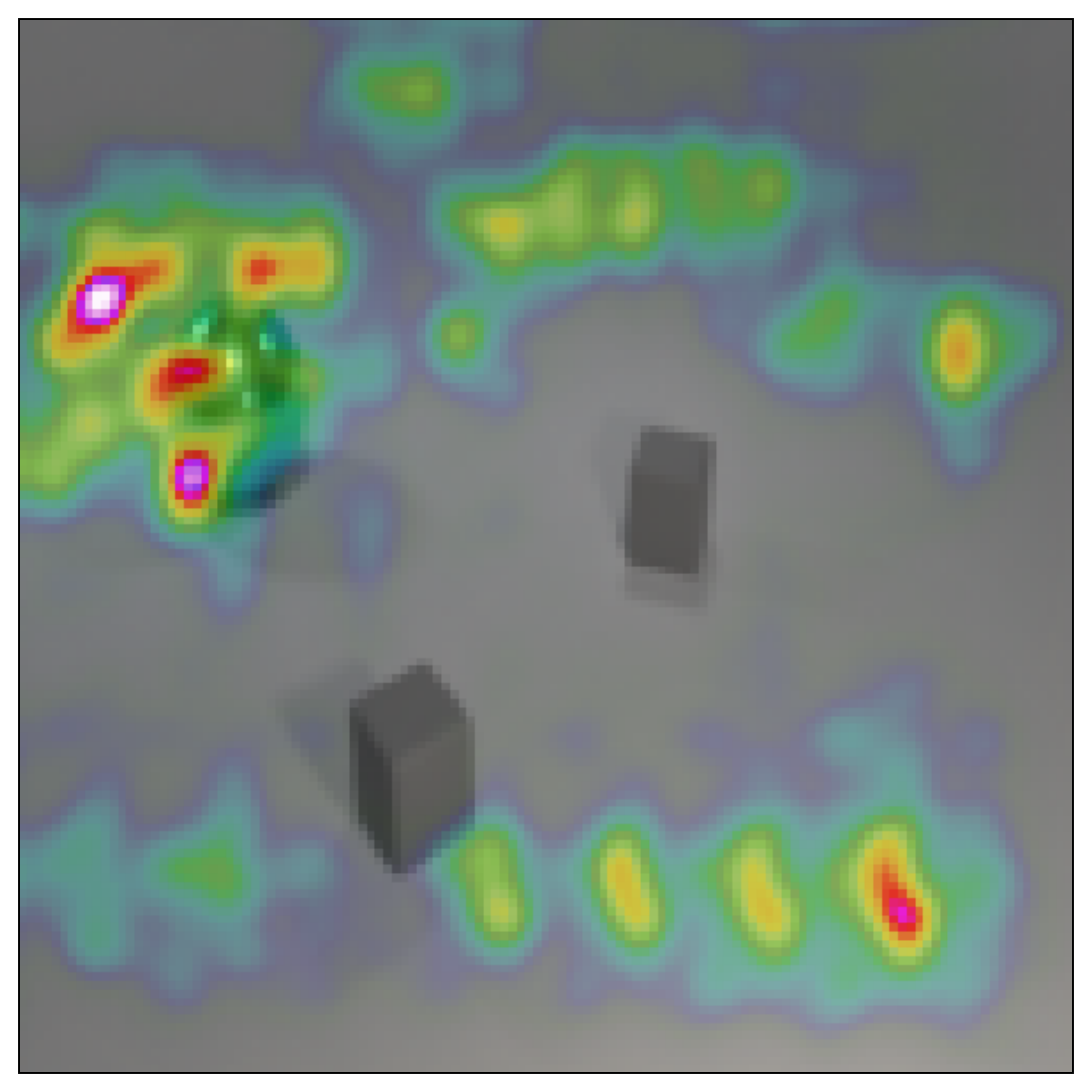} & \includegraphics[width=.12\linewidth,valign=m]{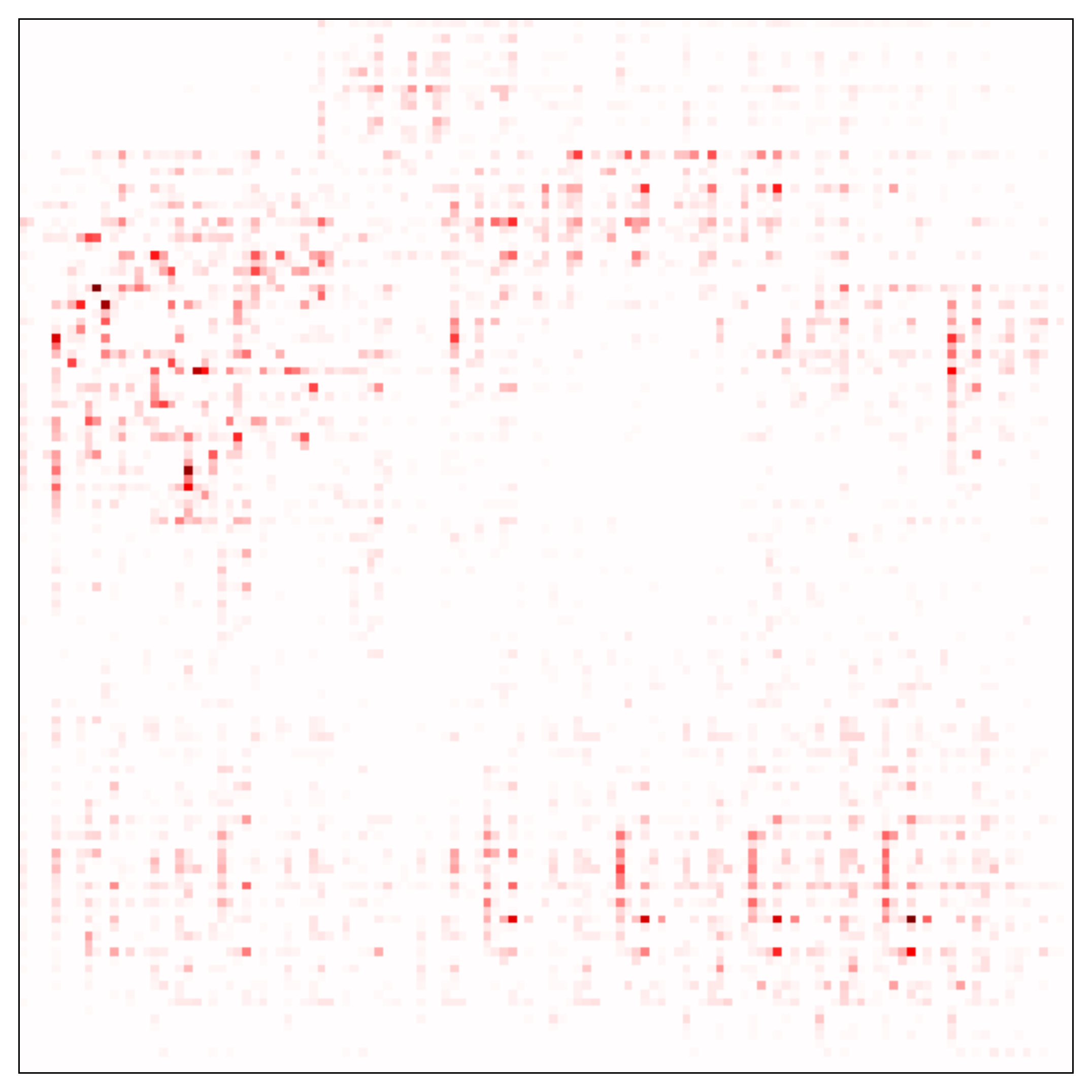} & 0.08 \\
Grad-CAM \cite{Selvaraju:ICCV2017}                  & \includegraphics[width=.12\linewidth,valign=m]{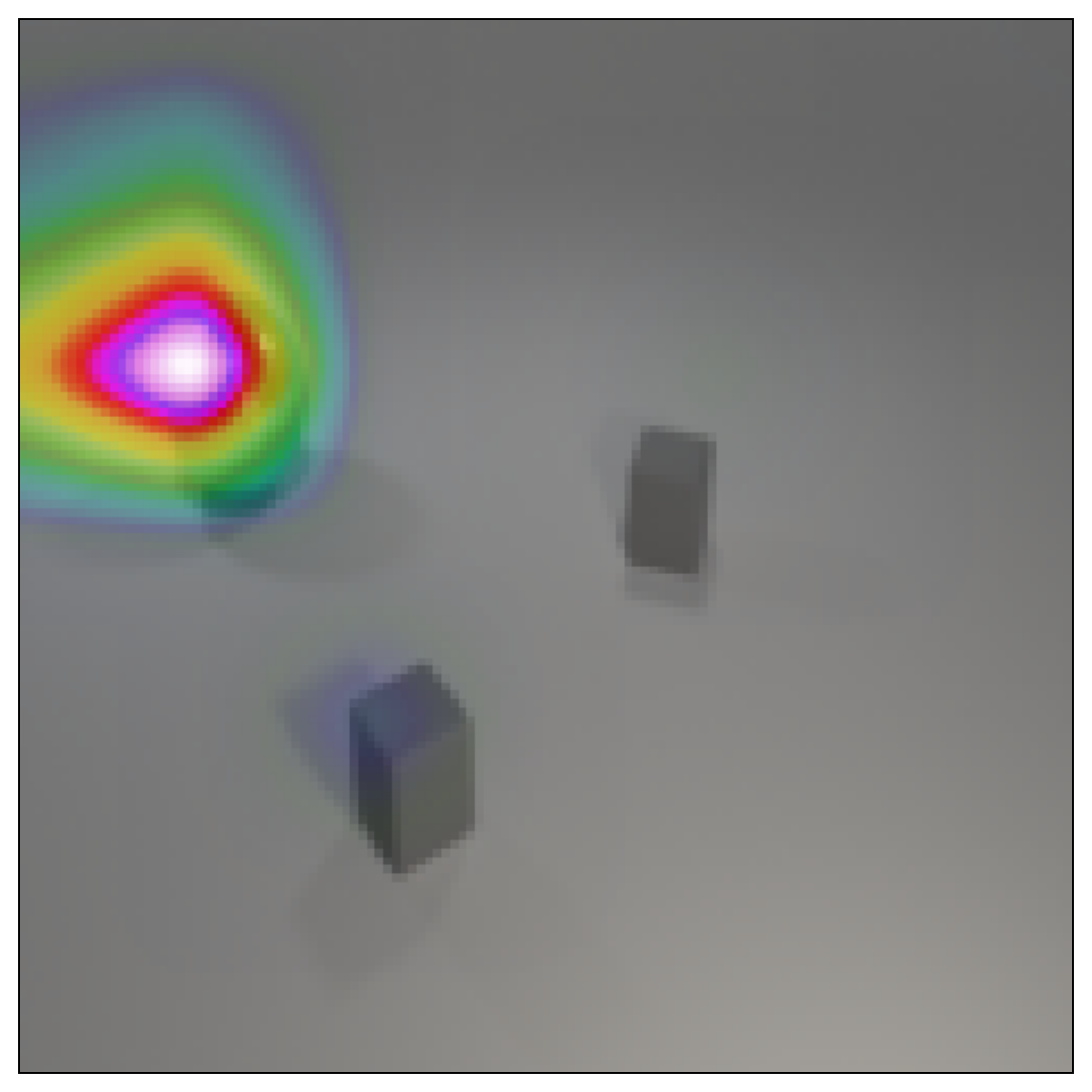} & \includegraphics[width=.12\linewidth,valign=m]{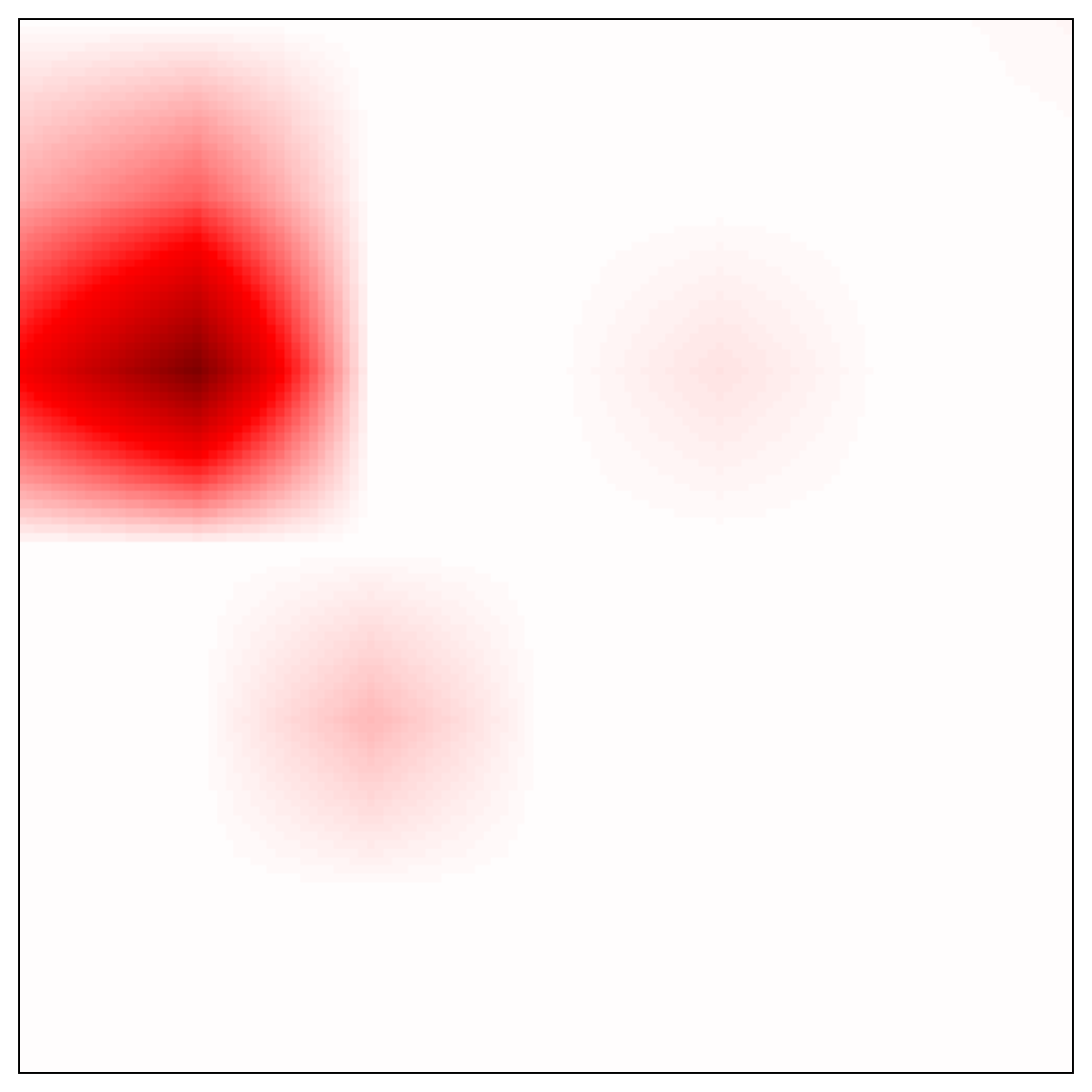} & 0.27 \\
\end{tabular}
\end{table}

\begin{table}
        \scriptsize
		\caption{Heatmaps for a correctly predicted CLEVR-XAI-simple question (raw heatmap and heatmap overlayed with original image), and corresponding relevance \textit{mass} accuracy.}
		\label{table:heatmap-simple-correct-17094}
\begin{tabular}{lllc}
\midrule
\begin{tabular}{@{}l@{}}What is the material \\ of the large block? \\ \textit{metal} \end{tabular}  & \includegraphics[width=.18\linewidth,valign=m]{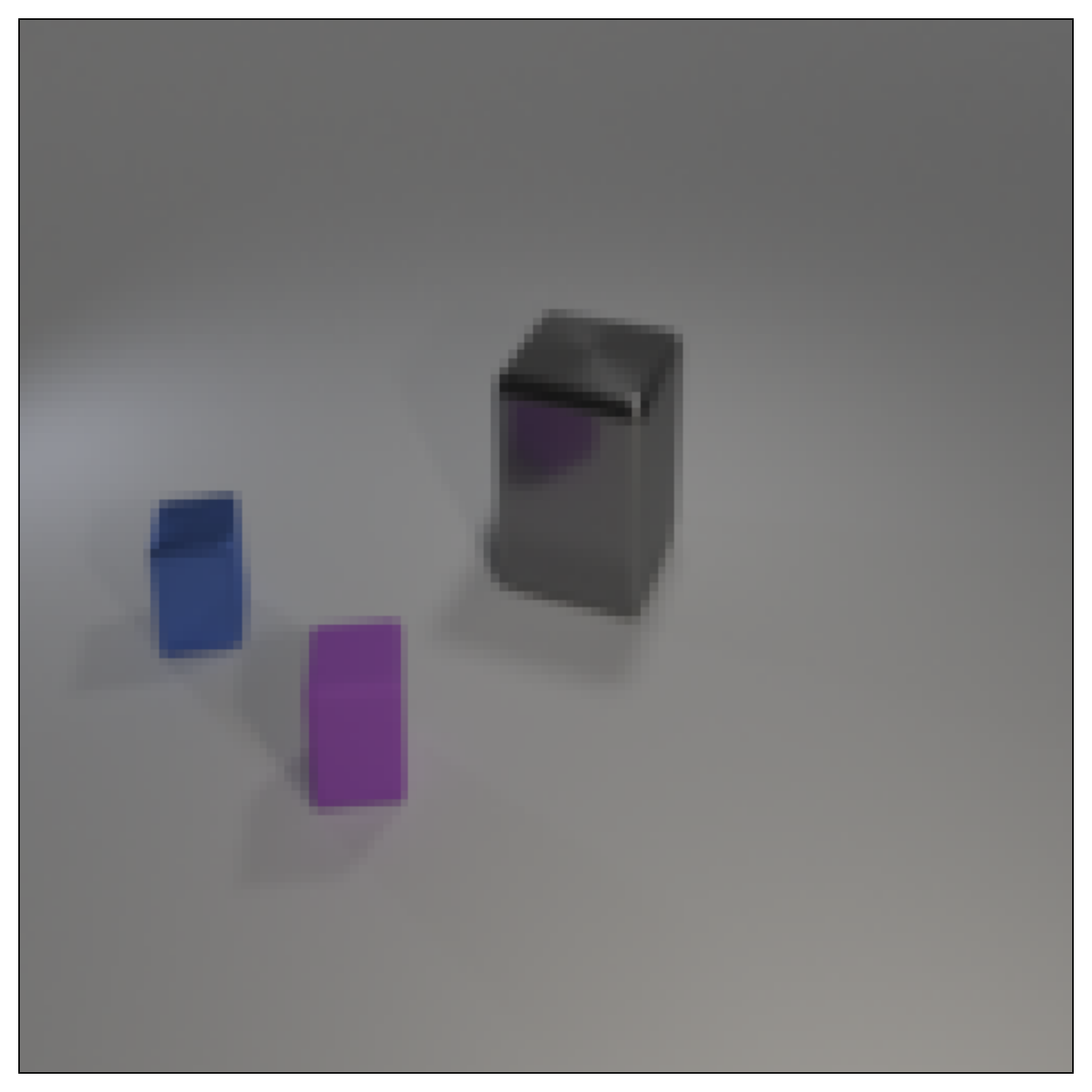} &
\includegraphics[width=.18\linewidth,valign=m]{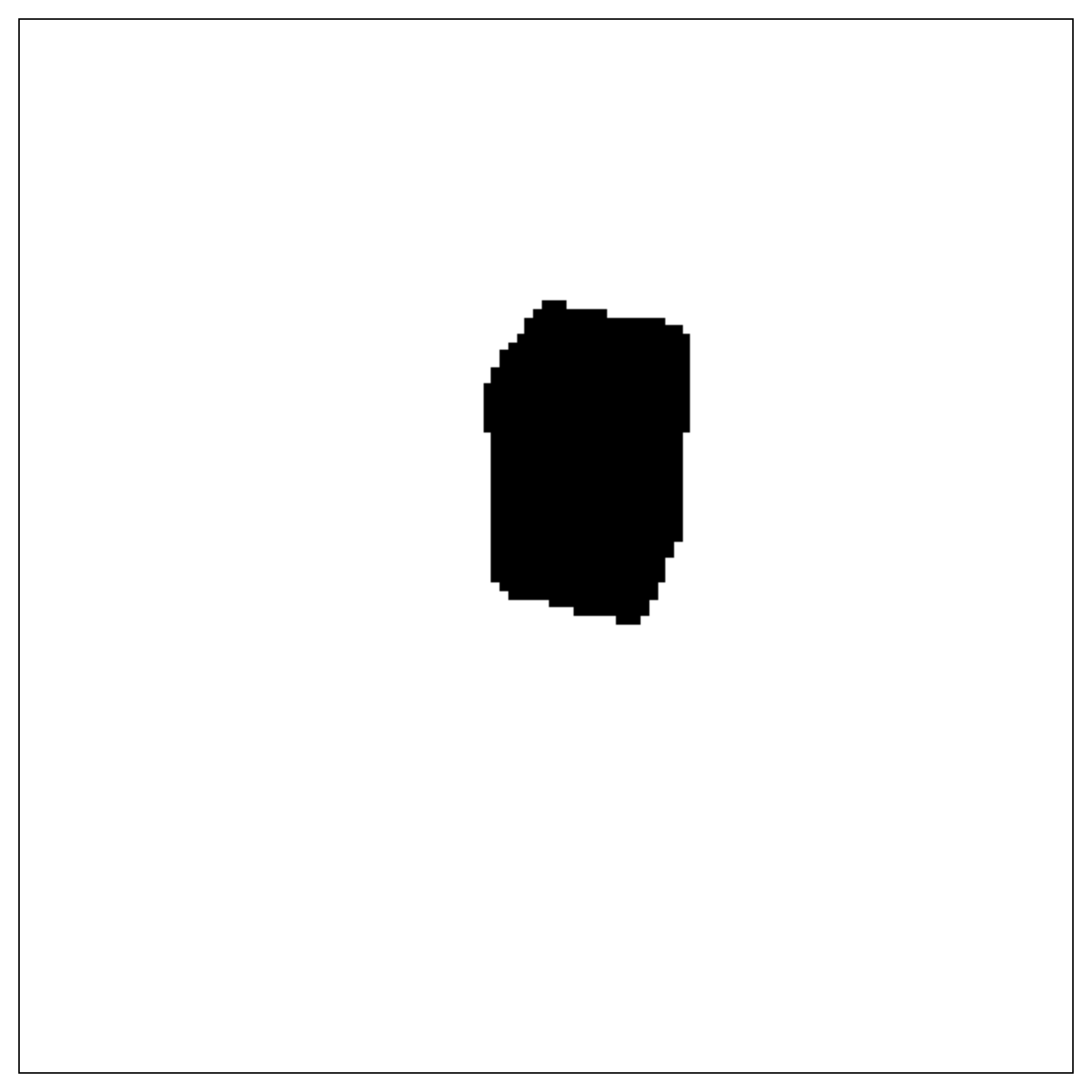} & GT Single Object \\
\midrule
LRP \cite{Bach:PLOS2015}                            & \includegraphics[width=.12\linewidth,valign=m]{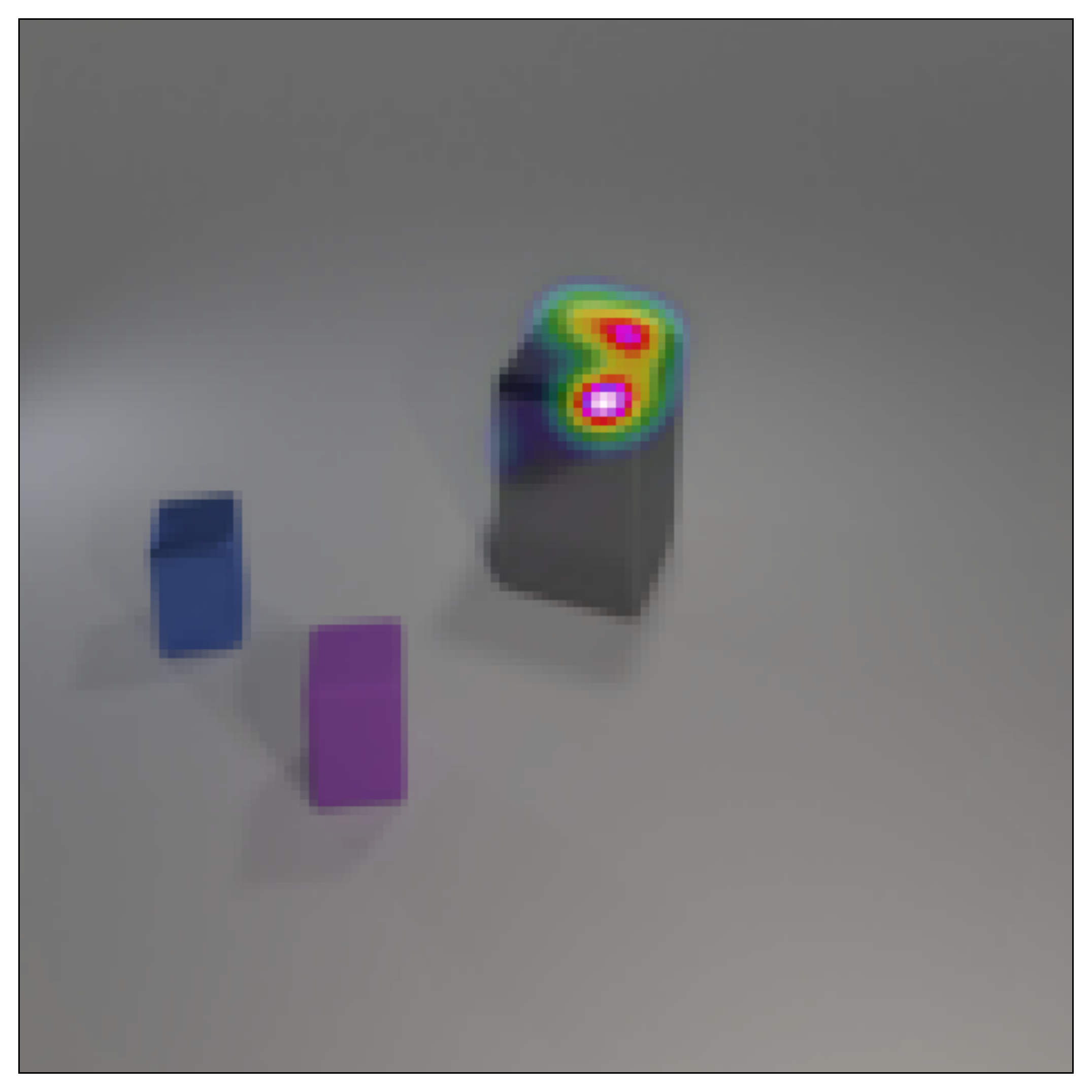} & \includegraphics[width=.12\linewidth,valign=m]{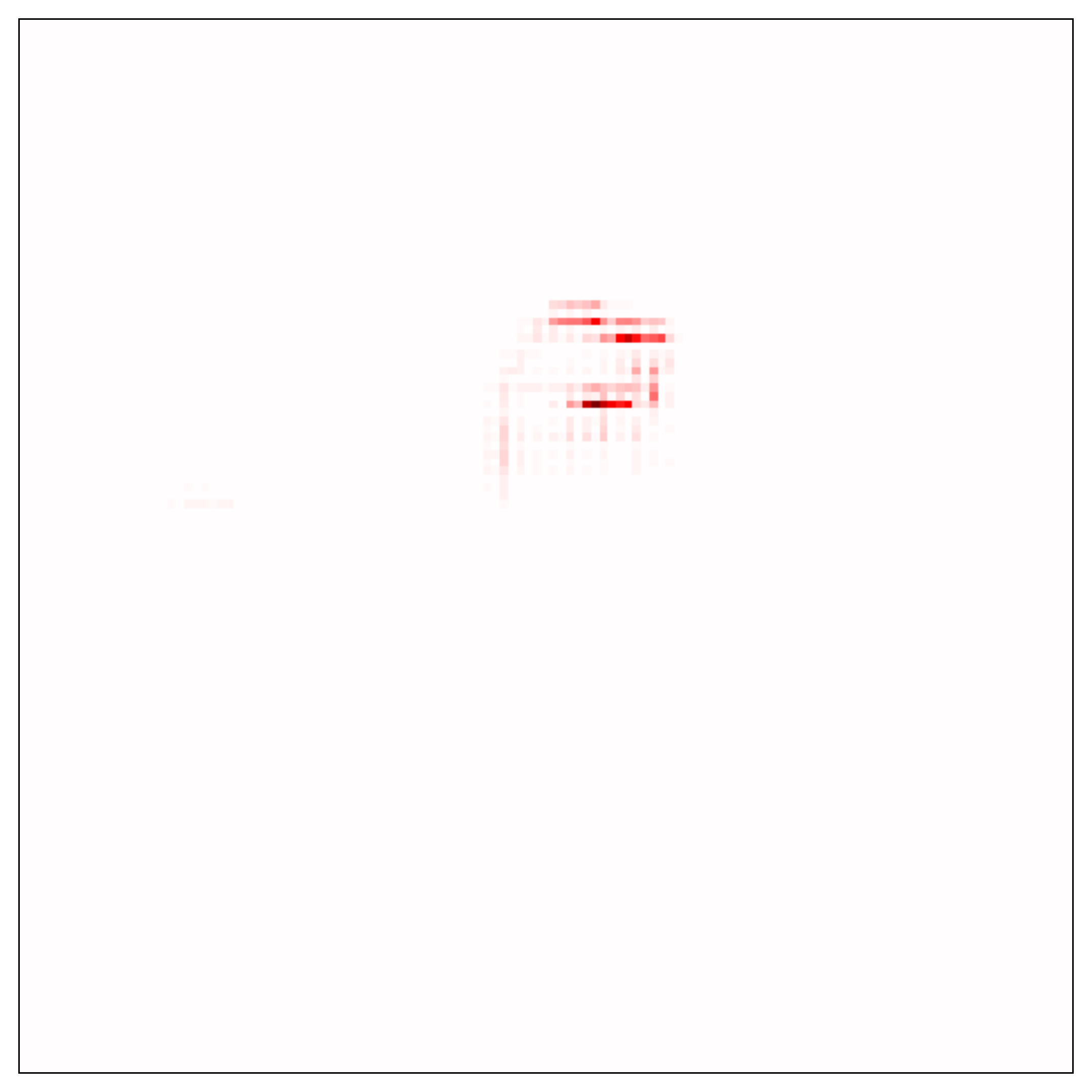} & 0.94 \\
Excitation Backprop \cite{Zhang:ECCV2016}           & \includegraphics[width=.12\linewidth,valign=m]{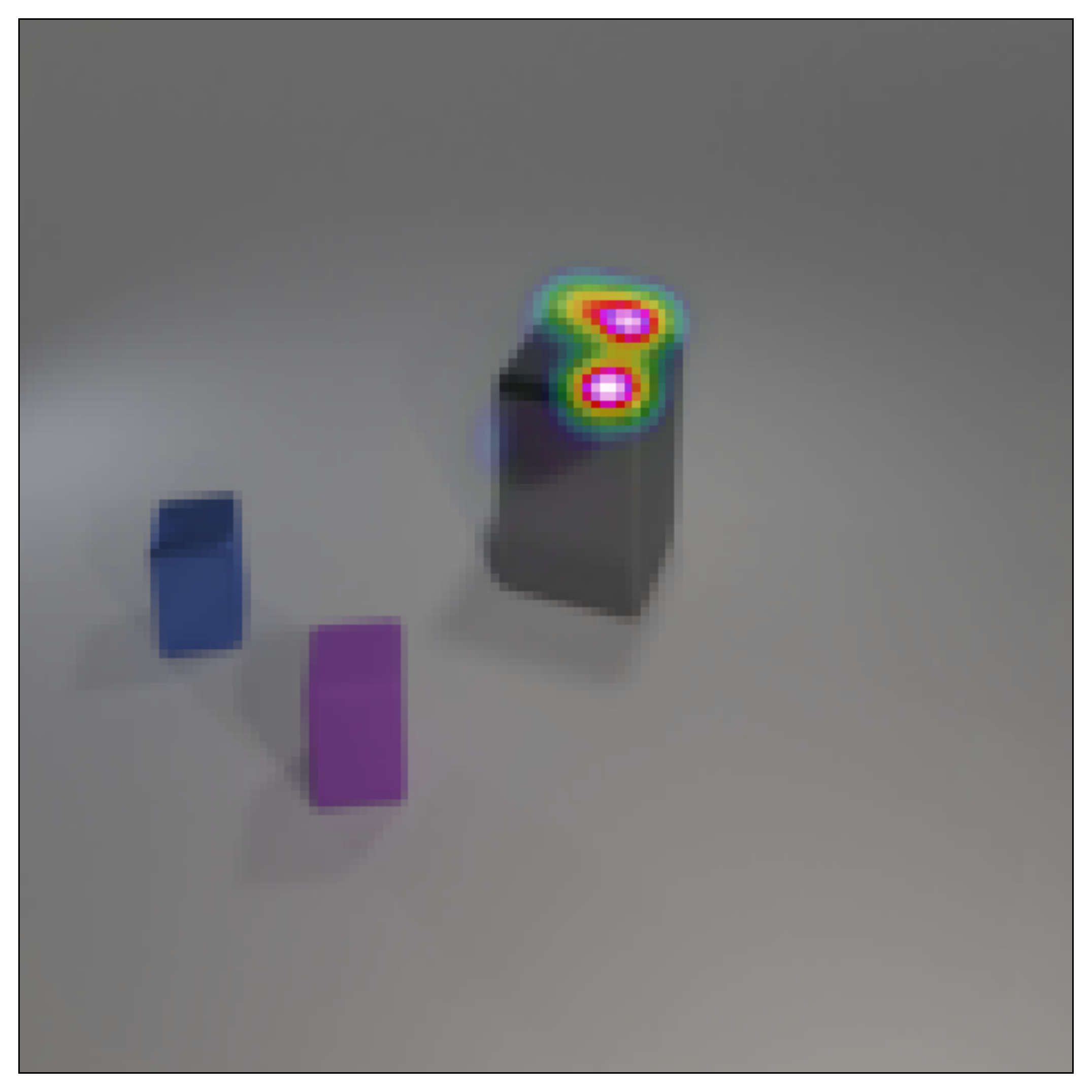} & \includegraphics[width=.12\linewidth,valign=m]{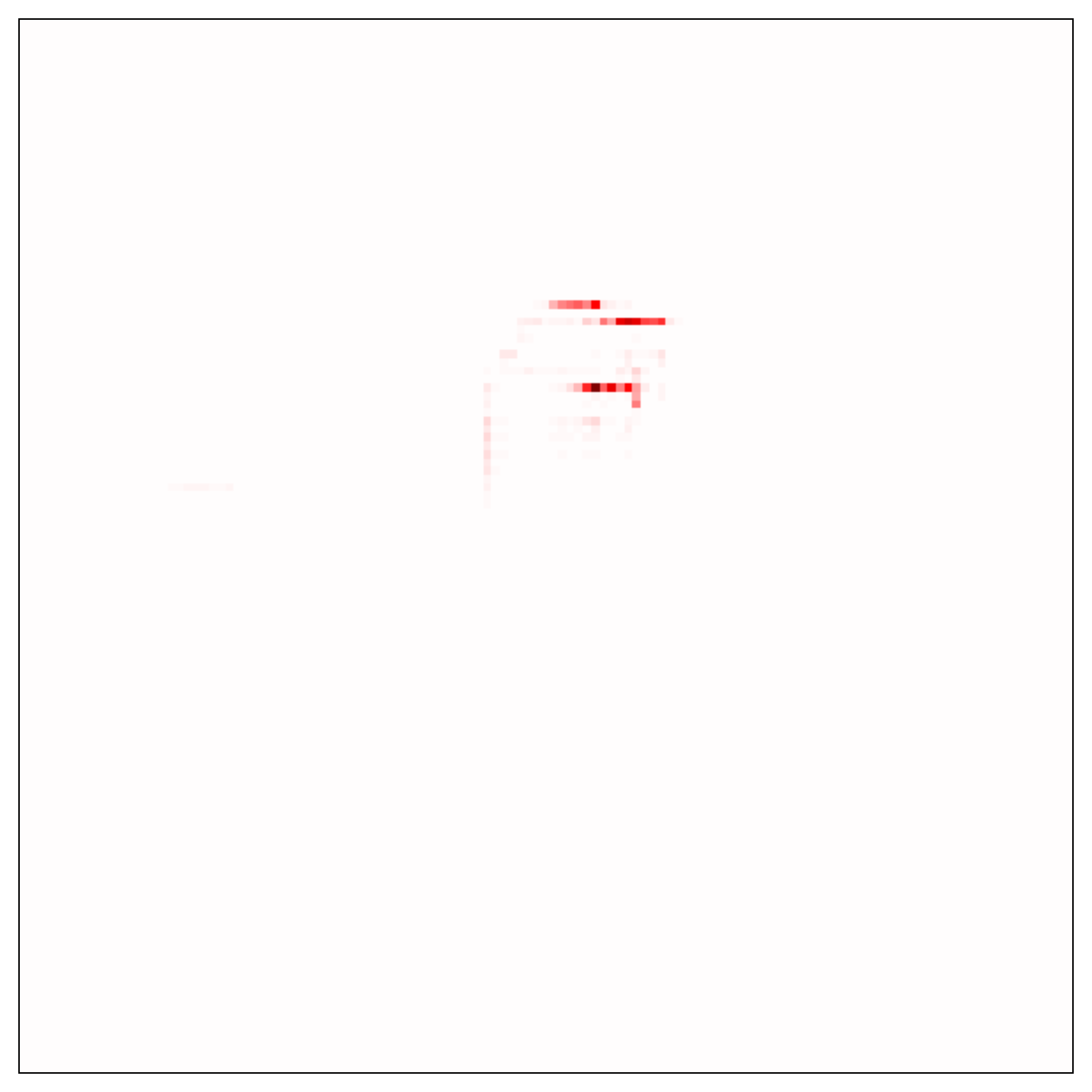} & 0.82 \\
IG \cite{Sundararajan:ICML2017}                     & \includegraphics[width=.12\linewidth,valign=m]{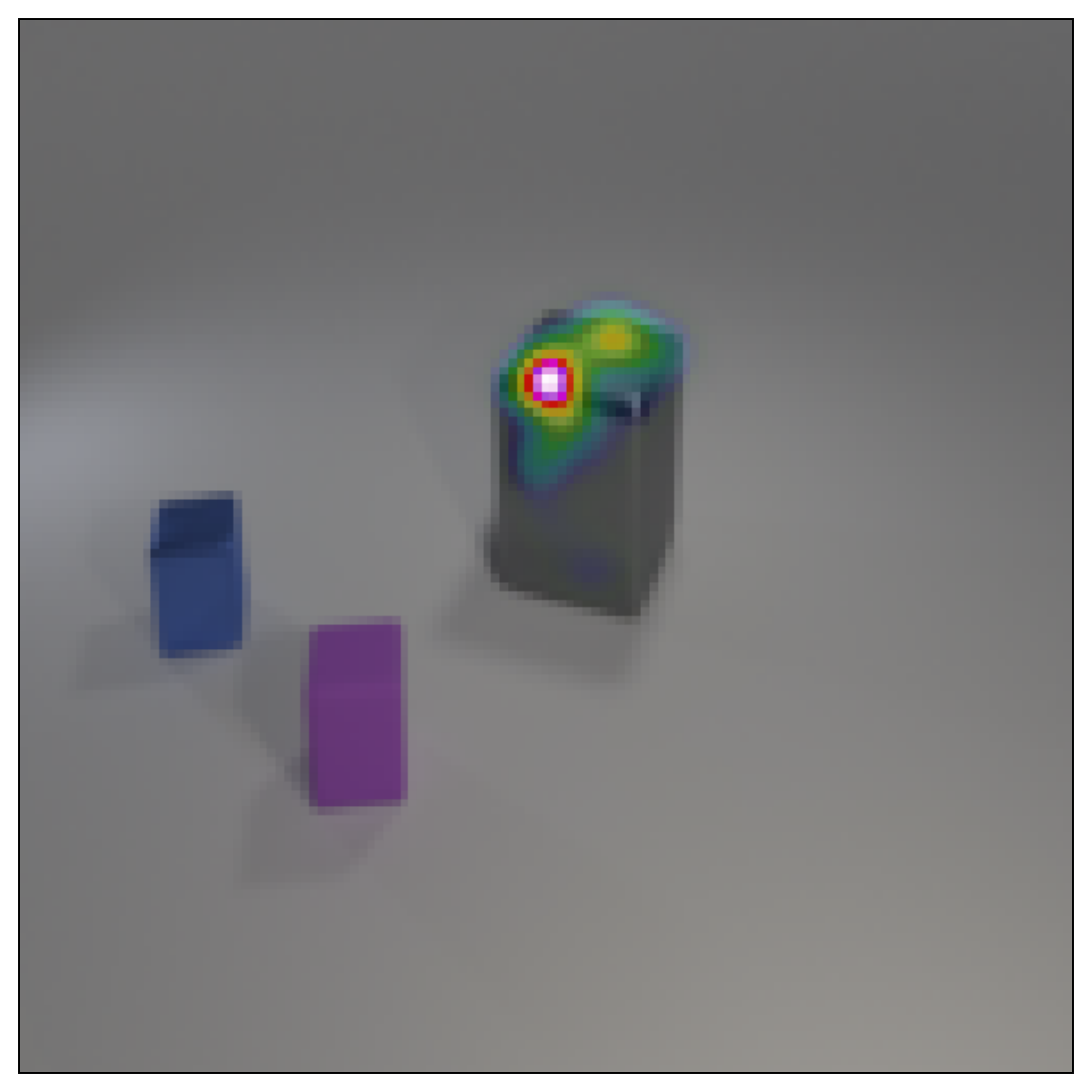} & \includegraphics[width=.12\linewidth,valign=m]{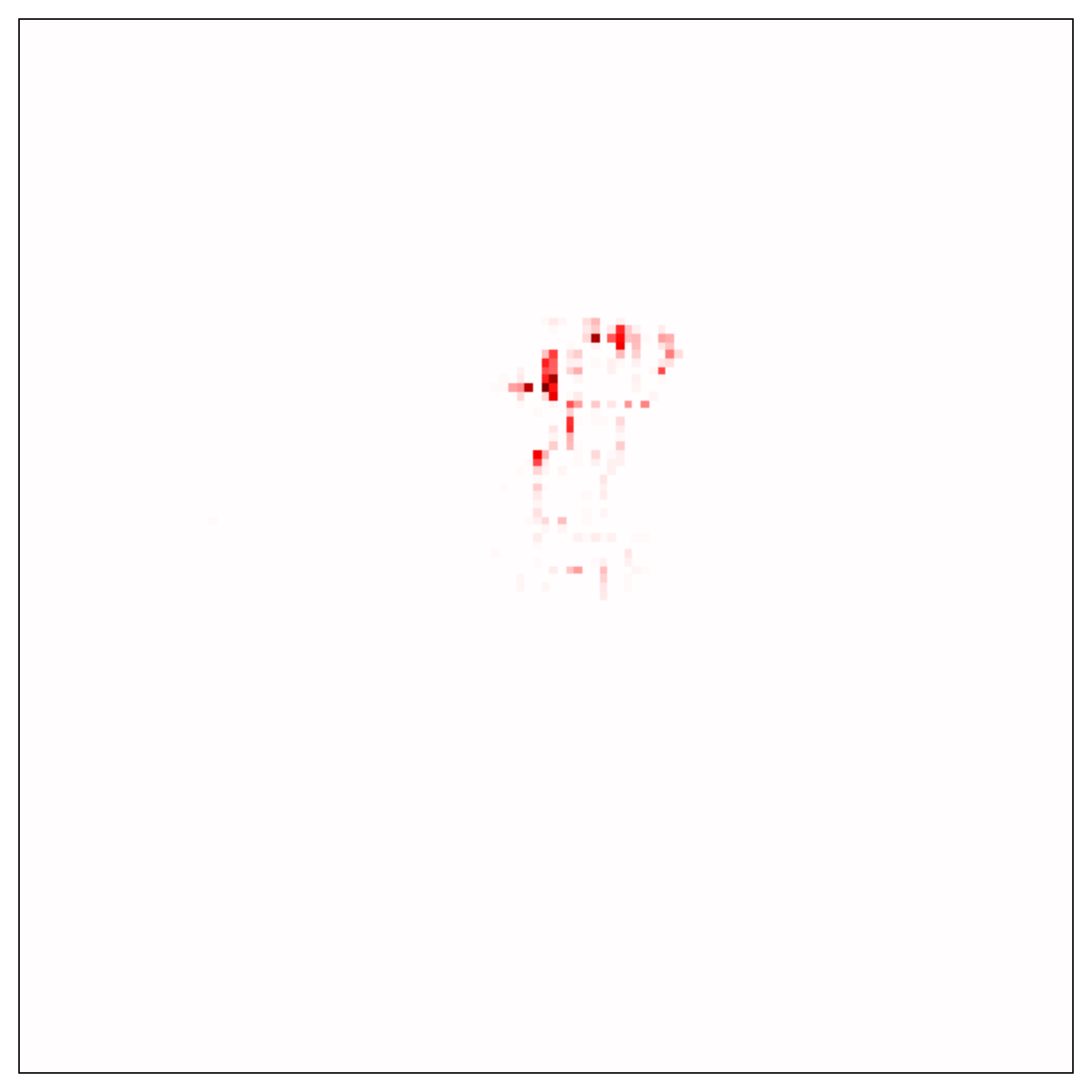} & 0.99 \\
Guided Backprop \cite{Spring:ICLR2015}              & \includegraphics[width=.12\linewidth,valign=m]{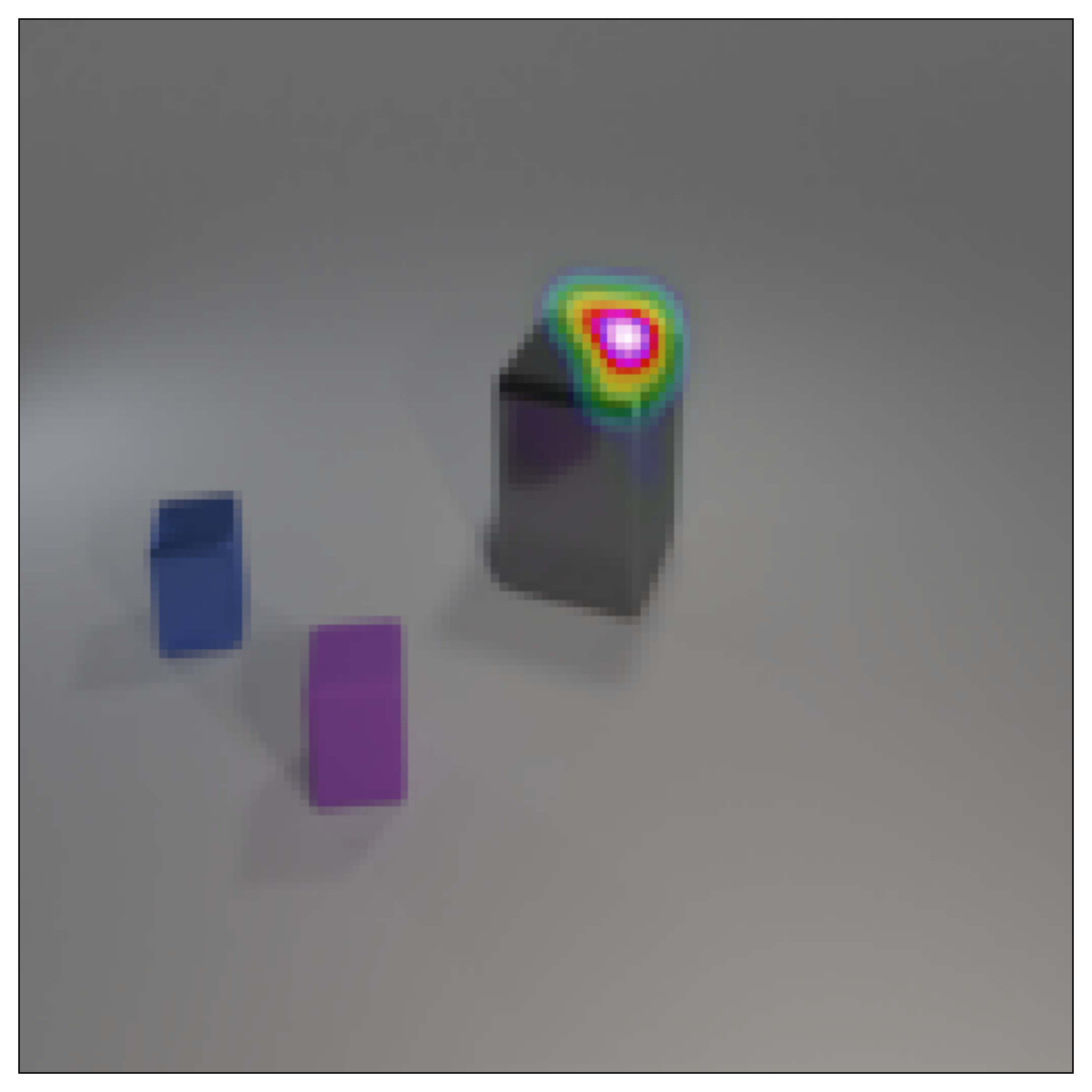} & \includegraphics[width=.12\linewidth,valign=m]{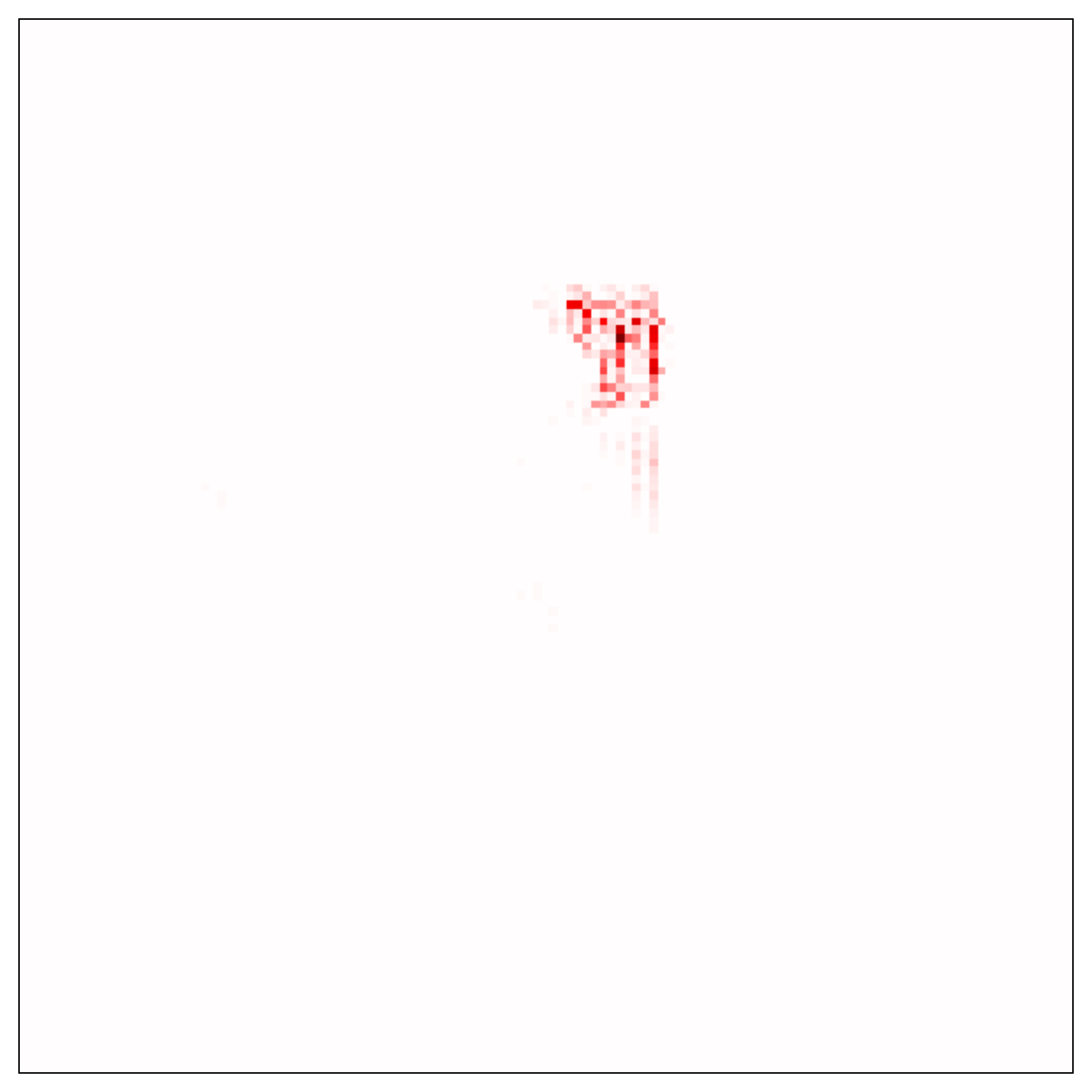} & 0.80 \\
Guided Grad-CAM \cite{Selvaraju:ICCV2017}           & \includegraphics[width=.12\linewidth,valign=m]{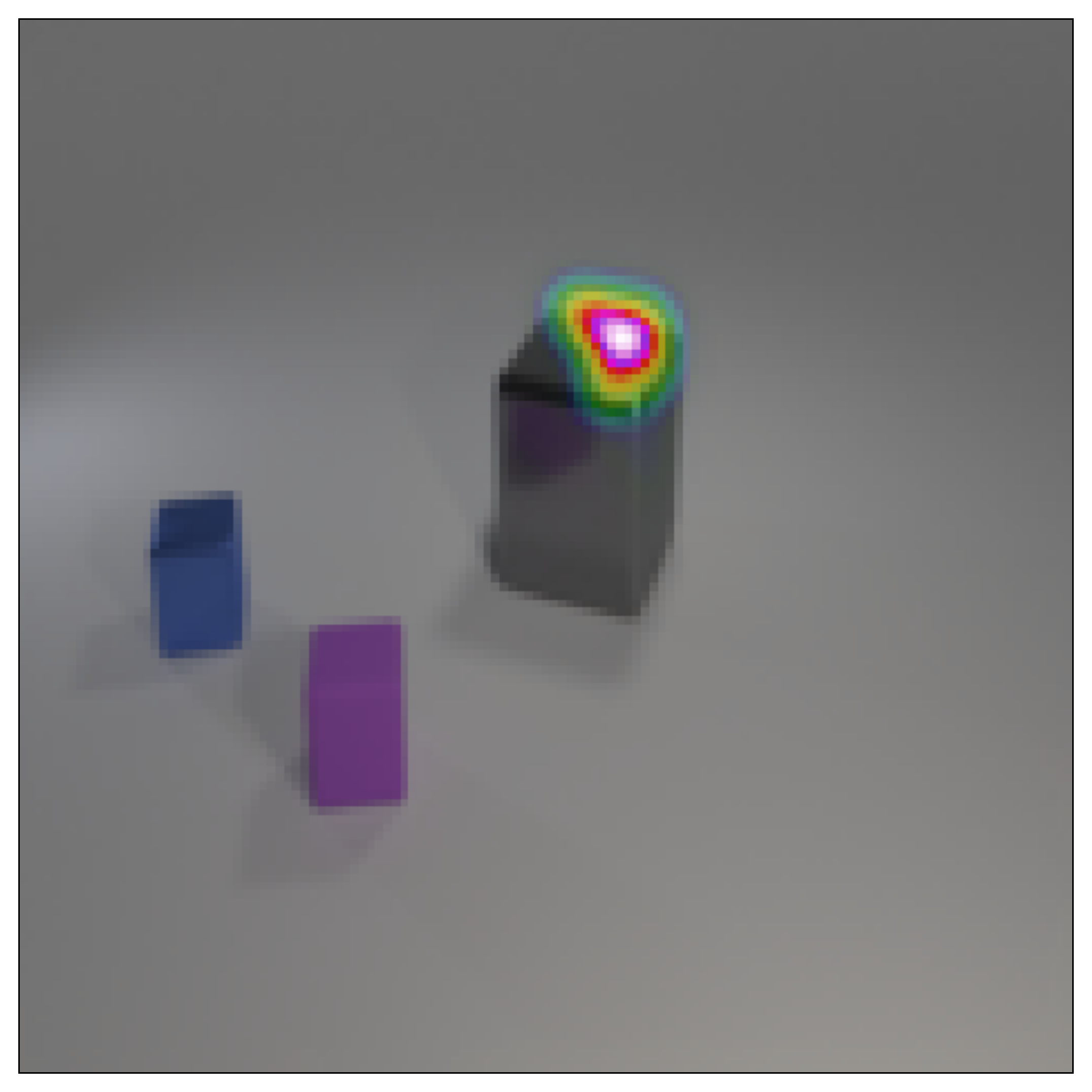} & \includegraphics[width=.12\linewidth,valign=m]{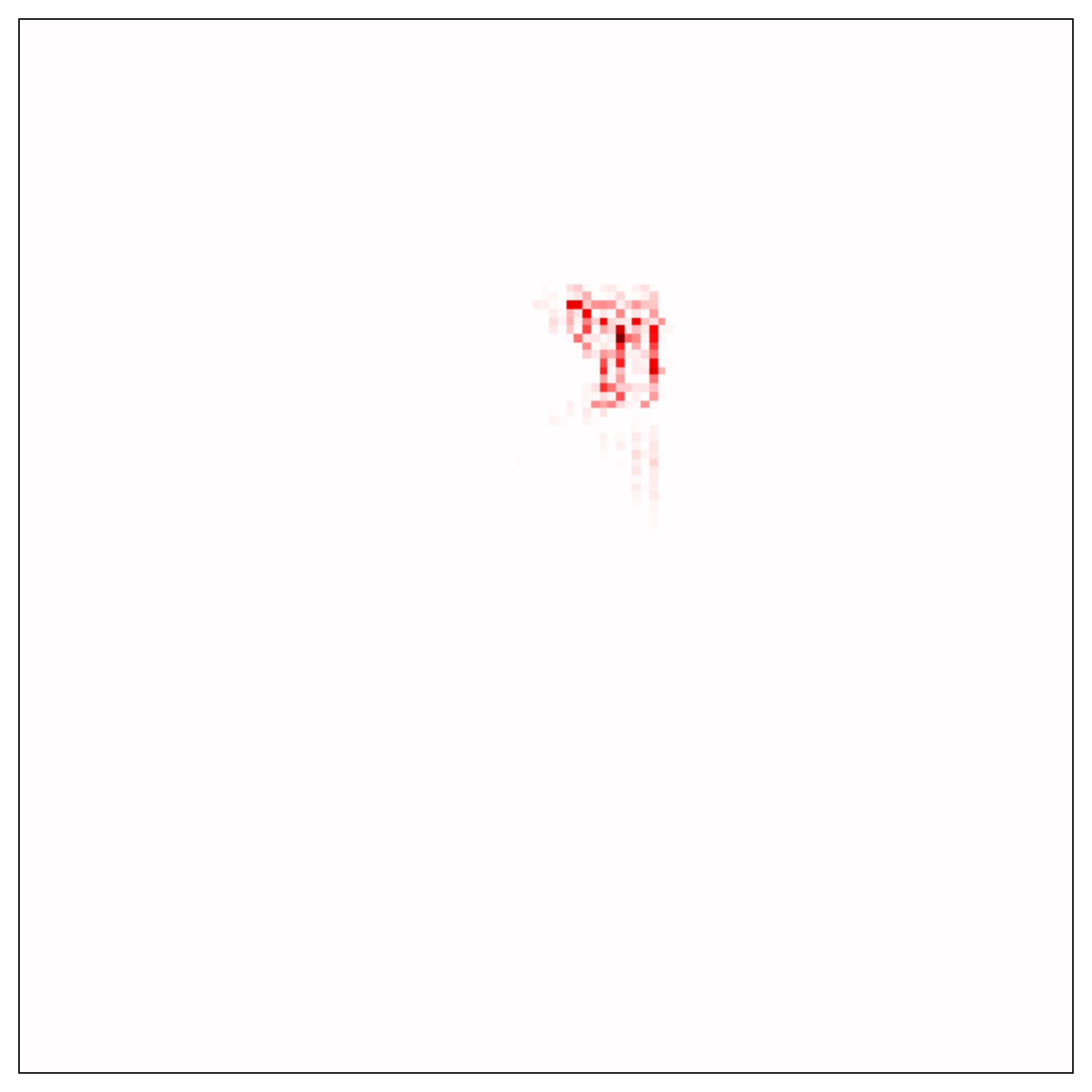} & 0.82 \\
SmoothGrad \cite{Smilkov:ICML2017}                  & \includegraphics[width=.12\linewidth,valign=m]{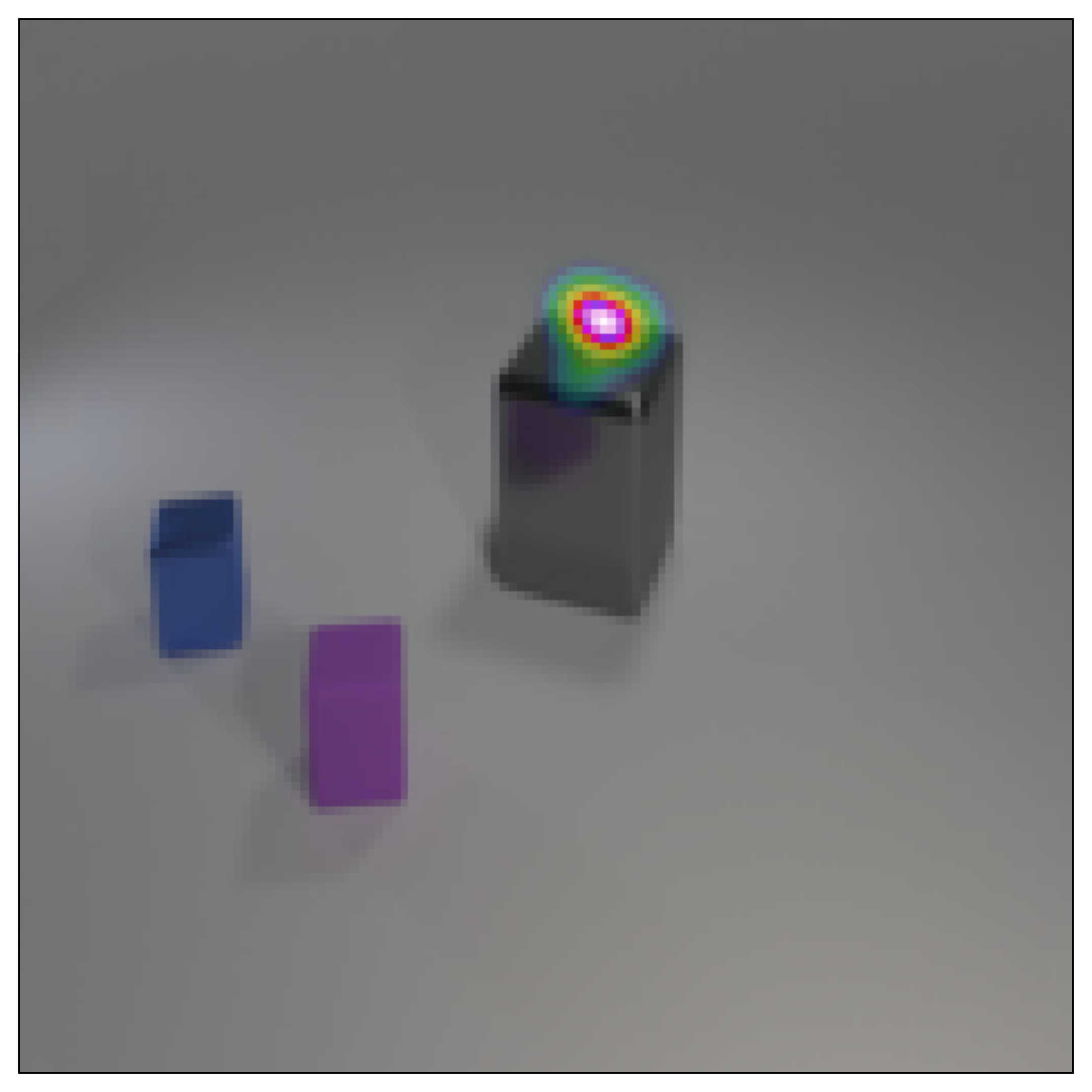} & \includegraphics[width=.12\linewidth,valign=m]{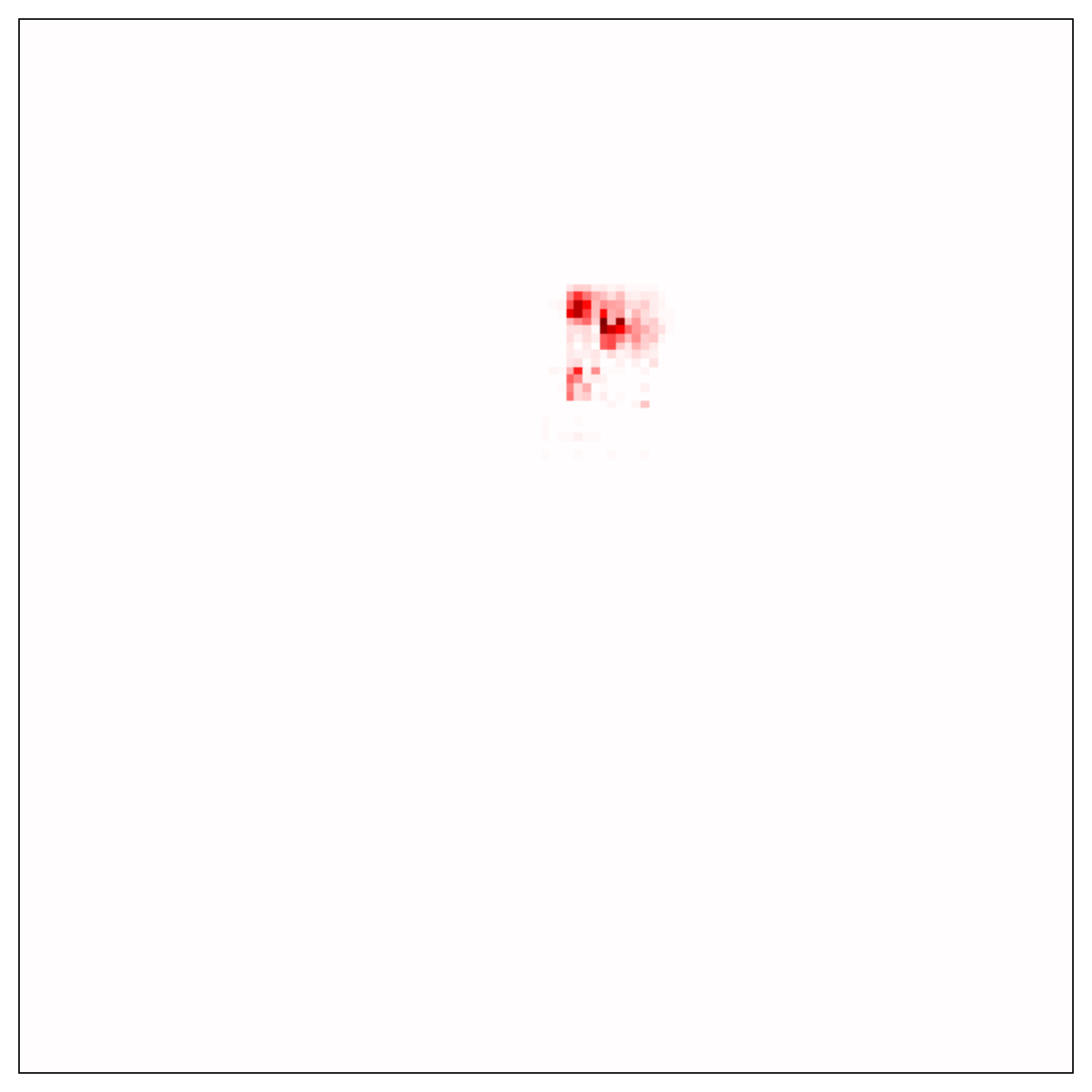} & 0.75 \\
VarGrad \cite{Adebayo:ICLR2018}                     & \includegraphics[width=.12\linewidth,valign=m]{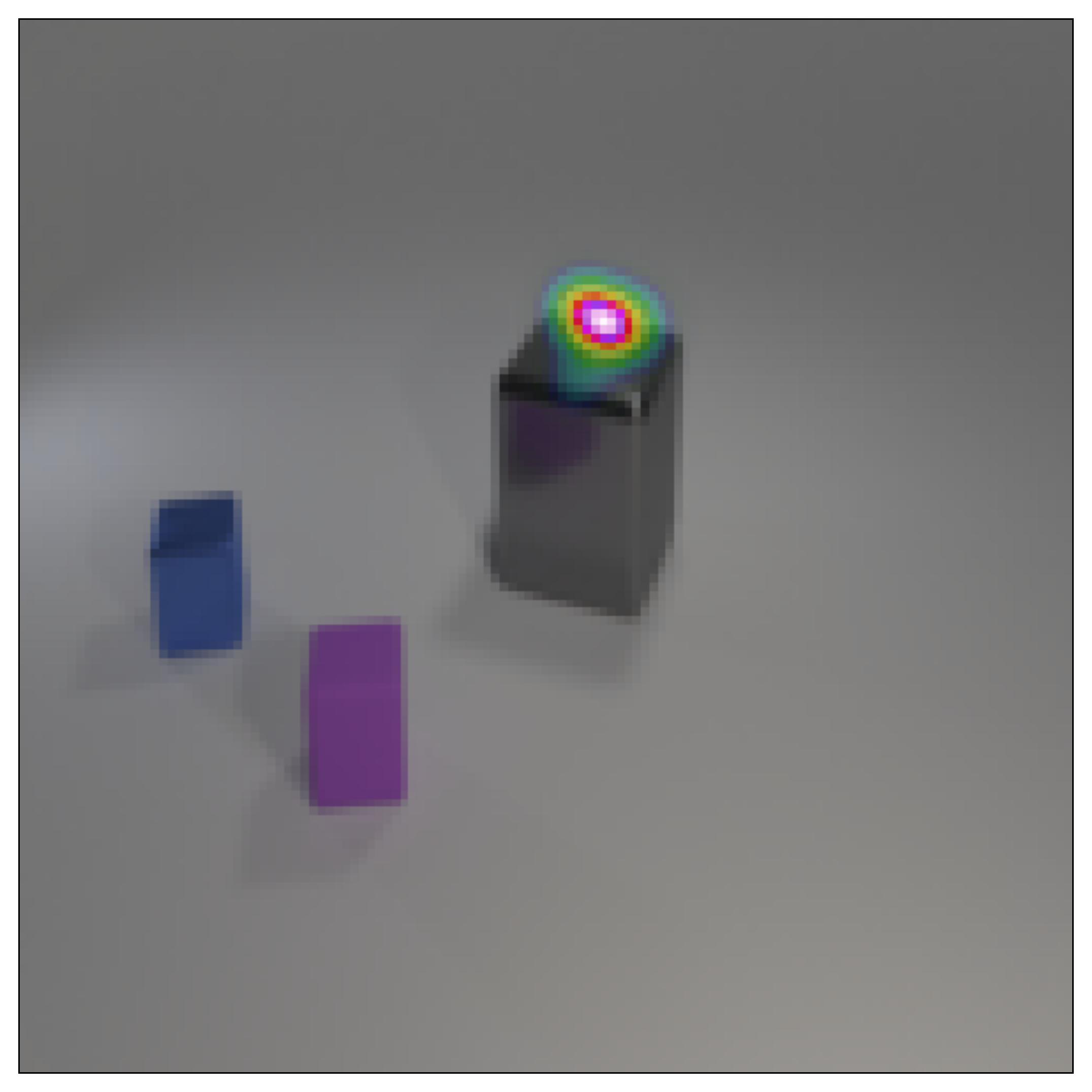} & \includegraphics[width=.12\linewidth,valign=m]{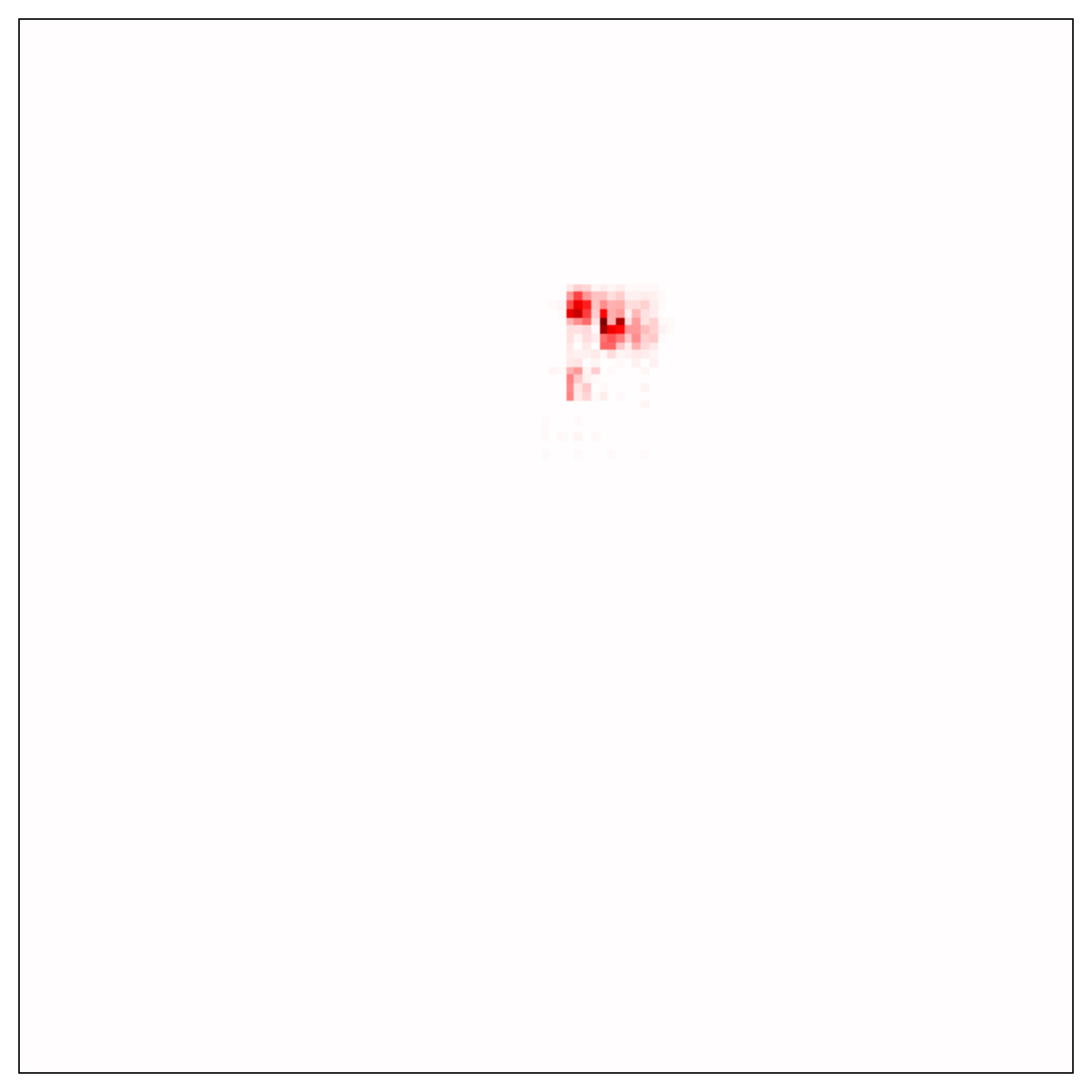} & 0.75 \\
Gradient \cite{Simonyan:ICLR2014}                   & \includegraphics[width=.12\linewidth,valign=m]{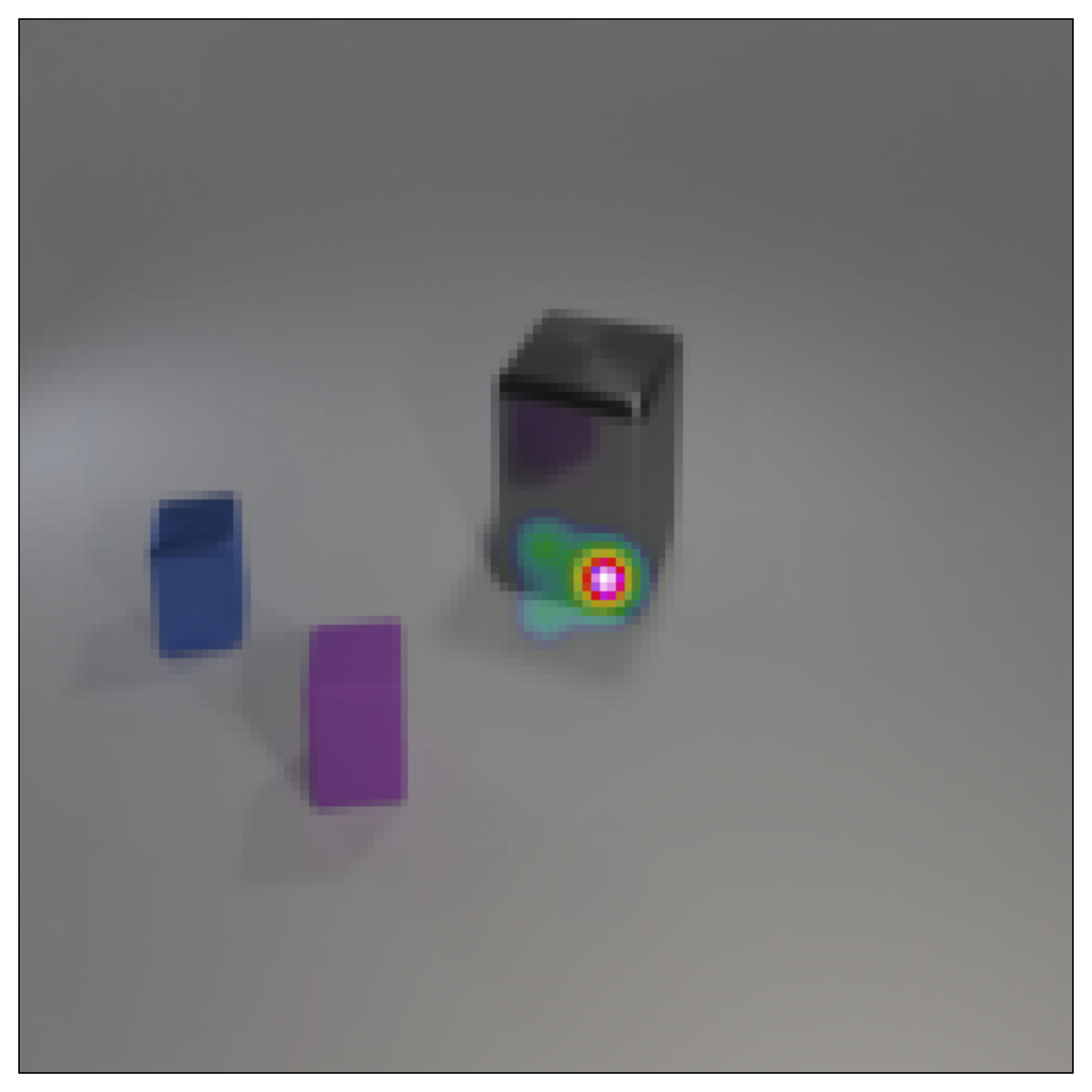} & \includegraphics[width=.12\linewidth,valign=m]{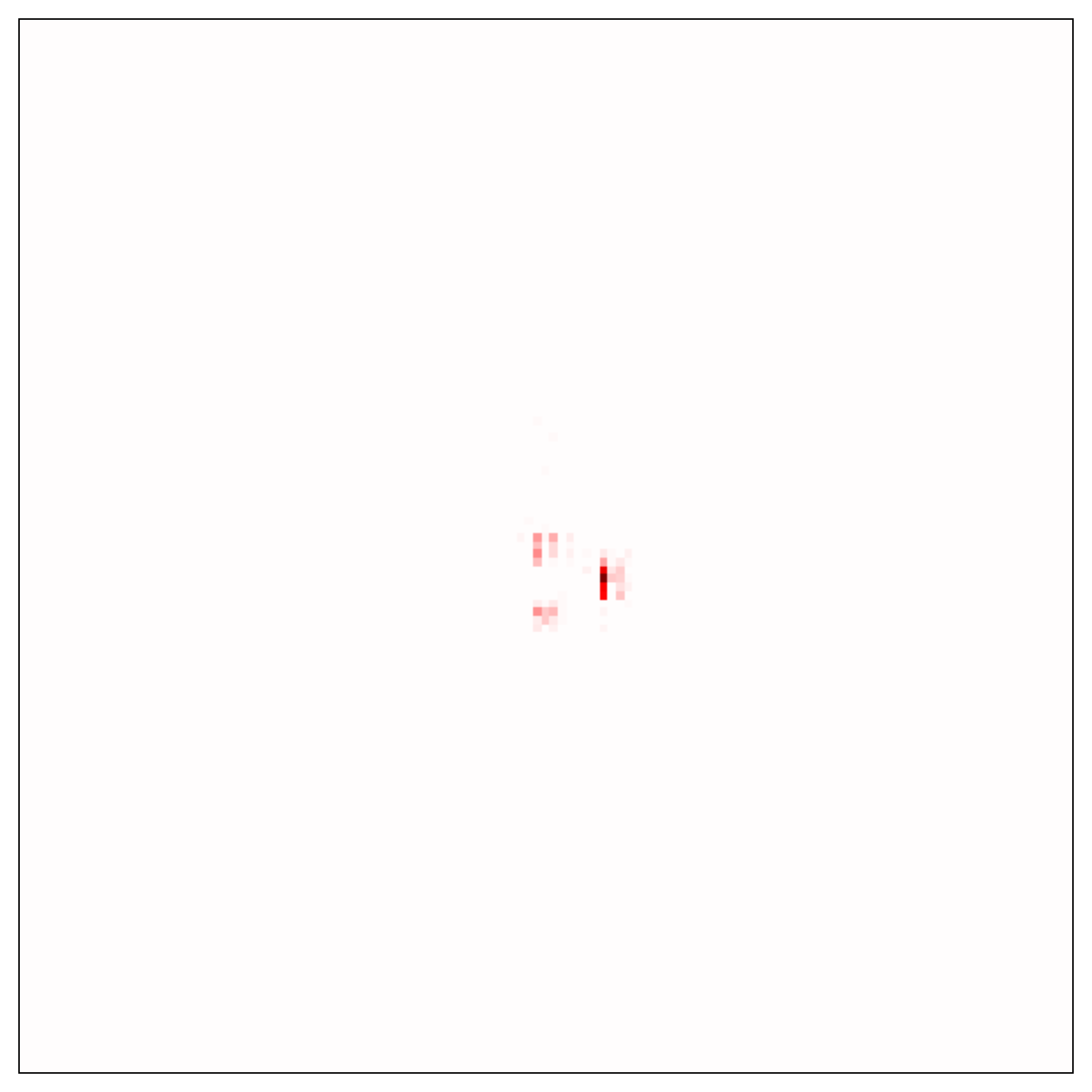} & 0.85 \\
Gradient$\times$Input \cite{Shrikumar:arxiv2016}    & \includegraphics[width=.12\linewidth,valign=m]{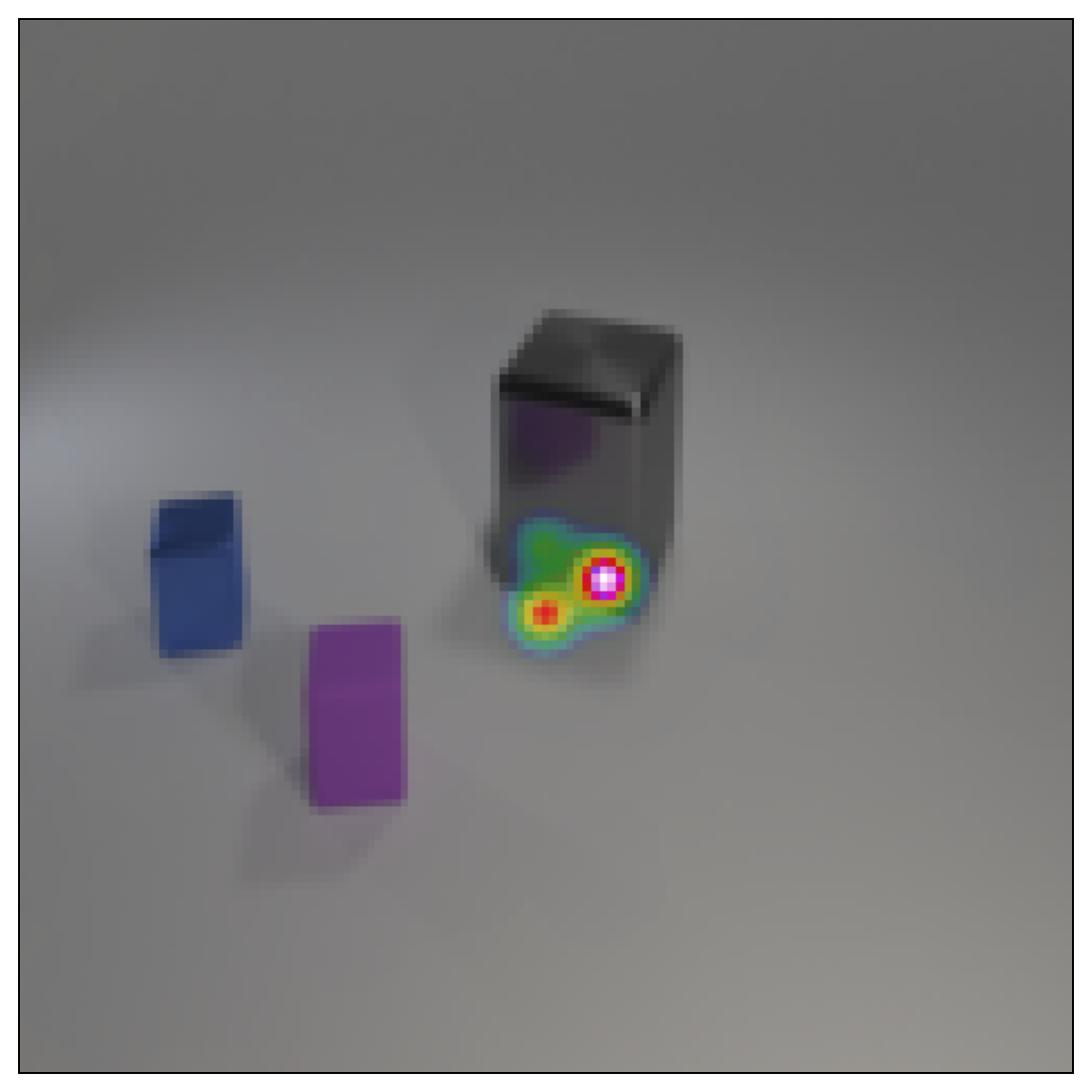} & \includegraphics[width=.12\linewidth,valign=m]{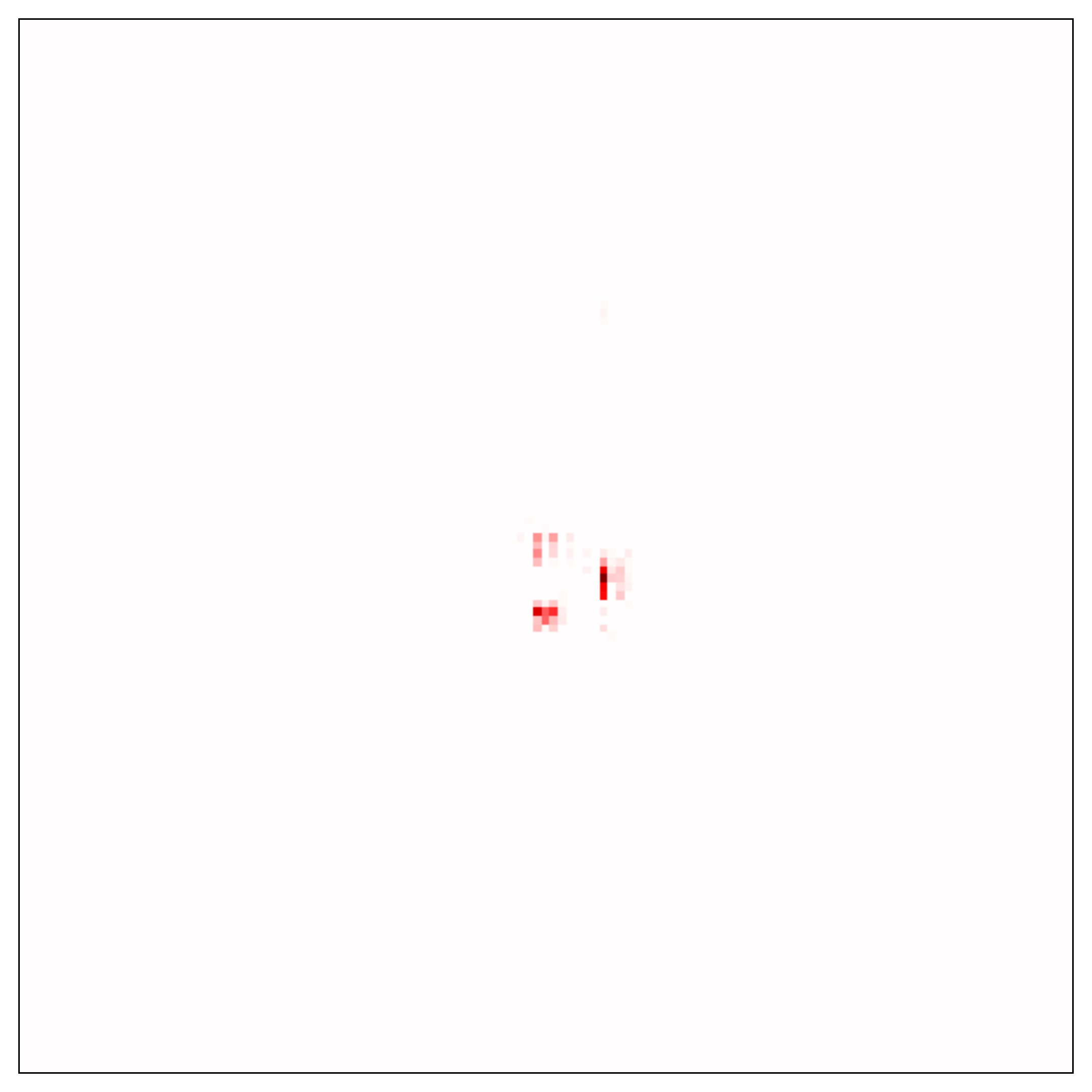} & 0.67 \\
Deconvnet \cite{Zeiler:ECCV2014}                    & \includegraphics[width=.12\linewidth,valign=m]{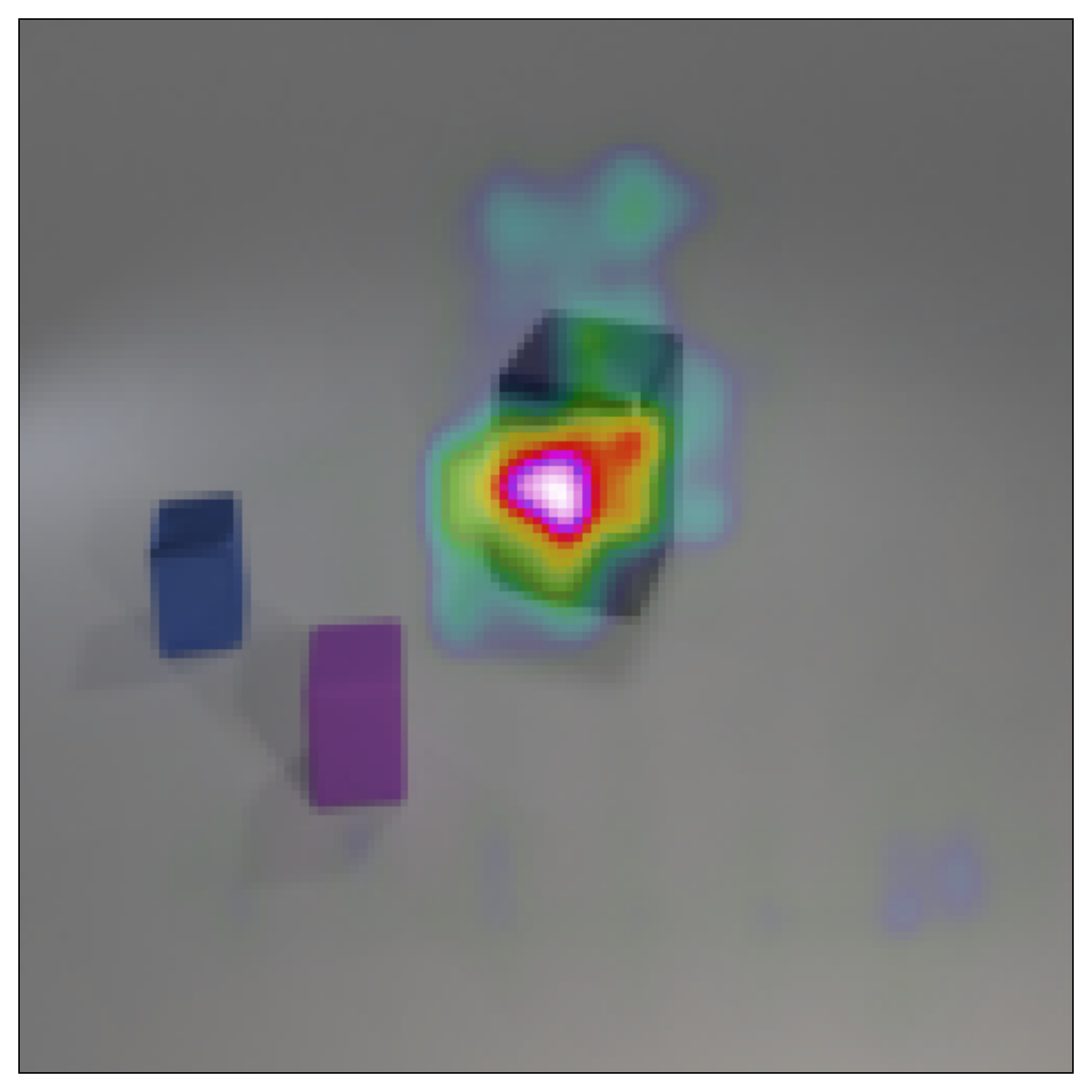} & \includegraphics[width=.12\linewidth,valign=m]{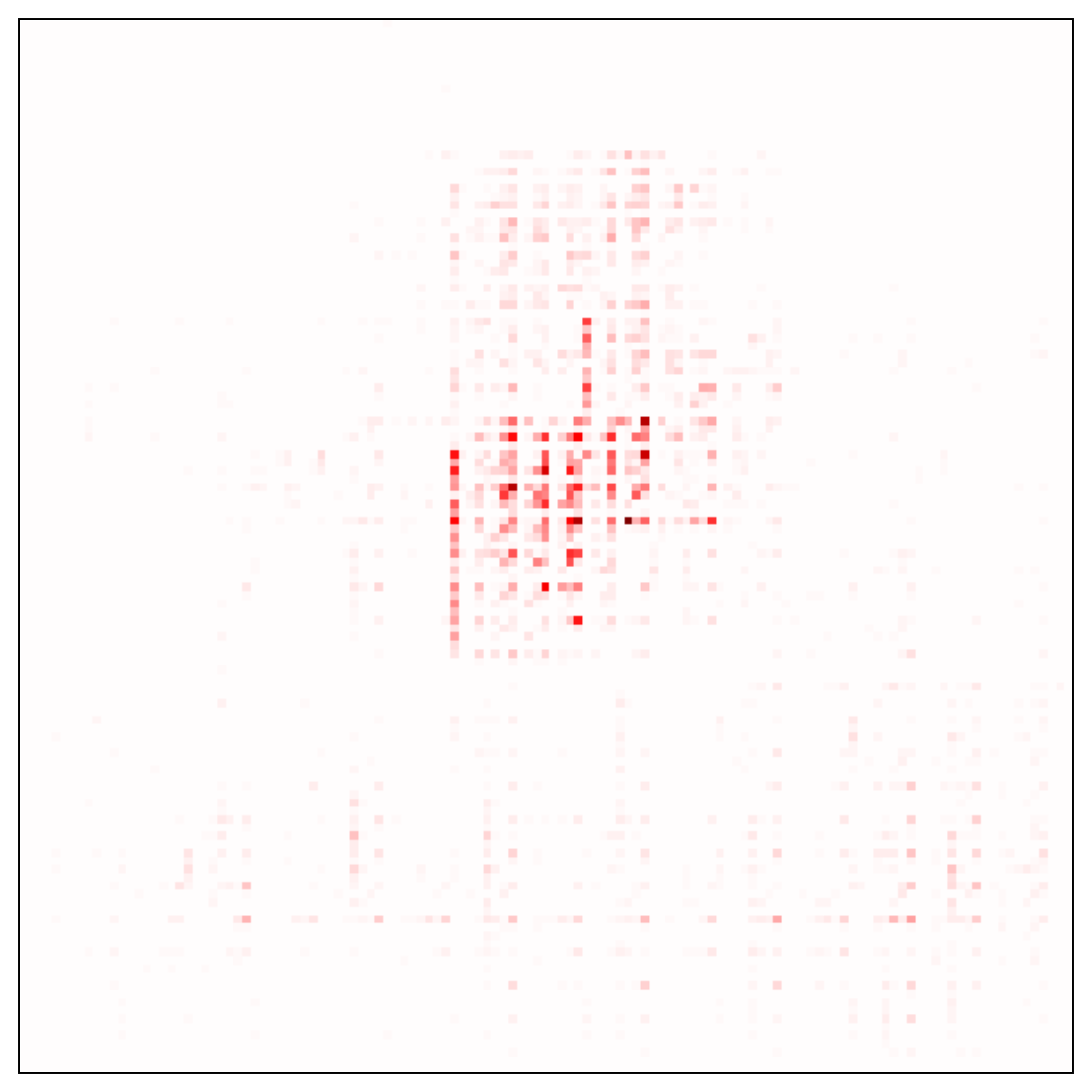} & 0.43 \\
Grad-CAM \cite{Selvaraju:ICCV2017}                  & \includegraphics[width=.12\linewidth,valign=m]{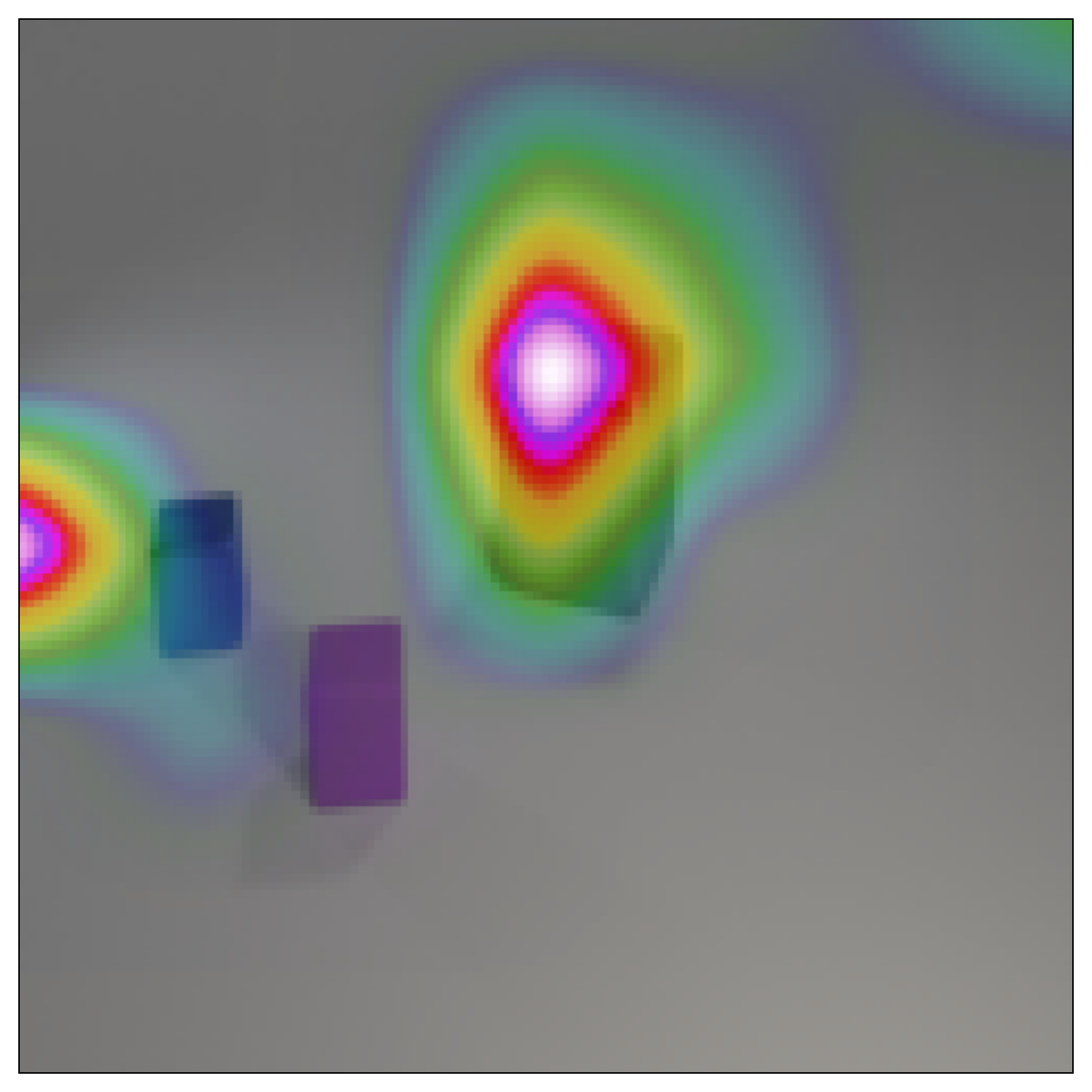} & \includegraphics[width=.12\linewidth,valign=m]{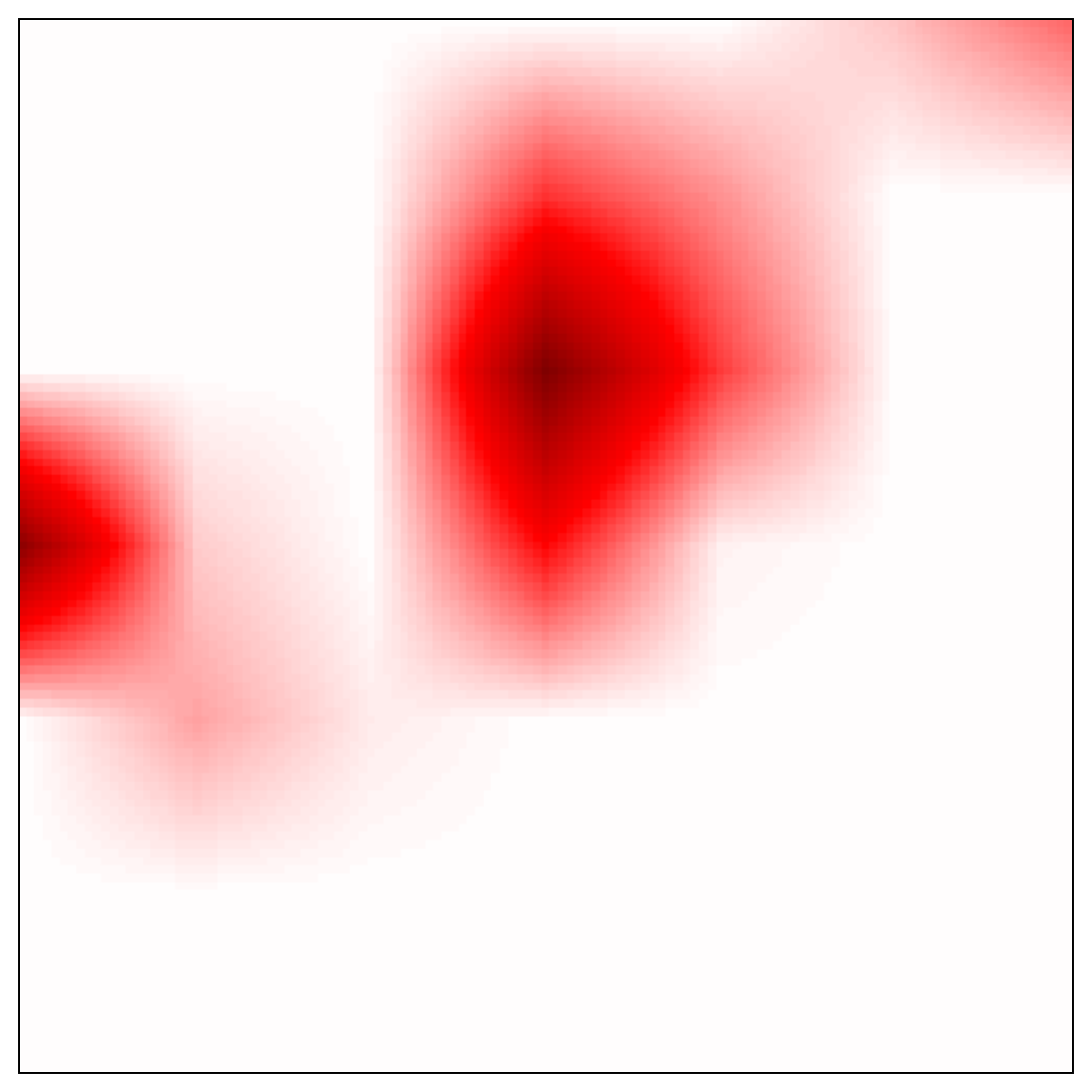} & 0.28 \\
\end{tabular}
\end{table}

\begin{table}
        \scriptsize
		\caption{Heatmaps for a falsely predicted CLEVR-XAI-simple question (raw heatmap and heatmap overlayed with original image), and corresponding relevance \textit{mass} accuracy.}
		\label{table:heatmap-simple-false-33038}
\begin{tabular}{lllc}
\midrule
\begin{tabular}{@{}l@{}}What is the color\\ of the rubber ball? \\ true: \textit{yellow} \\ predicted: \textit{brown} \end{tabular}  & \includegraphics[width=.18\linewidth,valign=m]{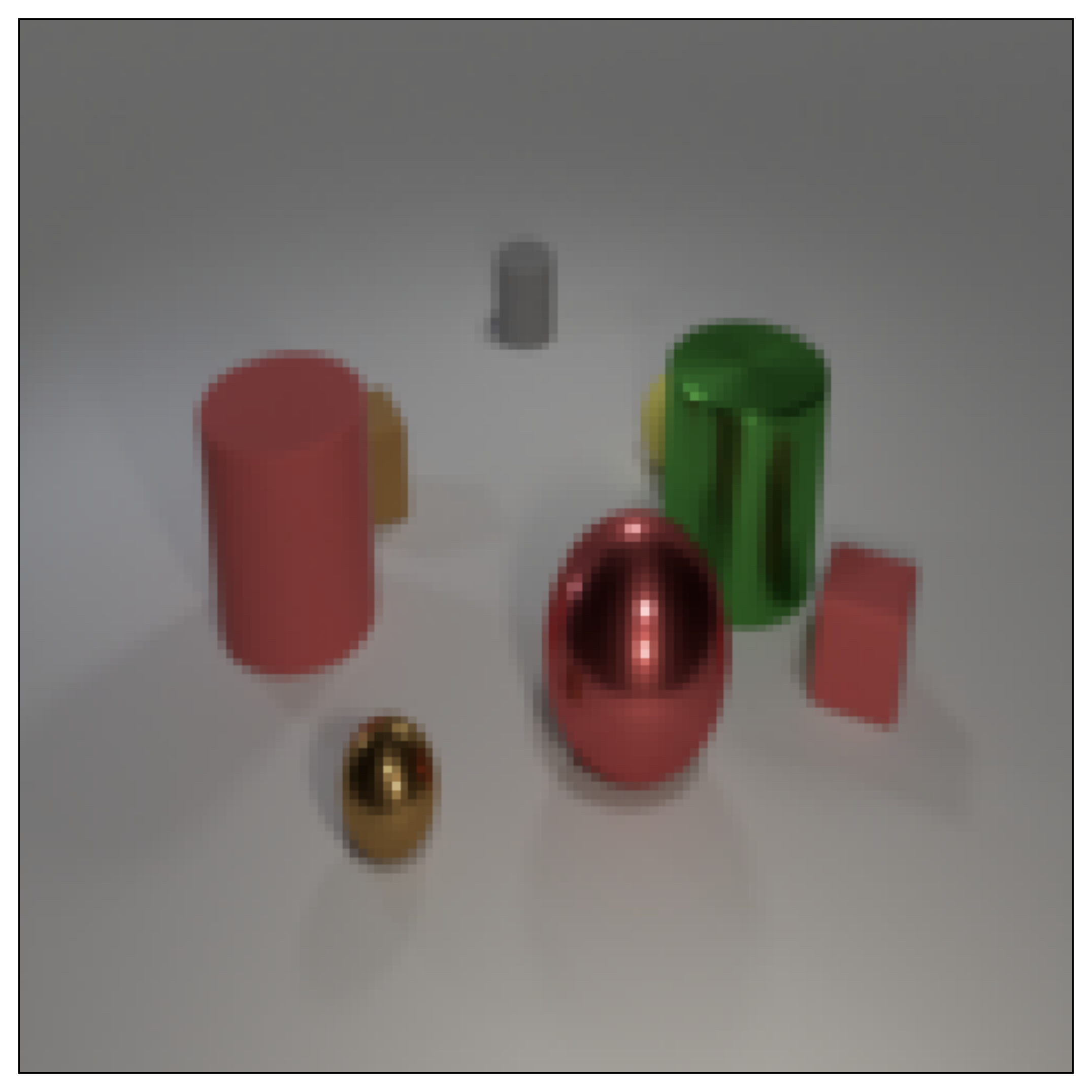} &
\includegraphics[width=.18\linewidth,valign=m]{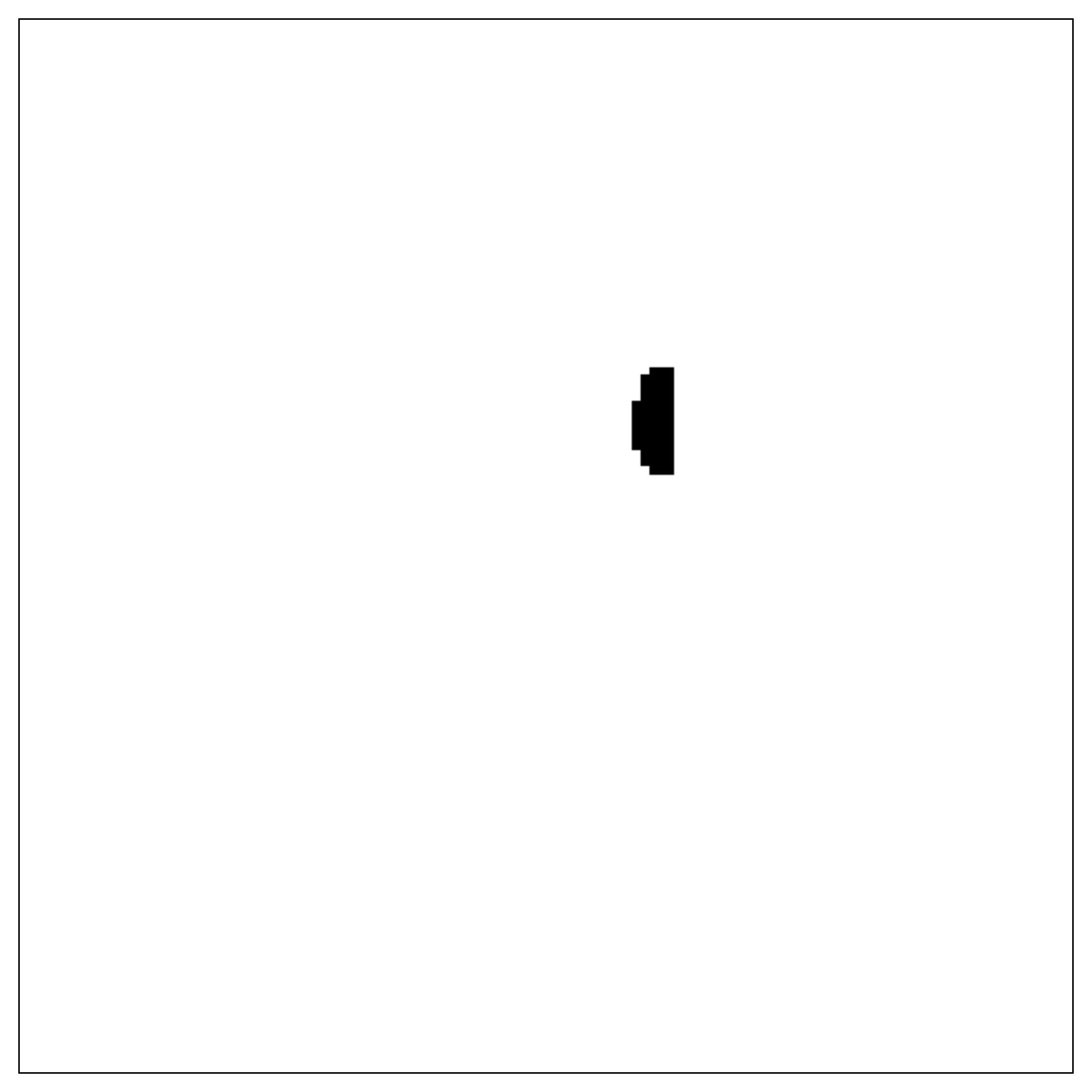} & GT Single Object \\
\midrule
LRP \cite{Bach:PLOS2015}                            & \includegraphics[width=.12\linewidth,valign=m]{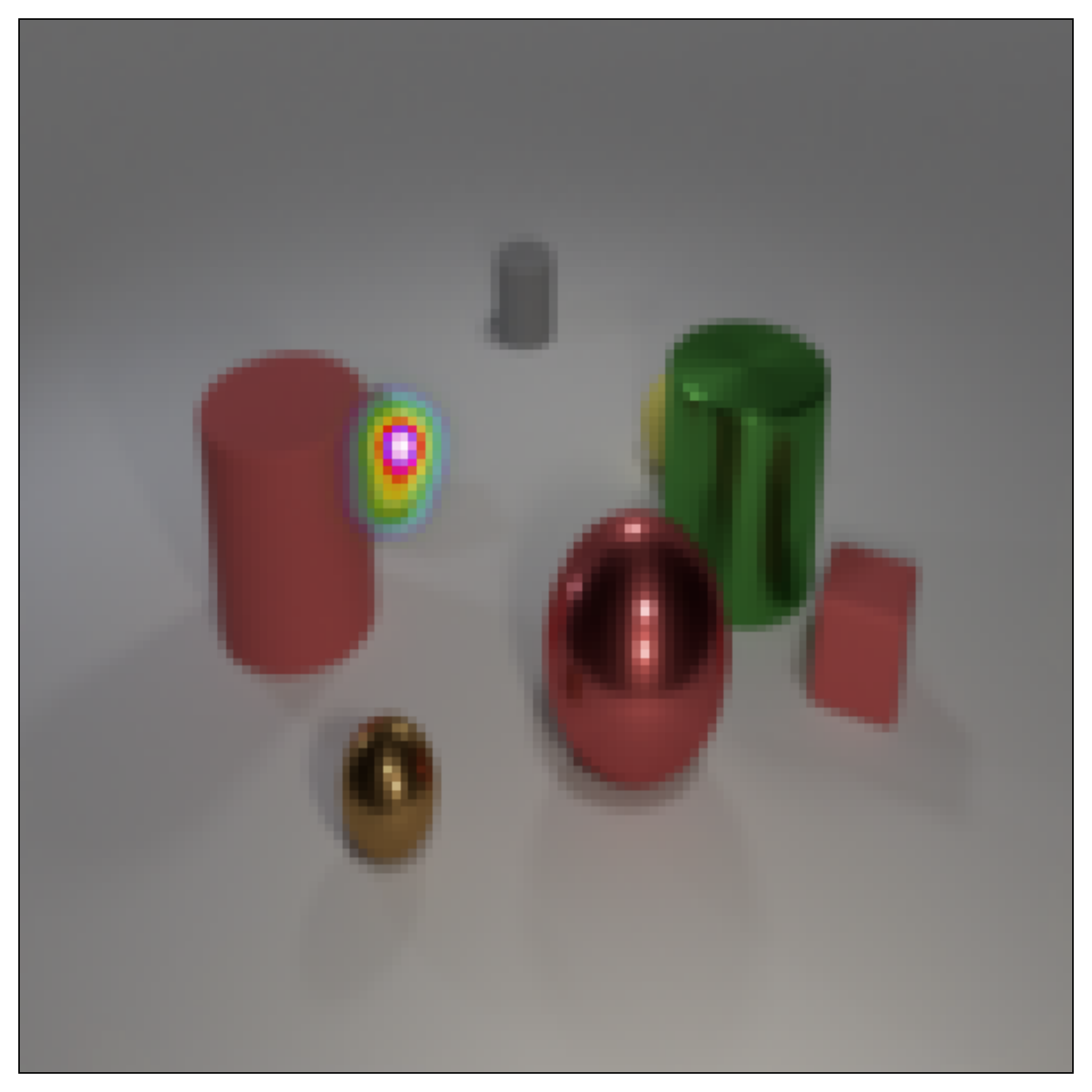} & \includegraphics[width=.12\linewidth,valign=m]{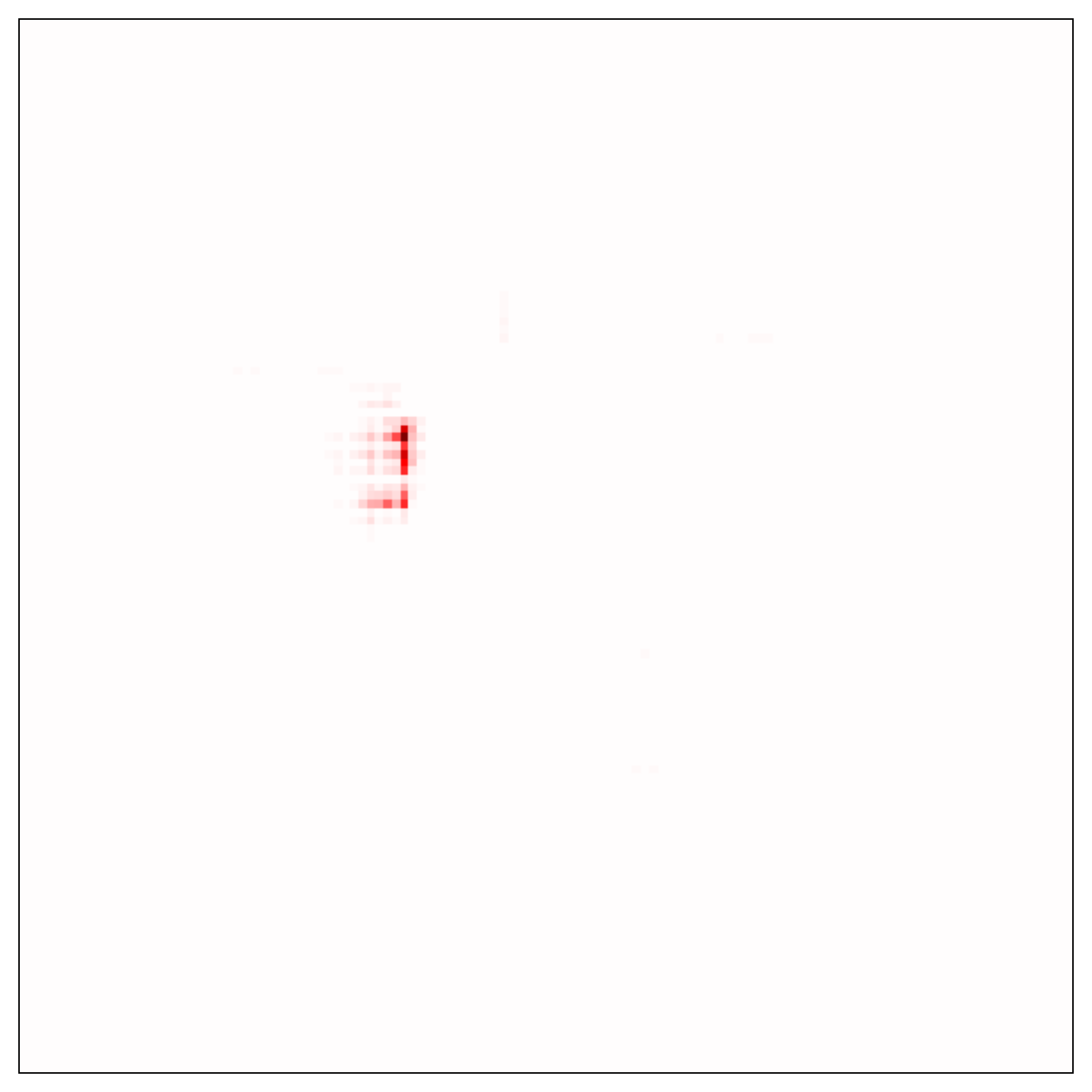} & 0.0 \\
Excitation Backprop \cite{Zhang:ECCV2016}           & \includegraphics[width=.12\linewidth,valign=m]{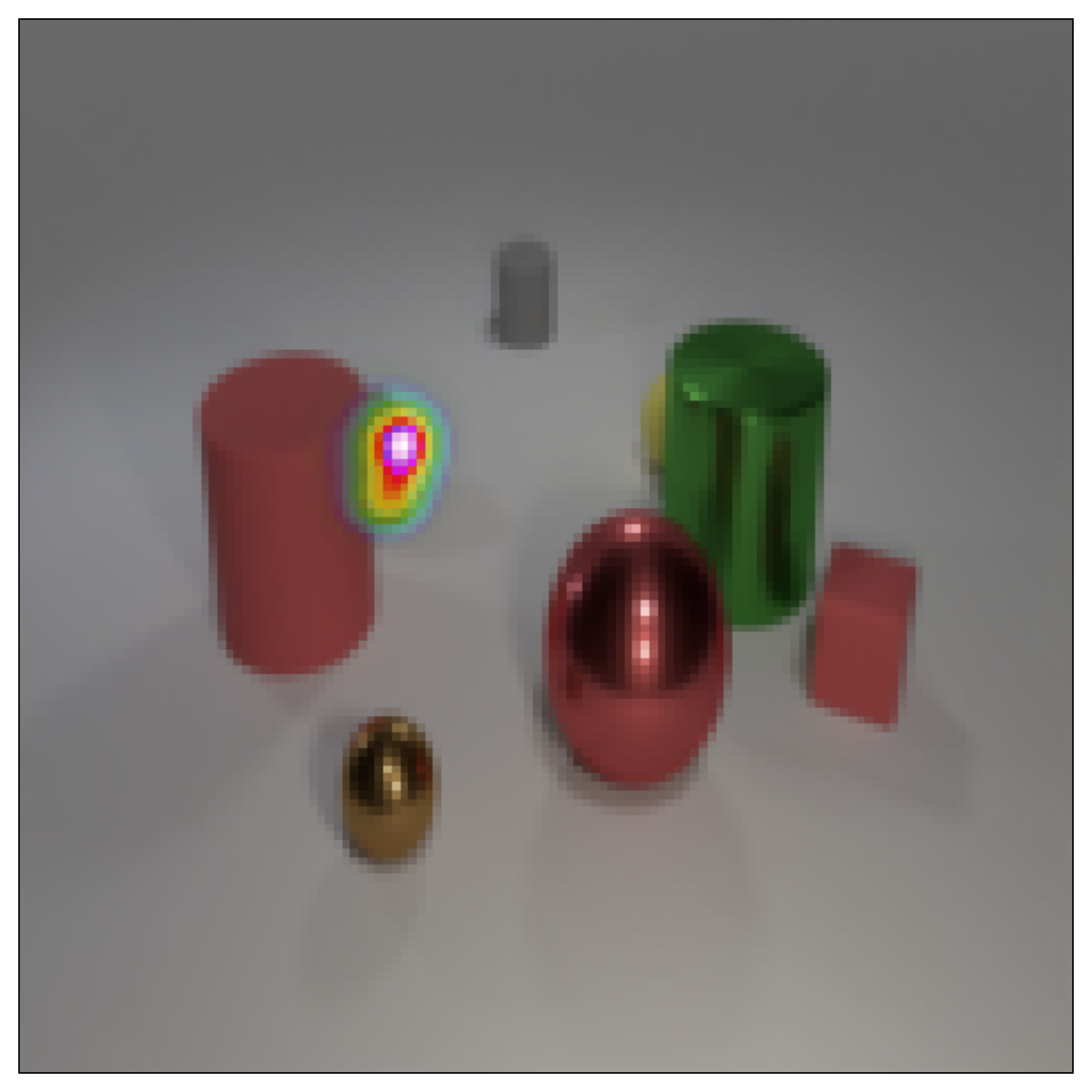} & \includegraphics[width=.12\linewidth,valign=m]{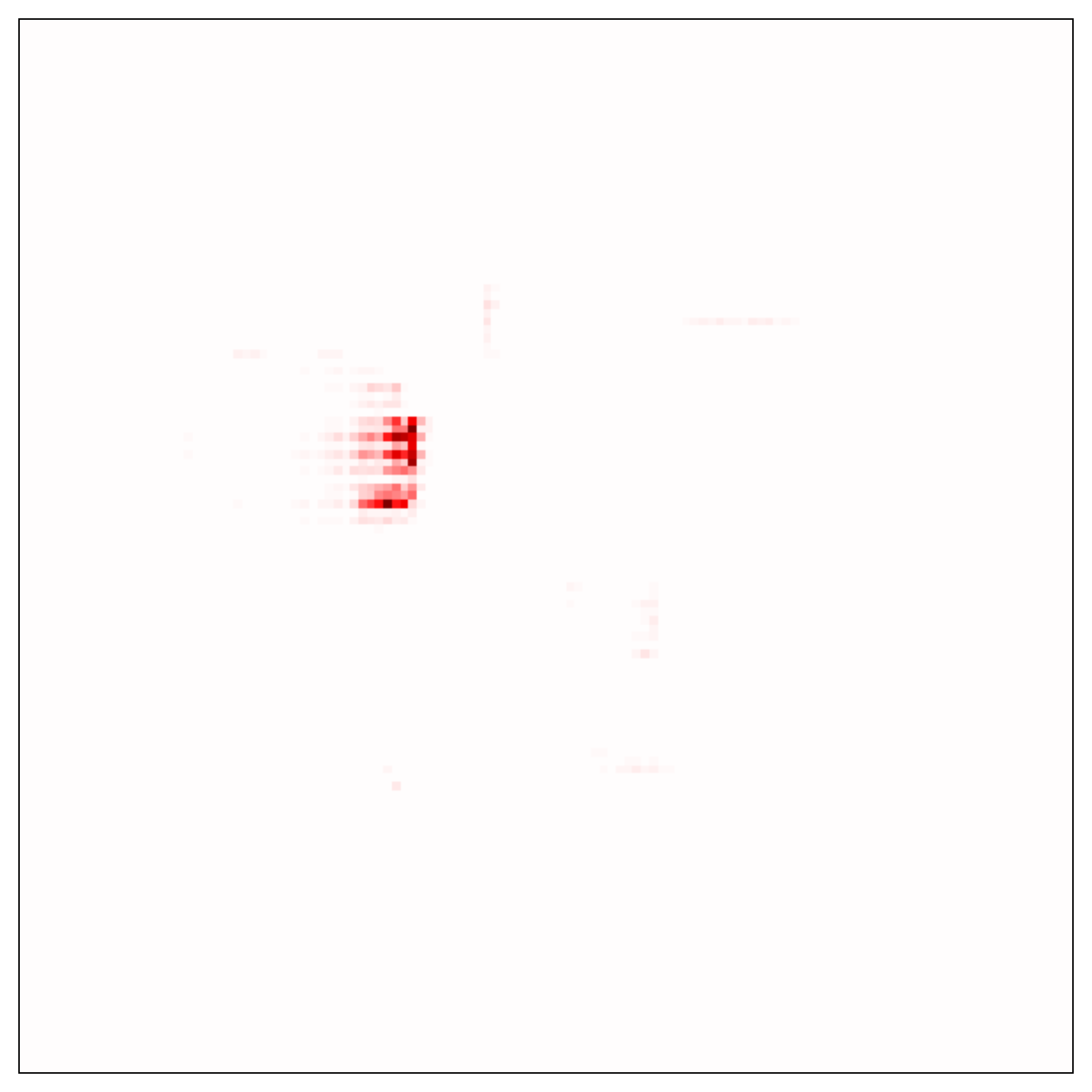} & 0.0 \\
IG \cite{Sundararajan:ICML2017}                     & \includegraphics[width=.12\linewidth,valign=m]{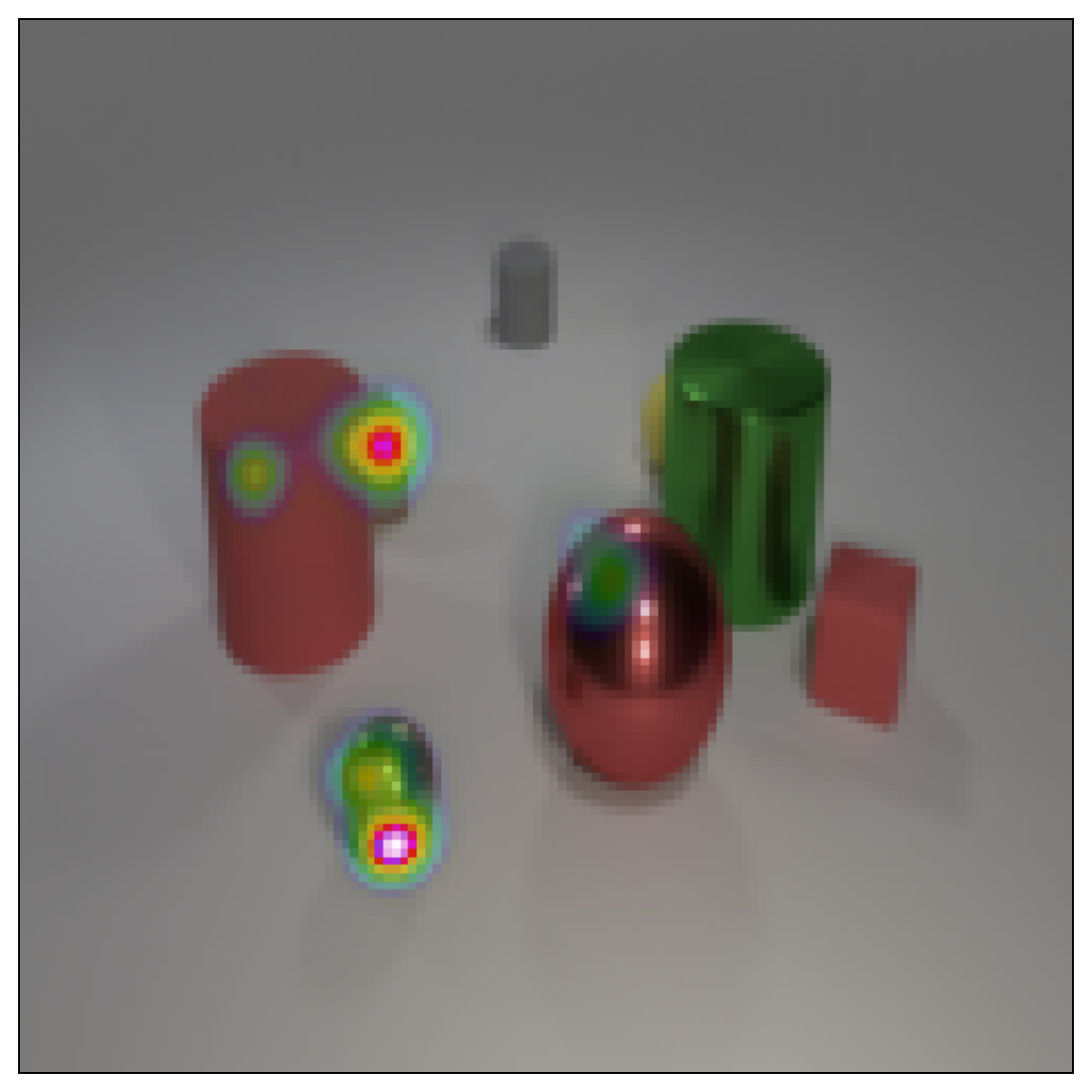} & \includegraphics[width=.12\linewidth,valign=m]{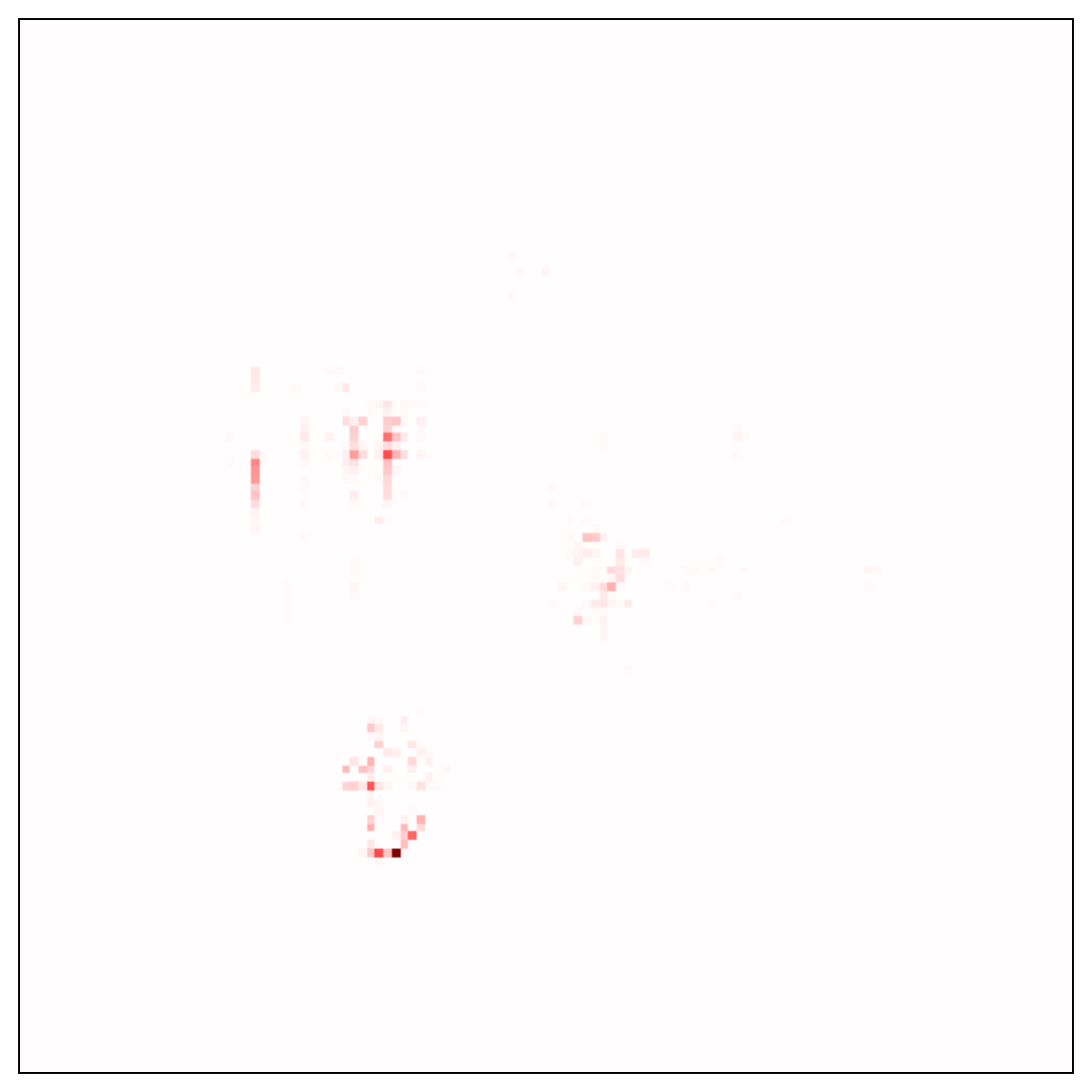} & 0.0 \\
Guided Backprop \cite{Spring:ICLR2015}              & \includegraphics[width=.12\linewidth,valign=m]{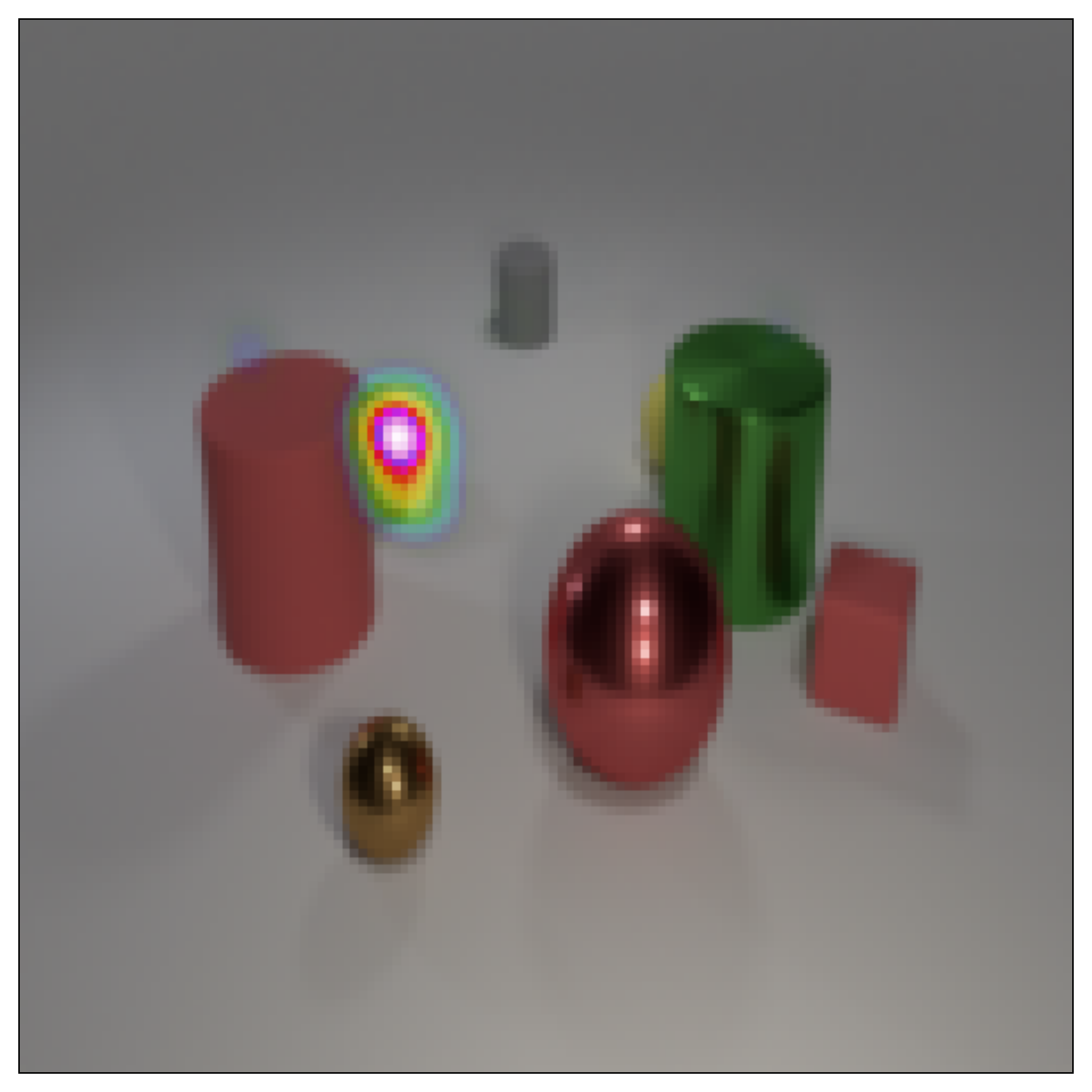} & \includegraphics[width=.12\linewidth,valign=m]{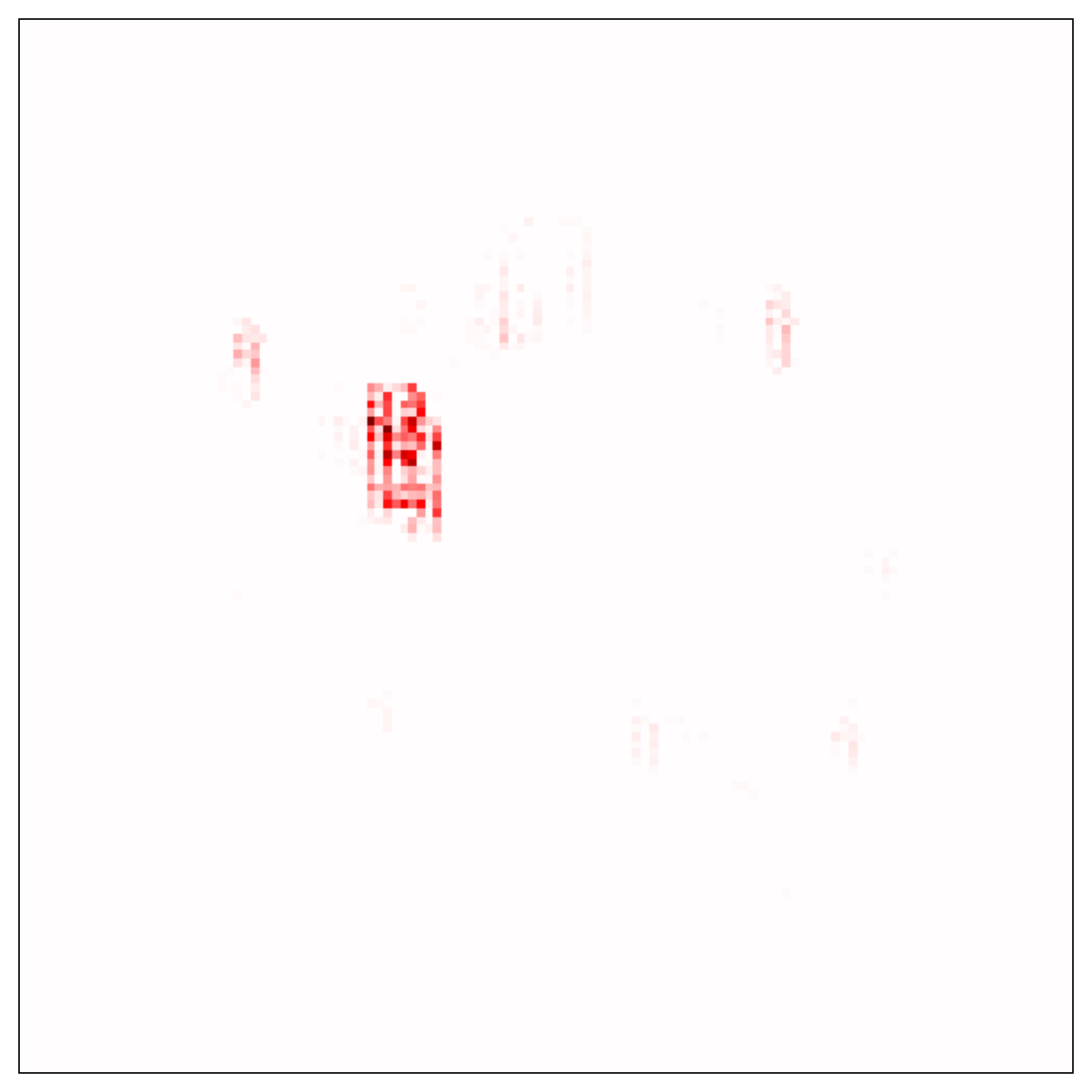} & 0.0 \\
Guided Grad-CAM \cite{Selvaraju:ICCV2017}           & \includegraphics[width=.12\linewidth,valign=m]{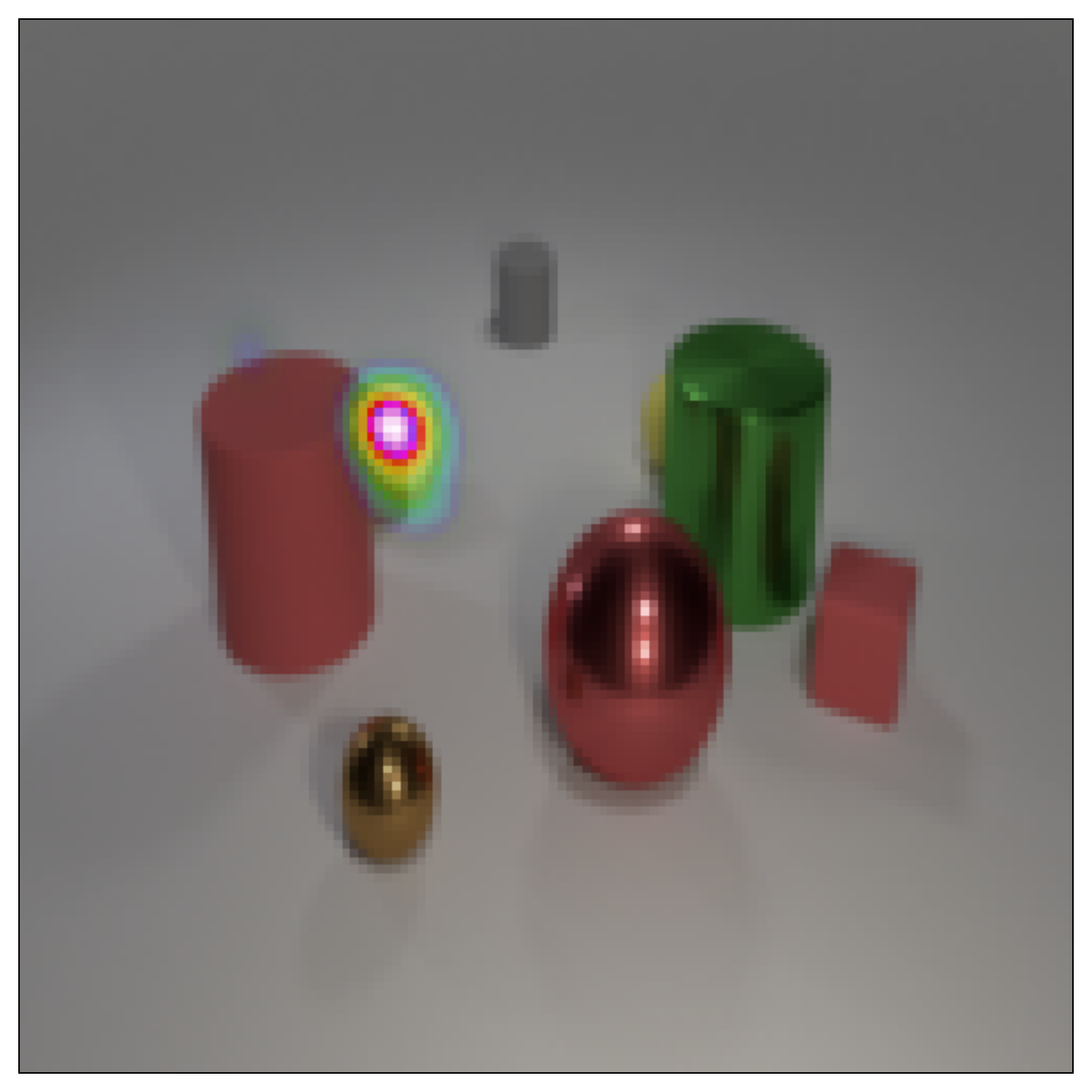} & \includegraphics[width=.12\linewidth,valign=m]{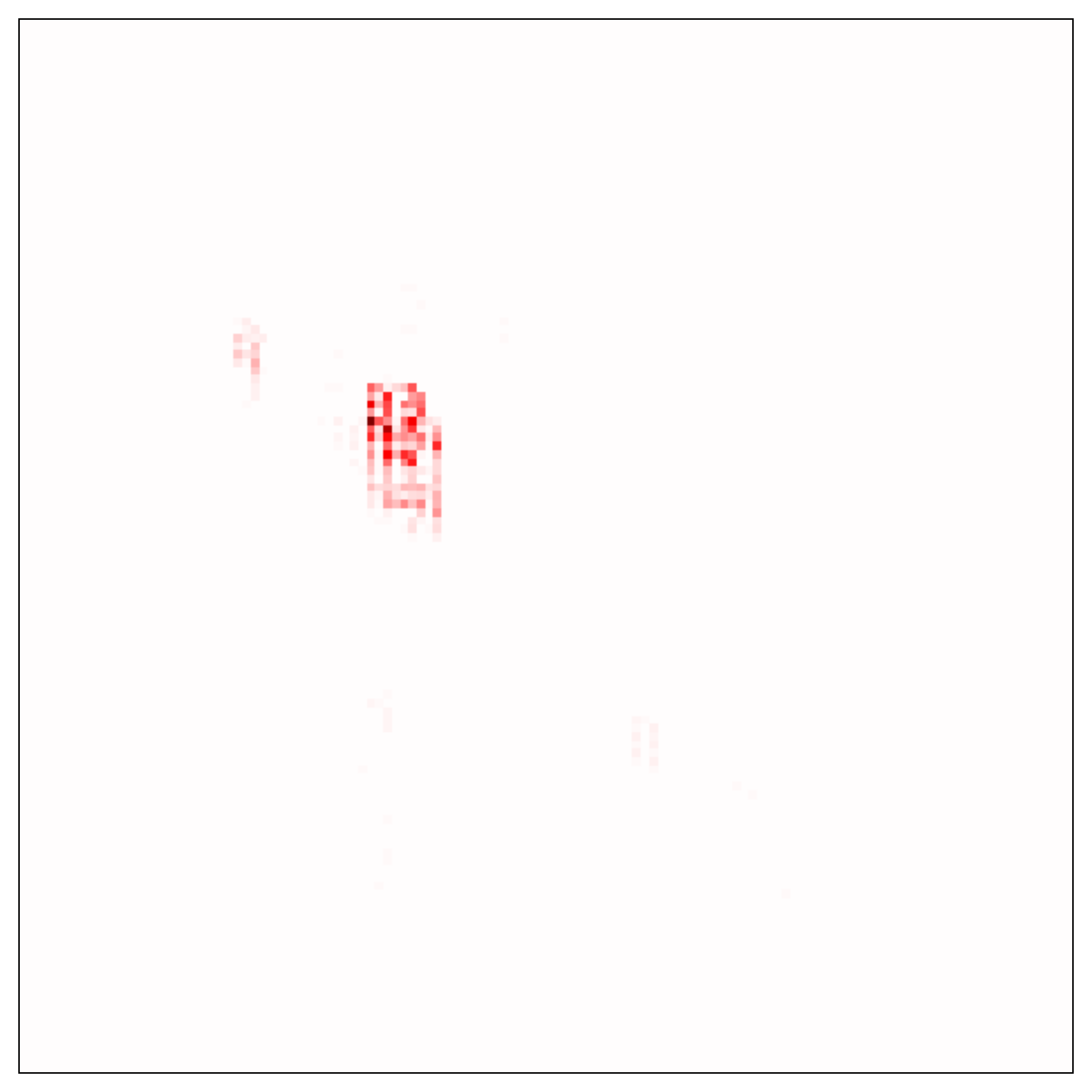} & 0.0 \\
SmoothGrad \cite{Smilkov:ICML2017}                  & \includegraphics[width=.12\linewidth,valign=m]{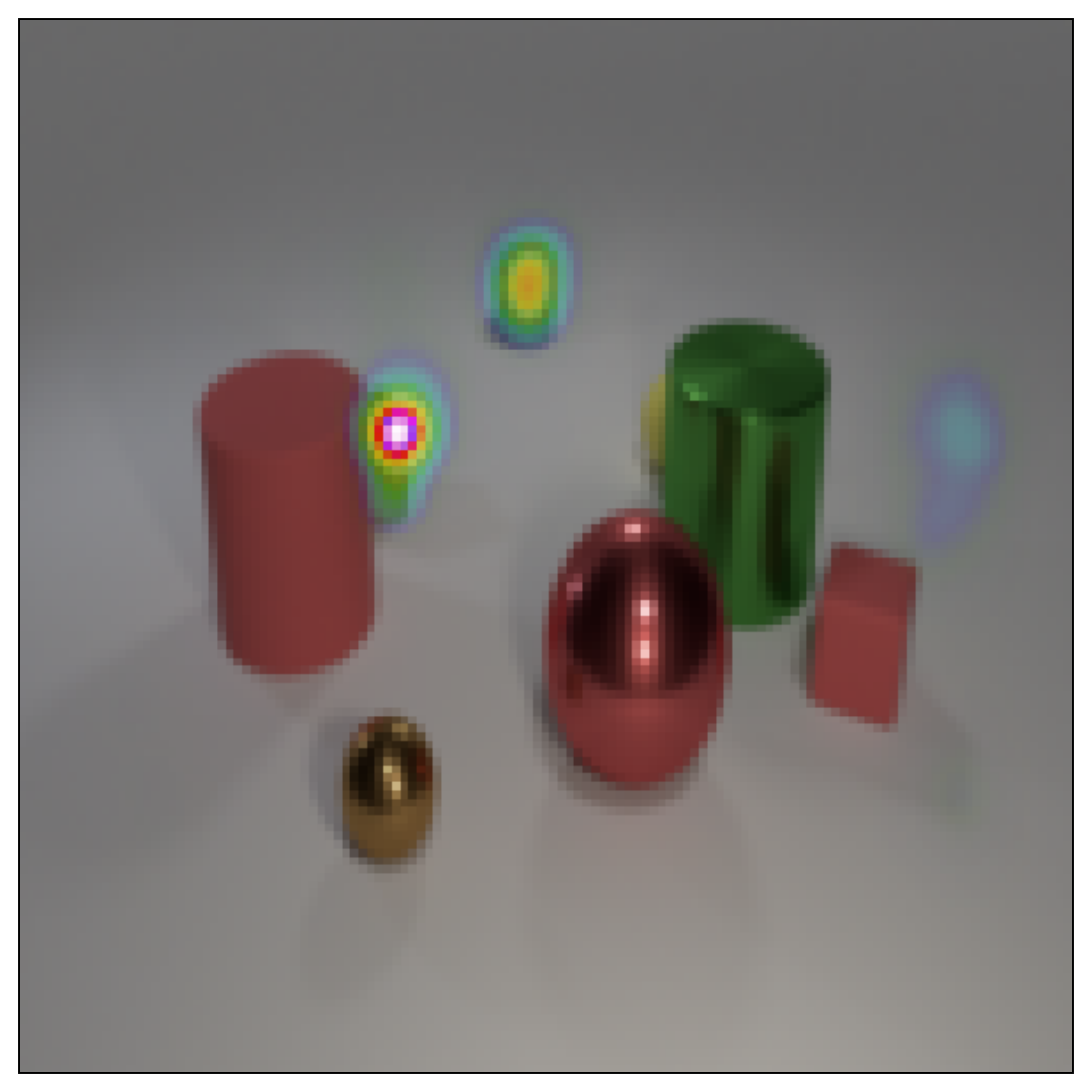} & \includegraphics[width=.12\linewidth,valign=m]{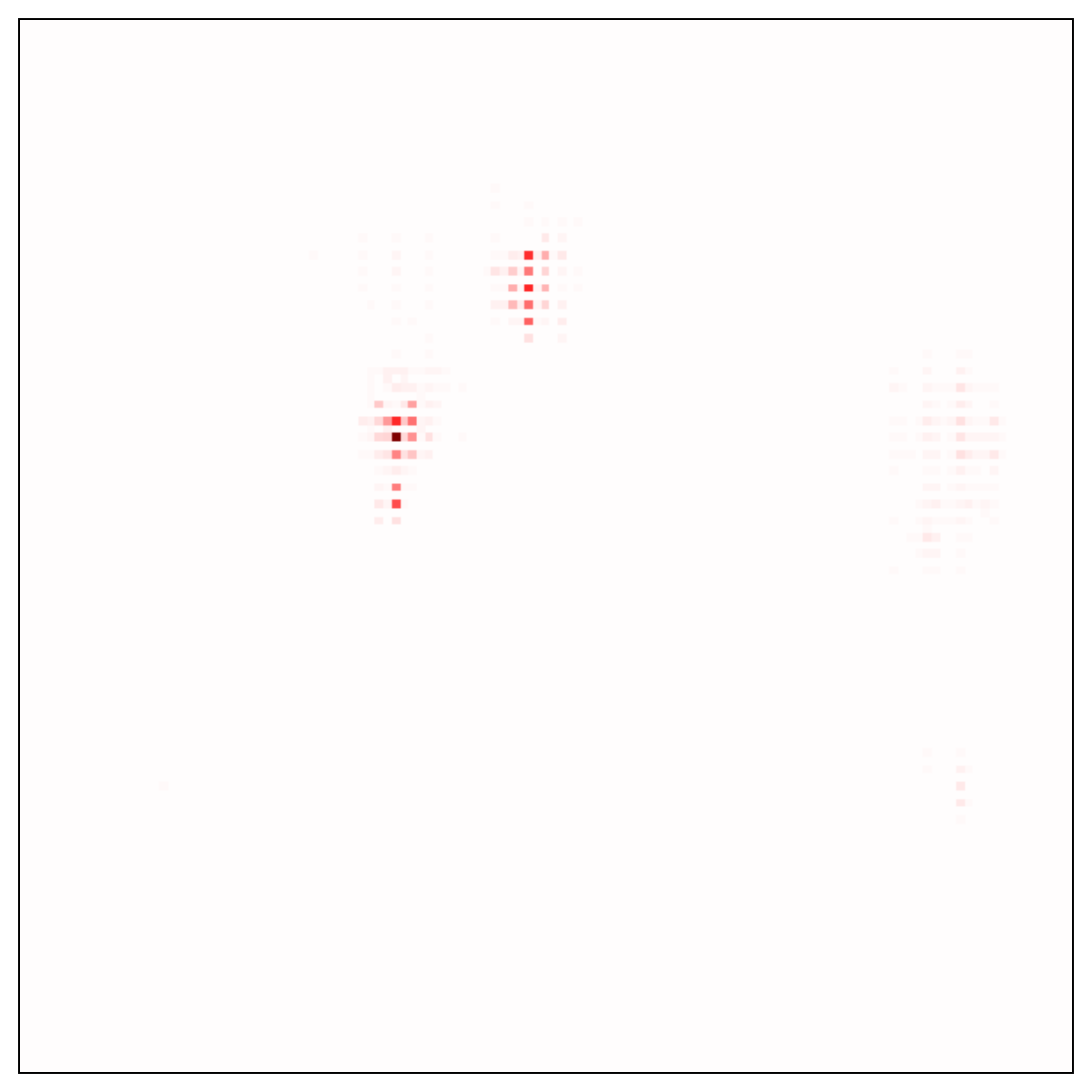} & 0.0 \\
VarGrad \cite{Adebayo:ICLR2018}                     & \includegraphics[width=.12\linewidth,valign=m]{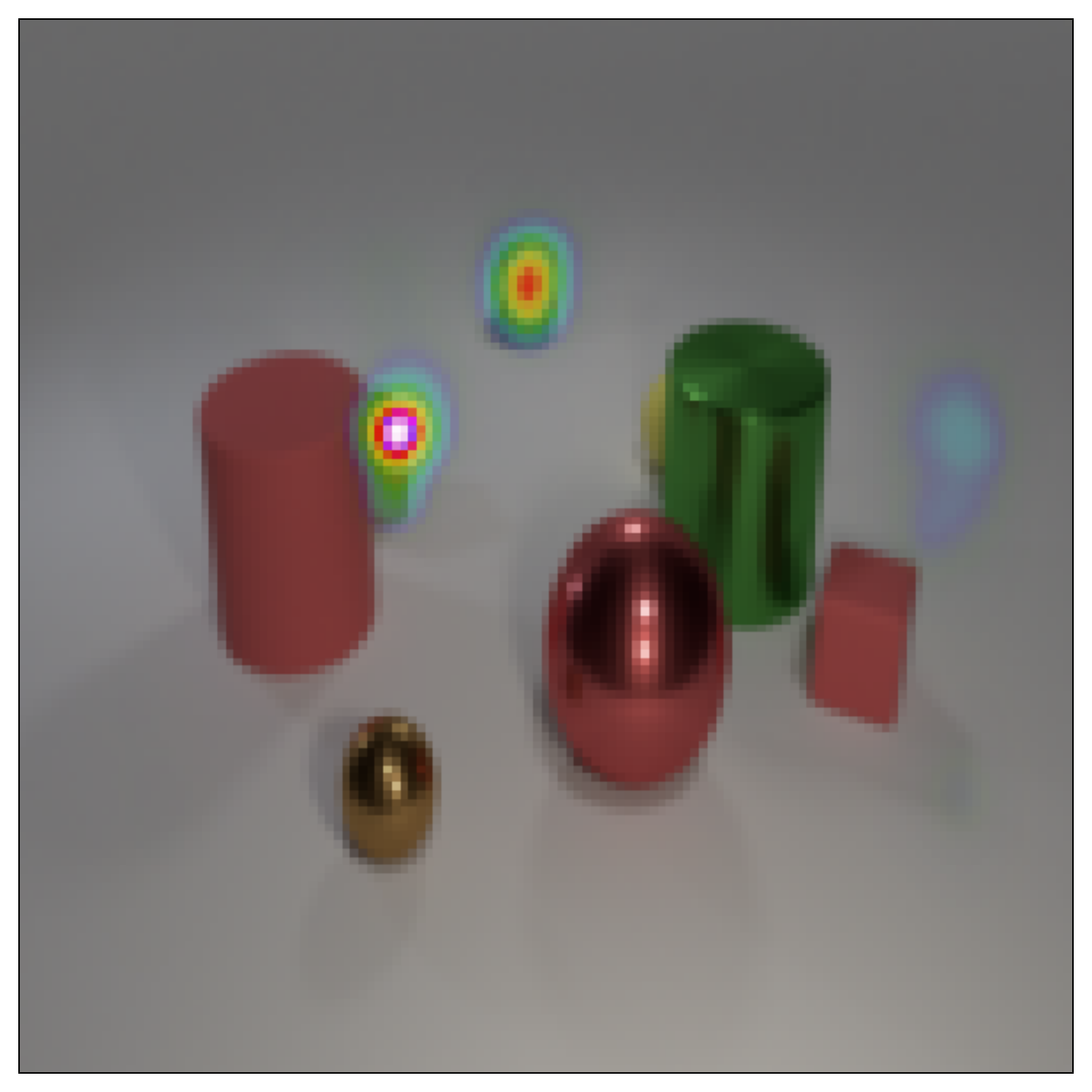} & \includegraphics[width=.12\linewidth,valign=m]{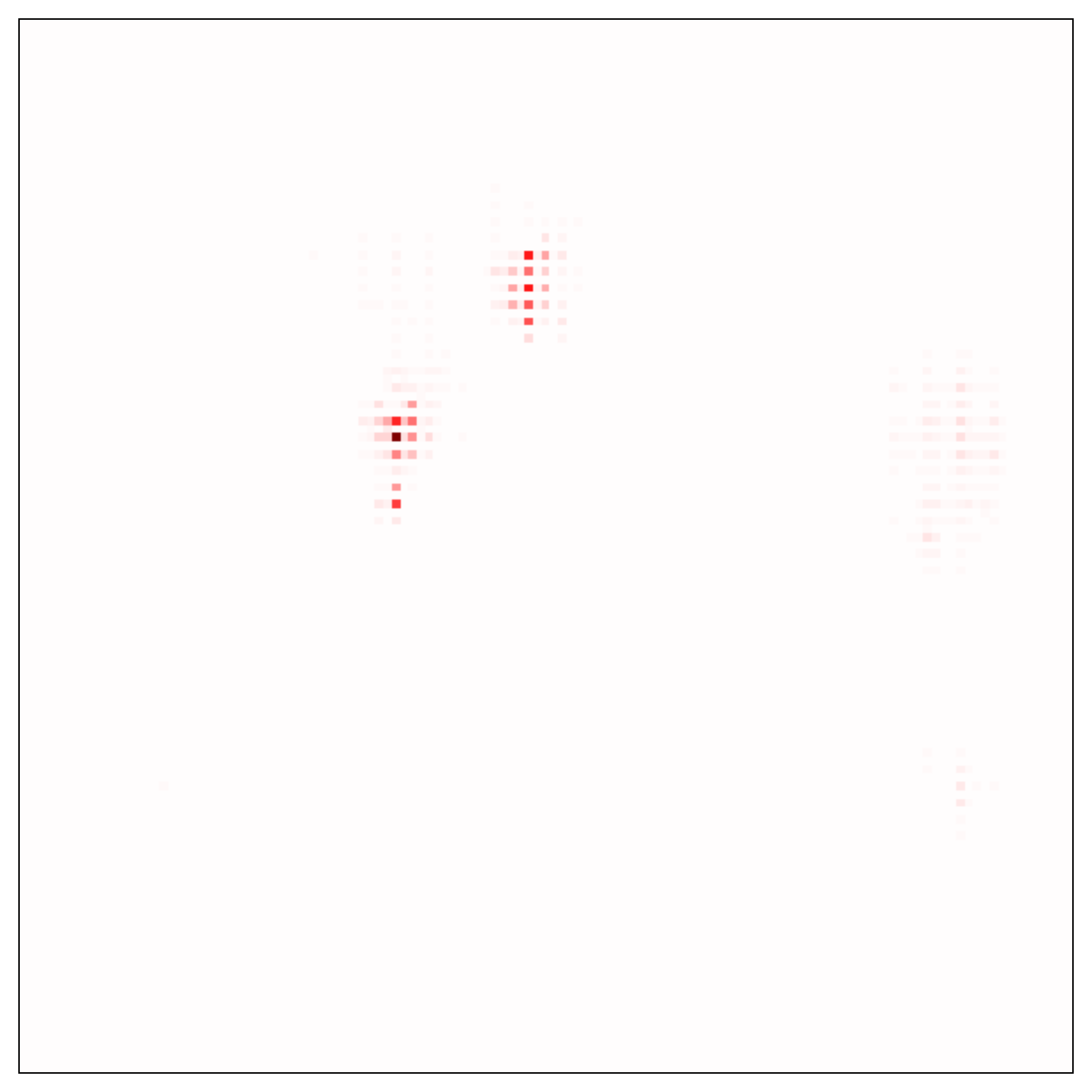} & 0.0 \\
Gradient \cite{Simonyan:ICLR2014}                   & \includegraphics[width=.12\linewidth,valign=m]{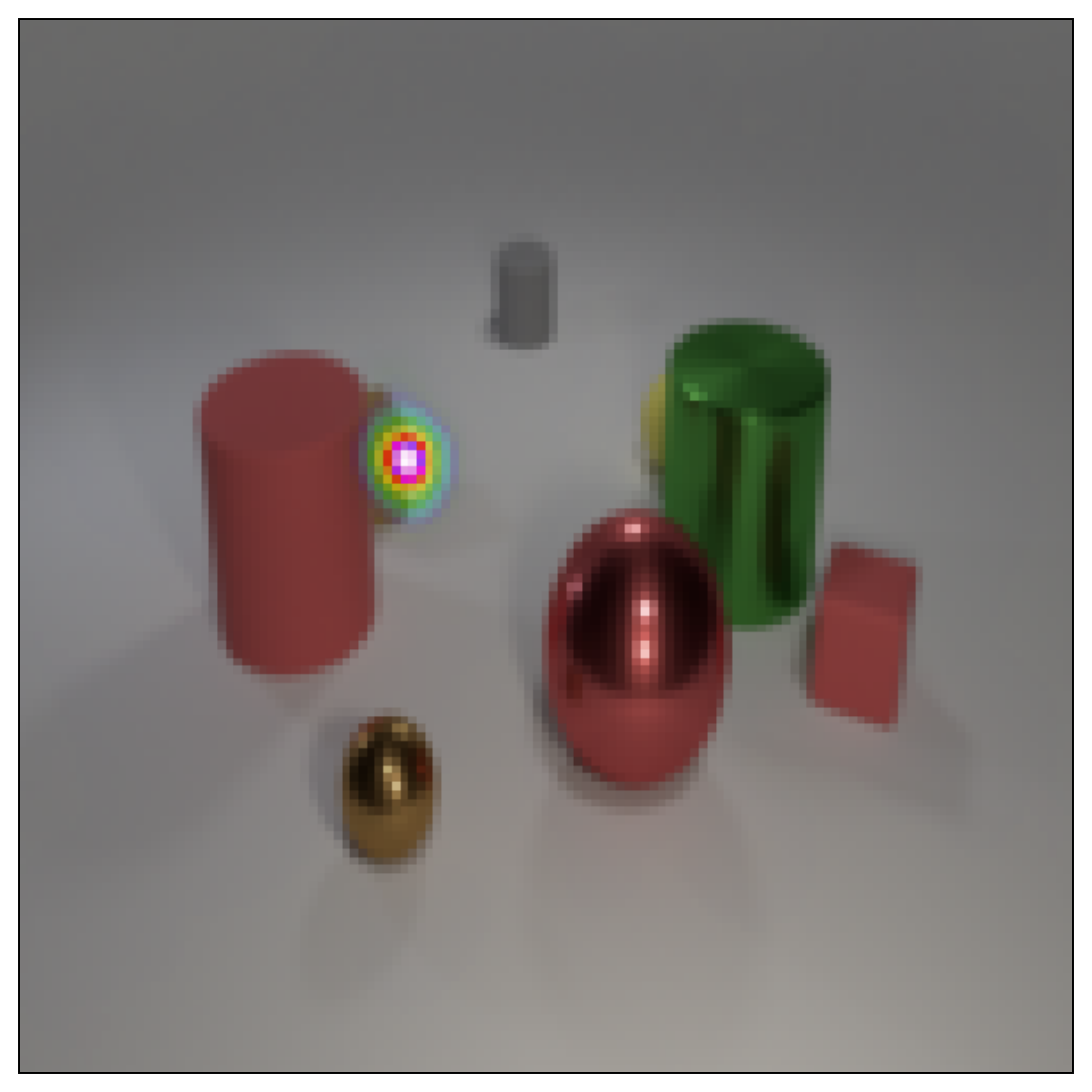} & \includegraphics[width=.12\linewidth,valign=m]{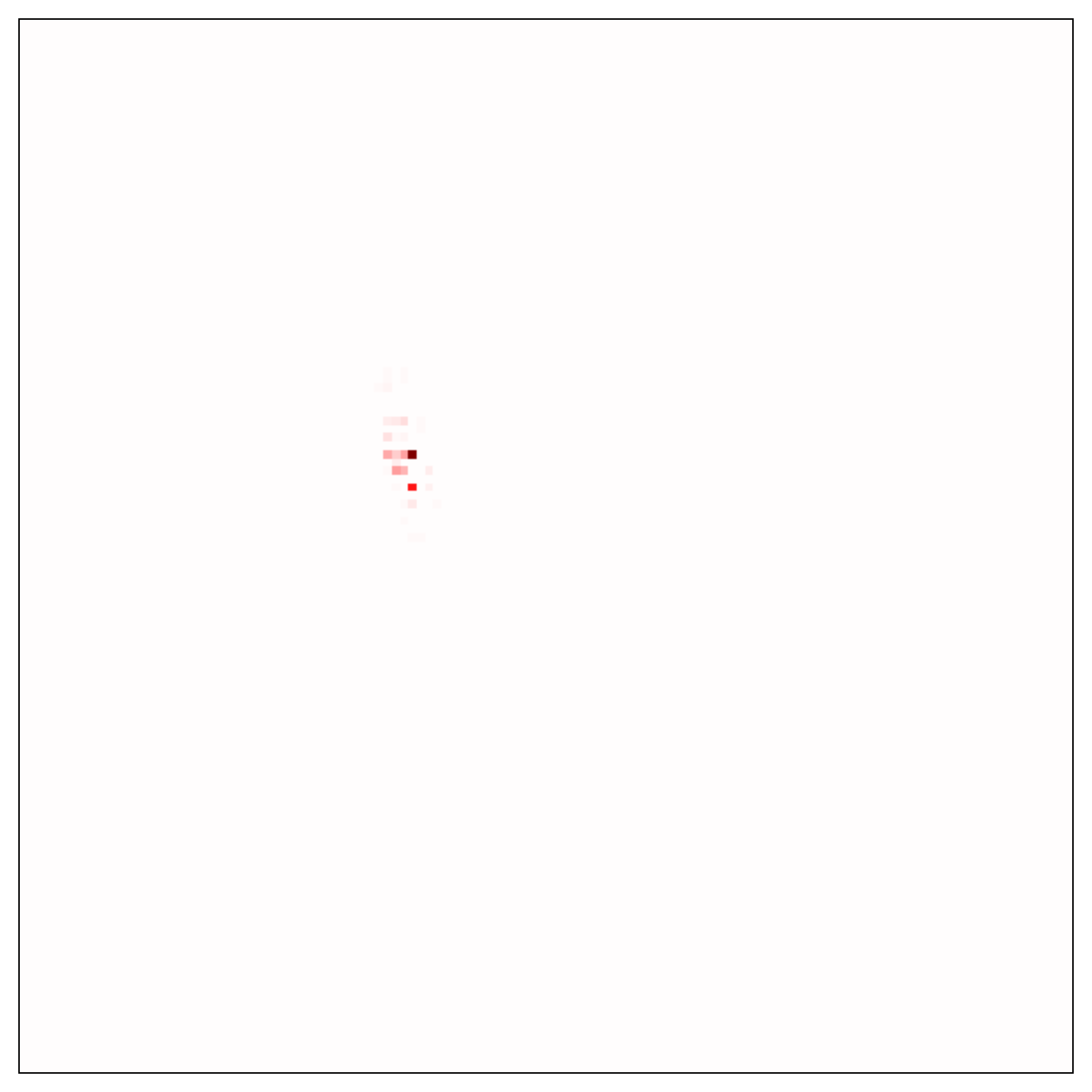} & 0.0 \\
Gradient$\times$Input \cite{Shrikumar:arxiv2016}    & \includegraphics[width=.12\linewidth,valign=m]{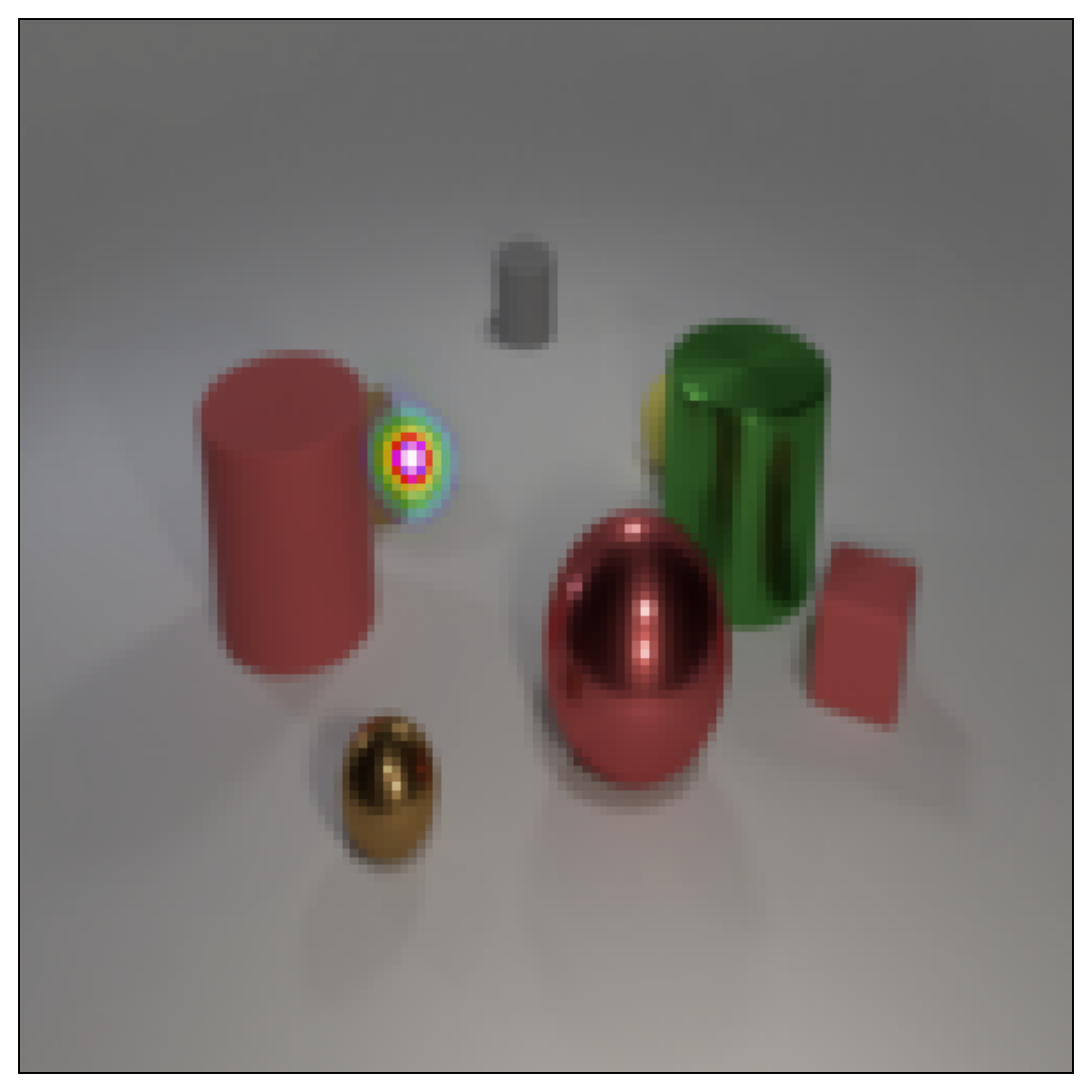} & \includegraphics[width=.12\linewidth,valign=m]{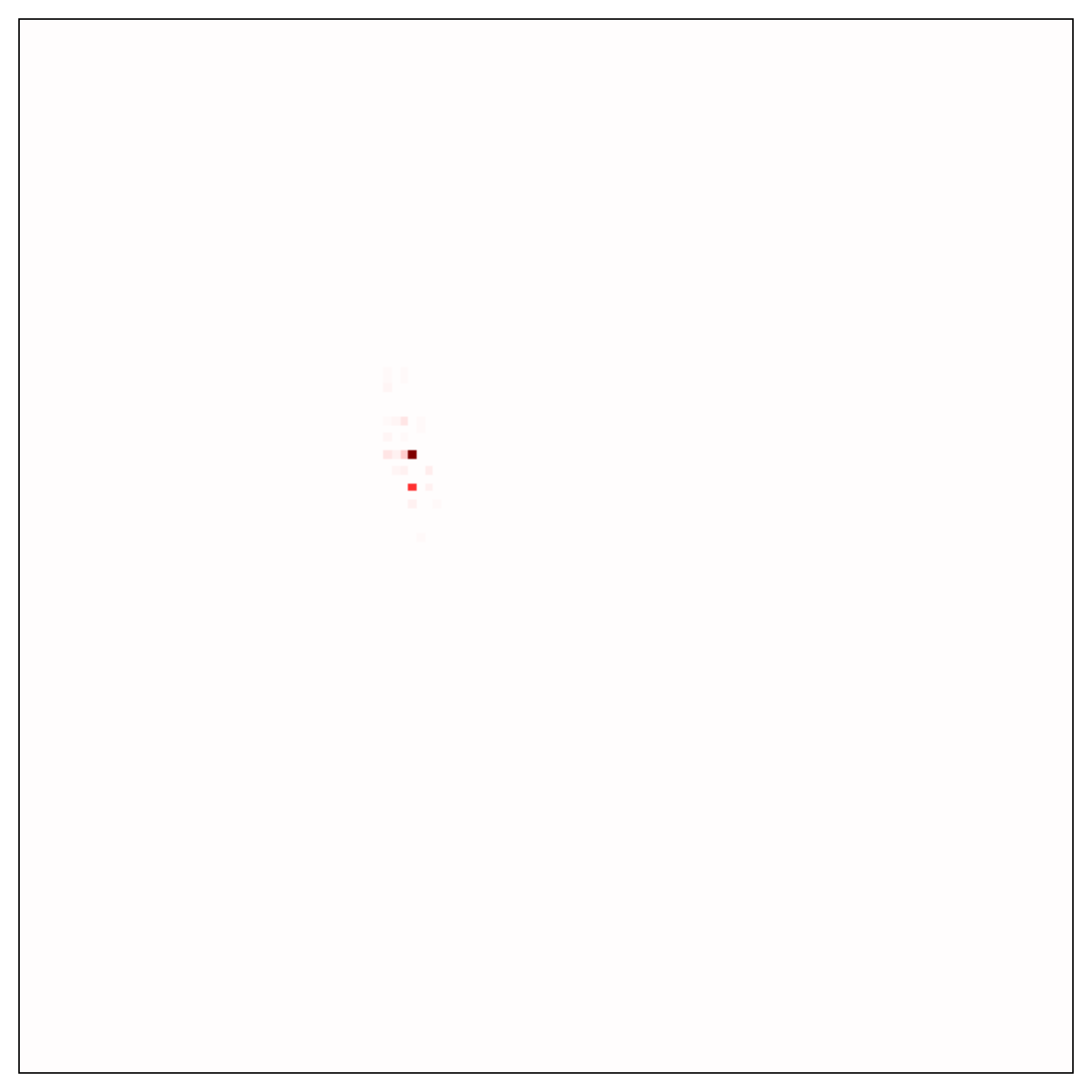} & 0.0 \\
Deconvnet \cite{Zeiler:ECCV2014}                    & \includegraphics[width=.12\linewidth,valign=m]{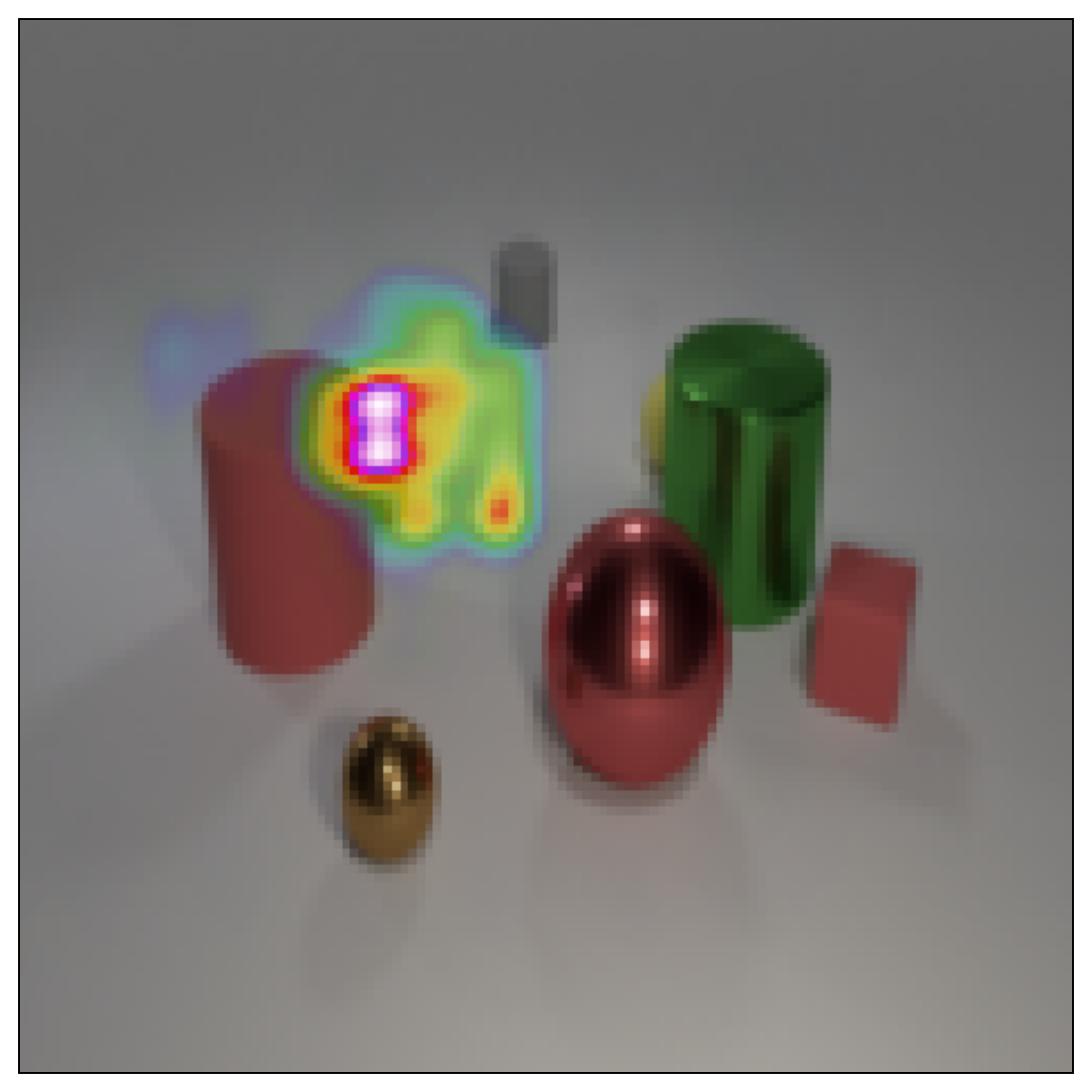} & \includegraphics[width=.12\linewidth,valign=m]{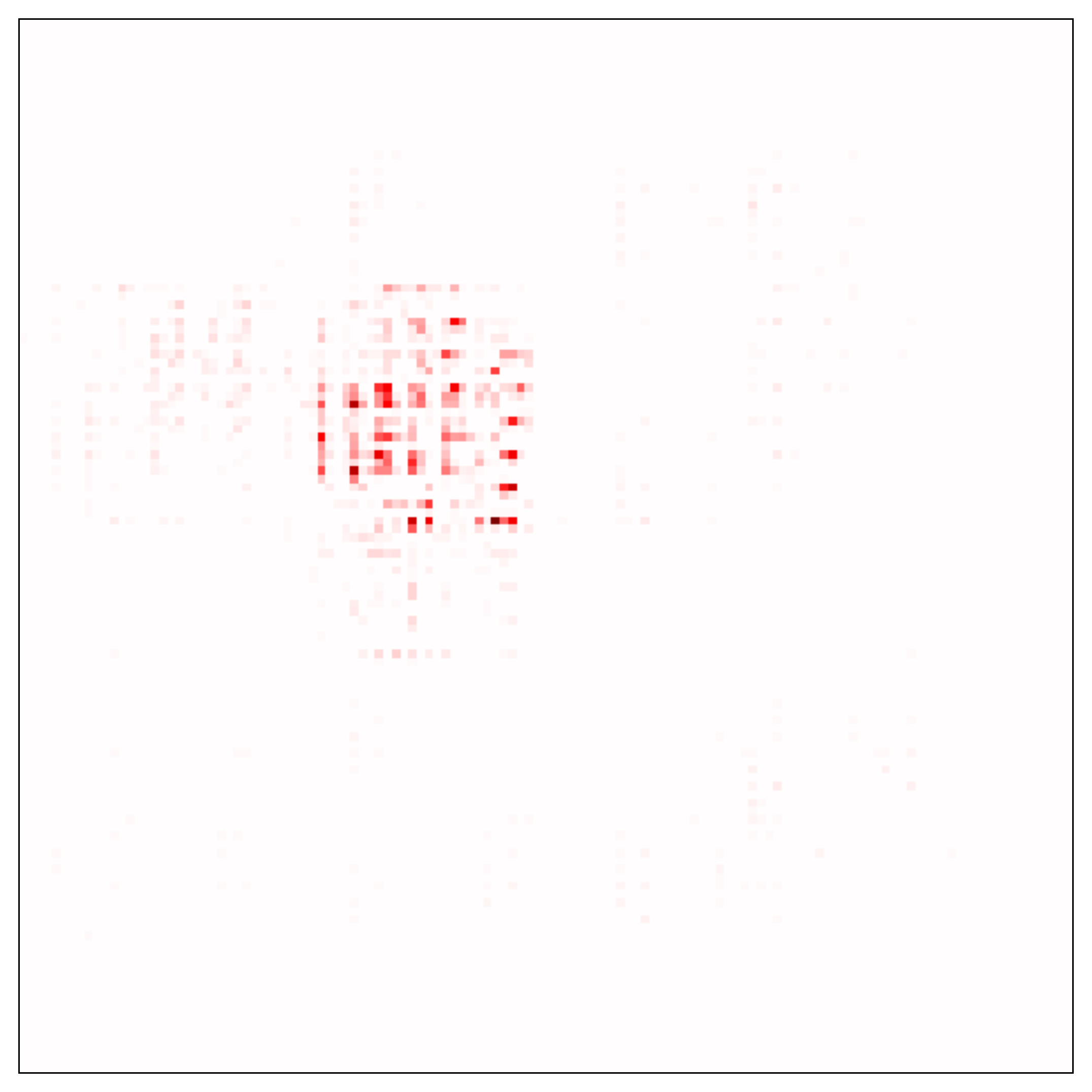} & 0.0 \\
Grad-CAM \cite{Selvaraju:ICCV2017}                  & \includegraphics[width=.12\linewidth,valign=m]{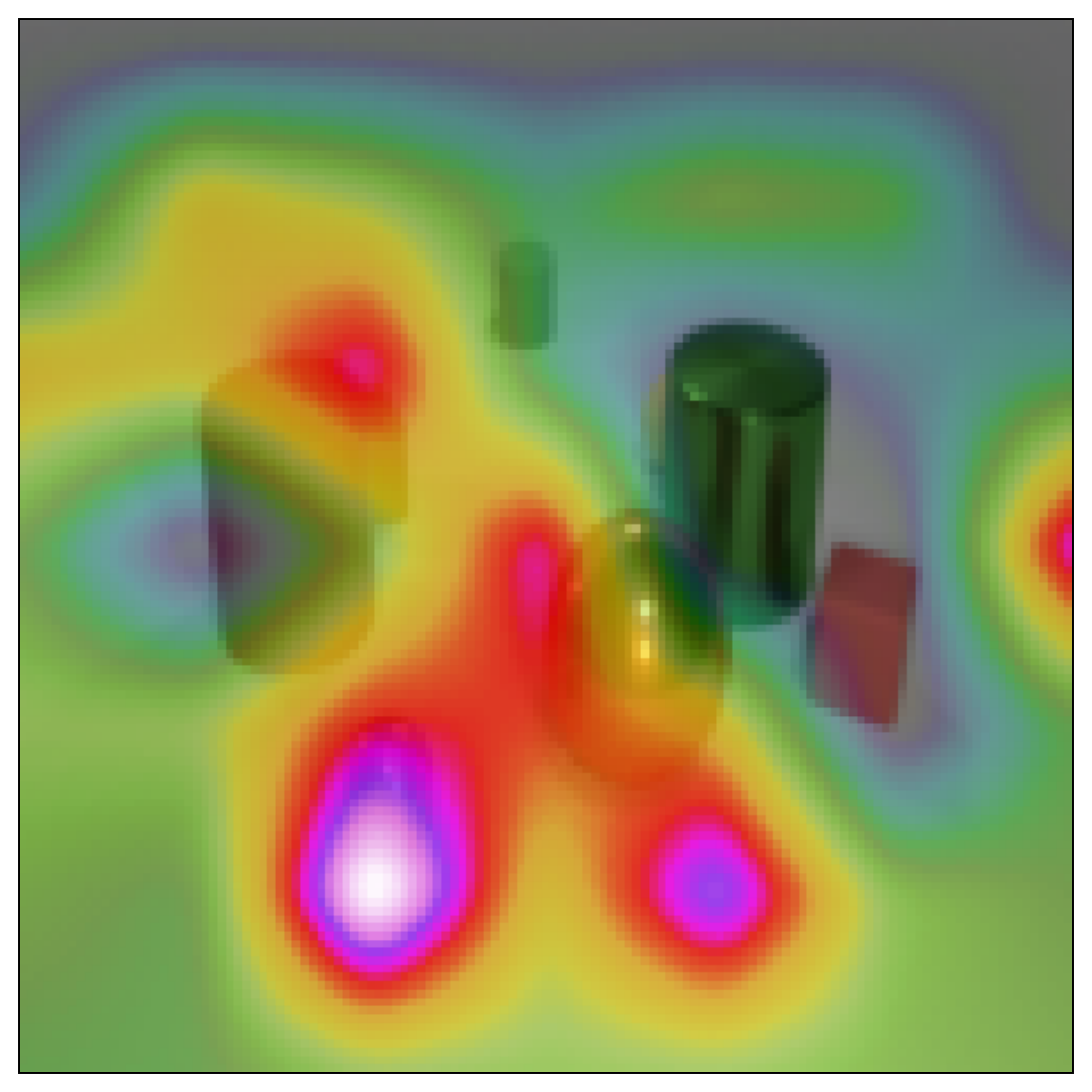} & \includegraphics[width=.12\linewidth,valign=m]{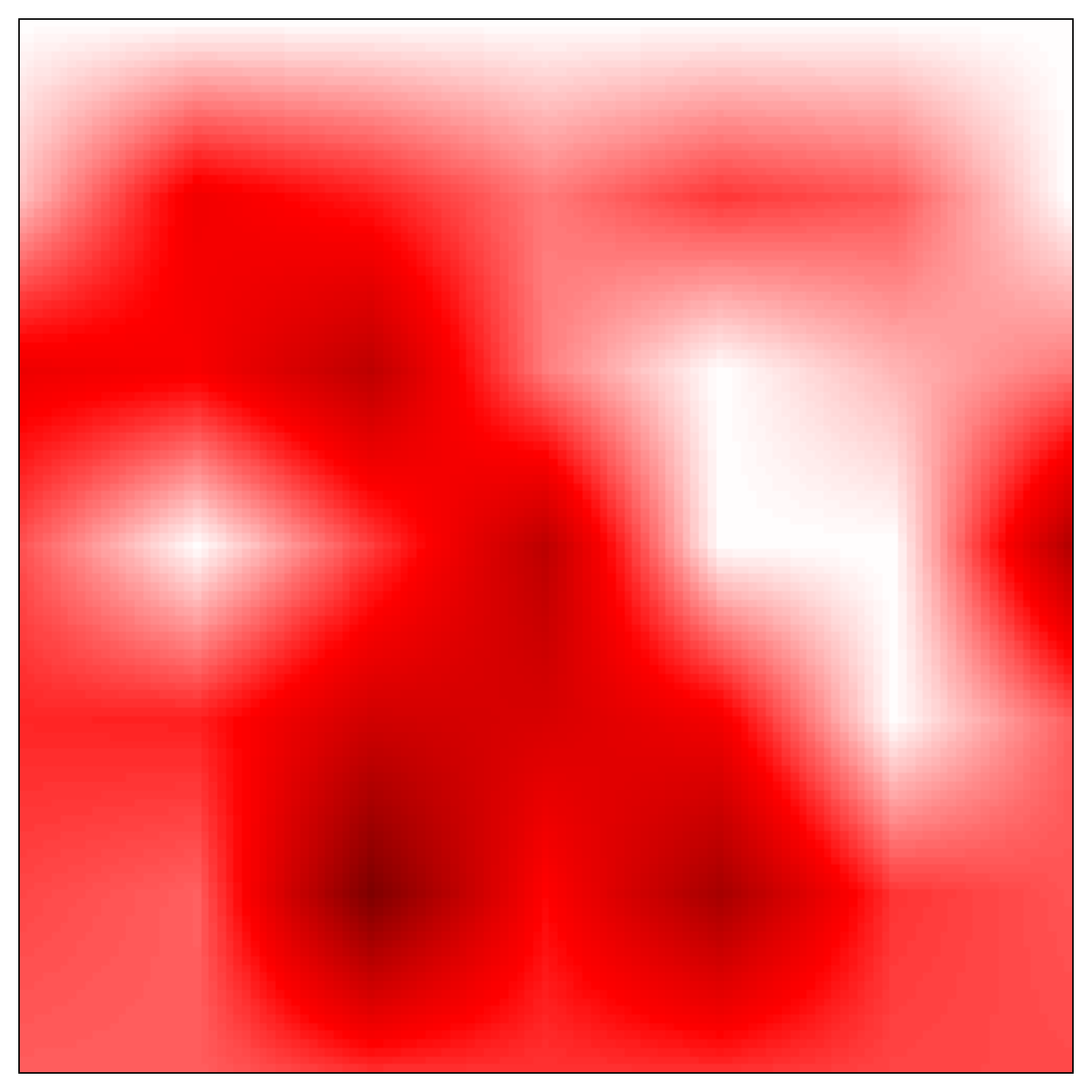} & 0.0 \\
\end{tabular}
\end{table}

\begin{table}
        \scriptsize
		\caption{Heatmaps for a falsely predicted CLEVR-XAI-simple question (raw heatmap and heatmap overlayed with original image), and corresponding relevance \textit{mass} accuracy.}
		\label{table:heatmap-simple-false-22552}
\begin{tabular}{lllc}
\midrule
\begin{tabular}{@{}l@{}}What is the color of \\ the small shiny thing? \\ true: \textit{red} \\ predicted: \textit{brown} \end{tabular}  & \includegraphics[width=.18\linewidth,valign=m]{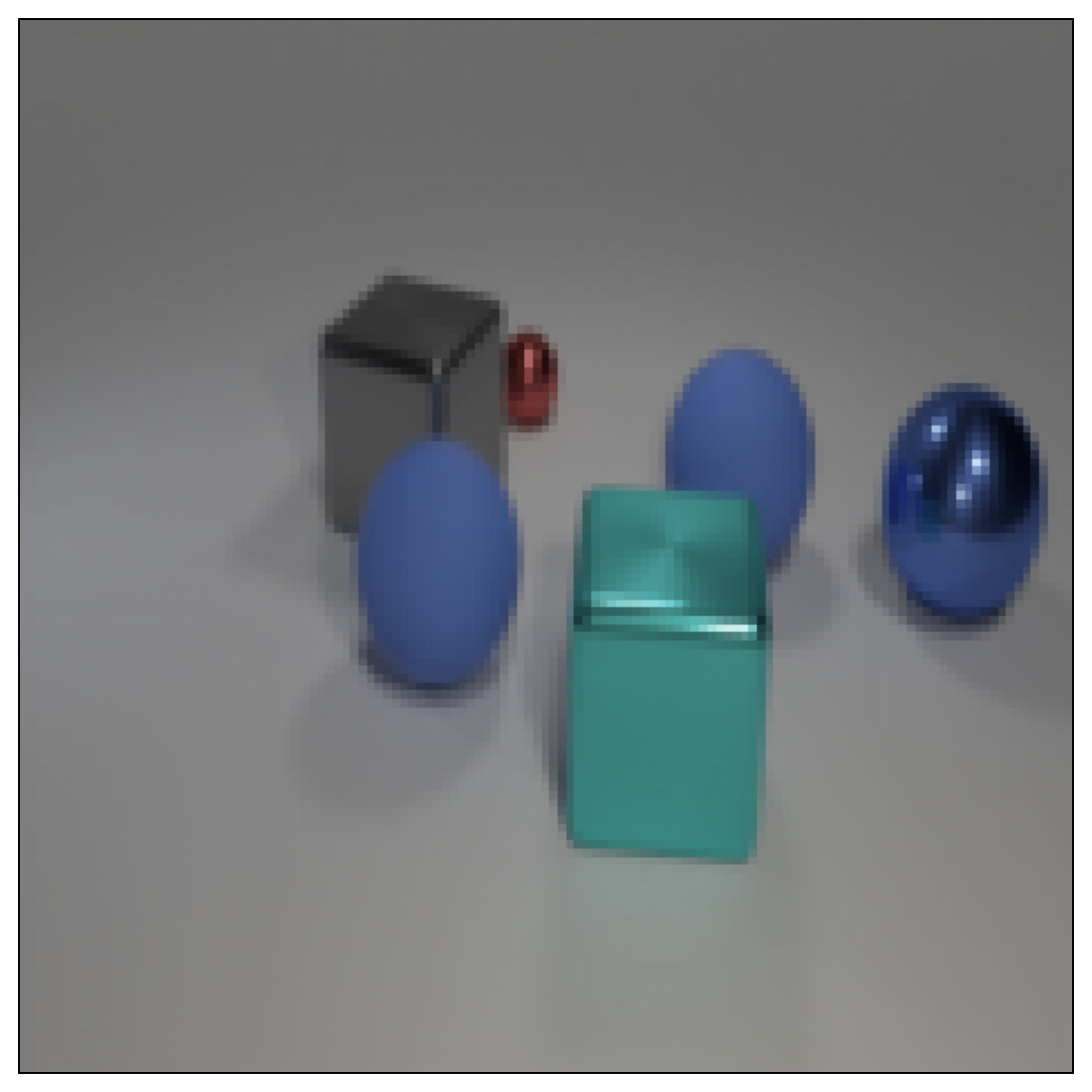} &
\includegraphics[width=.18\linewidth,valign=m]{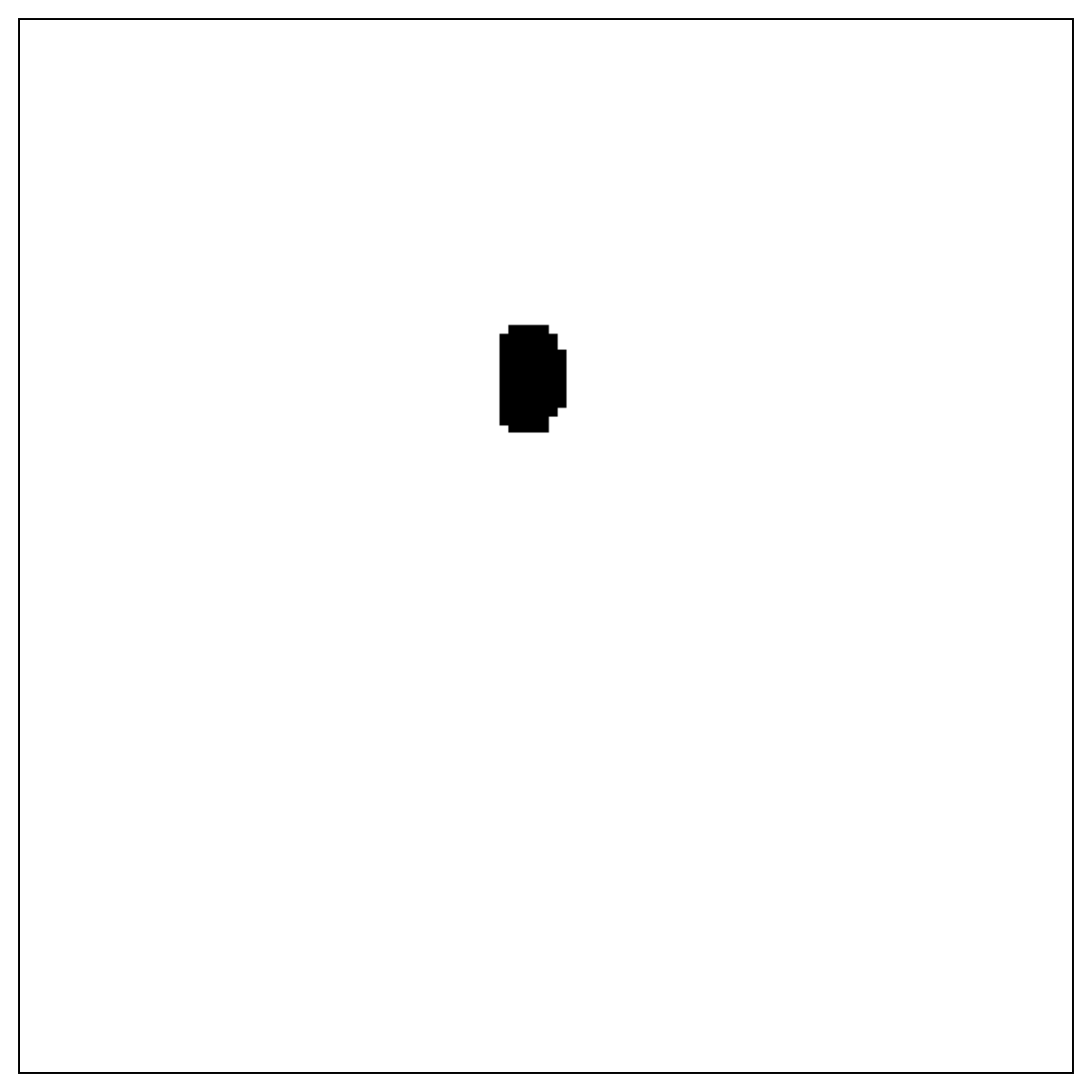} & GT Single Object \\
\midrule
LRP \cite{Bach:PLOS2015}                            & \includegraphics[width=.12\linewidth,valign=m]{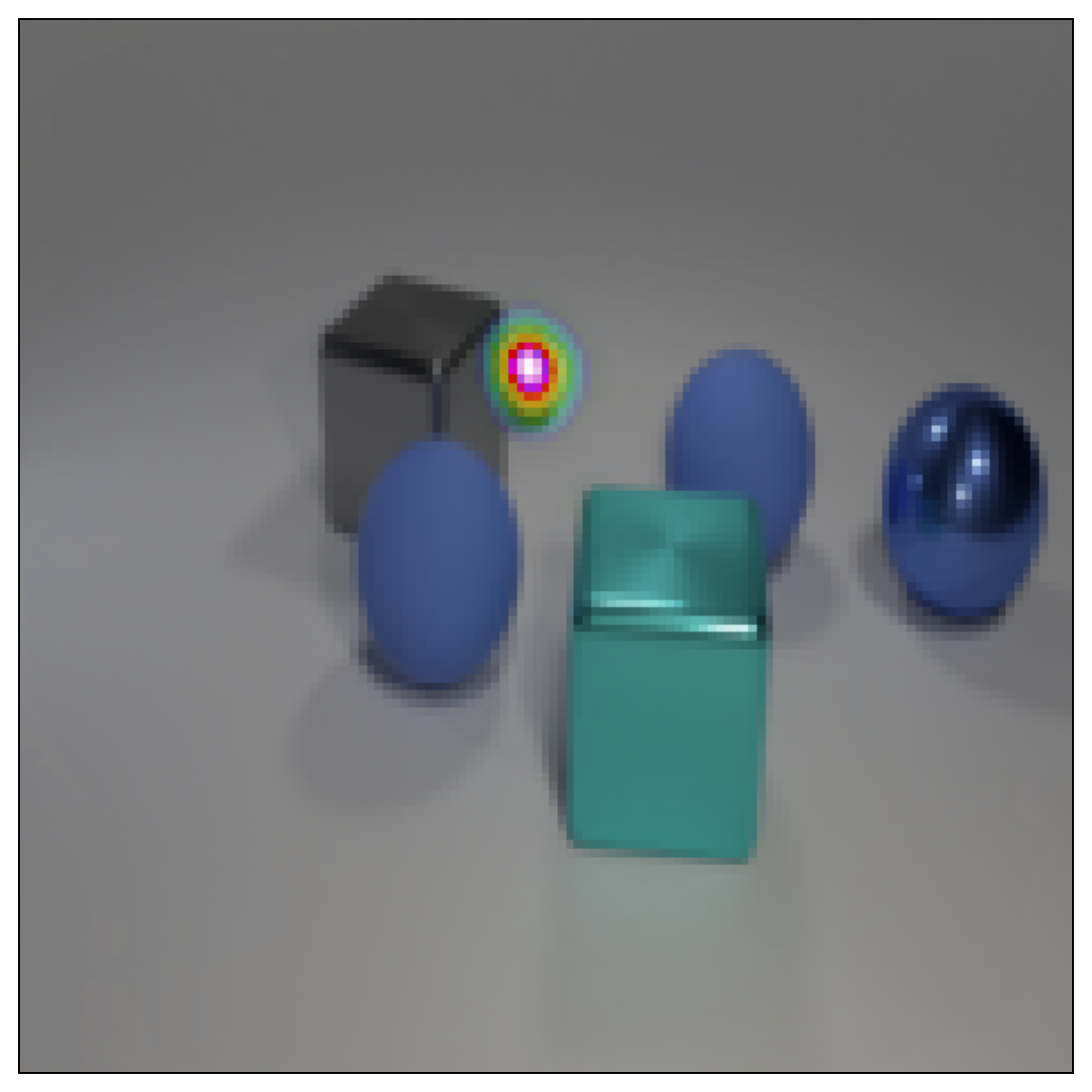} & \includegraphics[width=.12\linewidth,valign=m]{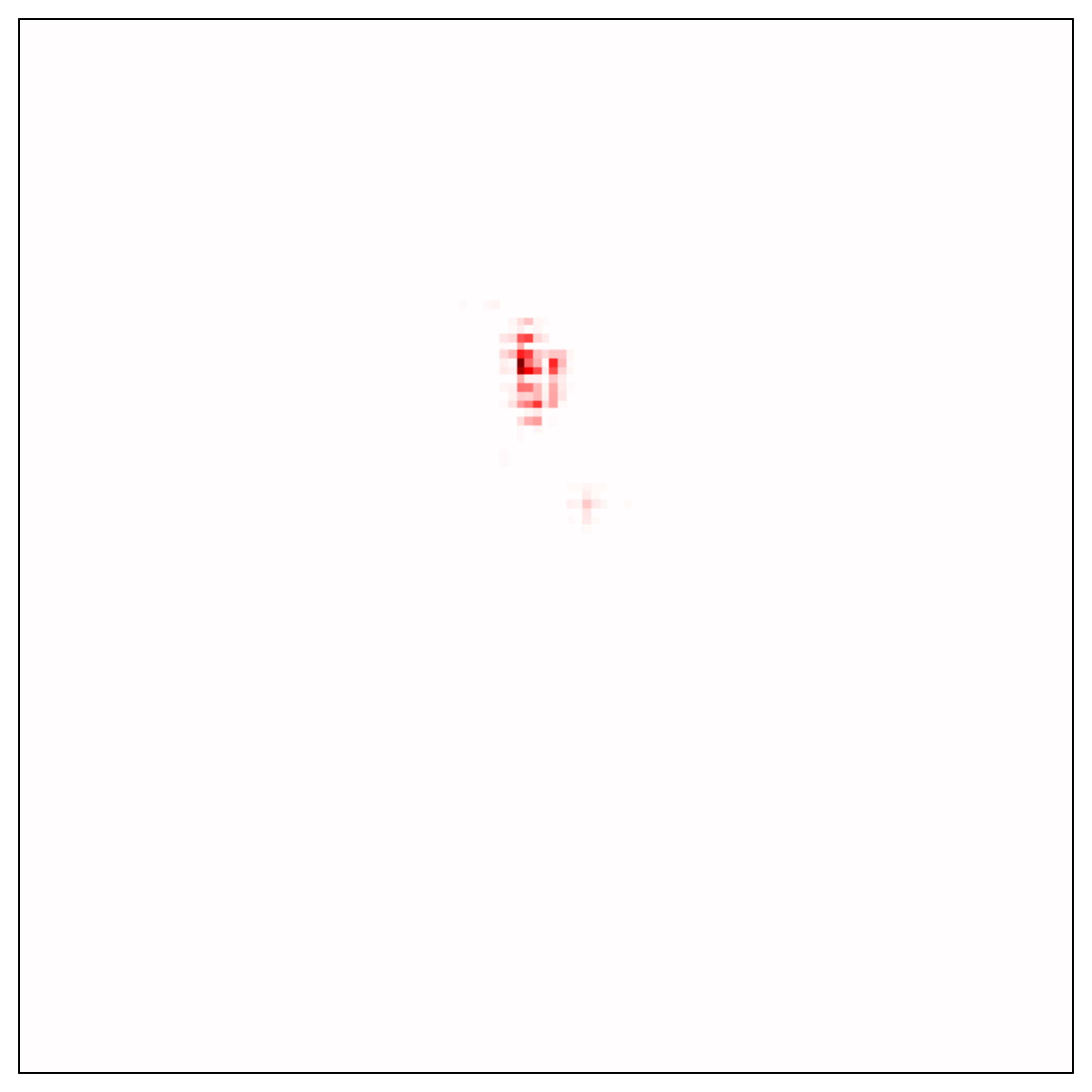} & 0.88 \\
Excitation Backprop \cite{Zhang:ECCV2016}           & \includegraphics[width=.12\linewidth,valign=m]{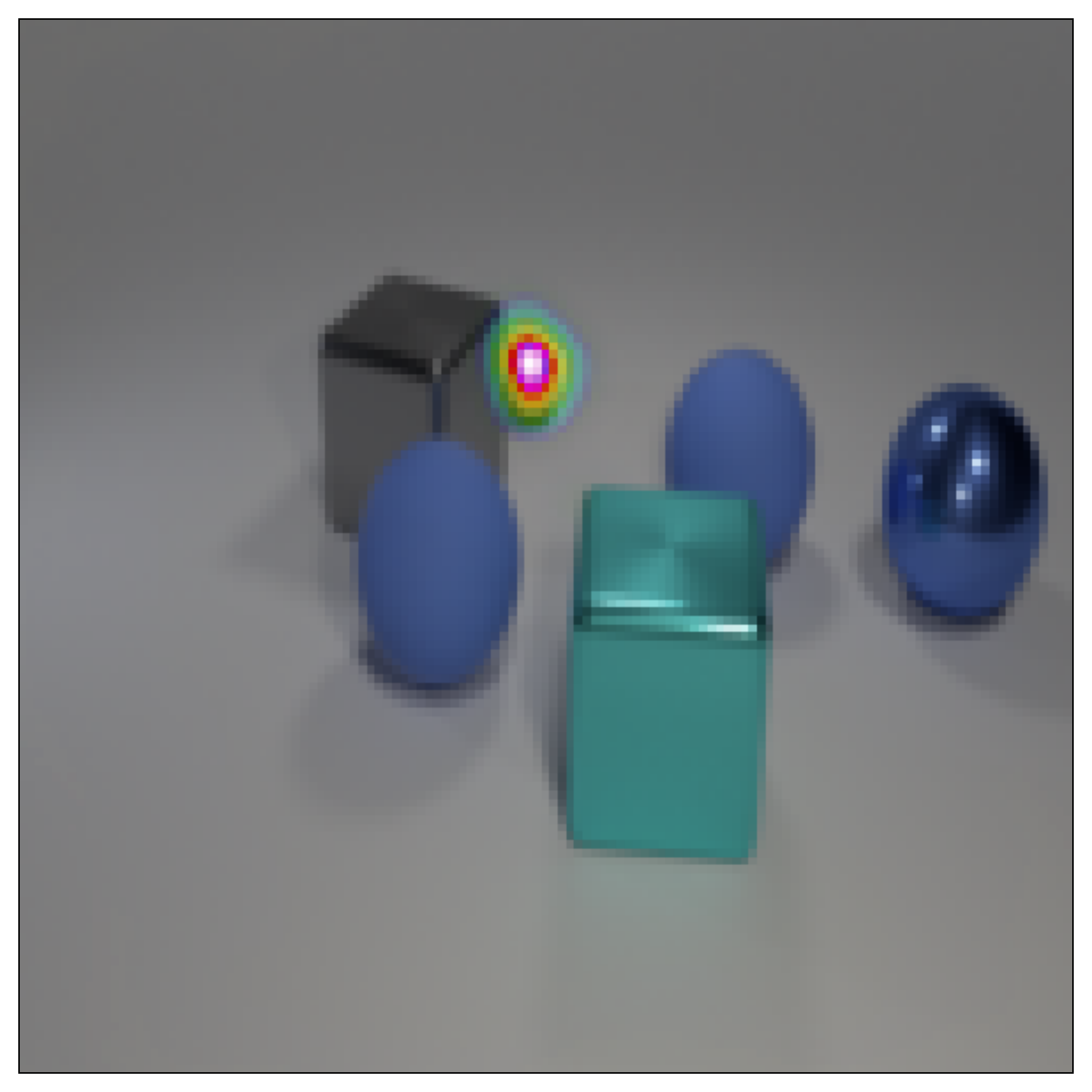} & \includegraphics[width=.12\linewidth,valign=m]{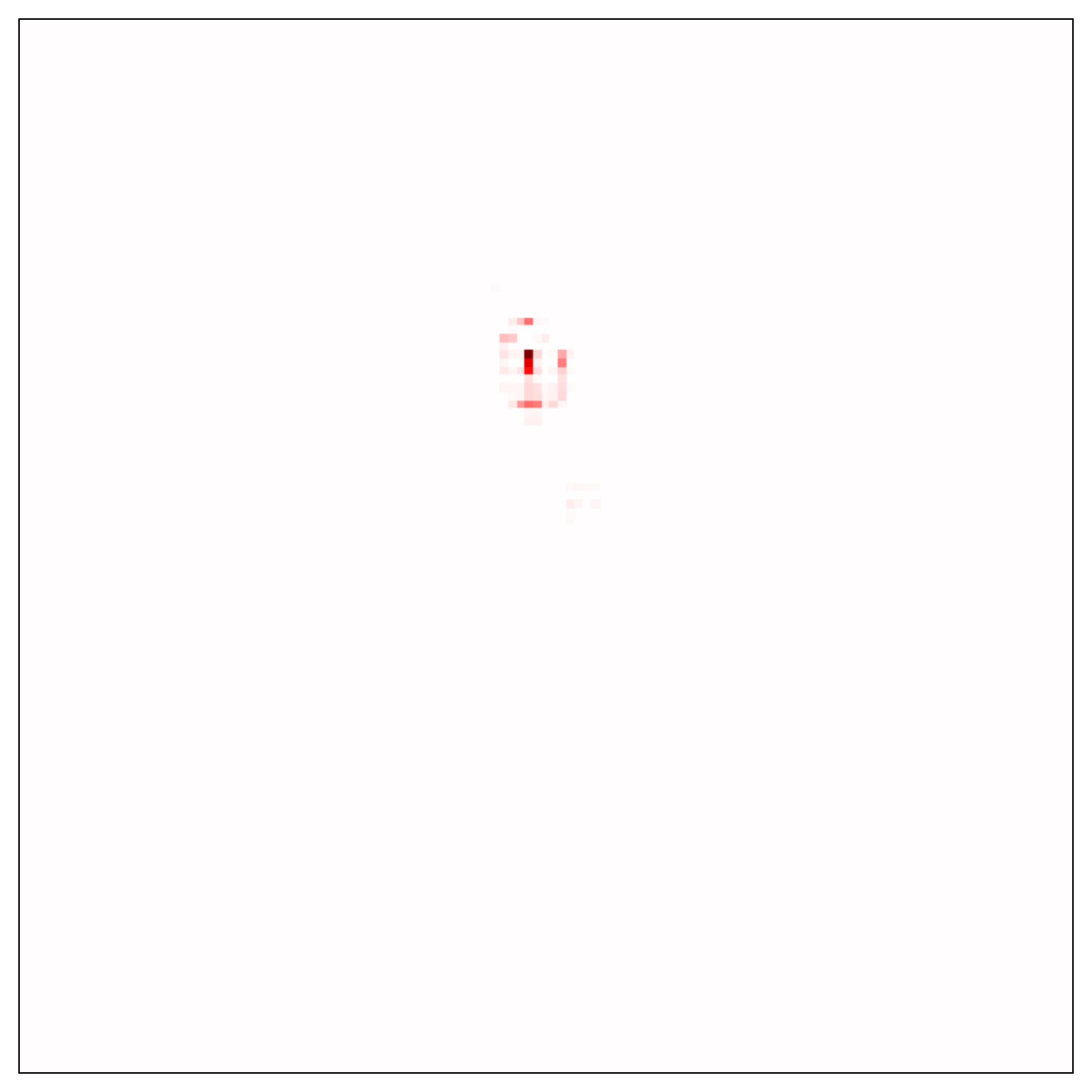} & 0.83 \\
IG \cite{Sundararajan:ICML2017}                     & \includegraphics[width=.12\linewidth,valign=m]{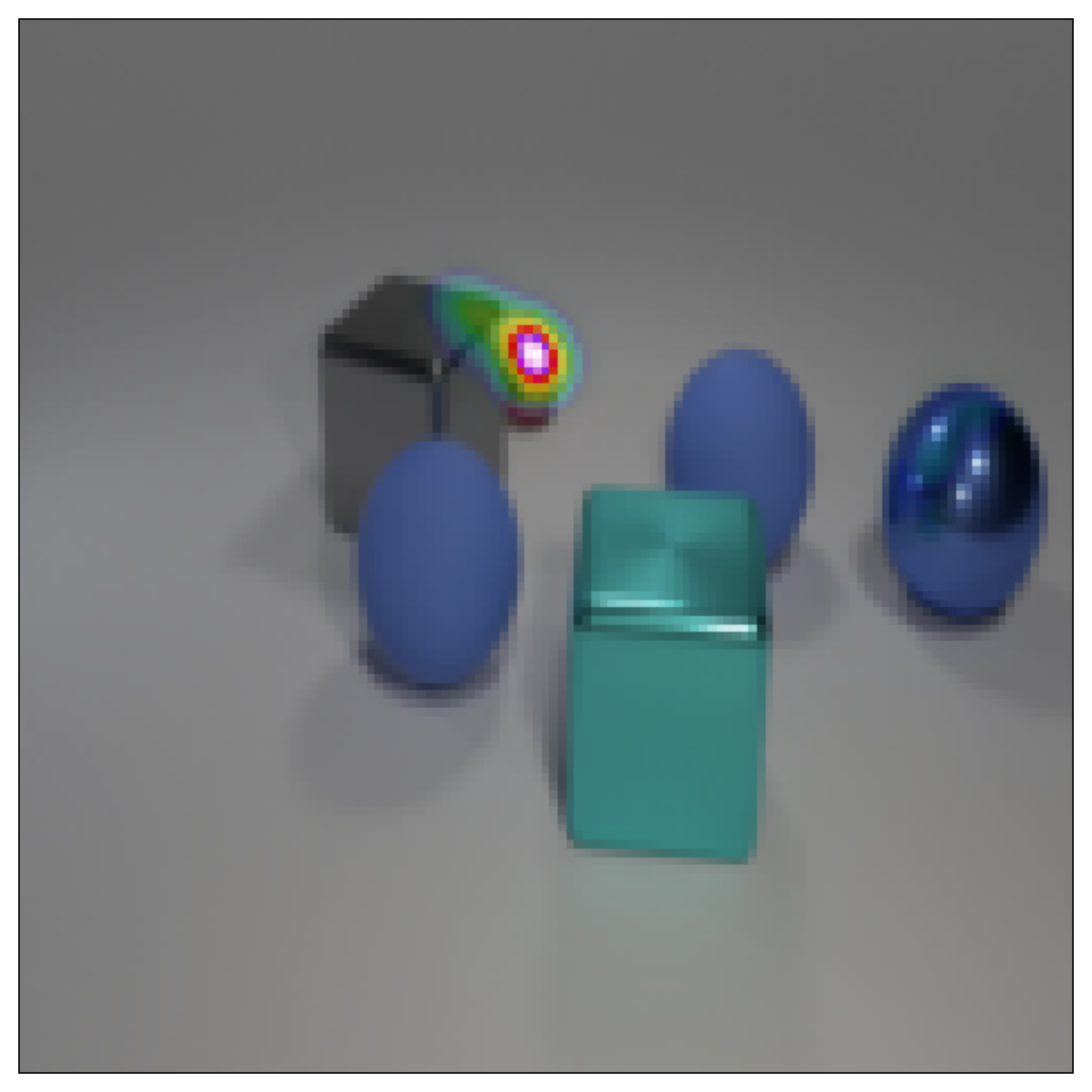} & \includegraphics[width=.12\linewidth,valign=m]{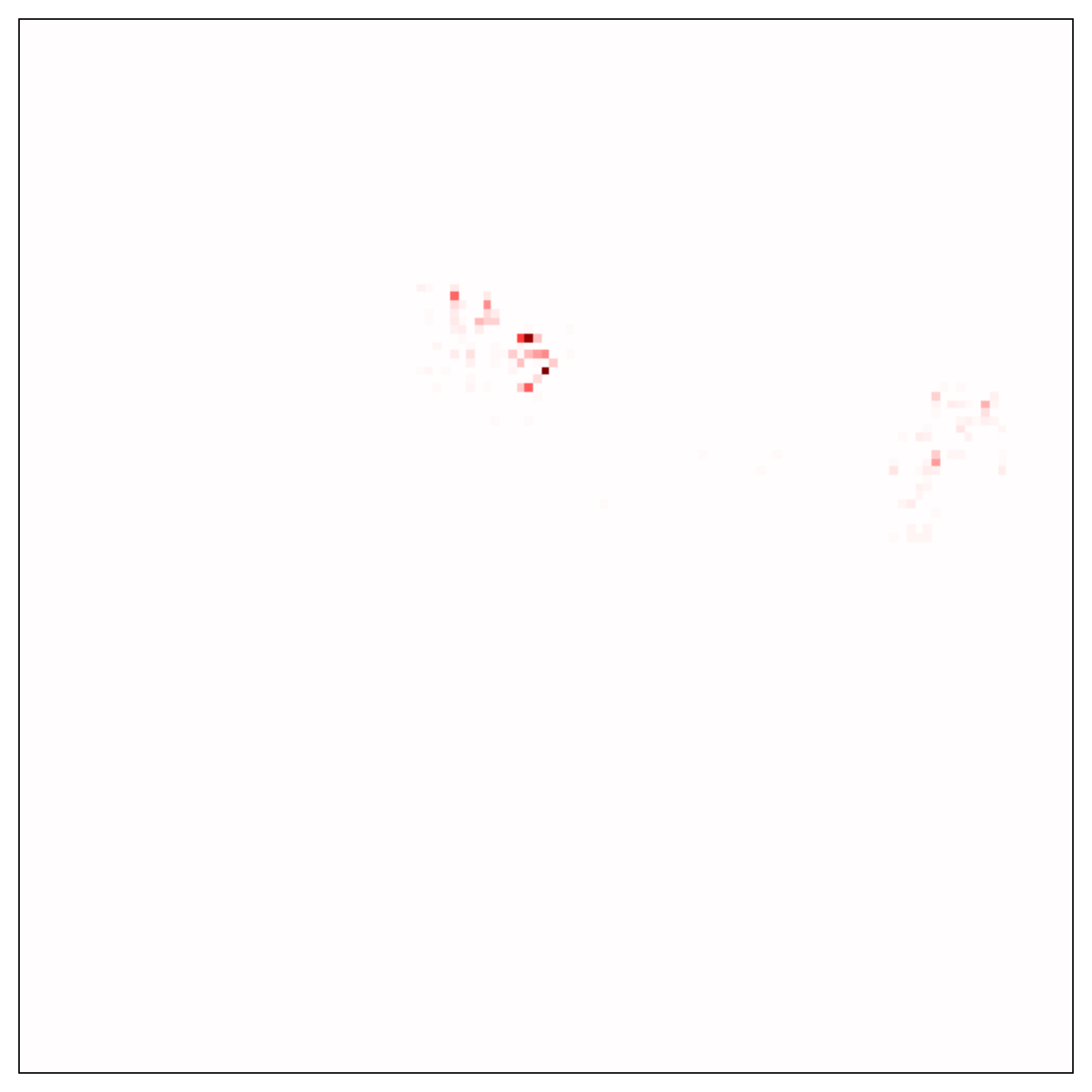} & 0.48 \\
Guided Backprop \cite{Spring:ICLR2015}              & \includegraphics[width=.12\linewidth,valign=m]{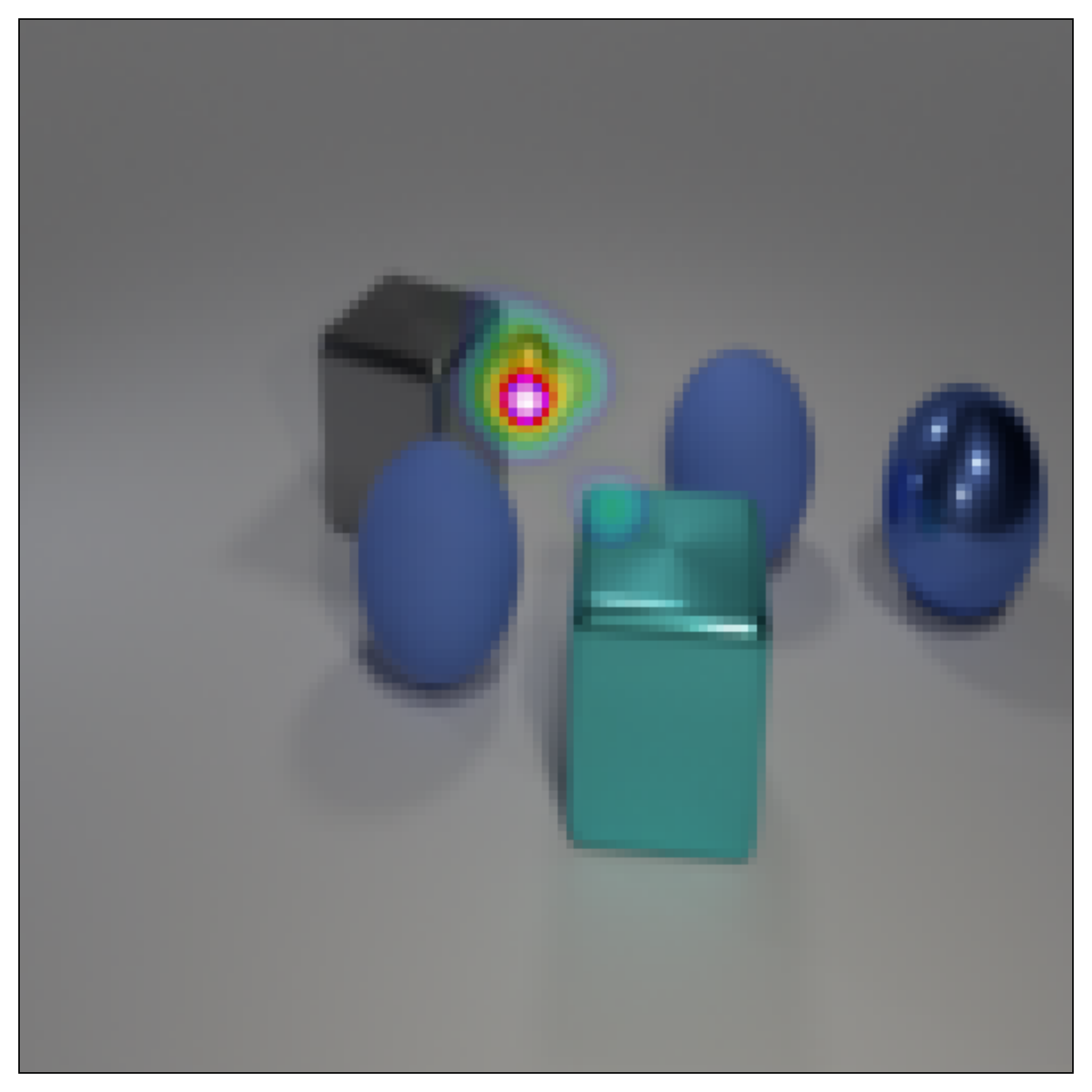} & \includegraphics[width=.12\linewidth,valign=m]{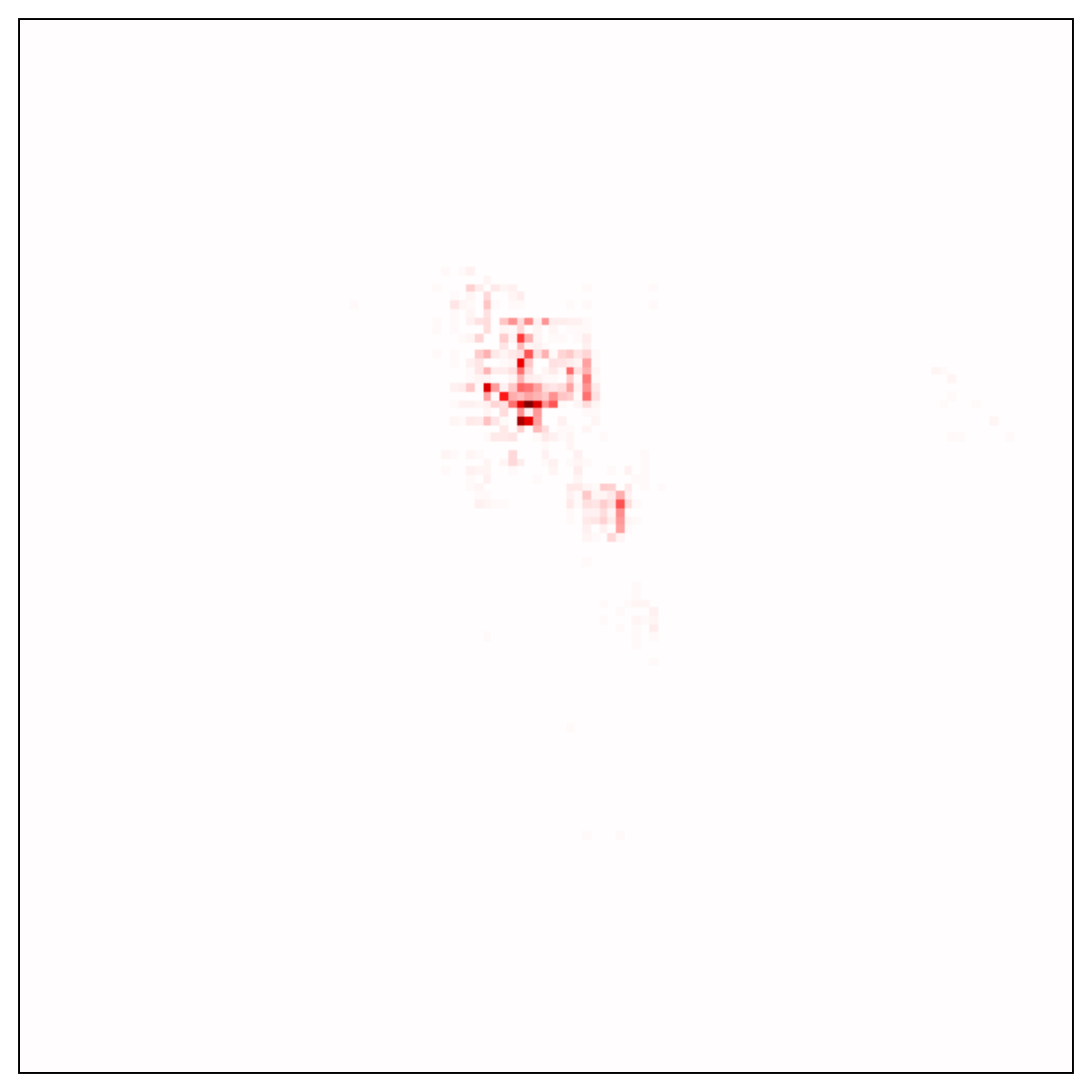} & 0.44 \\
Guided Grad-CAM \cite{Selvaraju:ICCV2017}           & \includegraphics[width=.12\linewidth,valign=m]{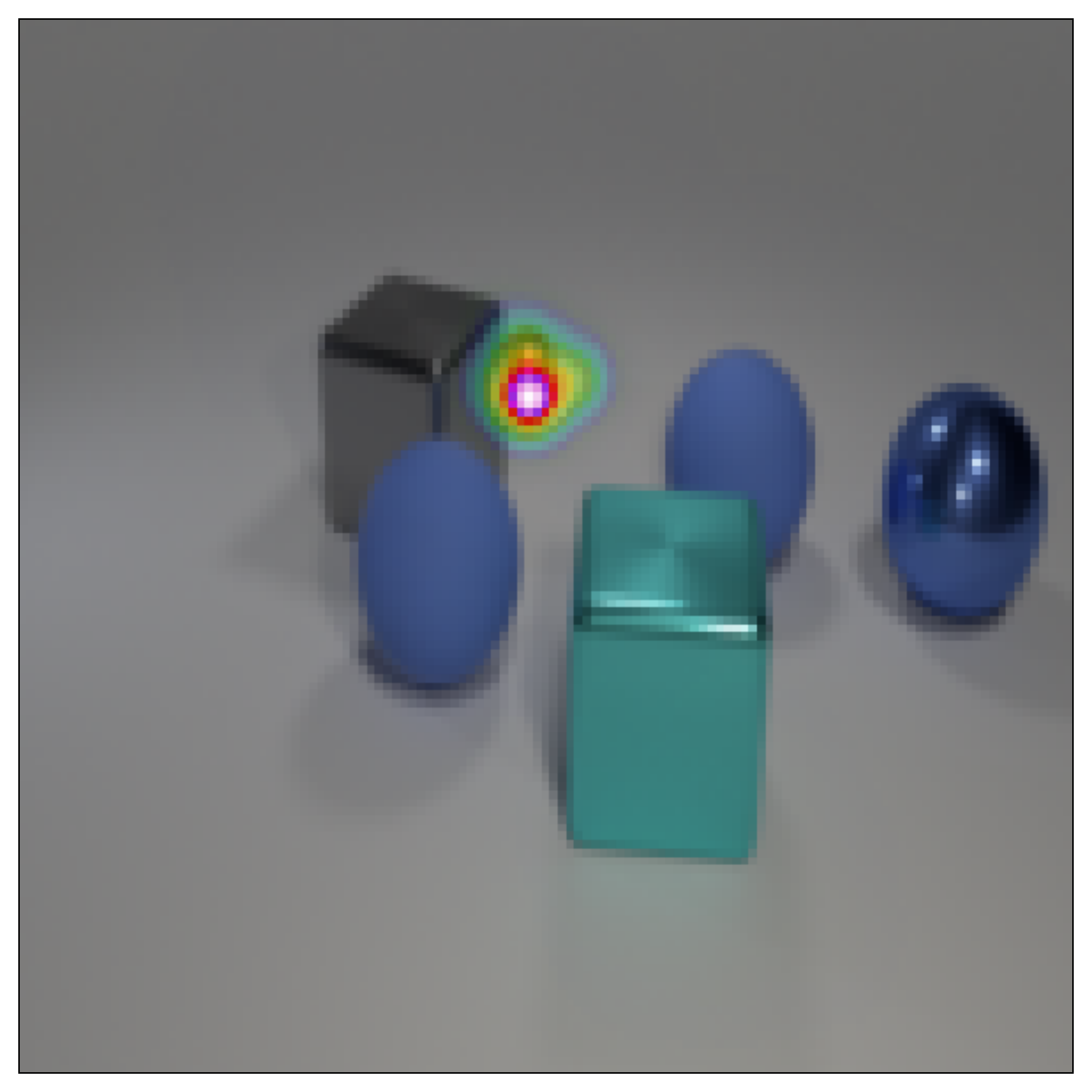} & \includegraphics[width=.12\linewidth,valign=m]{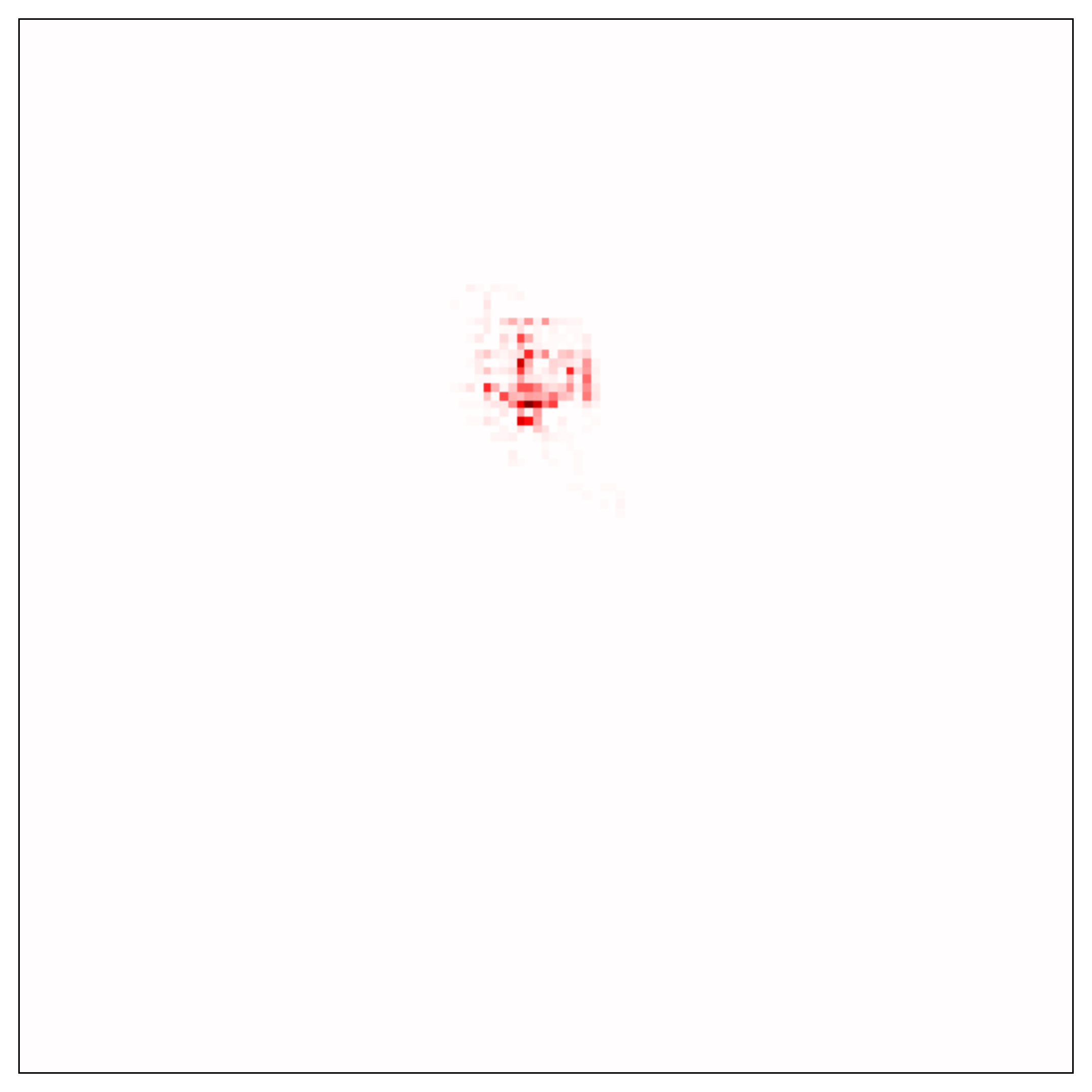} & 0.64 \\
SmoothGrad \cite{Smilkov:ICML2017}                  & \includegraphics[width=.12\linewidth,valign=m]{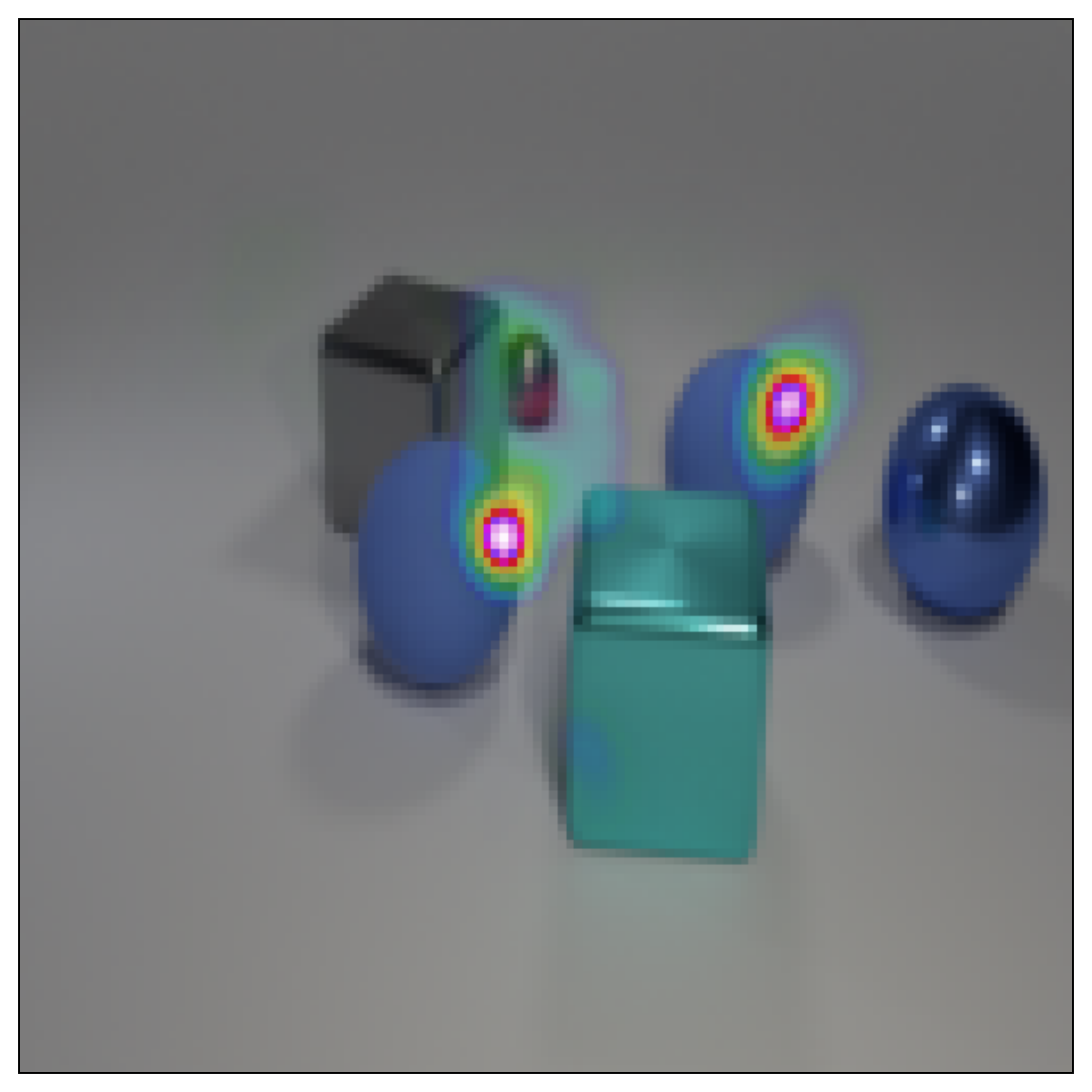} & \includegraphics[width=.12\linewidth,valign=m]{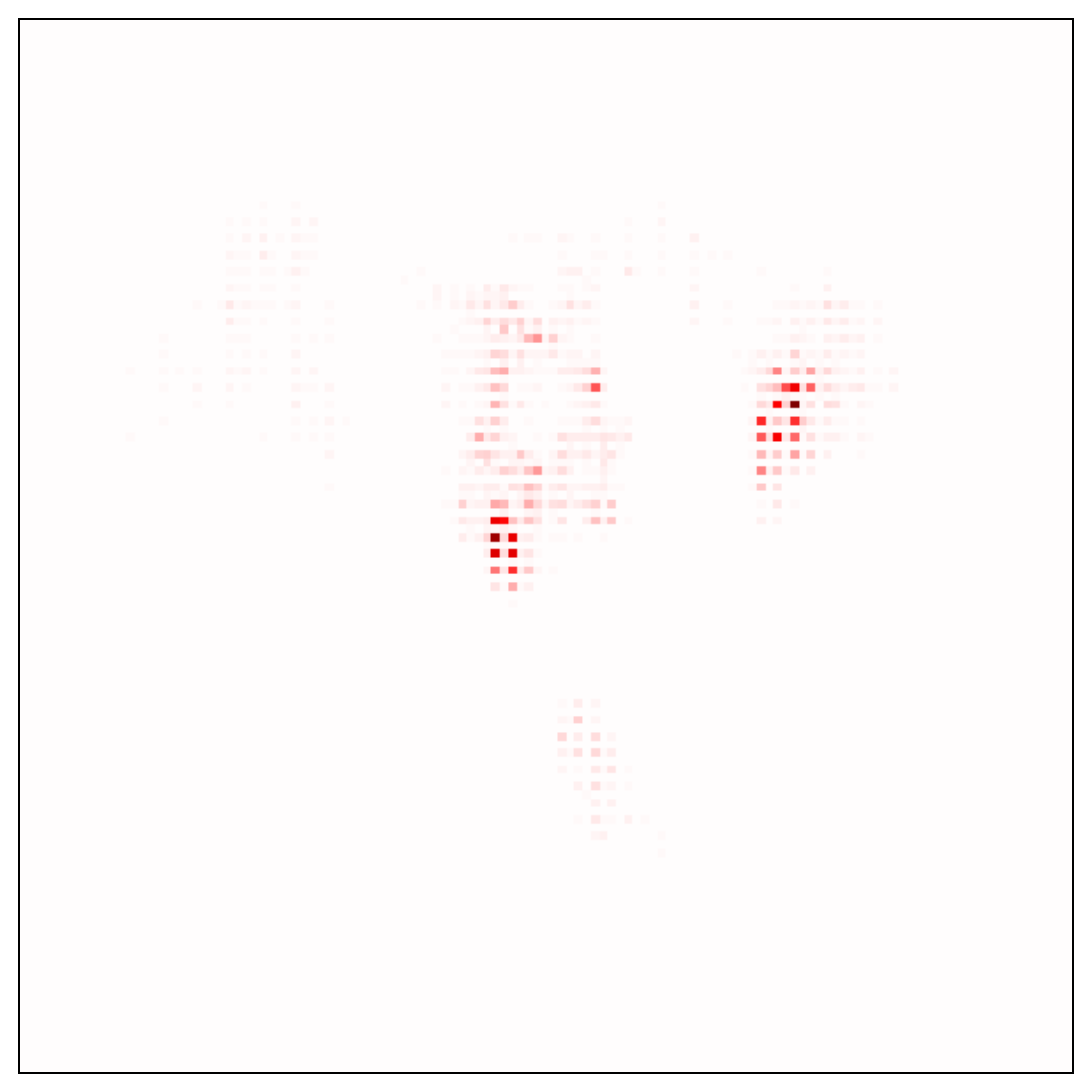} & 0.05 \\
VarGrad \cite{Adebayo:ICLR2018}                     & \includegraphics[width=.12\linewidth,valign=m]{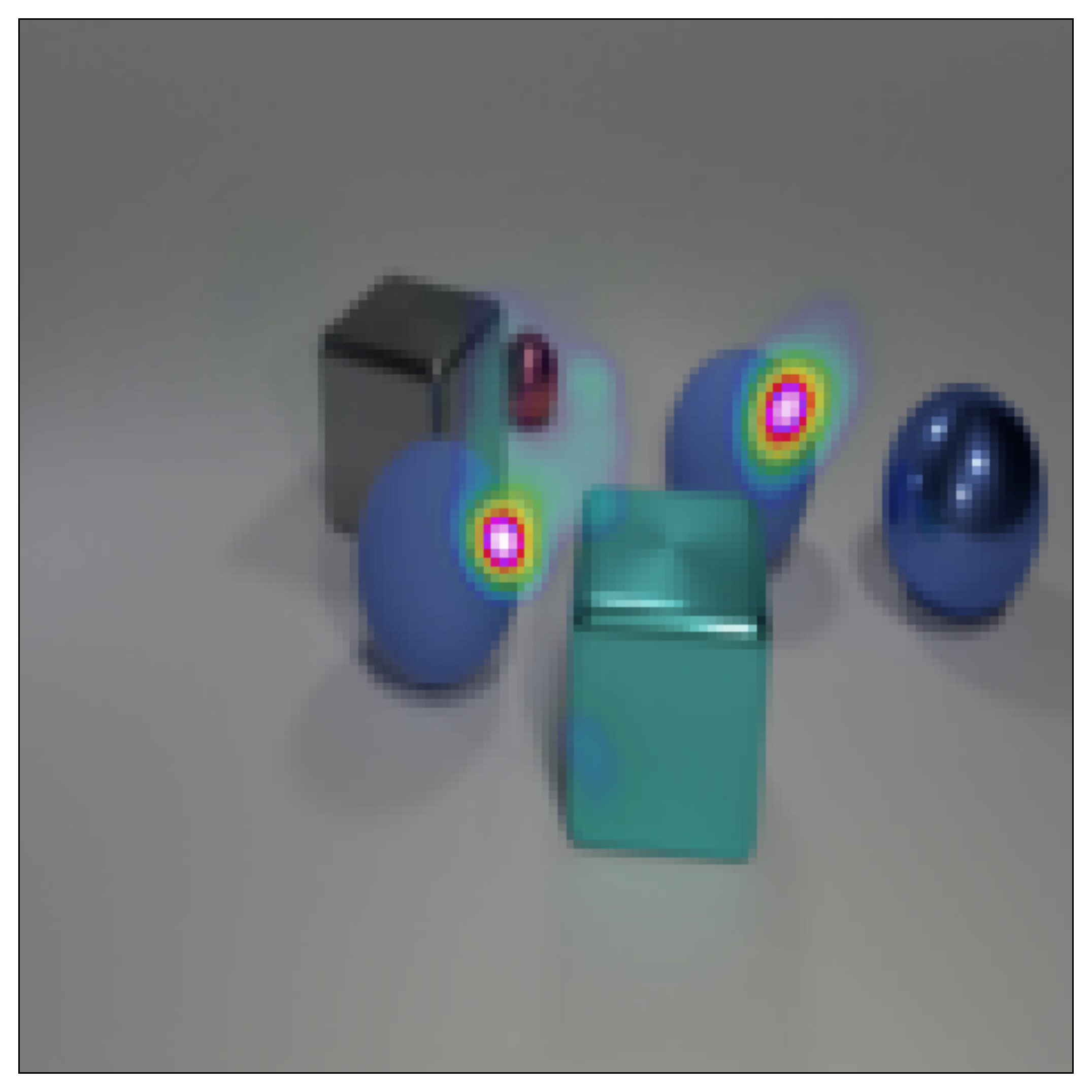} & \includegraphics[width=.12\linewidth,valign=m]{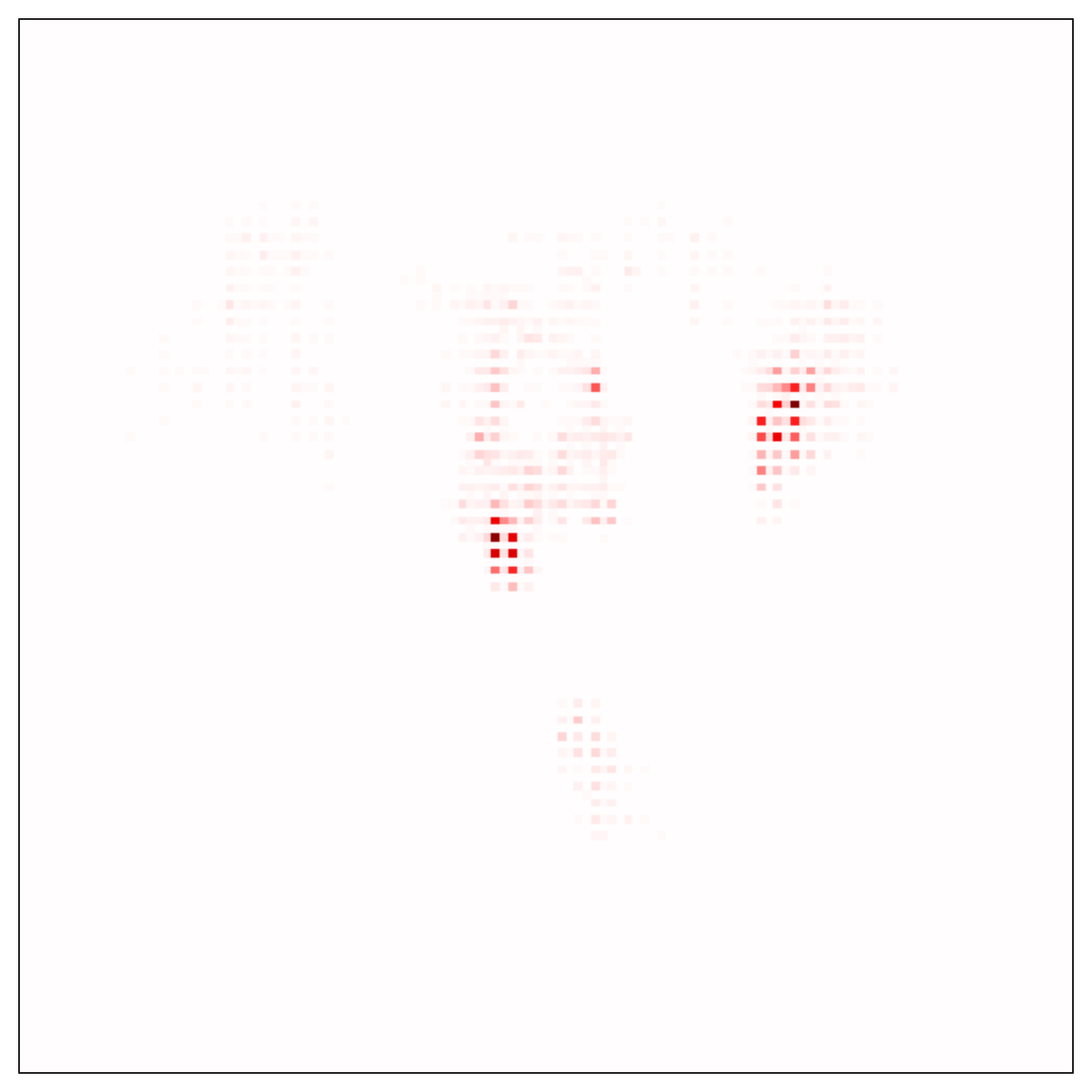} & 0.02 \\
Gradient \cite{Simonyan:ICLR2014}                   & \includegraphics[width=.12\linewidth,valign=m]{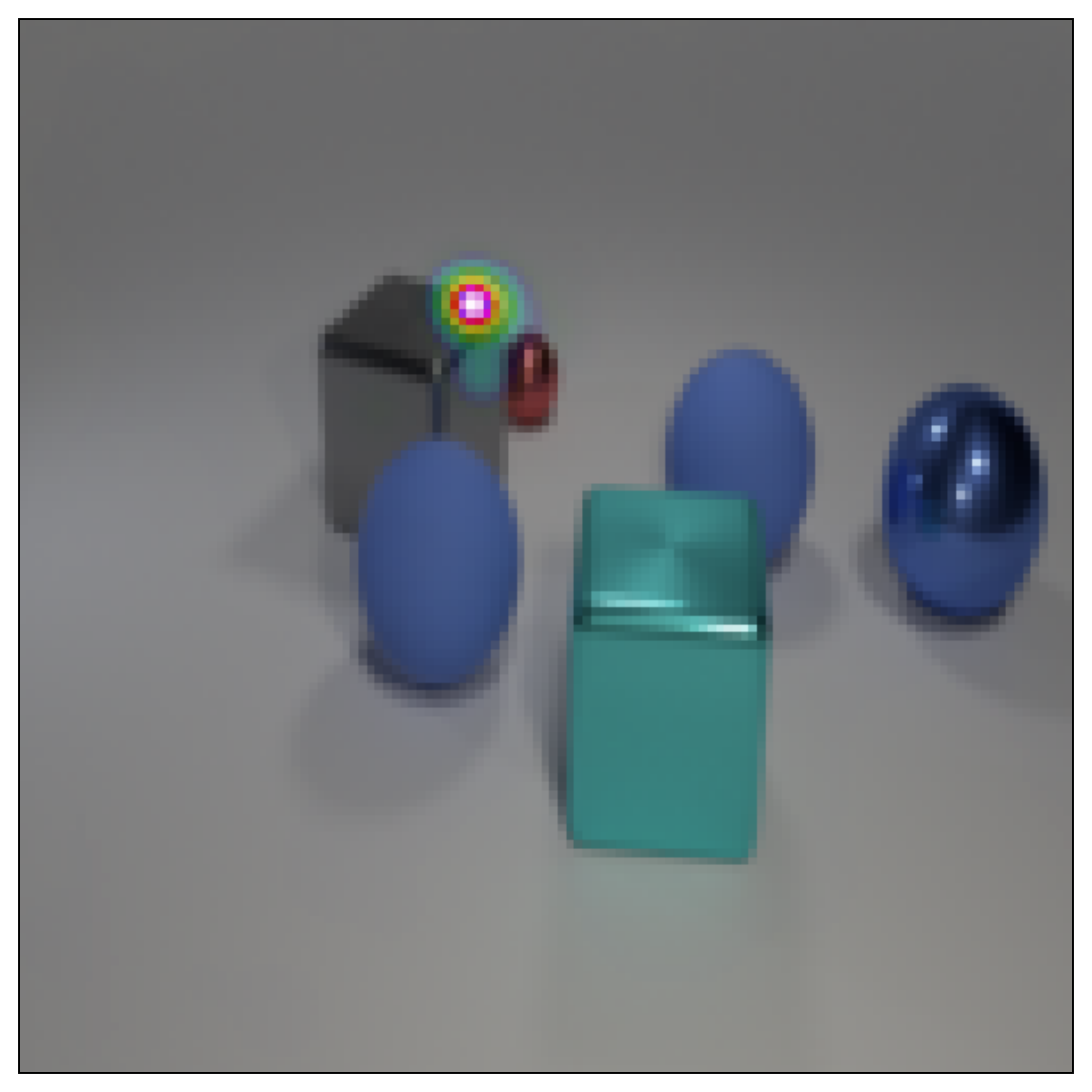} & \includegraphics[width=.12\linewidth,valign=m]{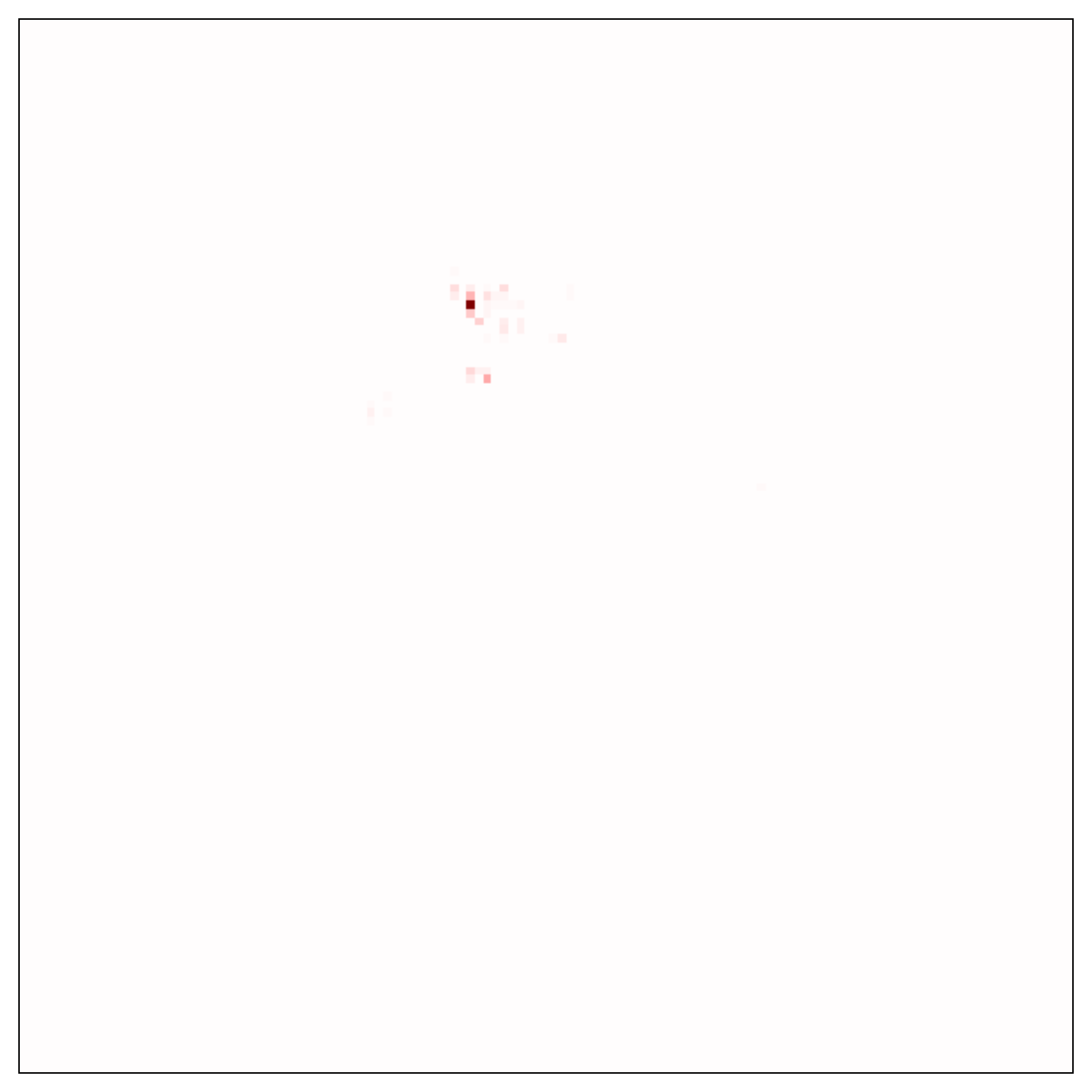} & 0.04 \\
Gradient$\times$Input \cite{Shrikumar:arxiv2016}    & \includegraphics[width=.12\linewidth,valign=m]{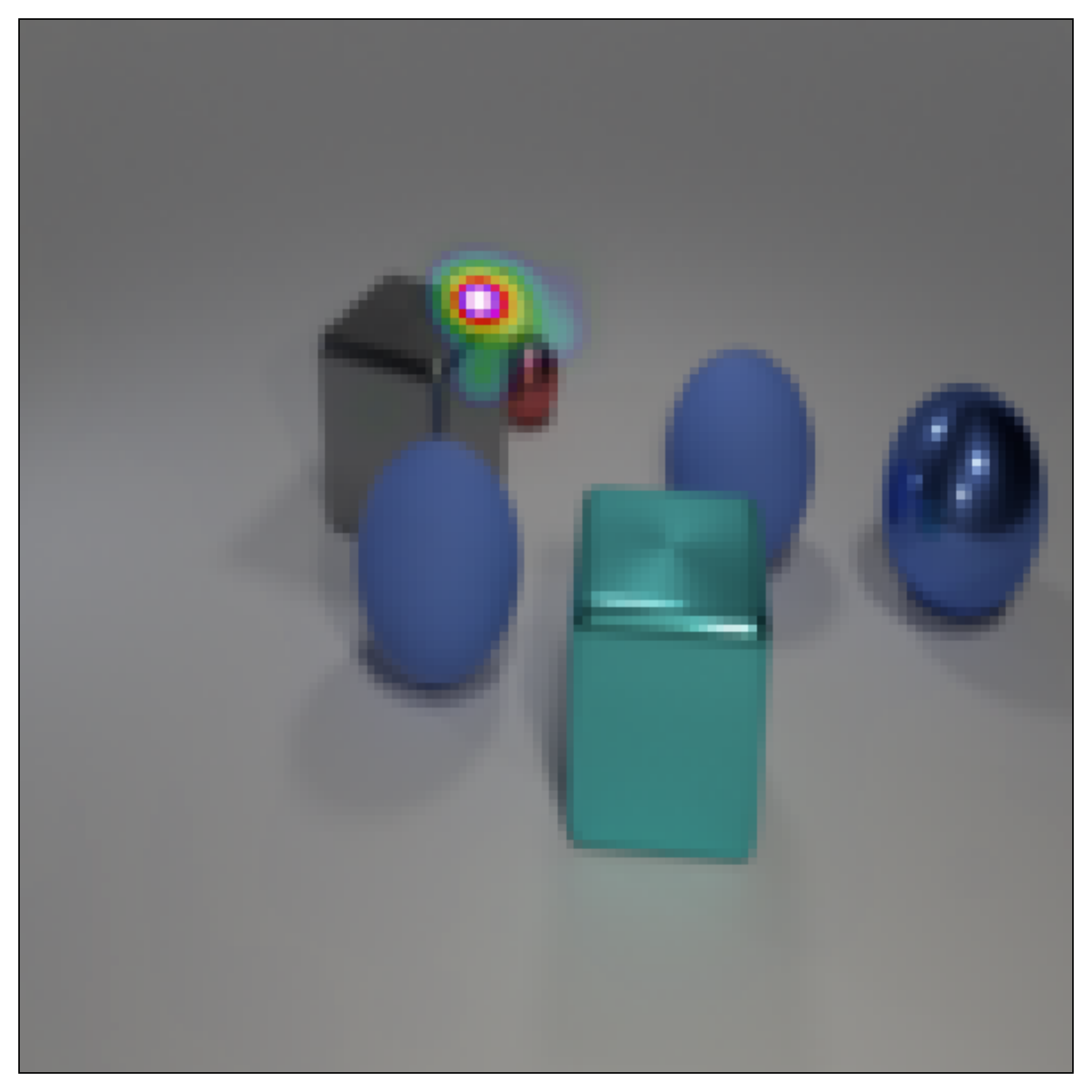} & \includegraphics[width=.12\linewidth,valign=m]{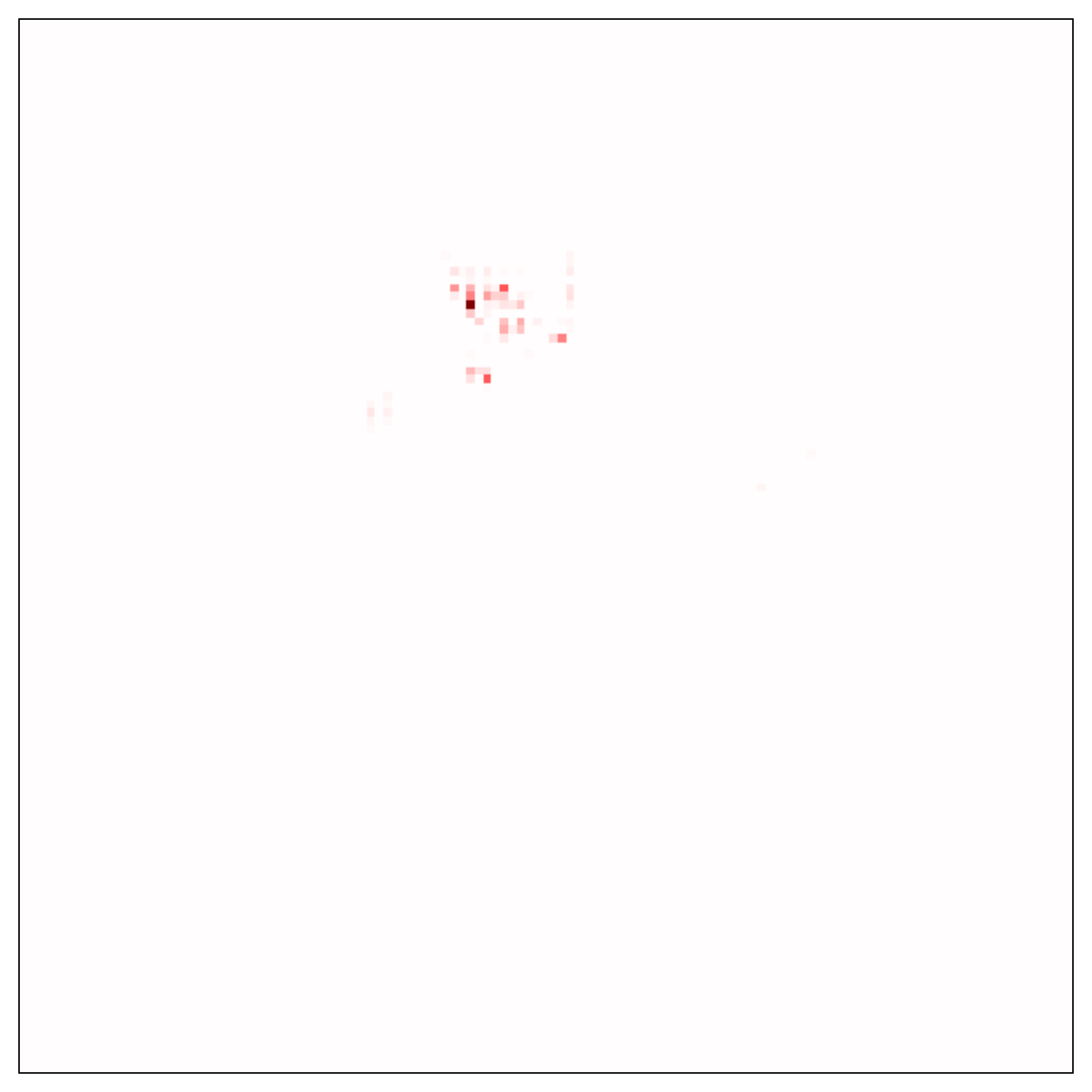} & 0.06 \\
Deconvnet \cite{Zeiler:ECCV2014}                    & \includegraphics[width=.12\linewidth,valign=m]{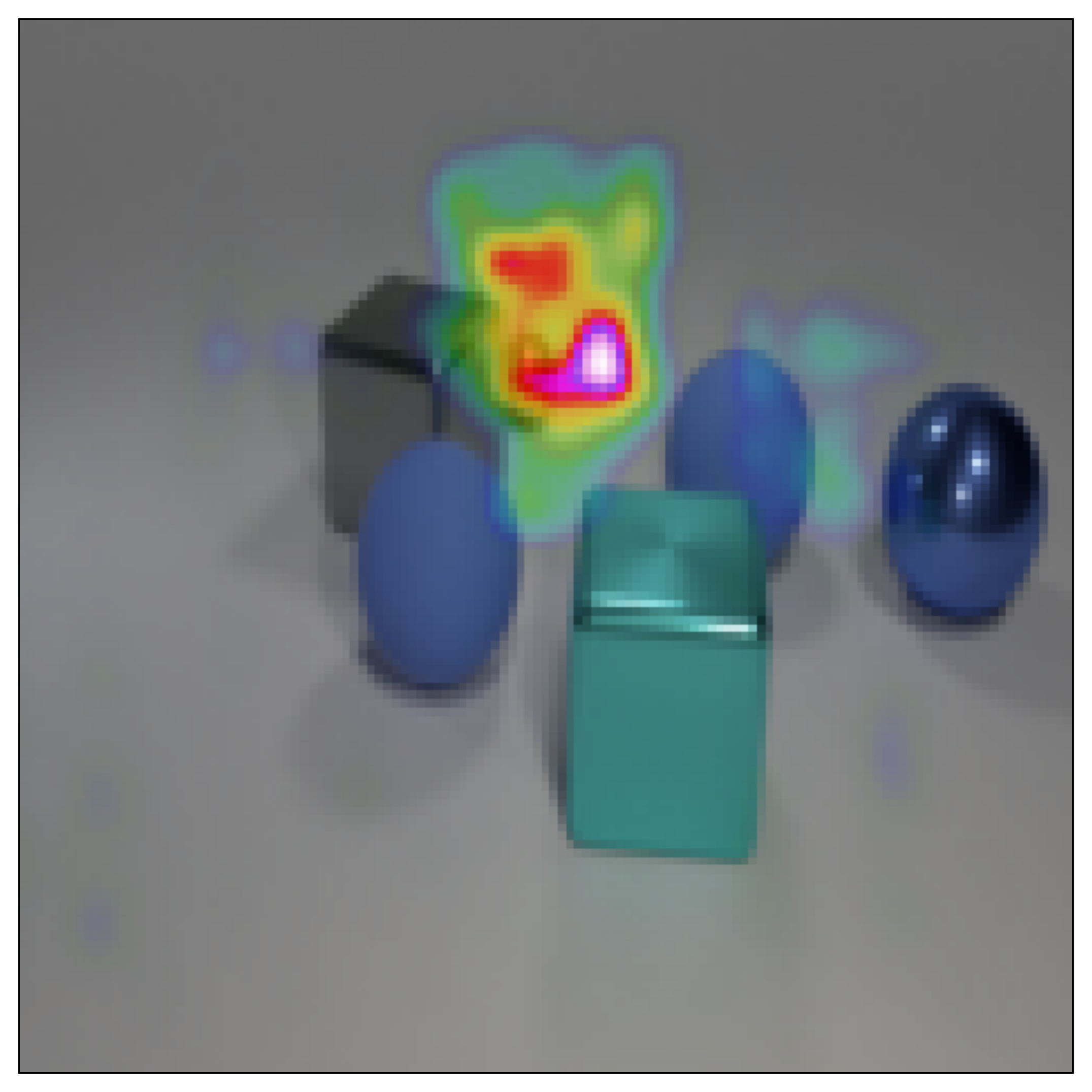} & \includegraphics[width=.12\linewidth,valign=m]{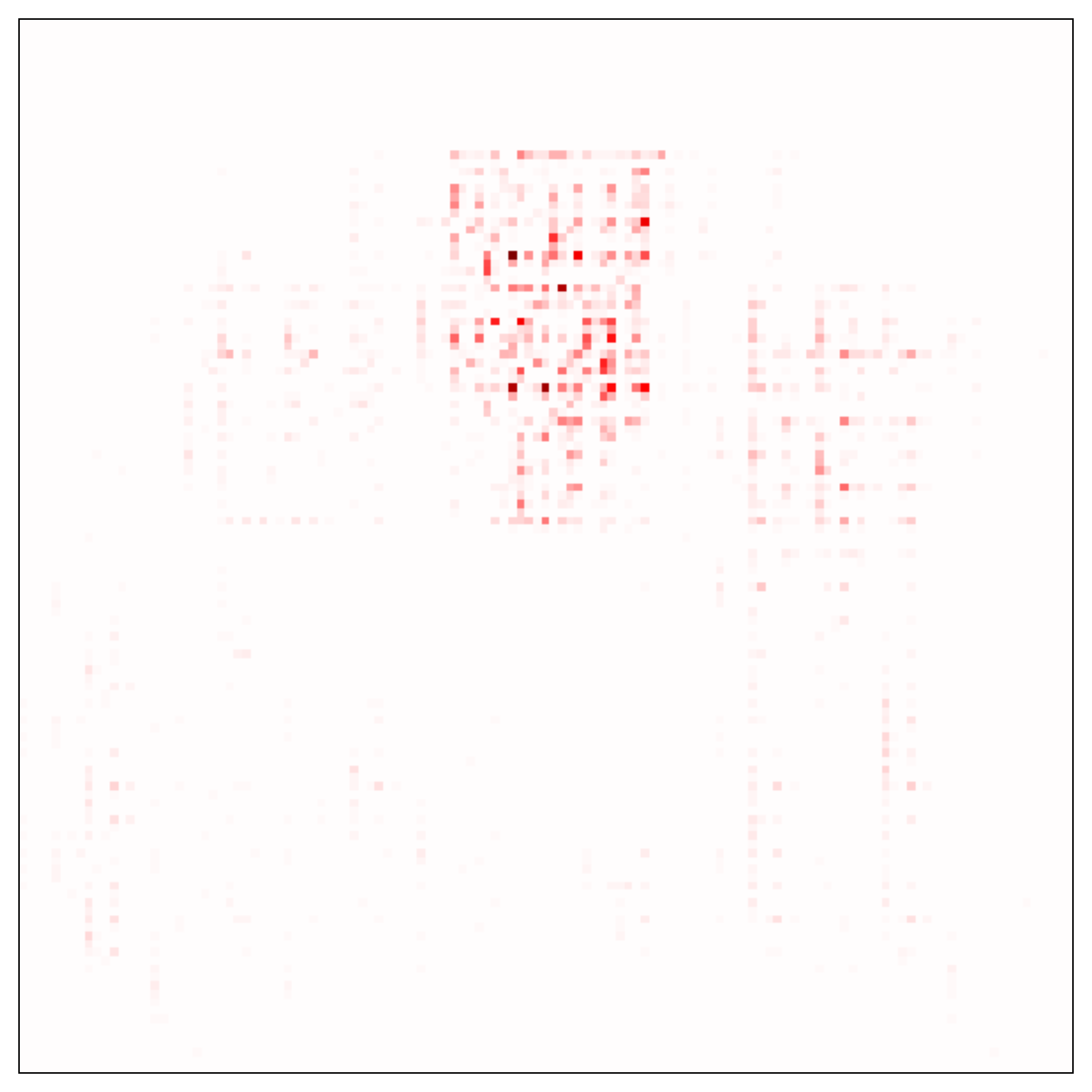} & 0.07 \\
Grad-CAM \cite{Selvaraju:ICCV2017}                  & \includegraphics[width=.12\linewidth,valign=m]{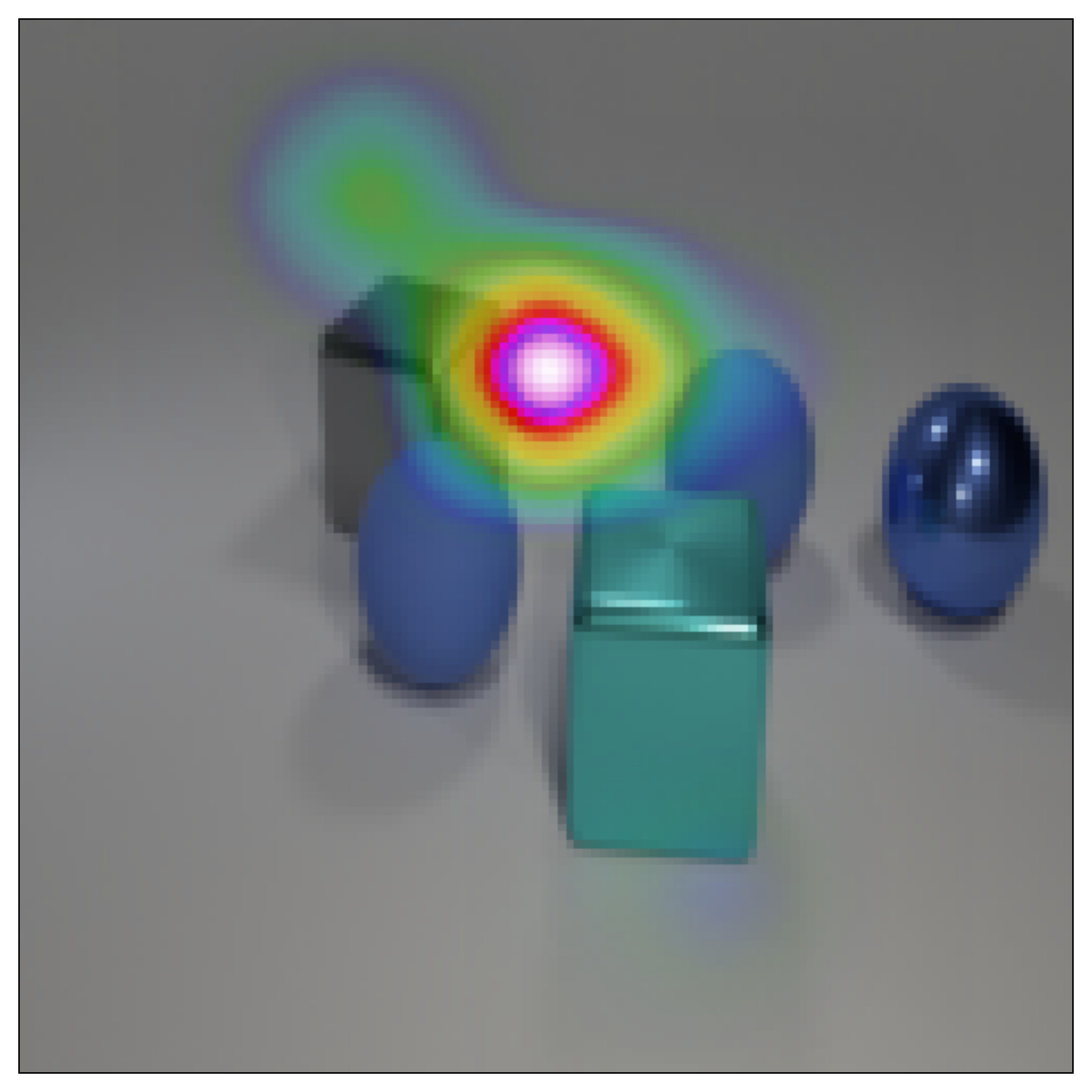} & \includegraphics[width=.12\linewidth,valign=m]{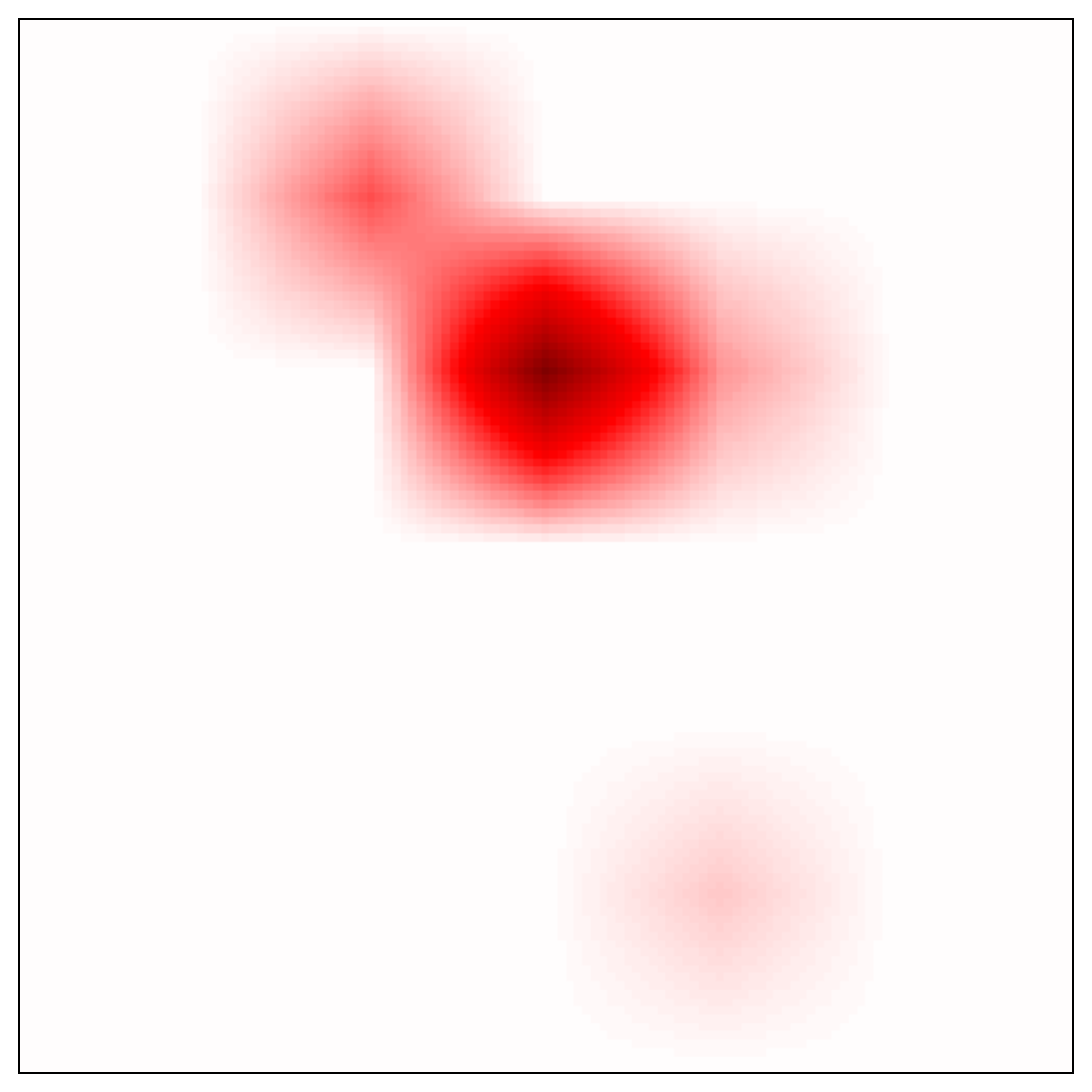} & 0.10 \\
\end{tabular}
\end{table}

\begin{table}
        \scriptsize
		\caption{Heatmaps for a falsely predicted CLEVR-XAI-simple question (raw heatmap and heatmap overlayed with original image), and corresponding relevance \textit{mass} accuracy.}
		\label{table:heatmap-simple-false-24357}
\begin{tabular}{lllc}
\midrule
\begin{tabular}{@{}l@{}}What color is the\\ small shiny block? \\ true: \textit{purple} \\ predicted: \textit{blue} \end{tabular}  & \includegraphics[width=.18\linewidth,valign=m]{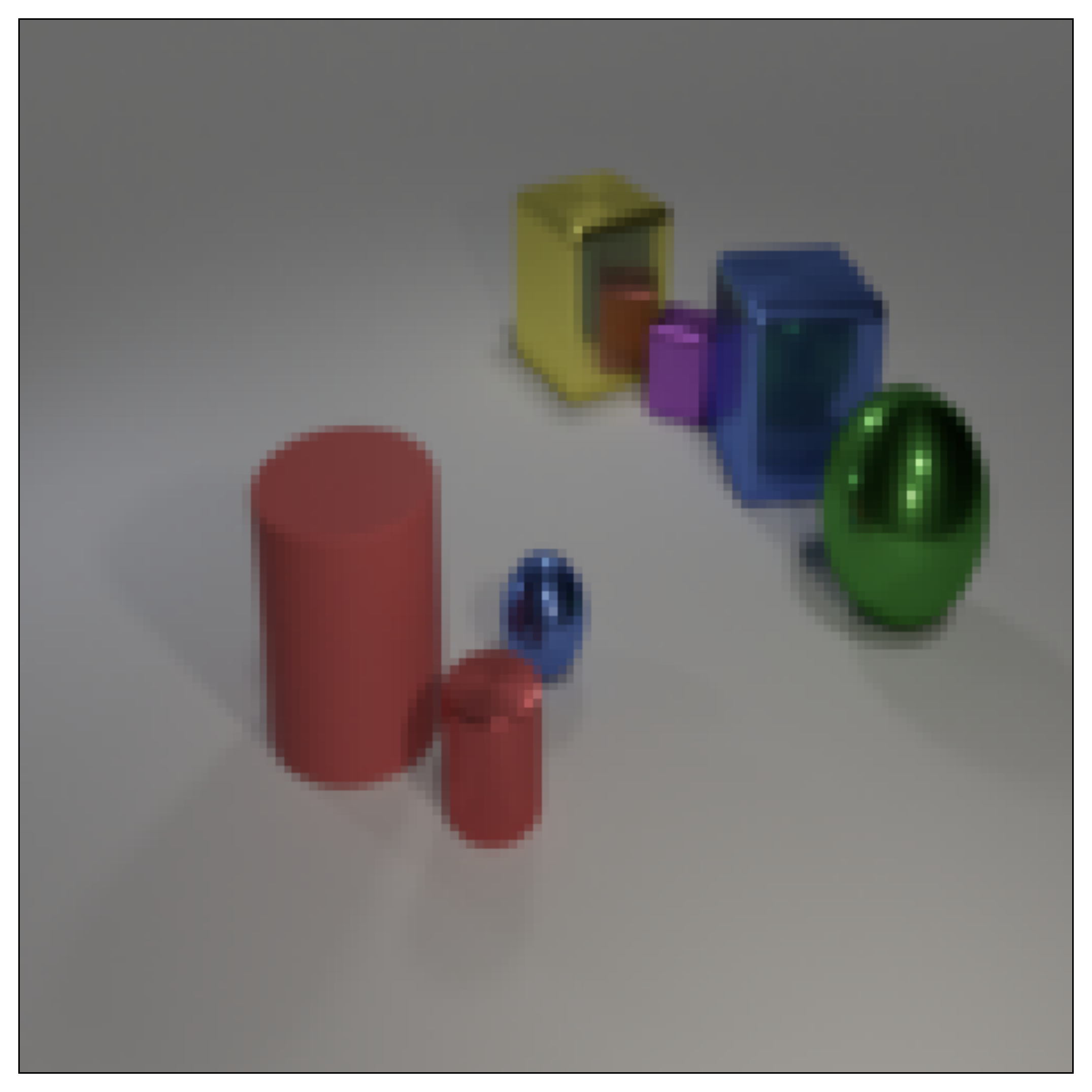} &
\includegraphics[width=.18\linewidth,valign=m]{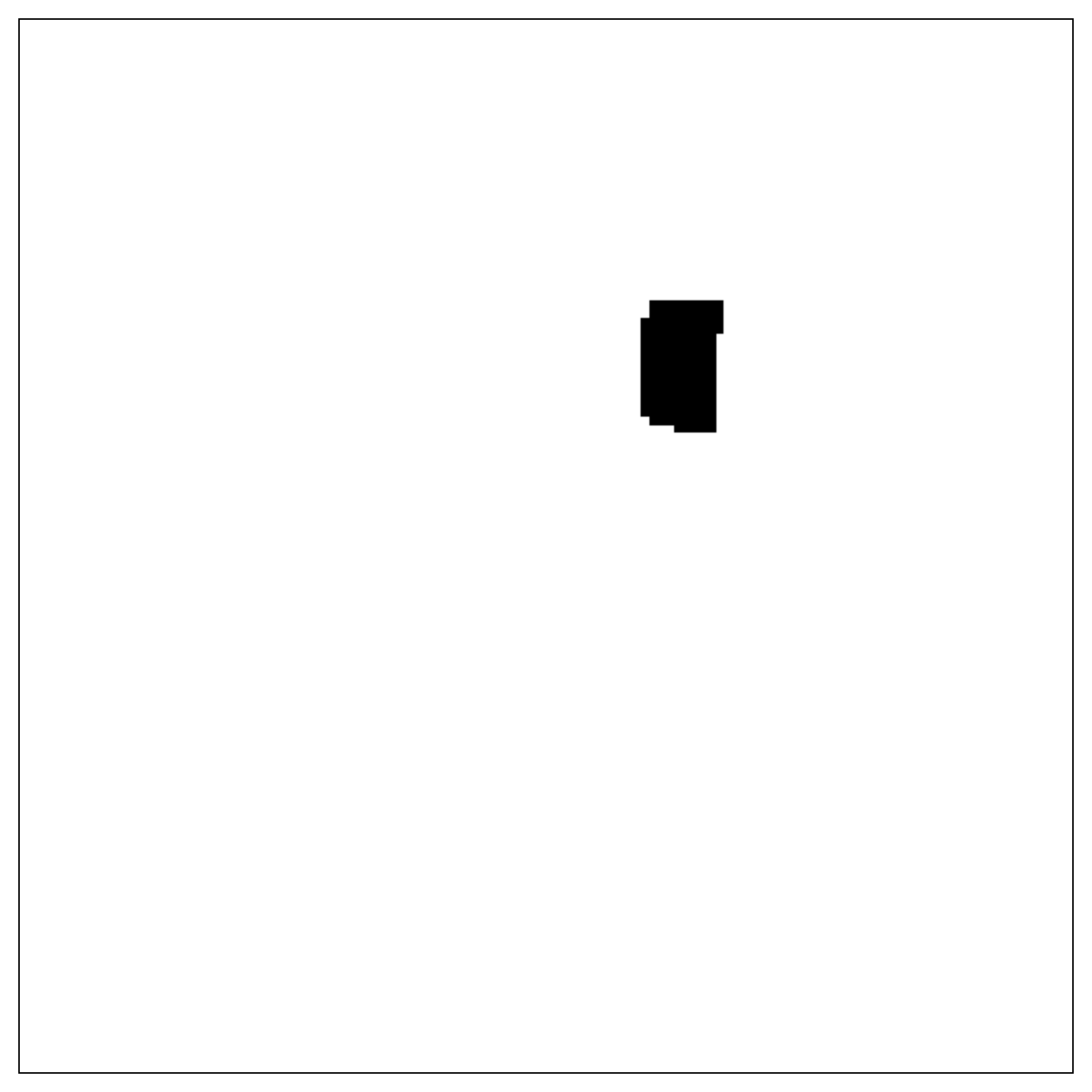} & GT Single Object \\
\midrule
LRP \cite{Bach:PLOS2015}                            & \includegraphics[width=.12\linewidth,valign=m]{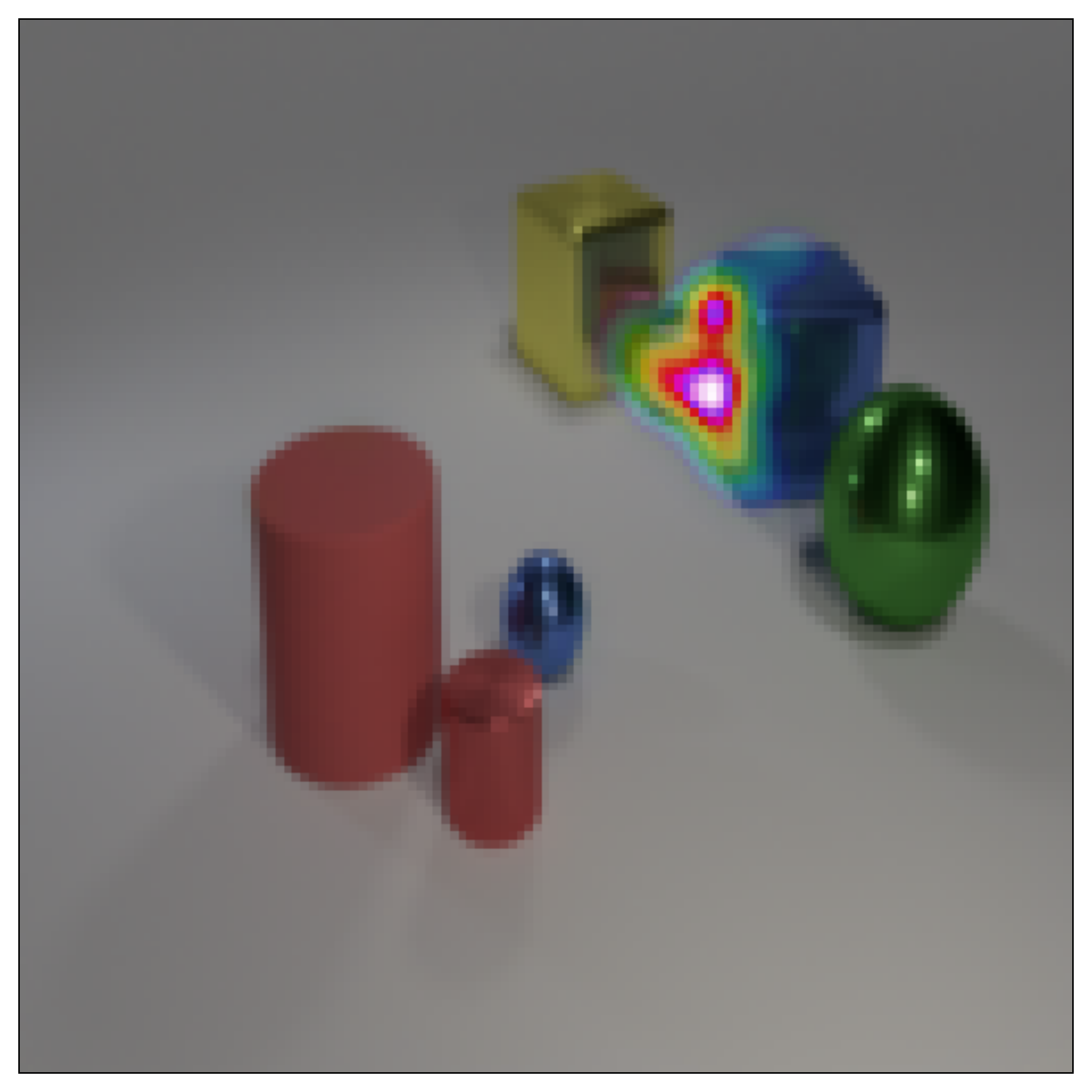} & \includegraphics[width=.12\linewidth,valign=m]{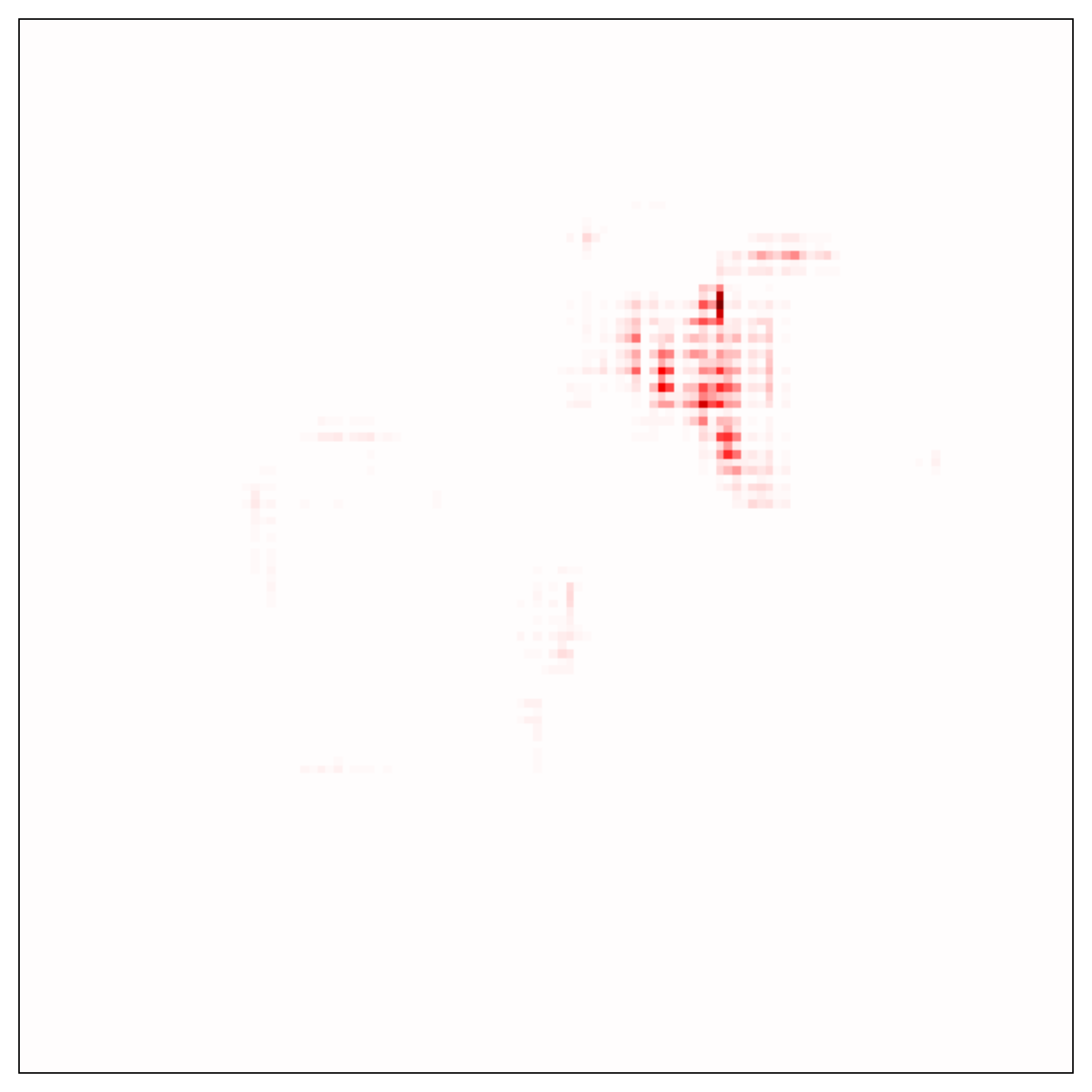} & 0.37 \\
Excitation Backprop \cite{Zhang:ECCV2016}           & \includegraphics[width=.12\linewidth,valign=m]{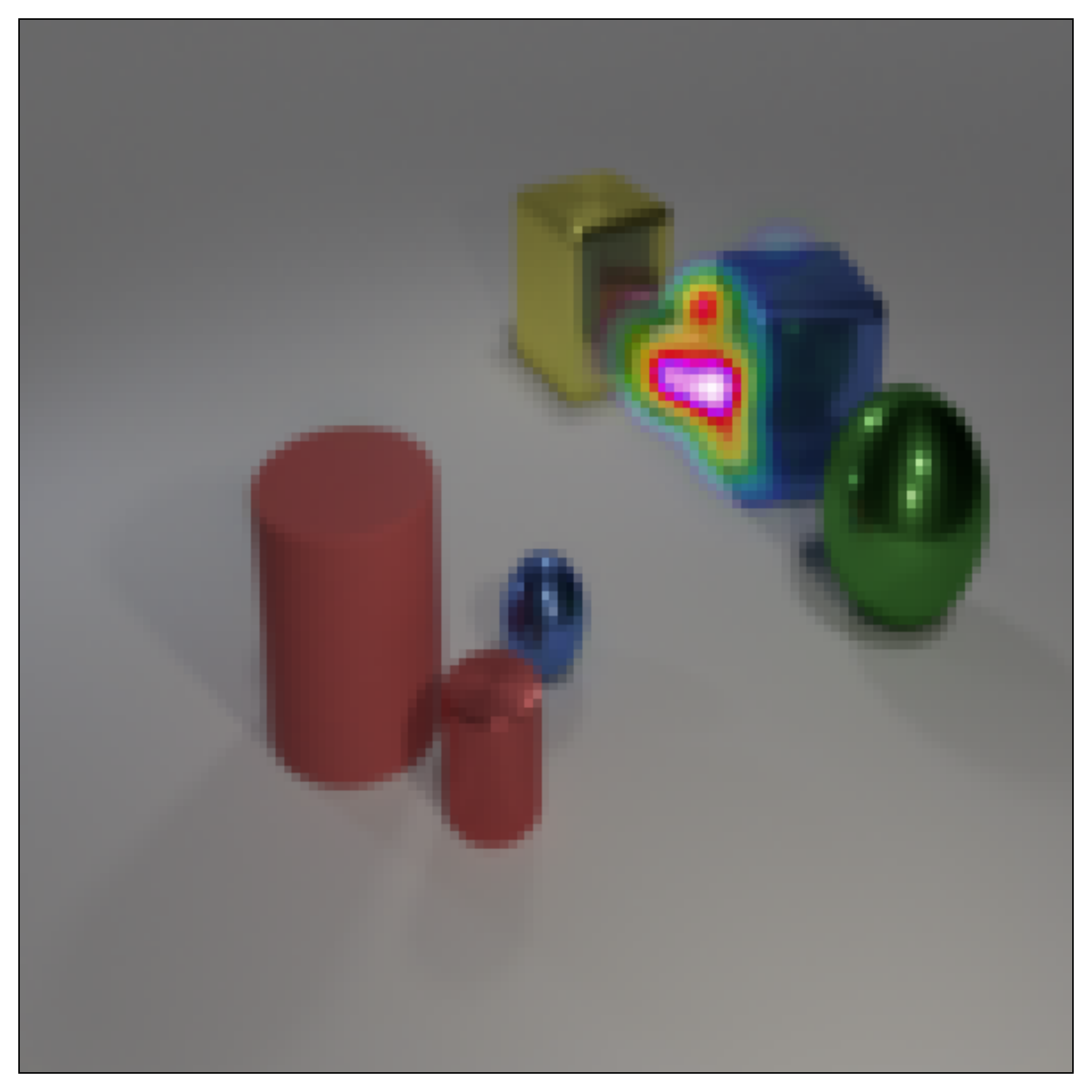} & \includegraphics[width=.12\linewidth,valign=m]{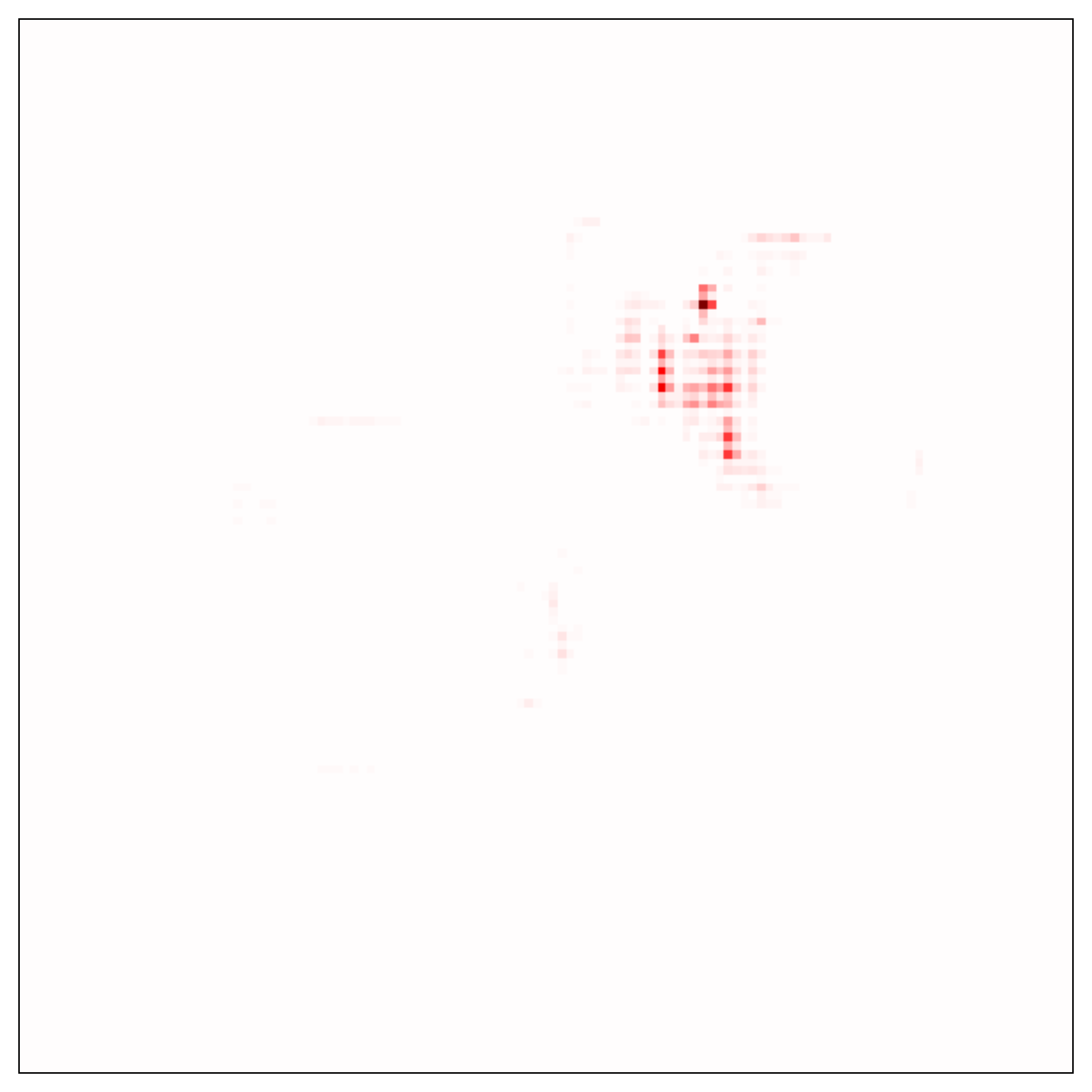} & 0.40 \\
IG \cite{Sundararajan:ICML2017}                     & \includegraphics[width=.12\linewidth,valign=m]{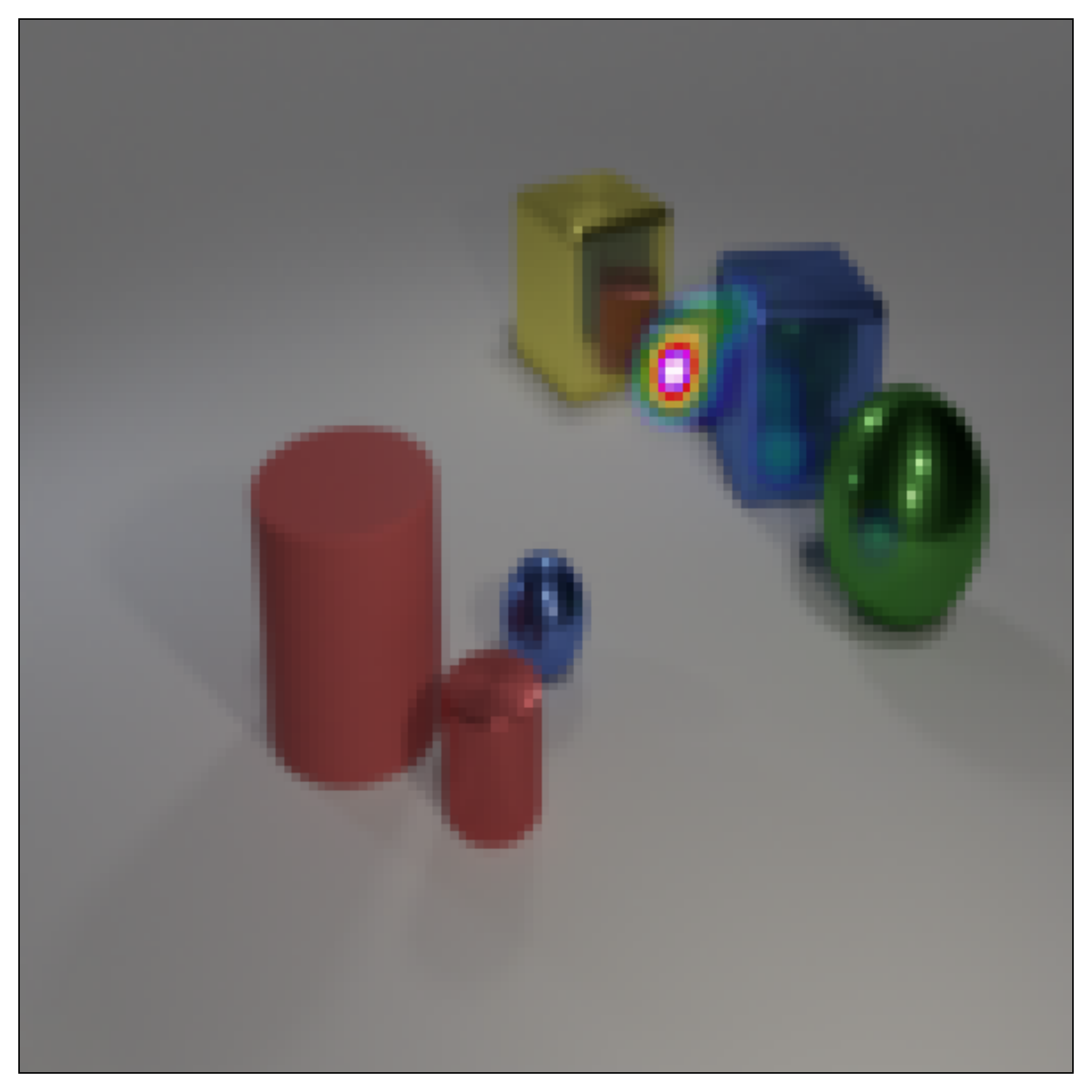} & \includegraphics[width=.12\linewidth,valign=m]{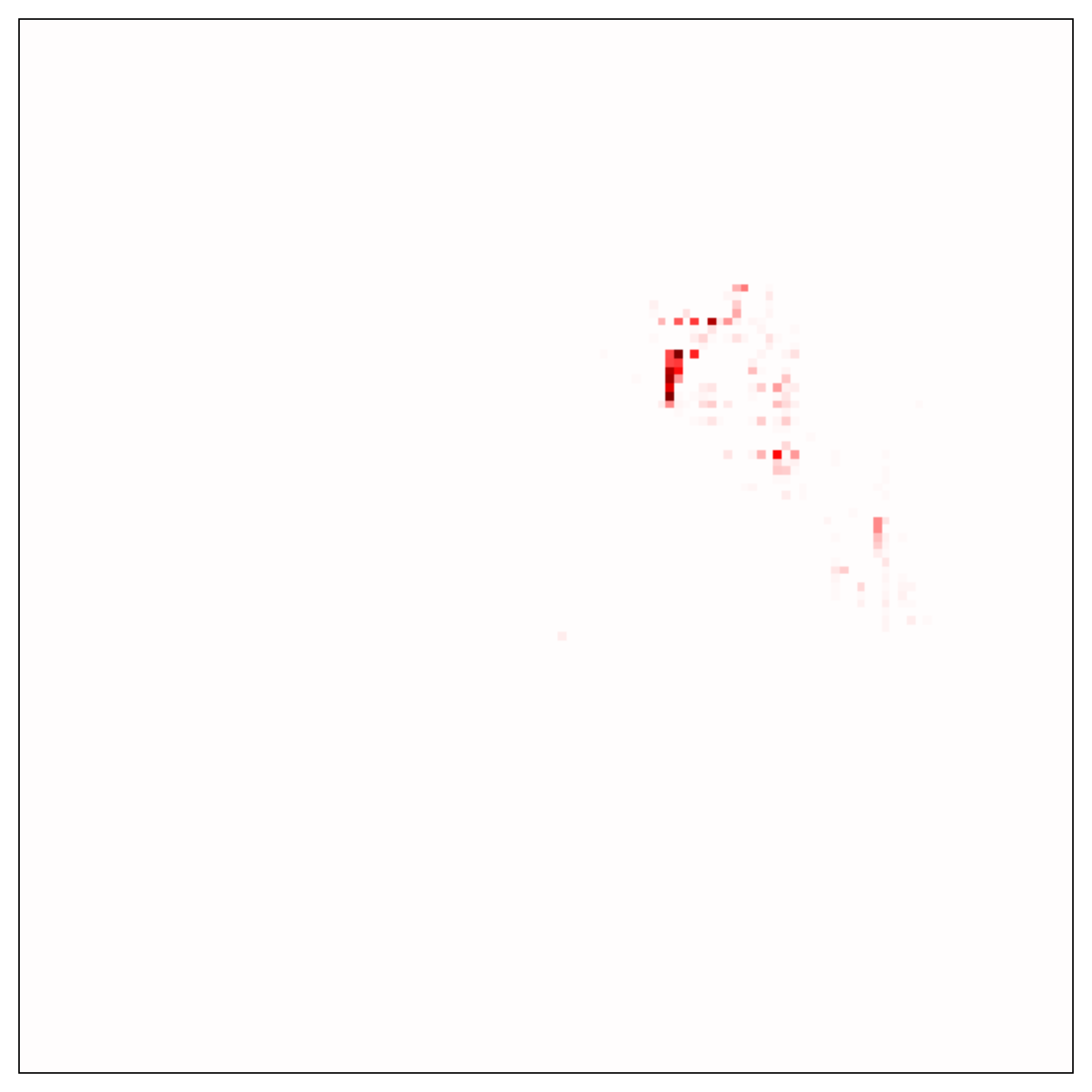} & 0.58 \\
Guided Backprop \cite{Spring:ICLR2015}              & \includegraphics[width=.12\linewidth,valign=m]{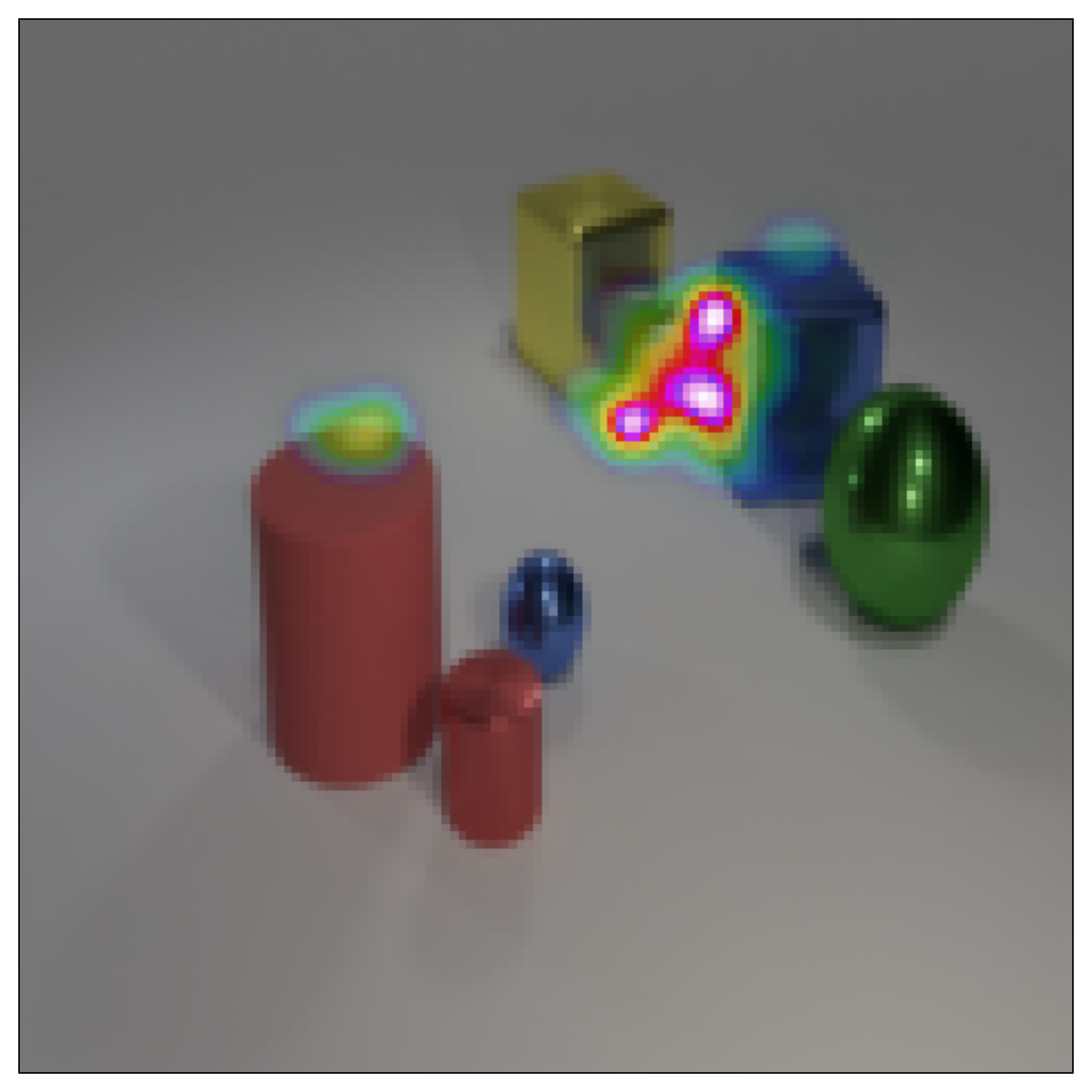} & \includegraphics[width=.12\linewidth,valign=m]{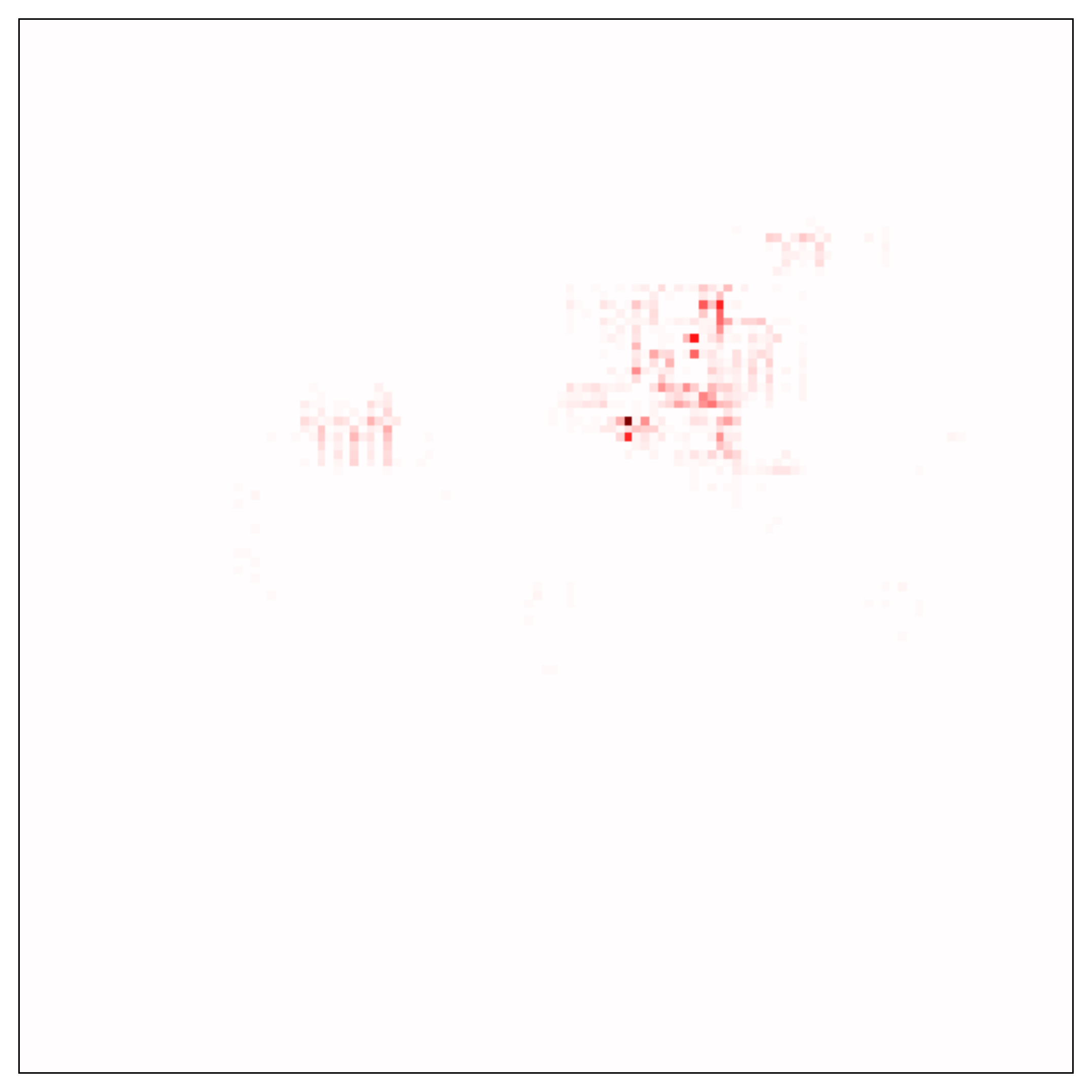} & 0.28 \\
Guided Grad-CAM \cite{Selvaraju:ICCV2017}           & \includegraphics[width=.12\linewidth,valign=m]{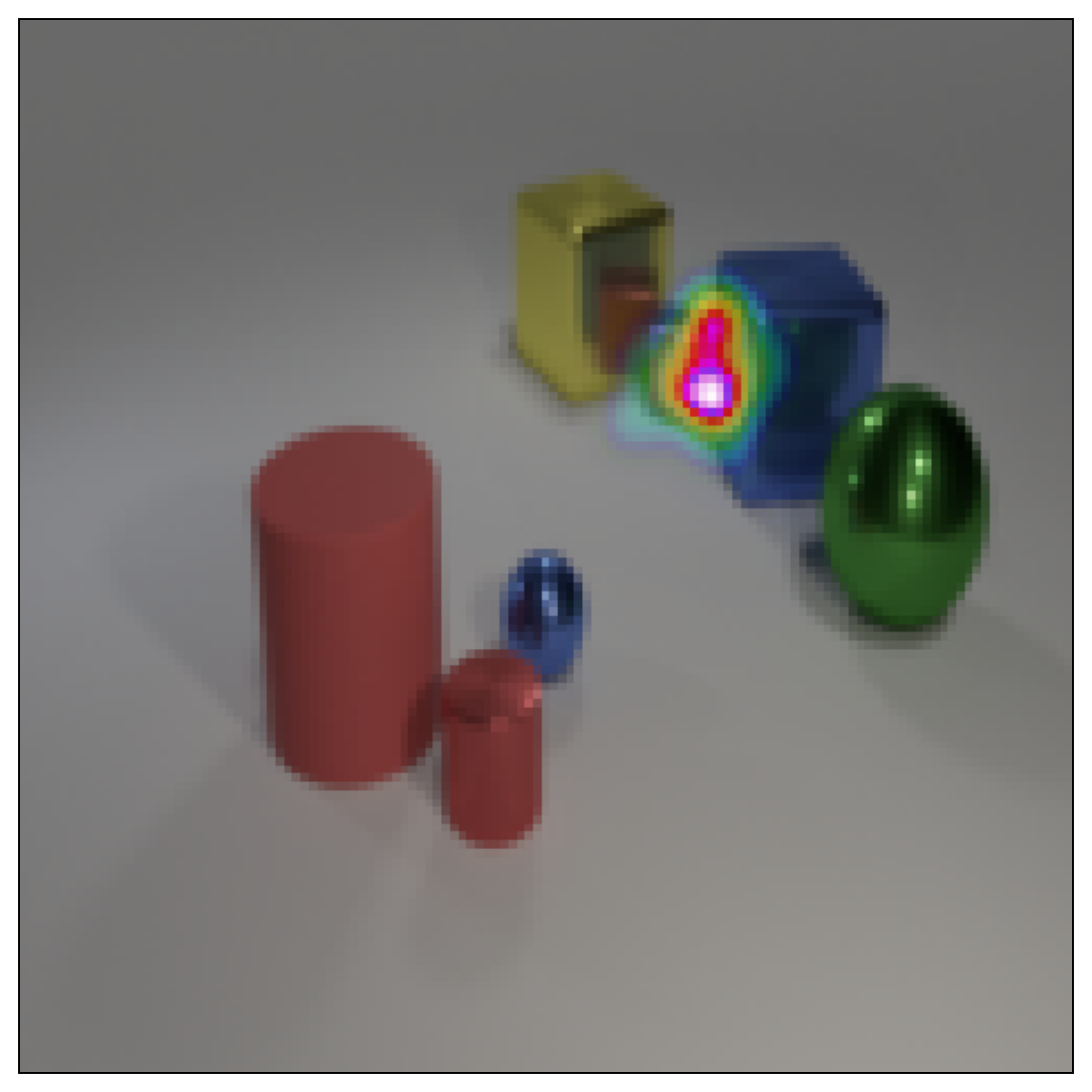} & \includegraphics[width=.12\linewidth,valign=m]{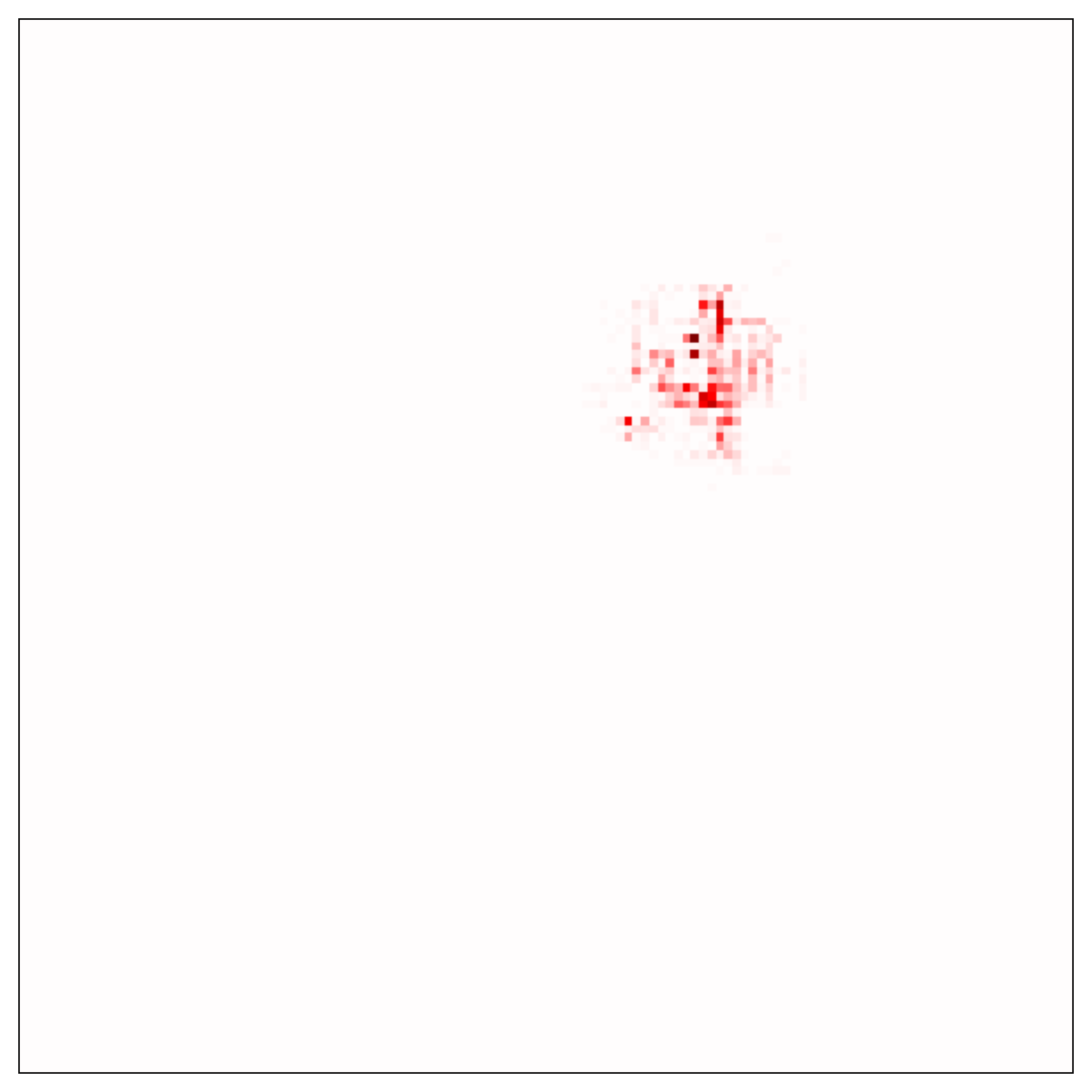} & 0.52 \\
SmoothGrad \cite{Smilkov:ICML2017}                  & \includegraphics[width=.12\linewidth,valign=m]{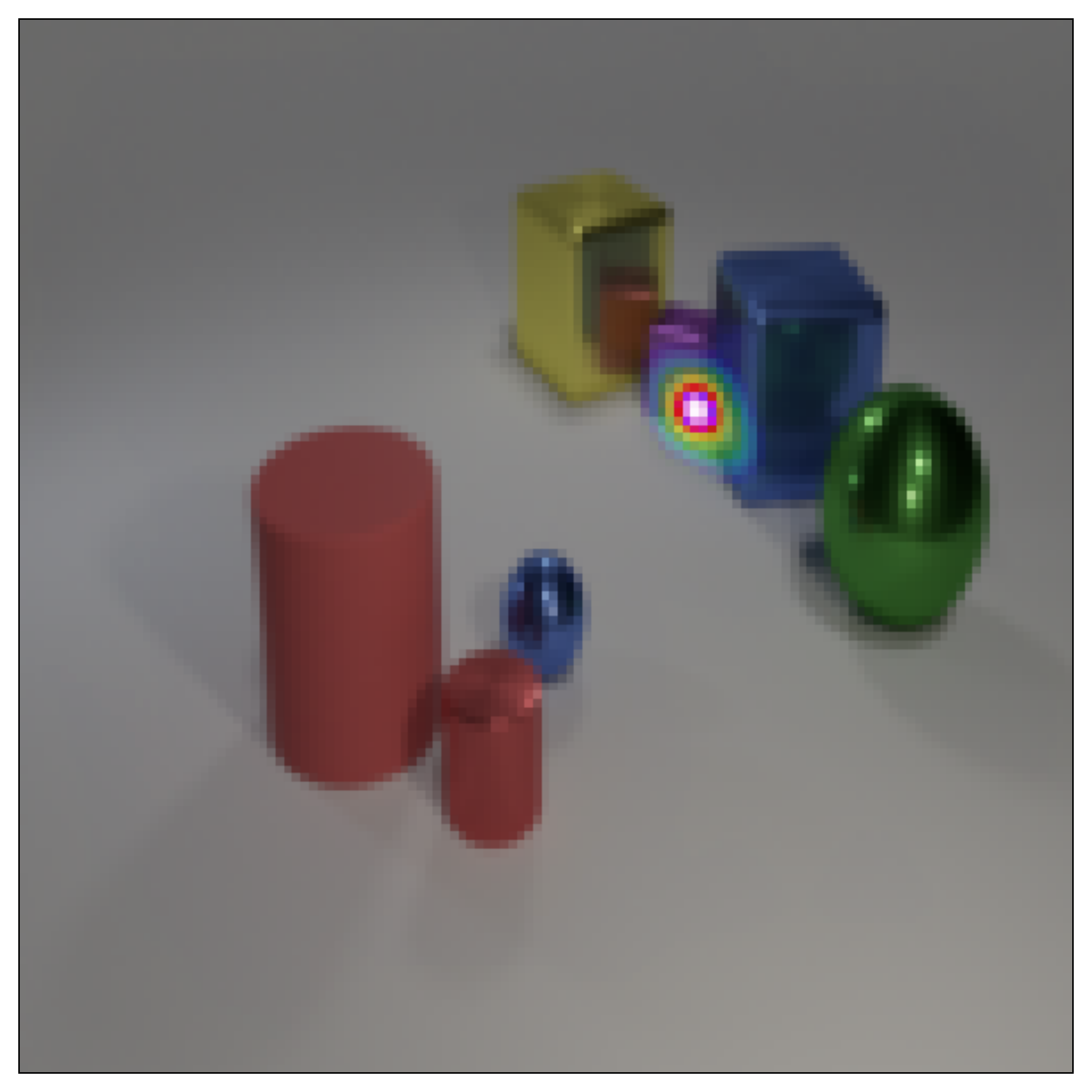} & \includegraphics[width=.12\linewidth,valign=m]{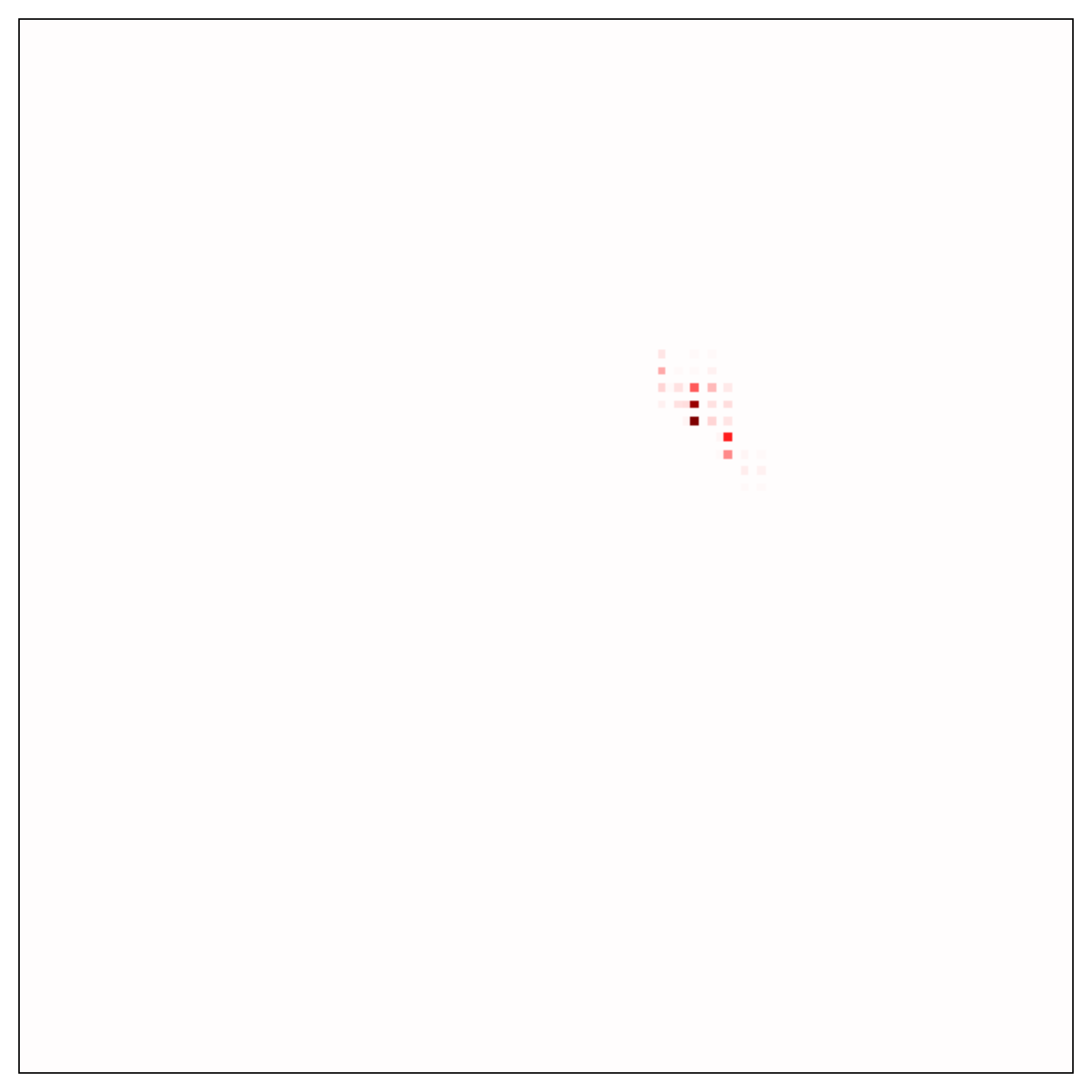} & 0.74 \\
VarGrad \cite{Adebayo:ICLR2018}                     & \includegraphics[width=.12\linewidth,valign=m]{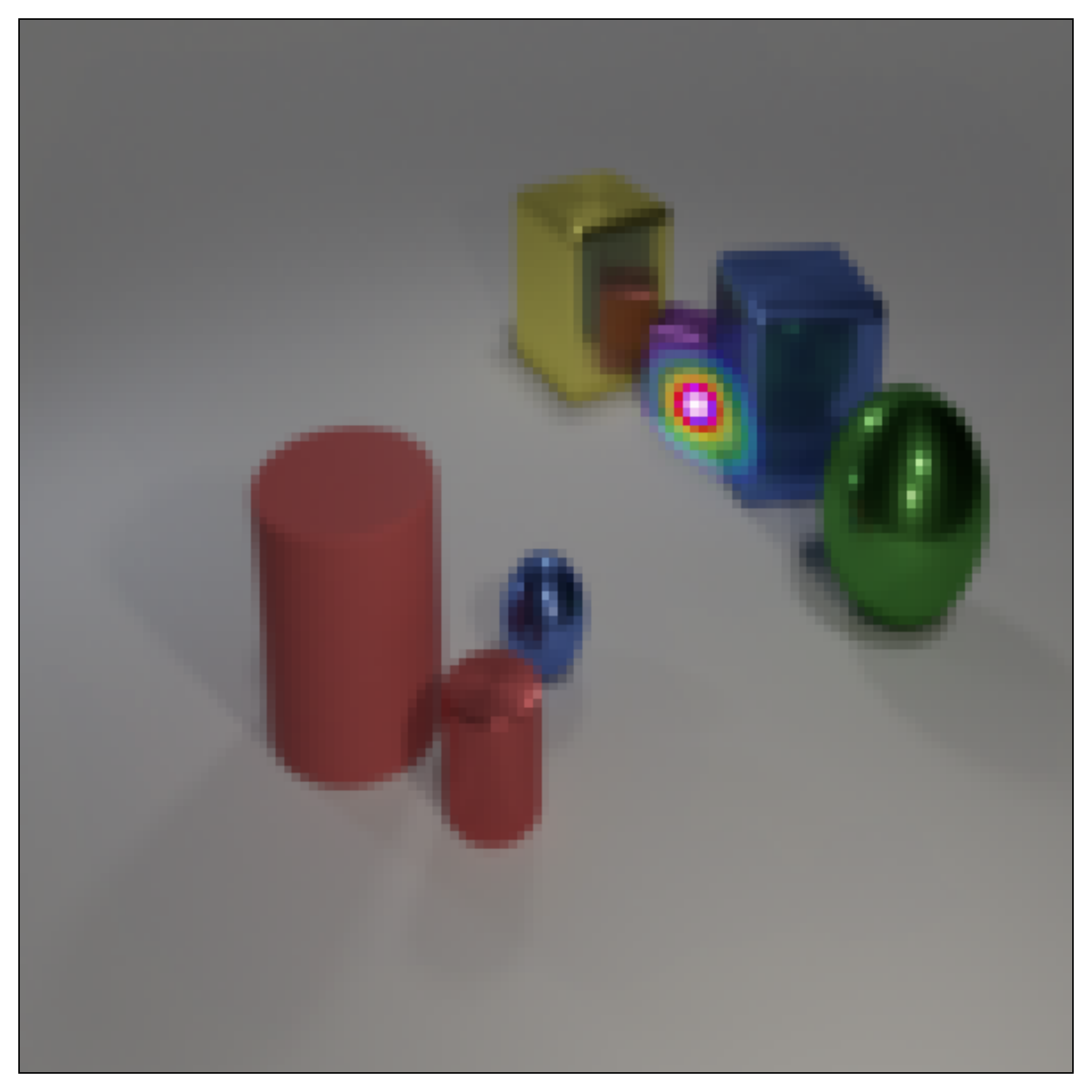} & \includegraphics[width=.12\linewidth,valign=m]{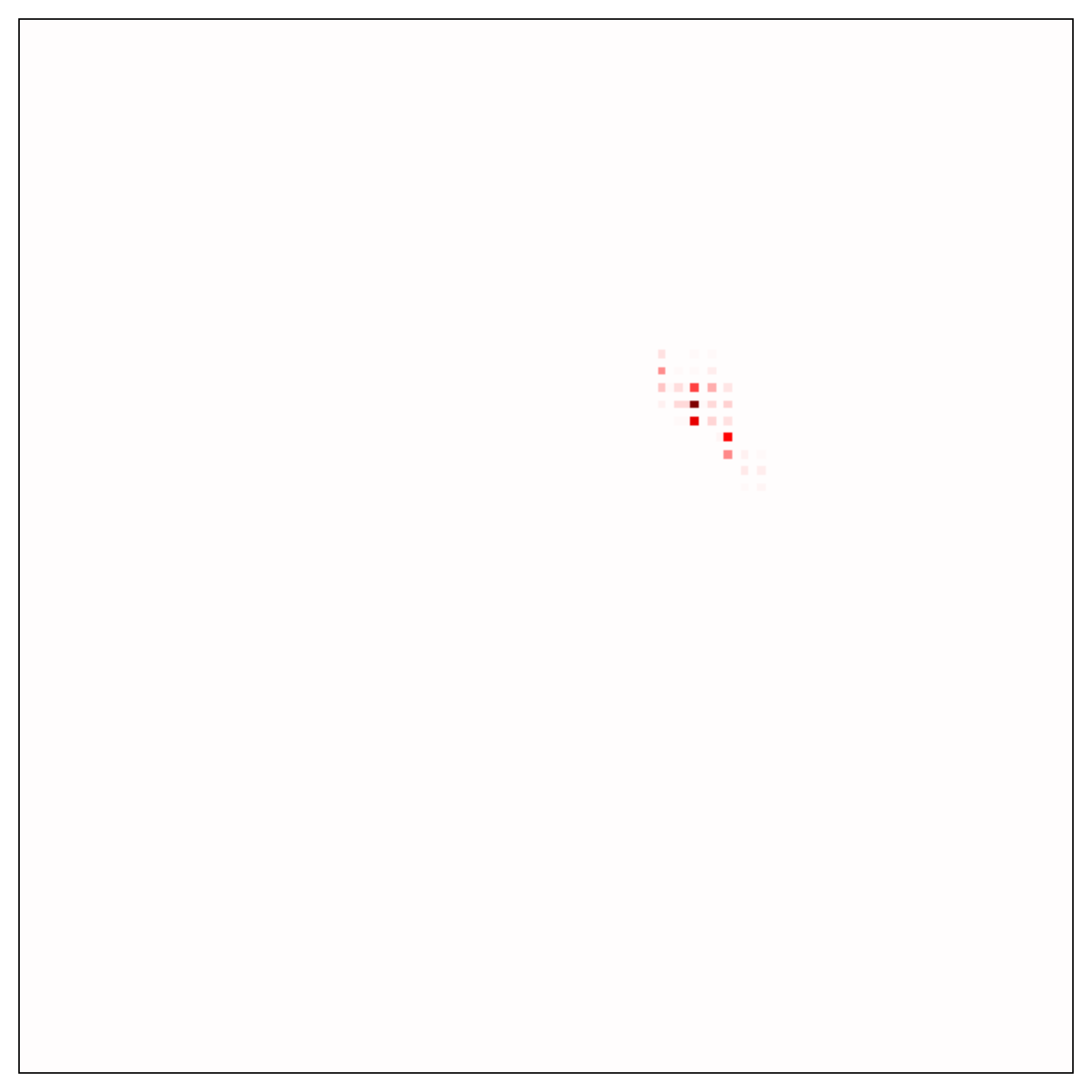} & 0.72 \\
Gradient \cite{Simonyan:ICLR2014}                   & \includegraphics[width=.12\linewidth,valign=m]{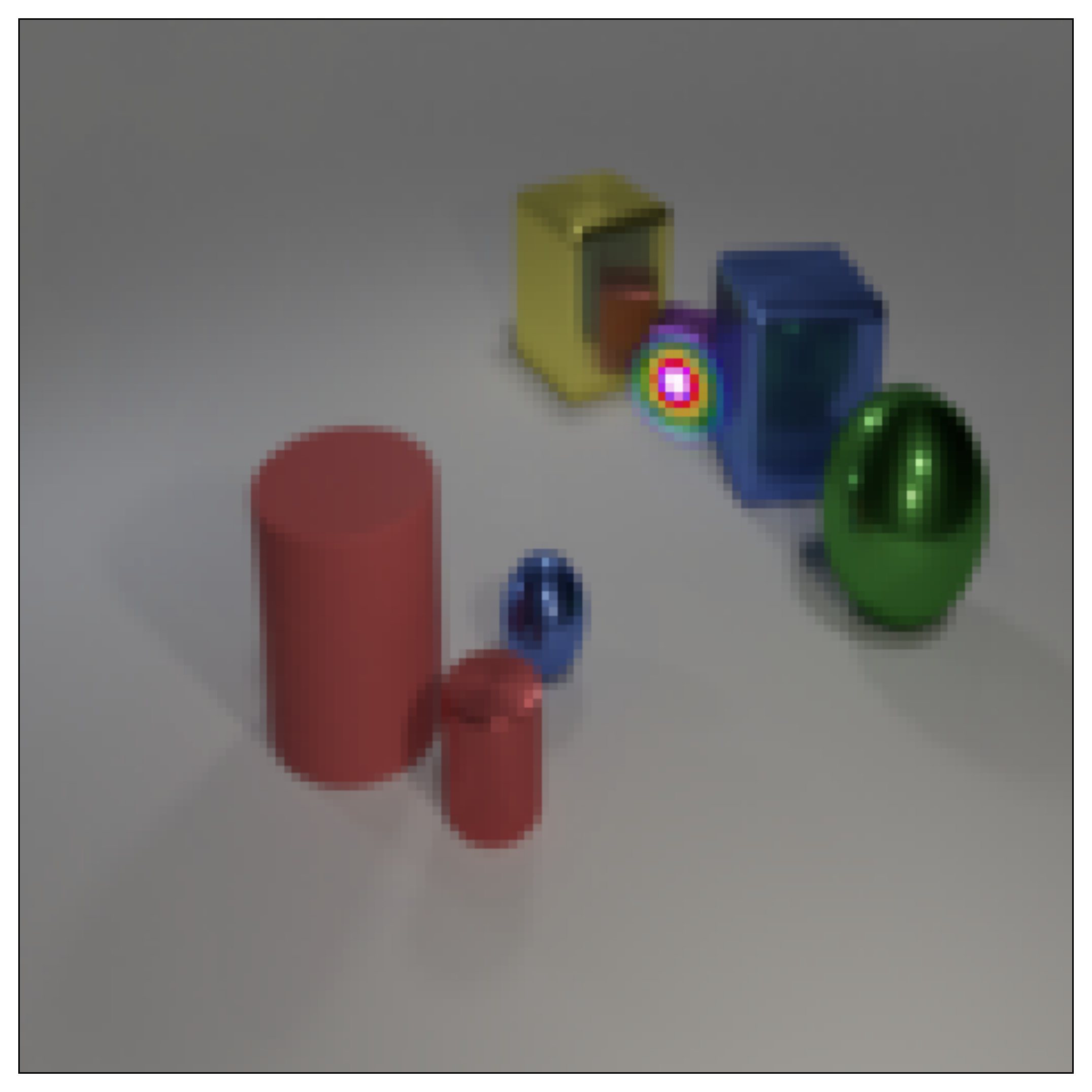} & \includegraphics[width=.12\linewidth,valign=m]{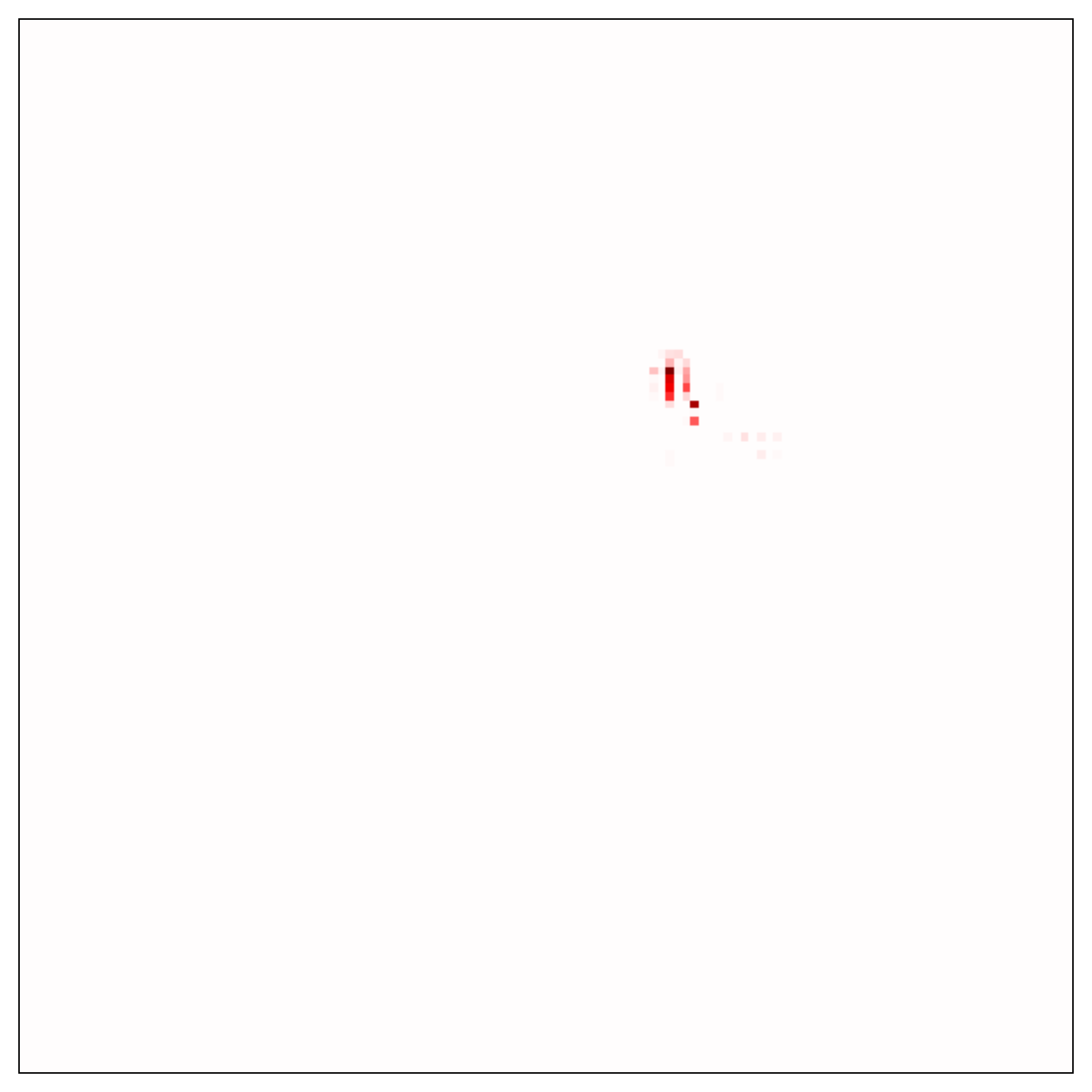} & 0.94 \\
Gradient$\times$Input \cite{Shrikumar:arxiv2016}    & \includegraphics[width=.12\linewidth,valign=m]{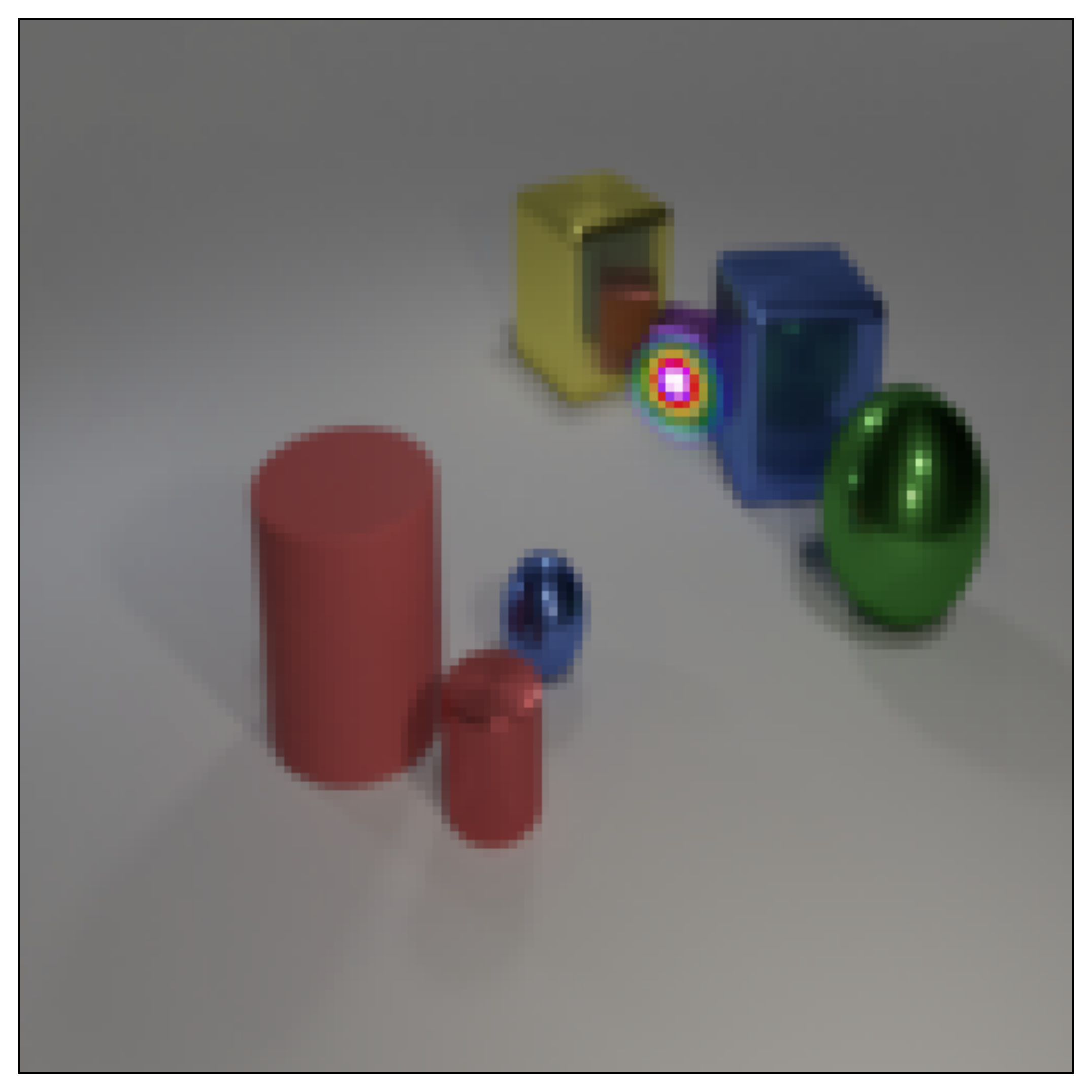} & \includegraphics[width=.12\linewidth,valign=m]{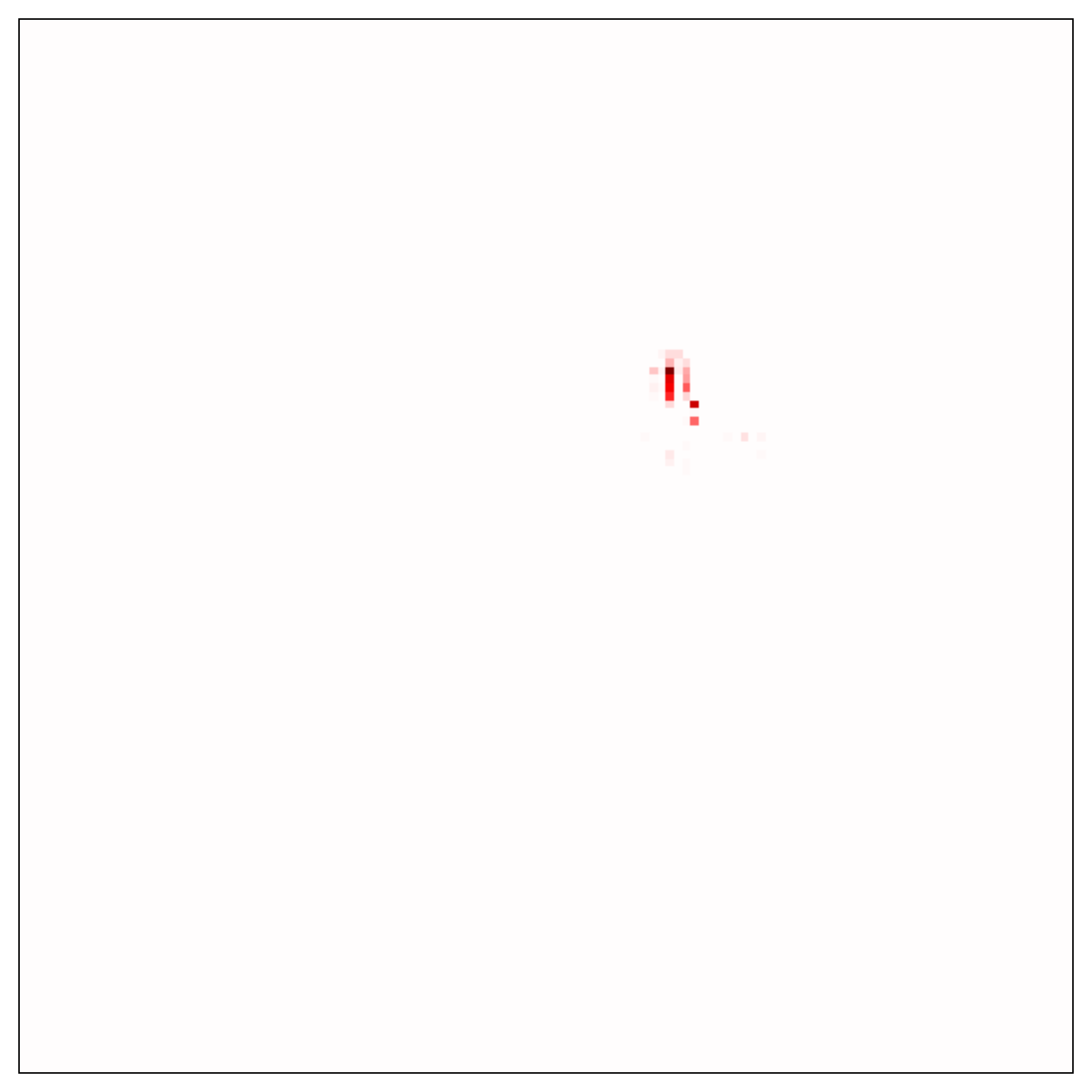} & 0.94 \\
Deconvnet \cite{Zeiler:ECCV2014}                    & \includegraphics[width=.12\linewidth,valign=m]{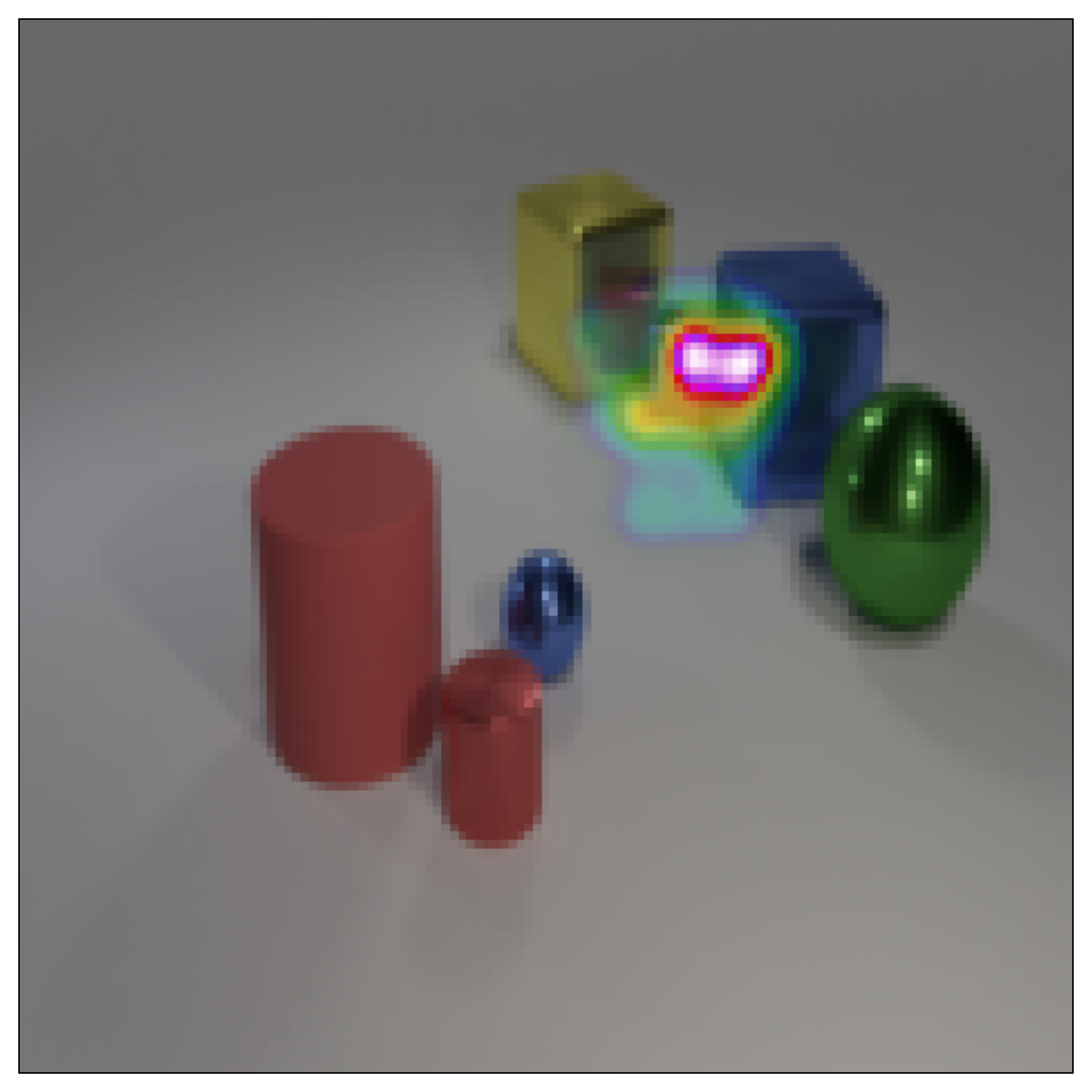} & \includegraphics[width=.12\linewidth,valign=m]{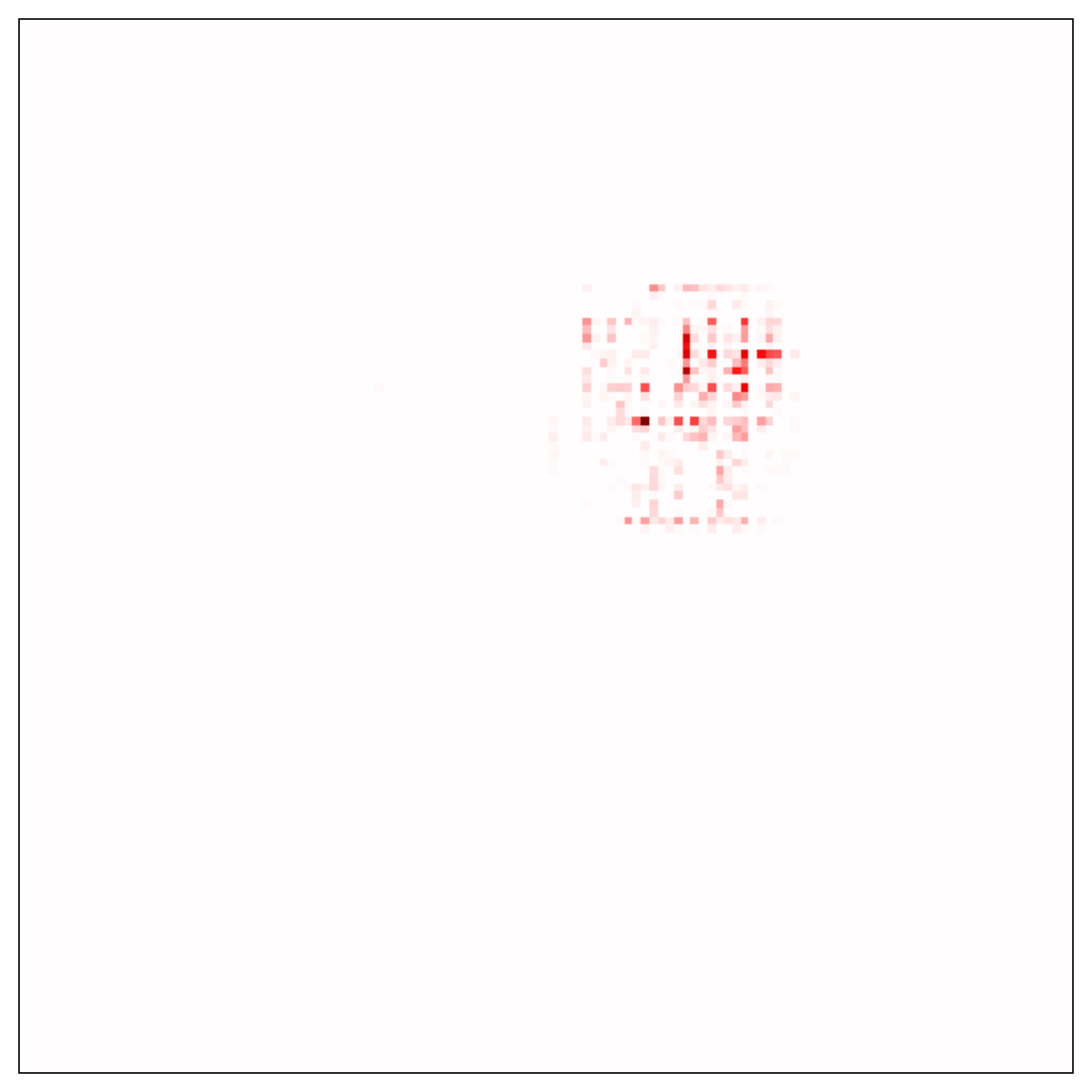} & 0.30 \\
Grad-CAM \cite{Selvaraju:ICCV2017}                  & \includegraphics[width=.12\linewidth,valign=m]{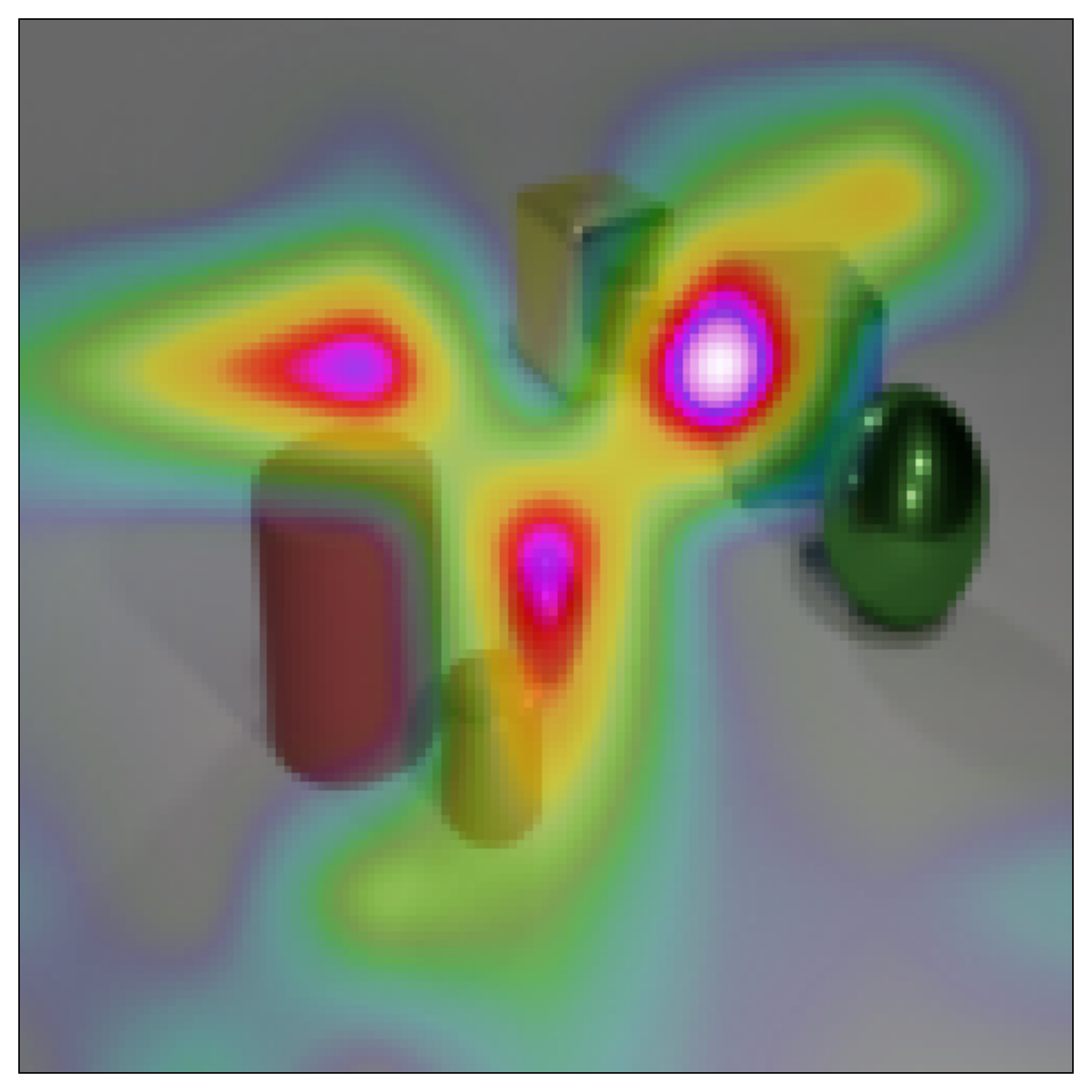} & \includegraphics[width=.12\linewidth,valign=m]{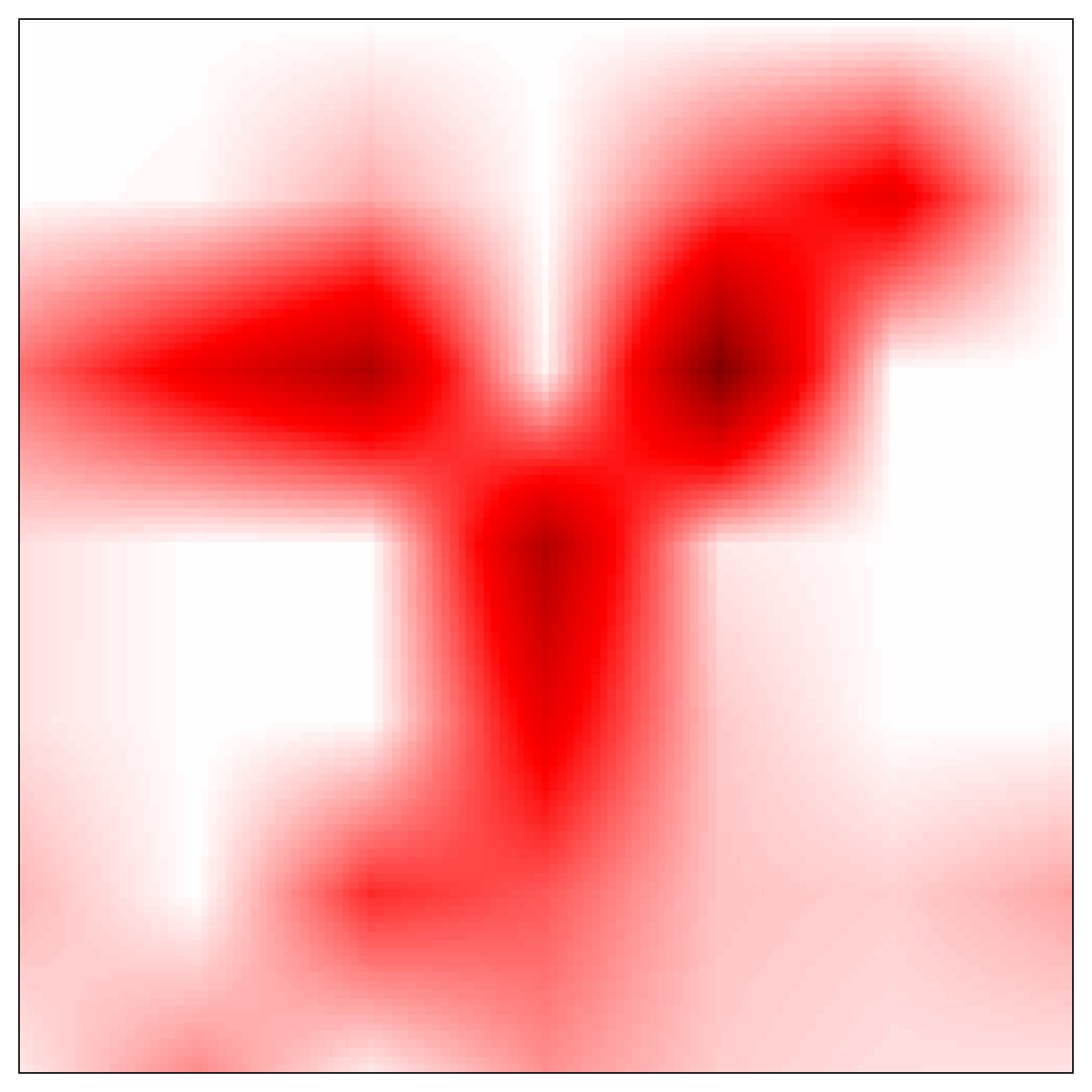} & 0.03 \\
\end{tabular}
\end{table}

\begin{table}
        \scriptsize
		\caption{Heatmaps for a correctly predicted CLEVR-XAI-complex question (raw heatmap and heatmap overlayed with original image), and relevance \textit{mass} accuracy.}
		\label{table:heatmap-complex-correct-65806}
\begin{tabular}{lllc}
\midrule
\begin{tabular}{@{}l@{}}\scriptsize What material is the large object that\\ is left of the big purple metallic ball? \\ \textit{metal} \end{tabular}  & \includegraphics[width=.18\linewidth,valign=m]{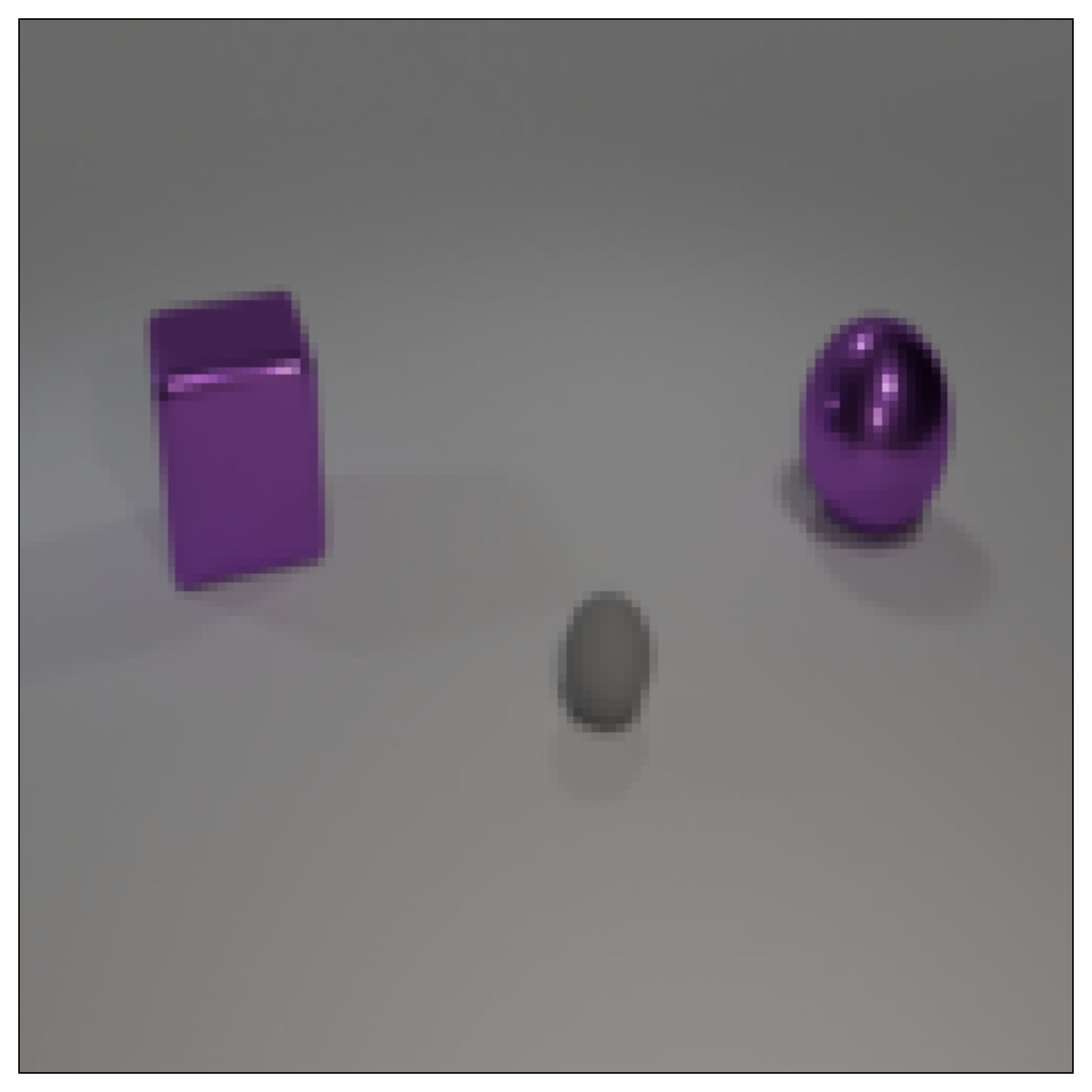} &
\includegraphics[width=.18\linewidth,valign=m]{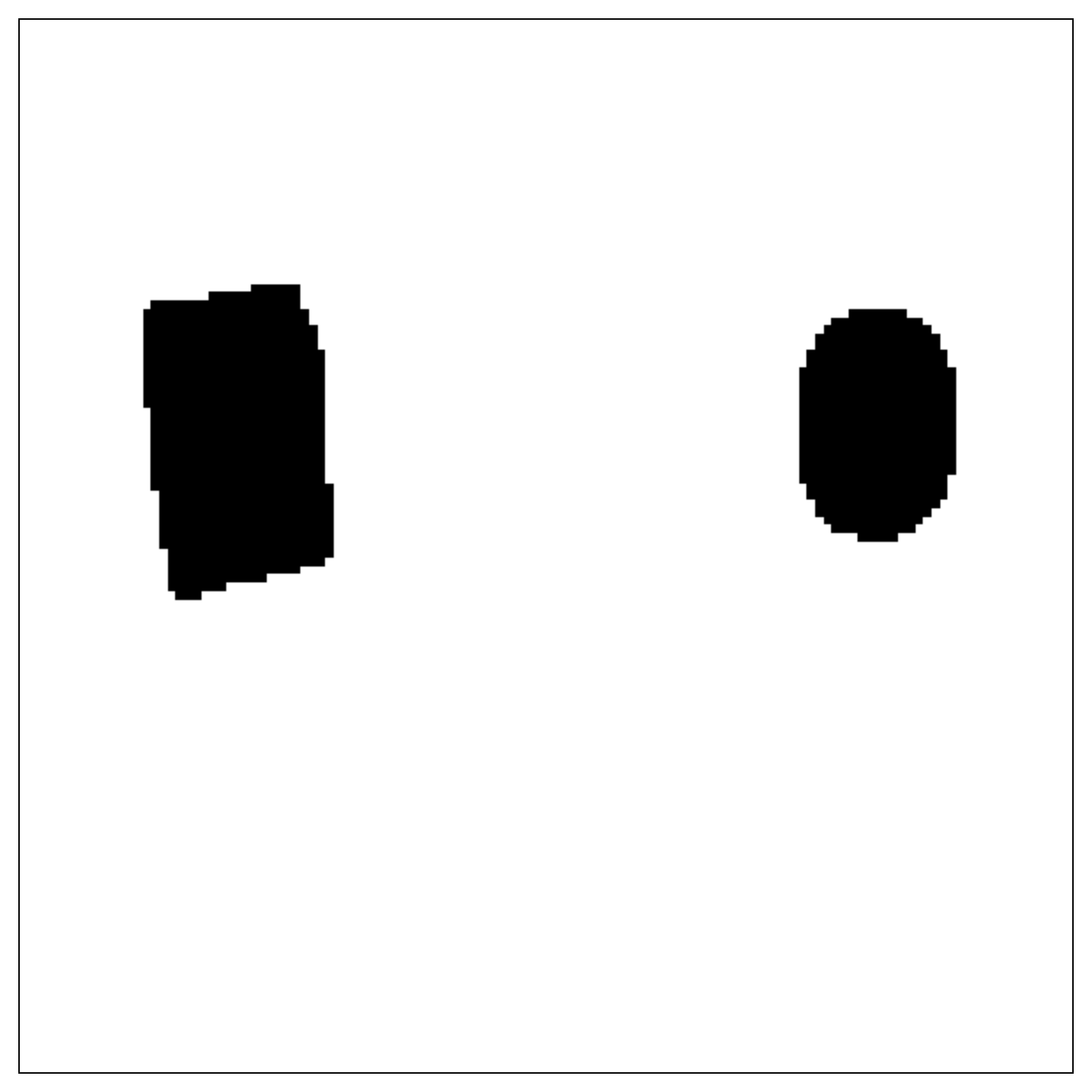} & {\tiny GT Unique First-non-empty} \\
\midrule
LRP \cite{Bach:PLOS2015}                            & \includegraphics[width=.12\linewidth,valign=m]{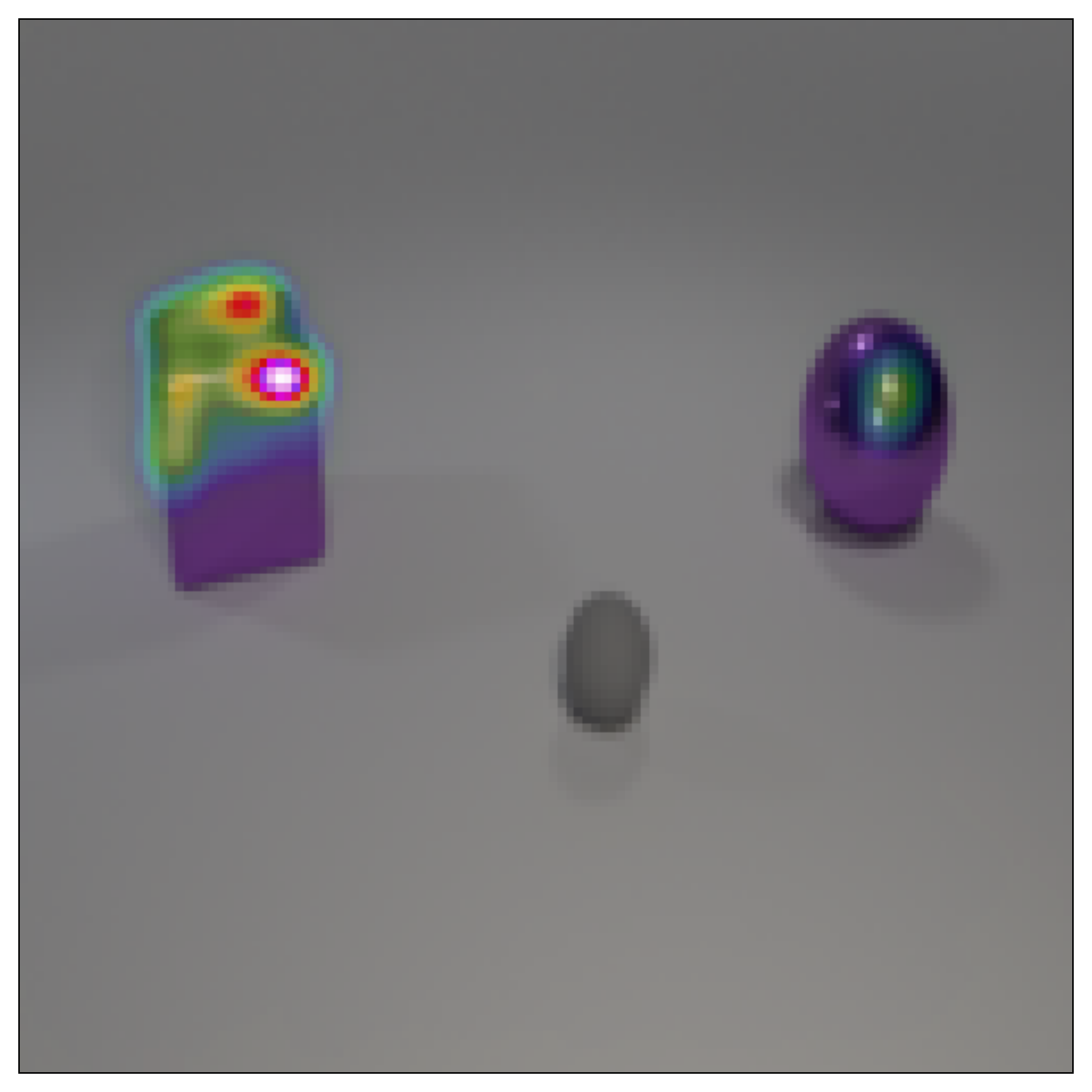} & \includegraphics[width=.12\linewidth,valign=m]{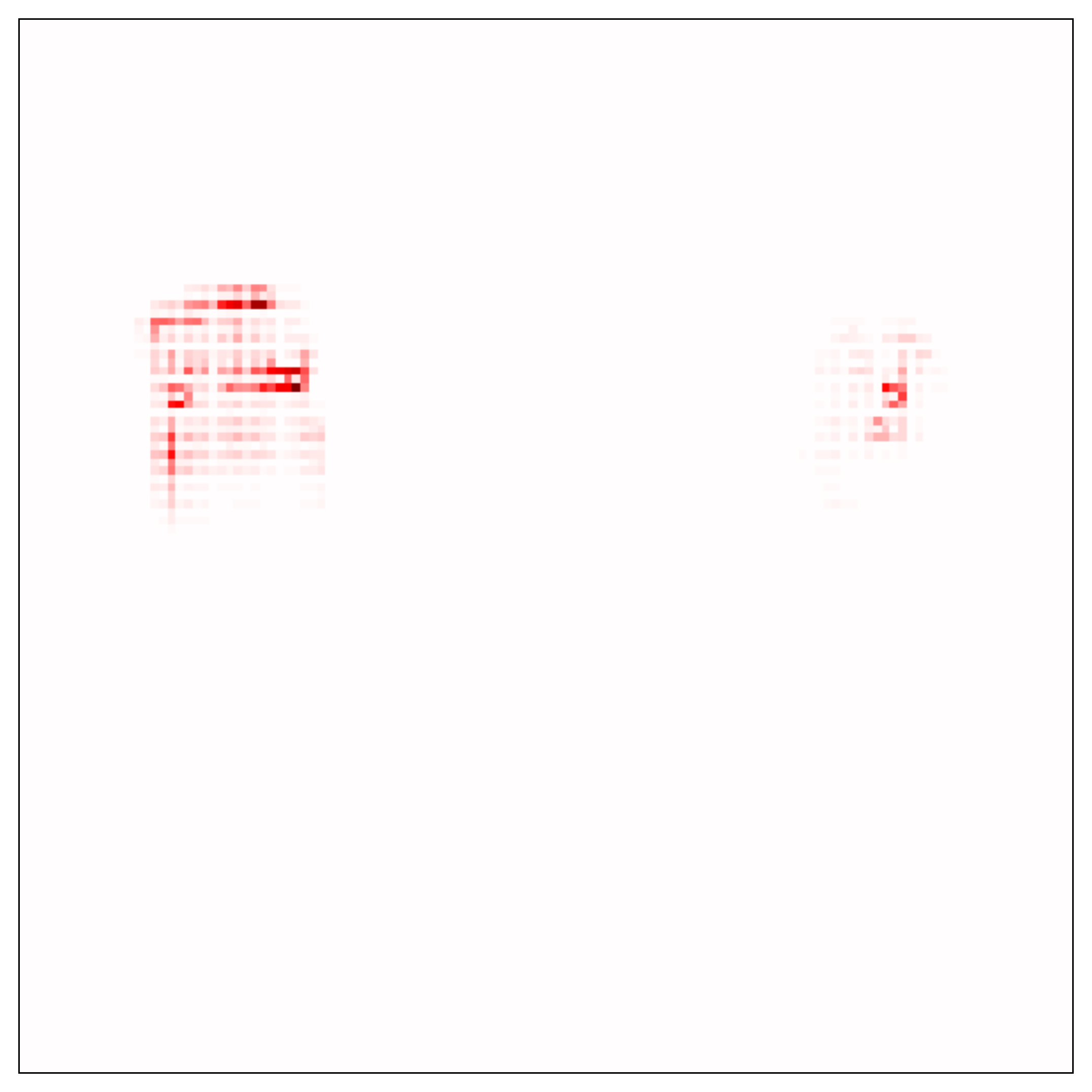} & 0.97 \\
Excitation Backprop \cite{Zhang:ECCV2016}           & \includegraphics[width=.12\linewidth,valign=m]{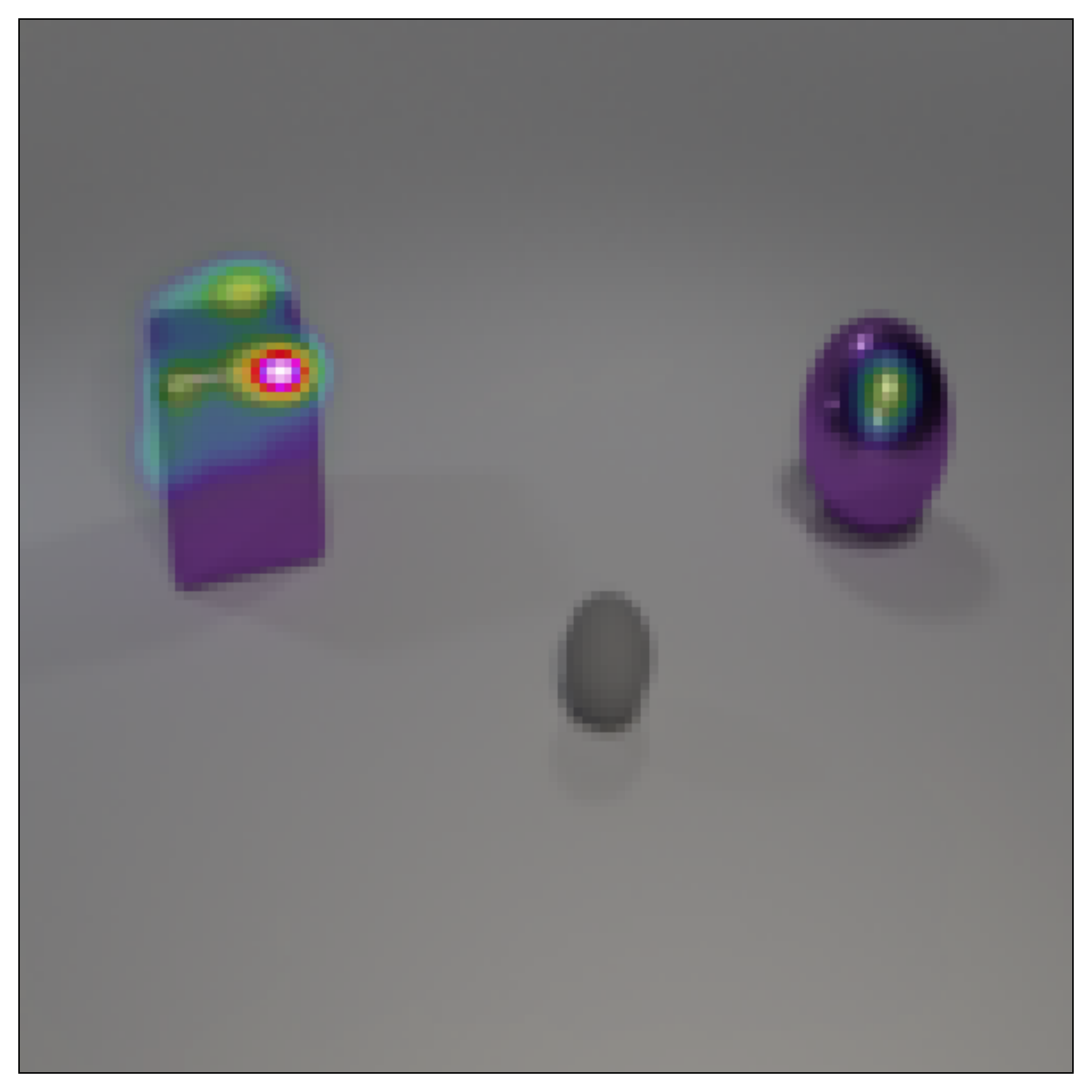} & \includegraphics[width=.12\linewidth,valign=m]{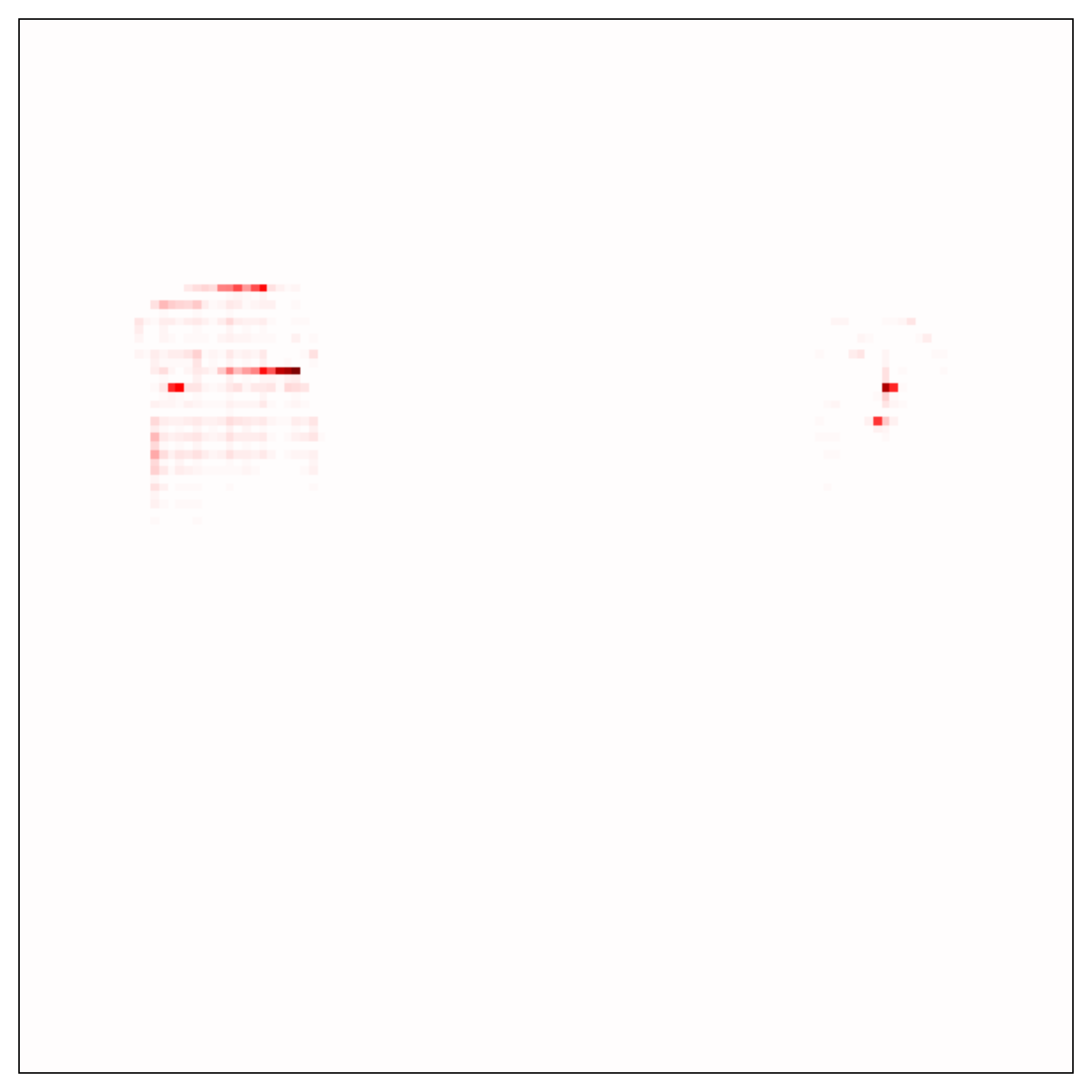} & 0.91 \\
IG \cite{Sundararajan:ICML2017}                     & \includegraphics[width=.12\linewidth,valign=m]{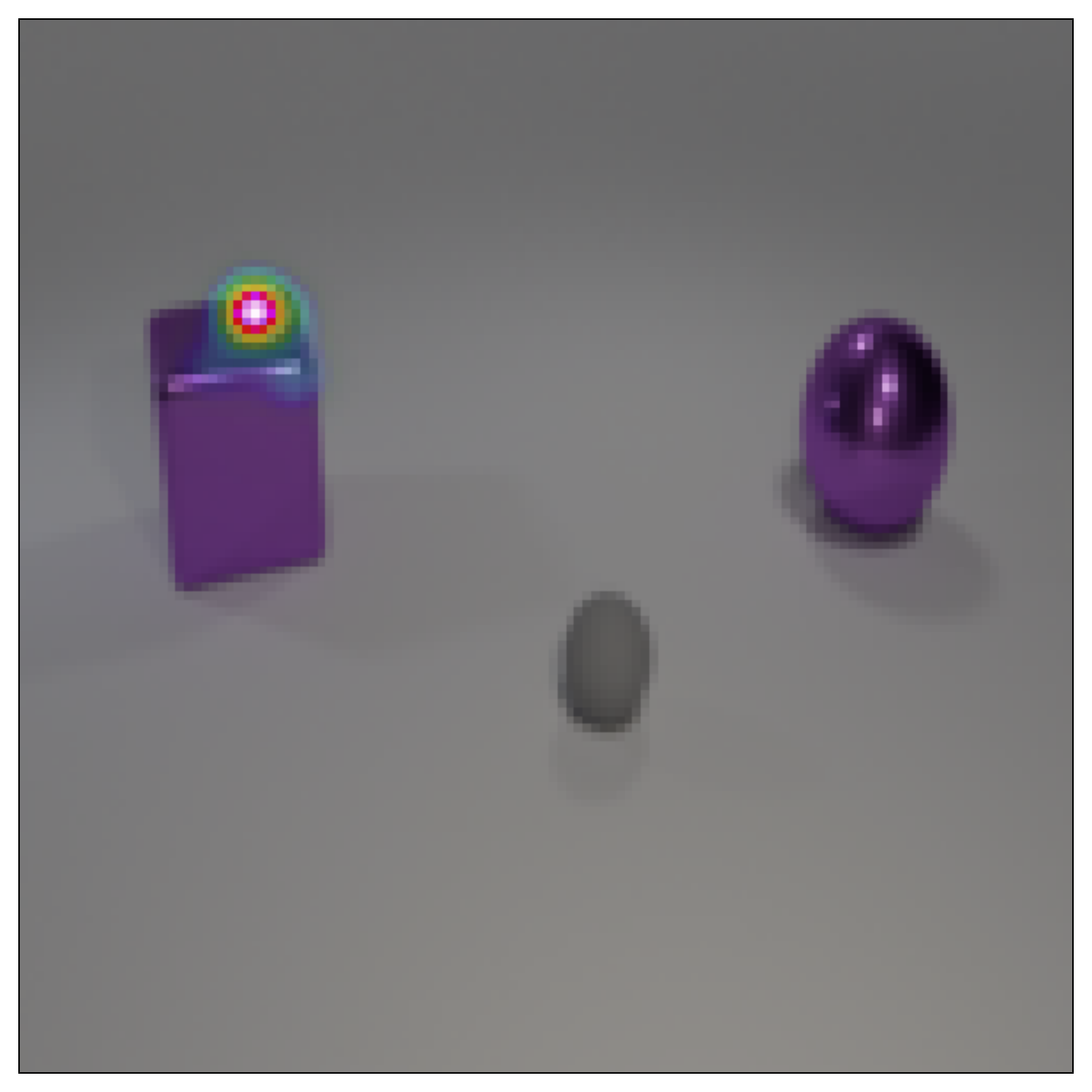} & \includegraphics[width=.12\linewidth,valign=m]{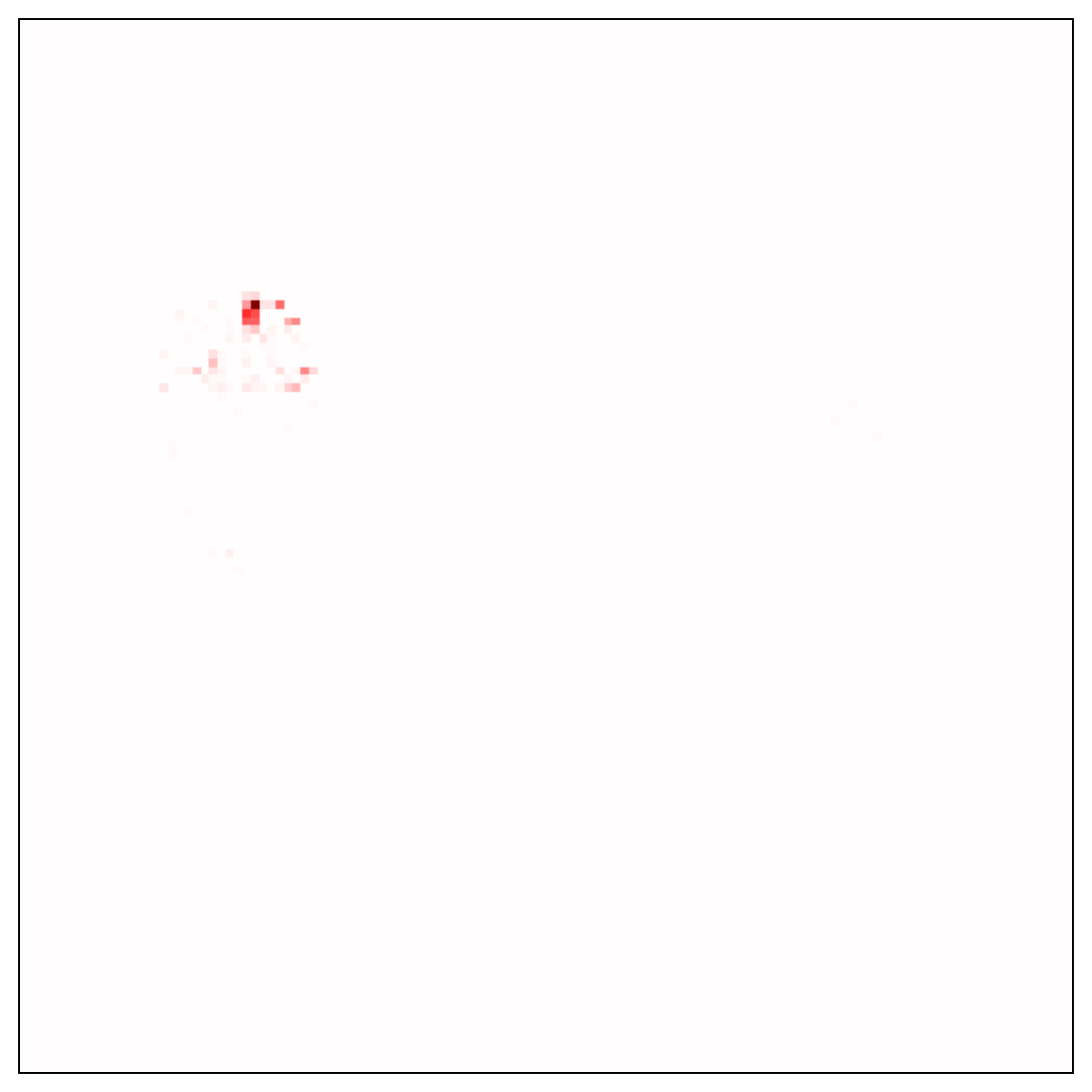} & 0.99 \\
Guided Backprop \cite{Spring:ICLR2015}              & \includegraphics[width=.12\linewidth,valign=m]{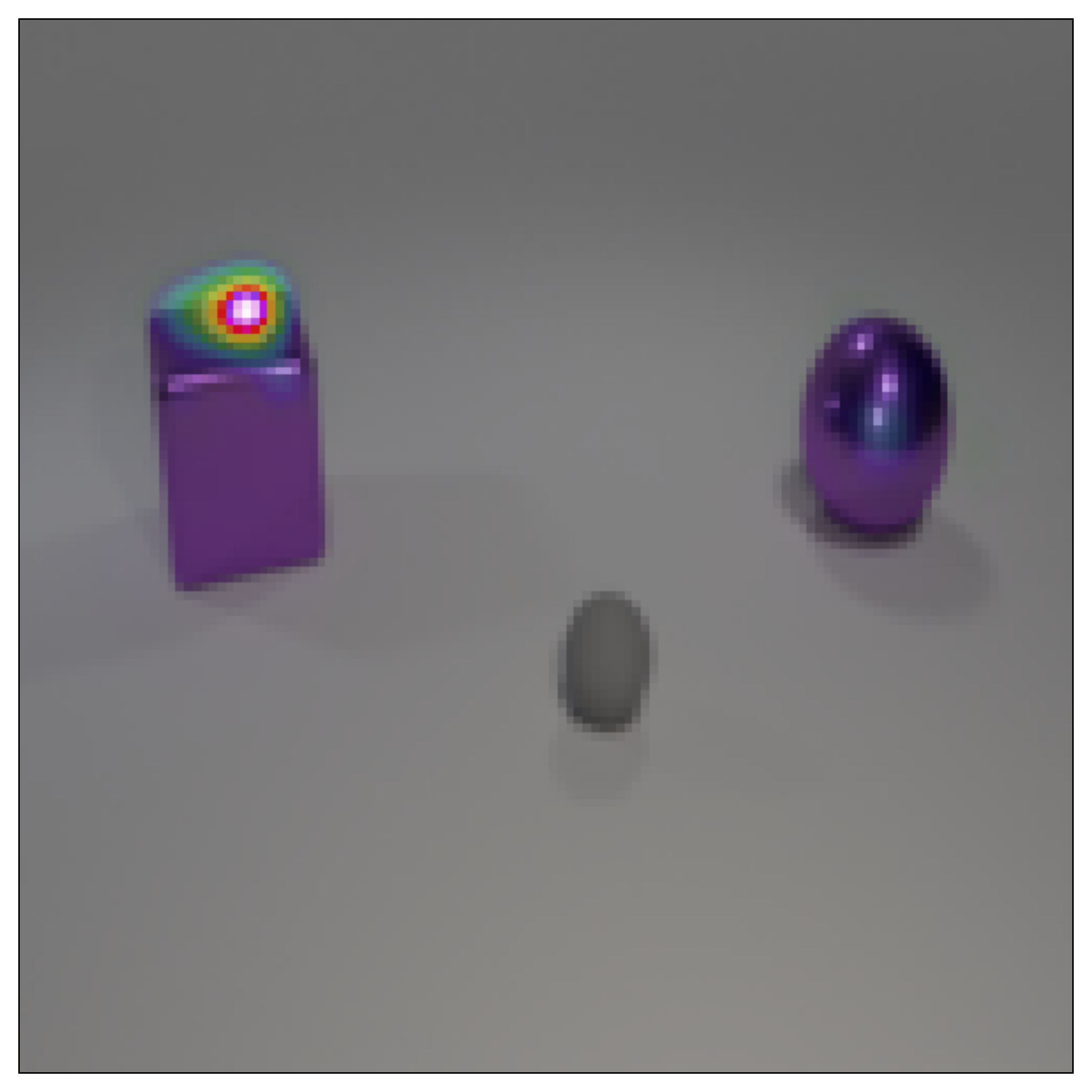} & \includegraphics[width=.12\linewidth,valign=m]{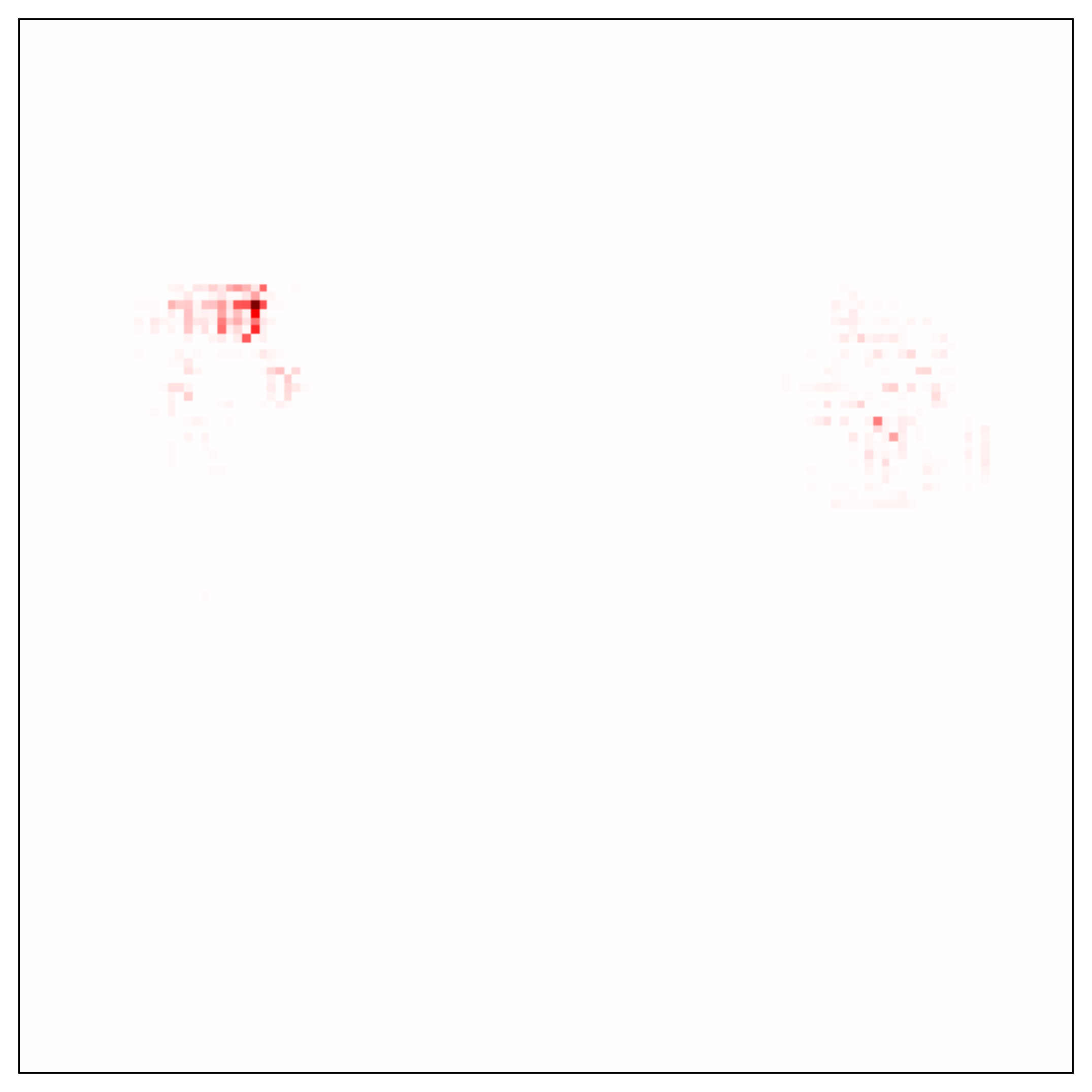} & 0.88 \\
Guided Grad-CAM \cite{Selvaraju:ICCV2017}           & \includegraphics[width=.12\linewidth,valign=m]{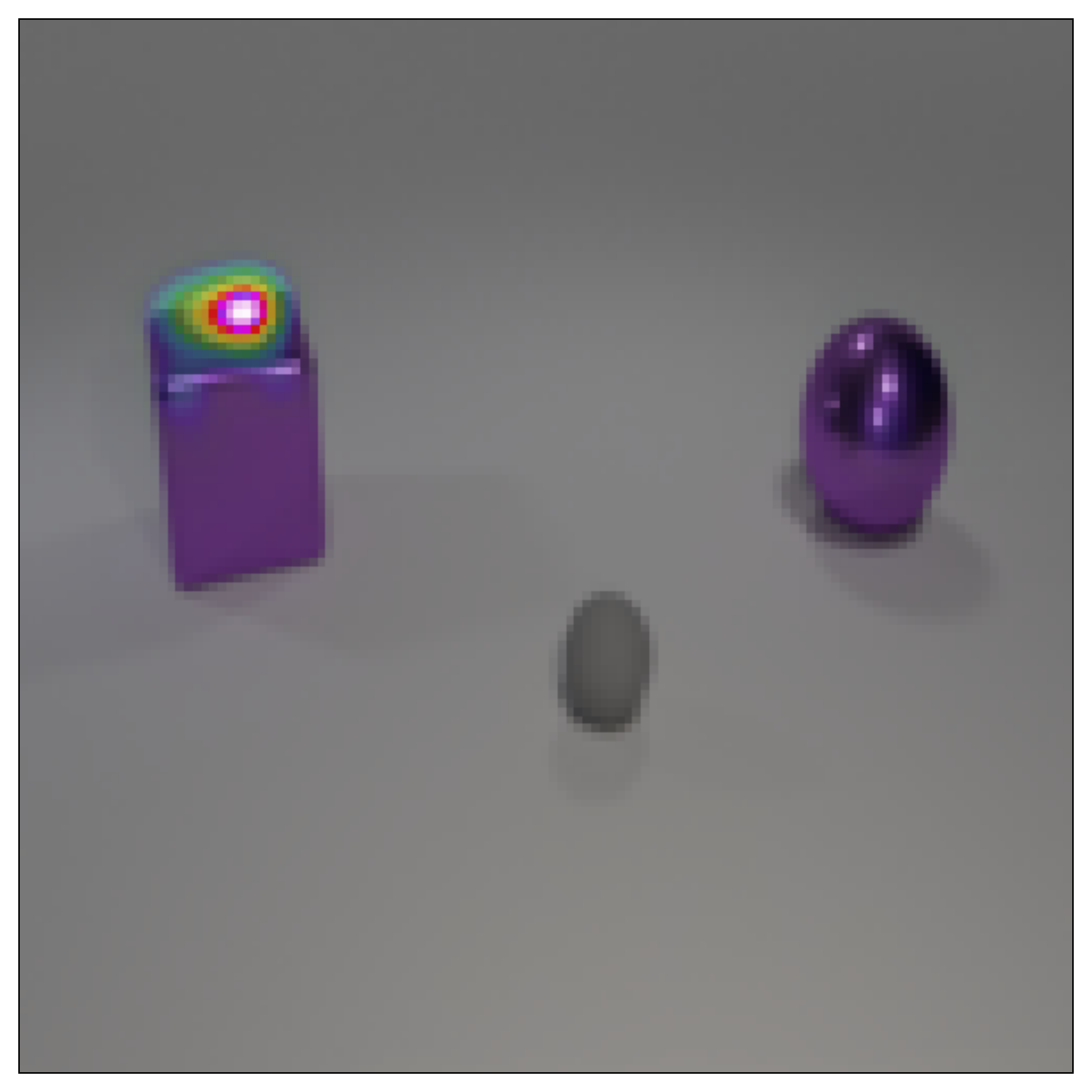} & \includegraphics[width=.12\linewidth,valign=m]{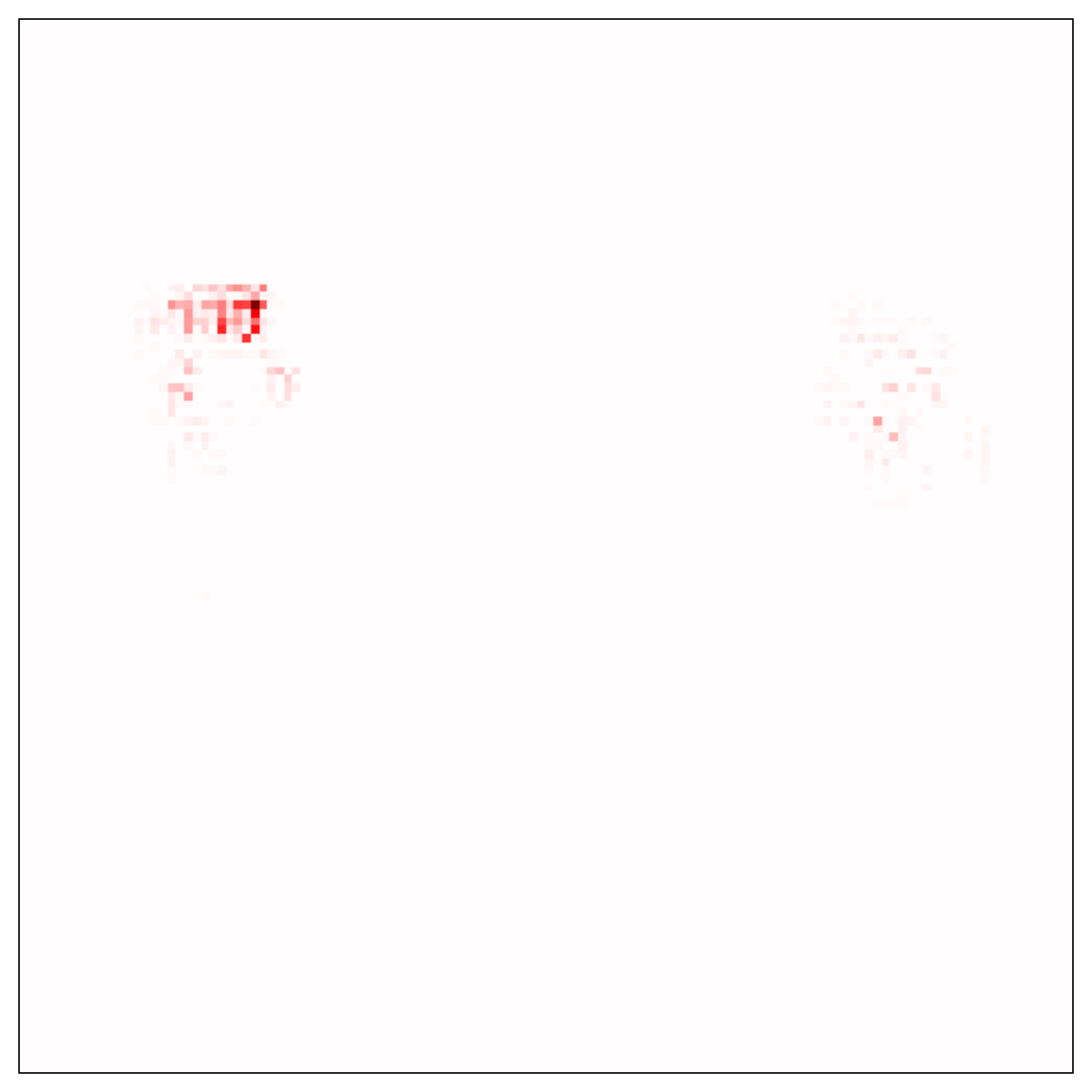} & 0.90 \\
SmoothGrad \cite{Smilkov:ICML2017}                  & \includegraphics[width=.12\linewidth,valign=m]{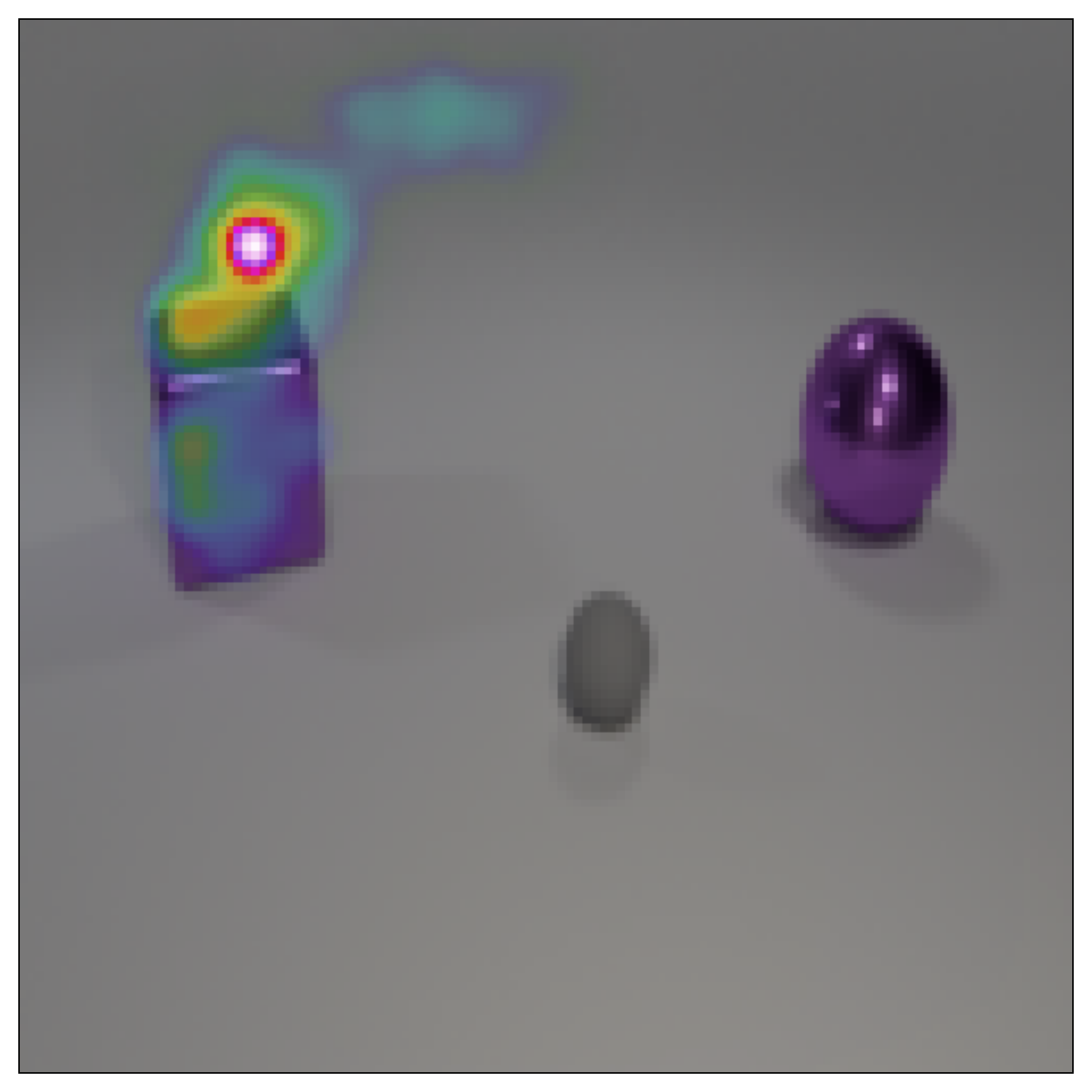} & \includegraphics[width=.12\linewidth,valign=m]{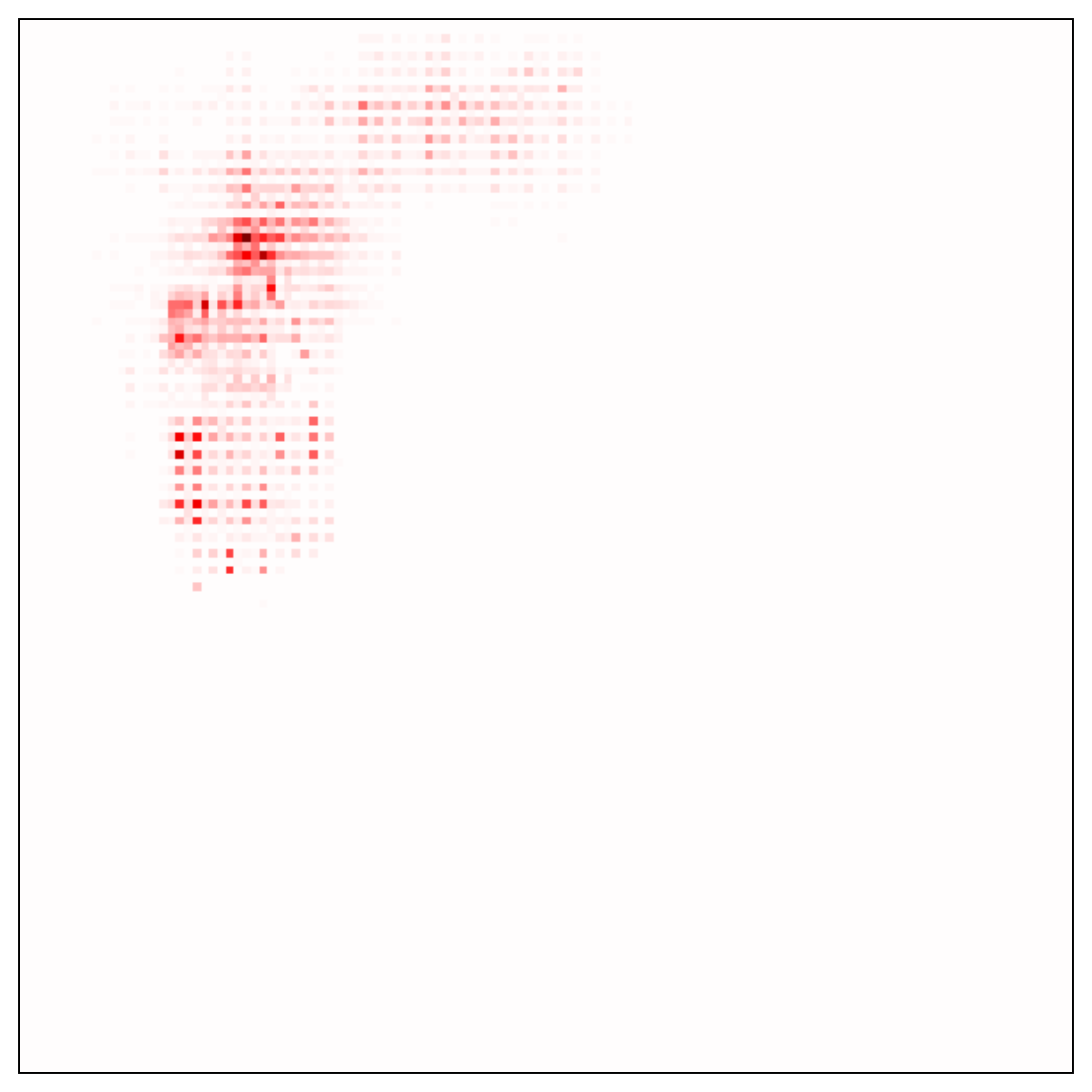} & 0.42 \\
VarGrad \cite{Adebayo:ICLR2018}                     & \includegraphics[width=.12\linewidth,valign=m]{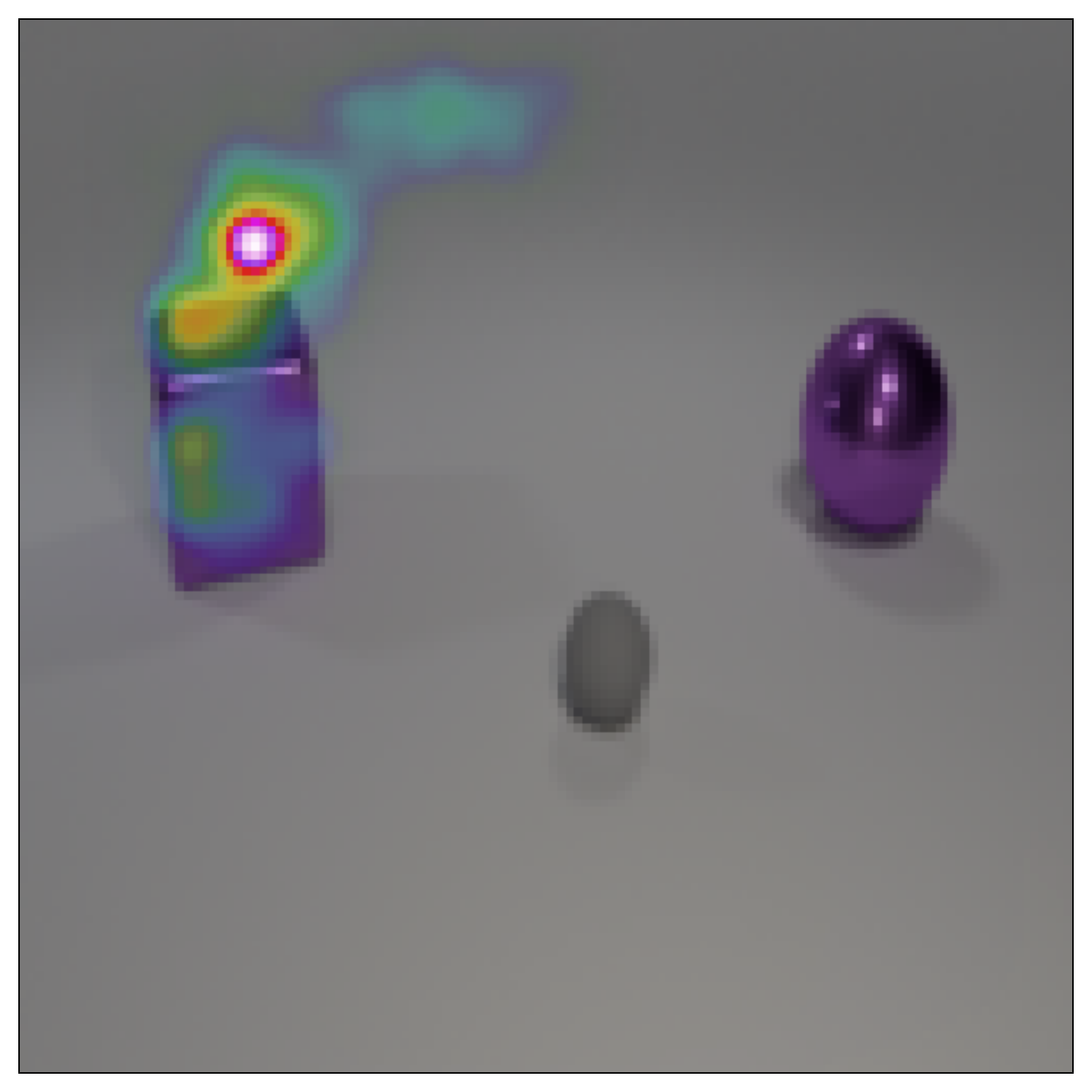} & \includegraphics[width=.12\linewidth,valign=m]{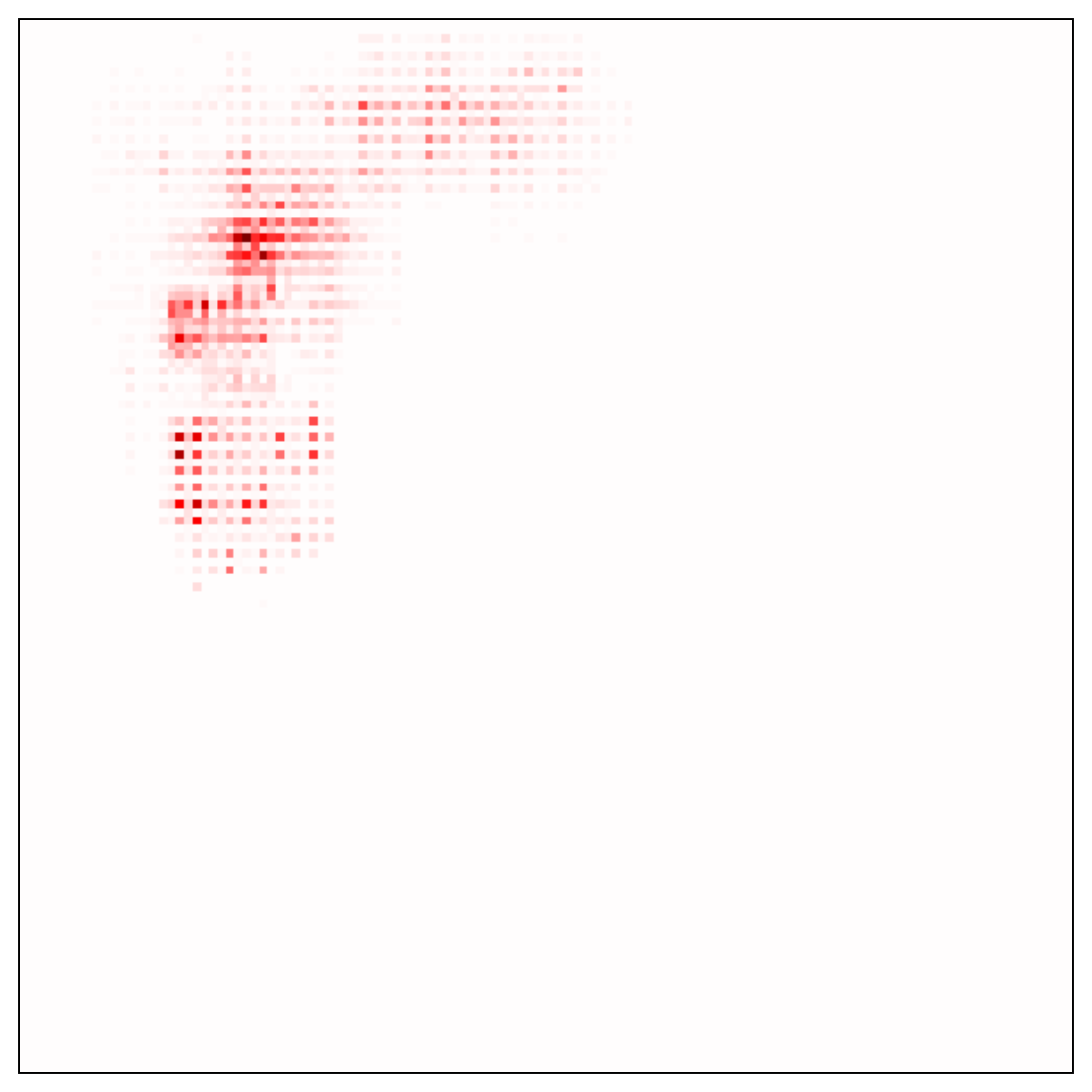} & 0.40 \\
Gradient \cite{Simonyan:ICLR2014}                   & \includegraphics[width=.12\linewidth,valign=m]{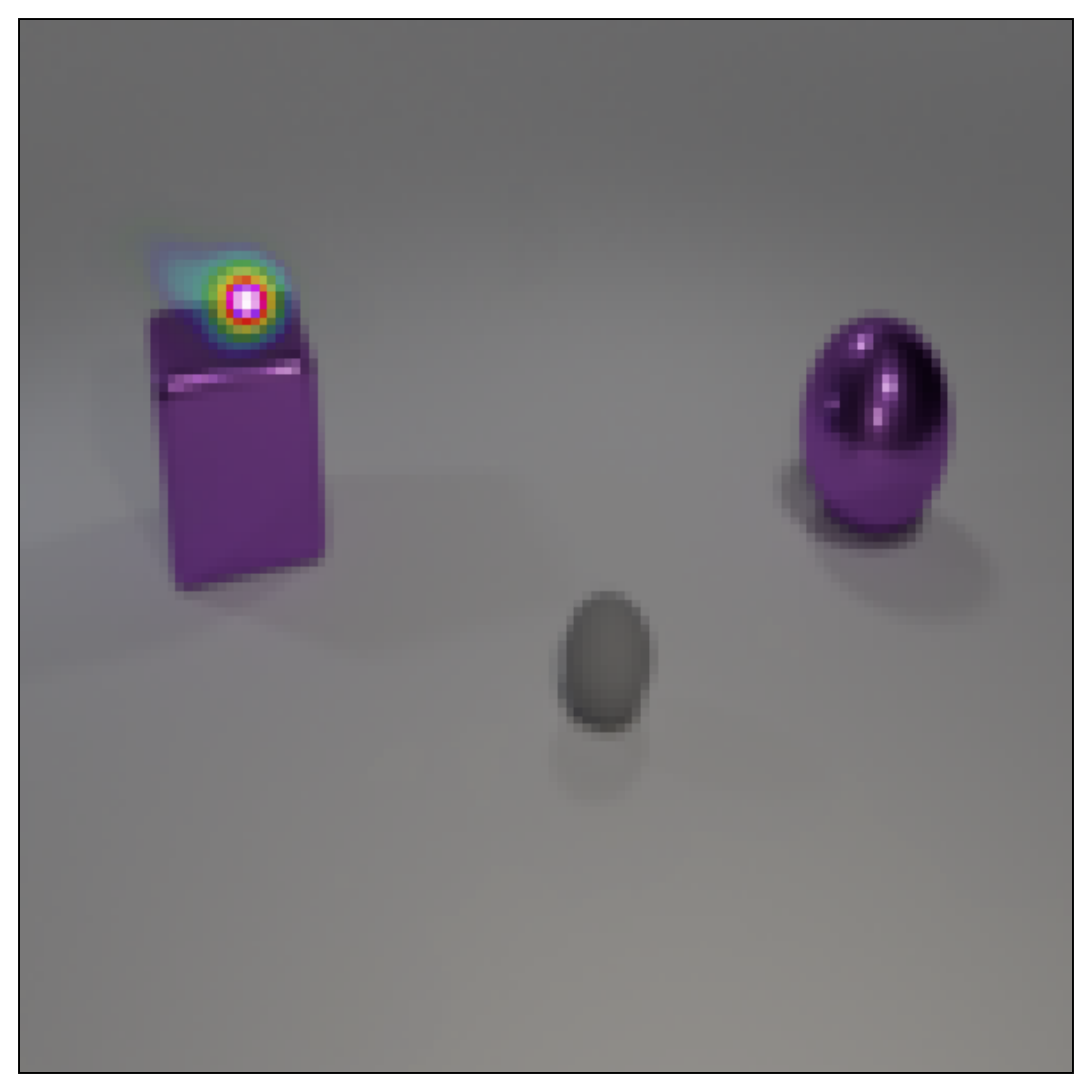} & \includegraphics[width=.12\linewidth,valign=m]{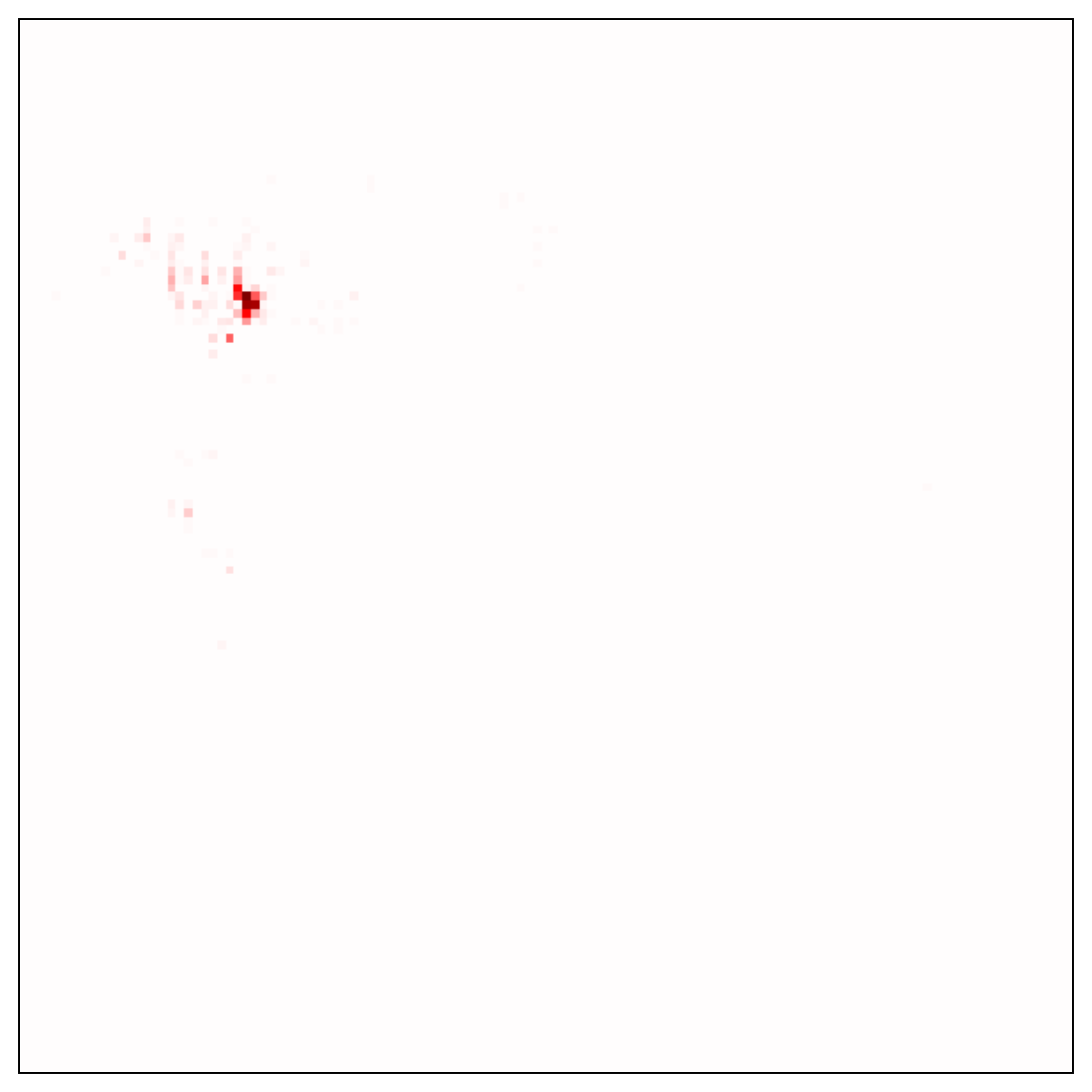} & 0.63 \\
Gradient$\times$Input \cite{Shrikumar:arxiv2016}    & \includegraphics[width=.12\linewidth,valign=m]{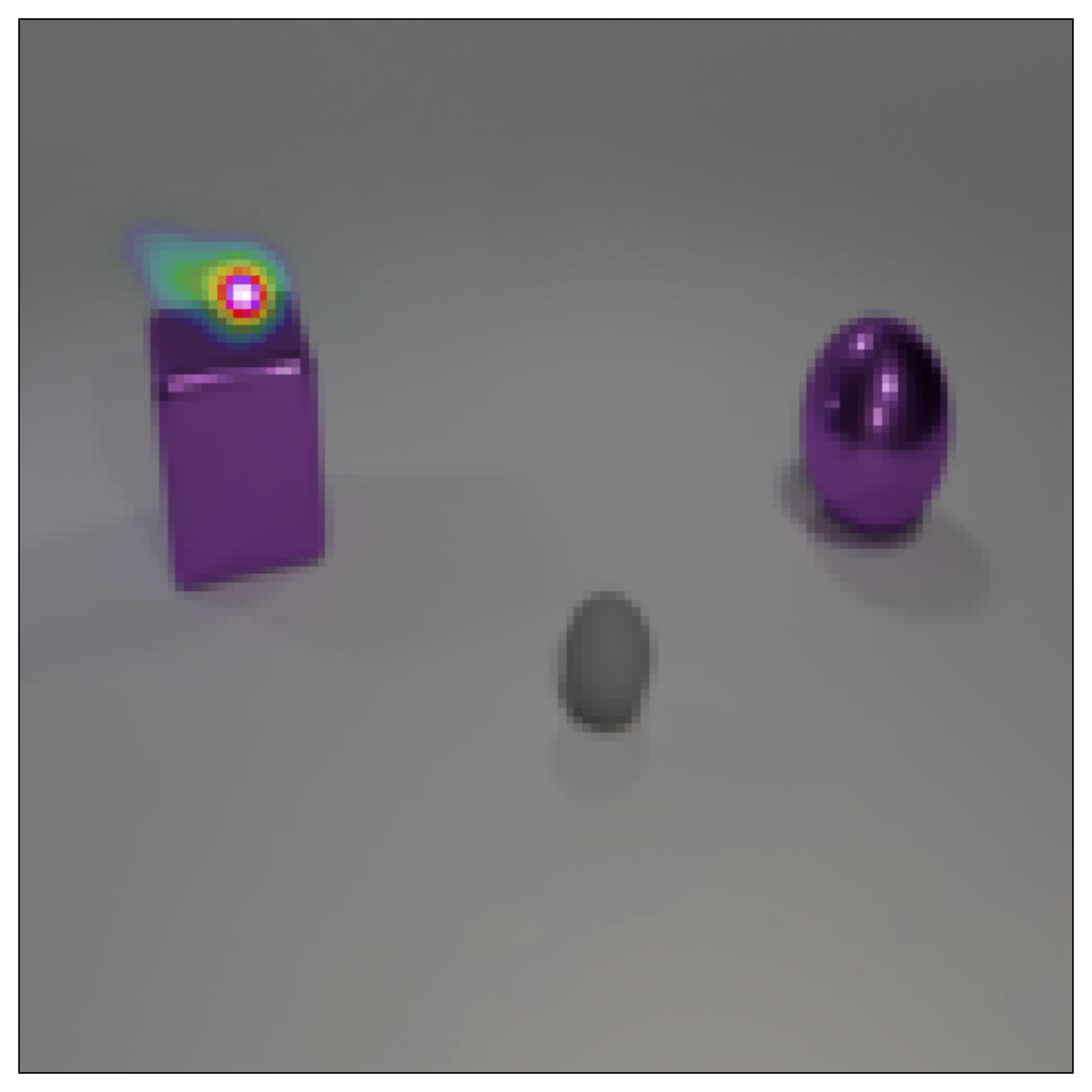} & \includegraphics[width=.12\linewidth,valign=m]{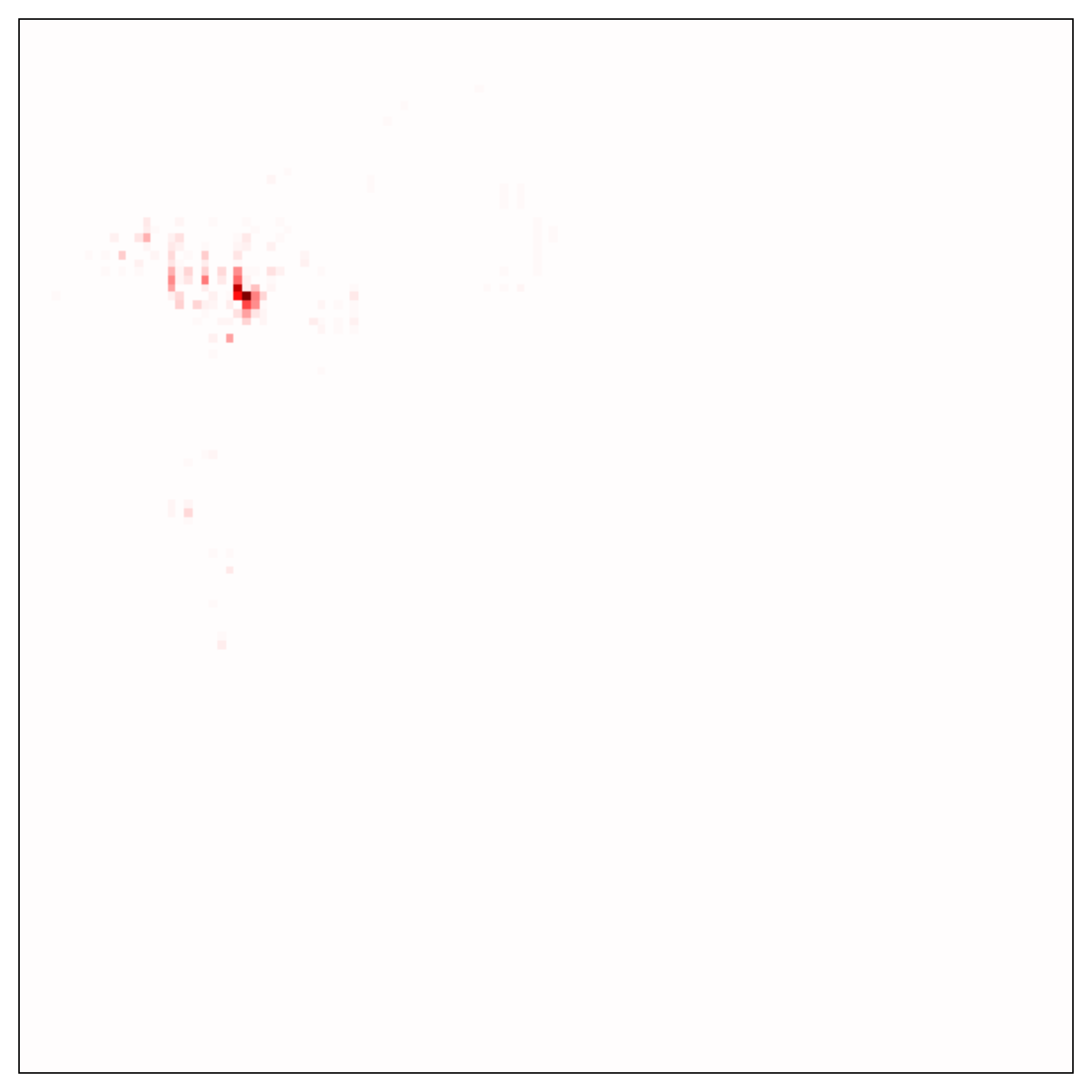} & 0.42 \\
Deconvnet \cite{Zeiler:ECCV2014}                    & \includegraphics[width=.12\linewidth,valign=m]{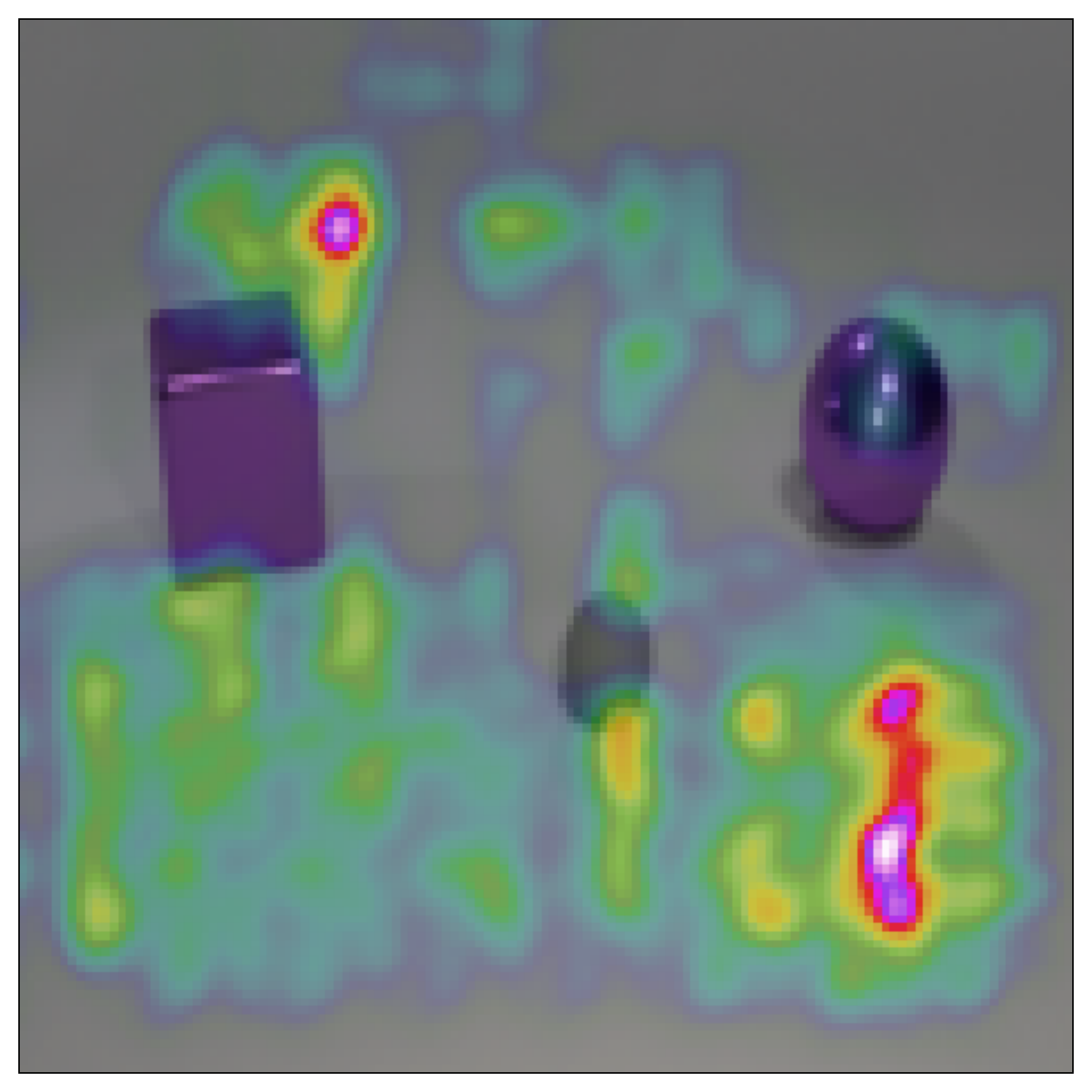} & \includegraphics[width=.12\linewidth,valign=m]{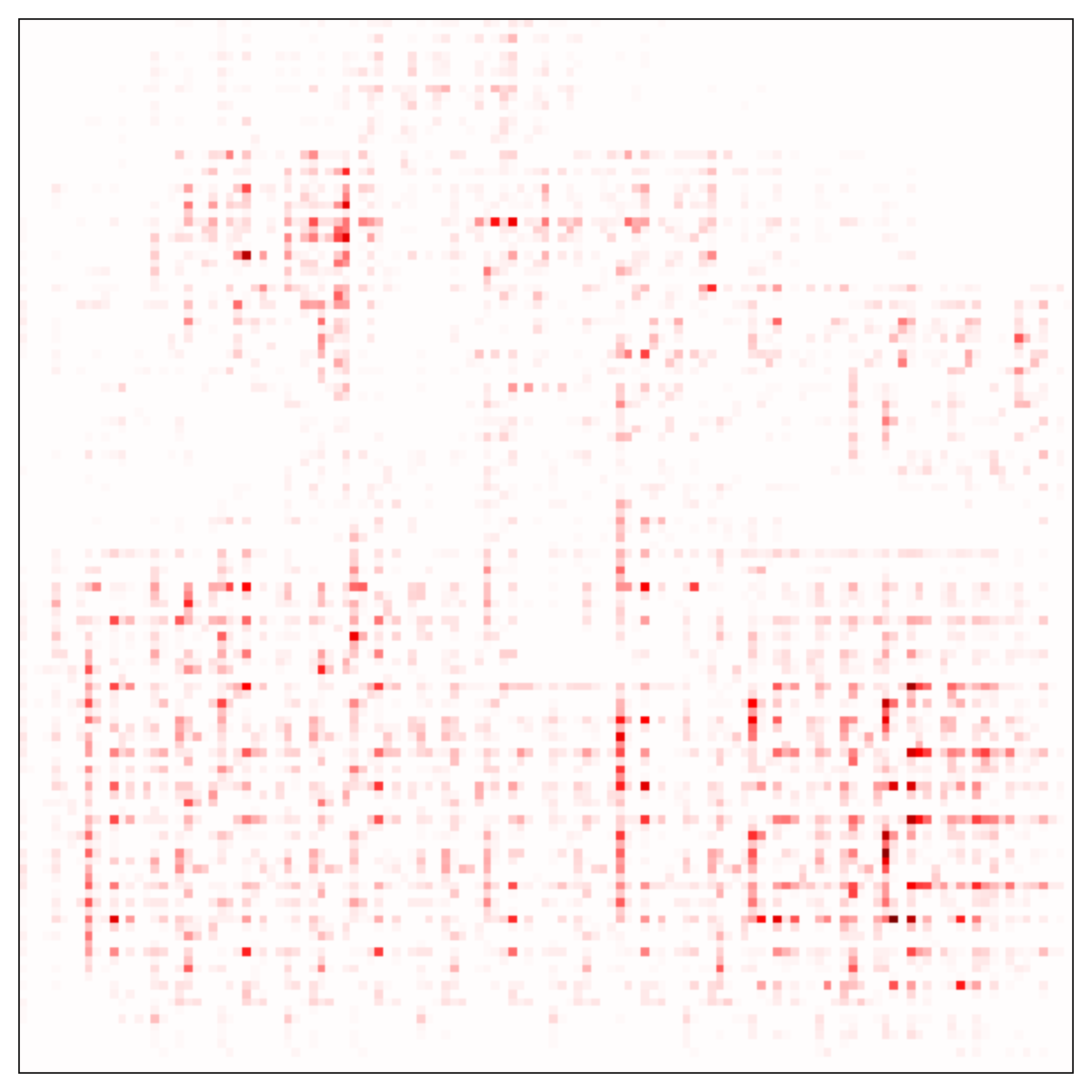} & 0.04 \\
Grad-CAM \cite{Selvaraju:ICCV2017}                  & \includegraphics[width=.12\linewidth,valign=m]{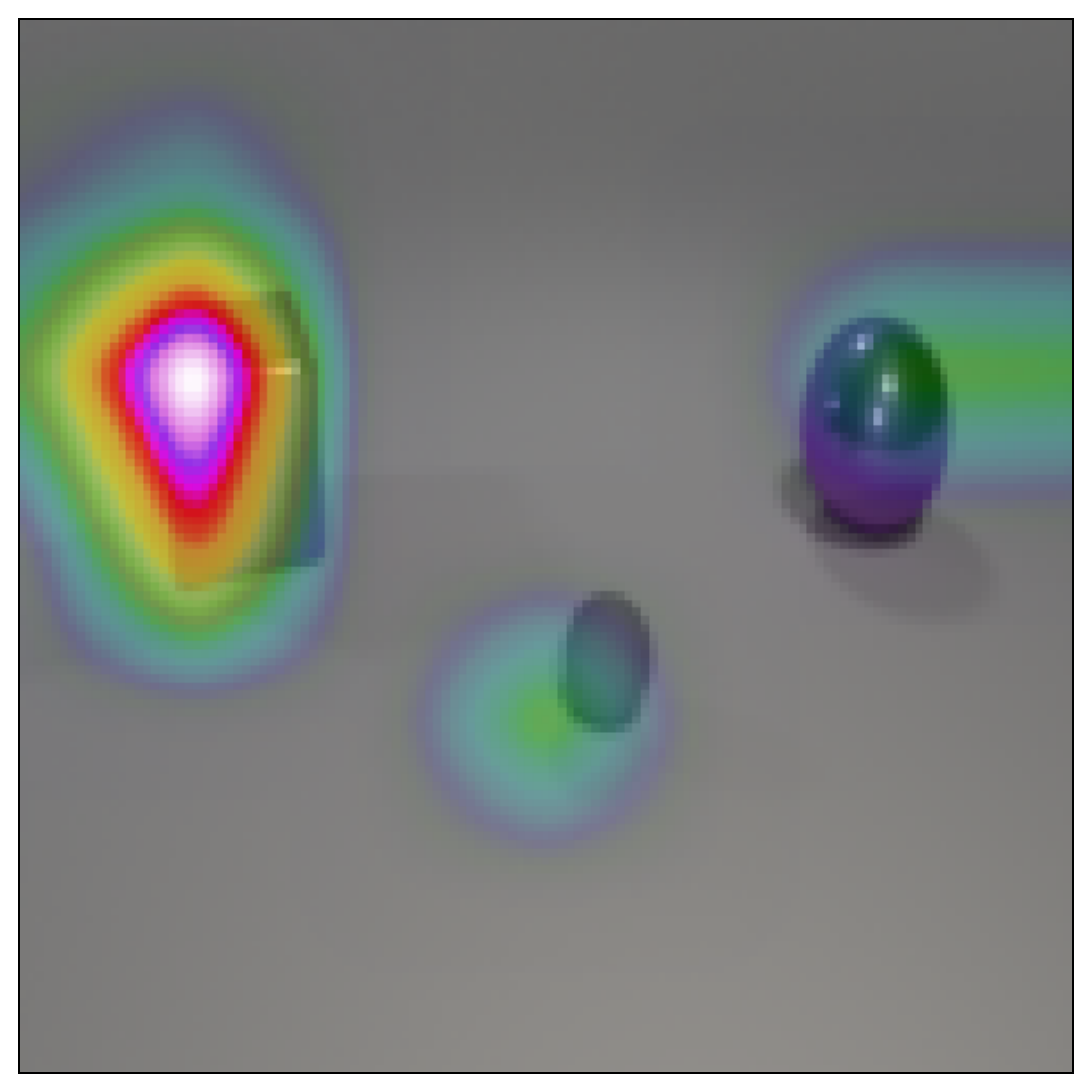} & \includegraphics[width=.12\linewidth,valign=m]{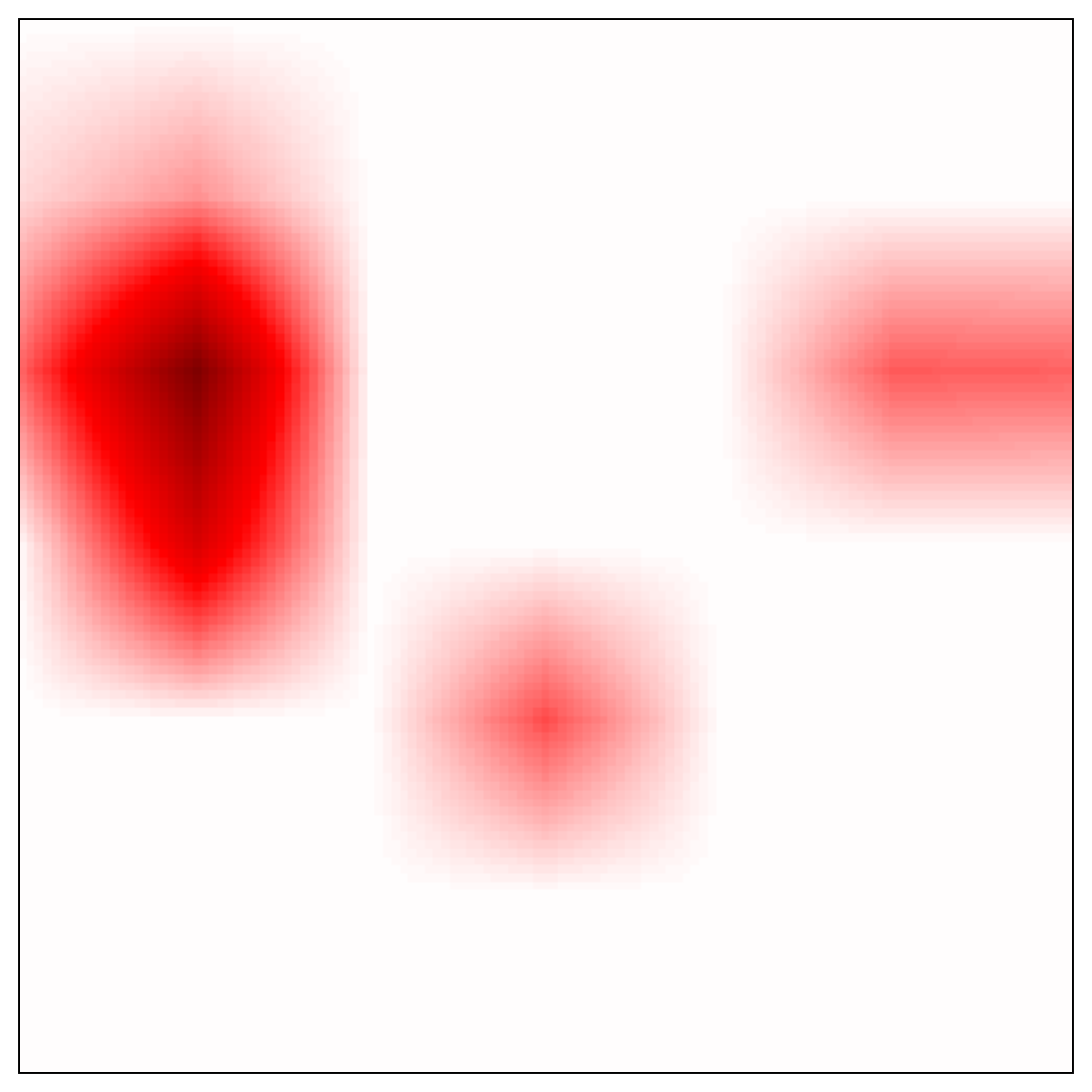} & 0.38 \\
\end{tabular}
\end{table}

\begin{table}
        \scriptsize
		\caption{Heatmaps for a correctly predicted CLEVR-XAI-complex question (raw heatmap and heatmap overlayed with original image), and relevance \textit{mass} accuracy.}
		\label{table:heatmap-complex-correct-12504}
\begin{tabular}{lllc}
\midrule
\begin{tabular}{@{}l@{}}\scriptsize There is a large purple cube that is behind\\ the brown shiny cylinder; what material is it? \\ \textit{metal} \end{tabular}  & \includegraphics[width=.18\linewidth,valign=m]{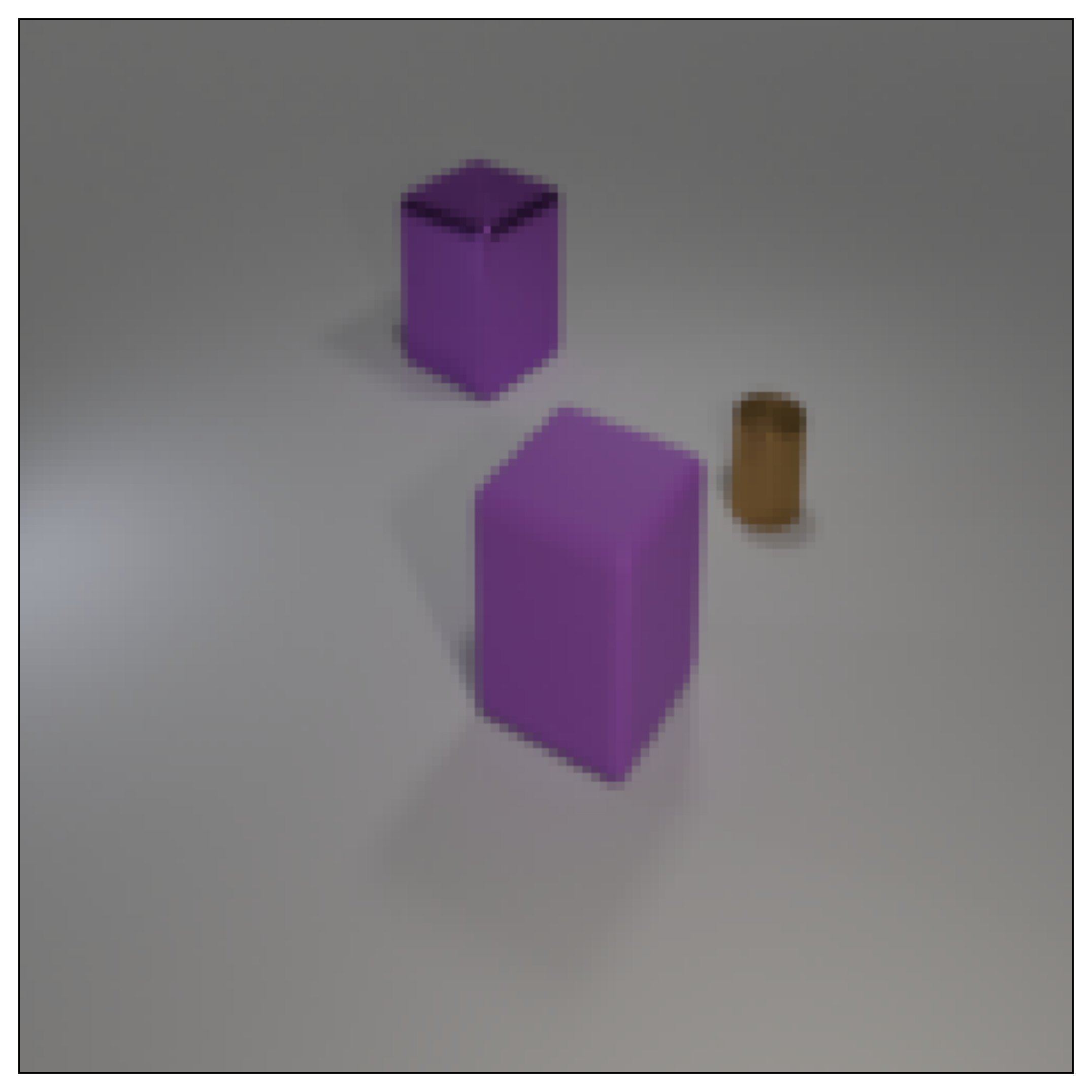} &
\includegraphics[width=.18\linewidth,valign=m]{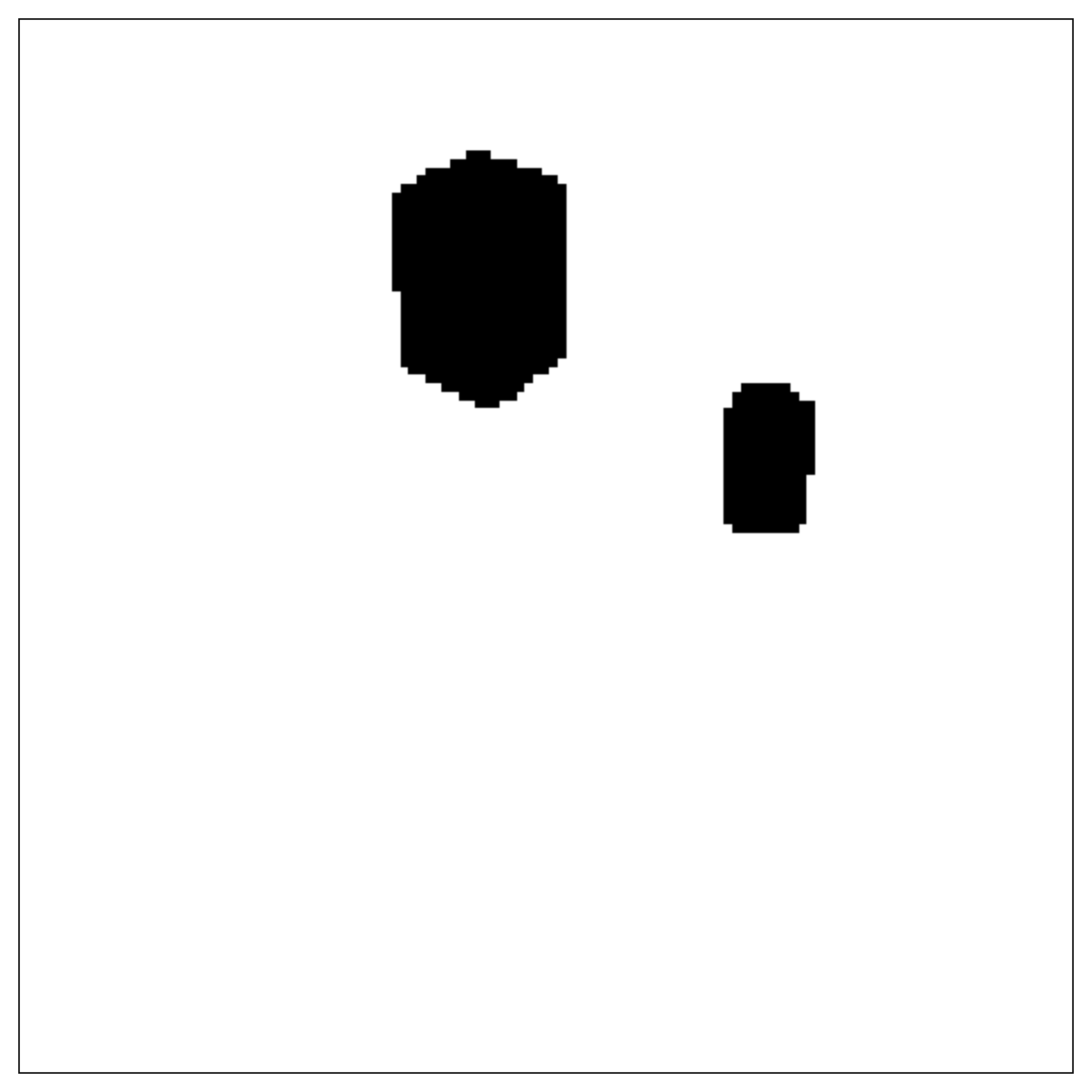} & {\tiny GT Unique First-non-empty} \\
\midrule
LRP \cite{Bach:PLOS2015}                            & \includegraphics[width=.12\linewidth,valign=m]{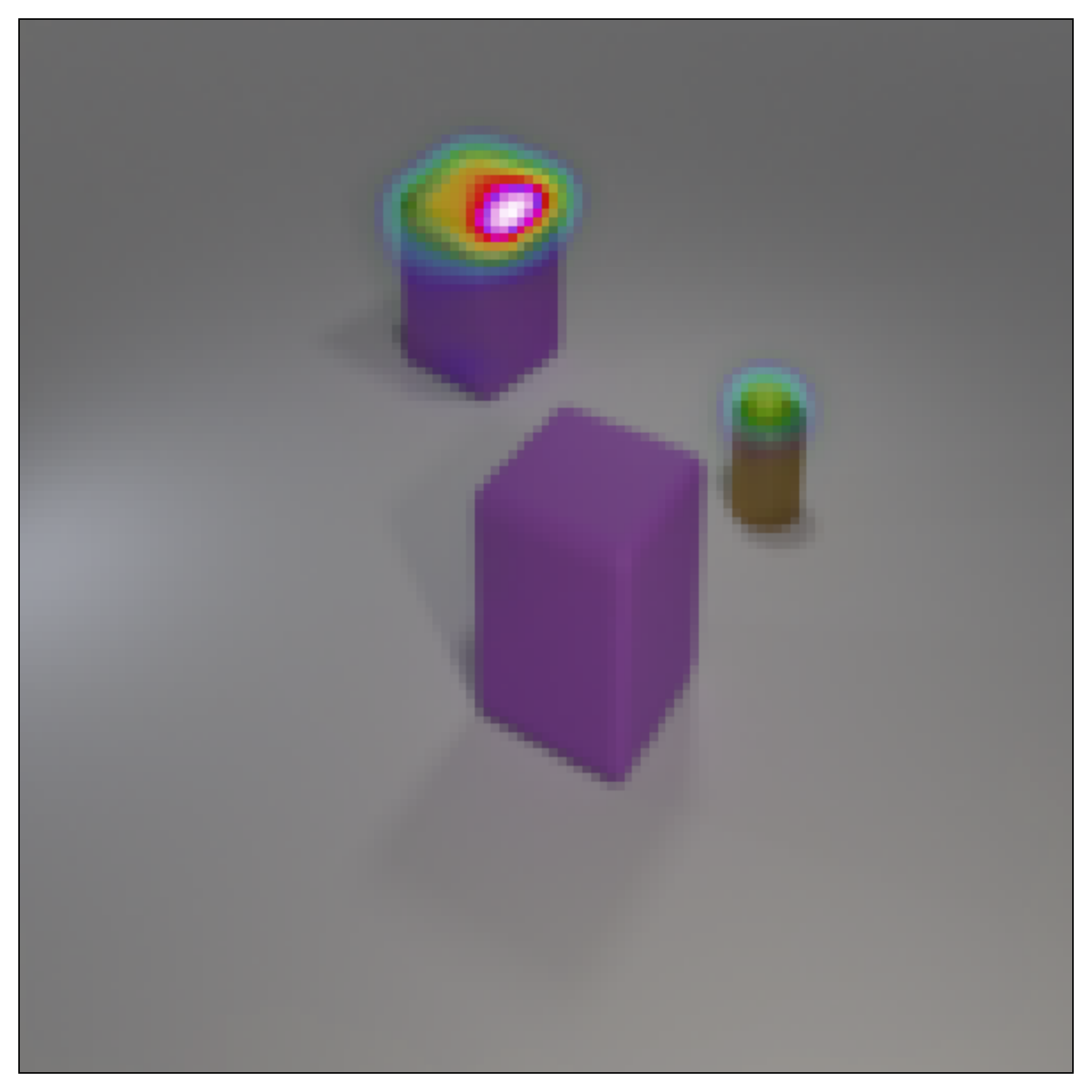} & \includegraphics[width=.12\linewidth,valign=m]{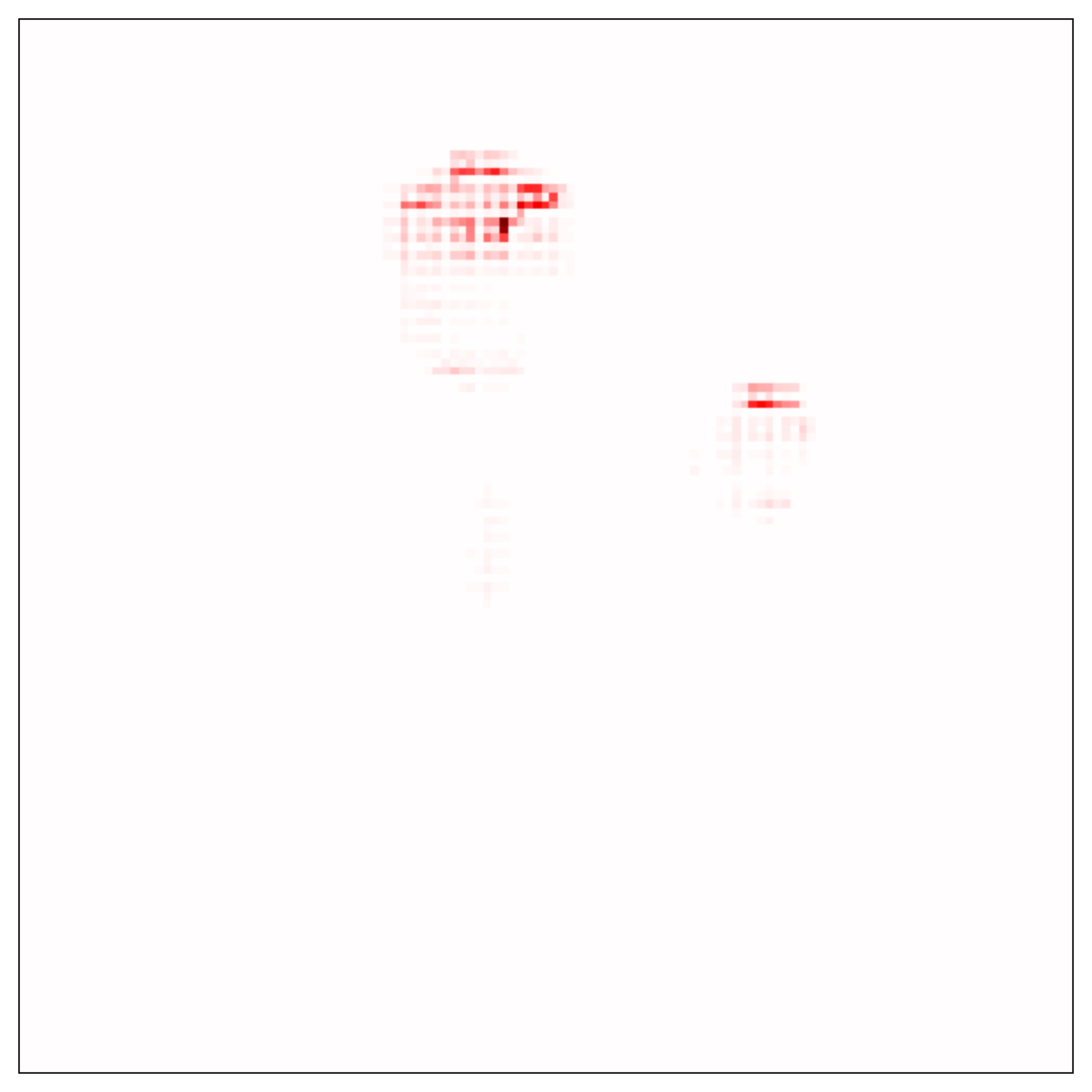} & 0.92 \\
Excitation Backprop \cite{Zhang:ECCV2016}           & \includegraphics[width=.12\linewidth,valign=m]{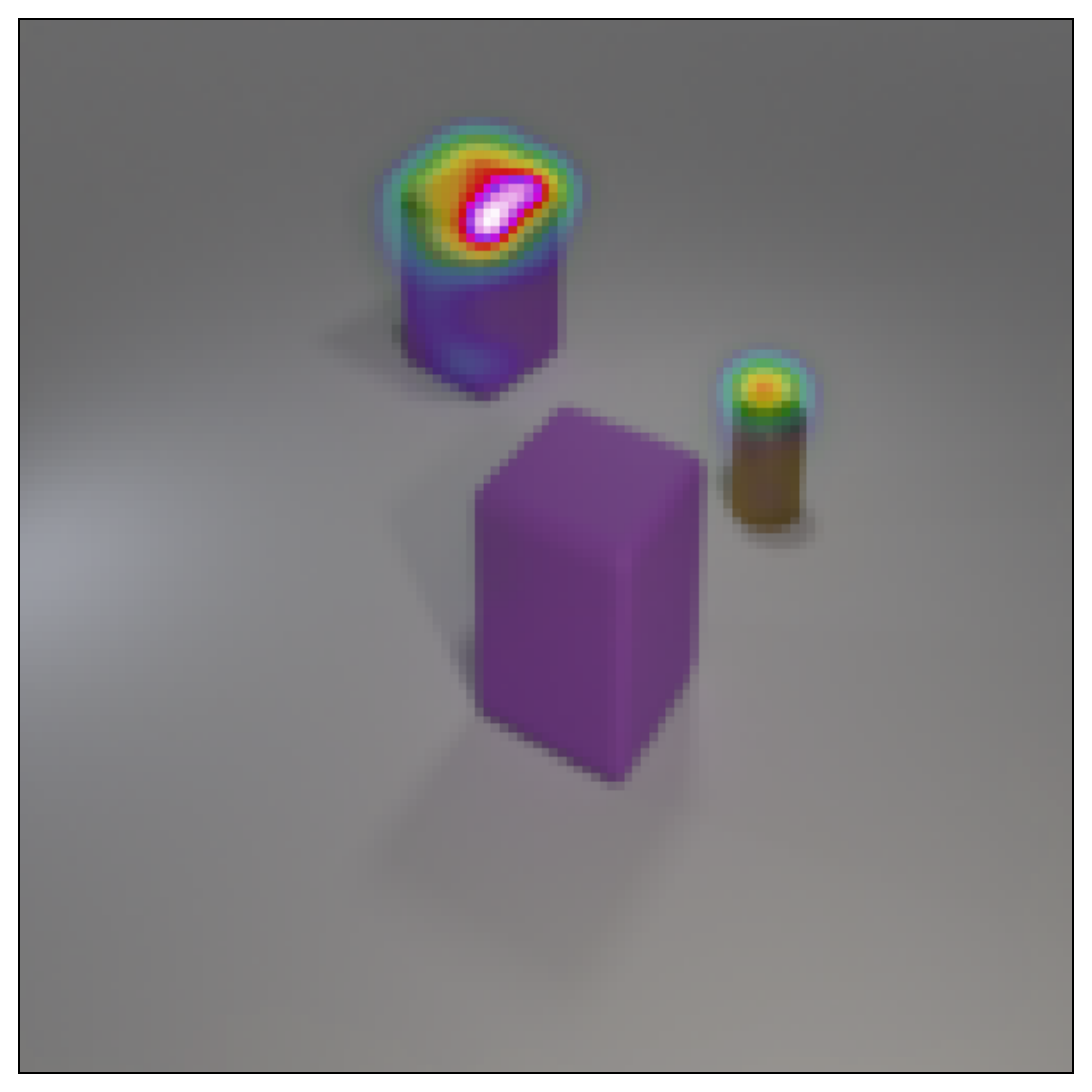} & \includegraphics[width=.12\linewidth,valign=m]{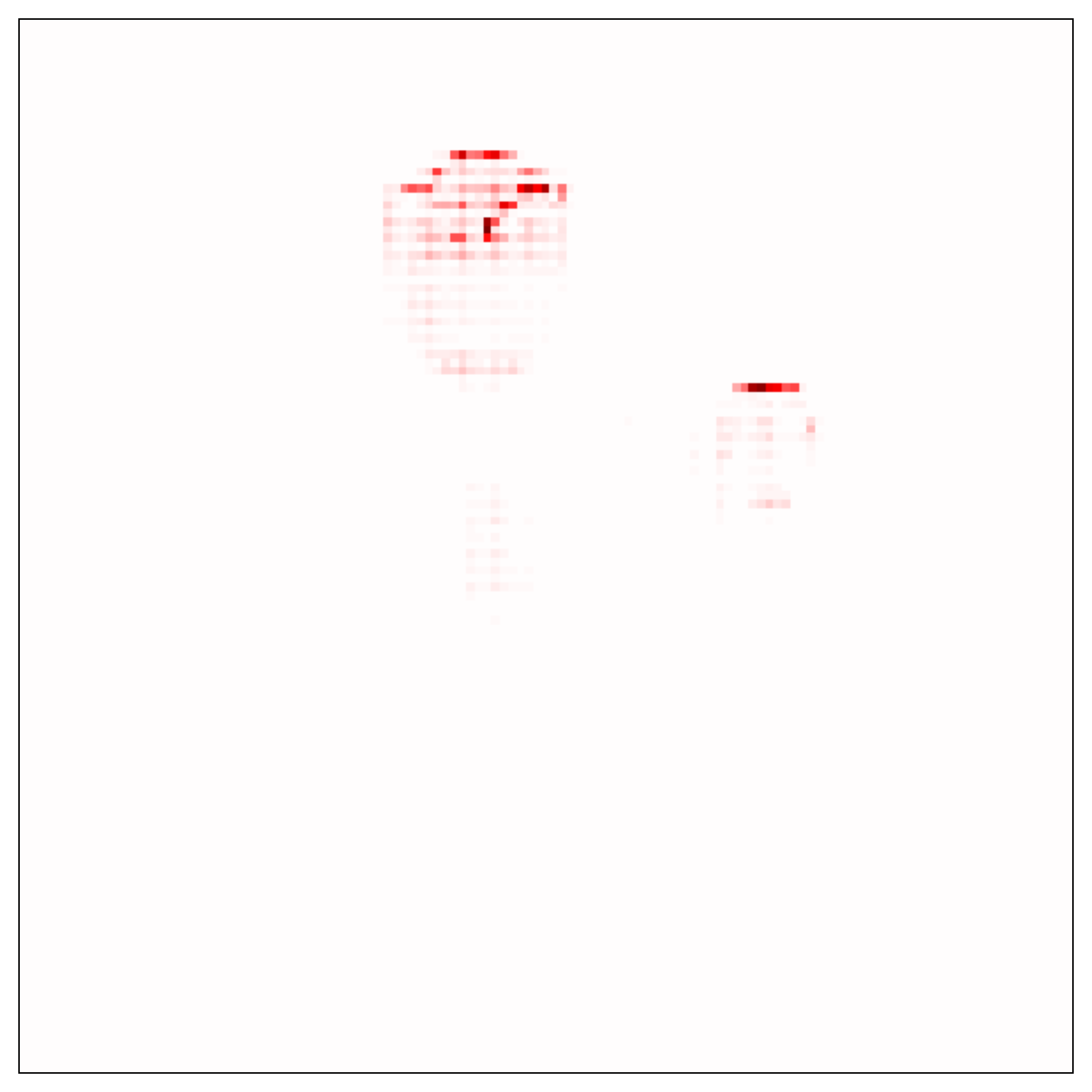} & 0.83 \\
IG \cite{Sundararajan:ICML2017}                     & \includegraphics[width=.12\linewidth,valign=m]{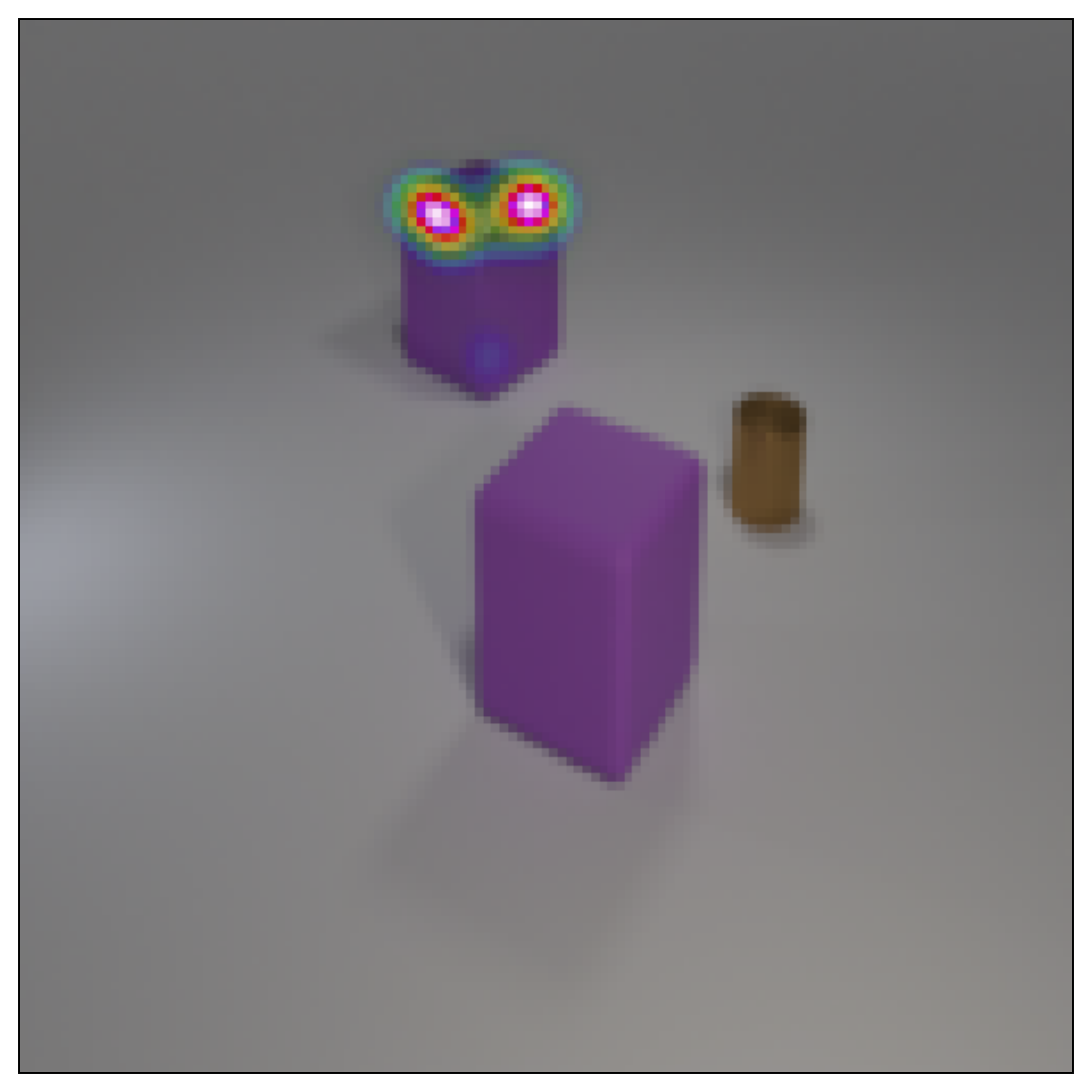} & \includegraphics[width=.12\linewidth,valign=m]{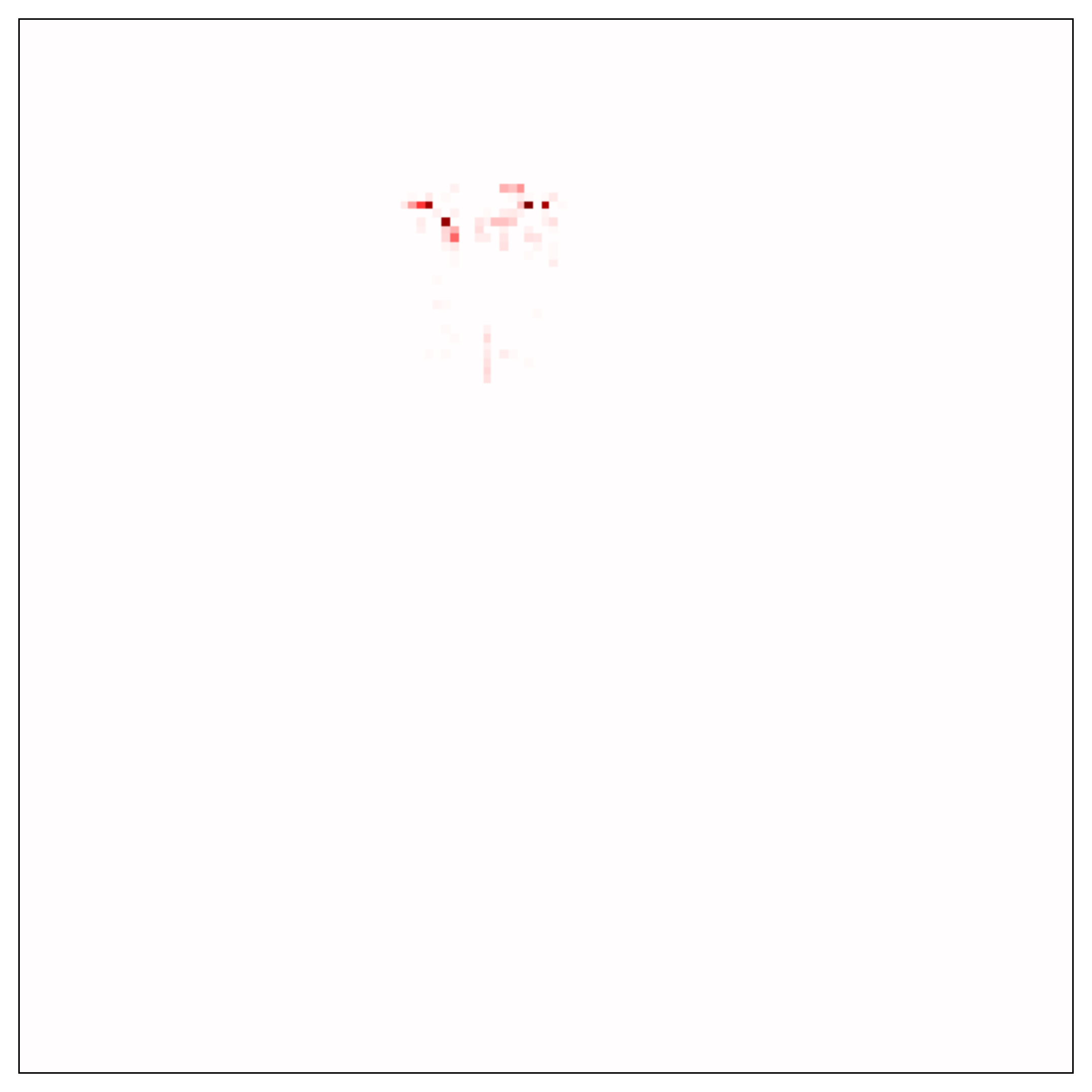} & 0.98 \\
Guided Backprop \cite{Spring:ICLR2015}              & \includegraphics[width=.12\linewidth,valign=m]{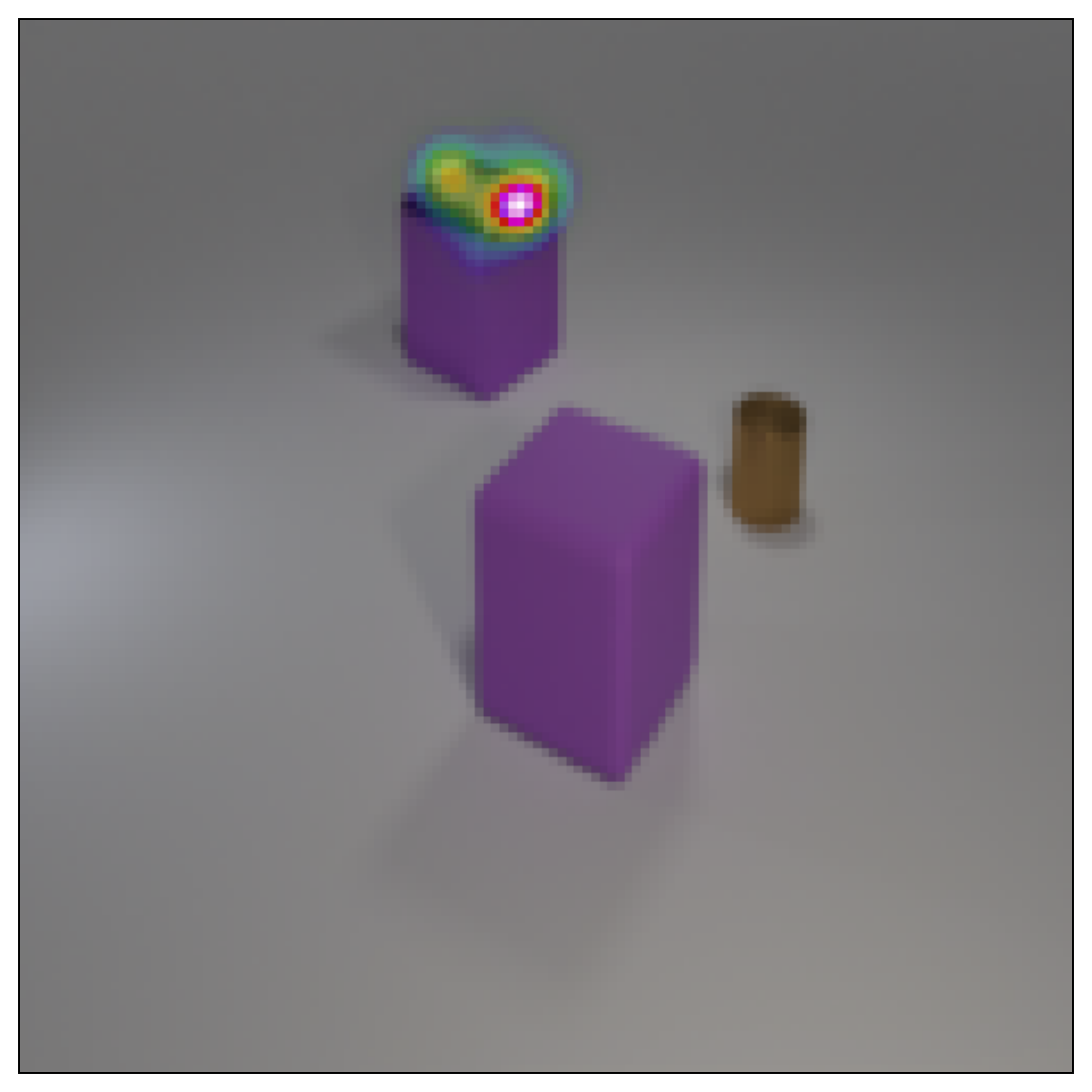} & \includegraphics[width=.12\linewidth,valign=m]{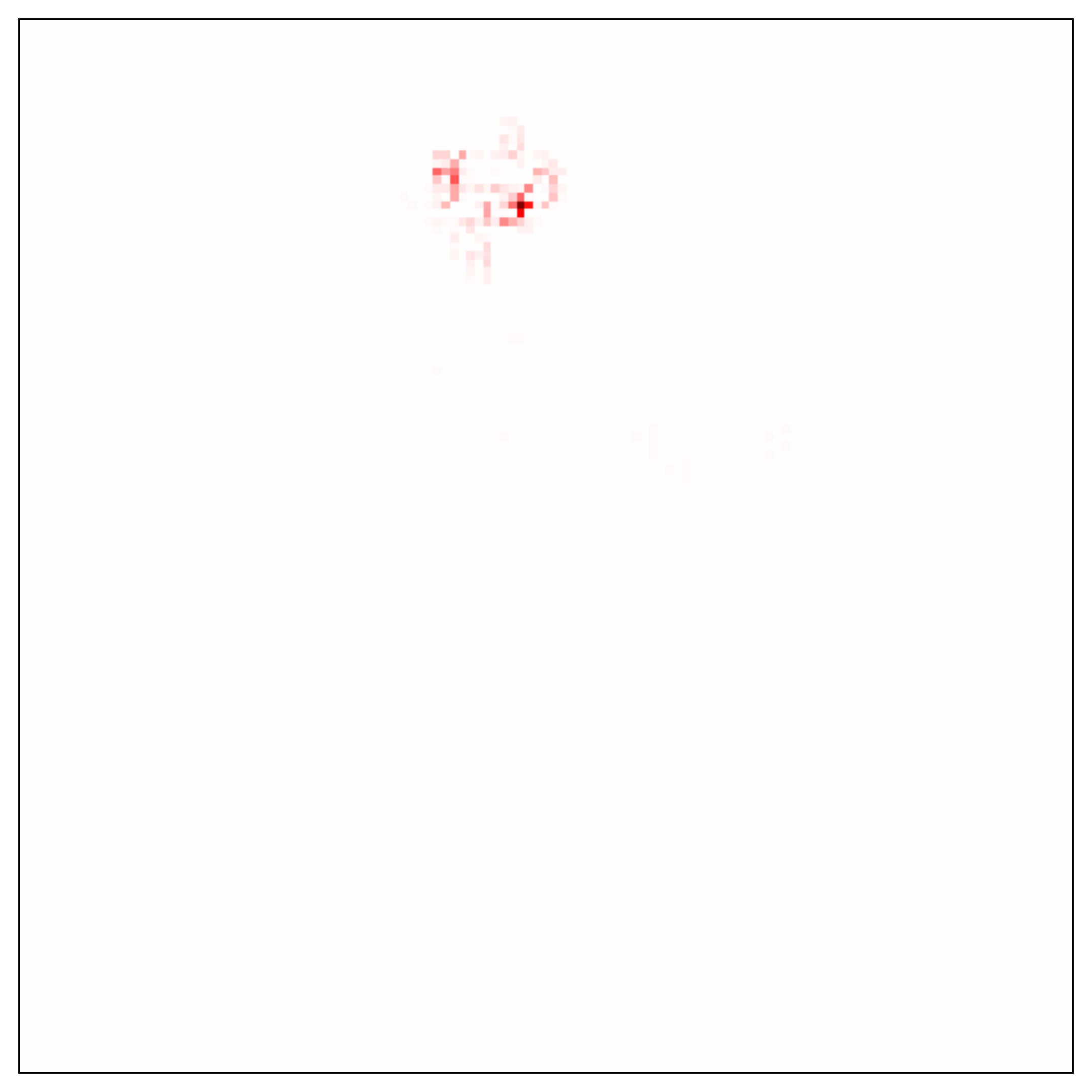} & 0.81 \\
Guided Grad-CAM \cite{Selvaraju:ICCV2017}           & \includegraphics[width=.12\linewidth,valign=m]{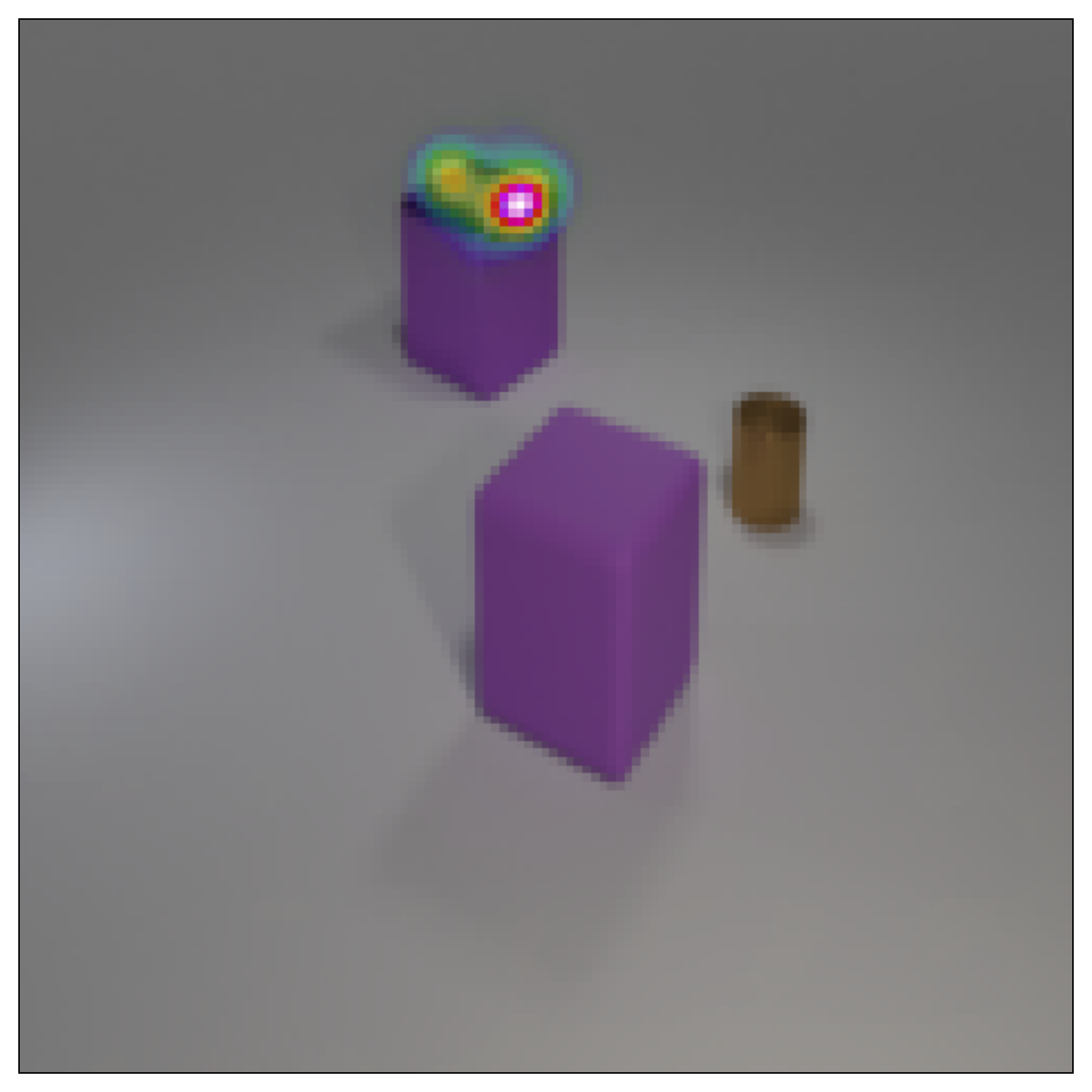} & \includegraphics[width=.12\linewidth,valign=m]{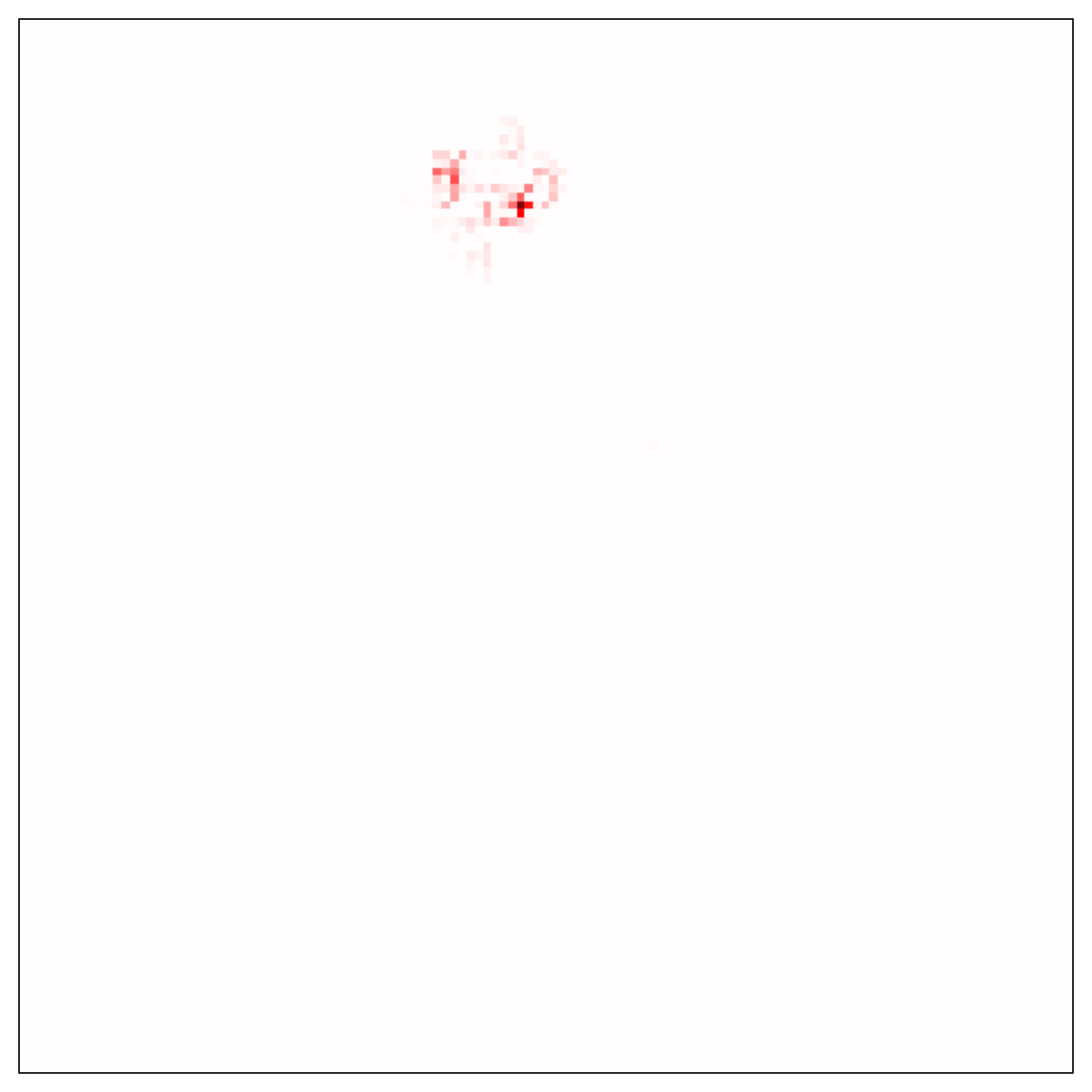} & 0.83 \\
SmoothGrad \cite{Smilkov:ICML2017}                  & \includegraphics[width=.12\linewidth,valign=m]{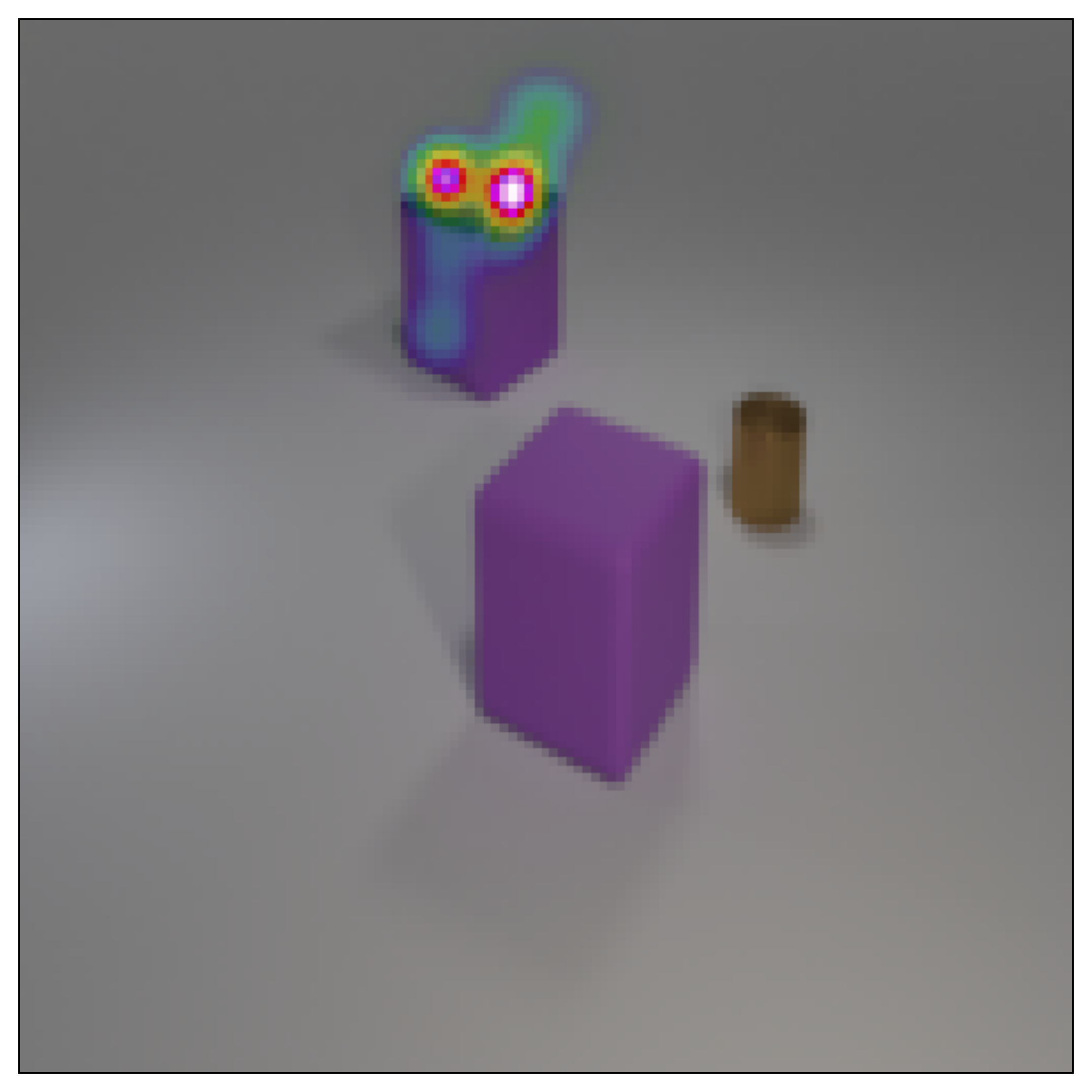} & \includegraphics[width=.12\linewidth,valign=m]{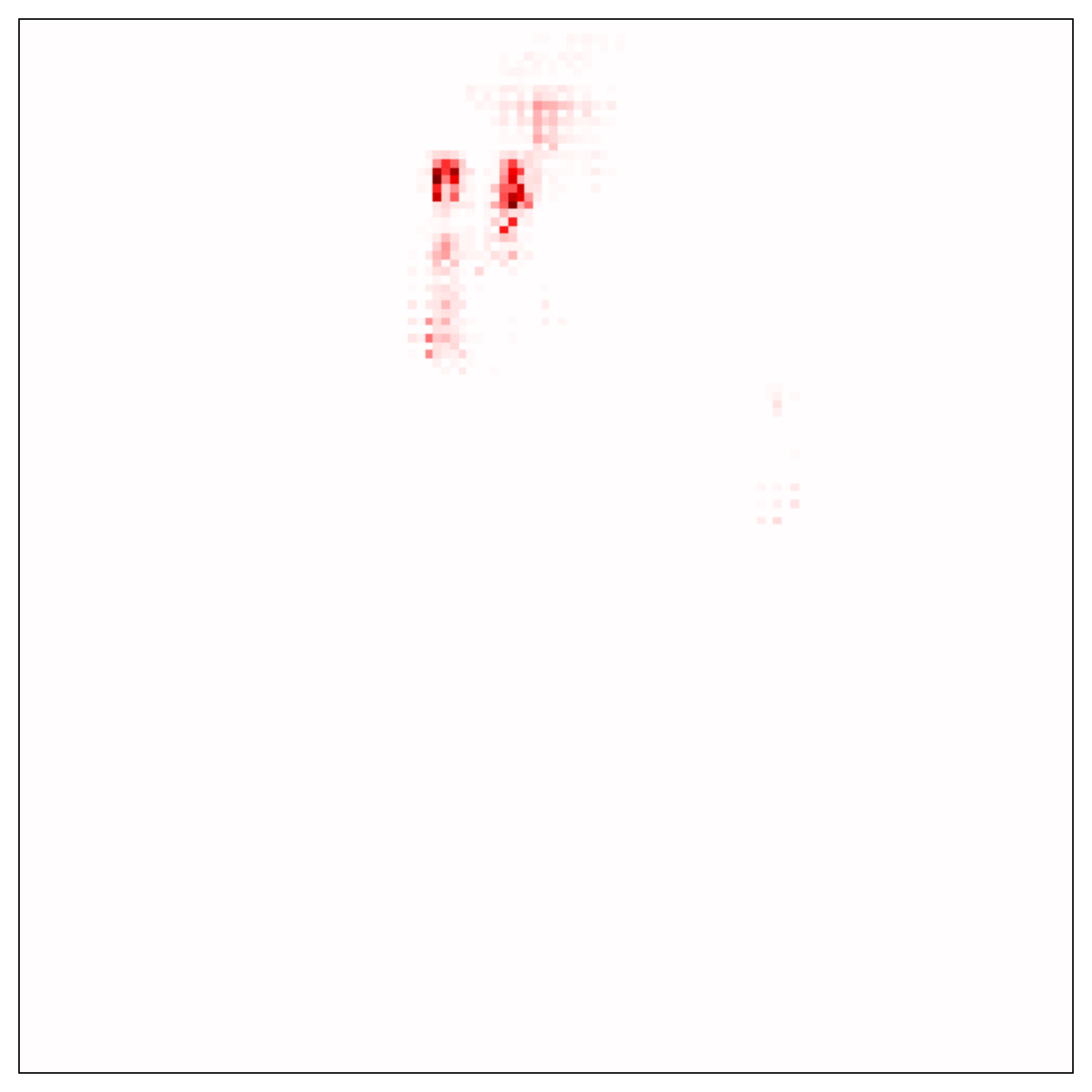} & 0.76 \\
VarGrad \cite{Adebayo:ICLR2018}                     & \includegraphics[width=.12\linewidth,valign=m]{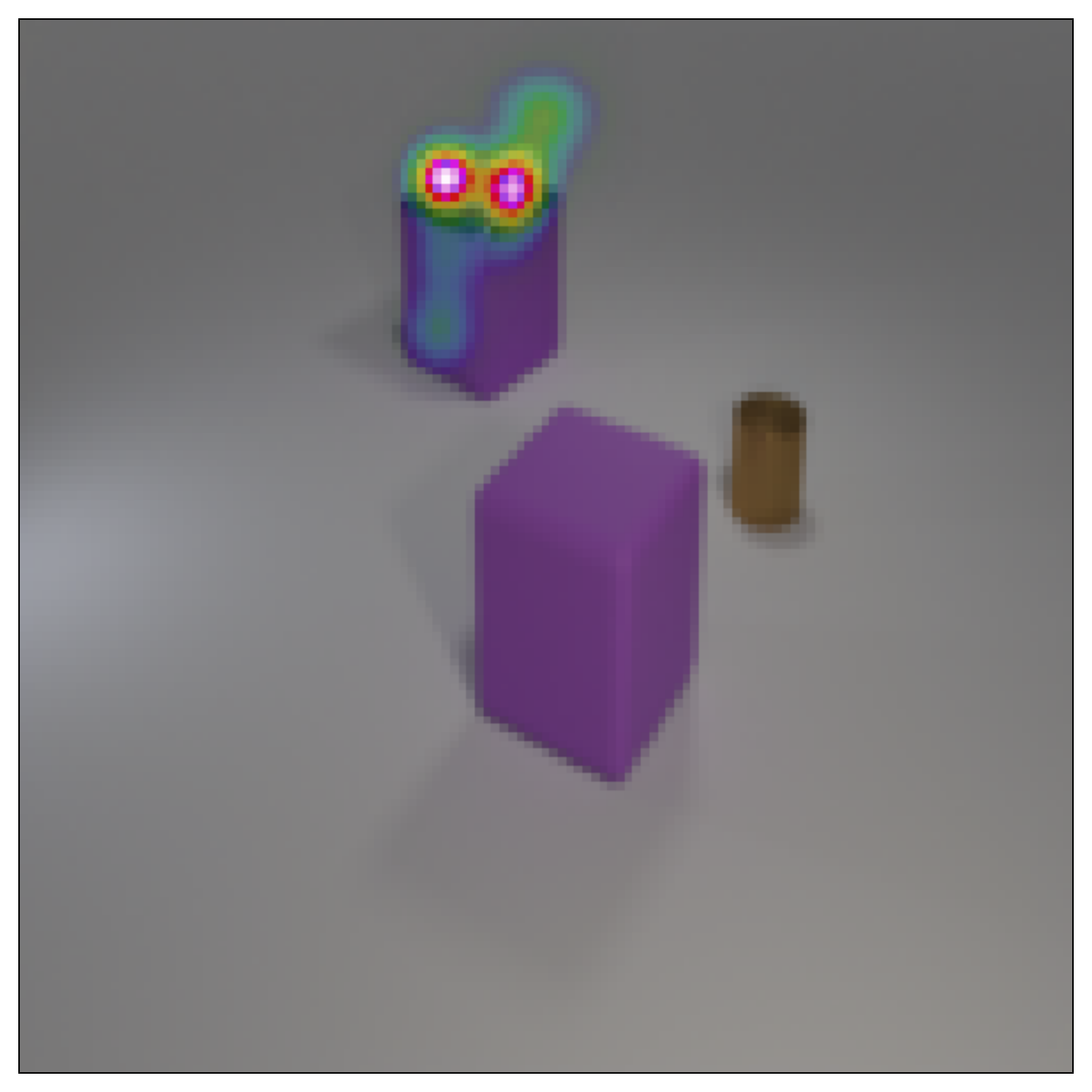} & \includegraphics[width=.12\linewidth,valign=m]{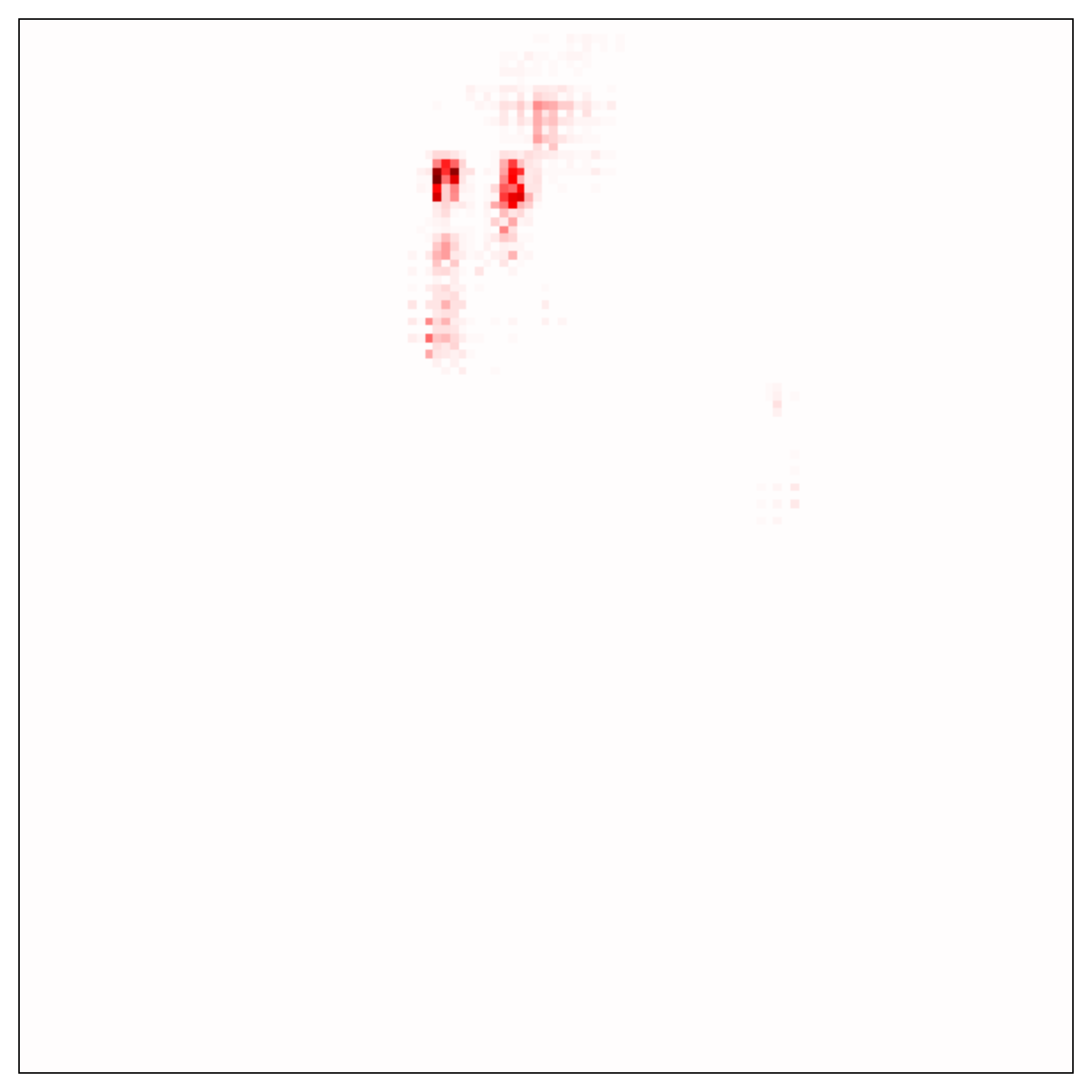} & 0.74 \\
Gradient \cite{Simonyan:ICLR2014}                   & \includegraphics[width=.12\linewidth,valign=m]{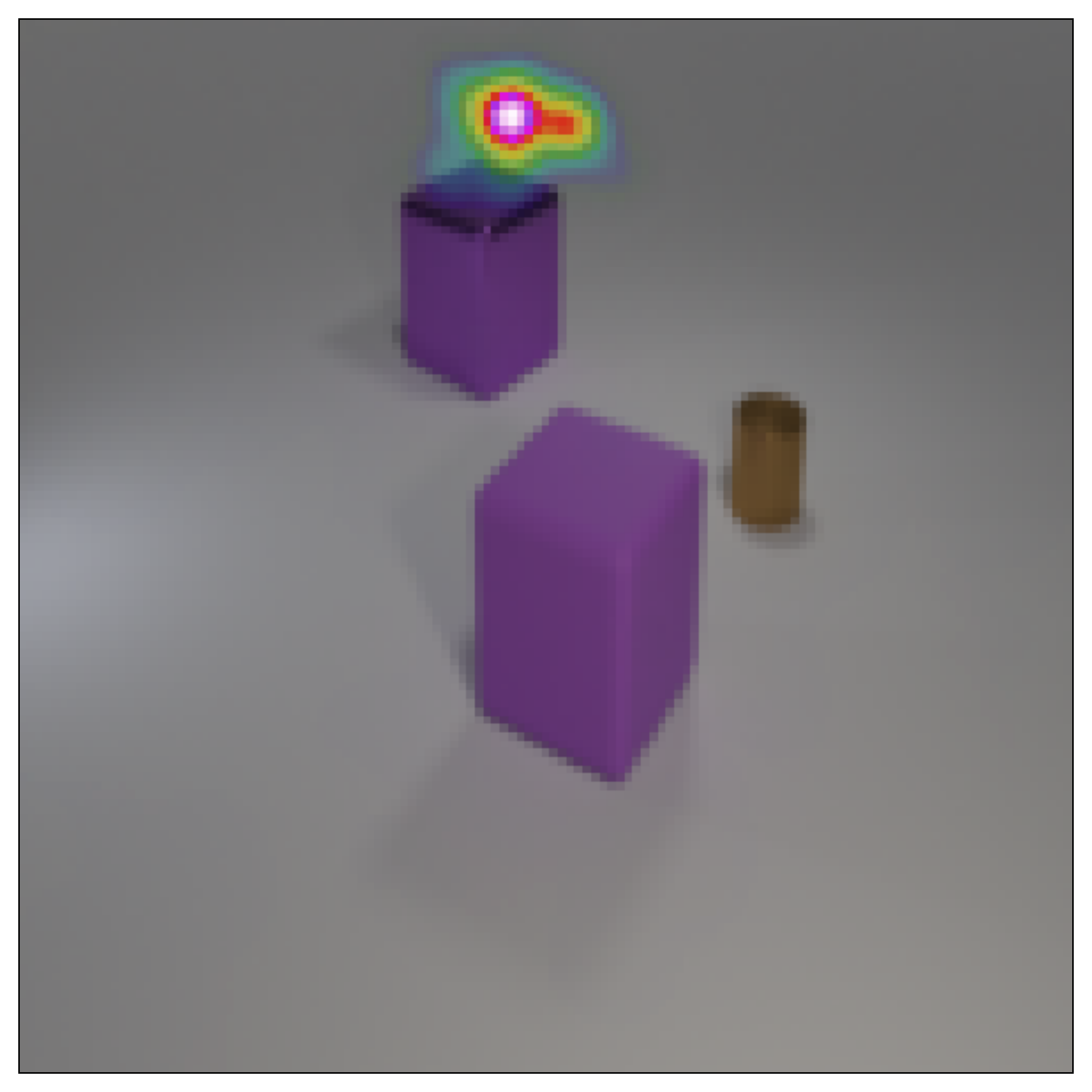} & \includegraphics[width=.12\linewidth,valign=m]{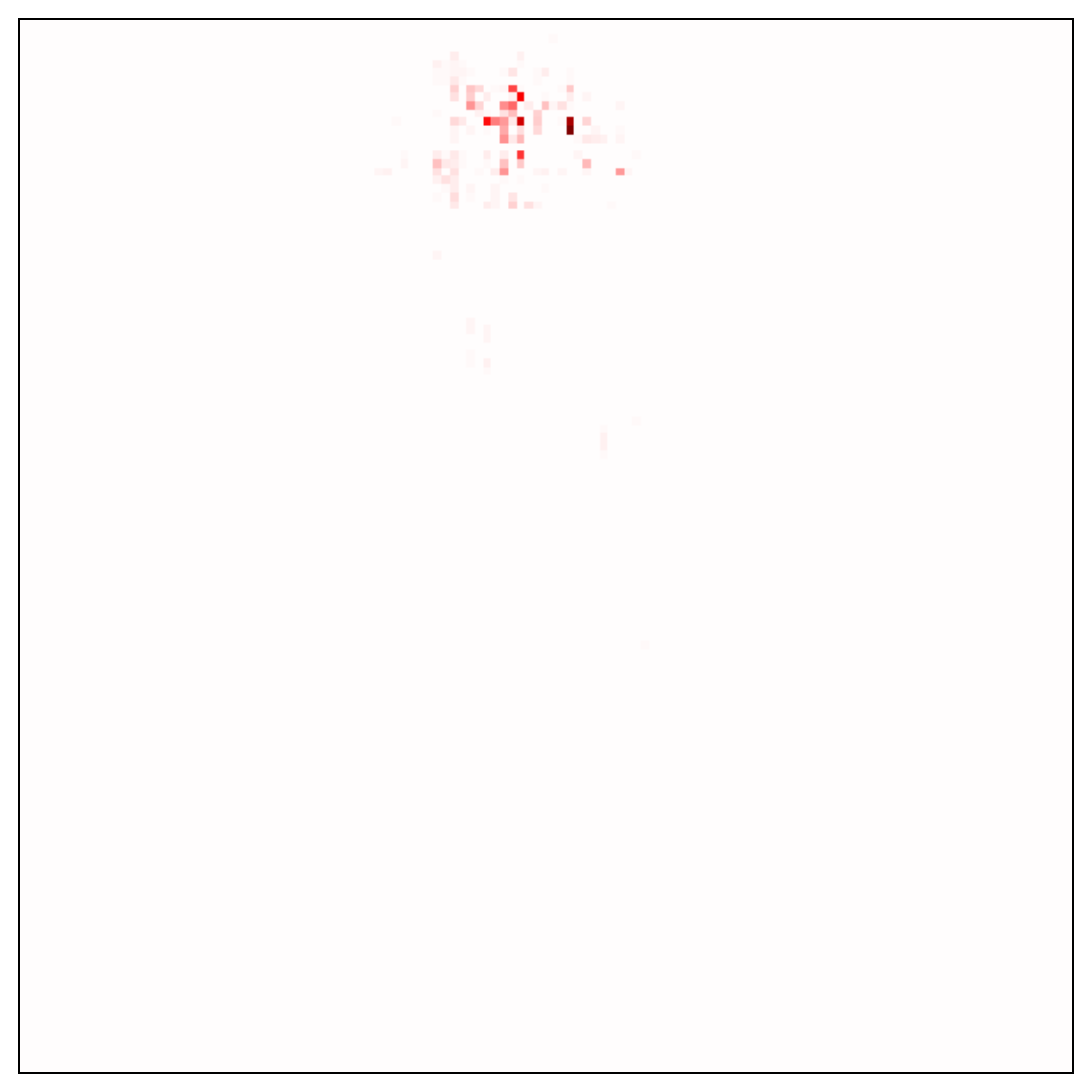} & 0.14 \\
Gradient$\times$Input \cite{Shrikumar:arxiv2016}    & \includegraphics[width=.12\linewidth,valign=m]{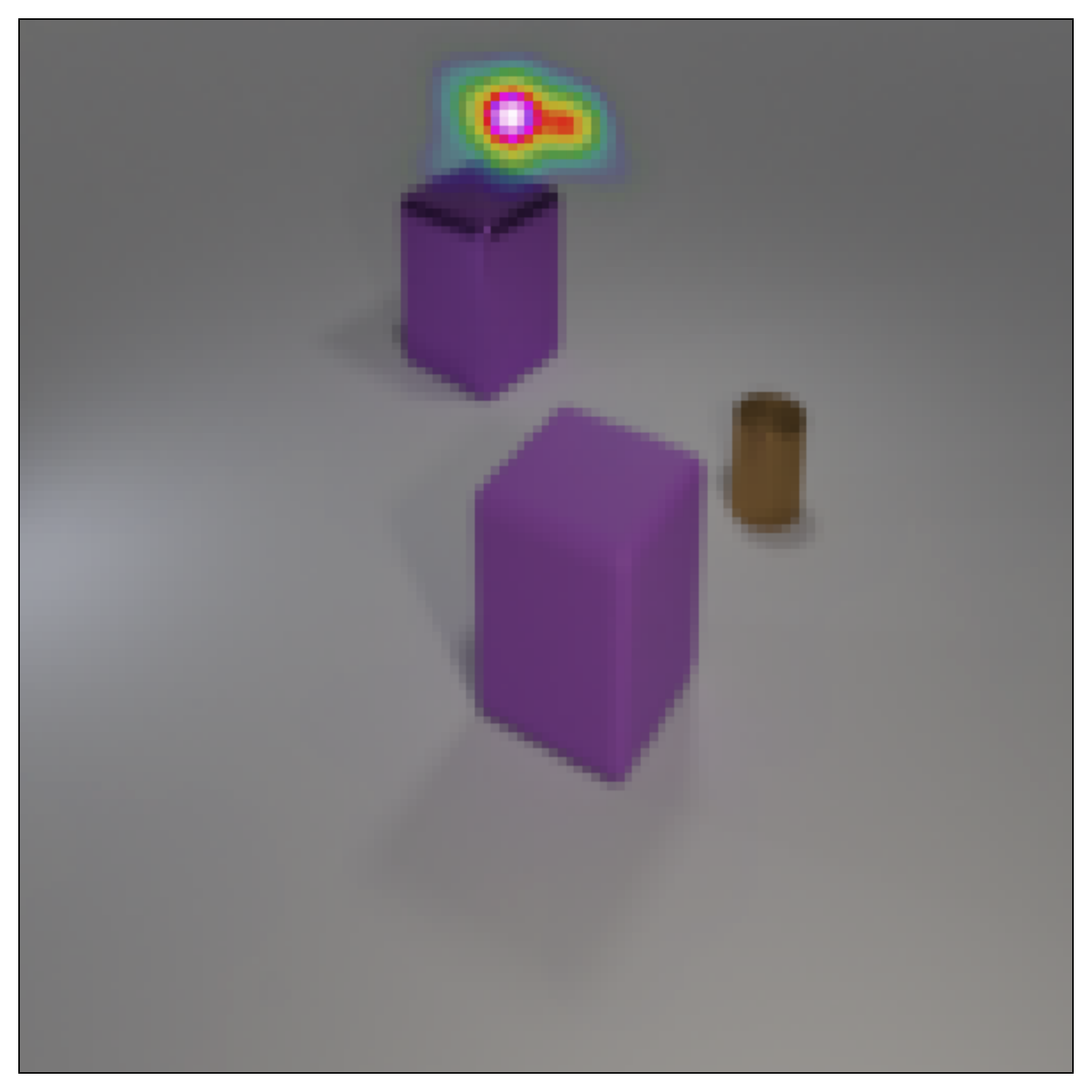} & \includegraphics[width=.12\linewidth,valign=m]{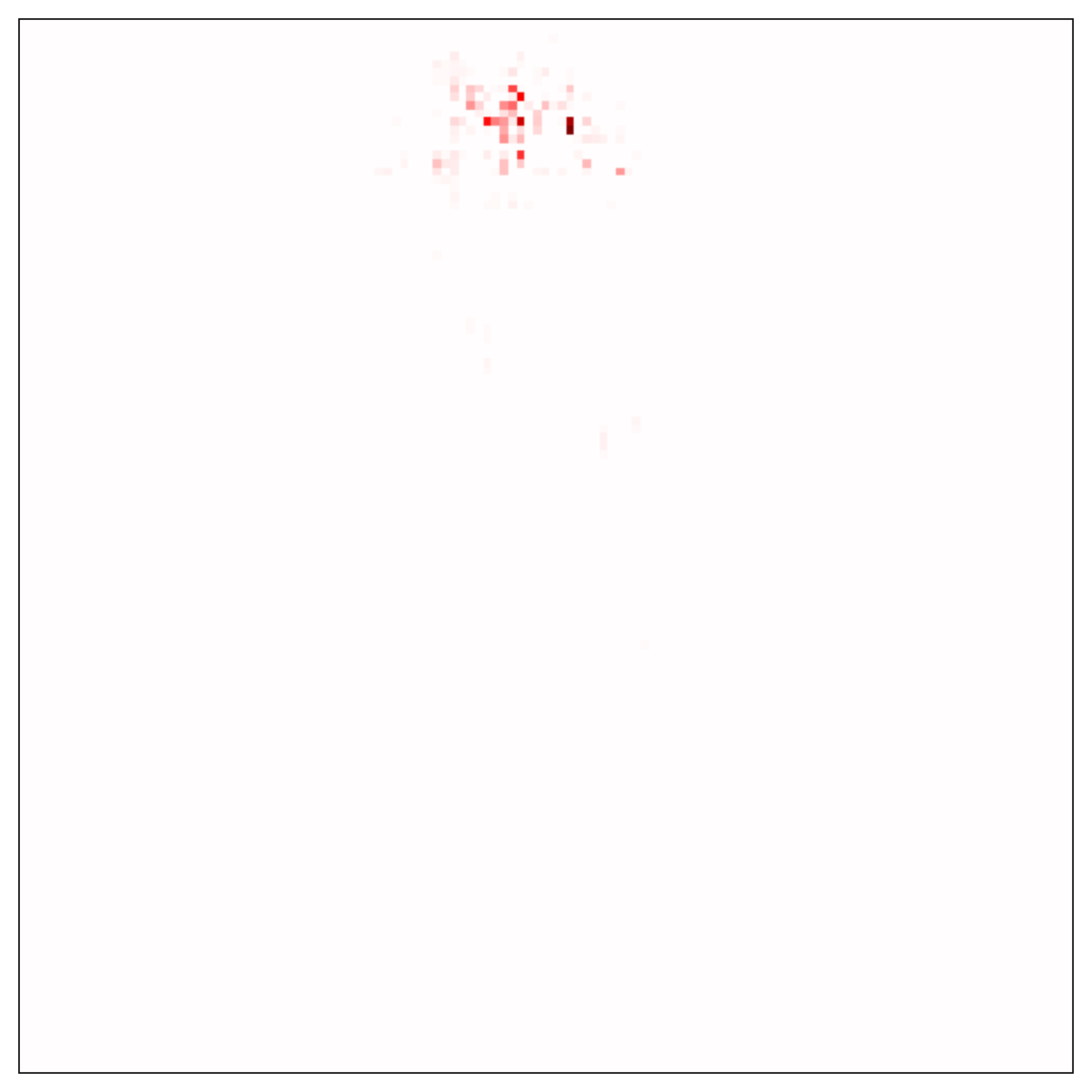} & 0.08 \\
Deconvnet \cite{Zeiler:ECCV2014}                    & \includegraphics[width=.12\linewidth,valign=m]{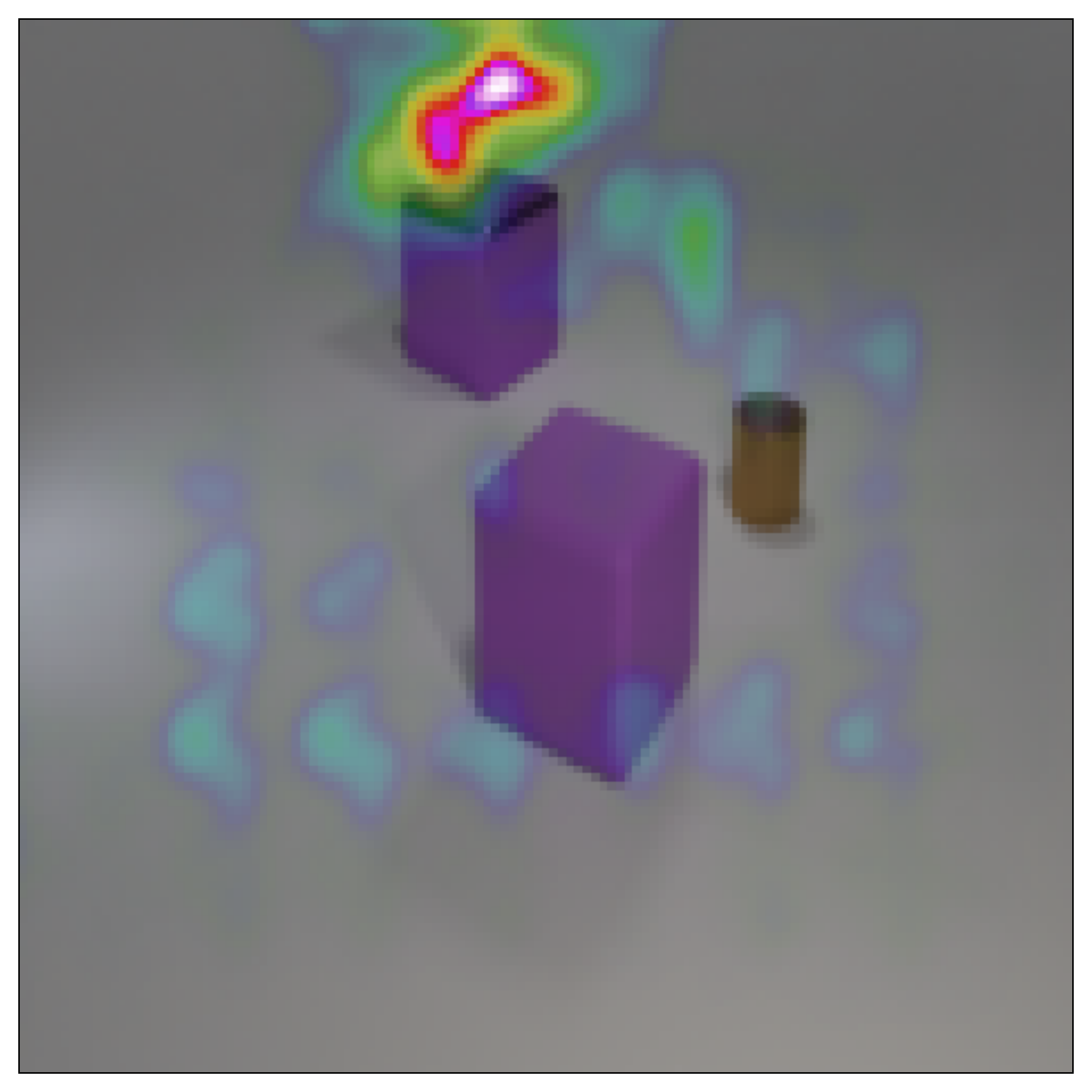} & \includegraphics[width=.12\linewidth,valign=m]{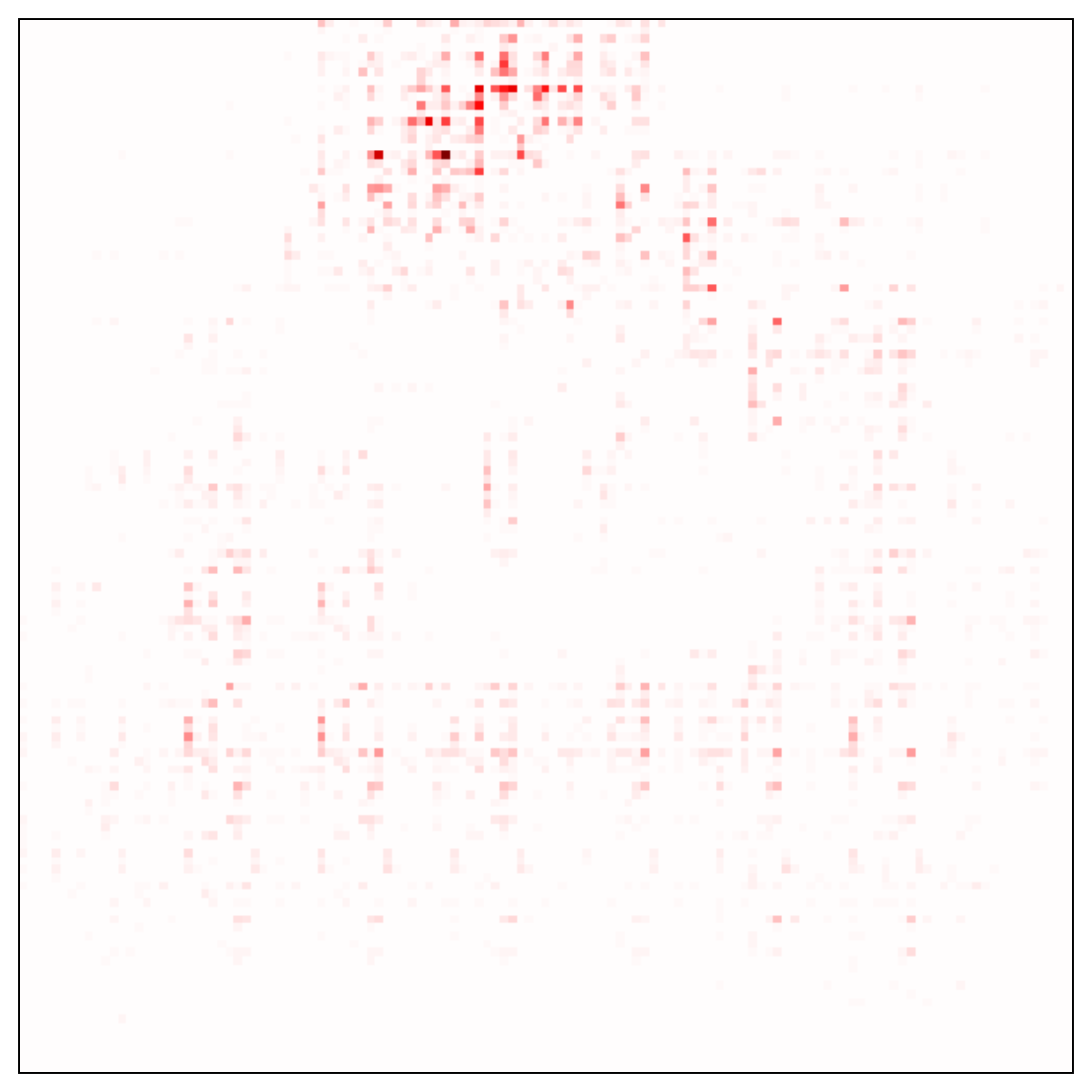} & 0.07 \\
Grad-CAM \cite{Selvaraju:ICCV2017}                  & \includegraphics[width=.12\linewidth,valign=m]{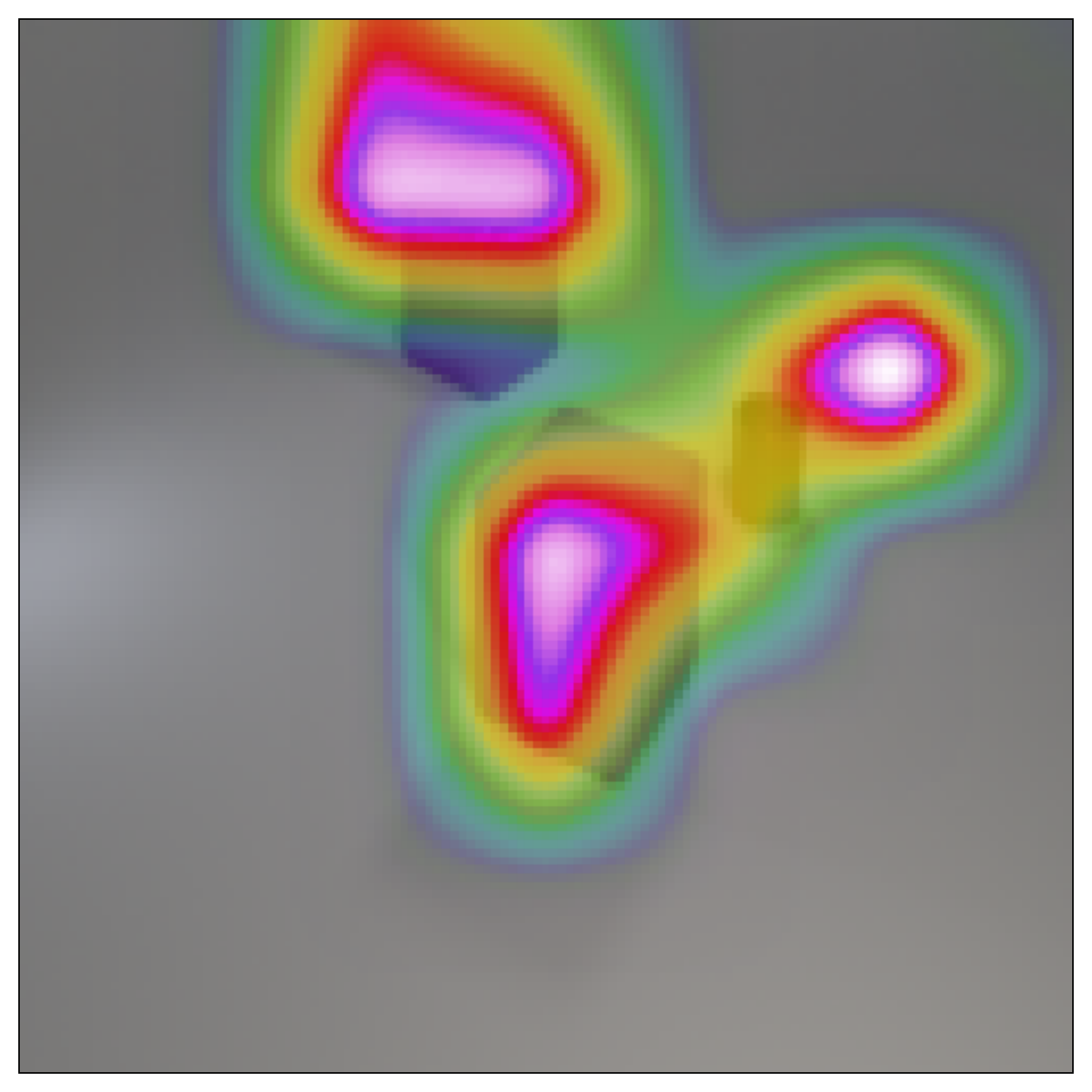} & \includegraphics[width=.12\linewidth,valign=m]{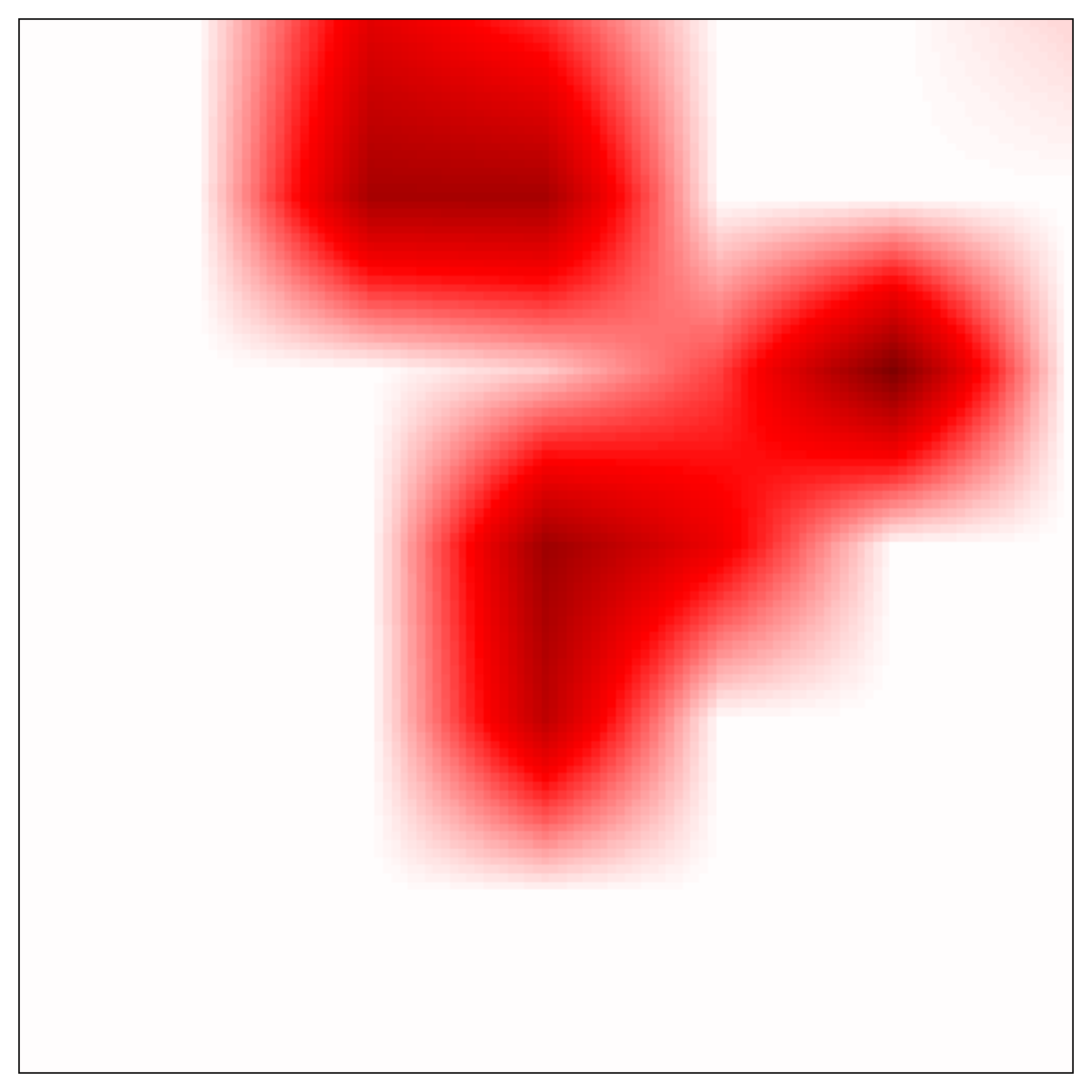} & 0.13 \\
\end{tabular}
\end{table}

\begin{table}
        \scriptsize
		\caption{Heatmaps for a correctly predicted CLEVR-XAI-complex question (raw heatmap and heatmap overlayed with original image), and relevance \textit{mass} accuracy.}
		\label{table:heatmap-complex-correct-49997}
\begin{tabular}{lllc}
\midrule
\begin{tabular}{@{}l@{}}\scriptsize There is a big ball on the right side of\\ the blue metal object; what is it made of? \\ \textit{rubber} \end{tabular}  & \includegraphics[width=.18\linewidth,valign=m]{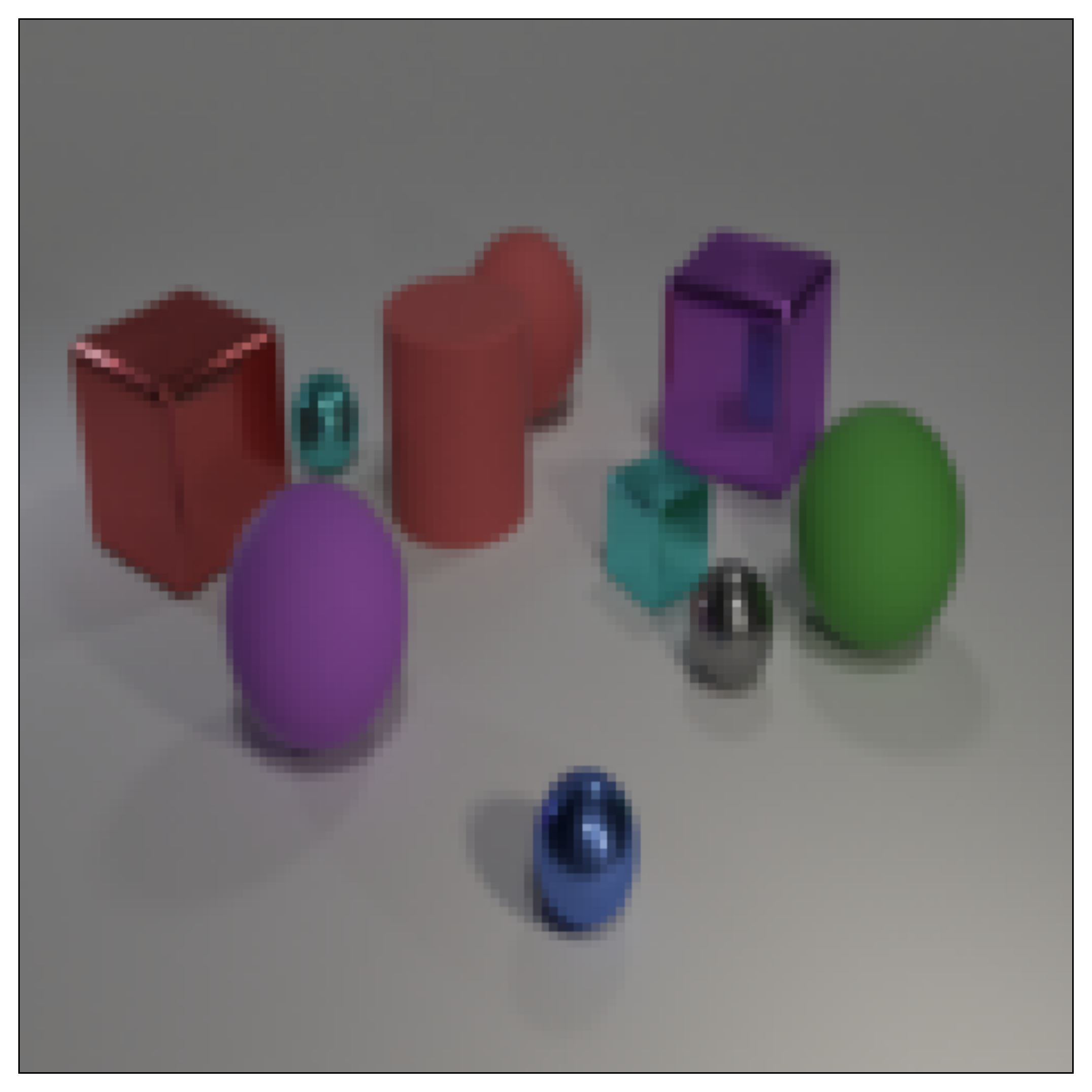} &
\includegraphics[width=.18\linewidth,valign=m]{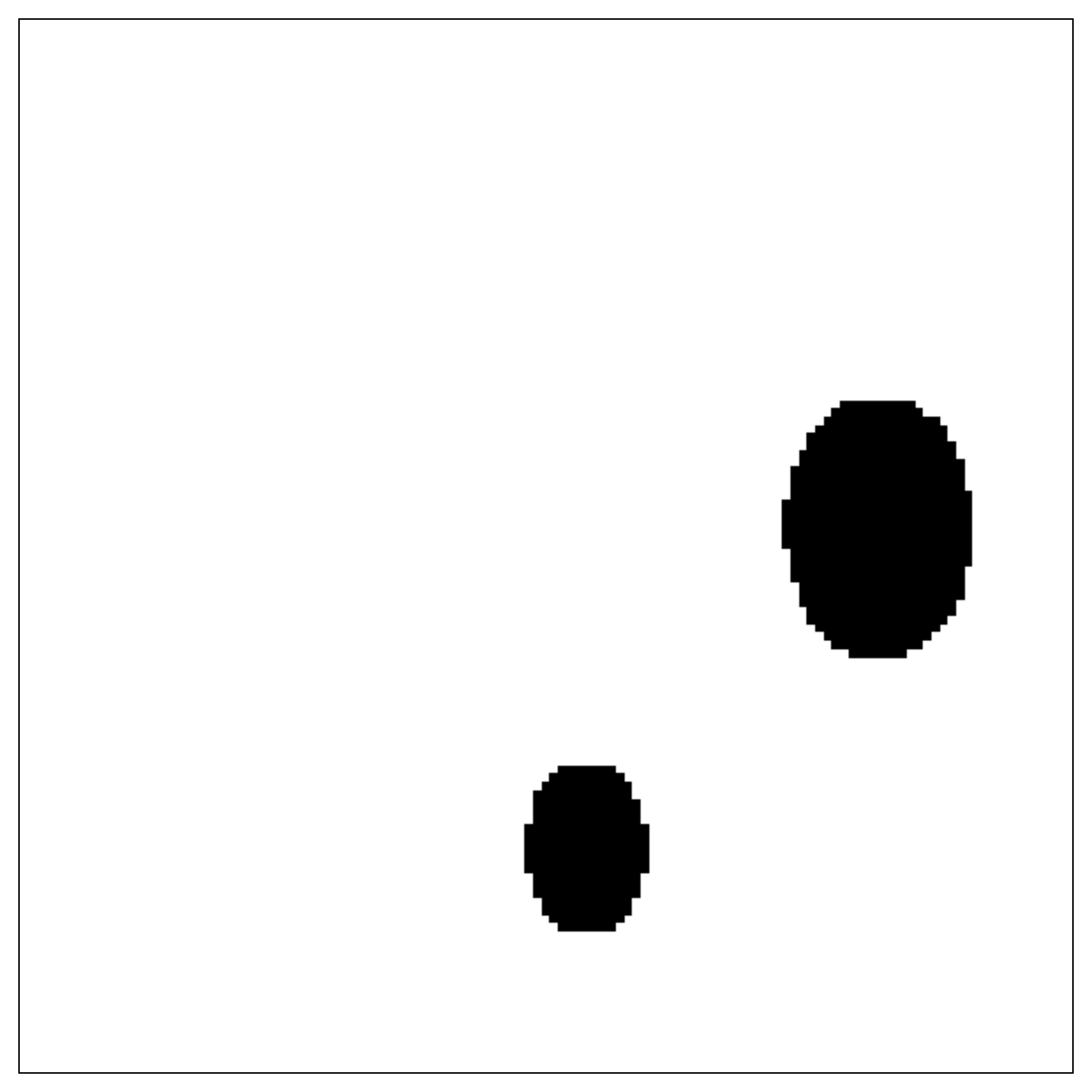} & {\tiny GT Unique First-non-empty} \\
\midrule
LRP \cite{Bach:PLOS2015}                            & \includegraphics[width=.12\linewidth,valign=m]{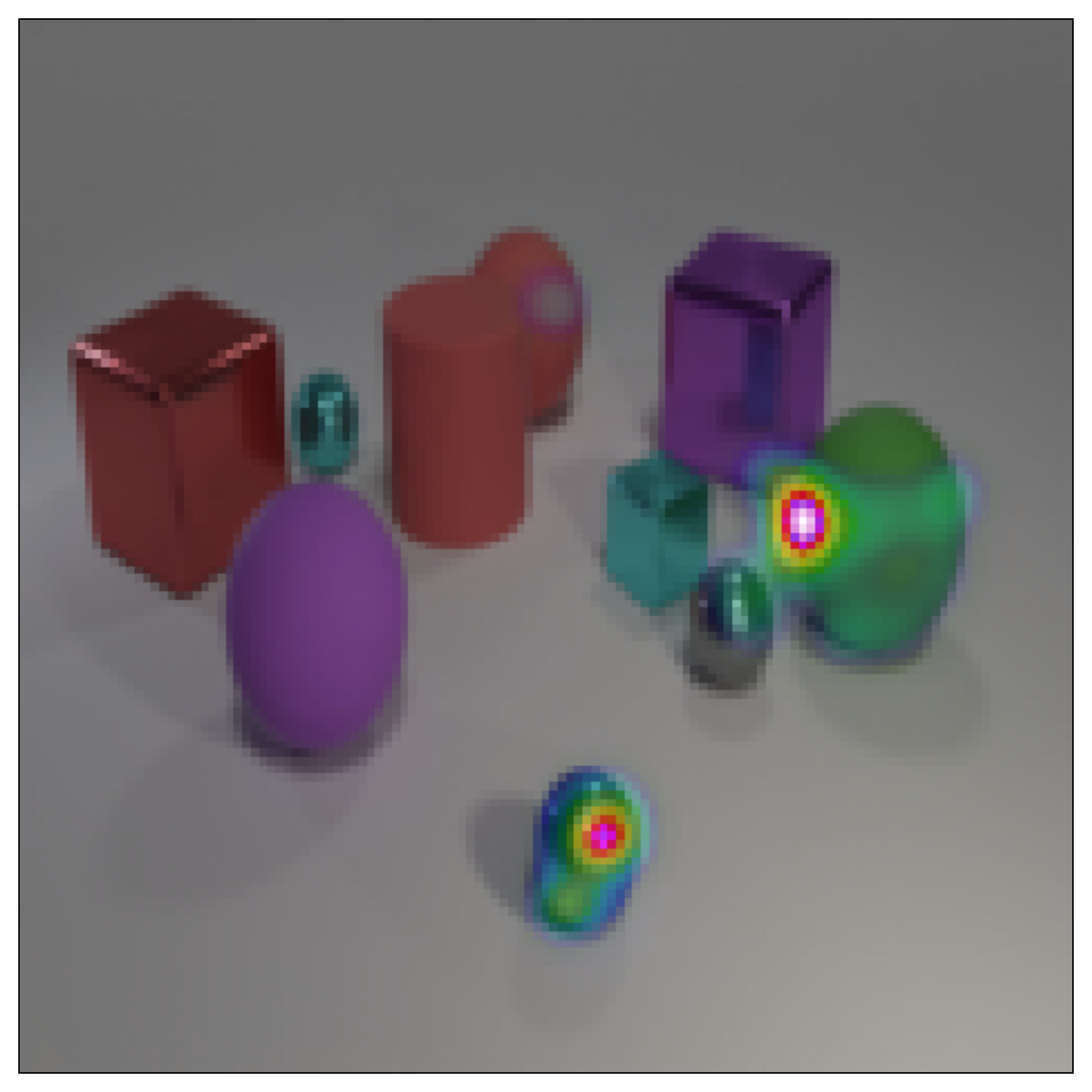} & \includegraphics[width=.12\linewidth,valign=m]{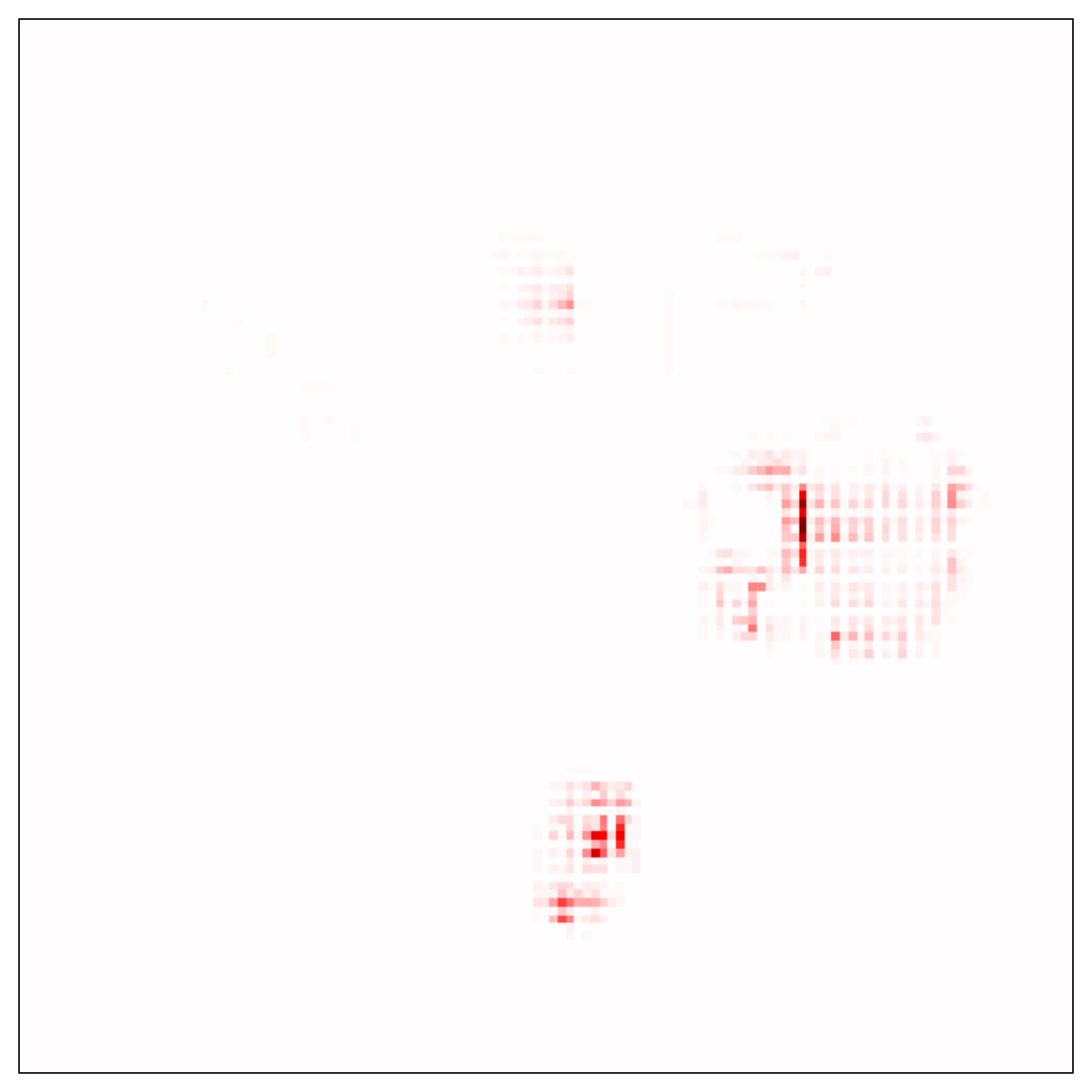} & 0.72 \\
Excitation Backprop \cite{Zhang:ECCV2016}           & \includegraphics[width=.12\linewidth,valign=m]{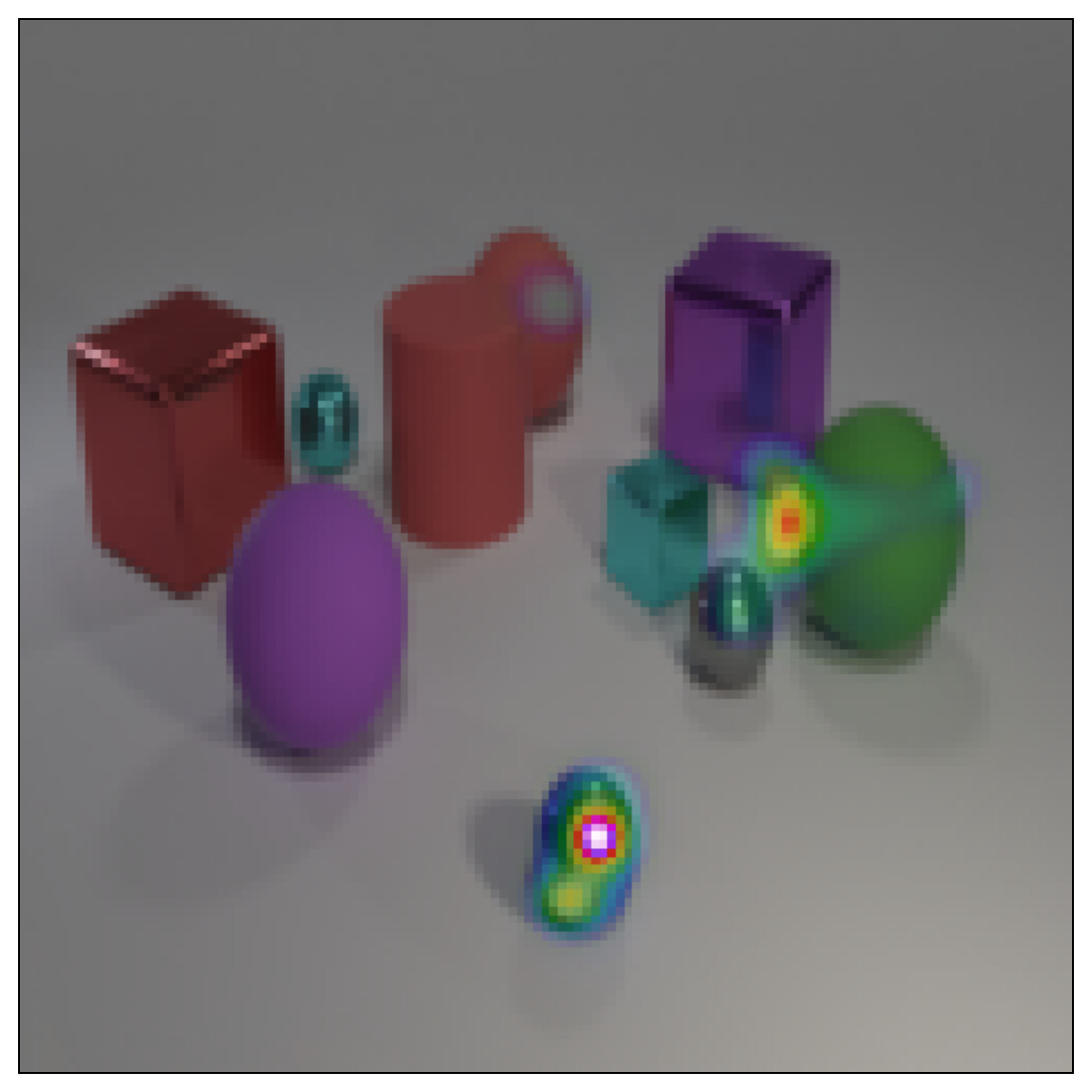} & \includegraphics[width=.12\linewidth,valign=m]{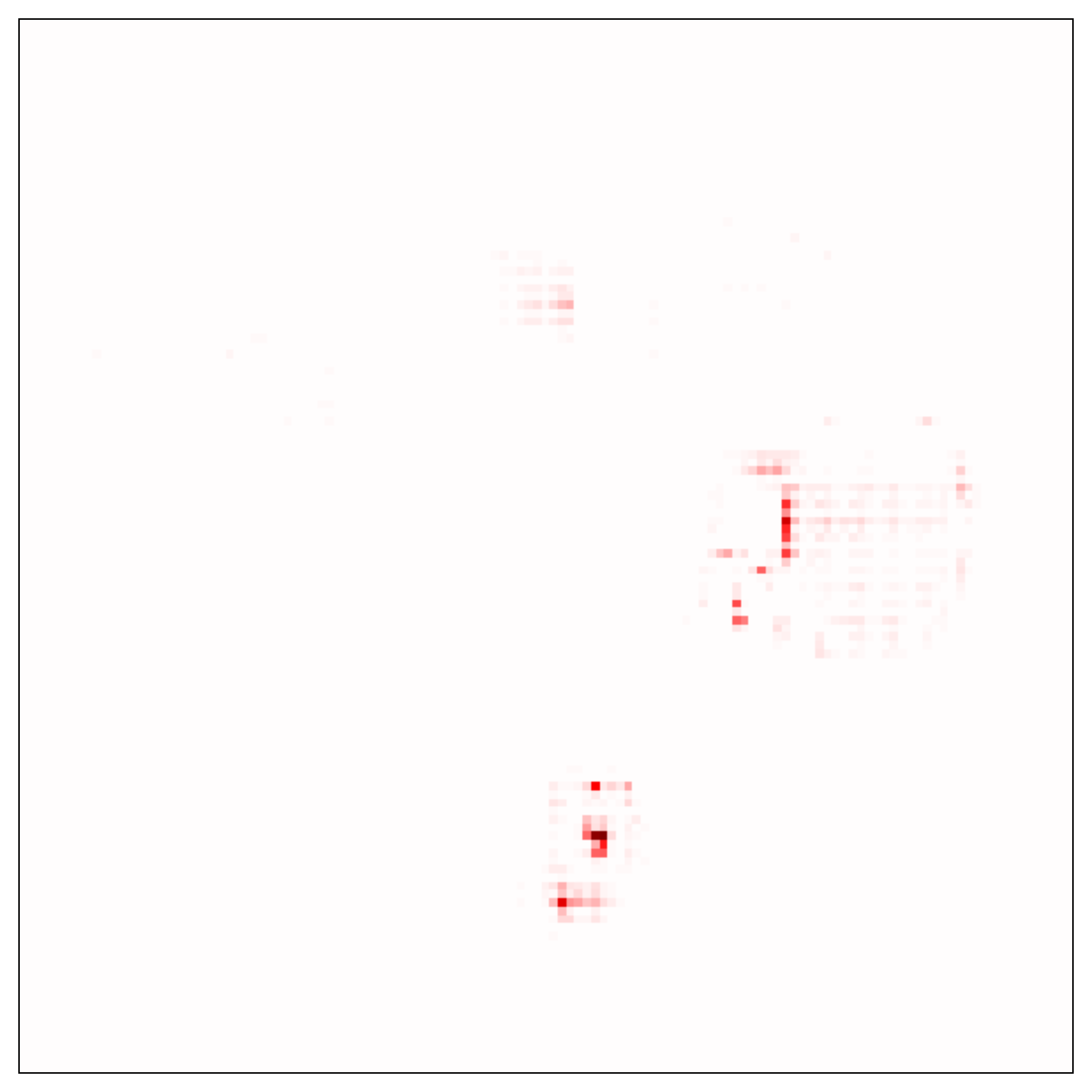} & 0.66 \\
IG \cite{Sundararajan:ICML2017}                     & \includegraphics[width=.12\linewidth,valign=m]{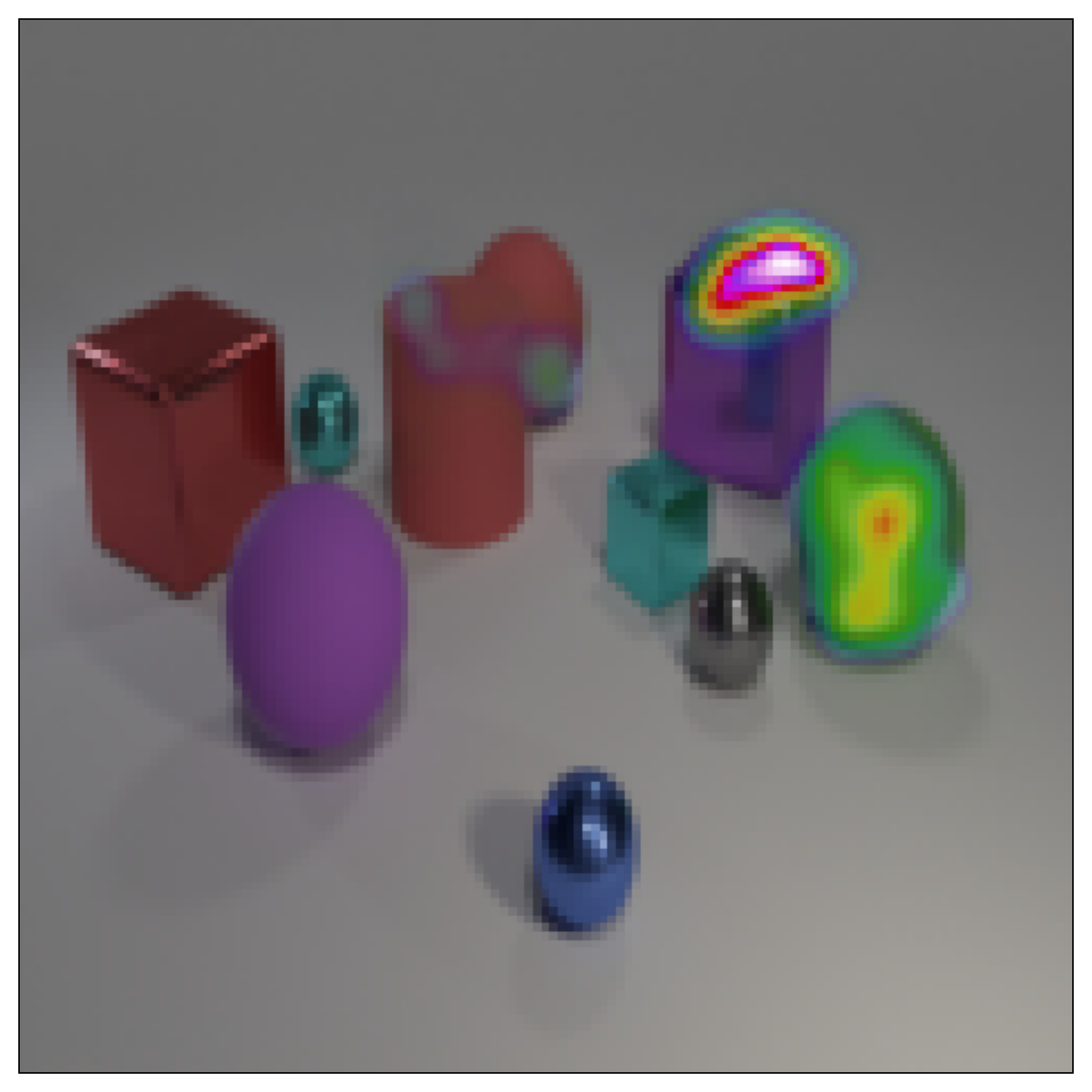} & \includegraphics[width=.12\linewidth,valign=m]{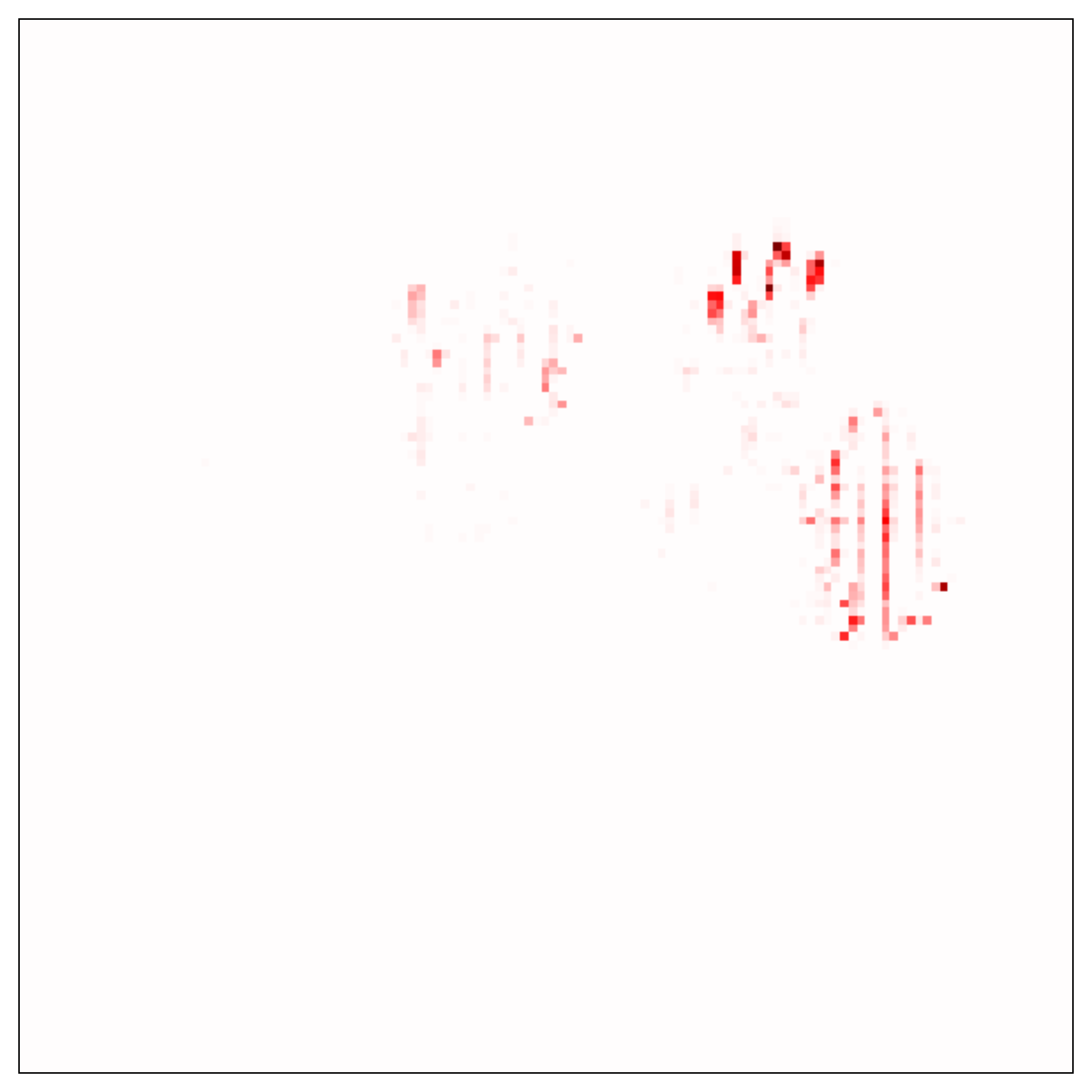} & 0.46 \\
Guided Backprop \cite{Spring:ICLR2015}              & \includegraphics[width=.12\linewidth,valign=m]{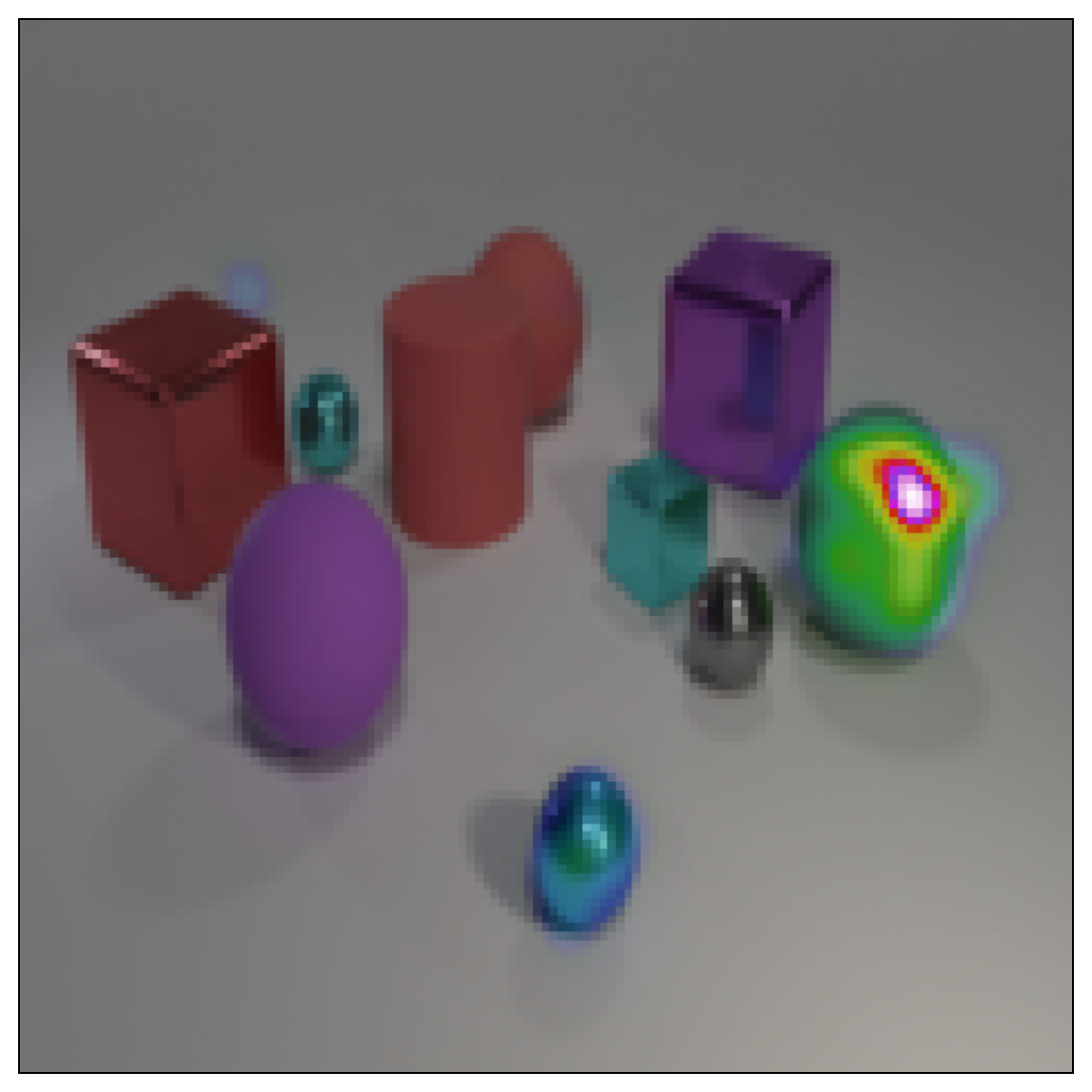} & \includegraphics[width=.12\linewidth,valign=m]{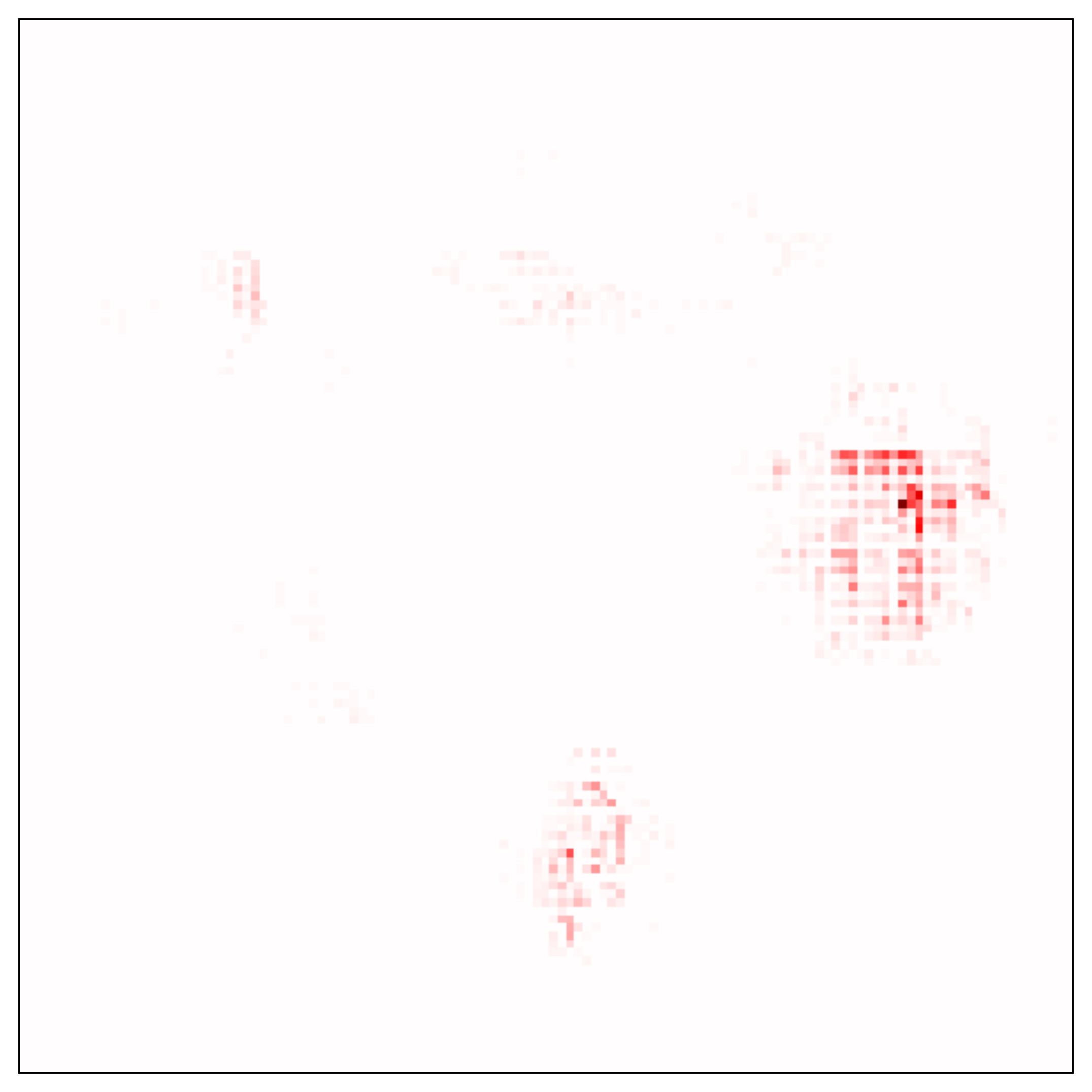} & 0.74 \\
Guided Grad-CAM \cite{Selvaraju:ICCV2017}           & \includegraphics[width=.12\linewidth,valign=m]{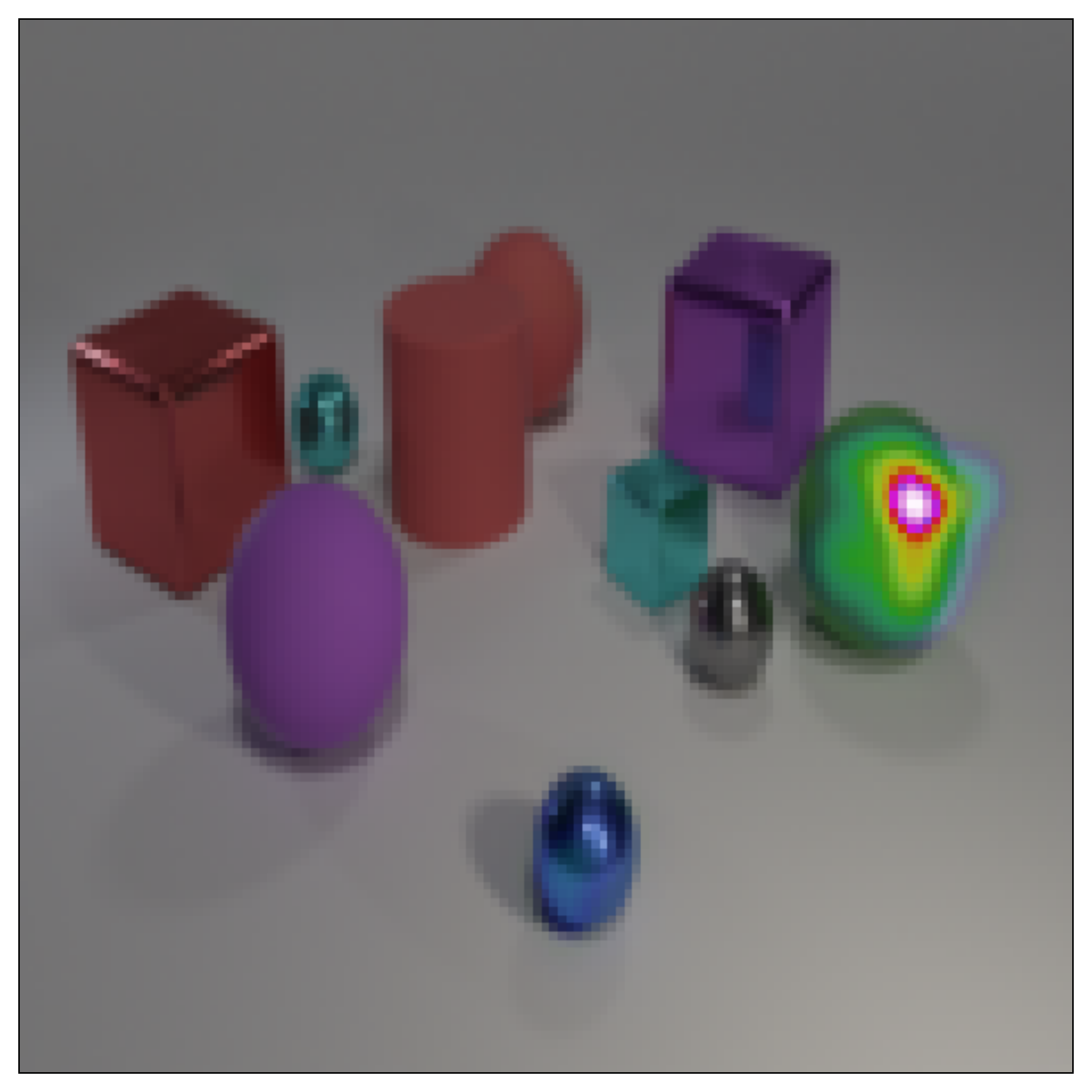} & \includegraphics[width=.12\linewidth,valign=m]{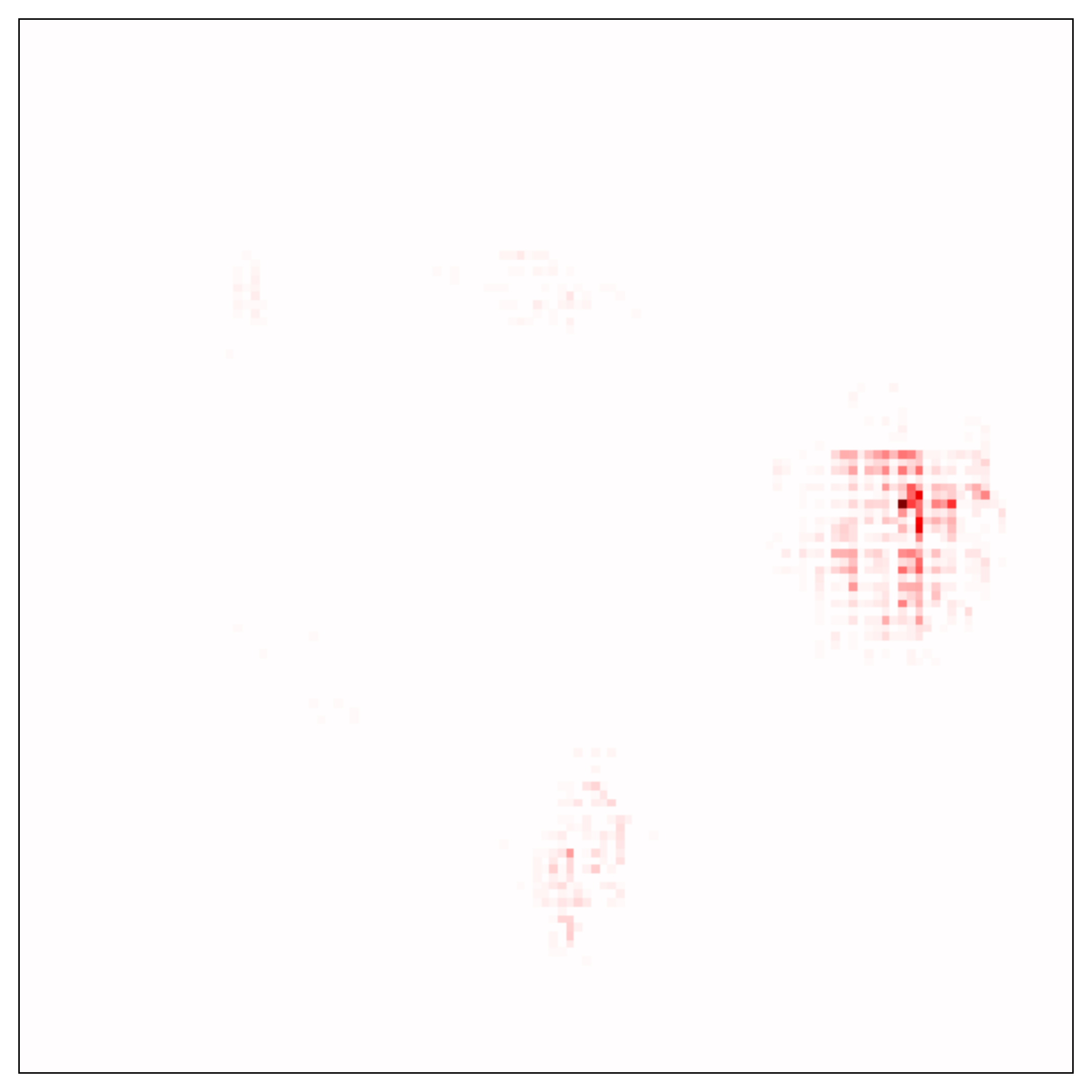} & 0.81 \\
SmoothGrad \cite{Smilkov:ICML2017}                  & \includegraphics[width=.12\linewidth,valign=m]{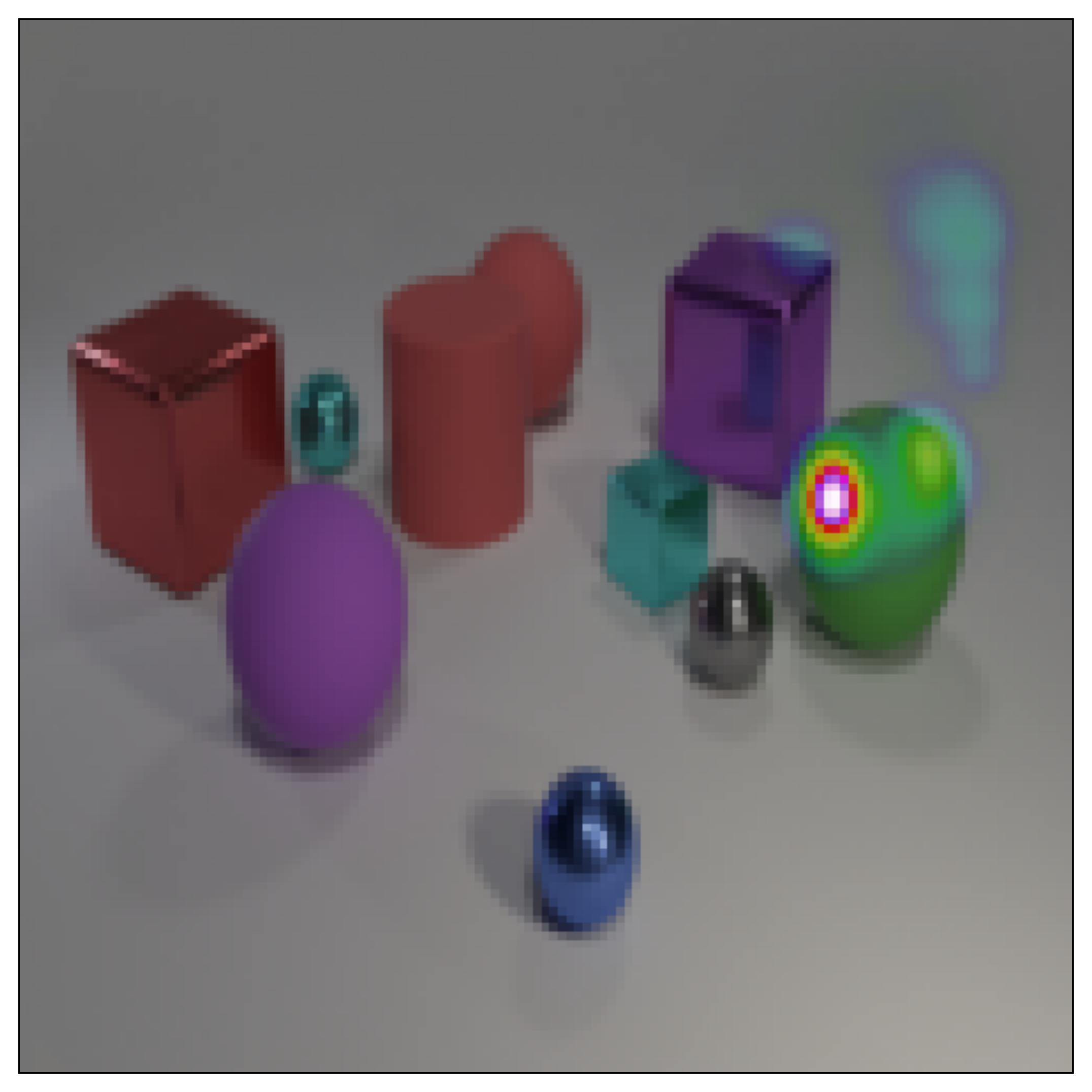} & \includegraphics[width=.12\linewidth,valign=m]{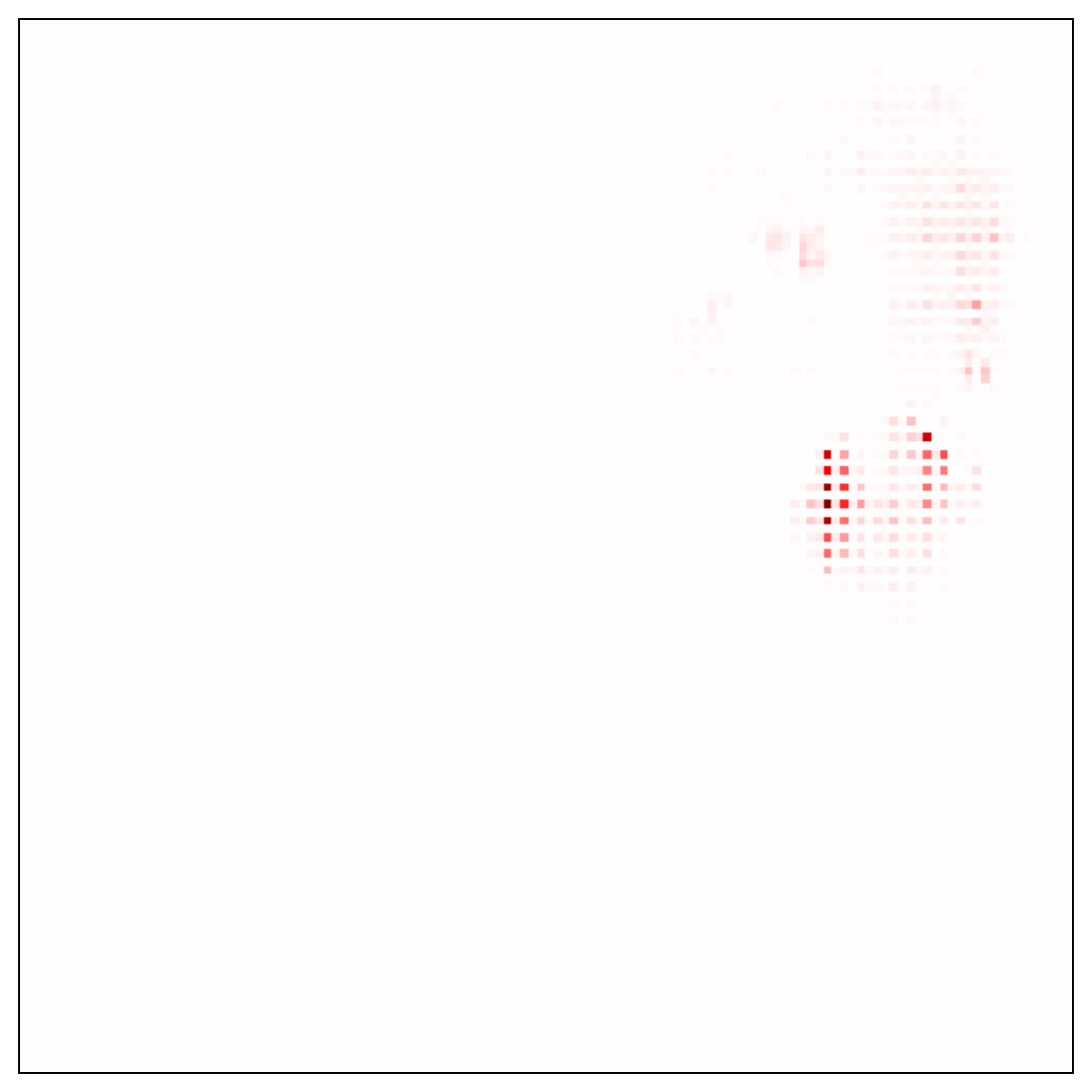} & 0.55 \\
VarGrad \cite{Adebayo:ICLR2018}                     & \includegraphics[width=.12\linewidth,valign=m]{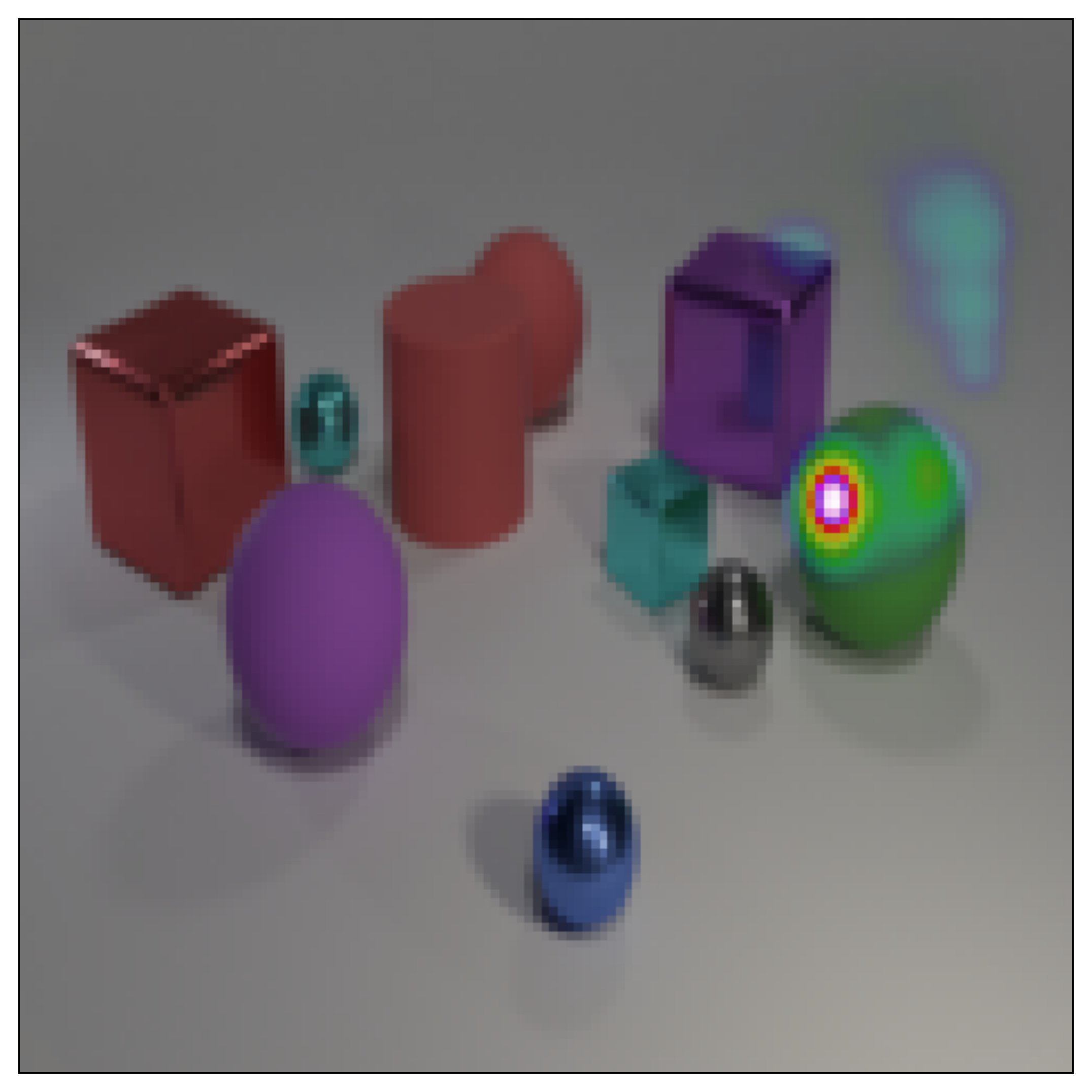} & \includegraphics[width=.12\linewidth,valign=m]{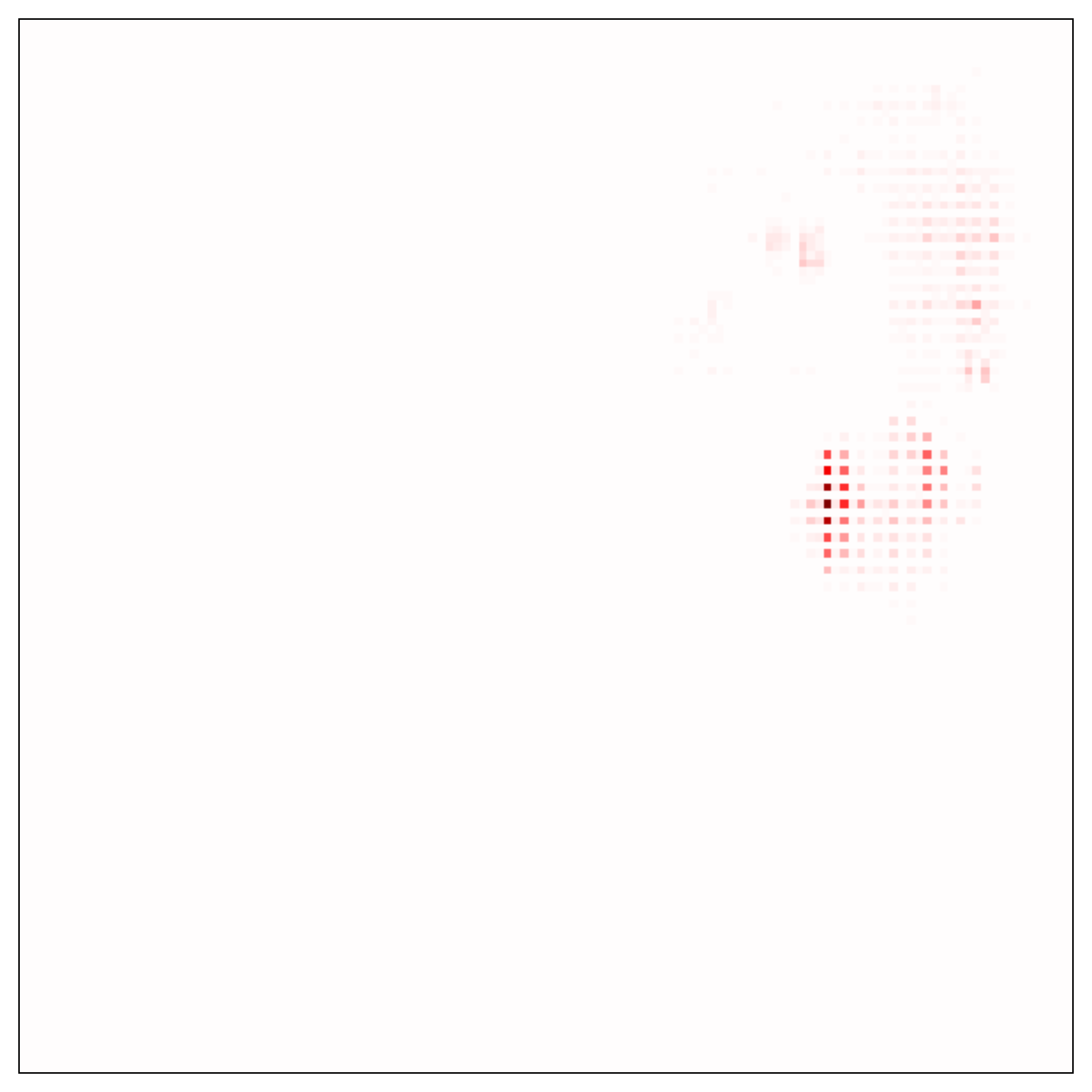} & 0.53 \\
Gradient \cite{Simonyan:ICLR2014}                   & \includegraphics[width=.12\linewidth,valign=m]{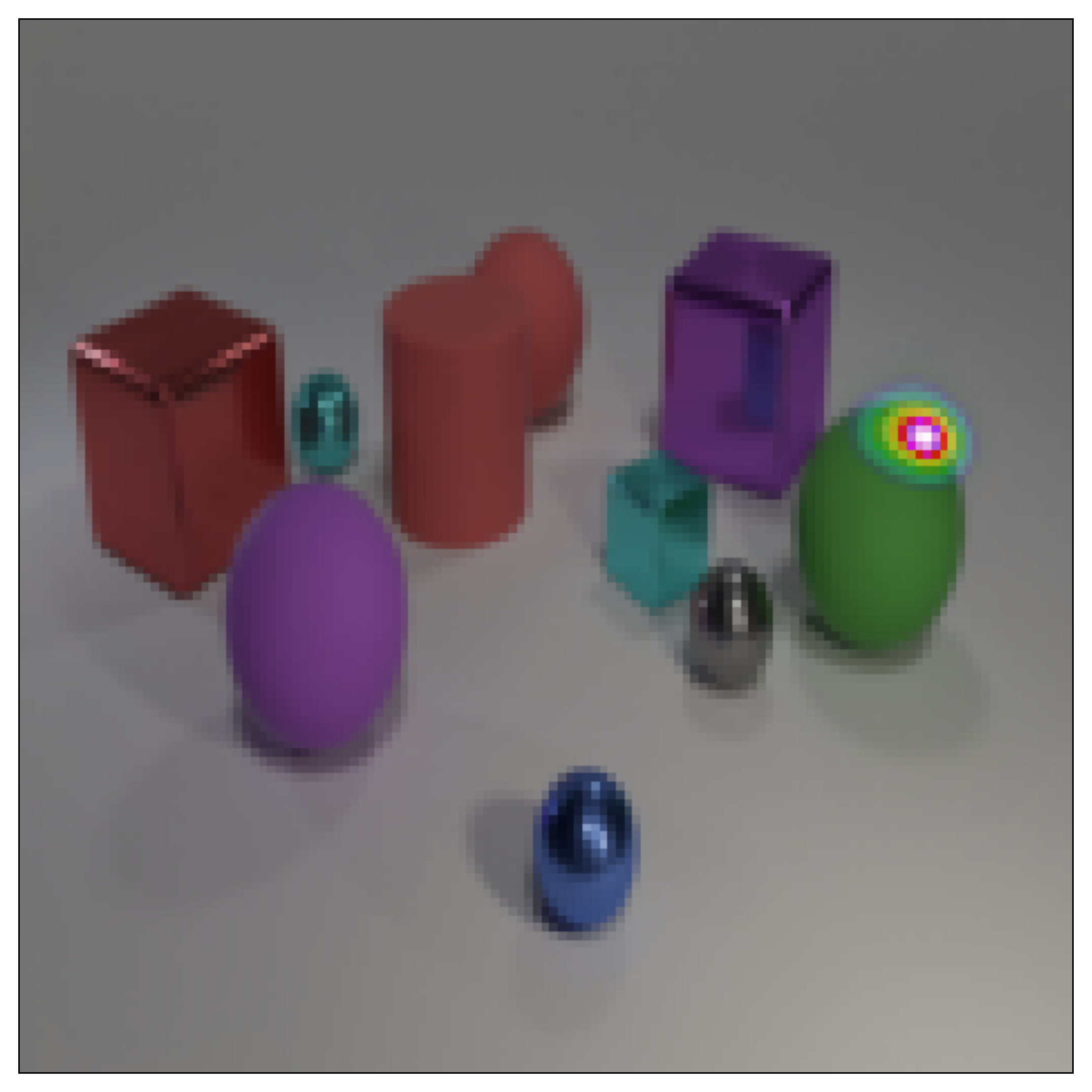} & \includegraphics[width=.12\linewidth,valign=m]{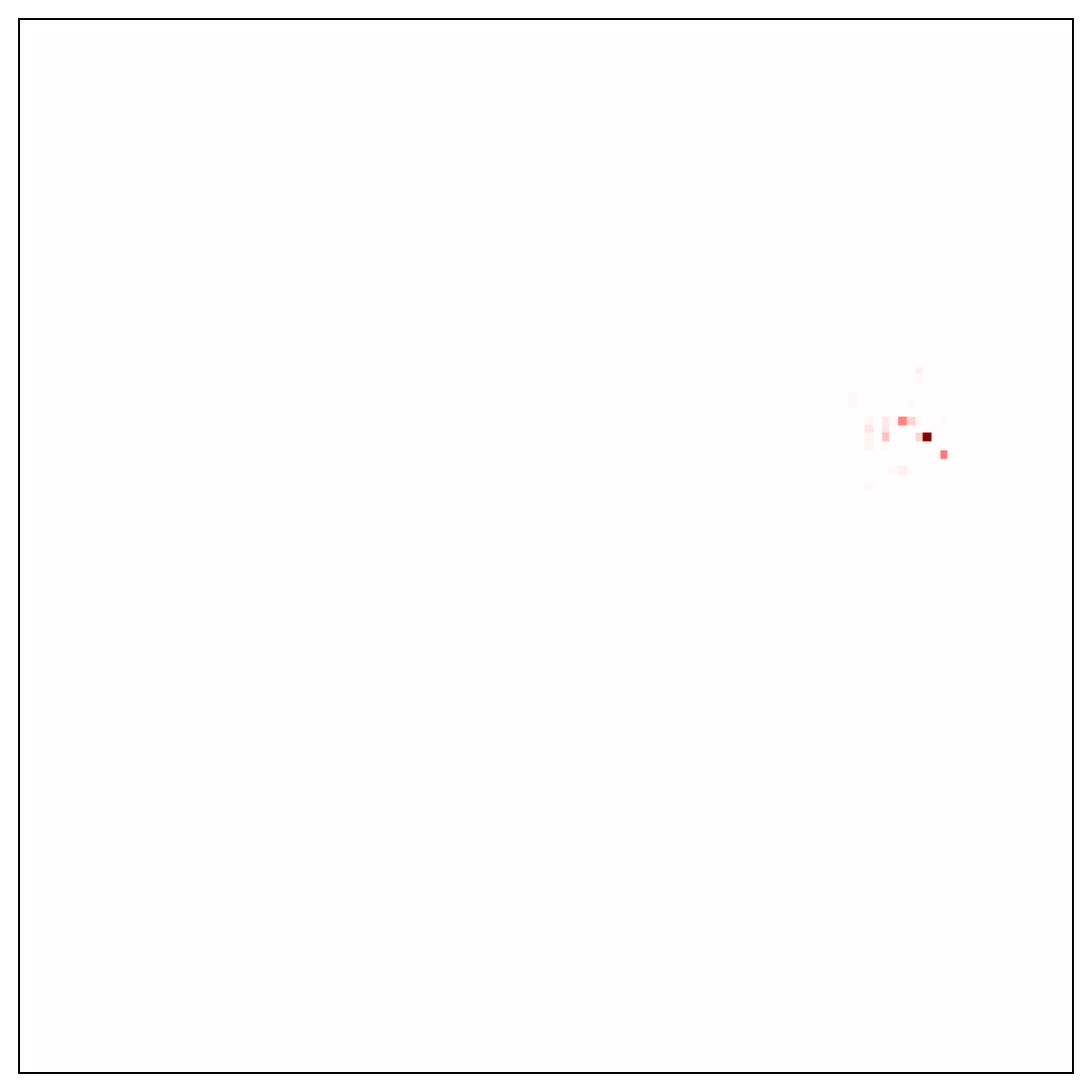} & 0.93 \\
Gradient$\times$Input \cite{Shrikumar:arxiv2016}    & \includegraphics[width=.12\linewidth,valign=m]{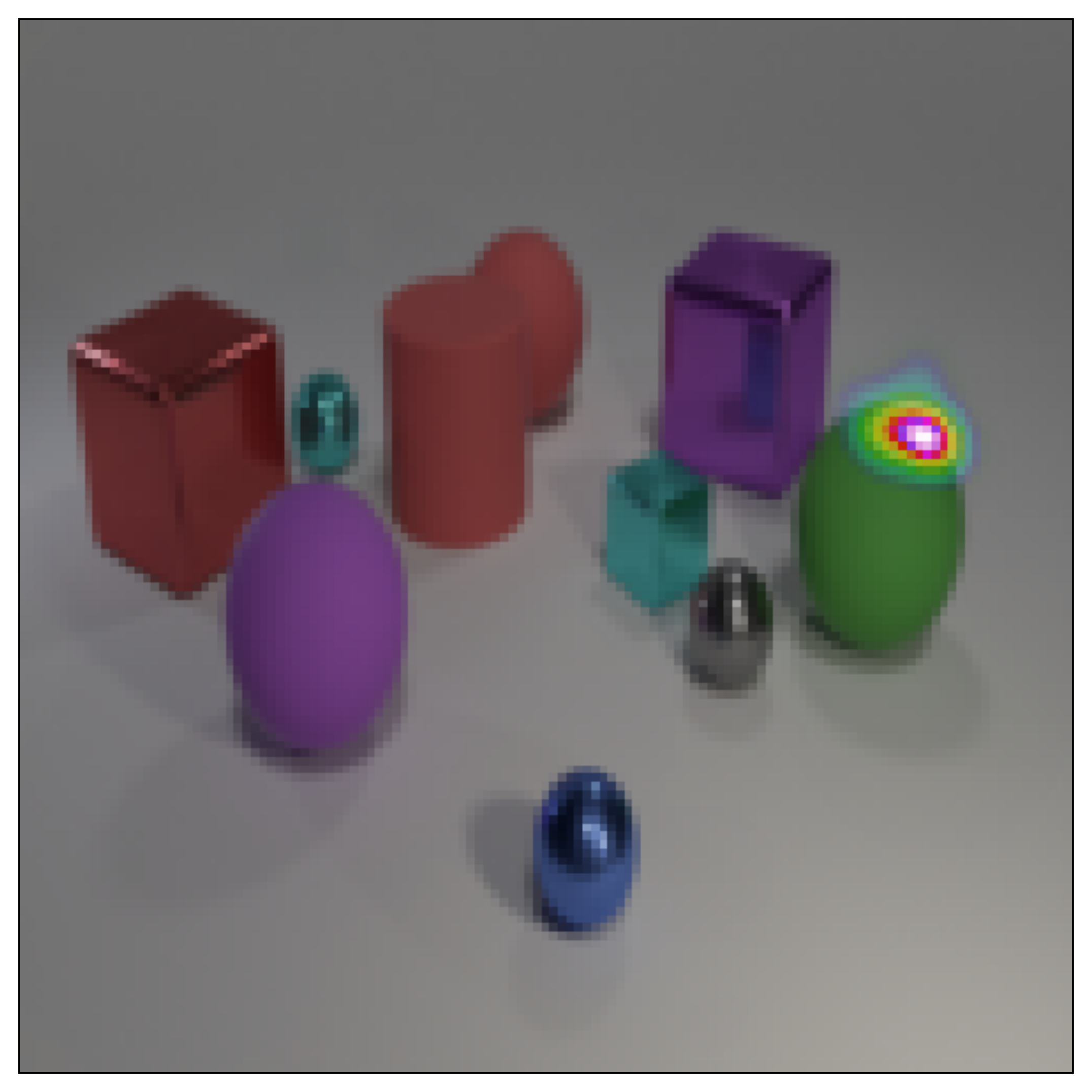} & \includegraphics[width=.12\linewidth,valign=m]{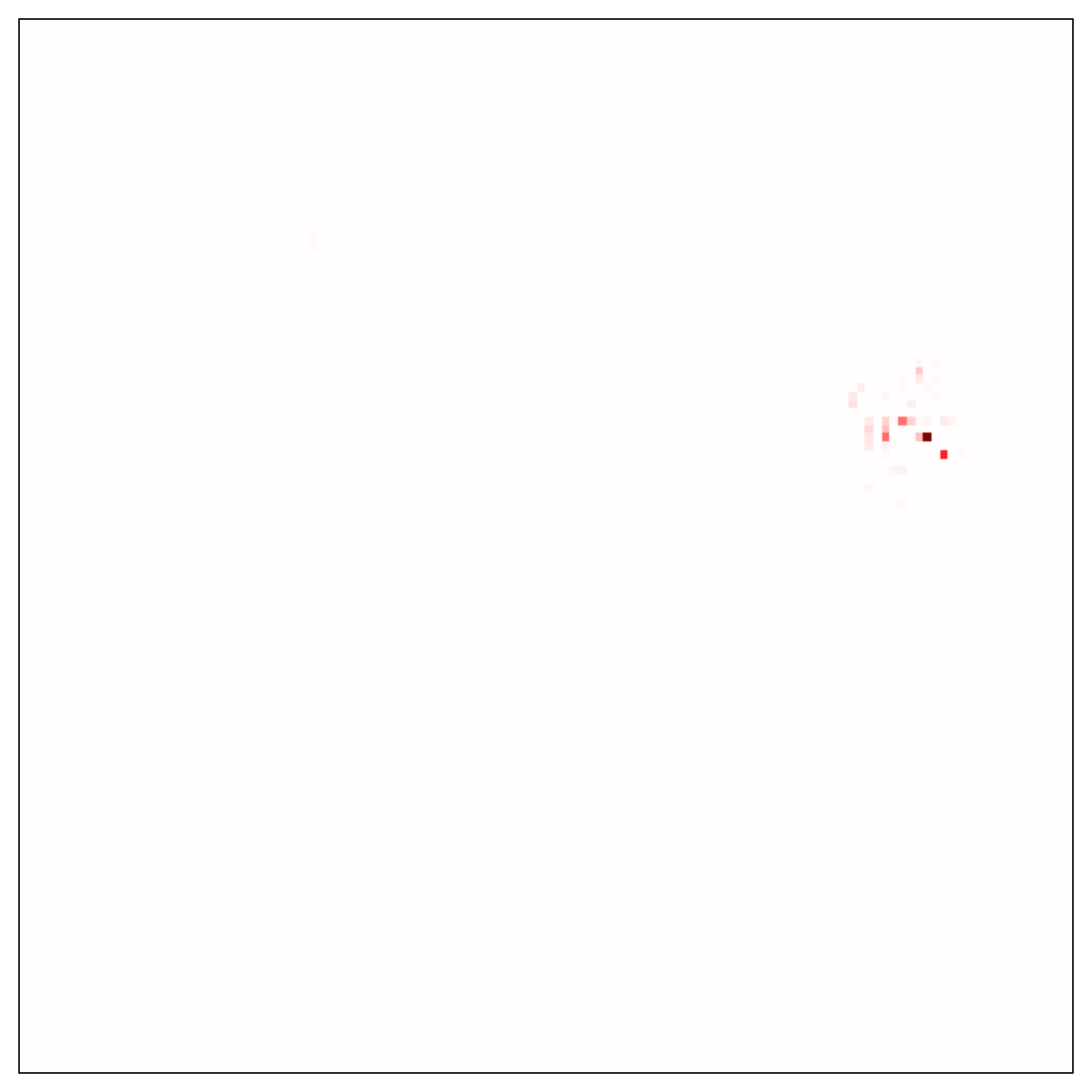} & 0.83 \\
Deconvnet \cite{Zeiler:ECCV2014}                    & \includegraphics[width=.12\linewidth,valign=m]{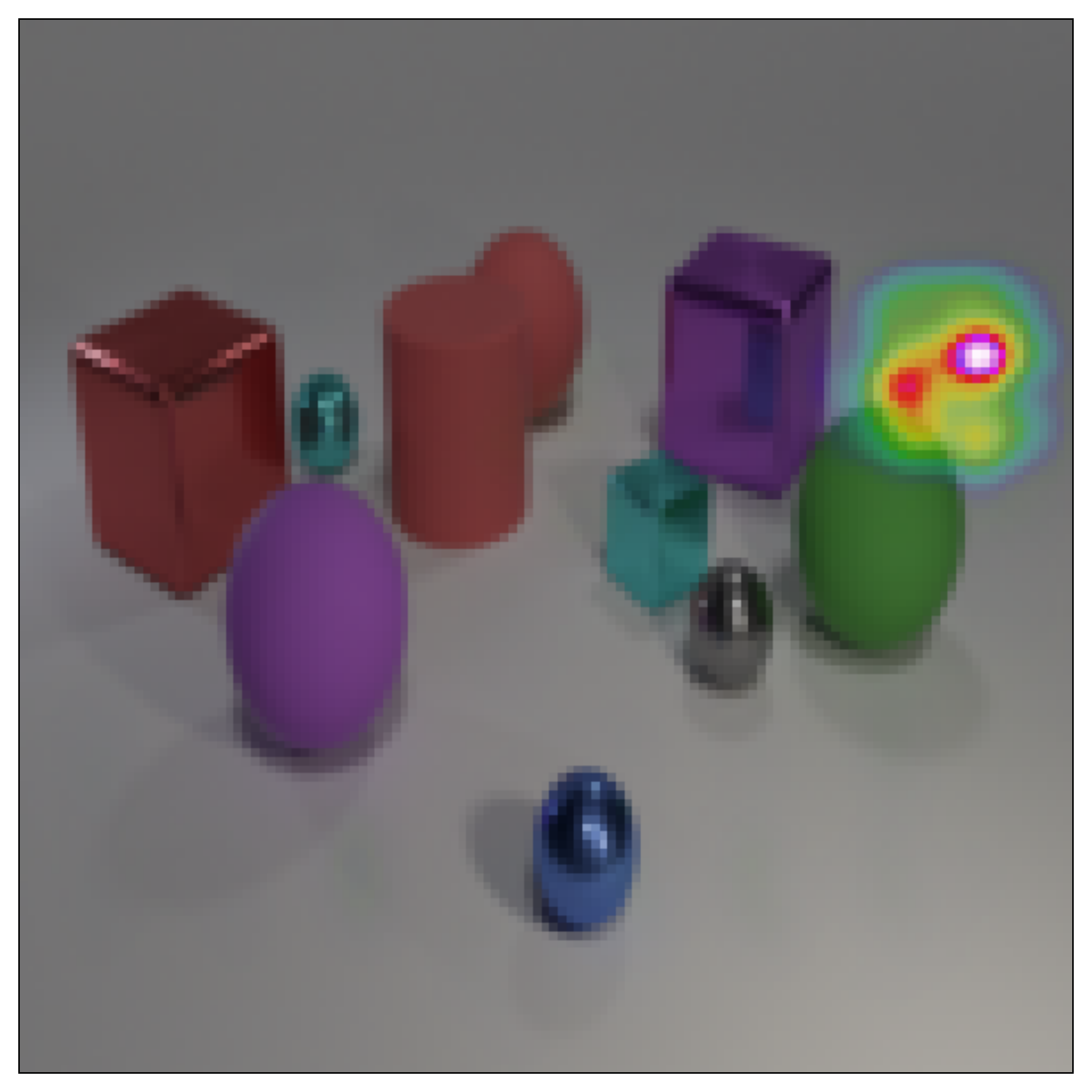} & \includegraphics[width=.12\linewidth,valign=m]{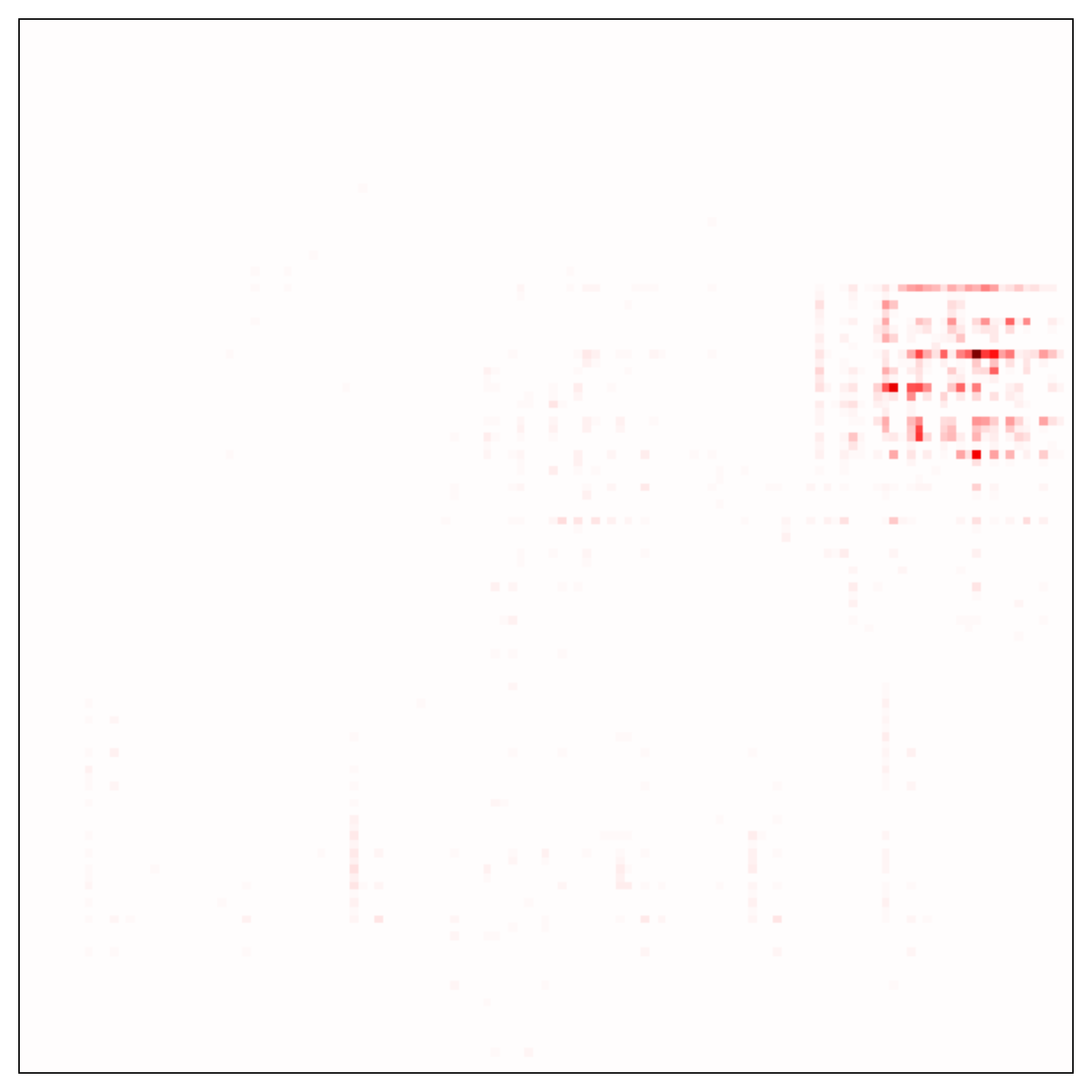} & 0.13 \\
Grad-CAM \cite{Selvaraju:ICCV2017}                  & \includegraphics[width=.12\linewidth,valign=m]{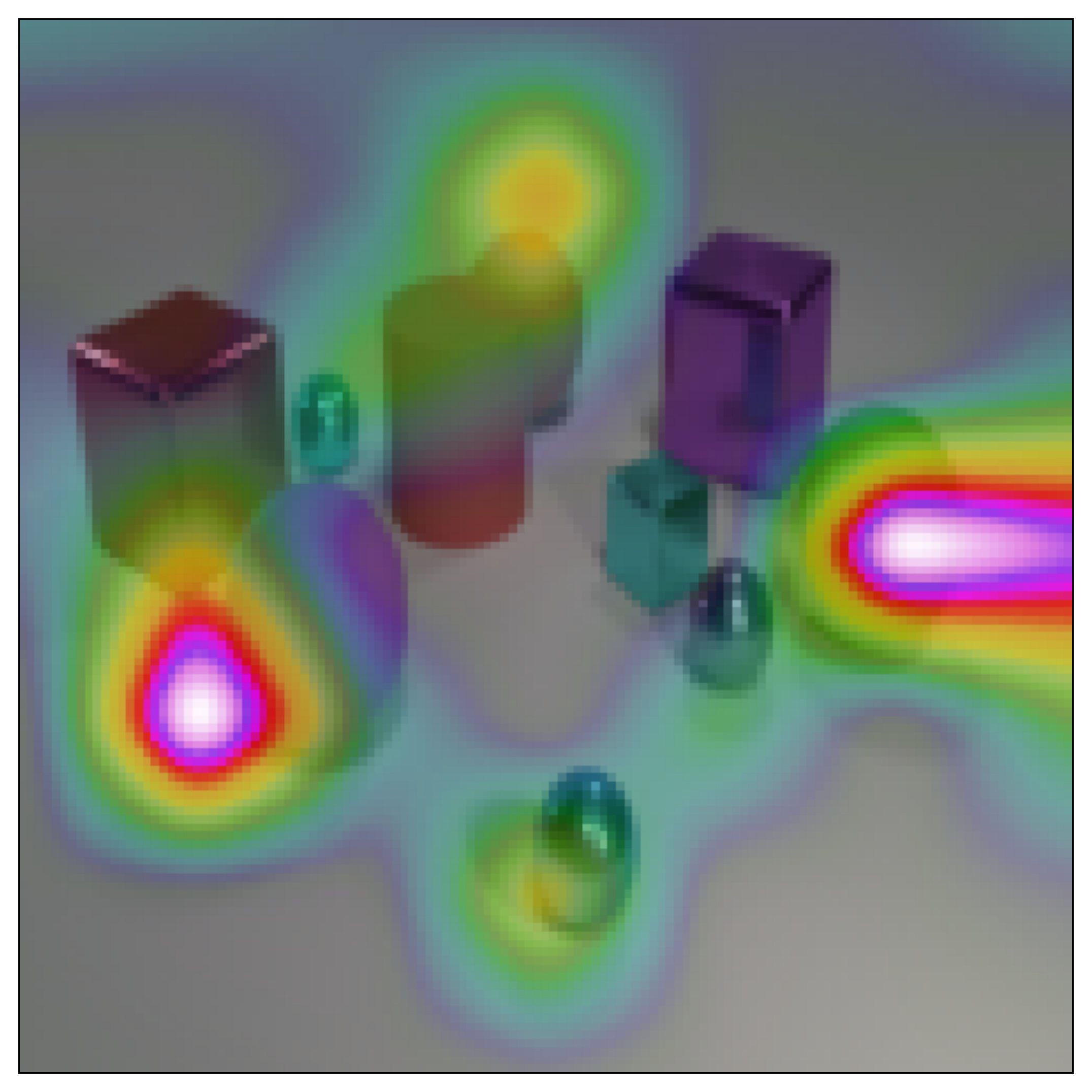} & \includegraphics[width=.12\linewidth,valign=m]{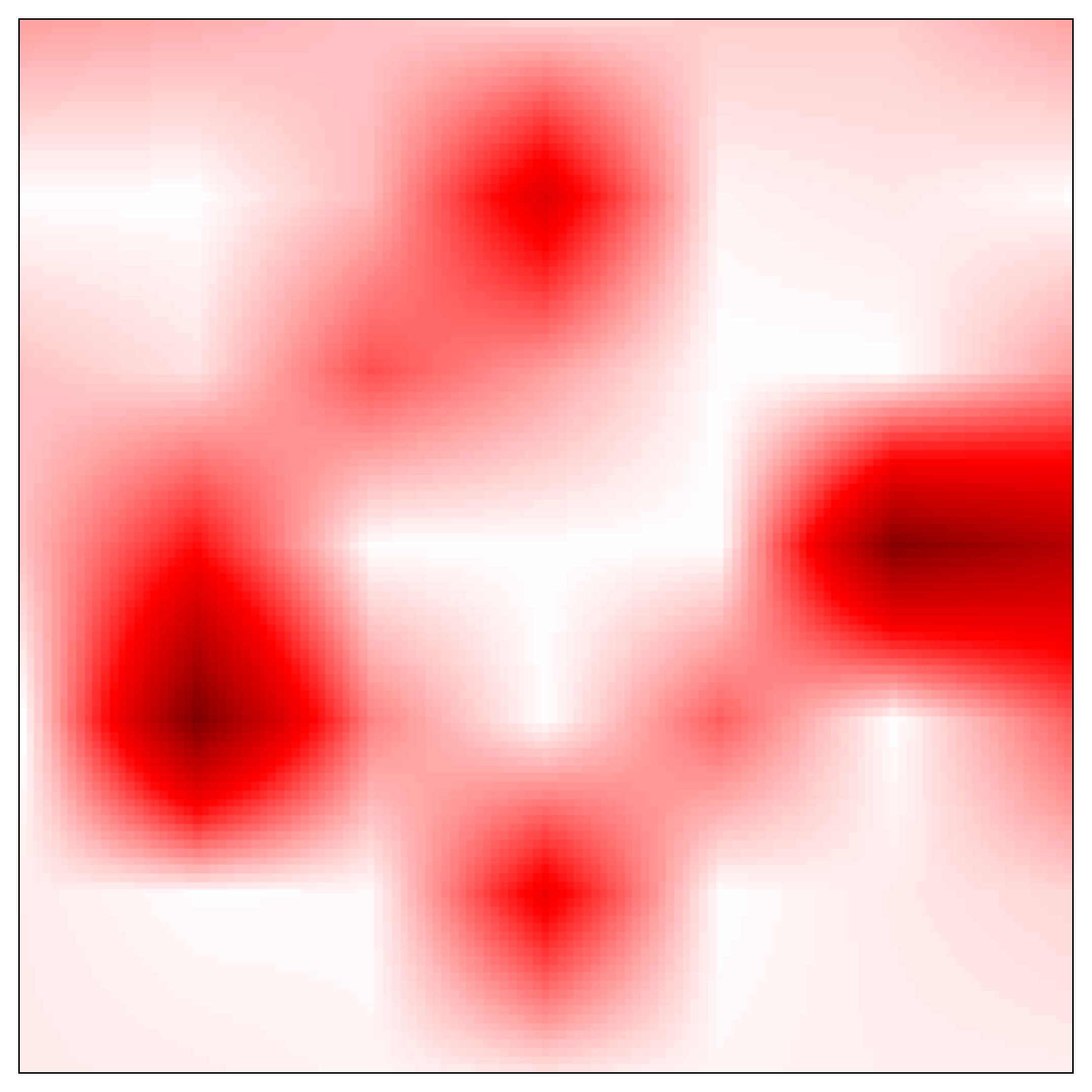} & 0.14 \\
\end{tabular}
\end{table}

\end{document}